\documentclass[english,12pt]{ucdenver-dissertation}
\usepackage[T1]{fontenc}
\usepackage[utf8]{inputenc}
\usepackage{textcomp}
\usepackage{floatrow}
\usepackage{url}
\usepackage{paralist}
\usepackage{xparse}

\usepackage[document]{ragged2e}
\RaggedRight
\usepackage{etoolbox}
\AtBeginEnvironment{figure}{\setlength{\RaggedRightParindent}{0in}}
\AtBeginEnvironment{table}{\setlength{\RaggedRightParindent}{0in}}
\AtBeginEnvironment{algorithm}{\setlength{\RaggedRightParindent}{0in}}

\usepackage{placeins}
\usepackage{ctable}
\usepackage{caption}
\usepackage{subcaption}

\usepackage{doi}
\usepackage[authoryear]{natbib}

\usepackage{amsmath,amssymb,amsfonts,mathrsfs}

\usepackage{color}
\usepackage{enumerate}

\usepackage{hyperref}

\usepackage[english]{babel}
\theoremstyle{plain}

\theoremstyle{definition}

\theoremstyle{remark}

\theoremstyle{definition}

\numberwithin{equation}{section}

\numberwithin{equation}{section}

\usepackage[utf8]{inputenc}

\DeclareFixedFont{\ttbb}{T1}{txtt}{bx}{n}{10} 
\DeclareFixedFont{\ttmm}{T1}{txtt}{m}{n}{10}  

\usepackage{color}
\definecolor{deepblue}{rgb}{0,0,0.5}
\definecolor{deepred}{rgb}{0.6,0,0}
\definecolor{deepgreen}{rgb}{0,0.5,0}

\usepackage{listings}

\newcommand\bashstyle{\lstset{
language=Bash,
basicstyle=\ttmm,
otherkeywords={self},             
keywordstyle=\ttbb\color{deepblue},
emph={python},          
emphstyle=\ttbb\color{deepred},    
stringstyle=\color{deepgreen},
frame=tb,                         
showstringspaces=false            %
}}

\lstnewenvironment{bash}[1][]
{
\bashstyle
\lstset{#1}
}
{}

\newcommand\bashinline[1]{{\bashstyle\lstinline!#1!}}

\graphicspath{{graphics/}}
\usepackage{lipsum}
\usepackage{color}

\usepackage{fancyhdr}
\usepackage{datetime}

\makeatletter
\title{Assimilation of Satellite Active Fires Data}
\authorLast{Haley}
\authorFirst{James}
\authorMiddle{D.}
\education{
B.S., Metropolitan State University of Denver, 2013 \\
Ph.D., University of Colorado, Denver, 2021
}
\school{University of Colorado: Denver, College of Liberal Arts and Sciences}
\program{Applied Mathematics}
\date{2021}
\submitDate{July 31, 2021}
\advisor{Jan Mandel}
\advisorTitle{Professor}
\committeeChair{Burton Simon}
\committeeMembers{
Troy Butler \\
Joshua French \\
Adam Kochanski
}

\preface{
\setlength{\parindent}{0.5in}

Wildland fires pose an increasingly serious problem in our society. The number and severity of these fires has been rising for many years. Wildfires pose direct threats to life and property as well as indirect threats through ancillary effects like reduced air quality due to smoke. The aim of this thesis is to develop techniques to help combat the negative impacts of wildfires by improving wildfire modeling capabilities by using satellite fire observations. Already much work has been done in this direction by researchers in the wildfire community. Our work seeks to expand the body of knowledge using mathematically sound methods to utilize information about wildfires that considers the uncertainties inherent in the satellite data.

In this thesis we explore methods for using satellite observations to help initialize and steer wildfire simulations. In particular, we develop a method for constructing the history of a fire, a new technique for assimilating wildfire data, and a method for modifying the behavior of a modeled fire by inferring information about the fuels in the fire domain. All three of these goals rely on being able to estimate the time a fire first arrived at every location in a geographic region of interest. Because detailed knowledge of the histories of real wildfire is typically unavailable, the basic procedure for developing and testing the methods in this thesis will be to first work with simulated data so that the estimates produced can be compared with known solutions. The methods thus developed are then applied to several real-world scenarios. Analysis  of these scenarios shows that the work with constructing the history of fires and data assimilation improves improves fire modeling capabilities. The research is significant because it gives us a better understanding of the capabilities and limitations of using satellite data to inform wildfire models and it points the way towards possible new avenues for modeling fire behavior.

}

\acknowledgements{
\setlength{\parindent}{0.5in}

I am thankful for my advisor Dr. Jan Mandel and his patience and wisdom. I am continually amazed by the breadth of his experience and knowledge.I am humbled by the experience of being his student.

I am also thankful for  Burt Simon, Troy Butler, Joshua French, and Adam Kochanski who served on my PhD committee. Their expertise, critiques, and advice was invaluable during the writing of this thesis. I appreciate the time and effort that each one of them devoted to this project. It should also be mentioned that of all the exams I took during my education, the hardest was given than none other than Troy Butler.

Special thanks also go to others I have met and been inspired by on this journey. The list of people who have helped me to do some task or inspired me to see it through is limitless. Included on this list are fellow students, professors, administrators, colleagues in the wildfire trade, and students of my own who it was a pleasure to interact with. Some of the names of these peoples are Chau Nguyen, Zina Stilman, Fareed Omar, John Ethier, William Emerson, Stephanie Puello, Maria Rase, Susan Rivera, Mike Kawai, Robert Rostermundt, Angelica Maramara, Tessa Rowe, Derek Mallia, Angel Farguell, Kyle Hilburn. I didn't know I knew so many people and have likely failed to list half of those who should be. 

And no list should fail to mention one's parents. I am eternally grateful to my parents, Jeff and Gale, for the boundless love and support they have given me over the years. I love them greatly and feel so lucky to be their son.

This work was partially supported by NSF grants DMS-1216481 and ICER-1664175.

\vfill
}

\usepackage{mathtools} 
\usepackage{array}

\makeatother

\begin{document}

\chapter{\uppercase{Introduction}} \label{chapter:01}


\section{The Problem and Proposed Solution Methods}\label{sec:problem}


\subsection{Wildfire Impacts and the Need for Improved Modeling}
Over the past several decades the numbers and sizes of North American wildfires have been increasing \citep{Schoennagel-2017-AMW}. Climate changes have resulted in longer fire weather seasons and increased burnable area on a global scale \citep{Jolly-2015-CVG}. For example, in California, the number of autumn days with extreme fire weather has doubled in the last 30 years \citep{Goss-2020-CCI}. These changes create a significant risk to life, property, natural habitat, and lead to increased economic costs in terms of fire suppression efforts, fuel management, property loss, and other societal costs \citep{Schoennagel-2017-AMW}. Additionally, the effects of smoke on air quality can extend well beyond the area directly affected by fire. These negative impacts on the health of people and on the air quality are of increasing concern as more fires occur \citep{Jaffe-2020-WPB}. To help combat the effects of increased wildfire activity, better modeling of wildland fires can be used to help fire suppression crews, emergency planners, and public health officials make more informed decisions \citep{Andrews-2007-PW}. Fire models can also be used to help with planning of preventative measures like prescribed-fire planning, assigning fire-danger ratings, and fire control planning \citep{Albini-1976-EWB,Andrews-2007-PW}.

\subsection{Towards Better Wildfire Modeling}

The purpose of the research presented in this thesis is to help achieve better modeling of fires. To accomplish this, we propose the following methods.
\begin{itemize}
\item Initialization of fire simulations from an estimate of the ignition point or fire arrival time. This task attempts to find the best estimate of the state of the fire before starting a simulation.
\item Steering the model by assimilation of satellite observations. This task keeps the simulation up-to-date with the latest information available about the observed state of the fire.
\item Adjustment of the fuel moisture content (FMC) used in the model. The FMC is a measure of how relatively wet or dry the fuels in the fire domain are. This task tries to optimize a key parameter of the model so that future predictions are more accurate.
\end{itemize}

Each of these tasks is briefly outlined below and full details of the methods and the capabilities will be covered in its own chapter.

\subsubsection{Estimation of the Ignition Point and Fire Arrival Time}
A method for estimating the fire arrival time from satellite data will be demonstrated. The fire arrival time is the time the time that fire first arrived at a location in an area of interest. Simulations of fire may be started from such an estimate and can provide better predictions than initializing a simulation from an ignition point. Because active fire detection data is relatively sparse, it is generally not possible to know when, or even if, a fire has arrived at every location within an area of interest. The method proposed in this thesis  estimates the fire arrival time at locations without active fire detections by comparison to nearby locations that do have recorded fire arrival times. In most cases, the known fire arrival times will be obtained from satellite active fire data, but other observations such as infrared fire perimeter data can be used as well. All that is needed are the GPS coordinates of  actively burning locations and the times the observation were made. The method will be discussed in Chapter \ref{chapter:03}. This chapter will also detail a method to use satellite data to estimate the ignition point of a wildfire. Having an estimate of the ignition point of wildfire can help in producing better fire forecasts in modeling or provide fire investigators additional information about the cause of the fire.

\subsubsection{Data Assimilation of Satellite Observations}
Steering the model with updated information about the fire location is in Chapter \ref{chapter:04}. This method will help keep the running model reflecting the latest information about the fire extent. The method is similar to that used for estimating the fire arrival time, but combines the model forecast with the fire data in a way that accounts for the known uncertainties inherent in the satellite data. Importantly, the properties of the satellite instruments have been studied in order to quantify those uncertainties with regard to the satellite fire observations. The process of combining satellite observations with the output of the wildfire model is used to make periodic restarts of the fire simulation when more data becomes available.

\subsubsection{Adjustment of Fuel Moisture Content}
The third goal of this research is to develop a method for adjustment of the fuel moisture content used by the model  by comparison of the model forecast with an estimated fire arrival time derived from satellite data.  The effect is to make the model fire spread faster or slower during the next simulation period to better match the conditions observed by satellite. This method will help the model make better predictions during the next simulation period after data assimilation has been performed and the simulation restarted from an updated state. The method will be discussed in Chapter \ref{chapter:05}

\subsection{Wildfire Modeling Approaches}

Wildfire modeling has a history dating back to the 1920s and continues to this day \citep{Sullivan-2009-RWF1}. There are currently many efforts being made to improve the accuracy and efficiency of computer wildfire models. An important component of a system is the spread model that shows how fire propagates given a set of inputs like local wind speeds, temperature, humidity, slope, and the properties of available fuels. One of the most widely used spread models is the Rothermel model \citep{Rothermel-1972-MMP}. It is a semi-empirical  model that that relies on information about winds, terrain, and fuels to estimate a rate of spread (ROS) of a fire at a point. Calculations can be made quickly and the model is appropriate for operational usage in the field. Other models such as FIRETEC \citep{Linn-2002-SWB} are fully physics-based and require high performance computing resources to run simulations due to the complexity of the calculations. A third approach is to model fire behavior as a stochastic process where a sequence of time steps determine how fire progresses from an initial point on a computational grid to other points in the fire domain. The  PROPAGATOR \citep{Trucchia-2020-POC} is one such model. The probability of fire spread from one location to another is computed from the properties of the fuels, winds, and terrain at locations in the fire domain. Computational times with the model are very fast and two days of simulation can be produced on a laptop computer in five minutes.
Wildfire Analyst (WFA) is a commercial software system designed for operational use in response command centers or in the field that is capable of running on a desktop or tablet computer \citep{Ramirez-2011-NAF}. The tool was first used in Spain but recent updates to its capabilities have made it suitable for use in other regions. The software provides real-time analysis of wildfires and can rapidly simulate the spread of wildfires \citep{WA-2021}). The fire spread model is based on the Rothermel model and incorporates tools for adjusting the fire rate of spread in the model, by using observations of the behavior of the actual fire \citep{Cardil-2019-ARS}. FlamMap \citep{Finney-2006-OFF} desktop software has been developed by the U.S. Forest Service as a package capable of simulating wildfire behavior by inclusion of the FARSITE \citep{Finney-1998-FFA} fire growth simulation model. FARSITE uses the Rothermel model to account for surface fire spread and also is capable of accounting for fire spread by crown fire and spotting, providing the capability to simulate an array of fire with different characteristics. A dead fuel moisture model accounts for slope, elevation, aspect, and weather at the site of individual pixels in the fire domain. A graphical user interface is capable of displaying fire perimeters at time steps used by the program. A comprehensive survey of wildfire modeling strategies developed between 1990 and 2007 is provided to the interested reader in \citep{Sullivan-2009-RWF}.

\subsubsection{Coupled Fire-Atmosphere Models}
Modeling of wildland fires is difficult since the phenomena is not completely understood \citep{Clark-1996-CAF}. The behavior of these fires is greatly affected by available fuels, topography, and weather \citep{Rothermel-1972-MMP}. For example, a fire will spread faster uphill through dry grass and in the presence of winds blowing up that the hill than than it would spread over level ground through moist timber with no winds present \citep{Pyne-1996-IWF}. In turn, wildland fires can affect the weather. In particular,  wildland fires release heat, smoke, and moisture into the atmosphere. These releases can affect local weather by producing winds, contributing to instabilities in the atmosphere, and creating cumulus clouds capable of producing rain, lightning, and downburts \citep{Pyne-1996-IWF}.
To reflect the interactions between fire and atmosphere in wildland fire models, a coupled fire-atmosphere model can be used. In such a model, outputs from a weather model, such as wind speed, temperature, and humidity, are used as inputs into a fire spread model. Outputs from the fire spread model, such as heat and vapor  fluxes, are used as inputs to the weather model \citep{Mandel-2011-CAF}. 

Coupled models have been in use for many years. \citet{Clark-1996-CAF}  demonstrated with The Coupled Atmosphere-Wildland Fire Environment (CAWFE) the effectiveness of coupling a mesoscale weather model with a simple dry eucalyptus forest fire model to create a wildland fire simulation model useful for the investigation of fire dynamics. Since then, other coupled fire-atmosphere models have been implemented by various researchers.
For example, the MesoNH-ForeFire Model of Filippi \citep{Filippi-2011-SCF} couples the ForeFire fire area simulator, based on the Balbi fire spread model \citep{Balbi-2009-PMW}, to the Meso-NH  mesoscale weather numerical model \citep{Lafore-1998-MHS}  to simulate the interactions between fire and atmosphere that would not be possible without the fire-atmosphere coupling. Using this system, simulations of real world fires were able to qualitatively reproduce fire characteristics such as the characteristics of fire plumes as observed in the field. These simulations were able to be run in less than a single day on a dual-core computer in less than a day.
The WFDS model of Mell \citep{Mell-2007-PAM} couples a fire spread model to the Fire Dynamics Simulator (FDS) \citep{McGrattan-2013-FDS} in order to simulate the spread of fire in surface fuels over flat terrain. This physics-based model showed an ability to simulate the progress of the head fire observed in two Australian grassland experiments but overestimated the rate of spread in the flank fires. These experiments were performed on fire domains up to the size of $1.5 \times 1.5$ km using computers with less than a dozen processors and took many hours to complete simulations lasting less than two minutes. The complexity of calculations makes this tool more appropriate for research usage than for operational usage \citep{Sullivan-2009-RWF3}
Dahl et. al \citep{Dahl-2015-CFA} introduced a coupled fire-atmosphere model by the joining Discrete Event System Specification Fire model (DEVS-FIRE) \citep{Ntaimo-2004-EFC} and the Advanced Regional Prediction System (ARPS) \citep{Xue-2001-ARP} atmospheric model. The spread model employs a raster-based approach that accounts for fuel, terrain, and weather data in a cellular space where fire ignition moves from cell to cell according to rules that are informed by the Rothermel fire spread model. Fire–atmosphere interactions are made using heat output from the DEVS-FIRE fire spread component as an input to the ARPS atmosphere model. These inputs result in changes in near-surface winds in ARPs that are, in turn, used as an input to DEVS-FIRE.
The HIGRAD/FIRETEC \citep{Reisner-2000-CAM,Linn-2002-SWB} is a coupled atmospheric transport–wildfire behavior model from Los Alamos National Laboratory. HIGRAD is a hydrodynamic model that solves a version of the Navier-Stokes equations \citep{Reisner-2000-CAM} and FIRETEC is the fire model that emulates the average behavior of fuels and gases in wildires.  The combined system is a physics-based model which solves equations for the conservation of mass, momentum and energy, and has managed to capture fine-scale processes through subgrid modeling. HIGRAD/FIRETEC has been used to study the dynamics of fire spread in grass fires \citep{Cunningham-2007-NSG}.

\begin{figure}[!ht]
\centering
\includegraphics[width=0.30\textwidth]{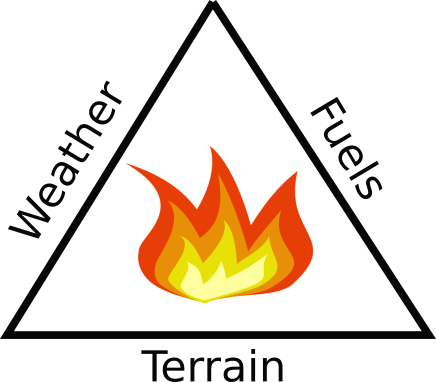}
\caption[The fire behavior triangle]{The fire behavior triangle. Weather, fuels, and the terrain are the three biggest factors affecting the behavior of wildfires. The Rothermel model used in WRF-SFIRE makes predictions about the fire based on these three inputs.}
\label{fig:fire_triangle}
\end{figure}


\subsection{Overview of the Modeling Strategy}
\begin{figure}
\centering
\includegraphics[width=0.8\textwidth]{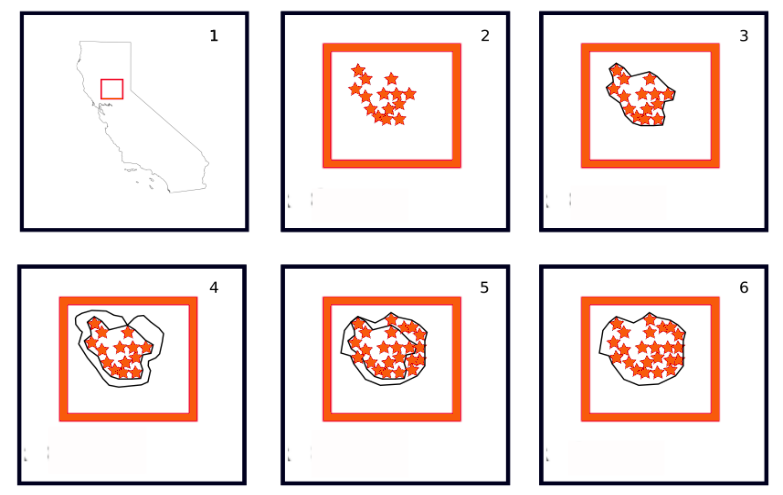}
\caption[Overview of the wildfire modeling strategy.]{Overview of the modeling strategy. Orange stars indicate locations of satellite acive fire detections and black curves represent estimate of the fire's outer perimeter. (1) Set up the fire domain. (2) Collect available observations of the the fire. (3) Estimate the current state of the fire from the observations. (4) Run the model forward in time from the estimate. (5) Collect the most recent observations and update the estimate of the fire arrival time. (6) Restart the model using the most up-to-date observations.}
\label{fig:flow_high}
\end{figure}

Modeling of a fire is performed in a cyclic fashion. The simulation is initialized from an estimate of the ignition point or an estimate of the state of the fire sometime after it has started. The model is then run forward in time, producing a forecast about where and when the fire will move. As more data about the fire becomes available, the model state is adjusted and restarted from an updated estimate  that combines the previous model output with the new data. The model is then run forward until new data becomes available and the the process repeats when more data is assimilated. See Figure \ref{fig:flow_high} for a pictorial representation of the process outlined in the steps below.
\begin{enumerate}
\item Set up the fire domain as a location in space and time. This can be done by hand or by using the WRFX web-based system. This process will bring in additional information about weather, terrain, and other inputs to the fire model. A brief summary of the modeling environment is given in section \ref{sec:modeling}.
\item Collect observational data such as satellite fire detections or infrared perimeter observations. Frequently, the infrared perimeter data needs to be checked for consistency.
\item Estimate the ignition point or fire arrival time from satellite data. Methods for accomplishing these tasks will be developed in Chapter \ref{chapter:03}
\item Run the model forward from the estimated initial conditions. 
\item Collect further observations and use them to adjust the fire arrival time as well as possibly adjust model parameters such as fuel moisture content. The methods used will be discussed in chapters \ref{chapter:04} and \ref{chapter:05}.
\item Restart the model from the updated estimate and continue the forecast. Steps 4, 5, and 6 get repeated.
\end{enumerate}

\section{Existing Methods and Literature Search}\label{sec:lit_search}

\subsection{Estimation of fire properties from data}
Satellite observations have been used to estimate properties of fires for many years. The information of where and when fires have been detected by satellite can be used to infer general properties about fires in  large geographic regions as well as detailed properties about specific fires of interest. These approaches help with fire response planning as well as responding to fires operationally \citep{Albini-1976-EWB}. The ability to estimate the ignition point of a fire or current area burned can be used to initialize computer simulations of fires and to issue more accurate short-term forecasts. Estimates of the current ROS and direction of the fire can be used to help inform such time-critical decisions such as evacuation orders or directing fire suppression efforts. Data about in-progress fires can be used to steer computers models through data assimilation techniques.

Various studies have attempted to use the sparse data from the polar orbiting satellites  to give a fuller picture of the characteristics of individual fires. In some cases, \citep{Benali-2016-DFD} the satellite data were used to infer only the most basic properties of wild fires such as the duration of the event or probable location of the ignition. Other researchers such as \citet{Sa-2017-EFG} have used satellite data to help validate models like the Fire Area Simulator (FARSITE). In \citet{Benali-2016-DFD} data from the Moderate Resolution Imaging Spectroradiometer (MODIS)  were used to estimate the time fire events began and ended as well as the location of the ignition point for fires in California, Greece, Alaska, Portugal, and Australia. The ignition time and location were estimated simply by using the first fire detection(s) associated with a particular fire. In some cases, this method could produce errors in location of more than a kilometer and an error in the time of ignition of up to a day. Given the temporal and spatial resolution of the MODIS instruments, a more accurate estimate would be hard to achieve.
Parks attempted to obtain day-of-burning (DOB) data (fire arrival time at a point with 24 hour temporal resolution) from MODIS satellite data. Ten different interpolation techniques were used to fill in missing DOB burning data for locations within known fire perimeters. There were 21 fires studied and comparisons were made with  the DOB data drawn from fire progression maps made by observers during fire events. MODIS fire data were accepted "as is" and no effort was made to account for uncertainties in this data \citep{Parks-2014-MDB}.
S{\'a} studied nine large fires in Portugal. The fires were modeled using the Fire Area Simulator (FARSITE) and comparisons were made to data from the MODIS platform \citep{Sa-2017-EFG}. The first and last fire detections from the satellites were used as the start and end times of the fire as in \citet{Benali-2016-DFD}. The goal of this research was to  evaluate  the effectiveness of the FARSITE fire modeling software based on the available MODIS data. Laboda et. al have used MODIS satellite data to map the location of fires in the boreal forests of Russia by clustering active fire detections spatially and temporally so that individual fire events can be identified and studied \citep{Laboda-2007-RFS}. A machine technique has been recently developed in \citet{Farguell-2021-MLE} for estimating the fire arrival time from satellite fire data. The technique uses the support vector machine (SVM) supervised classification model to define a fire arrival time from a boundary that separates the times and locations of burning and non-burning pixels as recorded by satellite fire detection algorithms. Importantly, the method uses information about where the satellite records no fire to be present as part of its calculations. The use of these ``non-fire pixels" has been used in this thesis as well.

\subsection{Data Assimilation of Fire Observations}
Data assimilation is the process of modifying the output of a numerical model by using subsequent observations of the real-world system to produce an optimal estimate of the true state of the system. In a typical situation, a computer model of a real-world system is first run, producing an estimate of the state of the modeled system. It is to be expected that this estimate will differ from reality in many ways. After some time, observational data about the real-world system becomes available and is combined with the model estimate to help it better reflect the true state of the system and to reduce the uncertainty of the model predictions. Since the data are utilized as they are collected the process has a relationship to sequential statistical estimation \citep{Ferguson-1967-MSD} and optimal interpolation \citep{Daley-1991-ADA}.

The use of data assimilation in the modeling of wildland fires is an active topic of research and some advances in the field follow. A method to assimilate sensor observations as part of a fire forecasting strategy was demonstrated in \citet{Jahn-2011-FFG}. An inverse modeling approach was used to determine key parameters that govern fire behavior that do not change over a certain length of time. The process involved assimilating data into a simplified model. The rate of spread computed from a Rothermel-based spread model was corrected by assimilation of infrared images taken at the fire front  by use of the Kalman filter algorithm \citep{Rochoux-2013-RSW}. Rochoux later expanded the work to make use of the ensemble Kalman filter \citep{Rochoux-2014-TPS}.
Rios et. al  propose to assimilate observations of fire front locations, obtained from overflight or by ground crews, in order to estimate key model parameters such as wind speed, wind direction, and fuel properties \citep{Rios-2014-FWW}. An inverse modeling approach, where a cost function is minimized, is the data assimilation strategy employed by Rios.
Srivas et. al \citep{Srivas-2016-WSP} propose to use observations for adjusting FARSITE \citep{Finney-1998-FFA} model output. Here, an ensemble Kalman Filter approach is used to update the fire perimeter when periodic observations with known uncertainties become available. In this study, no real-case scenario was explored. Instead, promising results were obtained by using one model run as the "truth", with noise added to it, over observational periods one hour apart in order to generate artificial observations that were subsequently assimilated. This strategy for using artificial data has been employed in much of the research presented in this thesis. 
Xue et. al  have shown the feasibility of assimilating real-time sensor data into large-scale simulations such as wildland fires \citep{Xue-2012-DAU}. Their approach uses a Sequential Monte Carlo (SMC) methods to update the forecasts produced by the DEVS-FIRE \citep{Ntaimo-2004-EFC} fire model. In this case, sample based methods like the SMC are chosen to overcome the non-Gaussian behavior of wildland fire \citep{Mandel-2008-WFM} and preclude the use of techniques such as the Kalman Filter which make a Gaussian assumption.
The research in this thesis follows closely from the work done by Mandel et  al.  in \citet{Mandel-2014-DAS}. In this research, a method was developed to for modifying the fire arrival time in the WRF-SFIRE model using satellite fire data by using a Bayesian approach to inverse modeling. The fire arrival time is adjusted by minimizing difference from the forecast fire arrival time while maximizing the likelihood of the fire detection data. 

\subsection{Modifying the Rate of Spread in the Model}
How far or fast a fire is able to spread depends on many factors. Aspects of the weather, fuel, and terrain determine the behavior \citep{Pyne-1996-IWF}. For example, fire spreads more rapidly in dry and windy conditions. One method to change the ROS in a fire model is to make changes to the fuel moisture content (FMC) in a way that causes the model to make forecasts more consistent with the observed behavior of the actual fire. Other techniques can also be used to modify the ROS.

Methods for adjusting the model parameters giving the ROS were developed by \citet{Rothermel-1983-FPV} and \citet{Finney-1998-FFA} for the Behave and FARSITE models, respectively. Adjustments to the Behave model were made by use of field observations used to make calculations of an ROS adjustment multiplier in the burn model. The FARSITE model uses the same approach to produce  rate of spread adjustment factors. In both cases, field personnel need to collect measurements of fire properties and perform calculations by hand to derive the desired adjustment factors, limiting the usefulness of the technique in cases where the observations may be difficult to acquire  or when rapid decision-making based on model output is required. 
\citet{Cardil-2019-ARS} have developed a method for adjusting the ROS in the Wildfire Analyst software \citep{Ramirez-2011-NAF} that promises quick results when data are available. The method finds ROS adjustment factors by minimizing the difference between the forecast fire arrival time and known fire arrival time of the actual fire at a set of control points within the simulation domain where the fire arrival time is known. Calculations can be made in real-time and the system has the capability to use an array of data sources \citep{WA-2021}).


\clearpage
\FloatBarrier

\chapter{\uppercase{Background: Widlfire Modeling and Data Sources}} \label{chapter:02}

This first part of this chapter gives some basic information about the WRF-SFIRE wildfire modeling environment that has been used in this research. The second part explains the satellite data sources that have been used in the effort to improve the modeling capabilities  of the model with attention paid to discussing its features and limitations. The final part of this chapter uses the known uncertainties of of the satellite data to provides a mathematically justified data likelihood function that will be used as a key component of a data assimilation strategy to be discussed in Section \ref{sec:data_assimilation}. 

\section{Modeling Wildfires}\label{sec:modeling}


\subsection{WRF-SFIRE}
The WRF-SFIRE \citep{Mandel-2011-CAF} system is the coupled fire-atmosphere model that was used in the research presented here. The WRF-SFIRE system combines the WRF mesoscale numerical weather prediction system \citep{Wang-2017-AUG} with SFIRE, a fire spread model based on the Rothermel semi-empirical surface fire spread model \citep{Rothermel-1972-MMP}. The system is capable of running in parallel on large computer systems with many processors. As a further development, data assimilation techniques are being incorporated into  and expanded within the WRF-SFIRE system. 

The Weather Research and Forecasting Model (WRF) is the weather component used in the coupled fire-atmosphere model \citep{Skamarock-2019-DAR}. The model has been under development for many years and finds use as both a tool for research and production of operational weather forecasts. WRF is usually run on a computational grid with distances between grid nodes on the order of kilometers. 

SFIRE (from spread FIRE) is the component of the system that models how fire spreads in the landscape. The state of model is encoded in the fire arrival time (FAT) that gives the time the fire first arrives at each location in the simulation domain. Output from the system is in files that contain scientific data sets. The fire arrival time can be investigated for all locations in the fire domain by examining three matrices in the data set that give the latitude, longitude, and fire arrival time for each node in the computational grid. In normal usage, SFIRE uses a computational grid with spacing between nodes on the order of tens of meters. The interested reader may consult \citet{Mandel-2014-DAS} and \citet{Mandel-2019-IDH} for more detailed information about the system. 

The fire spread model in SFIRE is based on the Rothermel semi-empirical equations \citep{Rothermel-1972-MMP}. The Rothermel model predicts the rate of spread at a point using the equation, here simplified,
\begin{equation}
\label{eq:rothermel}
R = \frac{1+\Phi_s +\Phi_w}{K},
\end{equation}
where $\Phi_s$ is a constant accounting for the effects of the slope of the terrain, $\Phi_w$ accounts for the effects of the wind, and $K$ is a combination of several constants that pertain to the properties of fuels.

It should be noted that wildfire modeling is a difficult problem and that all models have some limitations. Albini \citep{Albini-1976-EWB} lists three limitations inherent to wildfire modeling that we enumerate and give an example how they affect the capabilities of WRF-SFIRE.
\begin{enumerate}
\item The model may not be applicable to the situation. As an example, the Camp Fire of 2018 had periods of rapid growth, driven by winds carrying burning embers, but the WRF-SFIRE model cannot resolve ember transport and ignition
\item The model's accuracy may not be good. The Rothermel model used in the fire spread computations attempts to  simplify a complex phenomena and cannot be always expected to have high accuracy.
\item The model may make use of inaccurate data. Inputs like fuel moisture content or weather conditions are derived from sparse observations that are subject to measurement error.
\end{enumerate}

\subsection{WRFX}
Running the WRF-SFIRE model is a complicated process since setting up the weather model within the system is a non-trivial undertaking. The behavior of the model can be set by changing parameters in text files that are read by the program when a simulation is started. WRFX is a web-based interface for controlling the system that can be used more easily. Some of the tasks it is capable of automating and simplifying follow.
\begin{itemize}
\item Enables push-button initialization of simulations.
\item Can be used for setting up fire simulation domains and downloading data sets.
\item Has built-in routines for exporting fire visualizations. 
\end{itemize}

The WRFX environment is built with many modules for doing specialized tasks, but for the purposes in this thesis, we will refer to all of these modules collectively as WRFX.

\section{Data sources}\label{sec:data}
This section briefly details the data sources used in this thesis and the limitations of each. Of prime importance are data obtained from instruments aboard polar-orbiting satellites that have the ability to observe most location on Earth's surface twice each day.  Although the actual instruments differ in many ways, the Moderate Resolution Imaging Spectroradiometer (MODIS) and  Visible Infrared Imaging Radiometer Suite (VIIRS) share a similar design in their fire detection algorithms and therefore the data products from both platforms will be treated equally. Table \ref{table:data_sources} summarizes the key statistics of the data sources used for this research and the following sections highlight some of the advantages and disadvantages of using these sources. In this thesis, only MODIS and VIIRS fire products have been used for data assimilation. Infrared fire perimeters were used primarily for the purpose of model output evaluation. Characteristics of the satellites and the instruments aboard them such as altitude, flight path, or sensor capabilities constrain the amount of information available at any given time. Although, these instruments give us a consistent picture of what is occurring on the ground, that picture is far from complete.

\begin{figure}[!ht]
\begin{center}
  \includegraphics[width = 0.6\textwidth]{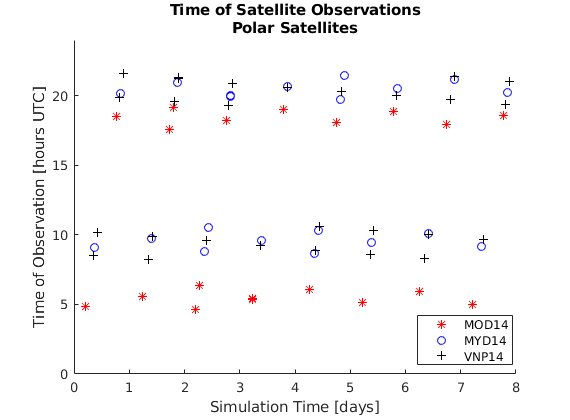}
  \caption[Timing of the satellite observations available for the Patch Fire]{Timing of the satellite observations available for the Patch Fire. The two MODIS and the single VIIRS observations are denoted by MOD14, MYD14, and VNP14, respectively. As can be seen, each platform made two passes over the fire domain daily, with observations on each pass separated by approximately 12 hours. Note the gaps in observations that occur daily between approximately 10:00 and 17:00 as well as 22:00 and 05:00 on the following day.}
  \label{fig:patch_sat_times}
  \end{center}
\end{figure}

Of special importance is the resolution of the observations made by the satellites. Taken together, the MODIS and VIRRS instruments provide several snapshots of fire conditions for any location, several times daily. Imaging by these devices is not continuous and there here are gaps in the coverage that last for many hours. Figure \ref{fig:patch_sat_times} shows a history of when these instruments observed a particular fire. During the time spanning 11 August, 2013 and 18 August, 2013 there were a total of 66 satellite granules whose observations intersected the area with the fire simulation domain. Each satellite made two passes over the fire domain daily, with observations on each pass separated by approximately 12 hours. In some cases, an instrument recorded two granules on a single pass. There are gaps in observations that occur daily, between approximately 10:00 and 17:00 as well as 22:00 and 05:00 on the following day.


\subsection{MODIS}
The two MODIS instruments are each mounted on a polar-orbiting satellite and deliver fire information with 1km resolution. These are the oldest data sources that will be used. These instruments follow a polar-orbiting track 800 km above the Earth's surface that allows them to pass over every location on the earth twice daily at approximately the same \citep{Giglio-2016-C6M}. The direction of the flight path relative to the ground is roughly in a North-South direction and the instrument scans back and forth in directions perpendicular to the flight path. The angle of the scan varies between $-55^{\circ}$ and $55^{\circ}$, producing a picture of the ground approximately 2000 km wide and 10 km thick on each scan. The instruments carry remote sensing equipment to record many wavelengths of electromagnetic radiation, but the fire detection algorithm primarily uses two infrared wavelengths of 4$\mu$ and 11$\mu$ for fire detection. Other wavelengths are used for the purpose of detecting clouds, sun glint, and other environmental factors that affect fire detections.

\subsection{VIIRS}
One polar orbiting satellite launched in 2011 with 375m resolution at nadir view \citep{Schroeder-2017-VII}. The satellite orbits at an altitude of approximately 800 km and produces a picture of the ground that is 3060 km wide. It uses a fire detection algorithm that is based on that developed for the MODIS instruments. The resolution of the fire product is 750m. Several pixels are aggregated for smaller scan angles so that the sizes of pixels at nadir and at the limit of of the scan in the across-track direction are more consistent. Figure \ref{fig:scan_angle} shows how the size of the fire pixels grows with the scan angle. The VIIRS instrument is also on board the NOAA-20 satellite which has recently become operational. Observations from this second VIIRS instrument have not been used in this thesis in order to maintain consistency among the data sets available that predate this new source. 

\subsection{Infrared Perimeter Observations}
Infrared perimeter observations provide sporadic information about the extent of selected fires. The resolution is on the scale of a few meters and each observation may include thousands of points describing the perimeter of a fire.  Handling of the data is sometimes problematic since dates and times can sometimes be mislabeled. These observations are typically made around midnight, local time, for fires of keen interest to fire management personnel. These observations have been used in this thesis both as source of additional information about wildfires that can be used to increase the effectiveness of the WRF-SFIRE model as well as source of information that can be used to asses the capability of the methods developed. Typically, the exact extent an location of a real-world wildfire is not known and infrared perimeter observations provide the the most accurate estimate at the time they are made.

\subsection{GOES}
Additional fire data are now becoming available from satellites in geosynchronous orbit. The time resolution is several observations per hour, but spatial resolution is around 4 km. Although not used as a data source  in this thesis, they are mentioned here because they have the potential to be used to augment the data sources outlined above. The increased temporal resolution of these observations has the potential reduce uncertainty in the fire arrival time at fire detections made by the polar satellite instruments.

\subsection{Information From Active Fire detections}

Satellite active fire detections tell us where the fire detection algorithm believes a fire to exist. The location is subject to geolocation error that grows with scan angle \citep{Nishihama-1997-ML1}.  Figure \ref{fig:geo_error} shows how the error in the reported location grows as the satellite looks further to the side from the nadir view, directly under the flight path. Associated with the location is the time of of the observation. If  fire is detected in a new location in the fire domain, most likely it did not arrive there at the exact moment of the satellite imaging. When using MODIS and VIIRS fire products, there can be many hours where the fire domain is unobserved by any instrument and the time at which a fire pixel becomes active is unknown. Without additional information, we cannot determine an exact fire arrival time and therefore we assume that the fire arrived at the pixel sometime in the previous six hours before the observation was made. This assumption follows from the shape of the function giving the probability of detection, seen in Figure \ref{fig:new_like}. The probability of detection is high and nearly constant for several hours after fire arrival and we treat fire arrival time like a uniform random variable. The future addition of GOES satellite observations, with a temporal resolution of approximately 15 minutes, will help reduce the uncertainty in these observations.

\begin{table}
\begin{tabular}{|l|c|c|c|c|}
\hline 
Platform & MODIS & VIIRS & GOES & Infrared Perimeters \\ 
\hline 
Spacial Resolution & ~ 1km  & 375m & 2 km & 10 m \\ 
\hline 
Time Resolution & Twice per day & Twice per day & 15 minutes & Sporadic \\ 
\hline 
Accuracy & Moderate & Good & Poor & Excellent \\ 
\hline 
3-$\sigma$ Error & 200m & ~350m & N/A& N/A \\
\hline
\end{tabular}
\caption[Available data sources. The final row gives  geolocation errors for MODIS and VIIRS products at nadir-view]{Available data sources. The use of GOES active fire products is currently in development. The final row gives  geolocation errors for MODIS and VIIRS products at nadir-view. Adapted from \citet{Baker-2011-JPS} and \citet{Nishihama-1997-ML1}.}
\label{table:data_sources}
\end{table}

\subsection{Limitations and Uncertainties of the Satellite Data}
\begin{figure}[!ht]
\begin{center}
  \includegraphics[width = 0.6\textwidth]{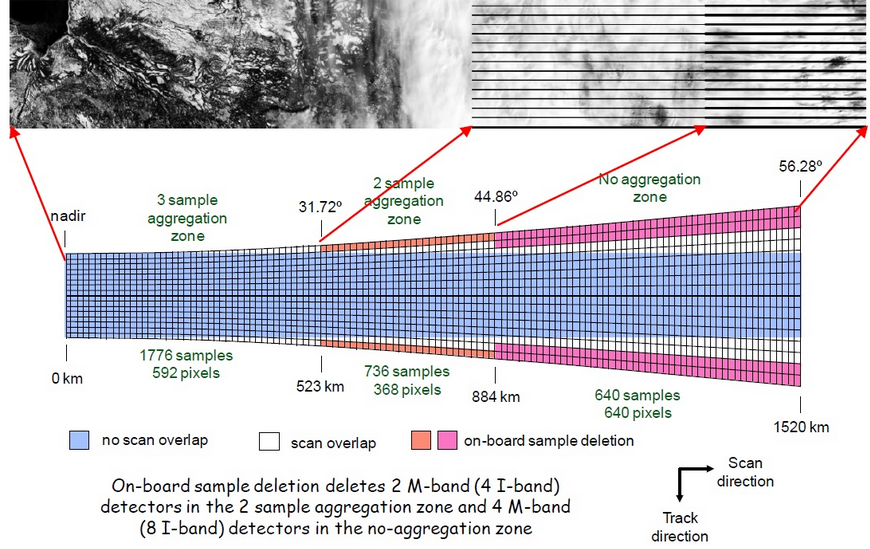}
  \caption[Effect of the scan angle on the size of fire pixels]{Effect of the scan angle on the size of fire pixels. The size of the pixels grow larger with increasing scan angle. Because of the ``bowtie shape" of the scans, there exists an overlap of areas image in successive scans. The VIIRS instrument deletes some of the overlapping pixels that are colored in the image. No deletion is performed by the MODIS instrument. From~\citet{Baker-2011-JPS}.}
  \label{fig:scan_angle}
  \end{center}
\end{figure}

\begin{figure}[!ht]
\begin{center}
  \includegraphics[width = 0.6\textwidth]{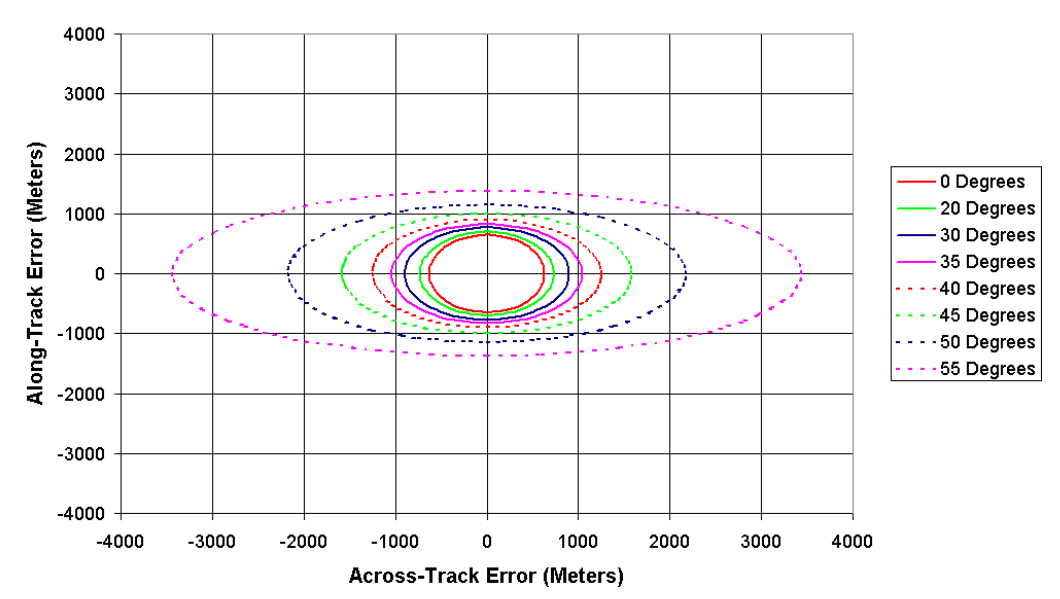}
  \caption[Geolocation error for satellite observations. As the scan angle increases, the uncertainty in the geolocation grows.]{Geolocation error for satellite observations. As the scan angle increases, the uncertainty in the geolocation grows. The ovals give an indication of the 3$\sigma$ geolocation error for the  given scan angle.  As can be seen, the uncertainty increases with scan angle and is most pronounced in the across-track direction. From~\citet{Baker-2011-JPS}.}
  \label{fig:geo_error}
  \end{center}
\end{figure}

\begin{figure}[!ht]
\begin{center}
  \includegraphics[width = 0.45\textwidth]{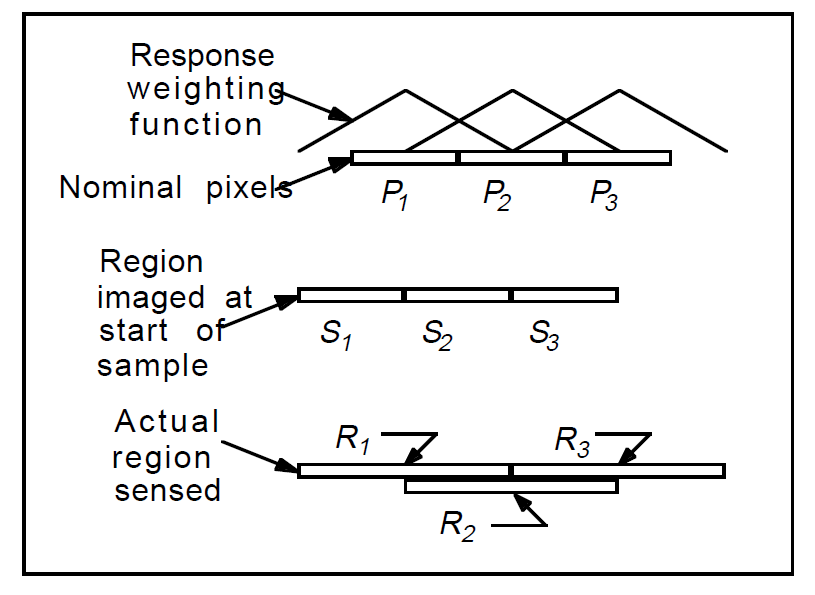}
    \includegraphics[width = 0.45\textwidth]{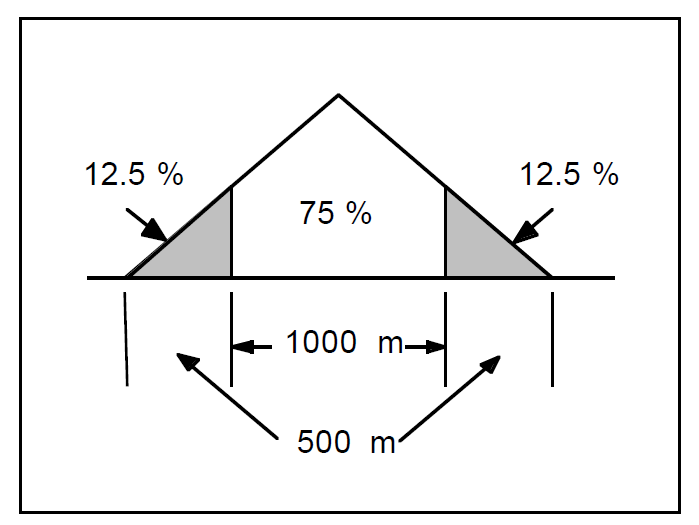}
  \caption[The triangular weighting function used by the VIIRS and MODIS instruments for collecting infrared observations.]{On the left, the triangular weighting function used in the MODIS and VIIRS instruments.  On the right, detail of how much neighboring pixels contribute to the input of  the nominal pixels in the MODIS instruments. From \citet{Nishihama-1997-ML1}.}
  \label{fig:weighting}
  \end{center}
\end{figure}

Although the satellite data available to fire researchers is impressive, care should be taken to remain aware of its associated limitations and uncertainties. Among the those are the the following.
\begin{enumerate}
\item Geolocation errors. Actual burning may be observed but the location of the burning is misreported or uncertain. The fire products delivered by satellite report an exact latitude and longitude of a fire pixel, but pixels may be large and a small fire can cause a fire detection.
\item Scan angle. Additionally, the scan angle of the observation affects the geolocation accuracy. As the satellites scans further to the side relative to the flight path, the size of area on the ground increases, leading to an increase in the uncertainty in locating small fires. Figure \ref{fig:scan_angle} shows how pixel size increases with the scan angle and Figure \ref{fig:geo_error} shows how scan angle effects the geolocation uncertainty. The oval shapes of increasing size represent the $3\sigma$ error associated with scan angles of increasing size. Thus, the smallest oval center, in the center, represents the nadir view with $0^\circ$ scan angle and the largest, outer oval corresponds to the geolocation uncertainty of an observation with a scan angle of $55^\circ$. 
\item There is an uncertainty of when the fire arrived at the reported location. The satellite tells when a fire was burning in a region but not when it arrived. A fire may have arrived at a location many hours before any satellite was making an overpass of the area in question.
\item The weighting function used by the detection algorithms implies dependence of neighboring pixels. As the instrument	scans in the direction perpendicular to the flight path, the CCD receptors in the infrared imaging device collects information from a strip of land on the ground. Because the actual region sensed is larger than the region underneath the nominal pixels, the reported location of a fire is subject to a geolocation error. For example, a hot fire near the edge of a nominal pixel may result in two fire detections being reported. Figure \ref{fig:weighting} shows the relationship between neighboring pixels.
\item Resolution. The fire data products available have resolutions of hundreds of meters but fire modeling typically takes place on grids with resolution of tens of meters.
\item Data availability. The data sets can be large and must be sent by first by satellite data link and the downloaded across the internet by the end user. In some cases, disruption of services can occur. Clouds and even smoke can obscure the satellite's view and the fire algorithm will not give an indication if areas on the ground are burning or not.
\end{enumerate}

\begin{figure}[!ht]
\begin{center}
  \includegraphics[width = 0.6\textwidth]{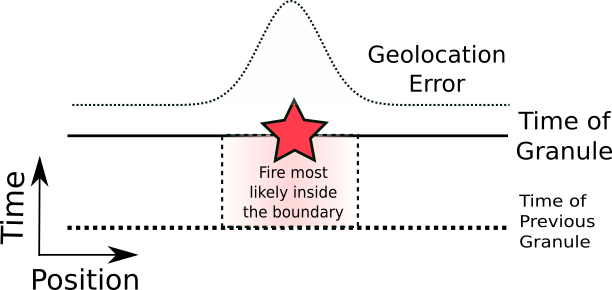}
  \caption[Active fire detections are imprecisely located spatially as well as in time.]{Active fire detections are imprecisely located spatially  as well as temporally. The star indicates the reported location of an active fire detection at the time indicated by the heavy line in the figure. The combined uncertainty means that there is a fire burning somewhere within the box shaded pink.}
  \label{fig:detection_location}
  \end{center}
\end{figure}

Figure \ref{fig:detection_location} illustrates the geolocation errors and the uncertainty in fire arrival time inherent in satellite fire detection data from polar orbiting satellites. The star in the center of the figure indicates the reported location of an active fire detection at the time indicated by the heavy line in the figure. The fire most likely arrived at the location some time before granule data was recorded, possibly as far back as the previous satellite overpass of the location on the ground. Additionally, the location of the detection is subject to an uncertainty, indicated by the ``bell curve" over the detection. The combined uncertainty means that the detected fire is most likely burning somewhere within the shaded box beneath the detection location. A more precise characterization of this spatial and temporal uncertainty will be developed in Section \ref{sec:ros_uq}.

\section{Data likelihood}\label{sec:likelihood}
This section begins the study of how the available satellite data sources are combined with the WRF-SFIRE model in order to adjust the state of a fire simulation by data assimilation techniques. The present work extends from that proposed in \citet{Mandel-2014-DAS} that adjusts the fire arrival time in the model by and additive correction found through the solution of a generalized nonlinear least squares problem. The method uses a Bayesian approach to find the maximum-a-posteriori (MAP) estimate for the fire arrival time. At the heart of the method is a data likelihood function that uses both satellite data and properties of the WRF-SFIRE burn model within the framework of Bayes Theorem to obtain the MAP estimate. The data likelihood function used for  data assimilation is a combination of two main two parts. One part is derived from an analysis of the the geolocation error associated with satellite observations and the second part is derived from the validation studies of the satellite fire detection algorithm that give a probability of detection of fire.


\subsection{Bayes Theorem}
The data assimilation framework used in this thesis follows from the typical Bayesian update problem: With a prior probability distribution describing the model output and a likelihood associated with real world observations of the system being modeled, Bayes' Theorem is used to obtain a posterior distribution that better matches the observations and contains less uncertainty than the prior.  

The process of data assimilation has a close relationship with the subject of Bayesian statistics. In particular,  Maximum A Posteriori Estimation, as detailed in \citet{Stuart-2010-IPB}, is closely related to the maximum likelihood estimation of unknown parameters in basic statistics. In Bayesian statistics a \textit{prior} is modified by using \textit{a likelihood} in order produce a \textit{posterior} which better represents reality. In the following equation, $p(A)$ represents our prior belief about the probability of some event $A$ occurring, event $B$ is some subsequent data pertaining to event $A$. The term $p(B|A)$ is the likelihood, the conditional probability of event $B$ occurring, given that event $A$ has occurred. The term $p(A|B)$ is the posterior probability, which is the conditional probability that event $A$ occurs given that event $B$ has happened. For events $A$ and $B$, Bayes Theorem states

\begin{equation}
\label{eq:bayes}
p(A|B) = \frac{p(B|A)p(A)}{p(B)}.
\end{equation}
%

In the context of wildland fire modeling, the probabilities of simple events $A$ and $B$ are replaced with probability distributions $f$. The model \textit{forecast} represents our prior belief about the system, real-world \textit{data} subsequently obtained gives us a data likelihood from which an updated estimation of the state of the fire, the \textit{analysis} is obtained, so that Bayes Theorem, Equation \ref{eq:bayes}, takes on the form
\begin{equation}
\label{eq:bayes_fire}
f(\text{Forecast}|\text{Data}) = \frac{f(\text{Data}|\text{Forecast})f(\text{Forecast})}{f(\text{Data})} .
\end{equation}

In practice, we often maximize over all possible forecasts, so the denominator, which is a normalizing factor independent of forecasts, does not need to be computed.

\subsection{Original Data Likelihood Function}
\begin{figure}[!h]
\begin{center}
\includegraphics[scale = 0.7]{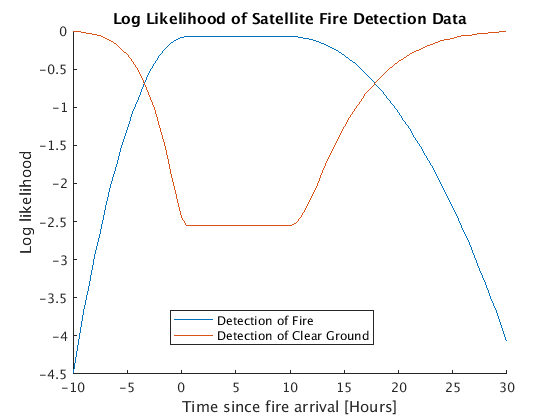}
  \caption[Log likelihood satellite detection data as function of time since fire arrival.]{Log likelihood satellite detection data as function of time since fire arrival. The log likelihood of a positive detection (in blue) is high for the first ten hours since fire arrival and then drops off slowly. Figure adapted from \citet{Mandel-2016-ASA}. Compare the shape with that of the newly proposed data likelihood function in Figure \ref{fig:new_like}.}
  \label{fig:old_likelihood}
  \end{center}
\end{figure}

The data likelihood function proposed by Mandel et al. in \citet{Mandel-2014-DAS}, and further developed in \citet{Mandel-2016-ASA}, serves as a motivation for developing the properties the likelihood function should have in order to be used in wildfire modeling. This likelihood function was designed in such a way that the likelihood of fire detection at the point $(x,y)$ is high when the fire has recently arrived, and is low otherwise. 
Conversely, the likelihood of no fire detection is high before the fire arrival time as well as for times much later, when the available fuels have been burned. Importantly, these properties of the data likelihood have been established by careful study of the validation studies pertaining to the observation instruments \citep{Schroeder-2008-VGM} . In Figure \ref{fig:old_likelihood}, we see the plot of the likelihood function $f(x,y,T^s-T)$, where $(x,y)$ represents a particular location on the ground and $T^s$ and $T$ are the time of the satellite imaging and the forecast fire arrival time, respectively. 
The likelihood of fire detection is high for the first ten hours after the fire arrival time and is low otherwise. This likelihood follows from the burn model seen in Figure \ref{fig:new_like} (a) which assumes an exponential decay in the fire heat flux as a function of time since fire arrival. Thus, given the forecast fire arrival time $T^f$, the analysis $p^a(T)$ is found by maximizing the expression over the set of satellite detection $S$

\begin{equation}
\label{eq:old_objective}
p^a(T) \propto \exp\left(\sum_S\sum_{(x,y) \in S}f_{S,x,y}(x,y,T^s-T) \right) \exp \left(-\frac{\alpha}{2}\|T-T^f\|_A^2 \right) \to \max_{T}.
\end{equation}

In practice, the log likelihood is formed from Equation \ref{eq:old_objective}, producing an equivalent minimization problem

\begin{equation}
\label{eq:old_log_like}
-\sum_S\sum_{(x,y) \in S}f_{S,x,y}(x,y,T^s-T)+\frac{\alpha}{2}\|T-T^f\|_A^2  \to \min_{T}.
\end{equation}

There are two parts to this equation. The first part is the expression
\begin{equation}
-\sum_S\sum_{(x,y) \in S}f_{S,x,y}(x,y,T^s-T)
\end{equation}
that gives the data log likelihood.  The second part,
\begin{equation}
\frac{\alpha}{2}\|T-T^f\|_A^2  \to \min_{T}
\end{equation}
represents the fire arrival time $T$ as a Gaussian function with mean $T_f$ and covariance $A^{-1}$, where $A$ is the discretized Laplacian operator 
\begin{equation}
\Delta = \frac{\partial^2}{\partial^2 x} + \frac{\partial^2}{\partial^2 y}
\end{equation}
that was chosen to enforce some smoothness properties on the analysis by imposing a penalty for a large second derivative in the analysis.

\subsection{Proposed Data Likelihood Function}
For this thesis, a newly developed data likelihood function is made from a combination of the probability that a satellite can detect a fire and the probability that a detected fire can be correctly geolocated. This function was designed by studying the fire detection algorithm validation studies \citep{Schroeder-2008-VGM} and geolocation algorithm \citep{Nishihama-1997-ML1} for the MODIS satellite products and captures the important feature of the originally proposed likelihood function, making the log-likelihood of fire detection to be low before the fire arrival time and again to be low some time after.

\subsubsection{Geolocation error}
As detailed in \citet{Nishihama-1997-ML1} there is a geolocation error associated with satellite observations. Sensor misalignment, inaccuracies in the  digital elevation model used, the triangular response function (see Firgure \ref{fig:weighting}), among other factors, lead to errors in the reported location of wildland fires. The geolocation quality of the various data sources used here is summarized in Table \ref{table:data_sources}.

We model the probability that the observed fire is at pixel $y$, given that it was reported to be at pixel $x$ as the Gaussian

\begin{equation}
P(y|x) = \frac{1}{2 \pi \sigma^2}\exp\left( -\frac{||x - y||^2}{2 \sigma^2} \right).
\label{eq:geo_error}
\end{equation}
The variance $\sigma^2$ is more complicated than just a number, as indicated here. In truth, the variance depends on the scan angle of the observation and other factors.  In this research we have used an estimate that $\sigma \approx 333$, approximating the variance seen in Figure \ref{fig:geo_error} for a scan angle of approximately $35^{\circ}$. 

\subsubsection{Probability of detection}
Following validation studies by \citet{Schroeder-2008-VGM} the probability of a satellite detection $d=1$ is modeled with the logistic curve
\begin{equation}
P(d=1) = \frac{1}{1+\exp(-aF + b)}
\label{eq:logistic_detection}
\end{equation}
where the constants $a$ and $b$ are determined by fuel characteristics and $d=1$ indicates a detection of a fire and $d=0$ indicates no fire was detected at the pixel location. This logistic curve can be seen in Figure \ref{fig:new_like} (b). We use $F = h(T)$, where $h$ is a function relating the fire arrival time to the heat flux in a fire pixel \citep{Haley-2018-DLA}. The function 
\begin{equation}
h(T) =
   \begin{cases} 
      e^{-\frac{T^s-T}{c}} & T^s \geq T \\
      0 & T^s < T
   \end{cases}  
\label{eq:heat} 
\end{equation}
assumes that the heat output in a fire pixel is zero before the fire arrival time $T$ and behaves  as decaying exponential otherwise \citep{Mandel-2011-CAF}. Here $T^s$ is the time the satellite recorded information about the pixel. The constant $c$ desribes how fast the decay in the heat output of the fire takes place and should depend on the underlying fuel properties at the observation location. The fires studied took place in regions predominately covered in timber so a single value of $c$ has been used in all applications in this work. 

The constant $a$ can be used to determine the probability of a fire detection, given a fixed amount of time since the fire arrival. Solving Equation \ref{eq:logistic_detection} for $a$, we get
\begin{equation}
\label{eq:detect_prob_a}
a = \frac{1}{h(T)}\left[ \log\left(\frac{p(d=1)}{1-p(d=1)} \right)-b\right].
\end{equation}
Setting $p(d=1) = 0.3$ and $T=24$, gives us the constant $a$ that will produce a 30\% chance of detection if the satellite images the fire location 24 hours after the fire arrival time in that location.
The parameter $b$ plays the role of determining the probability of a false detection. When, $h(T) = 0$, before the fire has arrived at the pixel coordinates, 
\begin{equation}
p(d=1) = \frac{1}{1+\exp(b)}.
\end{equation}
Solving for $b$ in terms of $p(d=1)$ gives us the probability of a false detection
\begin{equation}
b = \log\left(\frac{1-p(d=1)}{p(d=1)} \right).
\end{equation}

\subsubsection{Data likelihood function}
The data likelihood function combines the geolocation error with the probability of detection. Under this model, we assume the satellite is reporting data at location $x$ but is actually looking at location $y$.  Assuming that the probability of detection and the geolocation errors are independent random variables, we get a mixture probability 
\begin{equation}
P(d=1 \text{ at }x|T) = \int\int \frac{1}{2 \pi \sigma^2}\exp\left( -\frac{||x - y||^2}{2 \sigma^2} \right) \frac{1}{1+\exp(-ah(T(x)) + b)}dy_1dy_2.
\label{eq:pos_detection}
\end{equation}
The probability of a non-detection at pixel $x$ is then given by 
\begin{equation}
P(d=0 \text{ at }x|T) = \int\int \frac{1}{2 \pi \sigma^2}\exp\left( -\frac{||x - y||^2}{2 \sigma^2} \right)\left( 1 - \frac{1}{1+\exp(-ah(T(x)) + b)}\right)dy_1dy_2
\label{eq:neg_detection}
\end{equation}
since
\begin{equation}
\int\int \frac{1}{2 \pi \sigma^2}\exp\left( -\frac{||x - y||^2}{2 \sigma^2} \right)dy_1dy_2 = 1.
\end{equation}

\begin{figure}[!h]
\begin{center}
\includegraphics[scale = 0.5]{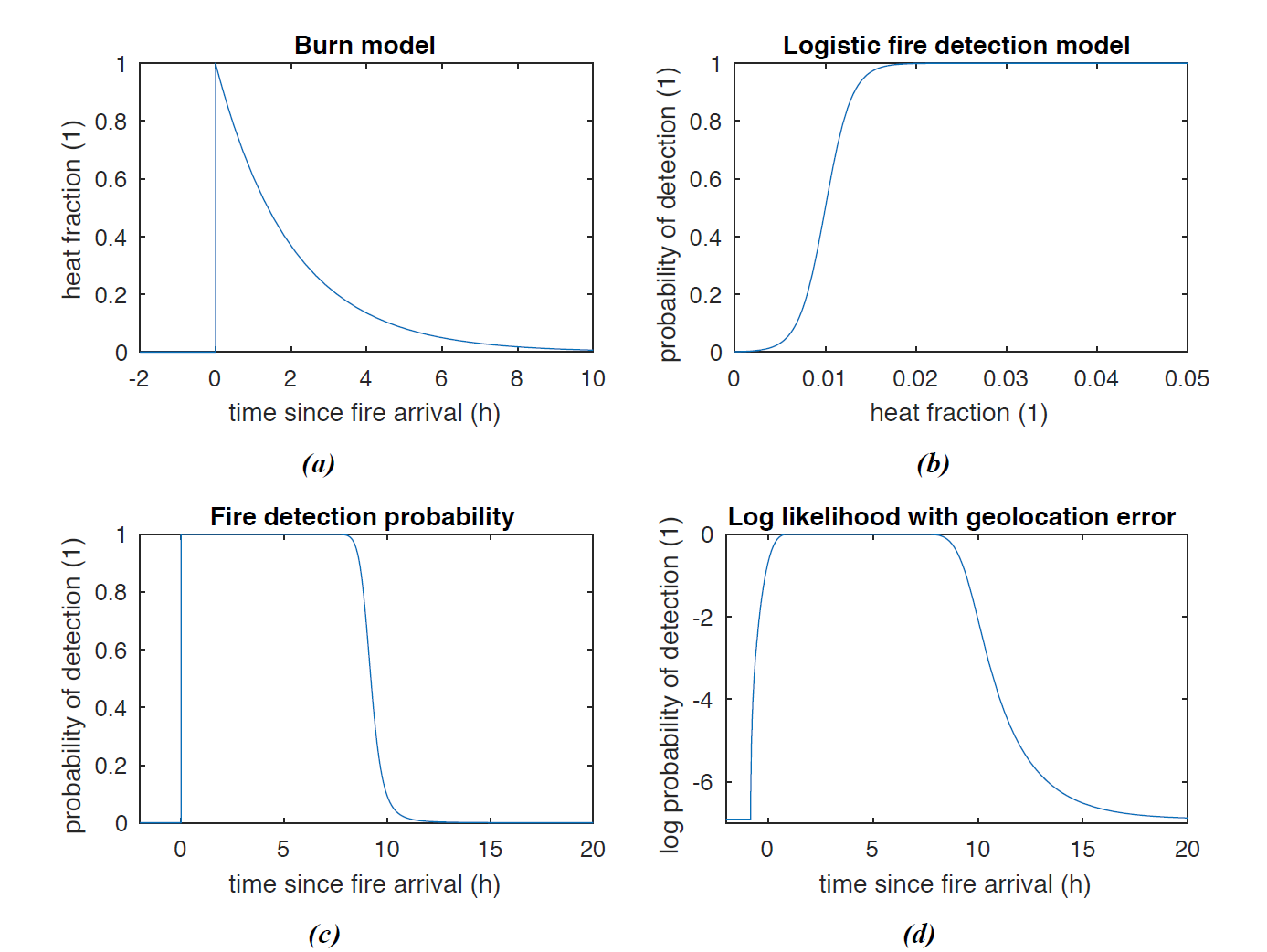}
  \caption[The likelihood function is a combination of the burn model, the fire detection model, and the geolocation error associated with each observation.]{The likelihood function is a combination of the burn model, the fire detection model, and the geolocation error associated with each observation. }
  \label{fig:new_like}
  \end{center}
\end{figure}

Figure \ref{fig:new_like} shows how the burn model, logistic curve describing the probability of detection, and the geolocation error of the observation are combined to form a curve with the properties that the probability of detection is high for a time after the fire arrival and low before and after this period. in the figure, the output of the burn model shown in panel (a) is used as an input into the fire detection model which gives the probability of detection as a function of time since fire arrival in panel (c). Multiplying by the Gaussian geolocation error and then taking the logarithm give the smoothed log likelihood  shown in panel (d). The likelihood shares the basic characteristic that likelihood is high for an initial period after fire arrival and low elsewhere. Comparison with Figure \ref{fig:old_likelihood} shows the main difference between the newer and older incarnations of the likelihood occur at the ``tails" of the function. In the older function, the likelihood shows no apparent minimum, approaching the log of a zero-possibility event. The new likelihood function has a minimum. Because of the possibility of false detection by the satellite, the probability of detection is never zero for any location in the fire domain and that difference causes a minimum of the log likelihood function to exist.

\FloatBarrier

\clearpage
\FloatBarrier

\chapter{\uppercase{Intitialization of the model from satellite data}} \label{chapter:03}




This chapter proposes methods for estimating some initial conditions of a fire from satellite data with the goal of using them to initialize a more accurate fire simulation than would otherwise be possible. The first method illustrates how the data likelihood function from section \ref{sec:likelihood} can be used to find an optimal time and ignition location for initializing a fire simulation. In practice, starting a fire simulation from an ignition point rarely used unless the fire is newly started and a location is known. In other cases, starting a simulation from an ignition point may be used as part of the process of performing prescribed burns, where advanced knowledge of the likely course of the burn is essential. The second method to be discussed uses satellite fire data to derive an estimate of the fire arrival time so that a simulation can be initialized from an advanced state that has already taken into account observations of the fire. This second method of initialization is preferred because there is an increased accuracy made available by being able to issue shorter forecasts that have less tendency to develop inaccuracies. Additionally, starting simulations from an estimate of a fire arrival time is computationally less expensive since the computational complexity of obtaining the estimate of the fire arrival time is much less than that involved with running the coupled fire-atmosphere model.


%

\section{Estimation of Ignition Point}\label{sec:ignition_estimation}

This section describes the first of two methods for obtaining initial conditions of a fire capable of producing better forecast when used in a wildfire simulation. Having a good estimate of the time and location of the ignition point of the fire gives a simulation a better chance of making an accurate forecast. Usually the location of the ignition cannot be known precisely. Unless the start of the fire was directly observed and its location recorded and shared by a human observer, the first indication of a wildfire starting might come from satellite observation. In \citet{Benali-2016-DFD} the first satellite detection associated with a particular fire even was taken to be the ignition point. Both the spatial and temporal resolutions of the satellite fire detection products limit the precision of the location and time of the ignition point using such a method. The proposed method for estimating the ignition point uses a grid search technique to find the latitude, longitude, and time of ignition from a larger set of fire data than a single active fire detection recorded by satellite. An ensemble of short term forecasts is made and the estimated ignition point is found from the forecast that has the maximum data likelihood,using the data likelihood function from Section \ref{sec:likelihood}. 

Because it is better to start wildfire simulations simulations from an estimate of the fire arrival time, this method for estimating the time and place of a fire ignition is unlikely to see much use in operational forecasting. However, the procedure may have other uses. For example, when collecting information about wildfires, the time and place of ignition is always a concern and something to be noted. Unless the start of the fire was observed, fire investigators may need to journey to the site to perform an investigation. In cases where the fire was especially remote, the investigation could be arduous and costly and the method outlined in this section could be used to obtain a satisfactory estimate.

\subsubsection{The Standalone Model}
The standalone model within WRF-SFIRE is used to make  short-term forecasts with minimum computational time. Using the standalone model, the fire spread component of WRF-SFIRE is not coupled to the weather component and can be used to quickly make short-term forecasts. Instead, one model run of the weather model is made for the duration of the simulation period and it is used for the weather input for each instance of an ensemble of fire forecasts. The assumption is that, for small periods of time and small fire regions, the effects on the fire will have a minimal impact on the weather since less heat will be released than would be the case for large fires. Running the standalone model allows for a large ensemble of forecasts to be run in the same time that only small number of forecasts could be run using the full, coupled model.

\subsubsection{Grid Search for Optimal Ignition Location}
The ignition point of a fire may be estimated by running an ensemble of forecasts and finding the forecast that maximizes the likelihood of the satellite observations. To begin, a set of simulations is made using the standalone model, each beginning at a unique point on a two dimensional grid of latitude and longitude coordinates and at a specific time. Each simulation is allowed to run for a fixed time and then the  likelihood of the satellite data recorded during that time is computed. The ignition point may be estimated as the location whose forecast maximizes the data likelihood. If a more precise location is required, the process may be repeated on a refined grid in the neighborhood of the estimated ignition location.

The process for estimating the ignition point from satellite active fire detection data is as follows.
\begin{enumerate}
\item Set up the fire domain and run WRF-SFIRE without any ignition point specified. The fire spread model is  not run and this generates the weather conditions that will be used in the  standalone model.
\item Set up a grid of ignition locations. The locations should form an array of latitudes, longitudes, and times of the fire ignition point.
\item For each entry in the grid, run the standalone model forward and save the forecast fire arrival time for later analysis.
\item Compute the data likelihood of the satellite fire data, given the forecast fire arrival time, for each ignition point in the grid.
\item Find the location of the maximum data likelihood from all the computed data likelihoods. This is the estimated ignition point. If desired, the location of the maximum data likelihood can be used to make a refined grid and the process continued from step 2.
\end{enumerate}

\subsubsection{Example: Test Case of a Hypothetical Fire}
As a first test of the method, a simple, hypothetical fire was used as the ground truth and artificial  fire detections were put in the fire domain. The grid search method was used to estimate the ignition point using the data likelihood function. To test whether this likelihood function can be useful in practice we first used a grid search technique to find the time and spatial coordinates of the ignition giving the best data likelihood of a simulated fire occurring over flat topography with homogeneous fuels and no winds.  The perimeter of such a fire would expand outward from its ignition point the like the ripples caused by a stone dropped into a pond of water and we model the progression of the fire perimeter by the equation of a cone

\begin{equation}
\label{eq:first_cone}
T(x_1,x_2) = \sqrt{(x_1-500)^2 + (x_2-500)^2} + 30.
\end{equation}

\begin{figure}[!h]
\begin{center}
\includegraphics[scale = 0.6]{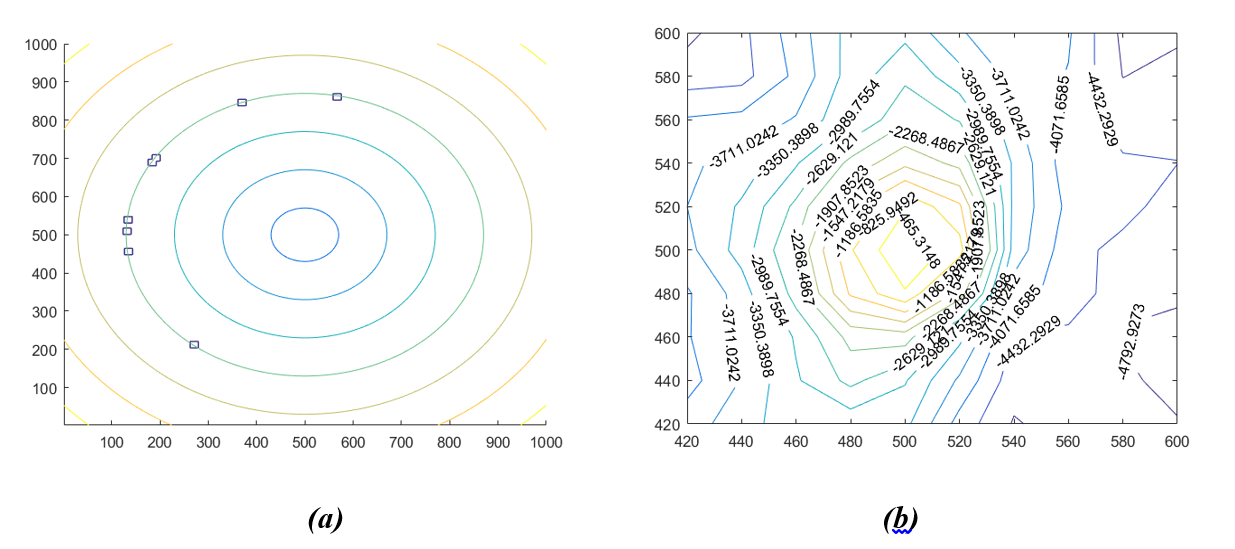}
  \caption[Estimating the ignition point for a fire with no wind in homogeneous fuel.] {Estimating the ignition point for a simulated fire with no wind in homogeneous fuel. On the left, the concentric circles represent the fire perimeter at various times and the squares represents locations of artificial fire detections. On the right is the contour map of data log-likelihood for simulated fire with ignition time $t = 20$.  The actual ignition point of the simulated fire was at the point (500,500).}
  \label{fig:exp_1}
  \end{center}
\end{figure}

Working on a $1000 \times 1000$ grid, this equation gives the fire arrival times for simulated fire starting a location $x = (500,500)$ at $T=30$. Several artificial satellite fire detections were then placed at various points near the simulated fire perimeter corresponding to the time $T = 300$ as shown in Figure \ref{fig:exp_1}. In a real-world setting, these detections might be all the information about a fire that is available. The left panel of Figure \ref{fig:exp_1} shows contour lines of the artificial fire arrival time made according to Equation \ref{eq:first_cone} with the small squares in the figure giving the location of the artificial satellite fire detections.

To determine the time and place of ignition of the simulated fire, we then duplicated the fire arrival time in Equation \ref{eq:first_cone}, with changed ignition point and time. The data log-likelihood of each of the perturbed fire arrival times was then computed using Equation \ref{eq:old_objective}. For this first test, a collection of 500 simulations was made over a 10-by-10 spatial grid at five separate times. The data log-likelihood for each simulation was then calculated. The maximum  data log-likelihood  was found to correspond to the simulation with ignition point $(500,500)$ and ignition time $T=20.$ The contour map of data log-likelihoods for the ignition time $t = 20$ is shown in Figure \ref{fig:exp_1}. This estimate of the time and place of ignition gives the correct spatial location  but misses the time of ignition, giving an earlier estimate of the true time of ignition. This is most likely due to the fact that areas within and close to a fire perimeter have a high probability of detection and areas outside the perimeter have a low probability of detection since they correspond to a forecast fire arrival time in advance of the time that fire was observed in the location.  Therefore an earlier ignition time would create a larger fire giving a perimeter that encompasses more of the area of the detection pixels, leading to a higher data likelihood.

\subsubsection{Example: Test Case in the WRF-SFIRE ``Hill Experiment"}

\begin{center}
\begin{figure}[h!]
  \includegraphics[scale = 0.7]{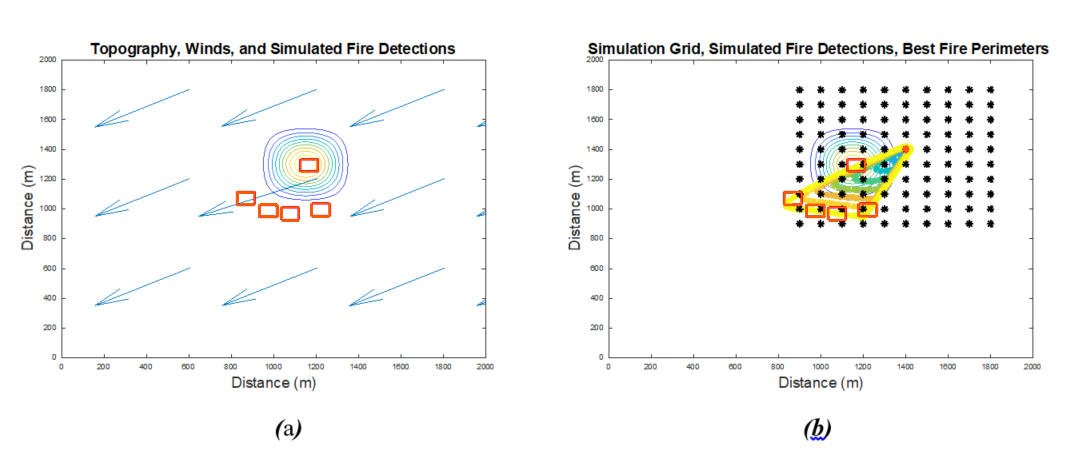}
  \caption[Estimating the ingition point using the ``Hill Exeperiment" distributed with the WRF-SFIRE modeling software.]{Estimating the ignition point using the ``Hill Experiment" distributed with the WRF-SFIRE modeling software. On the left, the topography, winds, and artificial satellite detections used for the ``Hill Experiment." The contour lines show the topography of the hill in the center of a square domain with winds blowing in the direction of the vector arrows. The red rectangles are the artificial satellite detections placed near the perimeter of a simulated fire. On the right, the grid of ignition locations and fire perimeter of the simulation with the largest data likelihood. Simulated fires were ignited at each of the black dots in the grid. The thick, colored contour lines correspond to the fire perimeters of the simulation with the largest data likelihood.}
  \label{fig:exp_2}
\end{figure}
\end{center}

A further test of the data likelihood function was made using the WRF-SFIRE model. This time, an experimental setup call the ``Hill Experiment" within WRF-SFIRE was used to created a simulated fire. The ``Hill Experiment" creates a simulation of a fire occurring in a square-shaped region with sides 2 km long, containing a small hill of a height 100 meters in the center. For the fire simulations, a constant wind blowing from the northeast was present. The left panel of Figure \ref{fig:exp_2} shows the topography, winds, and locations of the simulated fire detections used for this experiment. In this figure, the arrows give the wind direction, the concentric contour line give the locations of the hill, and the orange squares outline the location of the simulated fire detections. 

Like the first test, an initial fire was simulated and then artificial satellite detections were placed along the fire perimeter. In this experiment the ignition point was chosen to be at the coordinated $(x,y) = ( 1400m,1400m )$ at time $t=60 s$. Artificial satellite detections were then placed manually along the perimeter corresponding to the fire arrival time $t = 400s.$

With the artificial fire scenario established, we then ran 300 fire simulations, in a $10 \times 10$ grid, at three separate times, in order to determine if the method of estimating the fire ignition point and time by maximum data likelihood would work. The right panel of Figure \ref{fig:exp_2} shows the grid locations of the ignitions used in the 300 fire simulations as black asterisks. The orange asterisk in the center of the grid pattern shows the location of the ``true ignition" point and the colored contours emanating for this ignition point are the contours of the fire arrival time. 

\begin{figure}[h!]
\begin{center}
  \includegraphics[scale = 0.7]{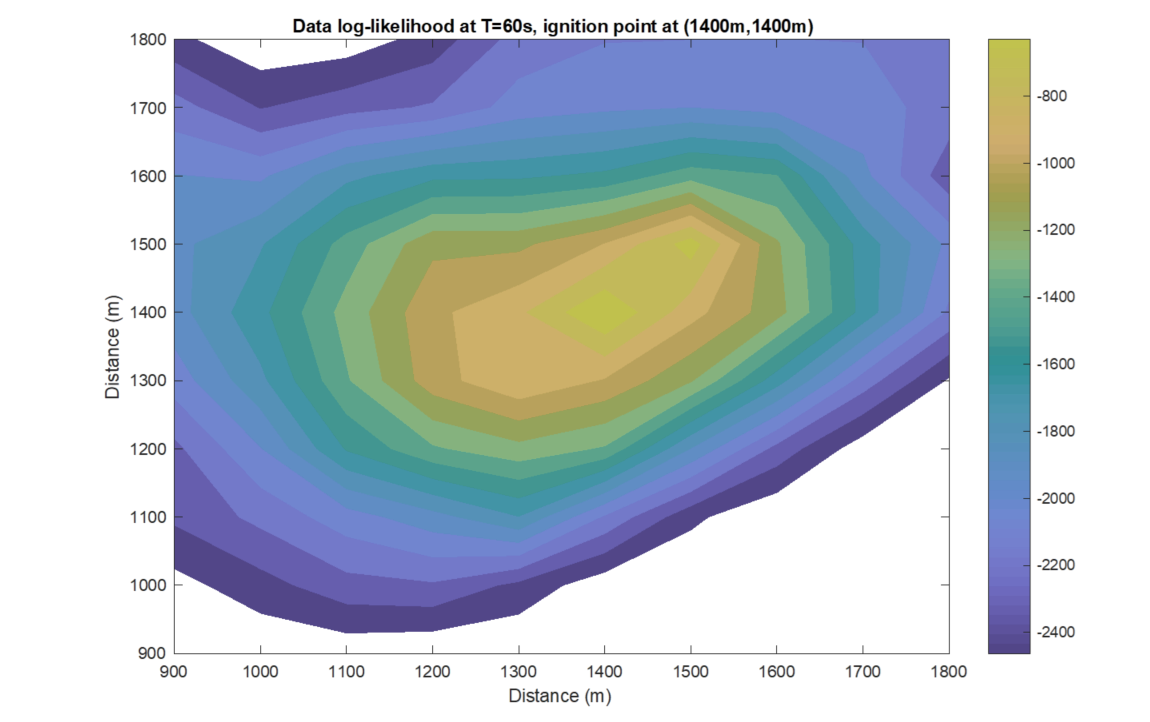}
  \caption[Contour map of data log-likelihood for ignition points of the WRF-SFIRE Hill Experiment.] {Contour map of data log-likelihood for ignition points of the WRF-SFIRE Hill Experiment corresponding to ignition time $t = 60s$. In this case, the maximum likelihood gives the correct time and location of the fire ignition, at the location with coordinates $(1400 m, 1400 m)$.}
  \label{fig:exp_2a}
  \end{center}
\end{figure}

With the 300 fire simulations completed, the data log-likelihood from each was computed and the maximum of these values was used to determine the estimated time and location of the ignition. In this experiment, the maximum data likelihood of all simulations belonged to the simulation which exactly matched the ignition point and time of the ``true fire" that was used to place the artificial satellite detections in the fire domain. Figure \ref{fig:exp_2a} shows a contour map of the likelihoods corresponding to ignitions with $t = 60s$.

\subsubsection{Example: Estimating the Ignition Point of the Patch Springs Fire}
As a final test of the method, we attempted to estimate the ignition location of a real-world fire using actual satellite observations from both VIIRS and MODIS instruments. The fire was known as the ``Patch Springs Fire" and occurred southwest of Salt Lake City, Utah in August of 1993. No ``true" location of the ignition location was known but the estimated ignition point  was compared to an ignition point estimated by fire investigators \citep{Inciweb-2013-Patch}). Our process of estimating the ignition point differed from that of investigators by several kilometers, but the time was off by one hour. In cases where it is difficult or expensive to have fire investigators determine the time and location of the ignition of a wildfire, the grid search method we have demonstrated could be used to good effect.

\begin{figure}[h!]
\begin{center}
  \includegraphics[width = 0.45\textwidth]{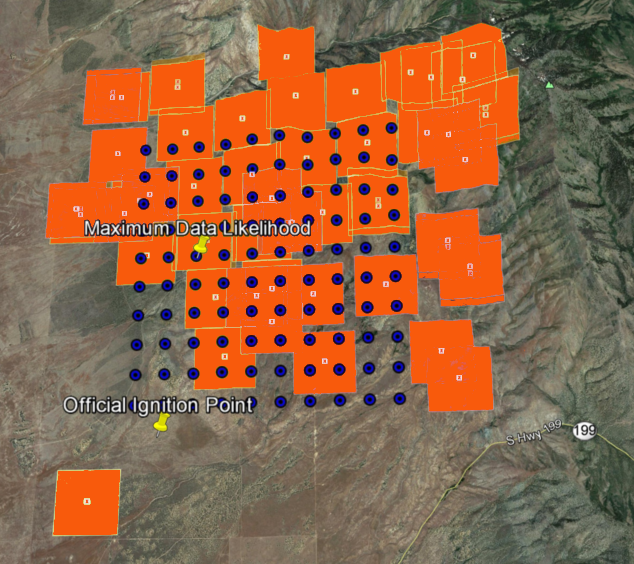}
  \includegraphics[width = 0.45\textwidth]{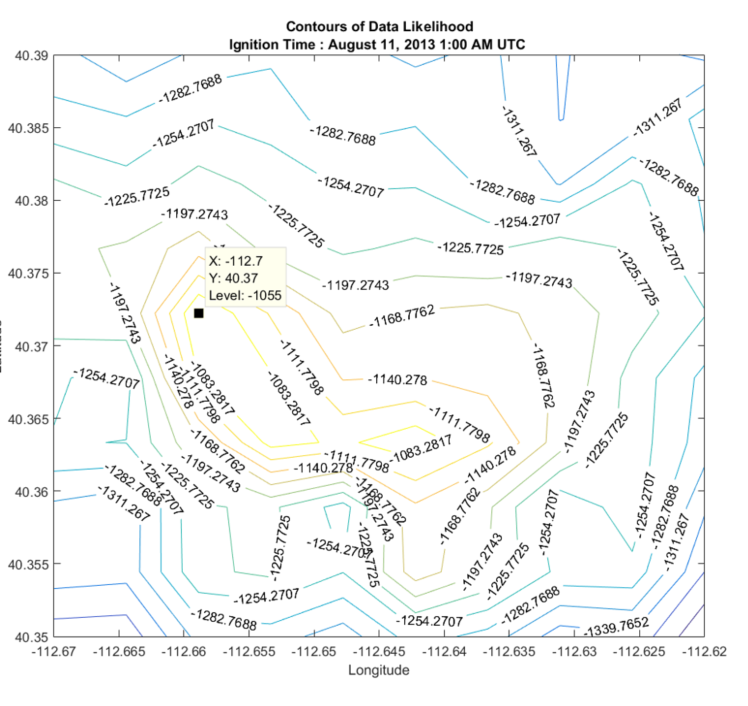}
  \caption[Satellite fire detections and WRF-SFIRE ignition locations used for estimating the true ignition point of the Patch Springs Fire.]{Satellite fire detections and WRF-SFIRE ignition locations used for estimating the ignition point of the Patch Springs Fire. The official estimate of the ignition point lies just outside of the grid of simulations and the estimate of the ignition point obtained by maximum data likelihood lies nearly 3 km to the north. On the right are the  contour lines of the computed data likelihoods for locations in the grid corresponding to the ignition time $1:00$ UTC.}
  \label{fig:exp_3a}
  \end{center}
\end{figure}

By the  process outlined in the previous experiment, an ensemble of simulations was run and the ignition point was estimated by the grid search method after evaluating the data likelihood for each simulation. For this experiment, the standalone model was used for producing the ensemble of forecasts after having first run WRF to produce the weather output for three days of simulation time. In this case, 1000 simulations were run using the standalone model on a $10 \times 10$ grid, over 10 different ignition times. Figure \ref{fig:exp_3a} shows the terrain, grid layout, and satellite fire detections (as orange squares) used to estimate the ignition point.

The ignition time and location of the simulation with the maximum data likelihood was found to be at $40.372^{\circ}$N, $-112.659^{\circ}$W at $01:00$ UTC on August 11, 2013. This is about 3 km from the official location of the ignition point determined by investigators at $40.341^{\circ}$N, $-112.67^{\circ}$W. The time of ignition we found differs from that estimated by investigators by one hour. The ignition time reported by investigators was $02:00$ UTC. The yellow pushpins in Figure \ref{fig:exp_3a} show the estimated ignition point, denoted as ``Maximum Data Likelihood" and the ignition location determined by fire investigators, denoted as ``Official Ignition Point." The right panel of Figure \ref{fig:exp_3a} shows the contour lines of the data likelihood for locations in the grid corresponding to the ignition time $1:00$ UTC.

The location determined by investigators lies outside the boundary of the grid of ignitions points used in this experiment. It is unlikely that expanding the grid to encompass this ignition point would have produced a different result. This fire exhibited slow growth for several days and then suddenly expanded rapidly on the third and fourth day. Most of the satellite detections collected during the first three days of the fire are at a distance from the ignition point determined by investigators, but the estimated ignition point was found by an ignition point closer to the locations of the fire detections recorded on the third day of the fire. Starting a simulation new where the bulk of the fire detections was recorded produced a better forecast in this case. 

\section{Estimating Fire Arrival Time and the Rate of Spread}\label{sec:fire_arrival}

The pattern of satellite active fire detections in a region gives clues about the behavior of a fire and its history in a region. Due to the resolution limitations of the data, both spatial and temporal, a complete picture of the fire's history cannot be known. For example, when using data from the polar orbiting instruments, there periods lasting many hours when a fire is unobserved. These gaps in the data, and those caused by other factors such as cloud cover, represent a challenge that many interpolation techniques cannot overcome. For example, ordinary kriging was used in \citep{Veraverbeke-2014-MDP} to derive spatially continuous fire arrival times using active fire detection data from the MODIS instruments. A machine learning technique was used in \citet{Farguell-2021-MLE} to find an estimate of the fire arrival time  of wildfires using fire data from both the MODIS and VIIRS instruments. Importantly, both detections of active fires and locations without fire are used in this process.  In this section, we describe a method to estimate the fire arrival time using the MODIS and VIIRS data. The method developed in this thesis uses mathematical concepts from graph theory to create a set of plausible paths that fire could take from an assumed ignition point to all satellite active fire detections in a prescribed fire domain. An interpolation technique is used to overcome the sparsity of active fire detection data by interpolation along the set of paths. Assignment of a fire arrival time to points in the fire domain near the data points on the fire paths is accomplished by a local averaging method. Finally, data about non-fire locations is used to constrain the outer perimeter of the estimated fire arrival time. In Chapter \ref{chapter:04} The estimated fire arrival time will be used to initialize fire simulations from an estimated perimeter. In Chapter \ref{chapter:05}, the ROS of the fire will estimated from the fire arrival time, allowing for adjustment to model parameters like fuel moisture content.

\subsection{Overview of Fire Arrival Time Estimation}\label{sec:fat_method}
We propose to estimate the fire arrival time in a region of interest by a method that attempts to trace the fire's progress from an assumed ignition point to all active fire detection locations in the domain. Interpolation and localized averaging of the fire arrival time at the data locations is then used to then provide estimates of the fire arrival time at nearby locations. The non-fire locations recorded by the satellite are used in this process to help establish the outermost perimeter of the estimated fire arrival time. The method will be described as an algorithm enumerated below. 

\begin{enumerate}
\item Organize and spatially cluster the fire detection data.
\item Construct a directed graph connecting detection locations. 
\item Make a shortest path from the assumed ignition location to all other fire detection locations in the domain
\item Interpolate extra points along the paths.
\item Adjust the fire arrival time over the whole fire domain with iterative interpolation.
\item Use non-fire pixels to constrain the outer perimeter of the fire arrival time.
\end{enumerate}

\subsection{Organizing the Data}

The first step to estimating the fire arrival time from observations is to obtain and organize the fire detection data. The following steps are taken.
\begin{enumerate}
\item Set up the fire domain. This entails setting geographic and temporal limits of the simulation. This can be done by hand or in an automated way by using WRFX system.
\item Collect the data and estimate the ignition point of the fire.
\item Interpolate the detection data onto a computational grid.
\item Cluster the data spatially. This is done to impose a structure on the data that will allow a plausible path for a fire to follow from the ignition point to other locations in the fire domain where fire was observed.
\item Make the set of shortest paths from ignition to all other active fire detections.
\end{enumerate}

\subsubsection{Set up the simulation domain}
This can be done by hand or by using the WRFX web interface. The domain will be defined by a bounding rectangle describing the minimum and maximum of both the latitude and longitude of the fire domain. Additionally, the time of the fire is considered and a starting and ending time are part of the domain boundaries. In principle, these boundaries are all that are needed to construct a plausible fire arrival time from detection data. The workflow used here read the domain boundaries automatically from a file output by the WRF-SFIRE model.

\subsubsection{Collecting data}
Several tasks are needed to acquire the data used by the method.
\begin{enumerate}
\item Download satellite detection data. This can be done using the tools in WRFX or the data can be downloaded manually from the NASA EarthData collection \citep{Nasa-2021-EOD}. The MODIS and VIIRS Level 2 (L2) fire products and their associated geolocation files have been used.
\item Subset the detection data for the fire domain. The L2 data products contain data in sets called granules that are much larger than typical fire domains used in simulations. Only data within the established fire domain need be considered.
\item Filter out low confidence fire detections. For each active fire detection, there is an associated confidence level assigned \citep{Giglio-2015-MC6}. Only detections with a confidence level of 70 and above have been used in this research.
\item Sort the data by increasing time of the observations. Each satellite fire detection will have an associated latitude,longitude, and time of the observation. 
\item If using infrared perimeters, decide how many points per perimeter to use. Perimeter observations can be obtained in many data formats but the perimeter data is usually stored as a list of ordered pairs giving the latitude and longitude of locations on the fire perimeter. There may be thousands of points in a single observation.
\end{enumerate}

\begin{figure}[!ht]
\centering
\includegraphics[width=0.45\textwidth]{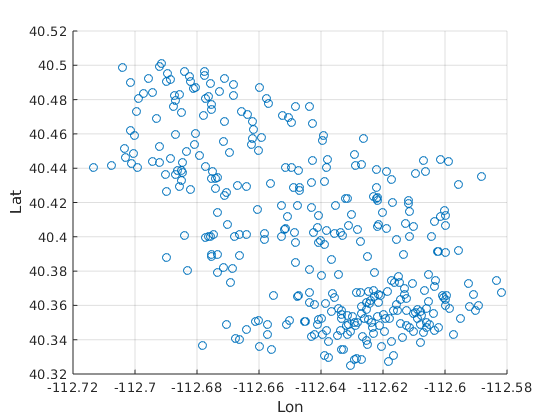}
     \hfill
\includegraphics[width=0.45\textwidth]{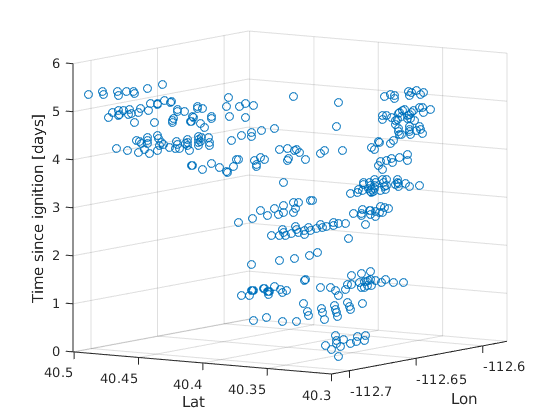}
     \caption[MODIS and VIIRS active fire detections for the Patch Springs fire of 2013.] {MODIS and VIIRS active fire detections for the Patch Springs fire of 2013. On the left, seen as a collection of points in a two dimensional grid of latitudes and longitudes. On the right, seen in a three dimensional setting with time along the vertical axis.}
\label{fig:patch_detections_no_graph}
\end{figure}

\subsubsection{Interpolate detection data onto a computational grid}
\begin{figure}[!ht]
\centering
\includegraphics[width=0.45\textwidth]{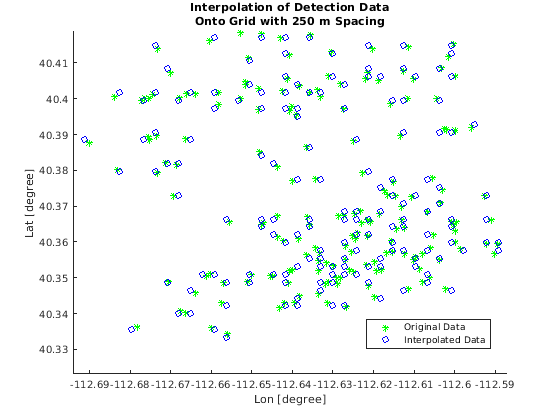}
     \hfill
\includegraphics[width=0.45\textwidth]{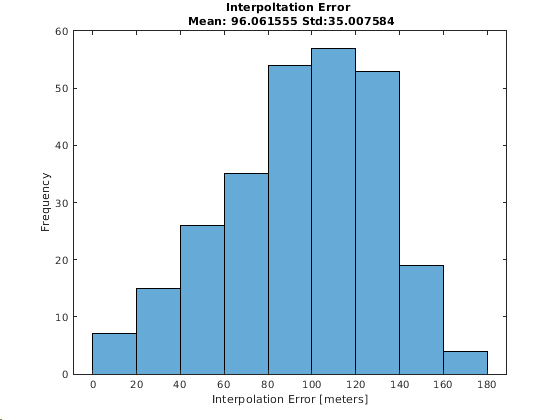}
     \caption[MODIS and VIIRS active fire detections interpolated onto a common grid.] {MODIS and VIIRS active fire detections for interpolated onto a common grid. On the left, the green stars in the figure show the locations of active fire detection pixels as reported by the satellites. The blue circles show the locations of these data points when interpolated onto a uniform grid with a 250 meter spacing between nodes. The right figure shows a histogram of the distances between the original data location and the location after the data points have been interpolated. }
\label{data_interp_error}
\end{figure}

The L2 fire product from the MODIS and VIIRS platforms is accessed through a set of two files. One file contains geolocation information and the other contains the fire products. The data for each satellite granule is scattered in arrays and must be interpolated onto a common computational grid for use. The interpolation of the satellite data onto a grid is accompanied by the interpolation of the relevant files output by WRF-SFIRE onto the same grid. Typically, the the fire model in WRF-SFIRE runs on a grid with spacing of tens of meters, but the default spacing for the interpolated data used by this method is 250 meters. This change of scale in the WRF-SFIRE fire model data does represent a loss of resolution. However, the resolution of the fire data is 750 meters and the choice of the default value for the interpolated grid size represents a compromise between maintaining the resolution of the WRF-SFIRE fire model and overfitting the uncertain fire detection data. Additionally, the coarser grid spacing of the interpolated data makes for large increases in the speed of the computations needed by the algorithm. Figure \ref{data_interp_error} shows how the interpolation data onto the computational grid with 250 meters introduces an error. The left panel gives a sense of how the locations of the data points are moved from the reported pixel location to the location of the nodes in the computational grid. The right panel is a histogram of the errors, reported as the distance the pixel moved, in a histogram. The mean of the errors in this example was found to be 96 meters. 

\subsubsection{Make a Directed graph}
\label{sec:graph}
The active fire detections are first organized into a directed graph so that a path taken by the fire from an assumed ignition point to all the fire detection locations in the domain may be constructed. For completeness, a brief overview of some of the basic concepts from graph theory is first given.

\begin{figure}[!ht]
\begin{center}
  \includegraphics[width = 0.6\textwidth]{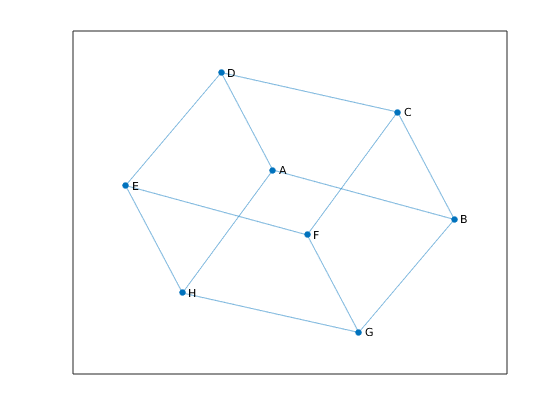}
  \caption{A graph derived from a cube. Each corner of the cube is connected to three others by an edge. }
  \label{fig:cube_graph}
  \end{center}
\end{figure}

Graph theory is the study of mathematical structures that define pairwise relationships between objects. A common notation for a graph is $G=(V,E)$ where $V$ is a set of vertices and $E$ is a set of edges that show a relationship between the vertices. For example a cube can be thought of a set of 8 vertices, representing the corners, connected by 12 edges. Each corner is connected to three others by an edge. In Figure \ref{fig:cube_graph}, $V=\{A,B,C,D,E,G,H \}$ and $$E = \{(A,B),(A,D),(A,H),(B,C),(B,G),(C,D),(C,F),(D,E),(E,F),(E,H),(F,G),(G,H)\}.$$

\begin{figure}[!ht]
\centering
       \includegraphics[width=0.45\textwidth]{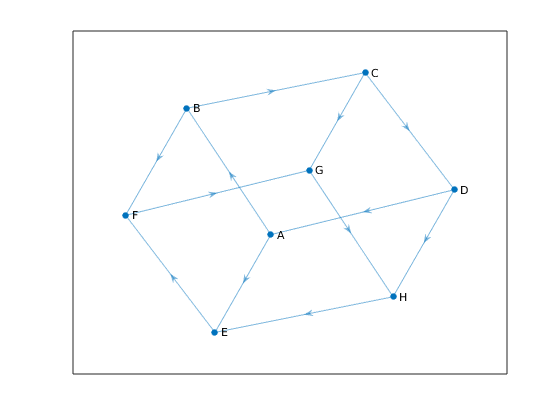}
     \hfill
       \includegraphics[width=0.45\textwidth]{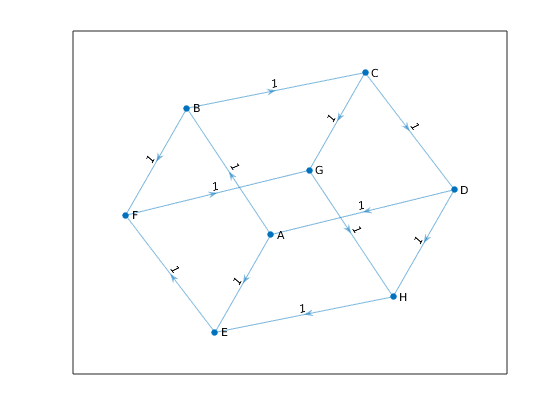}
     \caption{Directed graphs derived from a cube. Each corner of the cube is connected to three others by an edge. On the left no weight is assigned to any edge. On the right, all edges have a weight assigned.}
\label{fig:directed_graphs}
\end{figure}

A \textit{directed graph} imposes an orientation on the edges in a graph. Figure \ref{fig:directed_graphs} depicts a cube with edges that now have an associated direction. Each vertex is still connected to three others, but the direction along the edges denotes a more nuanced relationship between connected vertices. With arrows indicating a direction of travel, we see that direct travel is possible between vertices $A$ and $D$, but only in the direction from $D$ to $A$.

As a further refinement, each edge in the directed graph may be assigned a \textit{weight}. In the context of this work, the edge weights will thought of as a distance between connected vertices. Figure \ref{fig:directed_graphs} shows the directed graph derived from the unit cube. Table \ref{tbl:cube} characterizes the directed graph with edge weights as a matrix of values organized so that the ordered pairs of rows and columns give the edge weight of connected vertices. With directions and edge weights assigned to the graph, a key question will be how to travel from one vertex to another by the shortest path possible. In some cases, there may be more than one path between vertices. In other cases, there may be no path from one vertex to another.

\begin{table}
\begin{tabular}{|c|c|c|c|c|c|c|c|c|}
\hline 
 & A & B & C & D & E & F & G & H \\ 
\hline 
A & - & 1 & - & - & 1 & - & - & - \\ 
\hline 
B & - & - & 1 & - & - & 1 & - & - \\ 
\hline 
C & - & - & - & 1 & - & - & 1 & - \\ 
\hline 
D & 1 & - & - & - & - & - & - & 1 \\ 
\hline 
E & - & - & - & - & - & 1 & - & - \\ 
\hline 
F & - & - & - & - & - & - & 1 & - \\ 
\hline 
G & - & - & - & - & - & - & - & 1 \\ 
\hline 
H & - & - & - & - & 1 & - & - & - \\ 
\hline 
\end{tabular}
\caption[Table of edge weights for the directed graph derived from the unit cube. The rows and columns give the vertices in the directed graph.]{Table of edge weights for the directed graph derived from the unit cube. The rows and columns give the vertices in the directed graph of Figure \ref{fig:directed_graphs}. The table is organized so that the ordered pairs of rows and columns give the edge weight of connected vertices. For example, the first row may be interpreted as indicating that edges with weight 1 connect vertex A to both vertex B and vertex E.}
\label{tbl:cube}
\end{table}

The concept of the directed graph is used in the context of satellite active fire detections to help impose a structure on the data that can be used to infer properties about the fire behavior. Under the assumption that fire will spread from the ignition point to other fire locations in the domain by a shortest path that passes through intermediate fire detection locations, a directed graph is constructed with all of the detection locations used as vertices and the ``great circle" distance between them is given as the edge weight. The terms edge weight and distance will be used synonymously in the discussion that follows.

If there are $n$ active fire detections in set of active fire detections, the first step in making the set of shortest paths from the ignition point to all other detections begins with construction of a distance matrix $D$. The entries in this matrix contain the edge weights for a directed graph to be created. Each element in $D$ is a distance between two fire detections that are the vertices in the graph to be constructed. Thus, $D_{ij}$ is the distance between fire detection $i$ and fire detection $j$. We note that $D=D^T$ and at this point the matrix records the distance between every pair of fire detections in the domain. In principle, the distance can be measured in many ways. For the techniques presented here, the distance will be measured as the ground distance separating the locations of the active fire detections using the WGS84 Ellipsoid since the locations are derived from L2 data products using that coordinate system. 

\begin{figure}[!ht]
\begin{center}
\includegraphics[width = 0.40\textwidth]{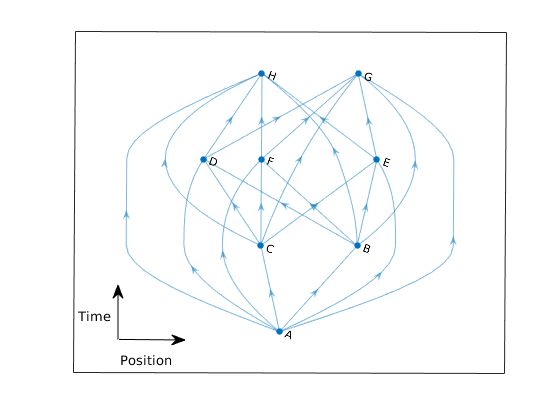}
\includegraphics[width = 0.40\textwidth]{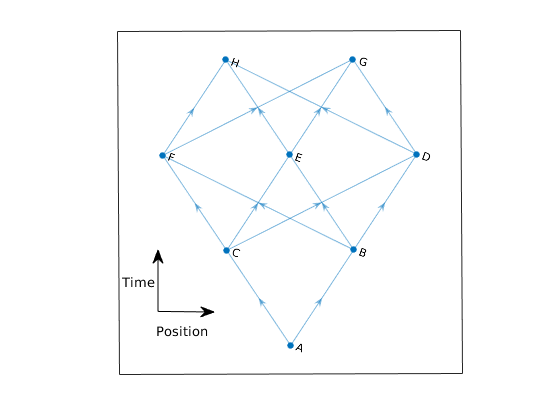}
  \caption[Imposing structure on a directed graph by limiting the allowable edges connecting vertices.]{Imposing structure on a directed graph by limiting the allowable edges connecting vertices. On the left, the graph has edges connecting to all vertices found above. The vertex $A$ is connected to all other vertices in the graph. If all edges in the graph have the same same weight, the shortest path from $A$ to $G$ is along the edge $(A,G)$. On the right, the edges connect only to vertices on the ``next level up." Vertex $A$ is connected only to vertices $B$ and $C$. The shortest path from $A$ to $G$ now must move along three edges. This kind of rule for controlling the can be considered a clustering of data temporally. The spatial clustering used here would allow for a path from $A$ to $G$ that possibly passes through point $E$.}
  \label{fig:graph_up_all}
  \end{center}
\end{figure}

\begin{figure}[!ht]
\centering
\includegraphics[scale=0.6]{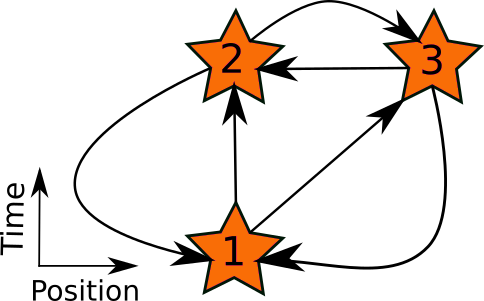}
\hspace{10mm}
\includegraphics[scale=0.6]{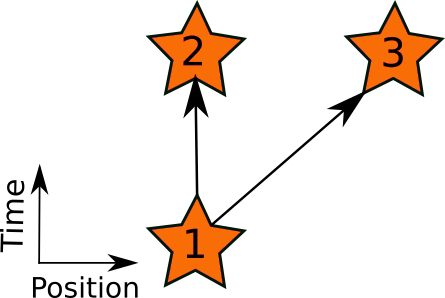}
     \caption[The direction  of edges in a graph of active fire detections.]{The direction  of edges in a graph of active fire detections. On the left, all possible edges connecting disting vertices are shown. On the right, a rule only allowing edges to connect early to subsequent detections simplifies the graph and corresponds to the way fire in one location spreads to other locations at a later time. Detections 2 and 3 were recorded at the same time and no edge connects them.}
\label{fig:edge_directions}
\end{figure}

\begin{figure}[!ht]
\centering
\includegraphics[width=0.45\textwidth]{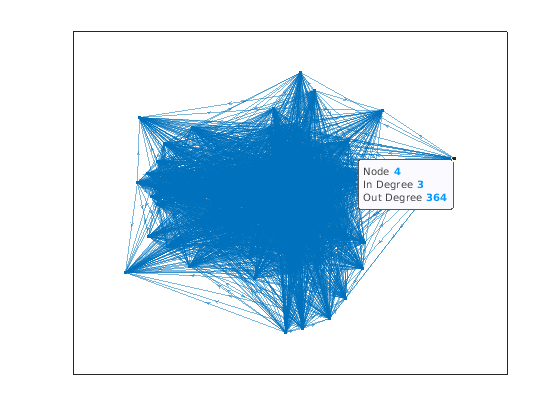}
\includegraphics[width=0.45\textwidth]{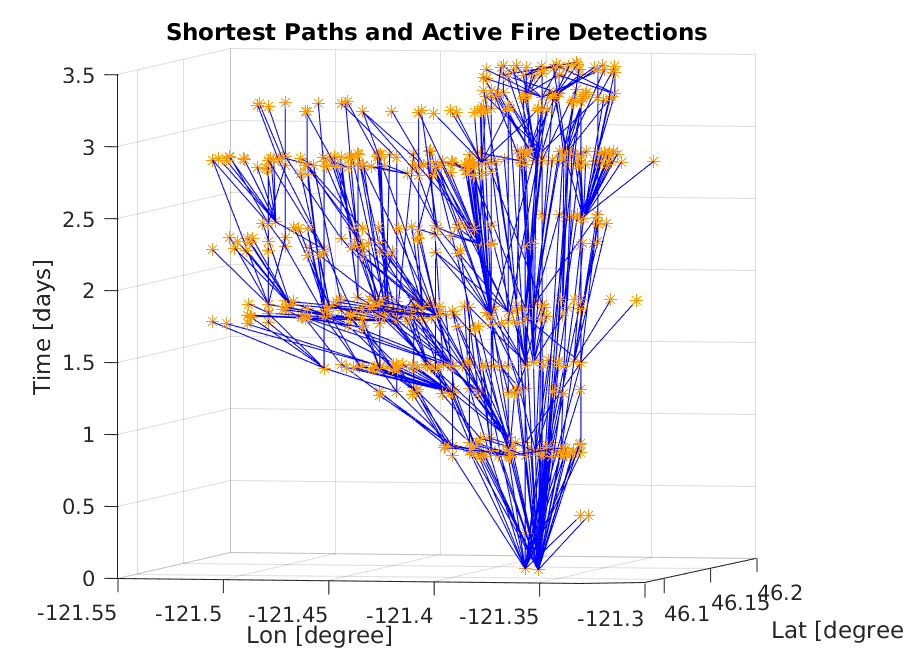}
     \caption[A directed graph of fire detections and the derived shortest path structure.]{A directed graph of fire detections and the derived shortest path structure. On the left is a directed graph showing the connections among 393 fire detections. Very little structure can be discerned from this image. On the right is the path structure showing the shortest paths from the ignition point to all other detections in the graph. The method used for imposing the structure seen on the right begins with spatially clustering the detection locations.}
\label{fig:abstract_paths}
\end{figure}

A second matrix $T$ is also constructed to store the time difference between vertices in the path. Thus, $T_{ij}$ records the time at which detection $j$ was observed to be burning minus the time at which pixel $i$ was observed to be burning. If $i < j$ then $T_{ij} \geq 0$. Importantly, in constructing the directed graph, the edge connecting two detections, $i$ and $j$ will have the direction from the earlier detection to the later detection. Figure \ref{fig:graph_up_all} shows how this restriction simplifies the graph and gives a relationship between the detection locations that corresponds to the idea that fire will spread from an early detection location only to other detection locations with a later time. The rate of spread $R_{ij}$ along a straight line between vertices $i$ and $j$ may be computed by $R_{ij} = D_{ij} / T_{ij}$. In many cases, the time of multiple fire detections in the domain is the same since the observations were recorded in a single satellite data granule. Thus, if detections $i$ and $j$ have the same recorded time, $T_{ij} =0$ and $R_{ij}$ is cannot be computed. For this reason, in making the edges between detection locations, the only allowed edges are those from a detection to a detection having a later recorded time. This concept is illustrated in Figure \ref{fig:edge_directions}. The left panel shows all possible connections between three detections and the right panel shows only those allowed by the logic explained above. Figure \ref{fig:abstract_paths} shows how real satellite data can be organized into a directed graph. The left panel shows the abstract directed graph and the right panel shows how it is simplified and  transformed into useful structure by finding the shortest path for the ignition point to all other active fire detections. How this is accomplished is explained in subsequent sections of this chapter.

\begin{figure}[!h]
\begin{center}
  \includegraphics[width = 0.49\textwidth]{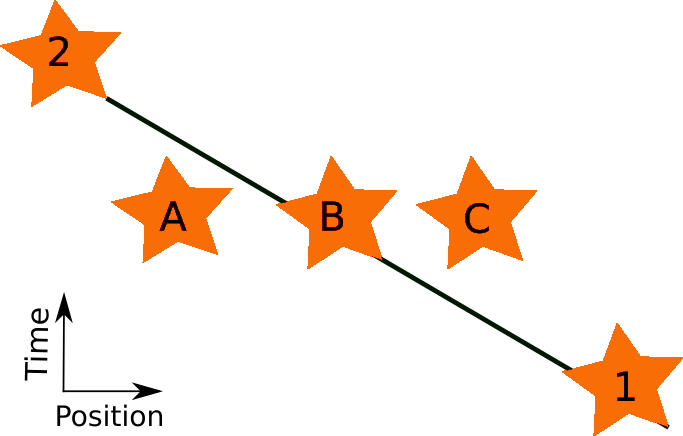}
  \caption[Multiple active fire detections in a hypothetical one-dimensional data granule demonstrates some of the uncertainties inherent in the observations.]{Multiple active fire detections in a hypothetical one-dimensional granule. The orange stars represent active fire detections and the solid line indicates the progression of the actual fire. Detections $A$,$B$,and $C$ were recorded at the same time. Detection $B$ is exactly located at the correct time and place. Detection $C$, above the line, has been burning for some time but was possibly not observed before. Detection $A$ is below the line, in a place where it should not yet be burning. Possibly a geolocation error has occurred and detection $A$ should be closer to detections $B$ or $C$.}
  \label{fig:same_granule}
  \end{center}
\end{figure}

\subsubsection{Cluster the data}

When the detection data has been organized into a directed graph with each detection connecting to all other detections with a later time, the ignition point is connected to all other detection locations in the graph. At this stage, the shortest path from the ignition point to any other detection location is a straight line. In the case of a fire that spread out at constant rate from the ignition point to a series of expanding, circular perimeters, these kinds of straight line would be consistent with what really occurred. We should expect real fires to behave differently. Fire should progress in a way that passes from the location where it was first was observed to a location where it was later observed, through detection locations that are intermediate both spatially and temporally.

\begin{figure}[!ht]
\centering
       \includegraphics[width=0.45\textwidth]{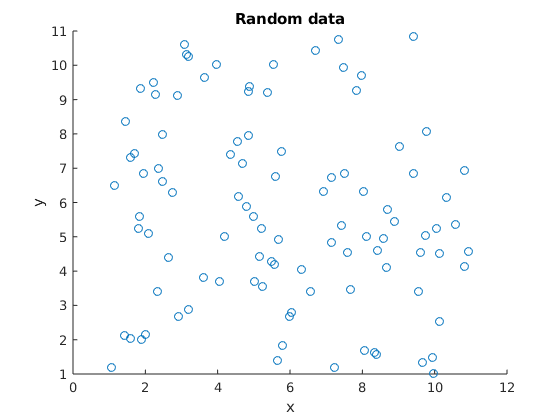}
     \hfill
       \includegraphics[width=0.45\textwidth]{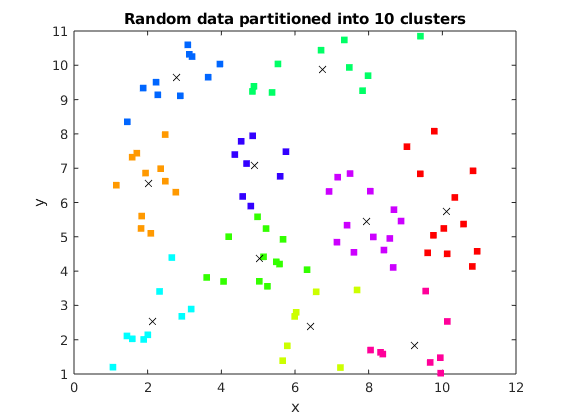}
     \caption[Spatial clustering of random data by the k-means method.]{Spatial clustering of random data by the k-means method. On the left, 100 points of random data have been plotted. On the right, the random data has been partitioned into 10 clusters. The black $\times$ shapes denote the centroids of the clusters.}
\label{fig:k_means_example}
\end{figure}

To accomplish the creation of paths that connect two detections via intermediate detections, the detection data is first spatially clustered. K-means clustering is an unsupervised machine learning algorithm that that classifies $n$ data points into $k$ clusters around a centroid \citep{Loyd-1982-LSQ}. Figure \ref{fig:k_means_example} shows an example where 100 points chosen randomly were categorized into 10 spatial groups by the k-means algorithm. For work with fire data, the value $k=20$ has been used. Section \ref{sec:cluster_test} shows the testing procedure used to arrive at that number. When the detection data has been clustered spatially, the distances between detections belonging to the same cluster are then shortened. This spatial clustering has the effect of causing the shortest paths from the ignition point to pass through intermediate detections that belong to the same clusters. This procedure is further explained in Section \ref{sec:shorten}.

\begin{figure}[!ht]
\centering
       \includegraphics[width=0.45\textwidth]{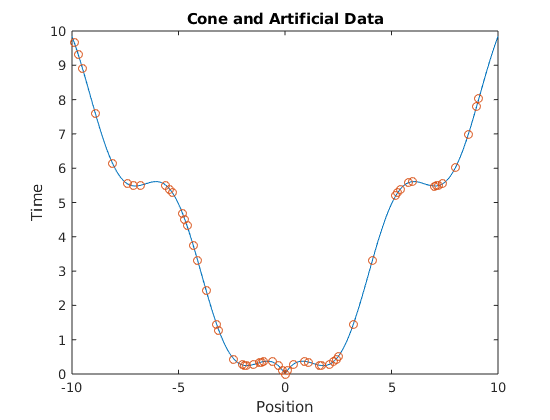}
     \hfill
       \includegraphics[width=0.45\textwidth]{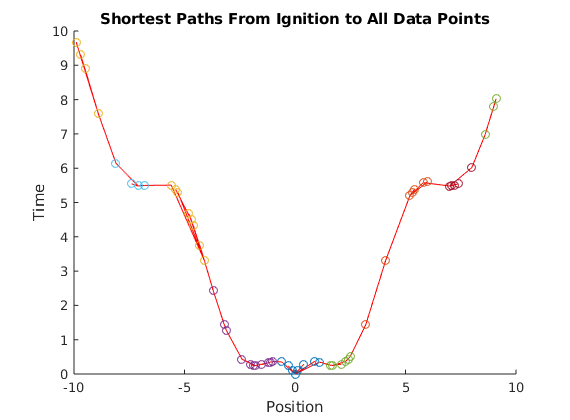}
     \caption[Illustrating the effect of spatial clustering on the set of shortest paths in a one-dimensional fire line] {Illustrating the effect of spatial clustering on the set of shortest paths in a one-dimensional fire line. On the left, the blue curve is taken to represent the progress of the fire and the colored circles on this line represent the time and location of ``fire detections." On the right, the detections have been clustered by spatial location and shortest paths have been drawn from the ignition point to all detections occurring later. Without clustering, all shortest paths would be straight lines to the individual detections, fanning outward from the ignition location. The clustering is essential for forcing the paths to maintain a gradient consistent with the data.}
\label{fig:2d_cluster}
\end{figure}

To show how clustering can be used to cause shortest paths in the data that follow the gradient, we first use an example of a two-dimensional fire line that can be thought of the cross section of a cone shape common to fire arrival times that follow diurnal cycles where the ROS increases in the day and slows at night. In Figure \ref{fig:2d_cluster} we see the fire cone and artificial data in the left panel and in the right panel we see shortest paths drawn from the ignition to all other points data points. The function used was
\begin{equation}
T(x) = |x| + 1.2 \cos x -1.
\end{equation}
The clustering of the data causes the shortest paths from the ignition point to other detection locations to pass through intermediate points. Without clustering, all paths would be straight lines, fanning outward from the ignition point.

\subsubsection{Using Infrared Perimeters as Data Source}
Infrared perimeter observation can be used as another source of information about the location of the fire front. This source of data is not always available and frequently is limited to one observation per day, but the spatial resolution is on the order of several meters. In this thesis, infrared observations are occasionally used as a data source for estimating the fire arrival time. When used, the perimeter observations were processed in a way that allowed them to be used in the same manner as satellite active fire detections. 

Infrared perimeter data come in various data forms and can be processed to give a list of GPS latitude and longitude locations of the fire perimeter. The time of the observation is also available. For use in making an estimate of the fire arrival time or use in data assimilation, the locations and times of the points on the perimeter of the fire can be used in the same way that satellite active fire detections are used. In many cases, the contours of a perimeter are distinct and easily interpreted in a way that makes it possible to form a polygon that encompasses the fire area neatly. In other cases, many points in the perimeter are close together and it is not clear where the actual fire perimeter location is. Thus, despite having good spatial resolution, some calculations involving perimeter data can be considered suspect because the perimeters are not clearly defined. Figure \ref{fig:cougar_moe_perim_5} shows an example where an algorithm for finding the interior of a perimeter has failed because of a gap in the points defining the perimeter.

\subsection{Making the Shortest Paths}
%
%
\subsubsection{Reducing the distances between detections within clusters}
\label{sec:shorten}
\begin{figure}[htbp]
\centering
\includegraphics[width=0.45\textwidth]{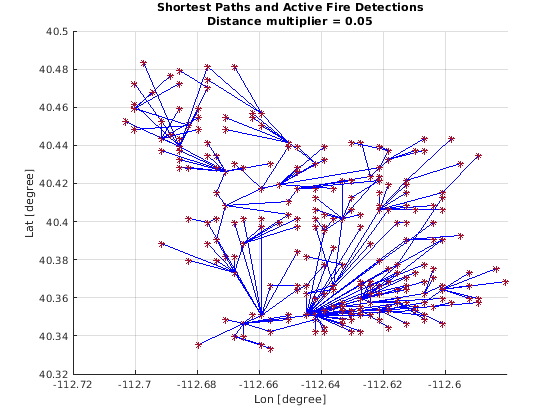}
\includegraphics[width=0.45\textwidth]{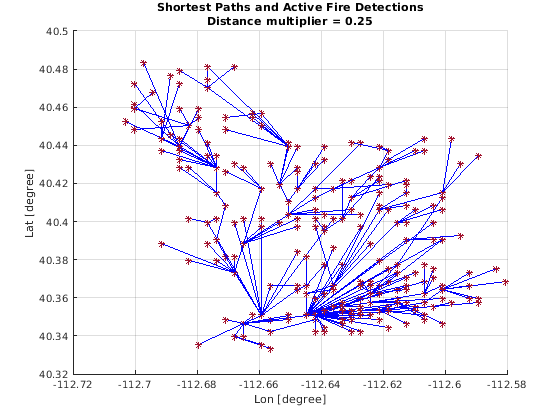}
          
\includegraphics[width=0.45\textwidth]{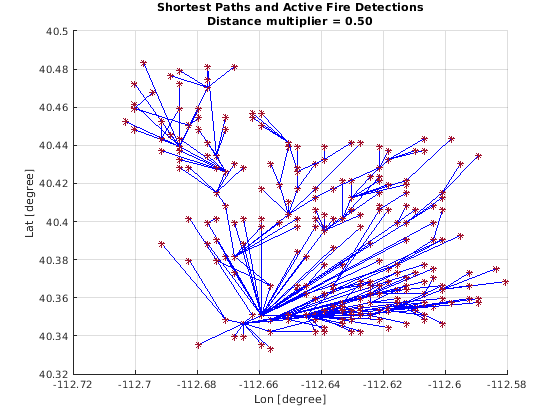}
\includegraphics[width=0.45\textwidth]{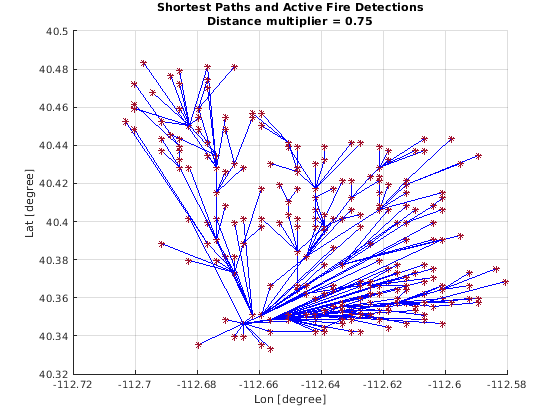}
 \caption[Various path structures obtained by changing the distance multiplier for detections within the same cluster]{Various path structures obtained by changing the distance multiplier for detections within the same cluster. The panel in the upper left and upper right show the path structures made when the distance multiplier was 0.05 and 0.25, respectively. The bottom left and bottom right used multipliers of 0.50 and 0.75, respectively. The two path structures in the low panels show fire moving along straight lines on more paths than the upper two panels. }
\label{fig:dm_paths}
\end{figure}

To make the shortest paths other than straight lines, the distances between detections within the same cluster are shortened to 1/4 of their nominal lengths.  This has the effect of breaking the triangle inequality for distances and causes a path from the ignition to the outer edge of a fire area to pass through intermediate detection locations. Shortening by any fraction of the nominal distance can produce the desired effect. The value of 1/4 was chosen by examining the effect of using values of of a distance multiplier $m$ with values $m \in \{0.05,0.25,0.50,0.75\}$. Figure \ref{fig:dm_paths} shows the path structures obtained using these values of $m$. The structures in the top  panels show less paths moving from the ignition point to other detections in the domain along straight paths and are seemingly to be preferred. Estimates of the fire arrival time were also made from these path structures and compared with each other. The largest relative difference in norms between any two of the resulting fire arrival times occurred between the pair made from using multipliers $m=0.05$ and $m=0.75$. This relative difference was 0.0045. Such a small difference indicates that the value of $m$ plays a minor role in how the method works. The important points is to break the triangle inequality and force the shortest paths to take courses other than straight lines.

\begin{figure}[!h]
\centering
    \includegraphics[width = 0.39\textwidth]{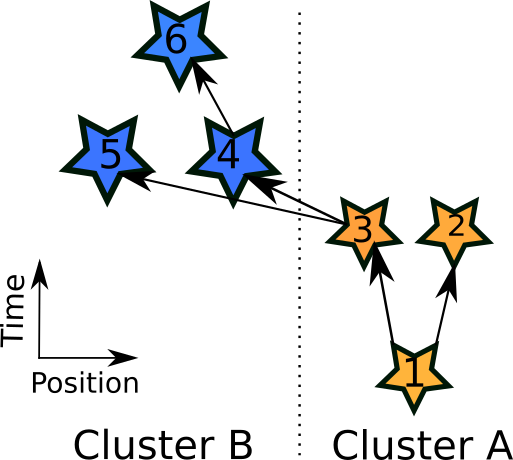}
    \caption[The effect of reducing the distance between active fire detections within the same cluster.]{The effect of reducing the distance between active fire detections within the same cluster. Shortest paths are drawn from the ignition point at detection 1 to all other detections.}
\label{fig:cluster_distances}
\end{figure}

Figure \ref{fig:cluster_distances} shows the effect of clustering spatially on the path creation using a hypothetical example. A clustering algorithm is first used to spatially divide the detections into 2 clusters and then the distances between detections belonging to the same cluster are reduced. The results are shortest paths that connect the ignition point to all other points in the graph in a way consistent with the idea that fire will move from an earlier detection location to other nearby locations that were recorded at a later time. Without clustering, the paths from detection 1 to detections 4,5, and 6 would be straight lines not passing through detection 3. Note that detection pairs 2,3 and 4,5 have been recorded at the same time and are therefore not connected in the graph.

A mathematical example with several random points in $\mathbb{R}^2$ illustrates how the triangle inequality is broken and shortest paths are made to differ from straight lines. We start with 5 random points in matrix $P$ whose rows are the ordered pairs of points
\begin{equation}
P=\begin{bmatrix}
0.3922  &  0.2769 \\
0.6555  &  0.0462 \\
0.1712  &  0.0971 \\
0.7060  &  0.8235 \\
0.0318  &  0.6948
\end{bmatrix}.
\end{equation}

Matrix $D_1$ then gives gives the Euclidean distance between each pair of points in $P$. The distance between points $i$ and $j$ is $D_1(i,j)$, the entry in row $i$, column $j$ of the matrix  
\begin{equation}
D_1 = \begin{bmatrix}
     0  &  0.3501  &  0.2849  &  0.6302  &  0.5518 \\
0.3501  &       0  & 0.4870   &  0.7789  &  0.8998 \\
0.2849  &  0.4870  &       0  &  0.9020  &  0.6137 \\
0.6302  &  0.7789  &  0.9020  &       0  &  0.6864 \\
0.5518  &  0.8998  &  0.6137  &  0.6864  &       0
\end{bmatrix}
\end{equation}

\begin{figure}
\centering
\includegraphics[width=0.6\textwidth]{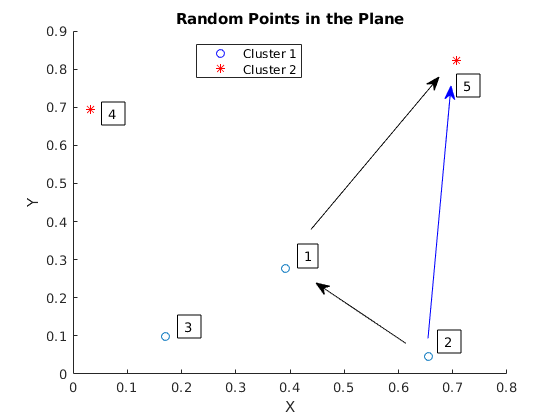}
\caption[How reducing the distances between points belonging to the same cluster causes shortest paths between two points to pass through intermediate points.]{How reducing the distances between points belonging to the same cluster causes shortest paths between two points to pass through intermediate points. The path between points 2 and 5 is shorter when making a detour through point 1 when distances between points belonging to the same cluster are reduced. This process of breaking the triangle inequality allows construction of plausible paths from an ignition point to all other fire detections in the domain that pass through nearby, intermediate fire detections in a way that is different from a straight line.}
\label{fig:triangle_break}
\end{figure}

The k-means clustering algorithm is then used to separate the points of $P$ into two clusters, with points ${1,2,3}$ belonging to one cluster and points ${4,5}$ belonging to the other cluster as shown in Figure \ref{fig:triangle_break}. The distances between points in $D_1$ are then reduced to 1/4 of their value if the points belong to the same cluster, giving distance matrix $D_2,$ where the entries belonging to cluster 1 are in the top left $3 \times 3$ diagonal submatrix of $D_1$ and those in cluster 2 are in the lower right $2 \times 2$ diagonal submatrix of $D_1$. The adjustment of distances gives us

\begin{equation}
D_2 = \begin{bmatrix}
     0 &   0.0875  &  0.0712  &  0.6302  &  0.5518 \\
0.0875 &        0  &  0.1217  &  0.7789  &  0.8998 \\
0.0712 &   0.1217  &      0  &  0.9020  &  0.6137 \\
0.6302 &   0.7789  &  0.9020  &       0  &  0.1716 \\
0.5518 &   0.8998  &  0.6137  &  0.1716  &       0
\end{bmatrix}.
\end{equation}


As an example, under the distances in $D_1$, the shortest path between points 2 and 5 is the straight line connecting them. When clustering is used to shorten inter-cluster distances, the path going from point 2 to point 1 to point 5 becomes shorter. Figure \ref{fig:triangle_break} shows the picture and a comparison of the distances is illustrated by the inequality
\begin{equation}
D_1(2,5) = 0.8988 \hspace{2mm} > \hspace{2mm} D_2(2,1)+D_2(1,5) = 0.0875 + 0.5518 = 0.6393.
\end{equation}

\subsubsection{Create shortest paths}

With a directed graph created and clustering of the data accomplished, the next step is to create a set of shortest paths connecting every detection point in the graph to the assumed ignition point. The Matlab function ``shortestpath" \citep{Matlab-2021-SPB}) uses the Dijkstra algorithm to accomplish this task, but first an ignition point must be chosen. Typically, information about the ignition point of the fire will be unknown and therefore it must be inferred from the data. The method for estimating the ignition point outlined in Section \ref{sec:ignition_estimation} is impractical for the present purpose since it require running the WRF-SFIRE model. Therefore, the proposed method is to use the first occurring fire detection in the fire domain as the assumed ignition point if its time is earlier than all other detections. If multiple detections were recorded at the same time, the ignition point may be estimated by taking the mean location of these first detections and then creating a new data point with a time six hours earlier. Figure \ref{fig:abstract_paths} shows the abstract directed path structure on the left and the resulting shortest path structure after clustering of the data has been achieved.

\subsection{Interpolation of the Fire Detection Data}
\begin{figure}[htbp]
\centering
          \includegraphics[width=0.45\textwidth]{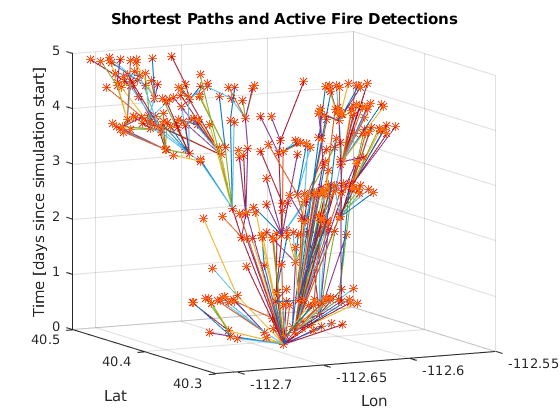}
          \includegraphics[width=0.45\textwidth]{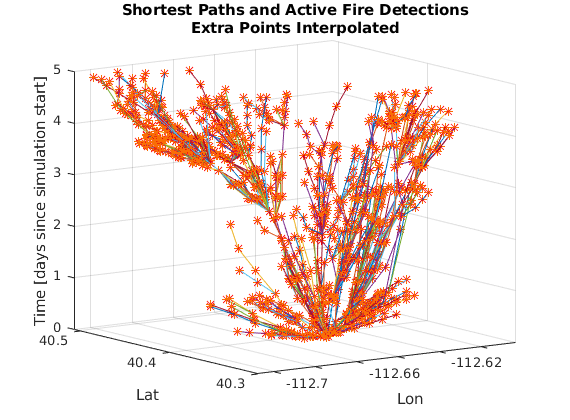}
 \caption[Interpolation of additional detection points along the shortest paths.]{Interpolation of additional detection points along the shortest paths. (Left) The shortest paths from an assumed ignition to all other points in the detection data. (Right)  New points have been interpolated between each node in all paths. }
 \label{fig:interp_new_points}
\end{figure}

With the satellite data organized into a set of shortest paths, already much of the history of the fire may be deduced by examining the structure of detections and paths. Figure \ref{fig:interp_new_points} shows such an object. This patch structure can be likened to the wire mesh constructions employed in computer aided design. Much of the shape and form of the fire is present, but more work needs to be done to provide essential details. That work begins with adding additional data points along the established fire paths. Then, an interpolation scheme is used to infer the fire arrival time for locations in the domain near to the points on the fire paths.
The method for creating an estimated fire arrival time from the detection data that has been clustered and organized into shortest paths follows the following steps that will be explained. 

\begin{itemize}
\item Interpolate extra points along the paths
\item Iterative interpolation of data
\item Smoothing and cleaning
\item Multigrid method for interpolation. Working on a sequence of meshes with increasing resolution.
\end{itemize}

\subsubsection{Interpolate points along the paths}
\label{sec:interp_points}
\begin{figure}[!h]
\begin{center}
  \includegraphics[width = 0.49\textwidth]{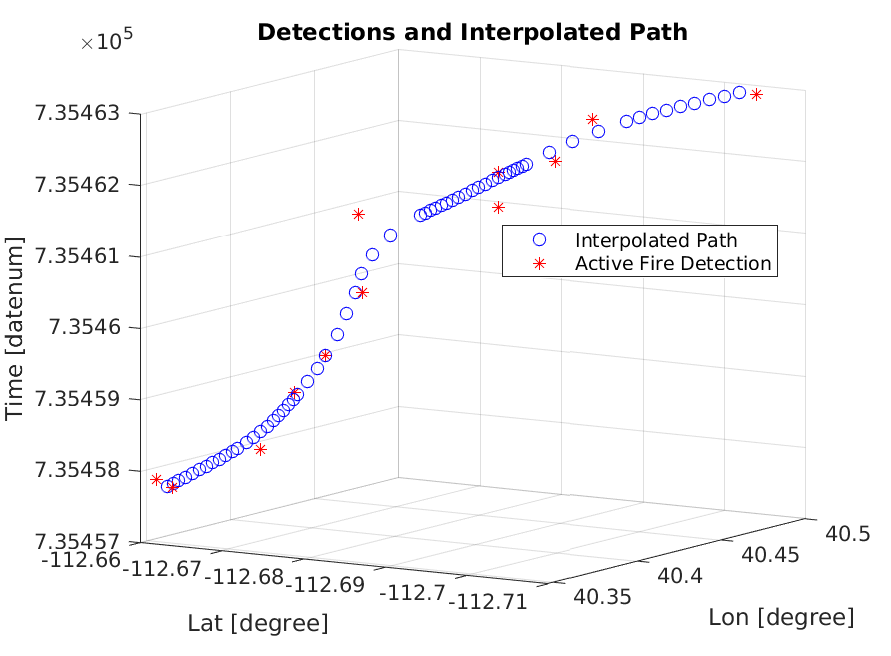}
  \caption[Interpolation of extra data points along a single shortest path.]{Interpolation of extra data points along a single shortest path. A smoothing spline was used to add more points along a path that connects the active fire detections seen as red stars. To avoid over-fitting of uncertain data and large oscillations in the path, the smoothing spline of Equation \ref{eq:spline} was used. The original path contained 12 data points and the resulting spline contains 63 data points. In practice, fewer additional points of data are added than are shown in this example.}
  \label{fig:spline_detect}
  \end{center}
\end{figure}

To provide additional data that can be used to estimate fire arrival times for locations near the paths, more data is interpolated at locations on or near the paths. Figure \ref{fig:spline_detect} shows how a set of sparse data along a path may be augmented with interpolated data. In this work, the  Matlab "Cubic Smoothing Spline" was used to interpolate extra points along the paths in the graph \citep{Matlab-2020-CSS}). The function seeks an interpolating function $f$ that minimizes  
\begin{equation}
\min_f \hspace{4mm} p\sum_{j=1}^n w_j |T_j -f(x_j)|^2 + (1-p)
\int\lambda(t) |D^2f(t)|^2dt.
\label{eq:spline}
\end{equation}
Here, $T_j$ is the time associated with the fire detection recorded at position $x_j.$ By changing the values of the parameter $p$, the spline can be made smoother at the expense of not passing exactly through the set of points $(x_j, T_j)$. Additionally, weights $w_j$ can be assigned at those points. The optional function $\lambda(t)$ can be assigned to weight the second derivative of the spline, allowing for a smoother or rougher fit, depending on location in the graph. In the work presented here, the value $\lambda(t) = 1$ was used consistently. The key parameter is $p$. When $p$ is small, more weight is given to the integral of the second derivative in Equation \ref{eq:spline}, causing the minimization to favor a smoother path at the expense of fitting the detection data exactly. The resulting, interpolated path may diverge widely from the the original. Conversely, small values of $p$ favor fitting the data at the expense of smoothness of the path. The plots in the top row of Figure \ref{fig:p_test_patch_paths_moe} show the effect of varying $p$ for a given set of detections.

When interpolating more points on the paths, the two main considerations are how many points should be added and what value of $p$ in the interpolation function should be used. Several tests were designed to find the best values to use. The tests show that using $p=0.9$ with interpolation of extra data points at a distance of 2 kilometers produced the best estimates of the fire arrival time. The testing methods will be described below in Section \ref{sec:find_p} and Section \ref{sec:ground_truth_1}.

\begin{figure}[!ht]
\centering
\includegraphics[width=0.45\textwidth]{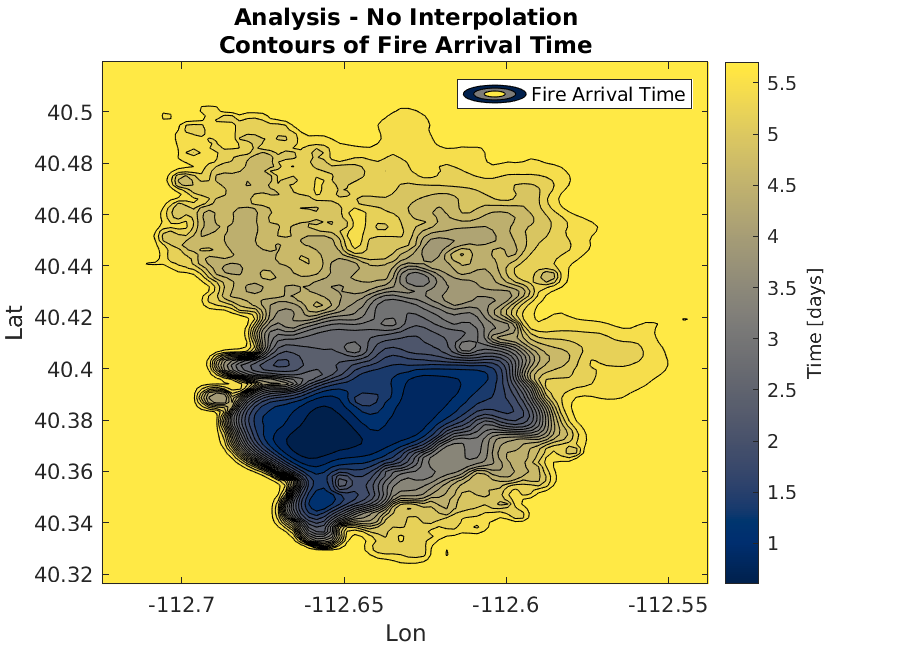}
     \hfill
\includegraphics[width=0.45\textwidth]{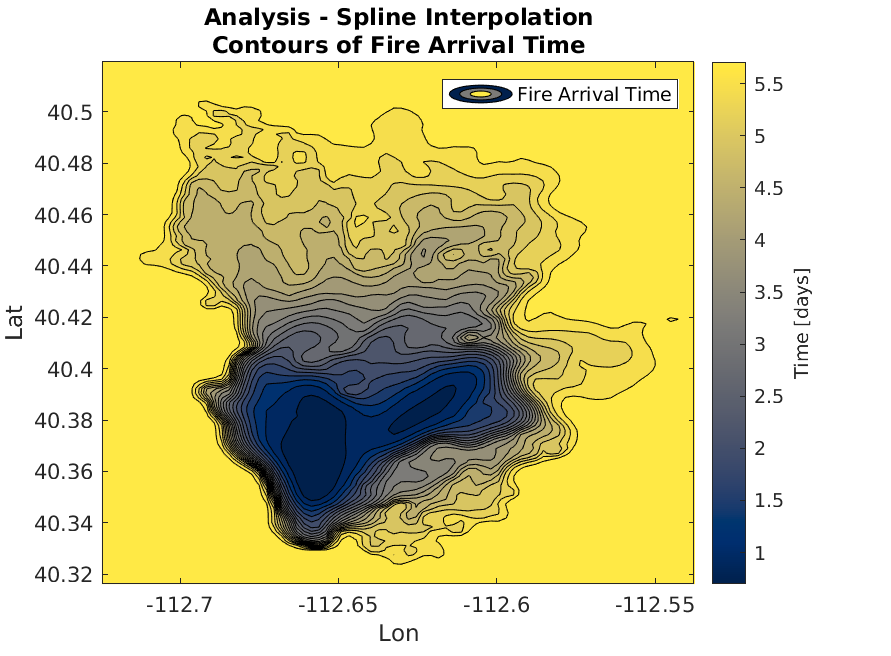}
     \caption[Comparison of the fire arrival time contours for an analysis made with or without interpolating additional points along the shortest paths.]{Comparison of the fire arrival time contours for an analysis made with or without interpolating additional points along the shortest paths. On the left, no additional points have been added to the paths. The contours show many peaks and valleys, indicating a fire that possibly spread by spotting. On the right, fewer peaks and valleys can be seen.  }
\label{fig:analysis_interpolation_compare}
\end{figure}

\begin{figure}[!ht]
\centering
\includegraphics[width=0.45\textwidth]{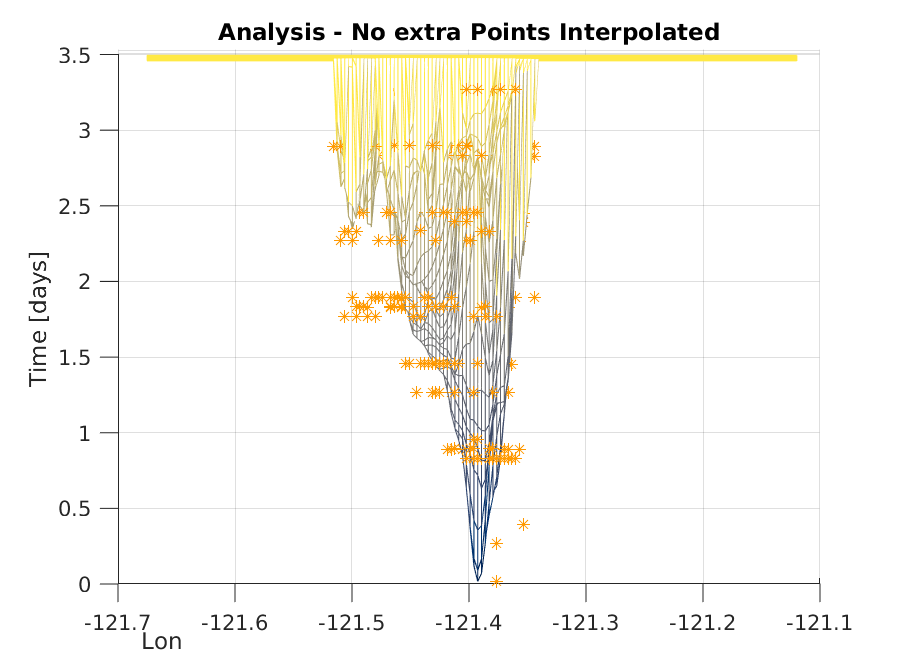}
     \hfill
\includegraphics[width=0.45\textwidth]{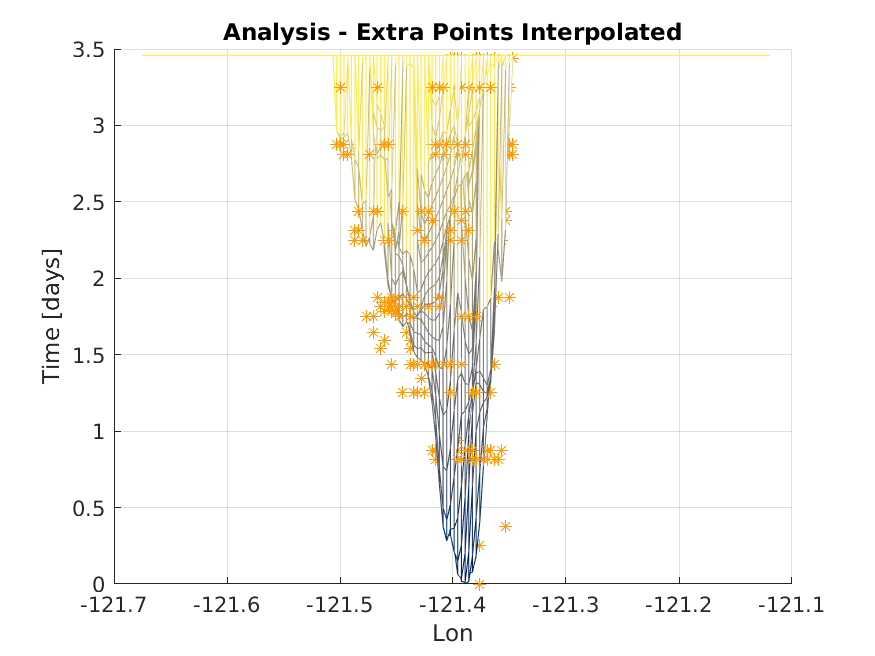}
     \caption[Comparison of the fire arrival time  for an analysis made with or without interpolating addition points along the shortest paths.]{Comparison of the fire arrival time  for an analysis made with or without interpolating addition points along the shortest paths. On the left, no additional points have been added to the paths. On the right, extra points have been interpolated along the shortest paths. The extra points allow the the analysis to capture more of the gradient information seen along the left side of the fire arrival time cone that is implied by the detection data. }
\label{fig:estimate_interpolation_compare}
\end{figure}


\subsubsection{Iterative interpolation of data}
\label{sec:iter_interp}

Let $G$ be the set of all shortest paths in the detection graph. Starting from a fire arrival time $T_0$, that is an initial estimate, along each  path $p_j \in G$ we make the fire arrival time $T$ equal to a weighted average of the $T_0$ and the time of the detection $T(d_i)$ for vertex $d_i$ in the graph as
$$T_1 = \alpha T_0 + (1-\alpha) T(d_i),$$
where $\alpha \in [0,1]$ is chosen, depending on application. For the process of estimating fire arrival time, $\alpha = 0$ may be chosen. For data assimilation, $T_0$ will be from a model forecast and the choice of alpha will be made from the data likelihood for each detection. An explanation of the method applied to data assimilation will be presented in Chapter \ref{chapter:04}.

The fire arrival time is constructed by first changing the fire arrival time at the locations of the active fire detections. If detection $d_i$ occurs on $n$ paths in the graph, the fire arrival time $T(d_i)$ at the detection location $d_i$ will be changed $n$ times. For example, the ignition point belongs to every path in the graph so the fire arrival time there gets changed once for every active fire detection in the graph. Active fire detections in final granule of the set appear only in one path since edges in the directed graph will only point towards them.  For detections that are part of $n$ paths, the fire arrival time at location $d_i$ will be computed as
%
%
      
\begin{equation}
\begin{split}
T_1 &= \alpha T_0 + (1-\alpha) T(d_i) \\
    &= \alpha (T_0 - T(d_i)) + T(d_i) \\
T_2 &= \alpha T_1 + (1-\alpha) T(d_i) \\
    &= \alpha \left(\alpha (T_0 - T(d_i)) + T(d_i)       \right) + (1-\alpha) T(d_i)  \\  
    &= \alpha^2(T_0 - T(d_i)) + T(d_i) \\
T_3 &= \alpha T_2 + (1-\alpha) T(d_i) \\
    &= \alpha \left(\alpha^2 (T_0 - T(d_i)) + T(d_i)       \right) + (1-\alpha) T(d_i)  \\  
    &= \alpha^3(T_0 - T(d_i)) + T(d_i),
\end{split}
\label{eqn:fat_move}
\end{equation}
eventually becoming
$$T_n = \alpha^n(T_0 - T(d_i)) + T(d_i), $$
when all the $n$ paths have been processed.

\begin{figure}[!h]
  \begin{center}
    \includegraphics[width = 0.45\textwidth]{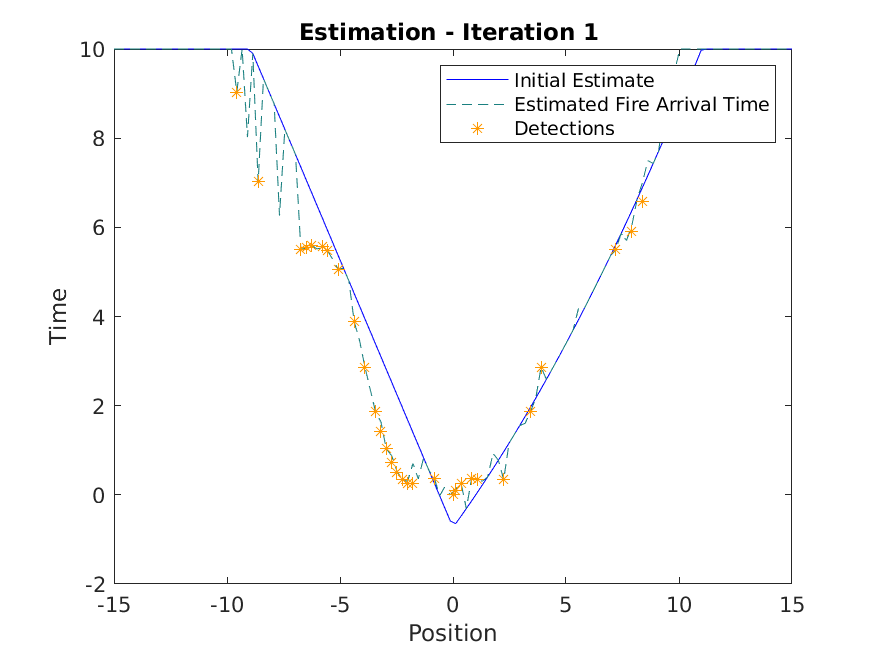}
         \hfill
       \includegraphics[width=0.45\textwidth]{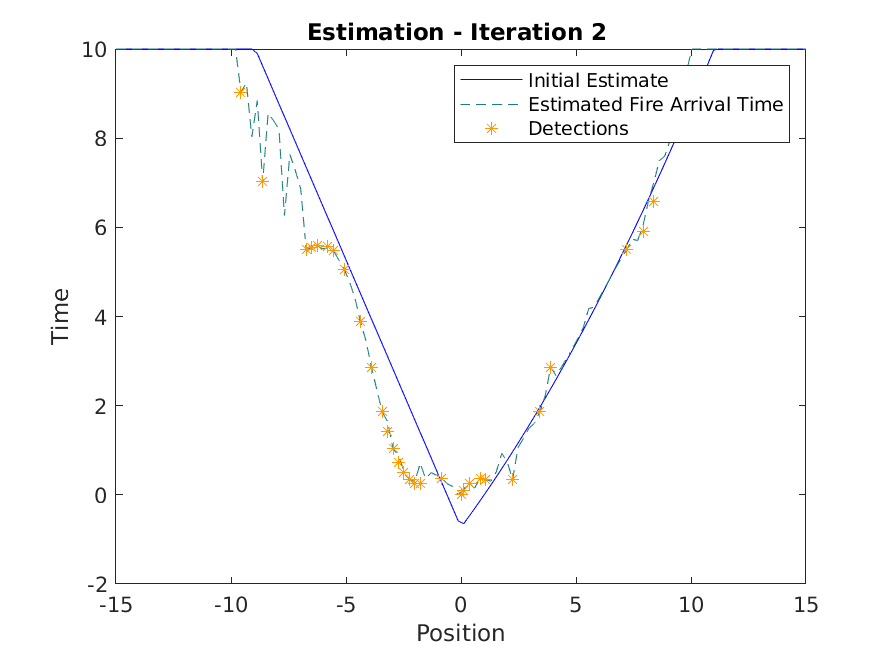}

    \includegraphics[width = 0.45\textwidth]{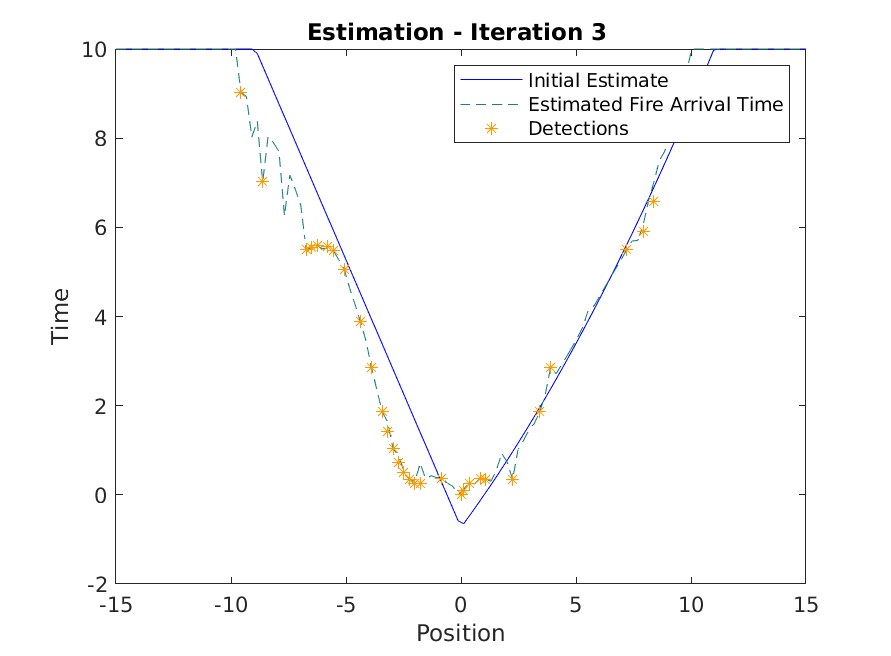}
         \hfill
       \includegraphics[width=0.45\textwidth]{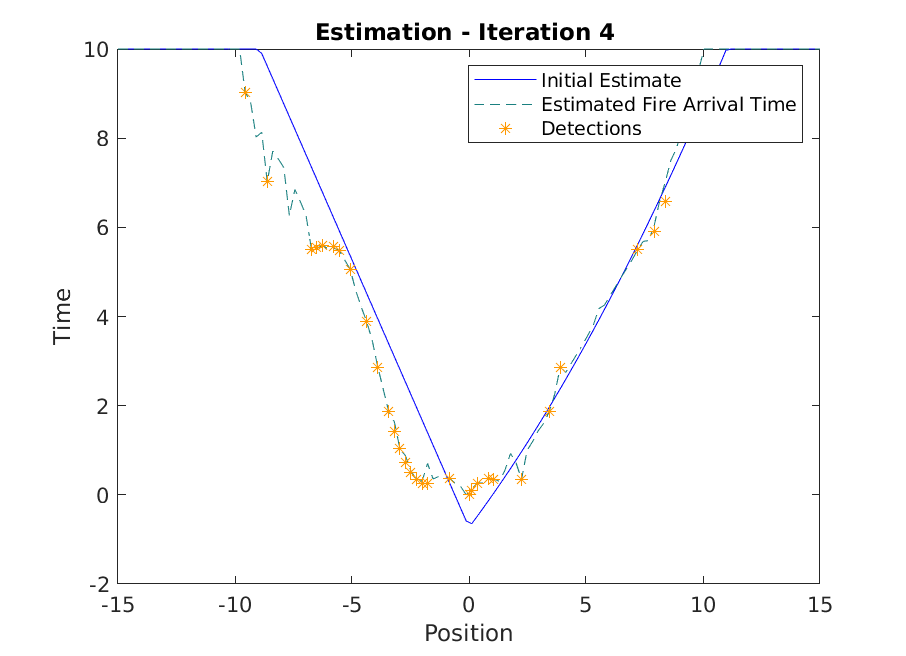}

    \includegraphics[width = 0.45\textwidth]{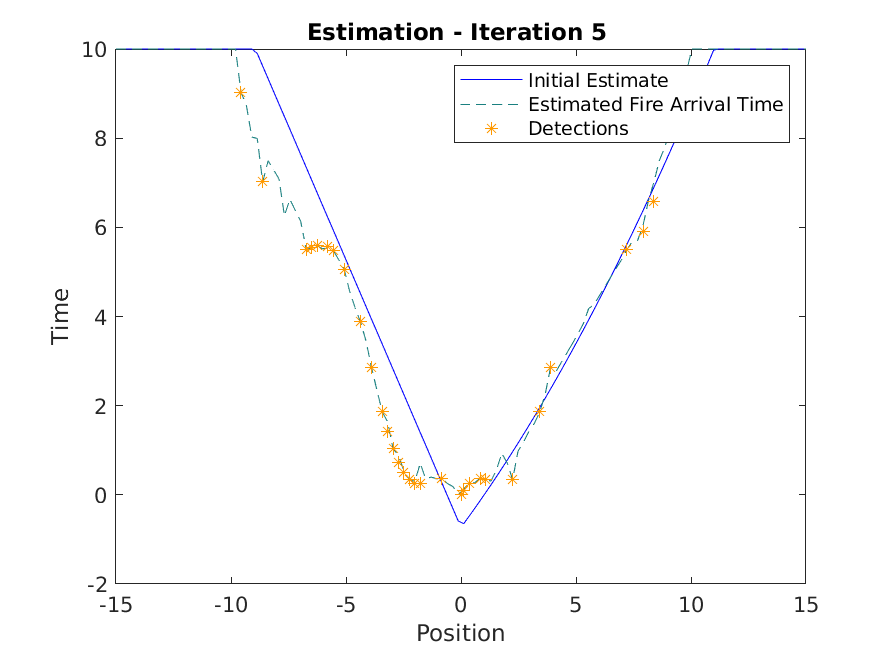}
         \hfill
       \includegraphics[width=0.45\textwidth]{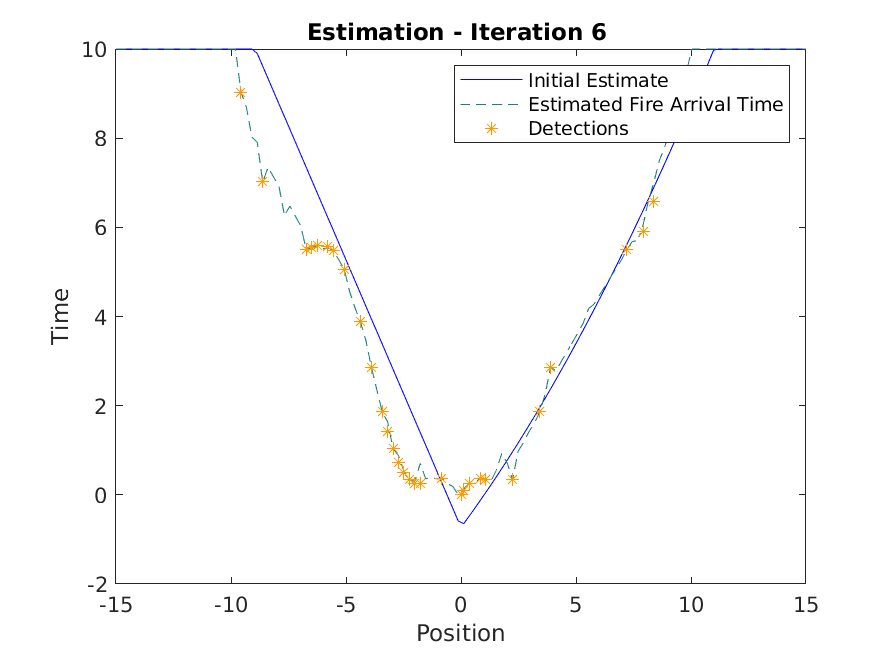}
       
    \caption{Iterative interpolation. The sequence of panels shows the data assimilation process of iterative interpolation. At each an estimate of the fire arrival time is made by adjusting the fire arrival time in places where detection data exists. A Gaussian smoothing is then applied to remove discontinuities in the data and to avoid over-fitting of uncertain data.}
    \label{fig:iter_interp}
    \end{center} 
\end{figure}

\begin{figure}[!h]
  \begin{center}
\includegraphics[width = 0.45\textwidth]{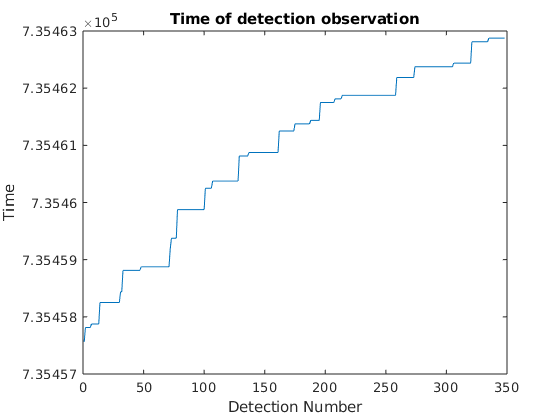}
\includegraphics[width = 0.45\textwidth]{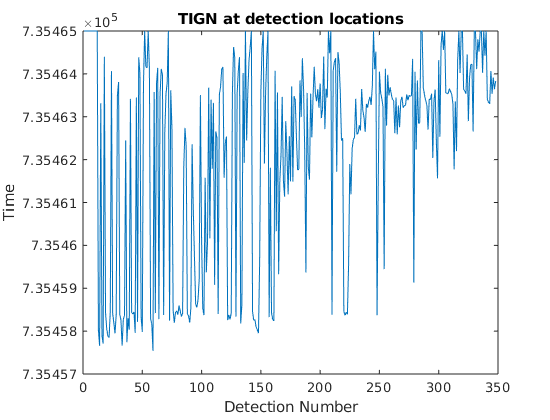}

\includegraphics[width = 0.45\textwidth]{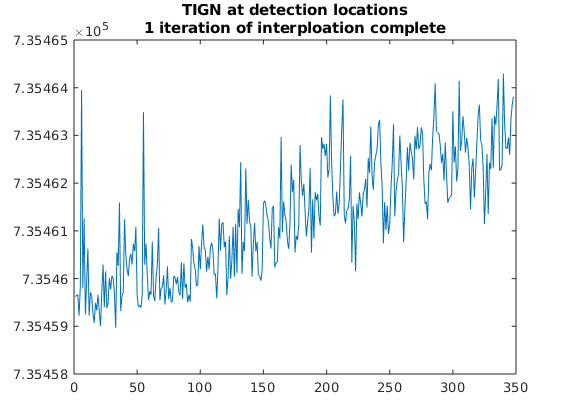}
\includegraphics[width = 0.45\textwidth]{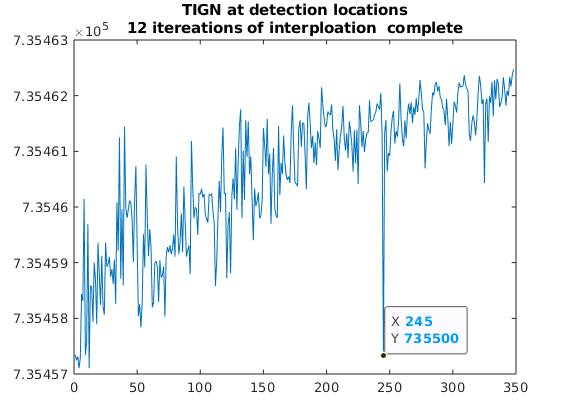}   
    \caption[Illustrating the change in the estimated fire arrival time at locations of active fire detections as the number of iterations increases. ]{In the upper left, the plot shows the active fire detections ordered in time. In the the upper right, is the forecast fire arrival  time at the locations of the active fire detections. The bottom row shows the adjusted fire arrival time after 1 and 12 iterations of the method, respectively. The plots of the adjusted fire arrival times show that difference between the forecast and the time of the detections decreases with more iterations. The large ``spike" that develops and is marked with a data label in the figure in the lower right corresponds to fire detection that was recorded several days after the fire began in a location near the estimated ignition point. }
    \label{fig:detection_times_sequence}
    \end{center} 
\end{figure}

A series of panels in Figure \ref{fig:iter_interp} illustrate the method using a hypothetical case of a one-dimensional fire line. The fire arrival time at each detection location is first changed from that of an initial estimate, introducing roughness and  discontinuities into the fire arrival time. A local averaging then smooths the fire arrival time over the whole domain. The total effect is that the fire arrival time at the detection locations and at nearby locations becomes closer to the time that the detection was recorded. The algorithm then repeats, with an adjustment of the fire arrival time at the detection locations followed by another local averaging. Starting from an initial estimate $T_0$, a sequence of fire arrival times ${T_1,T_2,T_3,...}$ is produced at each location in the fire domain. After iteration $i$ produces $T_i$, the relative difference between successive fire arrival times $T_{i-1}$ and $T_i$ is computed as
\begin{equation}
RD = \frac{\Vert T_{i-1}-T_i \Vert}{\Vert T_{i-1}\Vert}.
\end{equation} 
The algorithm can be halted when $RD$ falls below a chosen threshold or it can be halted after a fixed number of iterations. The left panel in Figure \ref{fig:squish_converge_figs} shows the decrease in $RD$ during the iterative interpolation process over 20 iterations. The plot shows large initial decreases in $RD$ over the first several iterations of the method, followed by smaller and smaller decreases. An alternative visualization of the process is  depicted in Figure \ref{fig:detection_times_sequence}. The upper left panel shows the active fire detections in the domain ordered in time. The upper right panel shows the times of an initial estimate of the fire arrival time, here taken to be a model forecast and titled ``TIGN", at the location of the active fire detections. If the forecast was in complete agreement with the detection data, these first two panel in the figure would be identical. Instead, we see a large number of places in the domain where the forecast fire arrival time is above or below that of the detection data. The remaining panels on the bottom show how the iterative interpolation process affects this difference, with the end effect being that the fire arrival time at detection locations moves closer to the time of satellite detection.

\subsubsection{Local averaging of the fire arrival time}
\begin{figure}[!h]
  \begin{center}
\includegraphics[width = 0.45\textwidth]{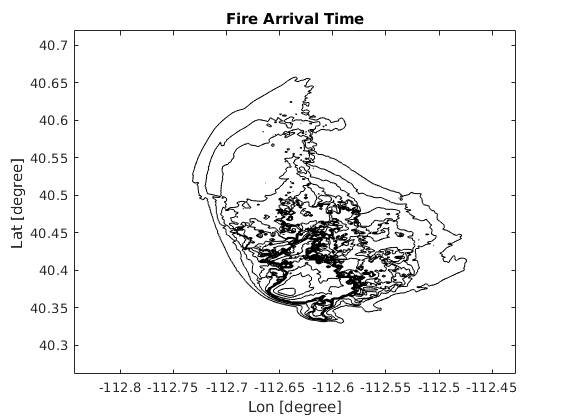}
\includegraphics[width = 0.45\textwidth]{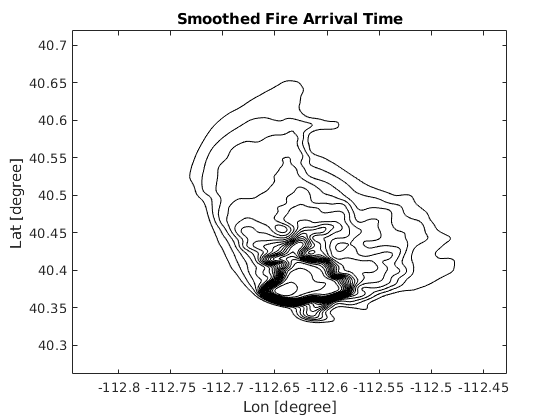}
    \caption[Local averaging of the fire arrival time with a Gaussian smoothing kernel.]{Local averaging of the fire arrival time with a Gaussian smoothing kernel. On the left are the contours of a fire arrival time of a simulation made with WRF-SFIRE. On the right, the fire arrival time has been smoothed.}
    \label{fig:imgauss}
    \end{center} 
\end{figure}

 After the fire arrival time has been set at all detection points along the paths in the graph, the fire arrival time will have developed roughness manifested as large jumps in the fire arrival time at detection locations when compared to neighboring locations. To smooth the data and change the fire arrival time at locations in the fire domain near the detection locations, a local averaging of the the fire arrival time is applied over the whole fire domain. The methods developed here use the Matlab function ``imgaussfilt" \citep{Matlab-2021-2GF}) to do this smoothing. The function uses a Gaussian smoothing kernel to produce a local averaging of the fire arrival time. Figure \ref{fig:imgauss} shows the effect of applying this smoothing to an example of a forecast fire arrival time. 

\subsubsection{Multigrid approach}
\begin{figure}[!h]
  \begin{center}
\includegraphics[width = 0.45\textwidth]{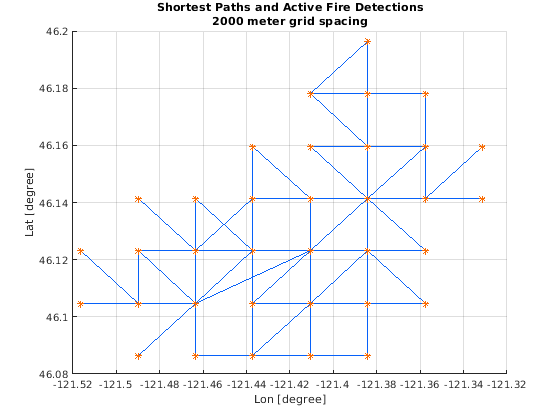}
\includegraphics[width = 0.45\textwidth]{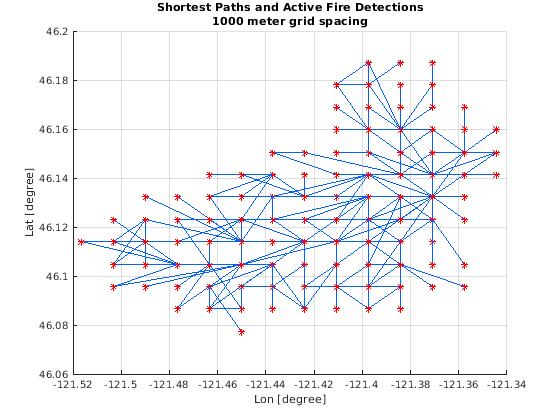}

\includegraphics[width = 0.45\textwidth]{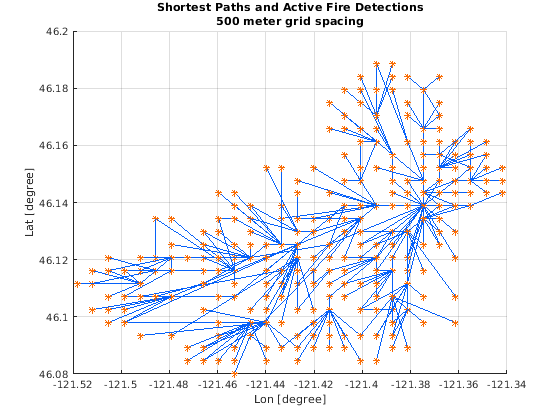}
\includegraphics[width = 0.45\textwidth]{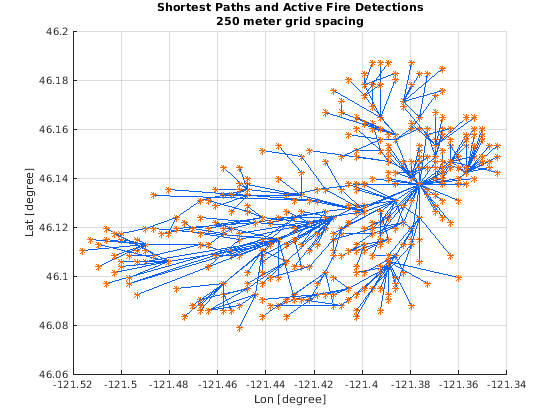}
    \caption[Satellite data and shortest paths interpolated on a sequence of computational grids of decreasing grid spacing.]{Satellite data and shortest paths interpolated on a sequence of computational grids of decreasing grid spacing. When using the multigrid approach for estimating the fire arrival time,  iterative interpolation is performed several times on computational meshes of increasing resolution. }
    \label{fig:patch_multi_grids}
    \end{center} 
\end{figure}

Even with local averaging of the fire arrival time by use of a Gaussian smoothing filter, roughness in the fire arrival time can develop. Changing the fire arrival time at scattered locations in the fire domain can introduce many local minima in the fire arrival time that are not consistent with a fire that has spread from point to point, outward from an ignition point. In extreme cases, the effect can make it appear that several separate ignitions have occurred in the fire domain. One method to help avoid introducing these spurious ignitions is to perform the iterative interpolation on a sequence computational grids that start with a coarse resolution and are progressively refined. The estimated fire arrival time obtained from iterative interpolation on a coarse grid becomes the initial estimate for iterative interpolation on the next, refined computational grid. Figure \ref{fig:patch_multi_grids} shows the interpolated fire detections and shortest paths on a sequence of grids with increasing spatial resolution.  This example shows the key point of the process; working on an initially coarse grid reduces the relative sparsity of the satellite data. Comparing the upper left panel (2000m resolution) with the lower right panel (250m resolution), we see a contiguous section of the fire domain with a fire detection at every grid point in the coarse resolution, but there are many grid locations in the fine resolution without any detections.   Figure \ref{fig:multigrid_progression} shows a sequence of estimates of a hypothetical one-dimensional fire arrival time that were made by interpolation on computational grids of increasing spatial resolution. The progression of images from left to right and top to bottom shows a refinement of the grid spacing used for interpolation.  In this example, no additional points were interpolated along the paths in the directed graph. Note that in the lower right, with 128 grid cells resolution, that the gradient in the ``flat part" on the left part of the domain at time $t\approx 5.5$ is only partially resolved when compared to the ``ground truth." Interpolating additional points along the paths can help overcome this apparent shortcoming. Figure \ref{fig:multi_1d_interp} shows the same hypothetical case, but additional points have been interpolated along the paths. In the final panel, we see the ``flat part" on the left part of the domain at time $t\approx 5.5$ is more resolved when compared to the example when no additional points were interpolated along the paths.

The process of using the multigrid approach is outlined below. 
\begin{enumerate}
\item Make an initial estimate of the fire arrival time.
\item Interpolate the estimated fire arrival time and detection data to a coarse grid. A 250 meter spacing was used.
\item Perform one step of the iterative interpolation as described in Section \ref{sec:iter_interp}. This entails adjusting the fire arrival times at the detection locations, followed by smoothing of the data.
\item Interpolate the fire arrival time back onto the original, fine grid and compute the relative difference between the initial estimate and the adjusted fire arrival time. If the relative difference is below a chosen threshold, end the multigrid interpolation routine.
\item If the relative difference is above a chosen threshold, interpolate the fire arrival time to a finer grid than used in the previous iteration and perform an additional step of the iterative interpolation method. The grid spacing was changed by  20\% on the successive iteration. 
\item Repeat the process until relative difference between estimates falls below the threshold or until the grid spacing gets to a minimal size that is chosen ahead of time. In this research, a minimal grid spacing of 250 meters was chosen since a smaller spacing would imply an overfitting of the satellite data that is of a coarser resolution.
\end{enumerate}

\begin{figure}[!ht]
\centering
\includegraphics[width=0.45\textwidth]{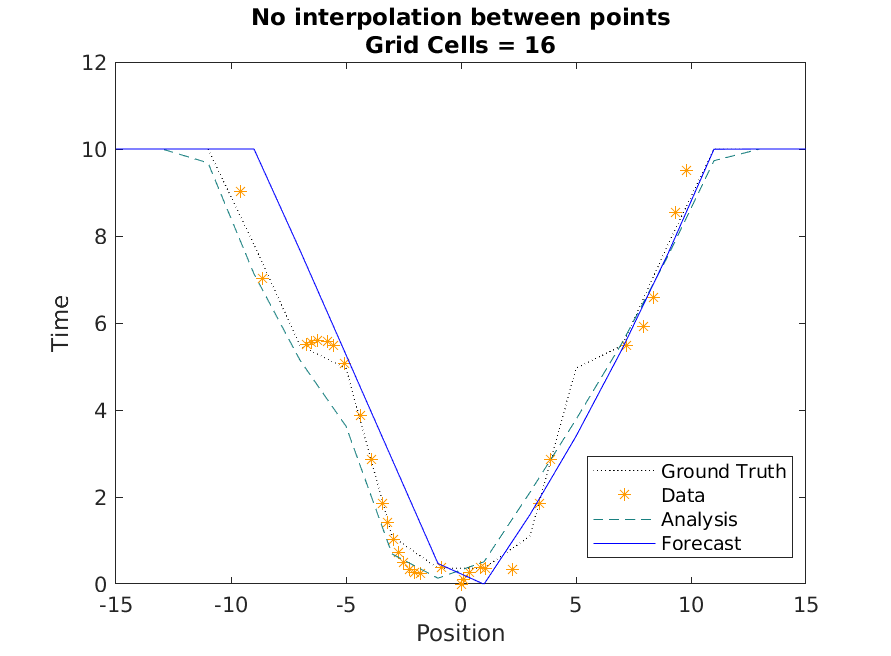}
\hfill
\includegraphics[width=0.45\textwidth]{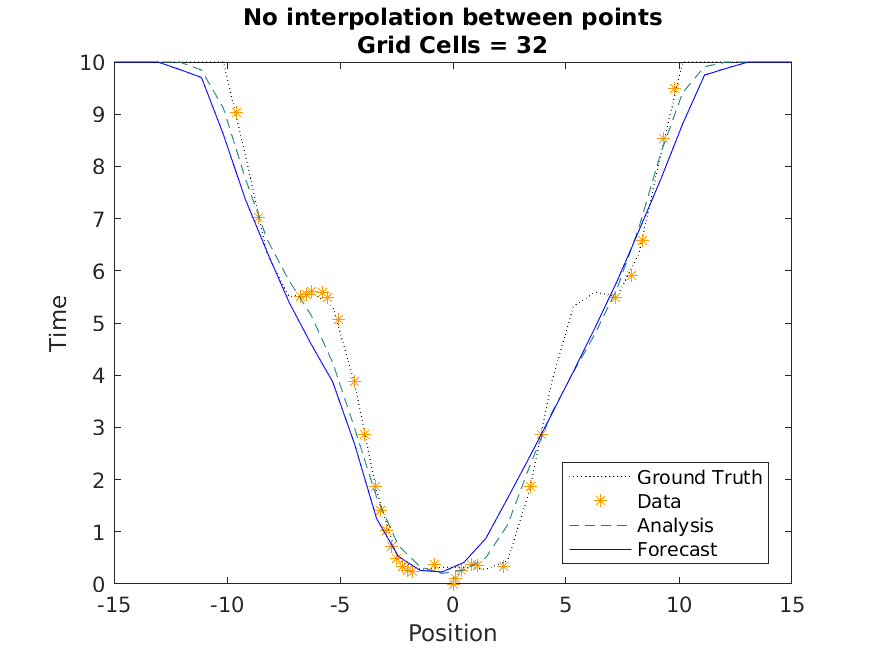}

\includegraphics[width=0.45\textwidth]{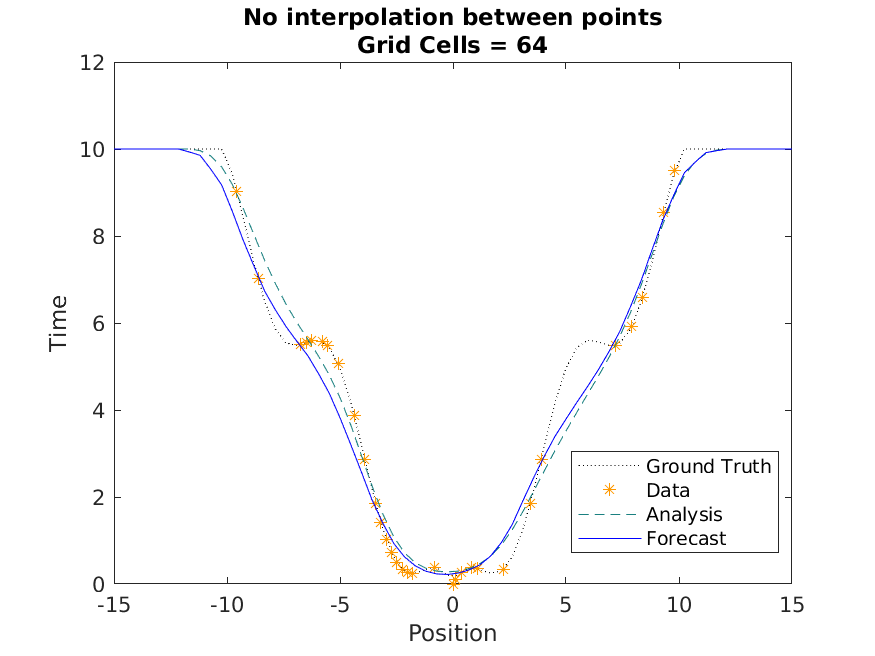}
\hfill
\includegraphics[width=0.45\textwidth]{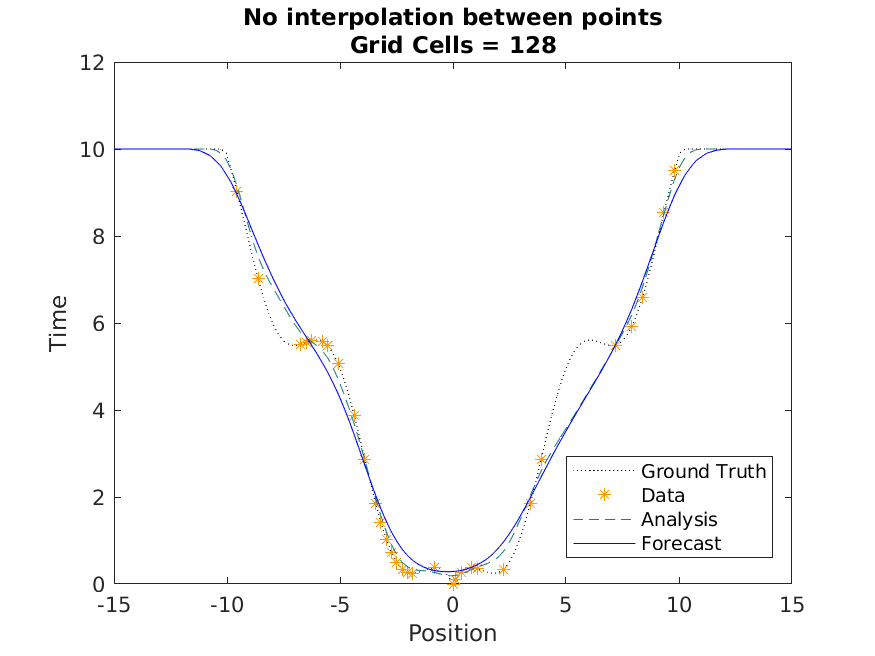}
\caption[Iterative interpolation with a multigrid strategy for data assimilation in a hypothetical one-dimensional fire without using interpolation of additional points.]{Iterative interpolation with a multigrid strategy for data assimilation in a hypothetical one-dimensional fire without using interpolation of additional points.}
\label{fig:multi_1d_no_interp}
\end{figure}

\begin{figure}[!ht]
\centering
\includegraphics[width=0.45\textwidth]{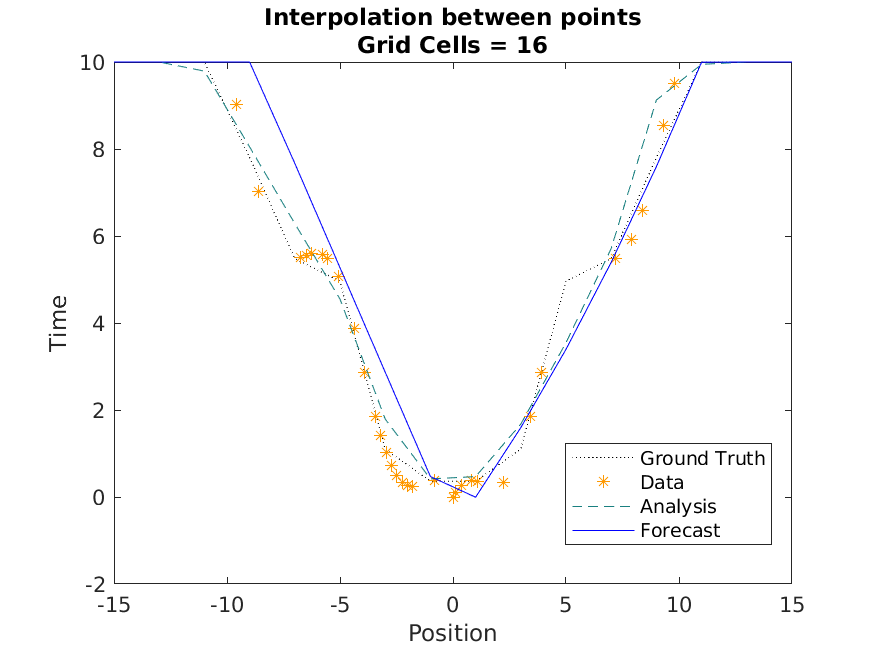}
\hfill
\includegraphics[width=0.45\textwidth]{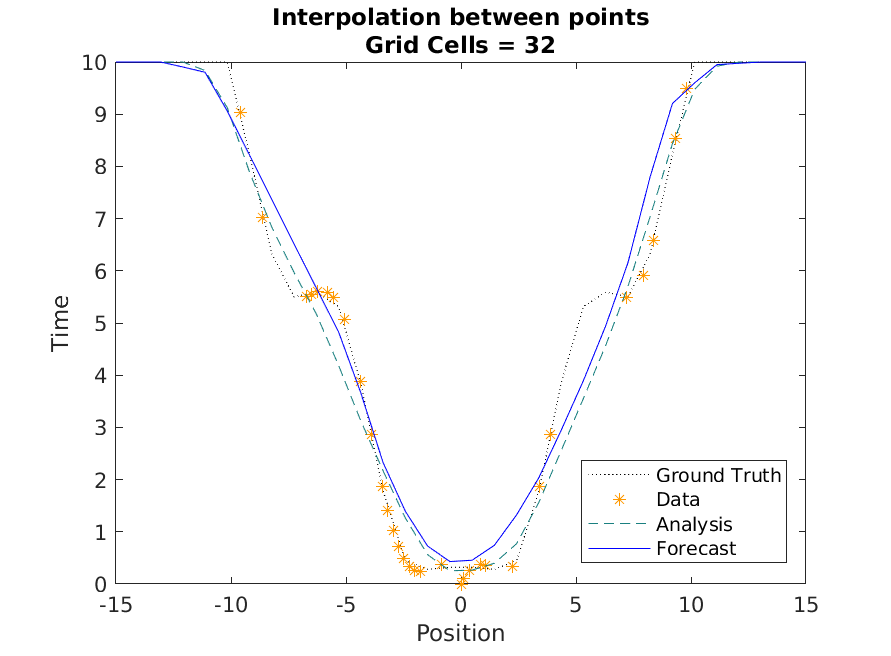}

\includegraphics[width=0.45\textwidth]{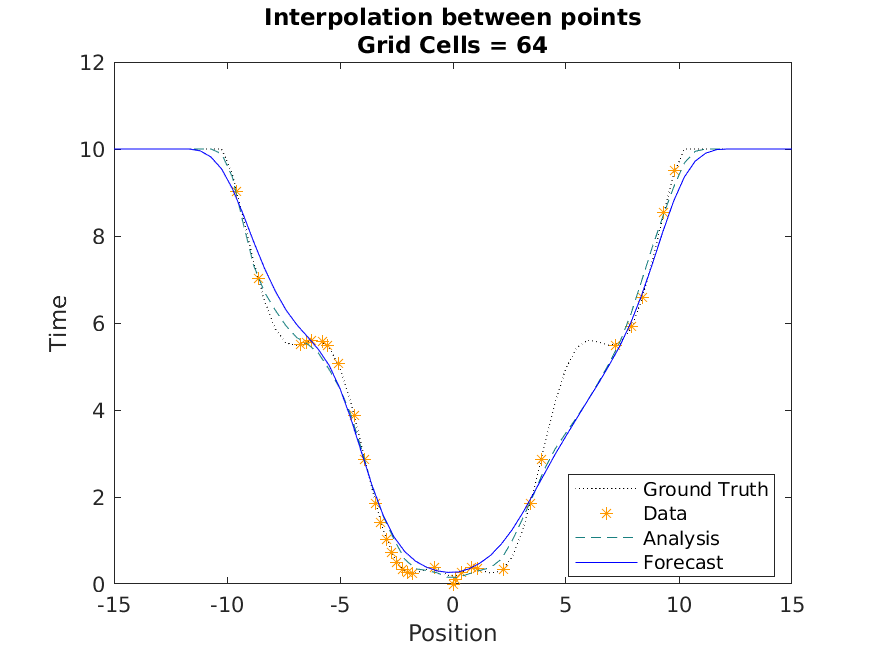}
\hfill
\includegraphics[width=0.45\textwidth]{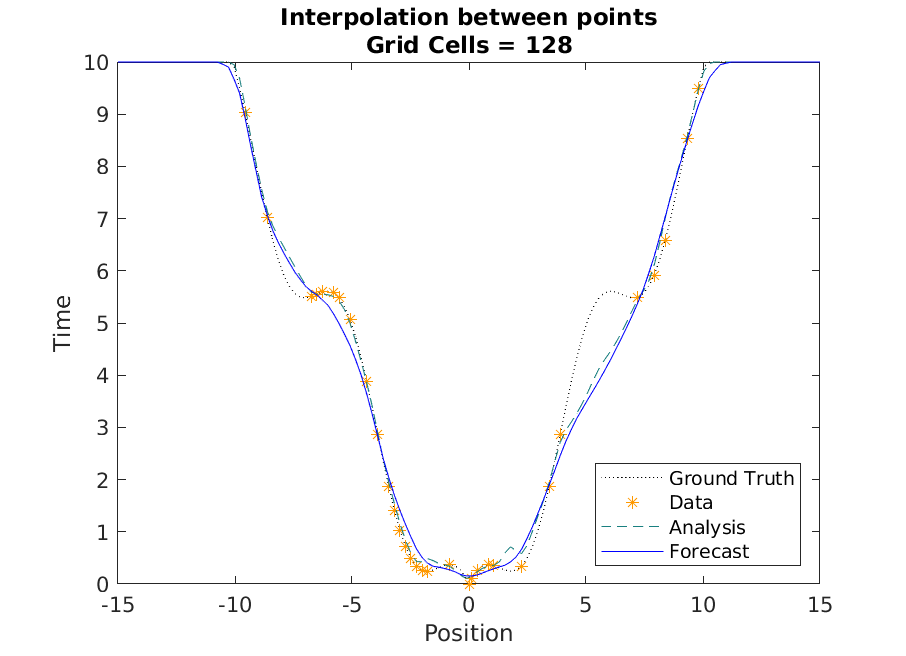}
\caption[Iterative interpolation with a multigrid strategy for data assimilation in a hypothetical one-dimensional fire using interpolation of additional points.]{Multigrid progression in a one-dimensional fire. In this example, additional points were interpolated along the paths in the directed graph. Note that in the lower right, with 128 grid cells resolution, that the gradient in the ``flat part" at time $t\approx 5.5$ on the left part of the domain is more fully resolved than when no interpolation is done as shown in Figure \ref{fig:multi_1d_no_interp}.}
\label{fig:multi_1d_interp}
\end{figure}

\subsection{Use of Non-fire Pixels}
The ``fire mask" that is part of the Level 2 fire data product gives locations of where the detection algorithm believes active fires to exist, but it also give locations for other kinds of observations as well. Importantly, the locations where the fire detection algorithm believes no fires are burning are reported. Additionally, locations where no determination can be made are reported. In these cases, possibly the presence of clouds or smoke may be obscuring the instrument's view. 

\begin{figure}[!h]
  \begin{center}
    \includegraphics[width = 0.45\textwidth]{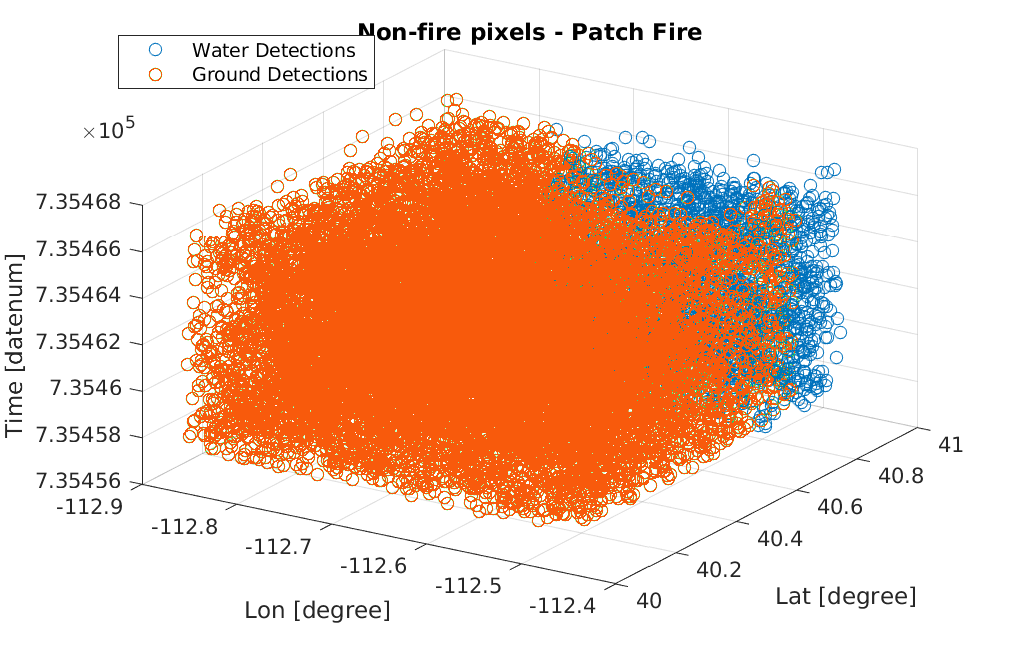}
        \includegraphics[width = 0.45\textwidth]{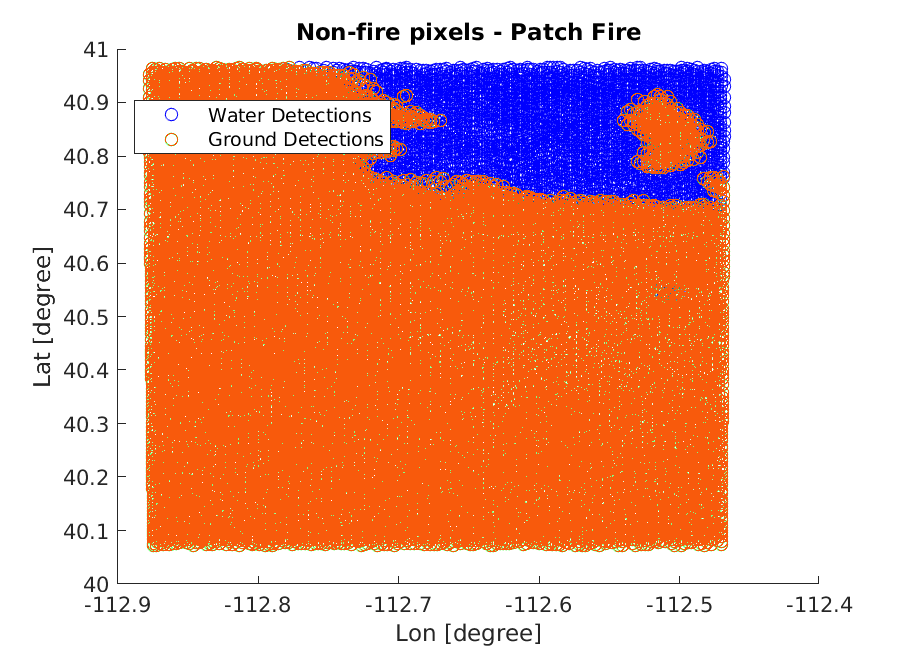}
    \caption[Non-fire pixels in the fire domain of a simulation of the Patch Springs fire.]{Non-fire pixels in the fire domain of a simulation of the Patch Springs fire.  The figure on the left shows the three-dimensional view. On the right, the data has been projected onto a plane.}
    \label{fig:patch_non_fire}
    \end{center}
\end{figure}

In this section we outline the method by which the locations where the detection algorithm has determined no active fire exist can be used to help shape the fire arrival time. The fire mask returned by the algorithm is a matrix of integers, 0-9, with specific meaning. Integers 7,8, or 9 indicate the presence of a fire with an associated confidence level of low, nominal, and high, respectively. Non-fire pixels are indicated by the integers 3 or 5 in the fire mask, indicating a non-fire water pixel or a non-fire land pixel, respectively. No confidence level is associated with these observations. For our purposes, the distinction between non-fire land or water pixels is unimportant, and we will treat them similarly. Figure \ref{fig:patch_non_fire} illustrates an aggregation of non-fire land and water pixels over a period of several days. To show the non-constant nature of the data, only 10\% of the detection pixels have been plotted in this figure. The blue pixels represent water detections most likely caused by proximity to the Great Salt Lake and orange pixels represent non-fire land detections.

As a first step towards using the locations of non-fire pixels, all satellite granules intersecting the the fire simulation domain are collected and examined. The times and the locations of the non-fire pixels are stored in a matrix for all granules whose times fall between the simulation start time and the simulation end time. This time period coincides with the time period over which the active fire data is stored for use in making the graph detailed in Section \ref{sec:graph}.

\begin{figure}[!ht]
\centering
       
       \includegraphics[width=0.45\textwidth]{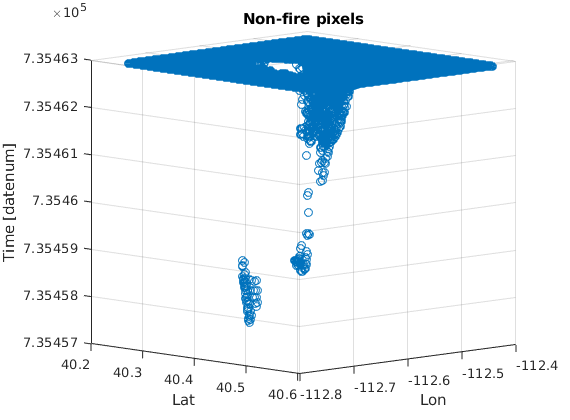}
       \includegraphics[width=0.45\textwidth]{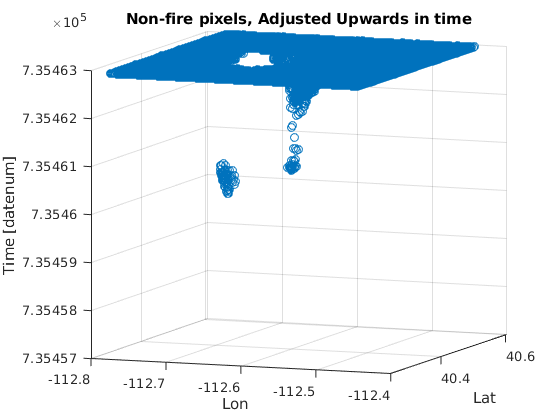}
       
       \includegraphics[width=0.45\textwidth]{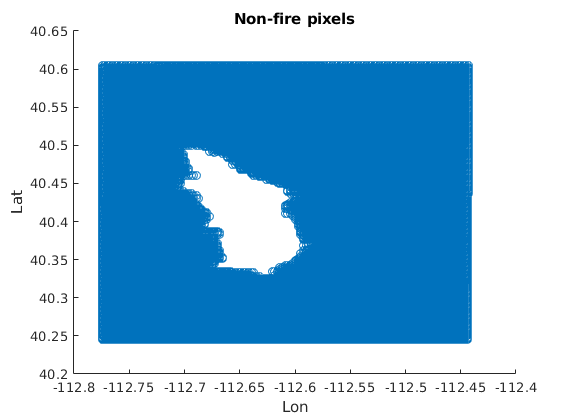}
     \caption[Adjustment of non-fire pixels during the interpolation of the fire arrival time from satellite data.]{Adjustment of non-fire pixels during the interpolation of the fire arrival time from satellite data. The white area in the center of the bottom panel is contained within a polygon drawn around the active fire detections. In the upper left are the non-fire pixels, scattered on the forecast fire arrival time surface. In the upper right, the fire arrival times have been moved upwards.}
\label{fig:patch_ground_detects}
\end{figure}

\begin{figure}[!ht]
\centering
       \includegraphics[width=0.45\textwidth]{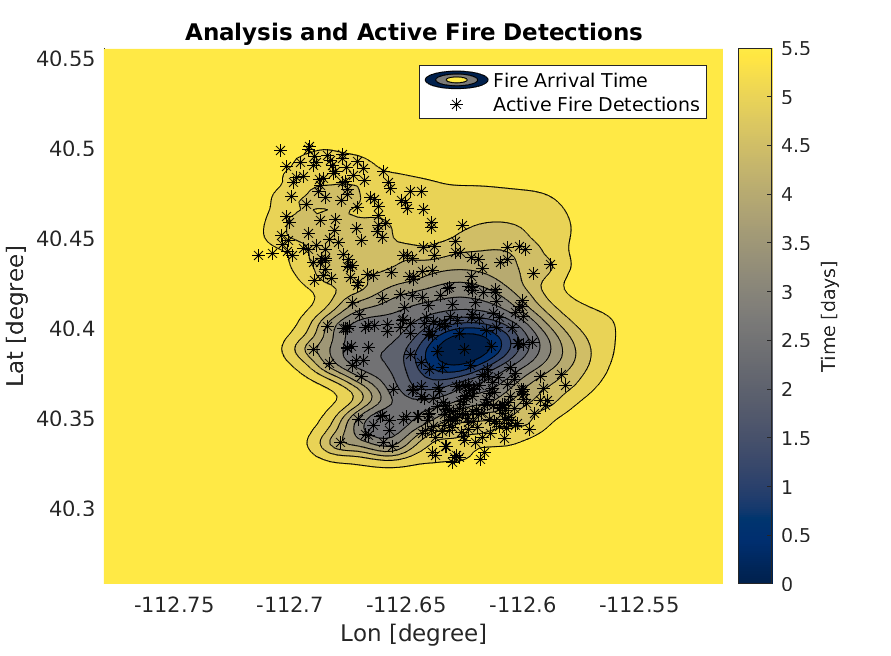}
       \includegraphics[width=0.45\textwidth]{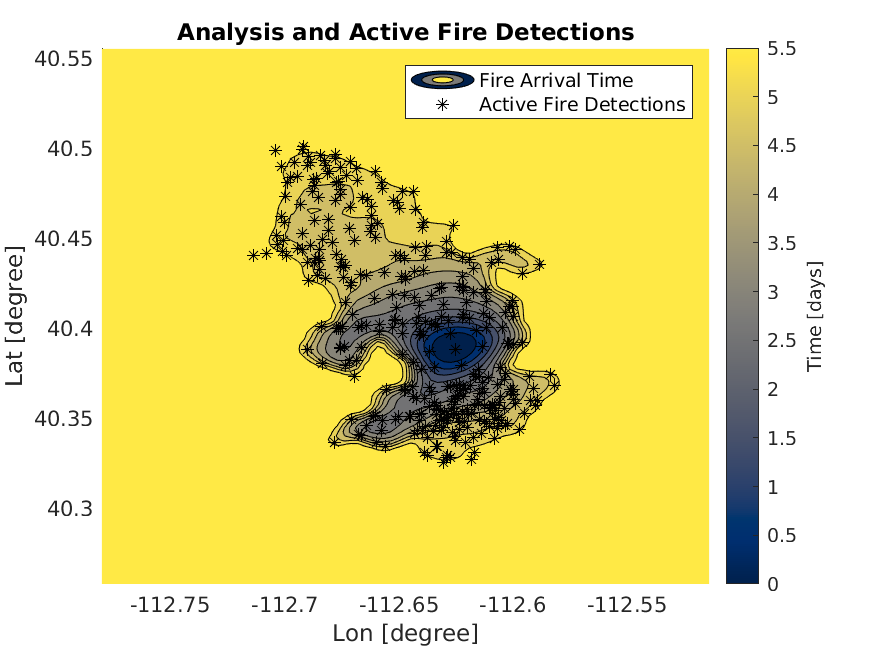}
     \caption[Comparison of the estimated fire arrival time made with and without use of non-fire pixel data.]{Comparison of the estimated fire arrival time made with and without use of non-fire pixel data. The left panel shows the estimated fire arrival time formed without use of non-fire pixels. Note the perimeters extend into regions not containing any fire detection pixels. The panel on the right shows the  estimated fire arrival time formed using non-fire pixel data. The outer perimeter more closely matches the locations where active fire detections are located. Compare with Figure \ref{fig:exp_using_ground} to see how the interpolation of fire arrival time is handled differently when data assimilation is being performed.}
\label{fig:exp_est_using_ground}
\end{figure}

When all fire pixels and non-fire pixels are collected, the fire pixels are aggregated and a polygon is drawn around them. Non-fire pixels outside of this polygon are used to adjust the fire arrival time upward towards the end time of the simulation. Non-fire pixels located within the polygon enclosing the active fire detections are ignored. In some cases, the non-fire pixels within the polygon may have been recorded in locations subsequently reached by the fire. In other cases, we cannot rule out the possibility that fire has never reached these locations. After the fire arrival time has been adjusted for the region outside of the polygon, local averaging in the whole fire domain is applied using a Gaussian smoothing kernel. Figure \ref{fig:patch_ground_detects} shows the effect of moving the fire arrival time of non-fire pixels upwards in the region outside of the boundary containing active fire detections. The bottom panel shows a view of the non-fire pixels outside of the boundary as seen projected onto a plane. The upper left and right panels show the three-dimensional view of the non-fire pixels before and after the adjustment, respectively. It should be noted that the local averaging of data applied after the adjustment of the fire arrival prevents an overfitting of that data that would occur if the the fire arrival times of grid locations outside of the boundary was simply set to the end time of the simulation. The difference between not using and using non-fire detection data for estimating the fire arrival time is illustrated in Figure \ref{fig:exp_est_using_ground}. The left panel shows an estimate made using only active fire detections and it can be seen that the outer perimeter of the fire extends beyond the fire detections in many places. The use of the non-fire detecions in the right panel shows the outer perimeter constrained to follow the data more closely.


\begin{figure}[!h]
  \begin{center}
    \includegraphics[width = 0.45\textwidth]{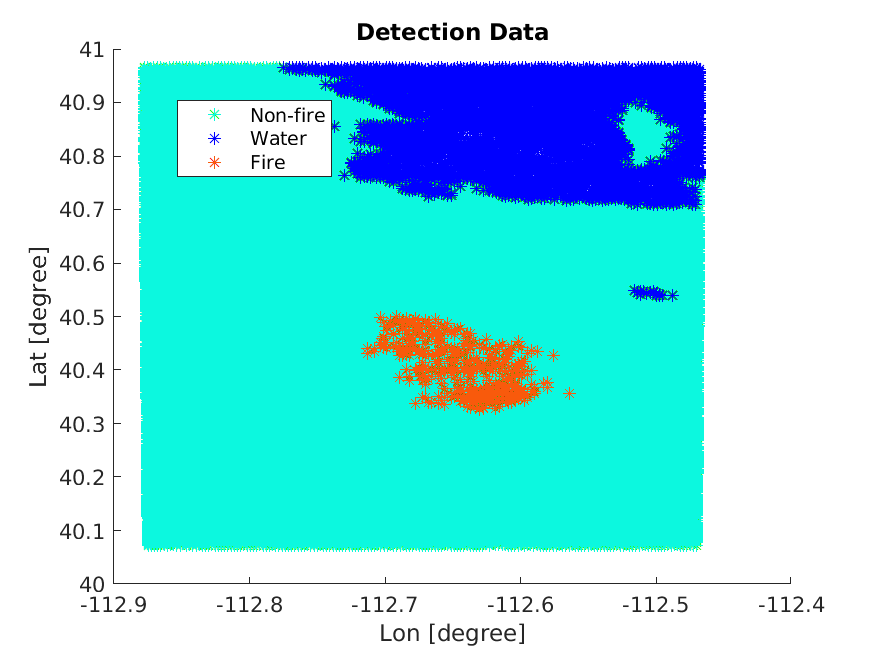}
    \includegraphics[width = 0.45\textwidth]{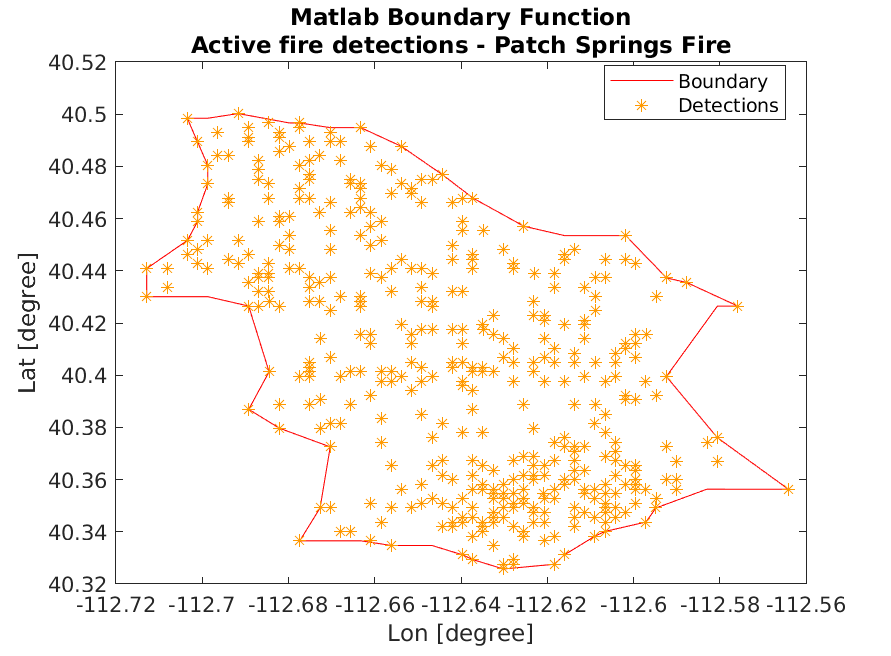}
    \caption[Showing fire and non-fire satellite data for the Patch Springs Fire.]{Showing fire and non-fire satellite data for the Patch Springs Fire. On the left, fire, non-fire, and water pixels from the fire mask are shown simultaneously . The boundary around active fire detections is shown on the right. Non-fire pixel data is used to adjust fire arrival times outside of this region.}
    \label{fig:patch_non_fire2}
    \end{center}
\end{figure}

\subsection{Estimating the Rate of Spread from Satellite Data}

In general, it is not possible to know the rate of spread of a wildfire at every point in a fire domain. However, knowing the ROS has many useful applications. For example, a high ROS might indicate the presence of unobserved winds in the fire domain or indicate that the underlying fuels are relatively dry.  Having an estimate of the fire arrival time at all points in the fire domain will allow us to find an estimate of the ROS at those points.

The ROS is determined by the gradient of the fire arrival time. Given that the fire arrival time $T = T(x,y)$ assigns a fire arrival time to each location in the fire domain,  the partial derivatives $T_x$ and $T_y$ are computed to obtain the change in fire arrival time per distance in the $x$ or $y$ directions. The reciprocals of these partial derivatives are the ROS components in the respective directions. With fire spreading in the direction normal to the level curves of the fire arrival time $T$, we see that the reciprocal of the ROS, $R$ is is magnitude of the gradient of the fire arrival time
\begin{equation}
\left\| \nabla T \right\| = \frac{1}{R}.
\label{eq:eikonal}
\end{equation}


%


\section{Tools for Assesing the Method}
\label{sec:assess_tools}
This sections presents some tools that will be used to assess the effectiveness of the proposed method for estimating the fire arrival time. The tools will be used in various tests to find the best values for important parameters used in the method. Later, the tools for assessment will be used to judge how well the method works for estimating the fire arrival time for real-world fire scenarios

\subsection{Average Growth Rate}
An important measure of how well a simulation or fire arrival time estimate agrees with the real-world fire is made by a comparison of the average fire growth rate. This measure gives a sense of how well the estimate manages to give the size of the fire over time. This measure is crude in the sense that it cannot say anything about how well the estimates manages to reflect the true state of the fire at specific locations. Still, knowing or predicting how many acres per day will be consumed by a fire can help fire responders allocate resources or issue public safety warnings.

To calculate the growth rate over a period of time, the starting and ending times of the even are partitioned into partitioned into discrete steps and the area of the fire at that each time step is estimated by one of the following methods.
\begin{enumerate}
\item If only satellite active fire detections are available, the fire arrival time is estimated as in section \ref{sec:fat_method} and the area is computed from the number of grid cells for which the fire arrival time is less than or equal to the discrete time step.
\item If, additionally, infrared perimeter data is available, a subset of the points in the perimeter are used in the same way as active fire detections to obtain an estimated fire arrival time.
\item When using artificial fire arrival times as the ``ground truth" for experiments, the area at each time step is computed from the number of grid cells for which the fire arrival time is less than or equal to the discrete time step.
\item If a large number of infrared perimeter observations are available for a real-world fire, time may discretized by the time of the observations and the area at each time step computed as the area within the polygon defined by the perimeter observation.
\end{enumerate}

\begin{figure}
\centering
\includegraphics[width=0.6\textwidth]{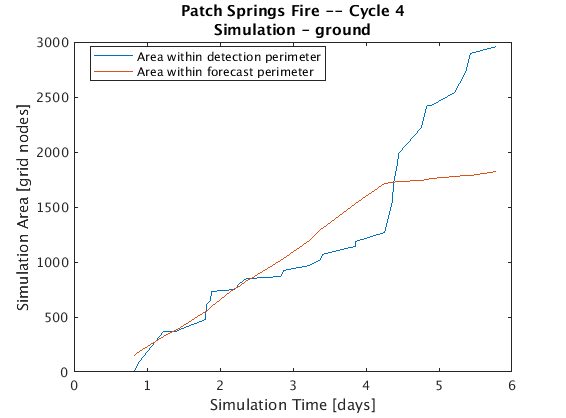}
\caption[Plot of estimated fire area over time.]{Plot of estimated fire area over time. The blue line indicates the size of the actual fire, estimated from satellite detection data. The red line is the size of the modeled fire. It can be seen that the model does not capture the explosive growth that occurred.}
\label{fig:growth_example}
\end{figure}

The growth rate of the actual fire and the model fire growth can be compared visually to give a sense of how well the simulation matched the actual fire. Figure \ref{fig:growth_example} shows an example of such a comparison where the slope of the curves depicting the size of the fire over time are roughly the same of the first four days of the fire, but are very different after that time. This divergence indicates that the model is failing to capture the behaviior of the real fire after the fourth day. For testing purposes, when evaluating many simulations or estimates of fire arrival times, for example, it is convenient to compute a single number, the relative growth error \begin{equation}\label{eq:relative_growth_error}
RGE =\frac{||A_e - A_g||_2}{||A_g||_2}
\end{equation}
where $A_e$ is the estimated fire area at each discrete time step, $A_g$ is the ``ground truth" fire area at the corresponding time step, and the norm is the Euclidean. The $RGE$ gives a sense of how well the estimate or simulation matches the real-world fire growth but does not give information about whether the estimated growth rate was too high or too low.

\subsection{Measure of Effectiveness}
\begin{figure}[!h]
\begin{center}
\includegraphics[scale = 0.5]{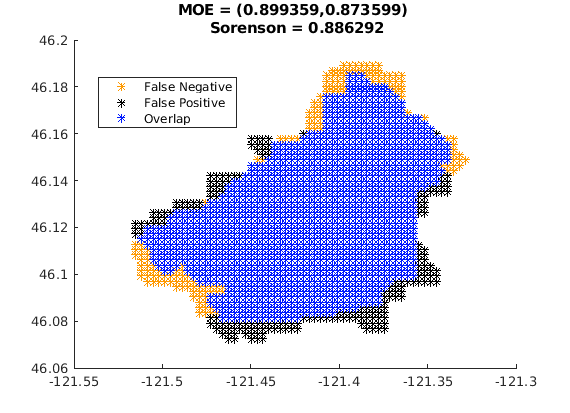}
  \caption[Illustrating the measure of effectiveness (MOE).]{Illustrating the measure of effectiveness (MOE). The MOE is computed from the areas of false positives, false negatives, and overlap. The overlap is the blue region where the model and real-world observations have both indicated burning. The observed area of the actual fire is the union of all blue and orange pixels. The area of the simulated fire is the union of the blue and black pixels. In this example, the area of the modeled fire is close to that of the observed fire.  }
  \label{fig:moe_example}
  \end{center}
\end{figure}

The measure of effectiveness (MOE) as presented in \citet{Warner-2004-UTM} is a two-dimensional statistic that evaluates a model output with a  comparison of the known state of the system, describing the decreasing amount of either false positives or false negatives in the forecast or estimate. In terms of fire simulations, a false positive represents the model predicting the fire to arrive at a location where no fire was actually observed. Conversely, a false negative indicates a location where fire was actually observed but the model indicated that location remained unburnt. It is computed as

\begin{equation}\label{eq:moe}
MOE = (x,y) = \left(\frac{A_{OV}}{A_{OB}},\frac{A_{OV}}{A_{PR}} \right) =
\left( 1 - \frac{A_{FN}}{A_{OB}}, 1 - \frac{A_{FP}}{A_{PR}}\right),
\end{equation}
where $A_{FN}$ and $A_{FP}$ given the area where false negatives and false positives occurred, respectively. $A_{OB}$ and $A_{PR}$ are the areas of the real-world observation and the model prediction, respectively, and $A_{OV}$ is the overlap area, where the observation and prediction coincide. Figure \ref{fig:moe_example} shows the relationship of these regions for a simulation that overestimated the size of the fire and had a score $MOE = (x,y) = (0.899359,0.873599)$.  Importantly, the region of the observed fire is the union of all blue and orange pixels and the region of the forecast fire is the union of all blue and black pixels. If the forecast and observations coincided exactly, the graphical representation would be a solid blue color. If the fire areas of the forecast and observations were disjoint, there would be no blue in the figure since all locations would indicate either a false positive or a false negative.

The MOE gives a good sense of how well a simulation or estimation of a fire arrival time matches the actual fire in terms of the locations where the fire occurred at a single time, but it cannot give a sense of how well the simulation evolved over the period of the event. When using the MOE to asses how well a method works it is possible to consider separately the tendency to produce false positive or false negative indications at locations in the fire domain. For operational use, it might be preferable to have a model that makes errors on the side of caution by producing more false positives than false negatives.

\subsection{S{\o}renson Index}
The S{\o}renson index is another method of assessment  made by comparison of regions contained within a fire perimeter. In the case of a real fire example, infrared fire perimeters as observed from overflying aircraft will be used to compare against the region of an estimated fire arrival time. For artificial data, perimeters are established by finding the location of all fire arrival times before a specified time associated with a perimeter. This method of evaluating simulations is algebraically equivalent to the method employed by in \citet{Cardil-2019-ARW} and was proposed by S{\o}renson as measure of the similarity between biological samples \citep{Sorenson-1948-MEG}. A score $S_a$ is assigned to each comparison of the forecast with the perimeter according to the formula

\begin{equation} \label{eq:sorenson}
    S_a = \frac{2\times A(forecast \cap perimeter)}{A(forecast)+A(perimeter),}
\end{equation}
where $A$ is the area of a region. Thus $A(observed)$ is the area of the fire contained within the observed perimeter, $A(forecast)$ is the area enclosed within the model forecast perimeter at the time the perimeter observation was made, and $A(forecast \ \cap \ observed)$ gives the area of the intersection of the forecast and observed perimeter areas. When the forecast and observed perimeters encompass disjoint areas, we have  $A(forecast \ \cap \ observed) = 0$ so that $S_a = 0.$ This is the lowest score which can be assigned. Conversely, when the forecast and observed perimeters exactly coincide, then $2\times A(forecast \ \cap \ observed) = A(forecast)+A(observed)$ so that $S_a = 1$, the maximum score possible.

In the case that there are $n$ perimeter observations during a simulation period, the score $S_a$ can be computed as the average

\begin{equation}
S_a = \frac{1}{n}\sum_{i=1}^n s_i.
\label{eq:area_score}
\end{equation}

The S{\o}renson index is similar to the MOE in that is gives a sense of how well a simulation or estimate of the fire arrival time compares with the actual fire in terms of area, but it does not give a sense of the goodness of fit in a temporal sense. The chief advantage of the S{\o}renson index is the simplicity of the measure. The single number is easy to interpret since it measures how well the forecast fire area matches that of the real fire but it does not provide information on whether the predicted fire area was overestimated or underestimated.

\subsection{Relative Error}
The relative error (RE) is the norm of the difference between the estimated and  actual fire arrival times divided by the norm of the actual fire arrival time. This score gives a good sense of how well the simulation or estimate of the fire arrival time was able to match the data. This method of assessment was used for tests using artificial fire scenarios to validate the methods being developed. Since the fire arrival times of real fires are generally not known, this method of assessment is not used when working with real fire scenarios. The relative error is calculated as 
\begin{equation}\label{eq:rel_error}
RE = \frac{||T-T_e||}{||T||}
\end{equation}
where $T$ is the known fire arrival time, recorded as days since simulation start, and $T_e$ is an estimate derived from observational data.

\subsection{Rate of Spread Error}
This assessment score seeks to measure how well the average ROS in a  forecast or estimate of fire arrival time compares with a known rate of spread. The ultimate intention is to be able to use this score to make a judgment about the accuracy of the burn model in SFIRE and improve forecasts by adjusting a key model parameter, the fuel moisture content (FMC). Because the total FMC is dependent on the underlying fuels in the fire domain, comparison between the forecast ROS and the known ROS are compared at each grid point in the fire domain where the forecast and the data both agree that fire had occurred. Thus, the score is computed only in the ``overlap region" that is used to  compute the MOE and corresponds to the blue region seen in Figure \ref{fig:moe_example}.

The ROS is computed using the Matlab function ``gradientm" \citep{Matlab-2021-GSA}) that uses the latitude and longitude of a location and a scalar field, in this case the fire arrival time, to compute the gradient with units in meters and seconds. Since $T = T(x,y)$, the gradient of the fire arrival time gives the reciprocal of the ROS, and is, in fact, the Eikonal equation

\begin{equation}
\left\| \nabla T \right\| = \frac{1}{R}.
\end{equation}

The ROS at a location can be found by the reciprocal of the gradient's norm. 

Although the ROS is computed from the fire arrival time $T= T(x,y)$ the rate of spread is more complicated than a function of only position. Other inputs such a fuel properties, weather, and terrain slope play a role in determining the ROS. Indeed, since the weather varies with time, the ROS is partially a function of the fire arrival time itself. Thus, making adjustments to the model based on an estimate of the ROS computed only from position is problematic.

\subsection{Spread Direction Error}
When the ROS is computed from the gradient of the fire arrival time, the partial derivatives $(T_x,T_y)=\nabla T$  allow for computation of the direction of the fire spread. The arctangent function is used to compute an angle
\begin{equation}
\theta = \arctan\frac{T_y}{T_x}
\end{equation}
that describes the direction of the gradient in radians counterclockwise from the direction east. 

\section{Examples and Testing the Method}\label{sec:example_test}
This section investigates how the method for estimating the fire arrival time works and shows some of the tests that were performed to find the best parameters that control how  different parts of the method are accomplished. Because the fire arrival time for real fires is usually not known, we use the strategy of constructing some artificial fire arrival time and artificial data so that a ``ground truth," known solution to a problem exists to be compared against the estimate produced by the proposed method. 

Artificial fire arrival times were constructed by a process that first created several cones with a common vertex and height, but varied in the width of the cone at the top. The cones were then combined by taking the maximum value among all the cones at all points in the domain. For example, if the cones could be determined by the functions $f_1(x,y),f_2(x,y),f_3(x,y)$, the artificial fire arrival time $T(i,j)$ would be determined at location $(i,j)$ in the fire domain by
\begin{equation}
T(i,j) = \max\{f_1(x,y),f_2(x,y),f_3(x,y)\}
\end{equation}
Artificial fire detections were randomly scattered on the surface of the artificial fire arrival time comes at locations where the fire arrival time was less than the time corresponding to the ``flat top" of the cone. A random number was generated for each potential location and if that number was below a chosen threshold, an artificial fire detection was created at that point. In the experiments that follow, roughly 5\% of the potential detection locations were assigned an artificial detection time. The top row of panels in Figure \ref{fig:new_test_2000_contours} shows a sampling of some these artificial data. The random process produced a number of different shapes of the artificial fires as can be seen. Some shapes were similar to simple cones, but others can be seen as disjoint regions implying the presence of more than a single fire in the domain. Figure \ref{fig:fat_227} shows one such example.

\subsection{Testing the Method}
\subsubsection{Choosing the Number of Clusters}
\label{sec:cluster_test}
When clustering data, the number of clusters $k$ must be chosen by the user. There exist algorithms to find an optimal number $k$ by seeking to minimize the squared distances from cluster members to the centroid of the cluster. Those algorithms have not been used in this research. For the purpose at hand, the optimal number of clusters to use will be that that allows the method for estimating the fire arrival time from satellite detections to be most accurate.

\begin{figure}
\centering
\includegraphics[width=0.6\textwidth]{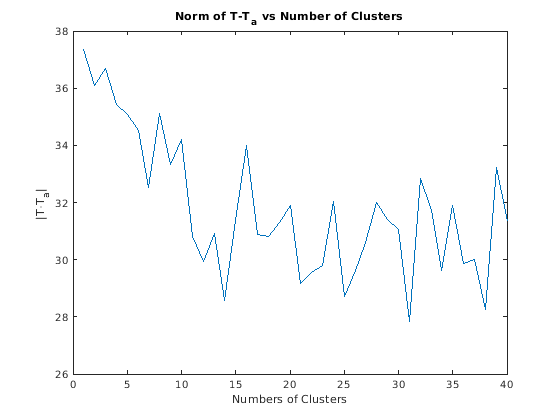}
\caption[The effect of cluster size on the  estimated fire arrival time produced from graphs derived using various numbers of clusters.]{The effect of cluster size on the  estimated fire arrival time of a single example case produced from graphs derived using various numbers of clusters. The horizontal axis shows the numbers of clusters used to partition the detection data and the vertical axis shows the norm of the difference between the ``ground truth" fire arrival time $T$ and the estimated fire arrival time $T_a$ derived from the detection data. The norm of the difference decreases generally when more clusters are used, but then reaches a plateau when approximately 15 clusters are used. Other example cases showed similar characteristics. }
\label{fig:clusters_plot}
\end{figure}

To find an optimal number of clusters to use, the method for estimating the fire arrival time was  repeatedly employed, using a varying number of clusters and the estimate produced was compared with the ``ground truth".  Directed graphs and sets of shortest paths using between one and forty clusters were used. For each cluster size, the norm of the difference between the ``ground truth" and the estimated fire arrival time was computed. Figure \ref{fig:clusters_plot} shows the results of varying the number of clusters in a single example case. In general, when the cluster size reaches approximately 15, the error reaches a plateau. 

This test was repeated for several different ``ground truth" scenarios, involving at least 100 points of data used as artificial detection locations and the results were similar. For that reason 20 clusters has been used as a default value for the work. There are a few exceptional cases where using a different number of clusters would make sense. If the number of fire detections is small, the number of clusters should be reduced. Indeed, if the number of fire detections is less than the number of clusters, the k-means algorithm cannot complete. If the number of detections equals the number of clusters, then each detection will belong to its own cluster and the purpose of clustering the data will not be fulfilled. For large fires, involving many thousands of fire detections, a larger number could be used although that has not been tested in the current work.

\subsubsection{Testing with Artificial Fire Arrival Time and Data}
\label{sec:ground_truth_1}
Because the fire arrival time of real-world fire is not known, to test the method we first begin with construction of artificial fire arrival times and artificial fire detection data. The method for estimation fire arrival time is then employed and the resulting estimated fire arrival time is compared with the artificial fire arrival time used as ``ground truth." In the first example, hundreds of artificial fire arrival time cones were generated, along with artificial detection data, and the interpolation method was performed using different parameters. The assessment methods outlined above were then used to find those parameters that gave better results.

The purpose of the experiment was to determine the optimal way to estimate the fire arrival time from only satellite data. In particular, choosing whether to use a multigrid approach or to work on a single grid was a goal. A secondary consideration was whether to interpolate extra points along the shortest paths in the directed graph and how many extra points to use. The conclusion of the test was that using the multigrid approach with extra data points interpolated along the paths at a distance of 2000 meters was the the best method.

In the experiment, 280 artificial fire arrival times were created. Artificial active fire detections were were randomly scattered across the fire arrival surface in such a way that approximately 5\% of the grid cells within the  outer perimeter would have an active fire detection. When the locations of the fire detections were determined, the time associated with each was adjusted upwards to a discrete set of times to mimic the way that real satellite data is received in granules.

The assessment of the estimated fire arrival times was done in two categories
\begin{enumerate}
\item Comparison of the fire arrival time in terms of its growth, final size and shape, as well as a relative error. 
\item Comparison of the rate of spread and the direction of the of the fire spread.  
\end{enumerate}

Acronyms for the assessment scores in first category described above.
\begin{itemize}
\item AGE - Average growth error. 
\item MRE - Mean relative error. 
\item SRE - Standard deviation of relative errors
\item MOE X - $x$-component of the measure of effectiveness. A higher score indicate less false negatives. 
\item MOE Y - $y$-component of the measure of effectiveness. A higher score indicates less false positives.
\item ||MOE|| - Norm of the the measure of effectiveness
\item S{\o}renson index
\end{itemize}

Acronyms for the assessment scores described above
\begin{itemize}
\item MRD - Mean ROS difference
\item SRD - Standard deviation of the differences in ROS
\item MDD - Mean direction of spread difference
\item SDD - Standard deviation  of the differences in the the direction of spread
\end{itemize}

Several different scenarios were tested. The goal was to find whether a multigrid approach was useful and to determine how interpolation of extra points along the paths in the graph affects the iterative interpolation process. This secondary goal was accomplished by making estimates of the fire arrival time without using interpolation along the paths as well as using extra points at various different spacings.  Spacing is taken to be the distance along the path between two original detection data. Extra points were interpolated along the paths so that the distance between any two points on the path was no greater than the spacing. For example, if the 1000 meter spacing was used but all detections on a path were 500 meters from their neighbors, no extra points would be interpolated. When the spacing indicated interpolation of extra points was to be made, three extra points were added to the path between the adjacent, original detections.

The following list describes them various strategies for estimating the fire arrival time used to test the method.
\begin{enumerate}
\item \textbf{Multi - NP}: Multigrid with no extra points interpolated
\item \textbf{Multi - 500}: Multigrid with extra points spaced at 500 meters
\item \textbf{Multi - 1000}: Multigrid with extra points spaced at 1000 meters
\item \textbf{Multi - 2000}: Multi grid with extra points spaced at 2000 meters
\item \textbf{Multi - 3000}: Multigrid with extra points spaced at 3000 meters
\item \textbf{Multi - 4000}: Multigrid with extra points spaced at 4000 meters
\item \textbf{Single - NP}: Single grid used with no extra points interpolated
\item \textbf{Single - 1000}: Single grid used with extra points spaced at 1000 meters
\end{enumerate}

Table \ref{tbl:multi_test} shows the results of this test relating the estimate to the ``ground truth" fire arrival time in terms of its growth, final size and shape, as well as a relative error. The rows of the table indicate various strategies for estimating the fire arrival time used to test the method. The columns of the table give the average test score for the estimated fire arrival time for all 280 artificial fire data scenarios created. The columns ``Best" and ``Worst" correspond to the specific artifical data scenario for which the strategy had the highest or lowest average of all test scores, respectively. At a glance, it can be seen that the mean relative error (MRE) is found to be lower when using the multigrid approach. Table \ref{tbl:multi_test_rank} ranks the averages of the scores according to which strategy produced the best results, with a lower rank indicating a better score. The column ``Rank Sum" adds up the test rankings for each strategy and the column ``Rank" rank shows these sums ordered by increasing values. The lowest value of this final rank corresponds to the strategy for estimating the fire arrival time by used of the multigrid approach with additional points interpolated along the paths with a spacing of 2000 meters.

\begin{table}
\begin{tabular}{|c|c|c|c|c|c|c|c|c|c|}
\hline 
Scenrario & AGE & MRE & SRE & MOE X  & MOE Y  & ||MOE|| & Sorenson &  Rank sum & Rank \\ 
\hline 
Multi - NP   & 2 & 6 & 6 & 1 & 6 & 5 & 5 & 31 & 4\\ 
\hline 
Multi - 500  & 7 & 1 & 4 & 8 & 1 & 8 & 8 & 36 & 6\\ 
\hline 
Multi - 1000 & 3 & 3 & 1 & 5 & 5 & 2 & 2 & 21 & 2\\ 
\hline 
Multi - 2000 & 1 & 2 & 5 & 4 & 2 & 1 & 1 & 16 & 1\\ 
\hline 
Multi - 3000 & 4 & 4 & 2 & 6 & 3 & 3 & 3 & 25 & 3\\ 
\hline 
Multi - 4000 & 6 & 5 & 3 & 7 & 4 & 4 & 4 & 32 & 5\\ 
\hline 
Single - NP  & 8 & 7 & 7 & 2 & 8 & 6 & 7 & 44 & 7\\
\hline 
Single -1000 & 5 & 8 & 8 & 3 & 7 & 7 & 6 & 44 & 8\\ 
\hline 
\end{tabular}
\caption[Ranking of the test results for estimating the fire arrival time from artificial data in terms of  growth, final size and shape, as well as relative error.  All of the 280 results of the experiment for estimating the fire arrival time from data are compiled into a single table.]{Ranking of the test results for estimating the fire arrival time from artificial data in terms of  growth, final size and shape, as well as relative error.  All of the 280 results of the experiment for estimating the fire arrival time from data are compiled into a single table.  The best method for estimating fire arrival time from data was to use the multigrid method with interpolation of extra points along the paths at a distance of 2000 meters. This table summarizes and ranks the results in Table \ref{tbl:multi_test}.}
\label{tbl:multi_test_rank}
\end{table}

Table \ref{tbl:ros_test} gives the results of this test in terms of the rate of spread and the direction of the of the fire spread. The rows of the table indicate various strategies for estimating the fire arrival time used to test the method. The columns of the table give the average test score for the estimated fire arrival time for all 280 artificial fire data scenarios created. The second and third columns  give the mean and standard deviation of the differences in the average ROS for each fire arrival time. The fourth and fifth columns give the mean and standard deviation of the differences in the angle of the gradient for each fire arrival time. It can be seen that the magnitude of the mean ROS difference (MRD) is lowest for the strategy using the multigrid approach and spacing of interpolated data points at 2000 meters.  Table \ref{tbl:ros_test_rank} ranks the average scores in the same was was done for the first part of the test. Again, it was found that using the multigrid approach with 2000 meter spacing of extra points produced the best results.

\begin{table}
\begin{tabular}{|c|c|c|c|c|c|c|}
\hline 
Scenario & MRD & SRD & MDD & SDD & Rank sum & Rank \\ 
\hline 
Multi - NP   & 4  & 6 & 2 & 5 & 17 & 3 \\ 
\hline 
Multi - 500  & 2  & 4 & 3  & 4  & 13   & 2 \\ 
\hline 
Multi - 1000 & 3 & 5  & 6  & 3  & 17   & 4  \\ 
\hline 
Multi - 2000 & 1 & 3  & 1  & 6  & 11   & 1  \\ 
\hline 
Multi - 3000 & 5 & 8  & 5  & 2  & 20   & 5  \\ 
\hline 
Multi - 4000 &  6 & 7  & 7  & 1  & 21  & 7 \\ 
\hline 
Single - NP  &  8 & 1  & 4  & 7  & 20  & 6   \\ 
\hline 
Single - 1000 & 7 & 2  & 8  & 8  & 25  & 8  \\ 
\hline 
\end{tabular} 
\caption[Ranking the estimation strategies by how well they were able to estimate the average ROS in the fire and how well the direction of the gradients matched that of the ``ground truth" fire arrival time.]{Ranking the estimation approaches by how well they were able to estimate the average ROS in the fire and how well the direction of the gradients matched that of the ``ground truth" fire arrival time. The best strategy was to use the multigrid approach with extra points interpolated along the paths at a 2000 meter spacing. This table summarizes and ranks the results in Table \ref{tbl:ros_test}.}
\label{tbl:ros_test_rank}
\end{table}

Finally, the conclusions from the first and second parts of this test, referred to as ``Area Tests" and ``ROS tests," respectively, are combined in Table \ref{tbl:all_test_rank}. The right column ranks the strategies for estimating the fire arrival time. The multigrid strategy using interpolation with a 2000 m spacing was found to produced the best results for this experiment using artificial data. Interestingly, not using the multigrid approach produced the worst results, even when extra points were interpolated along the paths.

\begin{table}
\begin{tabular}{|c|c|c|c|c|}
\hline 
Scenario & Area Tests & ROS Tests & Total & Rank\\ 
\hline 
Multi - NP   & 31&17 &48 &5\\ 
\hline 
Multi - 500  & 36&13 &49 &4 \\ 
\hline 
Multi - 1000 & 21& 17&38 &2  \\ 
\hline 
Multi - 2000 & 16&11 &27 &1  \\ 
\hline 
Multi - 3000 &25 &20 &45 &3  \\ 
\hline 
Multi - 4000 &32 &21 &53 &6 \\ 
\hline 
Single - NP  &44 &20 &64 &7   \\ 
\hline 
Single - 1000 &44 &25 &69 &8  \\ 
\hline 
\end{tabular} 
\caption{Table for all ranks for the experiment using artificial fire arrival times and fire detections. This table summarizes all results from Table \ref{tbl:multi_test} and Table \ref{tbl:ros_test}. Using a multigrid approach with interpolation of extra points with a spacing of 2000 meters provided the best estimates of the fire arrival time. }
\label{tbl:all_test_rank}
\end{table}

\subsubsection{WRF-SFIRE Output Files Used as Ground Truth}
\label{wrf_out_testing}
In this test, 115 fire arrival times output from WRF-SFIRE  were used to make ``ground truth" fire arrival times and artificial detections were scattered on the surface in the same way they were in the previous test, around 5\% of the possible locations had an artificial fire detection. Figure \ref{fig:gallery_wrf_test} shows the shape of 6 of these fire arrival times and the locations of the artificial detections. In comparison with the artificial fire arrival times used in the previous experiment, the fire arrival times from WRF-SFIRE output have rougher outer perimeters, but none indicate that more than a single fire was burning in the domain.

These simulations play the role of cross validation in the experiment, with the goal to determine that the methods developed do not result in an overfitting of data when applied to a different set of inputs. The testing method used was exactly the same as that producing the best results in the experiment using 280 artificial fire arrival times.  Only the multigrid method with interpolation of extra points along the paths with spacing of 2000 meters was used in this test. The results of the tests, in Table \ref{tbl:wrfout_test}, relating to estimated fire arrival time in terms of its growth, final size and shape, as well as the relative error the were similar to the tests using artificial fire arrival times, with the exception that the average growth error was worse. Table \ref{tbl:wrfout_ros_test} gives results for the tests relating to the ROS and the direction of the fire spread. In this case, we find that the difference in the average ROS is higher than it was for the tests using 280 artificial fire arrival times. 

The results in these tables represent averages of the scores for all 115 testing scenarios. To get a better sense of the associated variances of these statistics, Figure \ref{fig:moe_wrf_test} shows a scatter plot of the MOE, with MOE X on the horizontal axis and MOE Y on the vertical. Figure \ref{fig:re_wrf_test} and Figure \ref{fig:ros_wrf_test} show histograms of the relative errors and the differences in average ROS for all 115 tests.

\begin{table}
\begin{tabular}{|c|c|c|c|c|c|c|c|c|c|}
\hline 
Scenrario & AGE & MRE & SRE & MOE X  & MOE Y  & ||MOE|| & Sorenson & Best & Worst \\ 
\hline 
Multi - 2000 & 0.2690 & 0.0056 & 0.0015 & 0.9475& 
0.6453 & 1.15 & 0.7621 & 111 & 99 \\ 
\hline 
\end{tabular}
\caption[Results from estimating the fire arrival time using the multigrid approach with additional points interpolated along the paths at a distance of 2000 meters. The scores relate to the  growth, final size and shape, as well as relative error of the estimated fire arrival time. For this test, the ``ground truth" fire arrival time for each of the 115 estimates made was taken to be the output fire arrival time from a fire simulation made with the WRF-SFIRE model.]{Results from estimating the fire arrival time using the multigrid approach with additional points interpolated along the paths at a distance of 2000 meters. The scores relate to the  growth, final size and shape, as well as relative error of the estimated fire arrival time. For this test, the ``ground truth" fire arrival time for each of the 115 estimates made was taken to be the output fire arrival time from a fire simulation made with the WRF-SFIRE model. Compare with the results in Table \ref{tbl:multi_test} where comparison was made with artificial fire arrival times. The results are similar, with the exception of the average growth error score. The results for this score are worse. }
\label{tbl:wrfout_test}
\end{table}

\begin{table}
\begin{tabular}{|c|c|c|c|c|c|c|c|c|}
\hline 
Scenario & MRD & SRD & MDD & SDD & Best ROS & Worst ROS & Best $\theta$ & Worst $\theta$ \\ 
\hline 
Multi - 2000 & 0.1546 & 0.1027 & -0.0451 & 0.1463 & 111 & 6 & 61 & 90 \\ 
\hline 
\end{tabular} 
\caption[Results from estimating the fire arrival time using the multigrid approach with additional points interpolated along the paths at a distance of 2000 meters. The scores relate to the  ROS and direction of fire spread in estimated fire arrival time. For this test, the ``ground truth" fire arrival time for each of the 115 estimates made was taken to be the output fire arrival time from a fire simulation made with the WRF-SFIRE model.]{Results from estimating the fire arrival time using the multigrid approach with additional points interpolated along the paths at a distance of 2000 meters. The scores relate to the  ROS and direction of fire spread in estimated fire arrival time. For this test, the ``ground truth" fire arrival time for each of the 115 estimates made was taken to be the output fire arrival time from a fire simulation made with the WRF-SFIRE model. Compare the results to Table \ref{tbl:ros_test}. In this test, the difference in the estimated average ROS was much worse that when using artificial fire arrival times as the ``ground truth."}
\label{tbl:wrfout_ros_test}
\end{table}

%
%

\subsection{Examples and Testing with Real Fires}

\subsubsection{Finding the Parameter $p$ }
\label{sec:find_p}

\begin{figure}[!ht]
\centering
\includegraphics[width=0.45\textwidth]{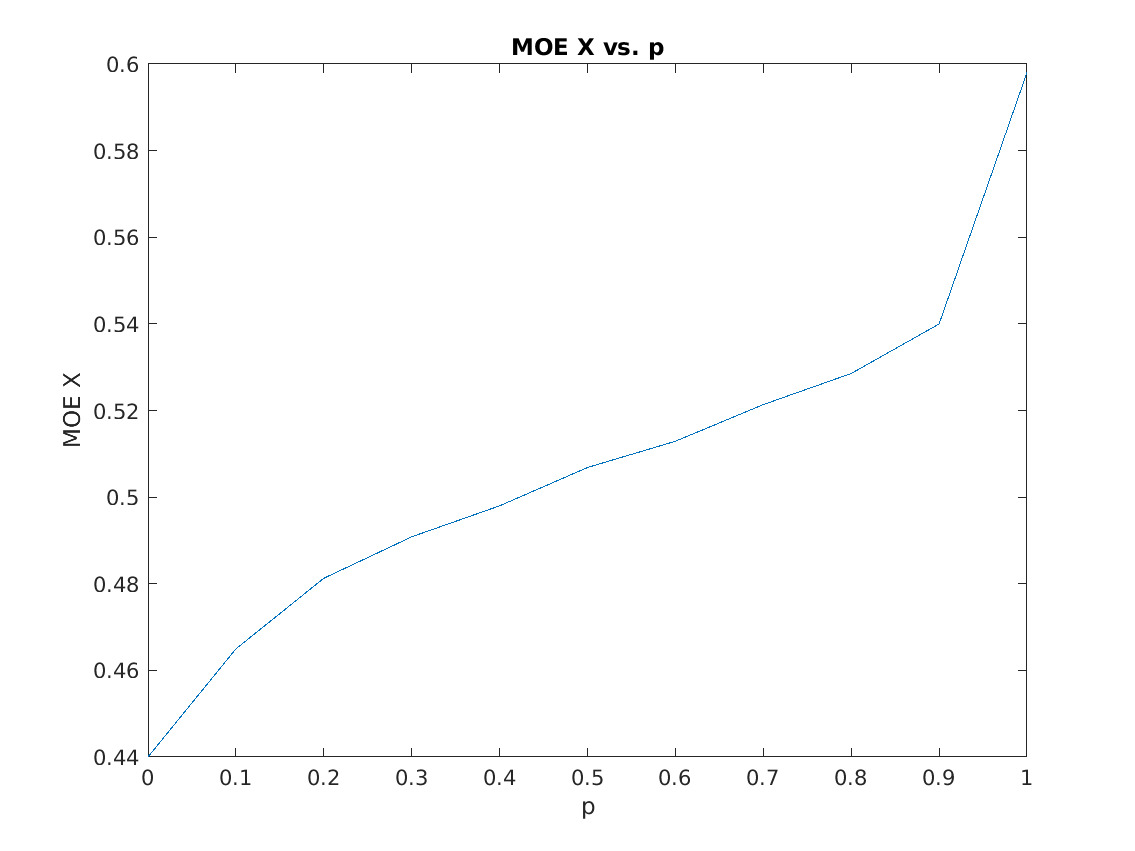}
\includegraphics[width=0.45\textwidth]{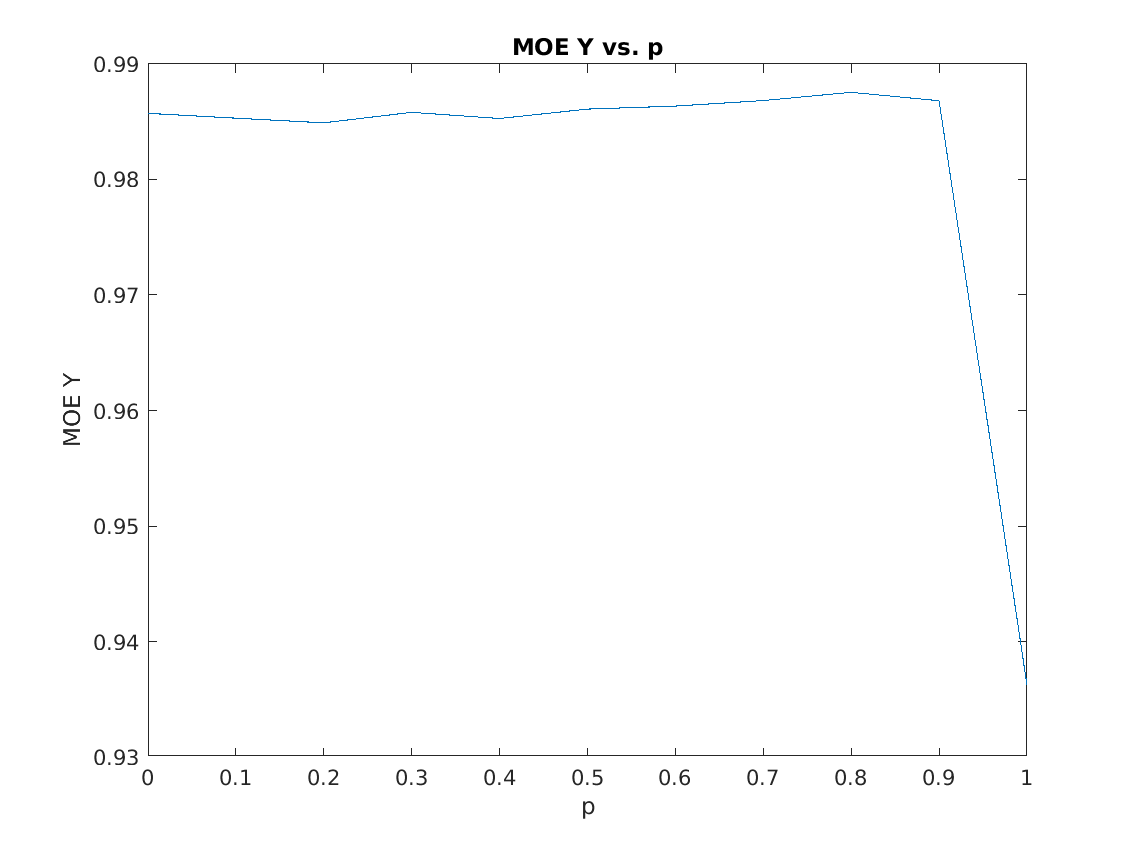}

\includegraphics[width=0.45\textwidth]{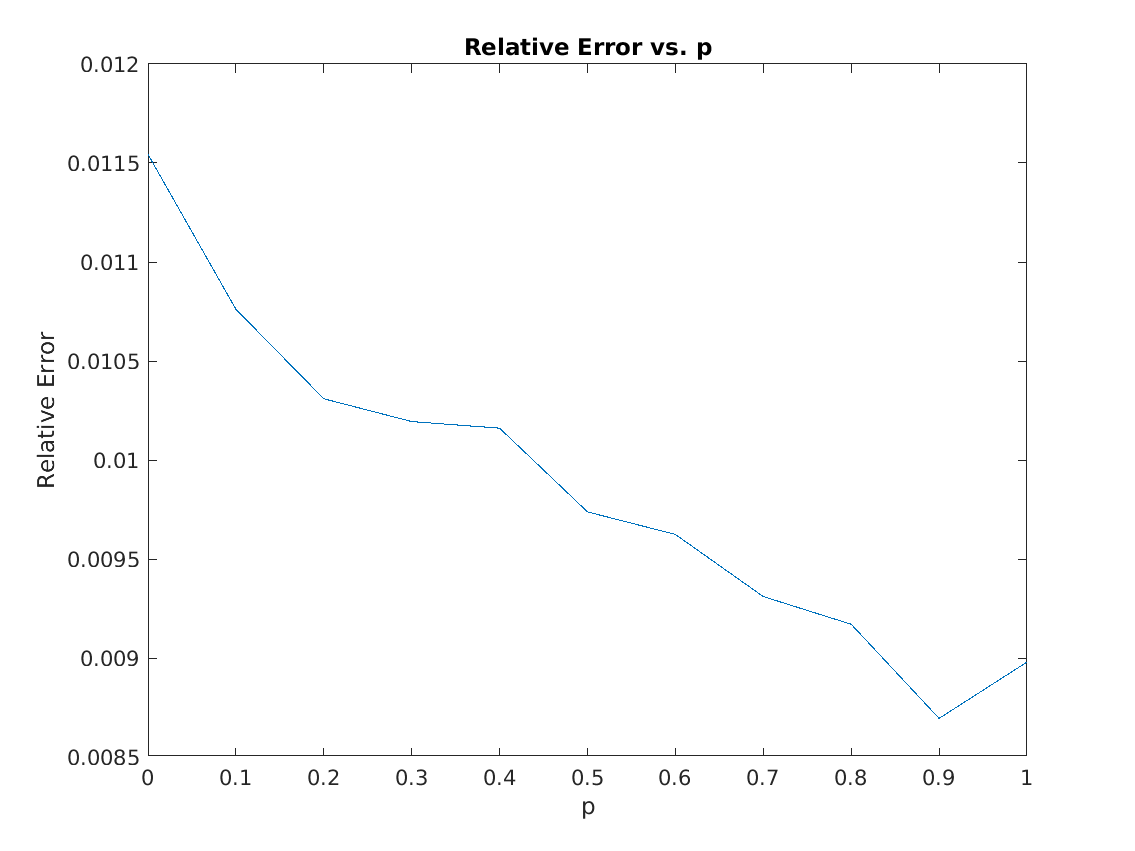}
\includegraphics[width=0.45\textwidth]{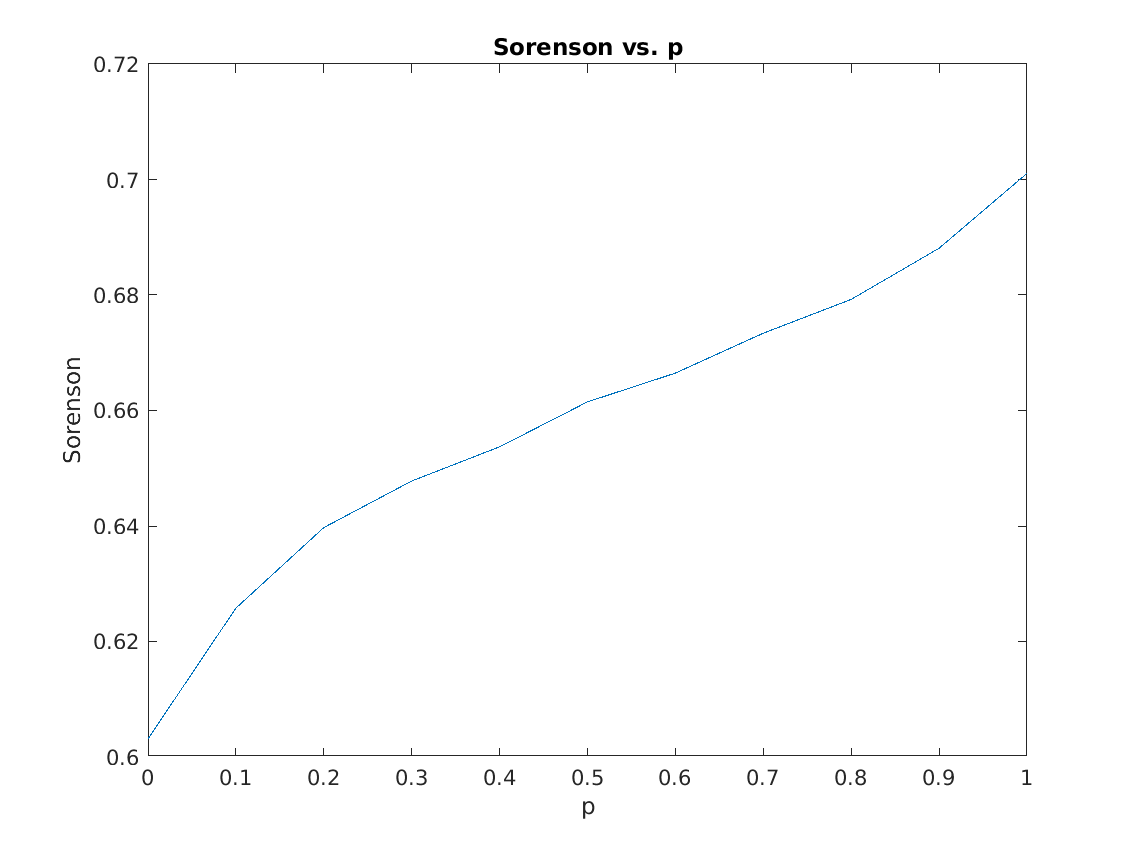}
     \caption[Assessment of estimated fire arrival times of 280 artificial fire arrival times using varied values of the interpolation parameter $p$.]{Assessment of estimated fire arrival times of 280 artificial fire arrival times using varied values of the interpolation parameter $p$. Each panel represents the average score of all 280 estimates for varying values of $p$.  On the top row, from left to right are the plots of MOE X and MOE Y. On the bottom are the plots of  relative error, and the S{\o}renson index. These plots suggests that using $p=0.9$ is optimal.}
\label{fig:p_test_artificial_figs}
\end{figure}

To determine the optimal value $p$ in the interpolation of new data points along the paths (Equation \ref{eq:spline}), a test was devised. As a first step, estimates of the fire arrival time were made for set of the artificial fire arrival times used in the experiments of Section \ref{sec:ground_truth_1} using various values of $p$. For each set of detection data, 11 estimates of the fire arrival time were made using values of $p$ ranging from 0.0 to 1.0 in increments of 0.1. All estimates were made on a common computational grid without using a multigrid approach to save time. Each fire estimated arrival time was then compared against the artificial ``ground truth" fire arrival time and the relative error, MOE, and the 
S{\o}renson index were computed. The average of each score for all tested values of $p$ was computed and the results plotted in Figure \ref{fig:p_test_artificial_figs}. The MOE and S{\o}renson index scores generally increase with $p$, except for MOE Y which exhibited a decease for $p=1.0$. The scores for the relative error generally decreased, indicating a better performance, when $p$ was increased, but and increase occurred for $p=1.0$.  Thus, it was determined that the optimal value is $p=0.9$.

Further investigation of the optimal value $p$ to use was accomplished by making estimates of the fire arrival times of real-world fires. Estimates of the fire arrival time for the the Patch Springs Fires\ and the Cougar Creek Fire were made using varying values of $p$ between 0.0 and 1.0. For each value used, the resulting fire arrival time was compared to an infrared perimeter observation and the MOE and S{\o}renson index were computed. The results are summarized in the tables and figures that follow. Tables \ref{tbl:patch_p_test} and \ref{tbl:cougar_p_test} show the results which are summarized in figures \ref{fig:p_test_patch_figs} and \ref{fig:p_test_cougar_figs}. The effect of $p$ in these individual cases was similar to that of the average effect determined in 280 cases using artificial data. The scores for the estimates generally increased with $p$ but showed a decrease for $p=1.0$. In both specific cases, using $p=0.9$ for the parameter was found to be the best, although results were similar for $p=1.0$. These results closely match those found when using the suite of artificial fire arrival times in the testing previously detailed.

\subsubsection{Real Fire Example: Patch Springs Fire with Perimeter Data}

\begin{figure}[htbp]
\centering
\includegraphics[width=0.45\textwidth]{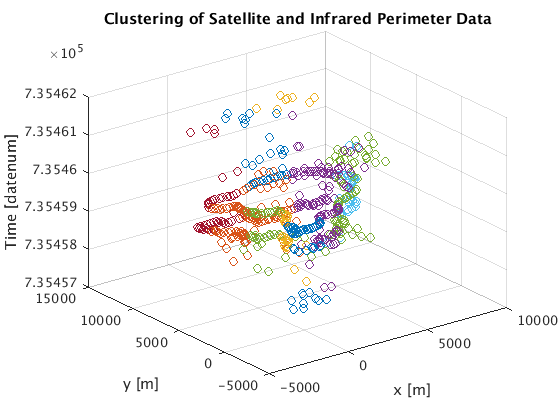}
\includegraphics[width=0.45\textwidth]{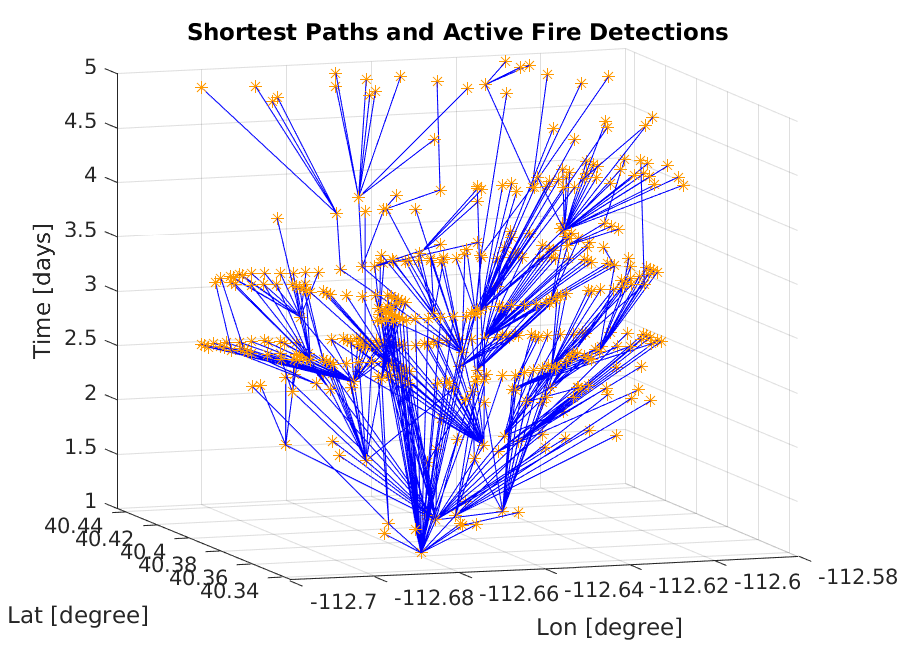}
\includegraphics[width=0.45\textwidth]{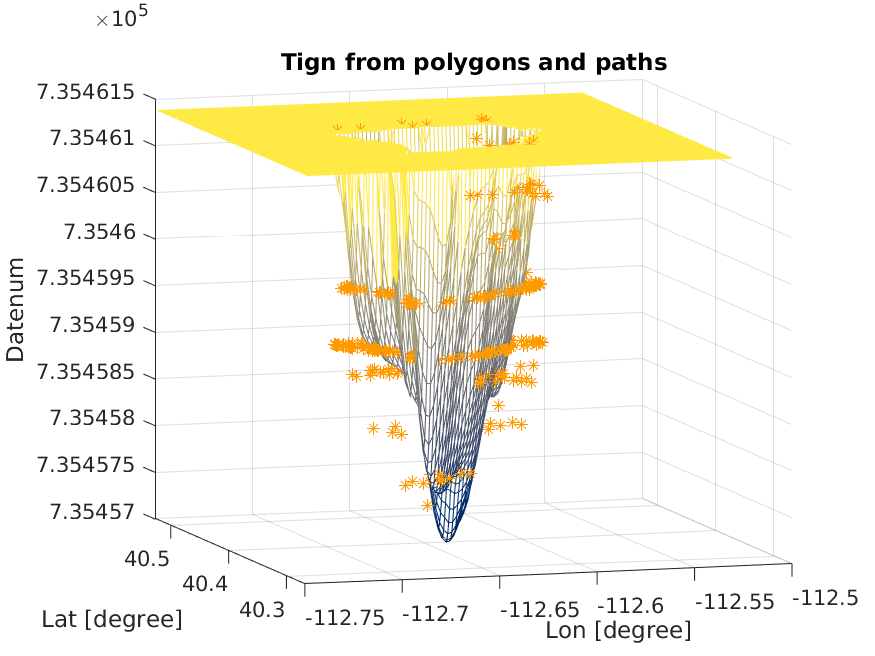}
 \caption[Estimation of the fire arrival time for the Patch Sprigs fire  using infrared perimeter observations.]{Estimation of fire arrival time for the Patch Sprigs fire  using infrared perimeter observations. In the upper left, active fire detection data and points from infrared perimeter observations used for estimation of the fire arrival time of the Patch Springs fire. In upper right, the shortest paths in the directed graph for the Patch Fire simulation, initialized from satellite data and infrared perimeter data. At the bottom, the estimated fire arrival time. 200 data points were used in each perimeter. The perimeters are visible as the rings with large amounts of data at approximately day 2.5 and day 3 of the estimate.}
 \label{fig:patch_estimate_simulation_graph}
\end{figure}

The Patch Springs Fire occurred in August 2013. In this section, we use the method of Section \ref{sec:fat_method} to create an estimate of the fire arrival time over the first six days of the event. The estimated fire arrival time used two infrared perimeters and 39 granules of  satellite data. For each perimeter observation, 200 fire data points were used. Although perimeter observations typically contain 1000 points or more, the larger grid spacing using for the interpolation technique makes use of more than a few hundred points unproductive.  Figure \ref{fig:patch_estimate_simulation_graph} shows the clustering of the data, the graph structure with shortest paths, and the resulting estimate of the fire arrival time. Comparison of the estimated fire arrival time was made with a perimeter observation from August 16 at 09:47 UTC. When compared with the estimated fire arrival times made without perimeter observations, in Table \ref{tbl:patch_p_test}, the simulation  here is marginally better. The results are summarized in Table \ref{tbl:patch_perim_test}. Without a known fire arrival time, only the MOE and S{\o}renson index have been used to evaluate the reults. For this test, the results are consistent with the experiments in Section \ref{sec:fake_perim_test}, indicating that the method of using infrared perimeters in addition to satellite fire detections is a useful strategy for estimating the fire arrival time of a real-world fire.

\begin{table}
\centering
\begin{tabular}{|c|c|c|c|} 
\hline
 MOE X  & MOE Y  & \textbar{}\textbar{}MOE\textbar{}\textbar{} & Sorenson  \\ 
\hline
 0.9232 & 0.8550 & 1.2858                                      & 0.9089    \\ 
\hline
\end{tabular}
\caption{Assessment of the estimated fire arrival time of the Patch Springs fire made using infrared perimeter data and satellite data. The addition of the extra data source has made the estimate better than any of the estimates in Table \ref{tbl:patch_p_test}. }
\label{tbl:patch_perim_test}
\end{table}


\begin{figure}[htbp]
\centering
\includegraphics[width=0.50\textwidth]{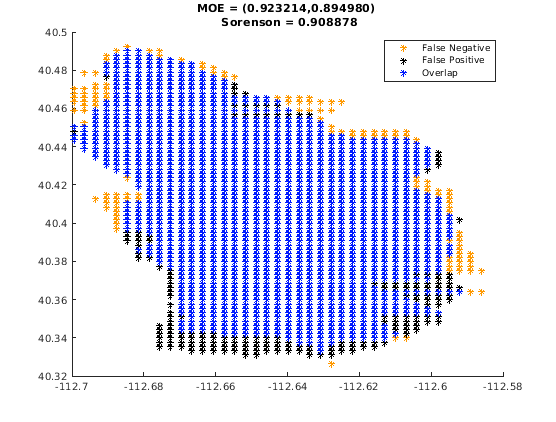}
 \caption[Assessment of the estimated fire arrival time for the Patch Springs fire using infrared perimeter observations as a data source.  ]{Assessment of the estimated fire arrival time for the Patch Springs fire using infrared perimeter observations as a data source. Two perimeter observations were used as a data source and the the estimate was compared to an infrared perimeter from August 16 at 09:47 UTC. The addition of the perimeter data made for a small improvement of the estimate in comparison with the estimates from Section \ref{sec:find_p} that only used satellite data.}
 \label{fig:patch_estimate_perim_moe}
\end{figure}

\subsubsection{Real Fire Example: Cougar Creek Fire}
In some cases, real fires are intensely observed and many infrared perimeter observations are made available. In this section, many estimates of the fire arrival time for the Cougar Creek Fire were made and evaluated by comparison with an infrared perimeter observation. Both a single-pass and the multigrid method with interpolation of points with 2000 meter spacing were used.  In each estimate, all satellite fire detection data up to the time of a perimeter observation was used to estimate the fire arrival time. In total, 27 perimeters were used to form comparisons. For each estimate, the MOE and the S{\o}renson index were computed. Additionally, the area within the perimeters of the observation and estimate were made at each of the 27 times. All result computations were made on grids of 250 meter resolution. This grid spacing is well below the resolution of the typical perimeter observation but above the resolution of the fire detection data.

\begin{figure}[htbp]
\centering
\includegraphics[width=0.45\textwidth]{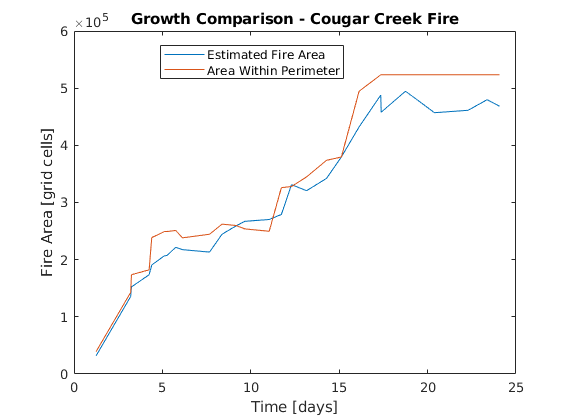}
\includegraphics[width=0.45\textwidth]{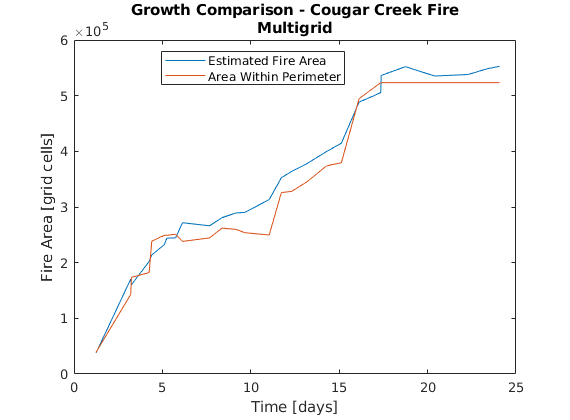}
 \caption{Comparison of fire sizes from fire arrival time estimates of the Cougar Creek Fire using a one-pass method (left) and multigrid method (right). The one-pass method  underestimated the size of the fire while the multigrid method overestimated it.}
 \label{fig:cougar_test_growth}
\end{figure}

\begin{figure}[htbp]
\centering
\includegraphics[width=0.45\textwidth]{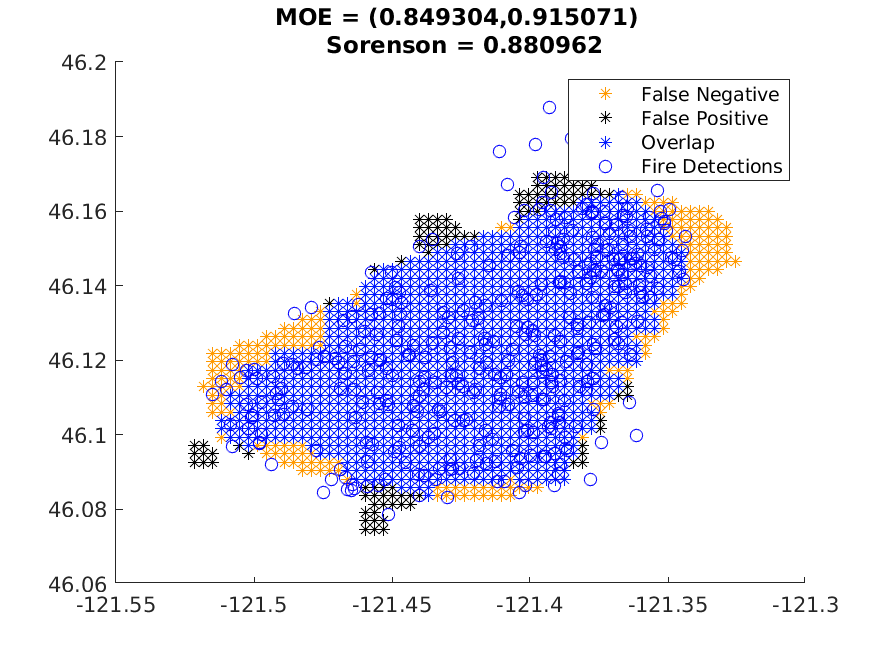}
\includegraphics[width=0.45\textwidth]{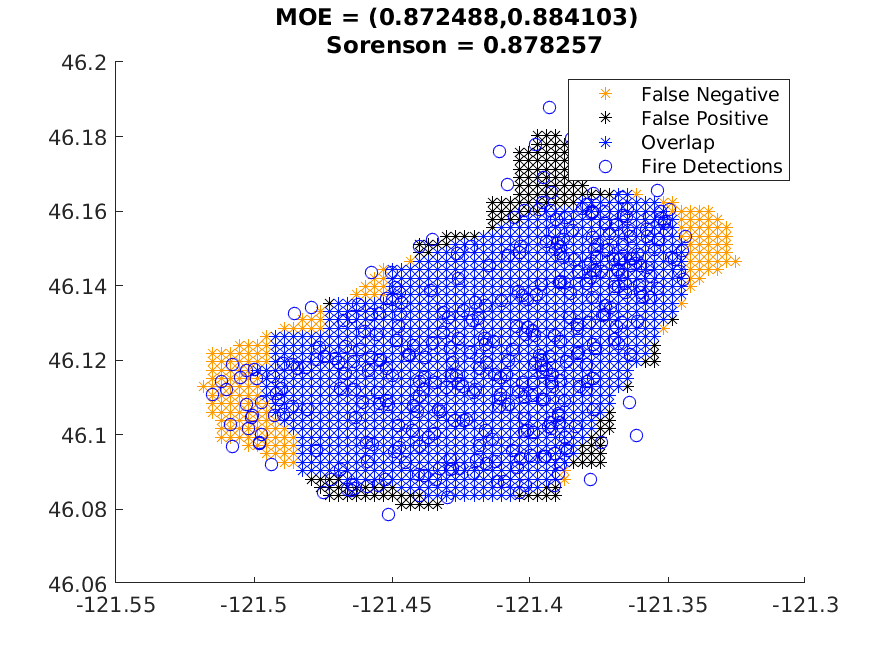}
 \caption[Assessing the estimated fire arrival time of the Cougar Creek Fire using a one-pass method (left) and multigrid method (right) by comparison with an infrared perimeter observation.]{Assessing the estimated fire arrival time of the Cougar Creek Fire using a one-pass method (left) and multigrid method (right) by comparison with an infrared perimeter observation. The estimates are derived for the available satellite data up to the perimeter time of August 14, 07:00:00 UTC. Both methods ted to underestimate the size of the fire. In this case, some satellite fire detections lie outside the infrared perimeter represented by the union of the orange and blue areas.}
 \label{fig:cougar_moe_perim_3}
\end{figure}

Given a sequence of infrared observations, its is possible to get an estimate of the growth rate of the real fire by computing the area within each perimeter observation. The area within these perimeters is plotted along with the area found by estimation using satellite data for both the single-pas and multigrid approaches Figure \ref{fig:cougar_test_growth}. The blue lines indicate area of the fire estimated from data and the red lines indicate the area within the infrared perimeters. Both estimation strategies show similar growth to that indicated by the infrared data, but the single-pass method tends underestimate the area and the multigrid approach tends to overestimate it. Figure \ref{fig:cougar_moe_perim_3} shows a comparison of the assessment of the tow methods for an estimate of the fire arrival time at August 14, 07:00:00 UTC. At this time step, both strategies both strategies had similar results, with the MOE and S{\o}renson index scores differing less than 0.05. In both cases, the strategies slightly underestimated the size of the fire. Figure \ref{fig:cougar_moe_perim_27} shows the final estimates of the sequence. Again, both strategies delivered similar results, but the multigrid method gives less false negatives and more false positives.

Table \ref{tbl:cougar_1_pass_results} and Table \ref{tbl:cougar_multigrid_results} show the results of assessment of the estimated fire arrival times using the single-pass and multigrid strategies, respectively. In general, both strategies had similar results with the typical score for both components of the MOE and the S{\o}renson index remaining above 0.80. Some lesser scores were recorded in the early stages of the fire and can most likely be attributed to the a relatively small amount of satellite data data being available. For example, the first estimate  had the score $S=0.6082$ using the single-pass strategy. The corresponding score for the multigrid strategy was also low, $S=0.6704.$ Scores for later estimates remained largely constant, implying that once sufficient satellite data were available, both strategies produced acceptable results. Table \ref{tbl:cougar_moe_summary} compares the average scores made by both strategies across all 27 estimated fire arrival times. The two strategies produced similar results. The largest differences exit in the scores for MOE X and MOE Y. The largest score for MOE X was achieved by the multigrid strategy, indicating it produced less false negatives. The single-pass strategy produced less false positives.

\begin{table}
\centering
\begin{tabular}{|c|c|c|c|c|c|} 
\hline
Method    & MOE\_X & MOE\_Y & \textbar{}MOE\textbar{} & S    \\ 
\hline
One Pass  & 0.8267 & 0.9051 & 1.2283                  & 0.8607           \\ 
\hline
Multigrid & 0.8898 & 0.8457 & 1.2299                  & 0.8640        \\
\hline
\end{tabular}
\caption{Comparison of the average assessment scores for the one pass method and multigrid method used to estimate the fire arrival time of the Cougar Creek Fire. The scores are very similar, with the multigrid showing a slight advantage in having a final burn area closer to that indicated by the infrared perimeter observations.}
\label{tbl:cougar_moe_summary}
\end{table}

\subsubsection{Real Fire Example: Camp Fire}
\begin{figure}[htbp]
\centering
\includegraphics[width=0.45\textwidth]{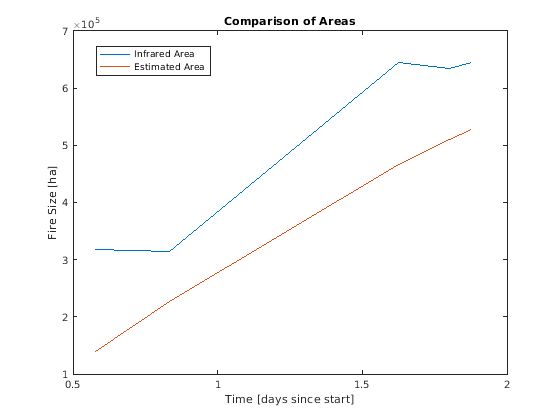}
 \caption{Growth of the Camp fire during the first several days. The growth rate of the area was 28,000 ha/day and the estimated growth rate was 30,000 ha/day. The estimated are of the fire was smaller than that observed, but the rate of growth was very similar. Interestingly, the computed area within the infrared perimeters showed a small decrease after 1.5 days.}
 \label{fig:camp_growth}
\end{figure}

\begin{figure}[htbp]
\centering
\includegraphics[width=0.45\textwidth]{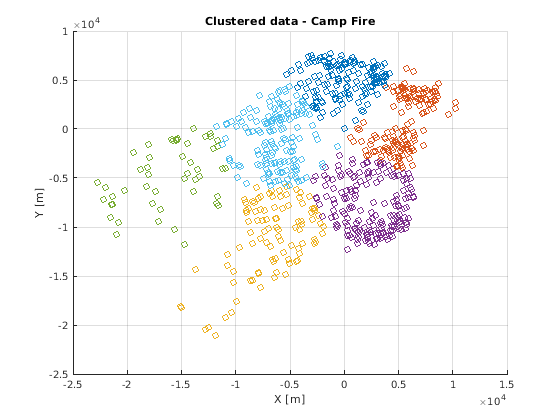}
\includegraphics[width=0.45\textwidth]{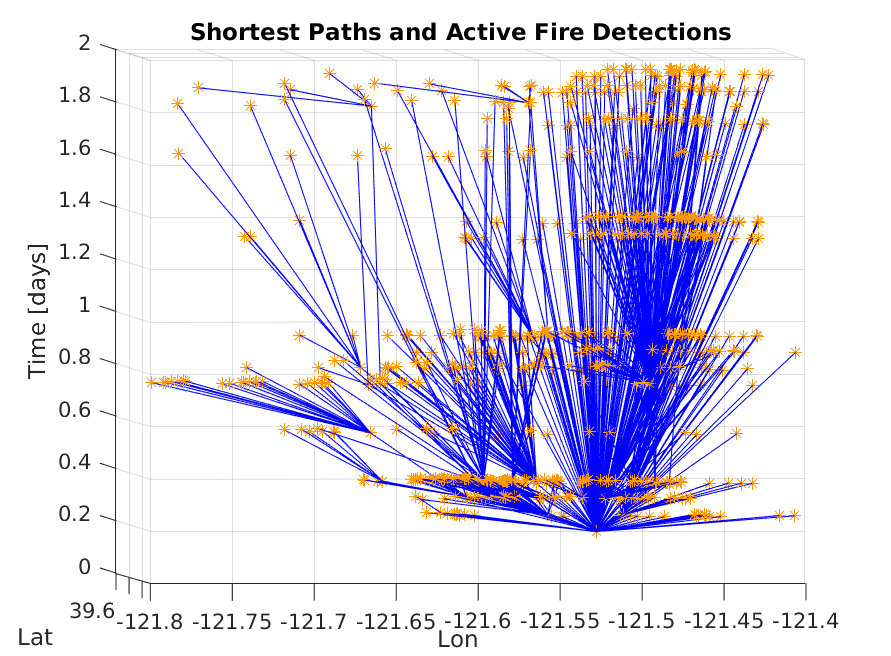}
 \caption{Clustered data for the Camp Fire (left) and the shortest paths in the directed graph (right). Note how the low angles of the paths near the bottom of the graph indicate the high ROS observed in the explosive early hours of this fire.}
 \label{fig:camp_data}
\end{figure}

\begin{figure}[htbp]
\centering
\includegraphics[width=0.45\textwidth]{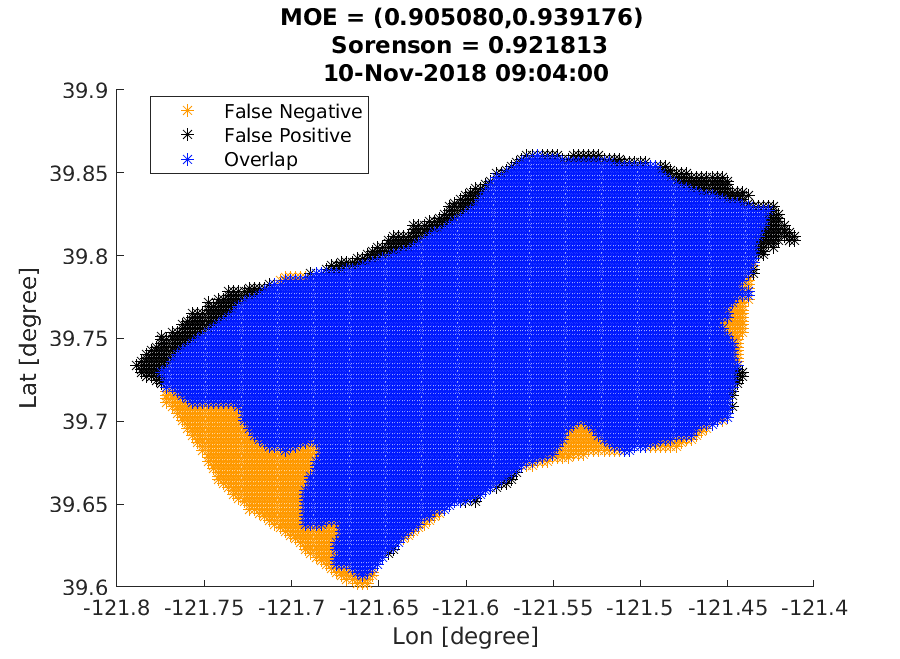}
 \caption{Assessment of the estimated fire arrival time of the Camp Fire using the MOE and the S{\o}renson index. The fire started in the northeast region of the domain and progressed rapidly towards the southwest.  The union of the blue and orange area indicate the region within the infrared perimeter.}
 \label{fig:camp_moe}
\end{figure}

The Camp Fire was a large and destructive fire that took place in Califormia during November, 2018. Extreme weather cause the fire to expand rapidly \citep{Brewer-2020-CFM} and it is thought that the spread mechanism was largely due to burning embers being carried by winds in advance of the main fire front \citep{Cal_fire-2018-GSB}). It is estimated that the fire grew to 28,000 ha in the first 24 hours. Figure \ref{fig:camp_data} shows a scatter plot of the clustering of the satellite data and the structure of shortest paths from an assumed ignition point. The ``flat bottom" of this graph is further evidence of the explosive growth of this fire in its early stages. Five infrared observations were made during the first two days of this fire and estimates of the fire arrival time were made to compare with each of them. Figure \ref{fig:camp_growth} shows how the area of the estimates compared with the are with in the infrared perimeters. In general, the estimated area was smaller, but the slope of the two curves are similar, indicating rates of growth that are largely the same. Figure \ref{fig:camp_perim_sequence} shows comparison of the estimated fire perimeters with the infrared perimeters at four times. The estimated perimeters differ from the infrared observations mostly in the southwest region of the fire that active fire front during the first two days. Figure \ref{fig:camp_moe} shows the assessment of the estimate made by comparison to the final infrared perimeter observation made on November 10, at 09:04 UTC. In the figure, the area of in the southwest showing false negatives may be due to the spread mechanism of the fire. Winds from the northeast carried burning embers ahead of the main fire front, causing many small spotting fires to occur. These spotting fires may have been observed by infrared camera onboard aircraft but not by satellites, given the region of false negatives at the edge of the fire front. The scores for this estimate were high, with both components of the MOE and the S{\o}renson index all exceeding 0.90. These scores were as good as those for estimates of both the Patch Springs and Cougar Creek fire previously discussed.

\subsection{Limitations of the Method}
The method currently cannot give good estimates of the fire arrival time under all circumstances. Some considerations are outlined in this section.

\subsubsection{More than one fire in the domain}\label{sec:detections_filters}
\begin{figure}[htbp]
\centering
\includegraphics[width=0.65\textwidth]{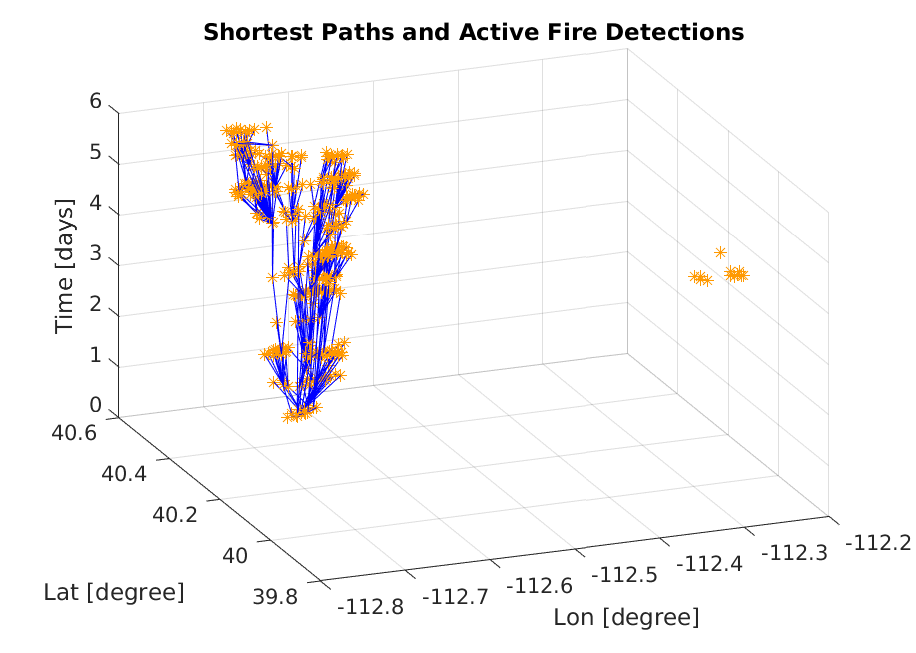}
 \caption[Two distinct fires in a single simulation domain.]{Two distinct fires in a single simulation domain. In total, there are 361 active fire detections in the fire domain. The active fire detections were first clustered into two sets using k-means clustering and then the members of the smaller cluster were automatically removed from the directed graph used to make the paths seen in the larger cluster.}
 \label{fig:patch_two_fires}
\end{figure}

On some occasions it can happen that more than one fire is present in the simulation domain but only one fire is being modeled. With very large simulation domains during fire season this is more likely. This presents problems to the method since the fire detections of both fires will be treated as belonging to a single fire with a unique ignition point. During the process of constructing the set of shortest paths from the ignition point to all the other fire detections in the domain, paths will be constructed to connect the two distinct fires. When more points are interpolated along these paths, the method will then be assigning fire arrival times to areas that have not burned. There are several options to avoid this kind of problems.
\begin{enumerate}
\item Attempt to detect distinct fires and ignore detections not belonging to the fire not being simulated. This has been explored in the process of the current research. Presently, after construction of the distance matrix, the active fire detections are clustered into two groups with k-means clustering. If the number of detections in one group is significantly smaller than the other, the distance between each detection in the smaller cluster and all other detections in the entire fire domain is set to zero. This has the effect of ``disconnecting" these detections from all others when the directed graph is constructed. Figure \ref{fig:patch_two_fires} shows a fire domain with two distinct fires inside it but the detections in the smaller cluster will not be considered as part of the fire being studied.
\item Eliminate connections between fire detections too far apart. This method attempts to exclude detections outside of the fire being simulated by imposing a ``speed limit" related to the path segements that could connect to fire detections in a directed graph. After the distance matrix is constructed, the ROS between any two possible connected detections can be computed by dividing the distance between them by the time the observations were made. Some threshold value can be chosen so that if the ROS between two detections is above the threshold value, the distance between them will then be set to zero, eliminating the possibility of a connection between them in the directed graph.
\end{enumerate}

These methods can work with a relatively small fire domain with just one or two fires inside it, but will not work on larger domains with more distinct fires inside. A first step to handling the problem would be devise a technique to identify individual fire complexes using the satellite active fire detections. If this could be done, then the methods developed in this research could be applied to the identified individual fires all at the same time. With parallel computing resources being used, it would be possible to work with all of the fires in a large region simultaneously.

\subsubsection{Size of the Fire}
The size of the fire is an important consideration. Large fires that have burned for a long time present a challenge because the amount of data can be quite large. For example, construction of the distance matrices used for making the path structure is expensive. This limitation could be overcome by only estimating the last several days of the fire arrival time for large, lengthy fires. Conversely, small fires may be easier to work with computationally but the sparsity of data is problematic. If only a small number of active fire detections are present, the method of spatially clustering the detection locations may fail.

\subsubsection{Ignition Point Estimation}
In some cases the estimation of an ignition point is difficult. This is likely to occur for fires that have an ignition followed by explosive growth, as was the case with the Camp Fire. In such cases, averaging the locations of the fire set of detections is likely to lead to an inaccurate ignition location estimate since the winds responsible for the explosive growth would indicate the location was near the edge of the region containing the detections used in the estimate. The method for estimating the ignition point by grid search to find the point giving the maximum data likelihood may not work with such cases because the WRF-SFIRE model is used to run an ensemble of short-term forecasts, but the spread mechanism of such a fire might be poorly modeled by WRF-SFIRE.


\section{Uncertianty in the Estimation of The Rate of Spread}\label{sec:ros_uq}

The path structure constructed as part of the method for estimating the fire arrival time serves as a convenient tool  for  estimating ROS at many locations within the fire domain. Under the assumption that fire has spread in a straight line between detection locations along a path in the structure, computation of the ROS is a simple matter of dividing the distance between the points by the difference of the times at which the locations were observed to be burning. However, as previous discussed in Section \ref{sec:data}, the actual location of fire on the ground is uncertain. Likewise, the time that fire arrived at the locations cannot be known precisely. Thus, the estimated ROS derived from satellite observations should be treated as a random variable. Some of the assumptions and uncertainties when working the estimated ROS are listed below.

\begin{itemize}
\item Geolocation errors. The geolocation error depends on the scan angle by which the satellite observed the pixel. Higher angles correspond to larger possible geolocation errors \citep{Nishihama-1997-ML1}. It will be assumed that the geolocation error for any observation follows a normal distribution with a standard deviation of 125 meters. This corresponds to an observation made with a scan angle of approximately 35 degrees. 
\item The geolocation error is more complicated than being assumed. The variance depends on the scan angle, but the components of this error are different in the track direction (parallel to the flight path) and the across track direction (roughly perpendicular to the flight path.) At nadir view (zero scan angle), the region containing the 3$\sigma$ error is roughly circular. As scan angle increases towards a maximum of $55^\circ$ the region containing the 3$\sigma$ becomes more elongated in the across track direction. 
\item It will be assumed that the uncertainty in the fire arrival time at any active fire detection will be uniformly distributed over the previous 12 hours.  This assumption follows from the probability of satellite detection seen in Figure \ref{fig:new_like}. The probability of detection is almost 100\% for many hours after the fire has arrived at a location but there is no reason to assume any time during this period is more likely than another.
\item In principle, the length of period for which the detection probability is high depends on factors not being accounted for. For example, fire in a dry, grassy region would be detectable for a relatively short time if windy conditions were present because the fuels would be consumed quickly. Fires moving slowly in regions with wetter fuels my be detectable for longer periods.  
\end{itemize}


\subsection{Estimation of Uncertainty for Points Along the Paths in the Graph}

\begin{figure}[!h]
\begin{center}
  \includegraphics[width = 0.79\textwidth]{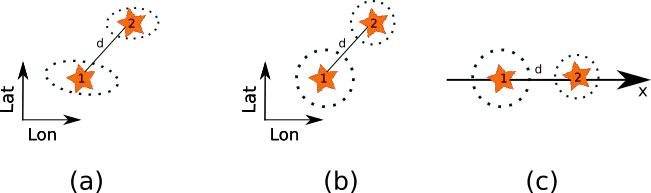}
\caption[Visualizing and simplifying the the uncertainty in the distance between two active fire detections.]{ Visualizing and simplifying the the uncertainty in the distance between two active fire detections.}
\label{fig:simplify_distances}
\end{center}
\end{figure}

With shortest paths connecting an assumed ignition point to active fire detections in the fire domain, an estimate of the rate of spread (ROS) between two can be calculated by dividing the distance between them by the difference in the times the observations were recorded. The calculated ROS is subject to error because the location of the fire detections always have an associated geolocation error. Further, even if there was no geolocation error involved, the time the fire arrived at any particular location cannot be known precisely. If a fire is reported at a position during the satellite overpass, there is no way to determine if the fire has just arrived there or if it has been there for many hours. Figure \ref{fig:simplify_distances} shows how the distance between two satellite observations will be handled. In the figure, the orange stars represent locations of active fire detections. The dashed lines around the detections represent the boundaries of 3$\sigma$ geolocation. Panel (a) shows two bivariate normal distributions with axes aligned according to the flight path of the observation platform. Panel (b) shows an approximation obtained by assuming equal variance in the across-track and along-track directions of each detection. Panel (c) shows the use of symmetry to reduce the problem to one dimension, resulting in reduced complexity by working with two normally distributed random variables.

Since the ROS between two active fire detections is always computed from observations made at different times (and possibly by different platforms) it will be assumed that the two observations are independent. The distance between the points on the ground can then be considered a random variable. Similarly, the difference in the time between observations will be considered a random variable derived from the difference of two iid uniform random variables. The ROS we will denote $R$ and the distance and time difference between observations will be denoted $D$ and $T$, respectively.
\begin{equation}
R = D/T
\end{equation}

To simplify the mathematics and make an analytic variance possible to compute, the assumptions shown in Figure \ref{fig:simplify_distances} are used to orient two fire detections in a way that the distance between them can be expressed as the difference of Gaussian random variables. The first detection is placed at the origin and has distribution $N(0,\sigma_1^2)$. The second is along the $x$-axis at location $(d,0)$ with distribution $N(d,\sigma_2^2)$. Thus the pdfs are 
\begin{equation}
f_1(x) = \frac{1}{\sqrt{2\pi}\sigma_1}\exp\left(\frac{-x^2}{2\sigma_1^2} \right) \hspace{5mm}
f_2(x) = \frac{1}{\sqrt{2\pi}\sigma_2}\exp\left(\frac{-(x-d)^2}{2\sigma_2^2} \right).
\end{equation}
The pdf of the distance between these points is computed by 
\begin{equation}
f(z) = \int_{-\infty}^{\infty}f_1(x-z)f_2(x)dx =
\frac{1}{\sqrt{2\pi(\sigma_1^2+\sigma_2^2)}}\exp\left(\frac{-(d-z)^2}{2(\sigma_1^2+\sigma_2^2)} \right)
\end{equation}

The random variable describing the time between fire arrival at the detection locations is the sum of two uniform random variables $T=Y+(-X)$ with $X\sim U(t_1-l,t_1)$ and $Y \sim U(t_2-l,t_2)$, where $t_1$ and $t_2$ are the times of the first and second detections and $l$ is the length of time before the reported time when the fire could have arrived at the location. The density is found using the convolution method for obtaining the sum of random variables.
\begin{equation}
f(t) = 
\left\{
  \begin{array}{ll}
 f_l(t) = \frac{1}{l^2}\left(t-[(t_2-t_1)-l] \right) &     \quad (t_2-t_1)-l \leq t \leq (t_2-t_1) \\
  f_r(t) = -\frac{1}{l^2}\left(t-[(t_2-t_1)+l] \right) &  \quad (t_2-t_1)\leq t \leq (t_2-t_1) +l \\
  0 & \quad \text{otherwise}
  \end{array}
\right.
\label{eq:time_pdf}
\end{equation}

To get the joint distribution describing the random variable that is the ROS between two active fire detections along a path, we need to divide a distance by the time. We now compute the reciprocal of the density in Equation \ref{eq:time_pdf}.
Letting $T$ be the random variable describing the time between two detections, define $S=\frac{1}{T}$. The PDF for $S$ is computed by first finding its CDF and then differentiating.

\begin{equation}
F_S(s) = P(S\leq s) = P\left(\frac{1}{T}\leq s\right) = P\left(\frac{1}{s}\leq T\right)
\end{equation}
The resulting CDF is found by integration of the  piecewise linear function in Equation \ref{eq:time_pdf}.

\begin{equation}
F_S(s) = 
\left\{
  \begin{array}{ll}
   0       &   s<\frac{1}{t_2-t_1+l}  \\
 \frac{\left(1/s-(t_2-t_1+l) \right)^2}{2l^2}   &  \frac{1}{t_2-t_1+l} \leq s \leq \frac{1}{t_2-t_1}   \\
 1 - \frac{\left(1/s-(t_2-t_1-l) \right)^2}{2l^2}       & \frac{1}{t_2-t_1} \leq s \leq \frac{1}{t_2-t_1-l}      \\
    1      &   \frac{1}{t_2-t_1-l} < s  \\
  \end{array}
\right.
\label{eq:recip_cdf}
\end{equation}
Differenting the CDF gives the PDF 
\begin{equation}
f_S(s) = 
\left\{
  \begin{array}{ll}
\frac{-1/s+(t_2-t_1+l)}{(ls)^2}  &   \frac{1}{t_2-t_1+l} \leq s \leq \frac{1}{t_2-t_1}  \\
\frac{1/s-(t_2-t_1-l)}{(ls)^2} &   \frac{1}{t_2-t_1} \leq s \leq \frac{1}{t_2-t_1-l}  \\
    0      &    \text{otherwise} \\
  \end{array}
\right.
\label{eq:recip_pdf}
\end{equation}

Letting $c=t_2-t_1$, the time between detections, the expected value of $S$ is found with
\begin{equation}
\begin{split}
E[s] =\int_{\frac{1}{c+l}}^{\frac{1}{c-l}}f_S(s)s ds & = \frac{c+l}{l^2}\ln\left(\frac{c+l}{c}\right) - \frac{c-l}{l^2}\ln\left(\frac{c}{c-l}\right) \\
   & =\frac{1}{l^2}\left[ (c+l)\ln\left(\frac{c+l}{c}\right) +(l-c) \ln \left(\frac{c}{c-l}\right)
   \right] \\
   & = \frac{1}{l^2}\left[ l\ln\left( \frac{c+l}{c-l}\right) +c\ln\left(\frac{c^2-l^2}{c^2} \right)  \right] \\
  & = \frac{1}{l^2} \ln\left[ 
\left(\frac{c+l}{c-l}     \right)^l 
\left(\frac{c^2-l^2}{c^2} \right)^c  
    \right]
\end{split}
\label{eq:expected_s}
\end{equation}

\begin{equation}
E[S^2] = \int_{\frac{1}{c+l}}^{\frac{1}{c-l}}f_S(s)s^2 ds = \frac{1}{l^2}\ln\left(\frac{c^2}{c^2-l^2}\right)
\label{eq:expected_s2}
\end{equation}

The variance is then 
\begin{equation}
\mathrm{Var}[S] = E[S^2]- E[S]^2 = \frac{1}{l^2}\ln\left(\frac{c^2}{c^2-l^2}\right) - \left(\frac{1}{l^2} \ln\left[ 
\left(\frac{c+l}{c-l}     \right)^l 
\left(\frac{c^2-l^2}{c^2} \right)^c  
    \right] \right)^2.
\label{eq:var_s}
\end{equation}

To get the variance of the ROS, expressed as  $R = DS,$ we compute
\begin{equation}\label{eqn:rosvar}
\begin{split}
\mathrm{Var}[R] &= E[R^2]- E[R]^2 \\
&= E[(DS)^2]- E[DS]^2 \\
&= E[D^2S^2]- E[D]^2E[S]^2 \\
&= E[D^2]E[S^2] - E[D]^2E[S]^2 \\
&= (\mu_1^2 + 
(\sigma_1^2 + \sigma_2^2))\frac{1}{l^2}
\ln\left(\frac{c^2}{c^2-l^2}\right) -
\mu_1^2  \left(\frac{1}{l^2} \ln\left[ 
\left(\frac{c+l}{c-l}     \right)^l 
\left(\frac{c^2-l^2}{c^2} \right)^c  
    \right] \right)^2.
\end{split}
\end{equation}

\begin{figure}[!h]
\begin{center}
\includegraphics[width = 0.45\textwidth]{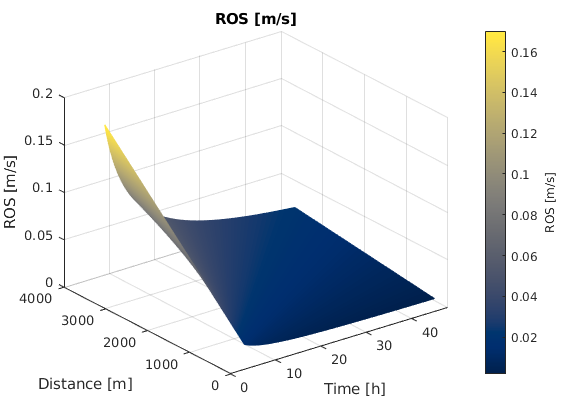}
\includegraphics[width = 0.45\textwidth]{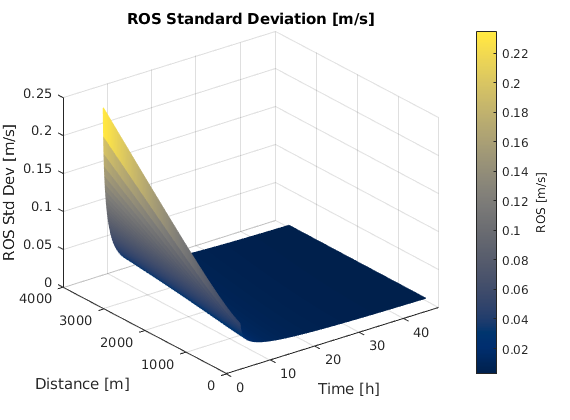}
\caption[The ROS and standard deviation of ROS, assuming the actual fire arrival time at the detection location is uniformly distributed over the  6 hours before the satellite imaging and assuming a geolocation error with standard deviation of 335 meters.]{The ROS (left) and standard deviation of ROS (right), assuming the actual fire arrival time at the detection location is uniformly distributed over the  6 hours before the satellite imaging and assuming a geolocation error with standard deviation of 335 meters. In both panels, the horizontal axes give the spatial and temporal distances between two fire detections. The standard deviation is high when the distance between detections is large but the time separating them is less than ten hours.}
\label{fig:ros_variance_3d}
\end{center}
\end{figure}

Figure \ref{fig:ros_variance_3d} explores the variance in the ROS graphically. The left panel shows a surface giving the ROS as a function of the differences in time and distance and the distance between detections. The panel on the right shows the standard deviation of the ROS, according to Equation \ref{eqn:rosvar}. It is notable that the standard deviation is large when the ROS is large. This should serve as a warning that a high estimated of the ROS should be treated with caution. High rates of spread between points on a path will be encountered when active fire detections a distant spatially but were recorded at similar times.

\begin{figure}[!h]
\begin{center}
\includegraphics[width = 0.45\textwidth]{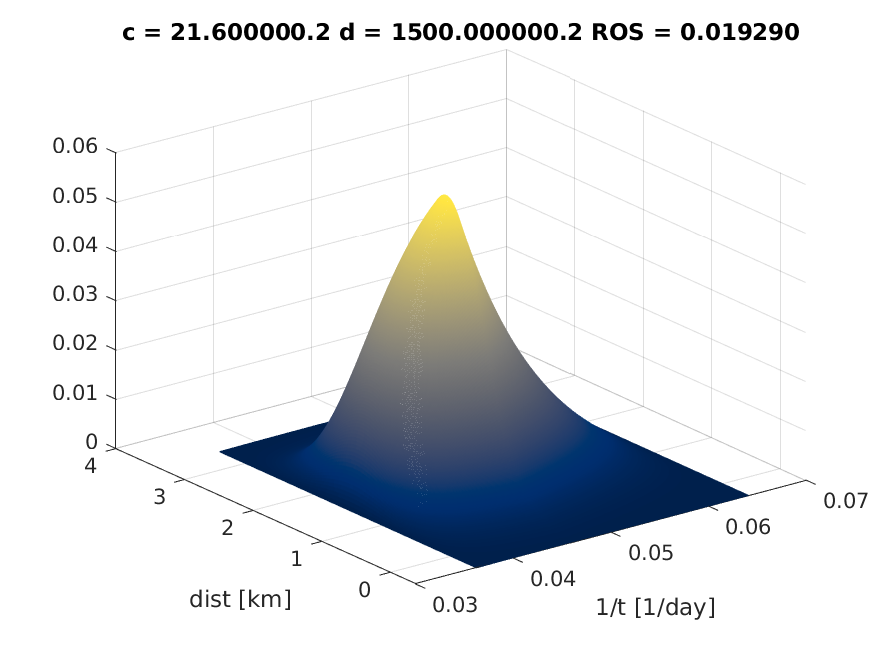}
\includegraphics[width = 0.45\textwidth]{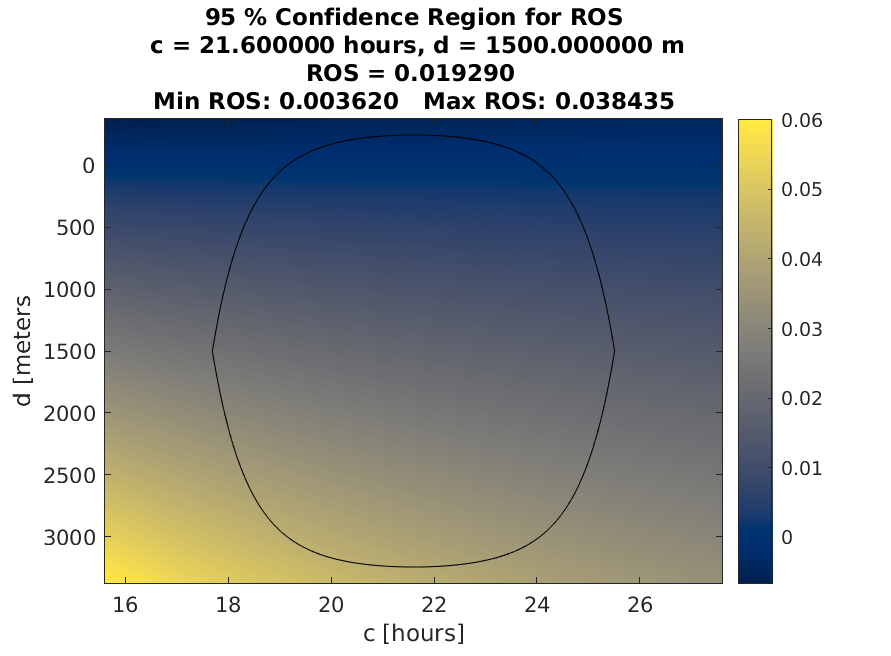}
\caption[The probability distribution and 95\% confidence region of the ROS with two fire detections separated by 21 hours and 1500 meters.]{The probability distribution and 95\% confidence region of the ROS with two fire detections separated by 21 hours and 1500 meters.
On the left is the probability distribution of the ROS when considered a random variable. On the right is a color plot of the ROS showing the 95\% confidence region for the ROS.}
\label{fig:ros_prob_95}
\end{center}
\end{figure}

\clearpage
\FloatBarrier

\chapter{\uppercase{Adjusting the Fire Arrival Time in the Model}} \label{chapter:04}
This chapter explains how satellite fire data is used to make corrections to a running wildfire simulation by a data assimilation technique that makes use of the data likelihood function and method for estimating the fire arrival time from data that were previously discussed. The first part of the chapter gives an explanation of the method and its rationale. Hypothetical examples are used to illustrate the method and a discussion of its limitations is given. The second part of the chapter makes use of the data assimilation method in modeling three real-world wildfires. The final part of the chapter details a workflow for modeling wildfires that uses infrared perimeter observations and satellite fire data to make an estimate of the fire arrival time to be used to initialize a wildfire simulation. After short-term forecast is made, data assimilation techniques are then used to update the model state and a longer forecast is then produced.




\section{DataAssimilation}\label{sec:data_assimilation}
Data assimilation seeks to make adjustments to the fire arrival time in a way that balances the uncertainties of the forecast with the uncertainties in the observational data. In a typical forecasting usage, a model is run forward for a length of time during which inaccuracies in predictions are likely to increase. During this initial model run, real-world observation of the wildfire may become available that can be used to correct the course of the model. The model is stopped and data is used to make corrections to the fire arrival time and the model is then restarted from the updated estimate. A cycling approach where a model is successively started and then corrected with new data can be used to help steer an in-progress simulation. An early example of the approach was used in \citet{Coen-2013-USR}. The Coupled Atmosphere-Wildland Fire Environment (CAWFE) \citep{Clark-1996-CAF} was used to simulate a fire from the ignition and fire perimeters estimated from VIIRS active fire detections were then used to re-initialize the fire simulation with each successive overpass of the satellite. In this experiment, the satellite data was used directly, without effort made to account for the inherent uncertainties in. Progress towards a more robust method using the cycling concept was made in \citet{Mandel-2014-DAS}. Instead of using satellite directly, a Bayesian approach was used to combine satellite data with the model forecast to obtain an updated state of the fire. This updated state of the fire was then used to form the restart conditions in a cycling routine similar to that used by Coen et al. The data assimilation method proposed in this thesis follows the cycling strategy employed by both Coen et al. and Mandel et al. but employs a different method for obtaining an update of the fire arrival time used to restart the simulation. The method follows closely from that described in Section \ref{sec:fire_arrival} that is used to estimate the fire arrival time from satellite data alone. Instead of using a rough estimate of the fire arrival time as a starting point for the iterative interpolation process, the model forecast fire arrival time is used and adjustments made to it according to the data the data likelihood function from \ref{sec:likelihood}. Both VIIRS and MODIS data are used as a source of observations.

\subsection{Method}

\begin{figure}[!ht]
\centering
       \includegraphics[width=0.45\textwidth]{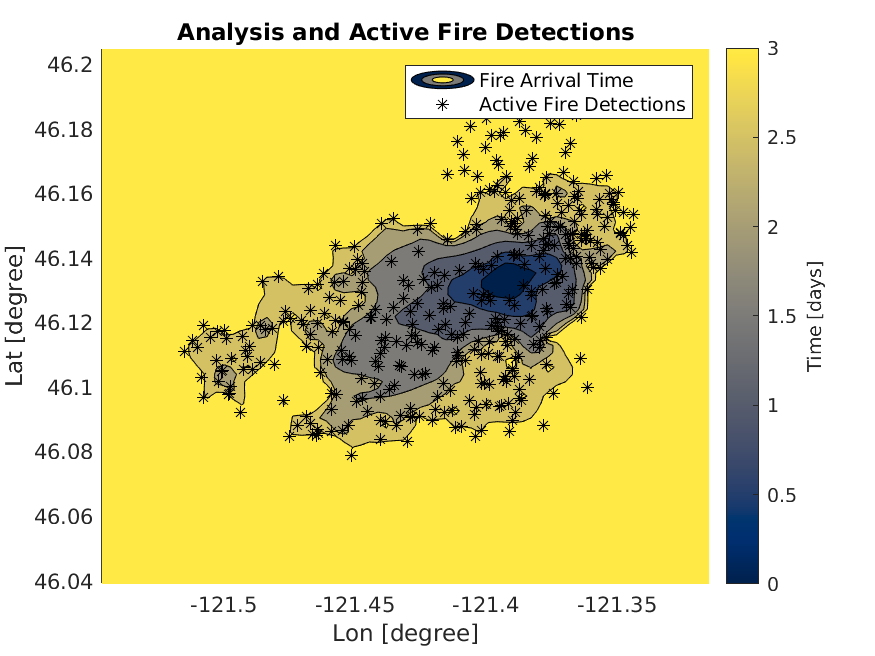}
       \includegraphics[width=0.45\textwidth]{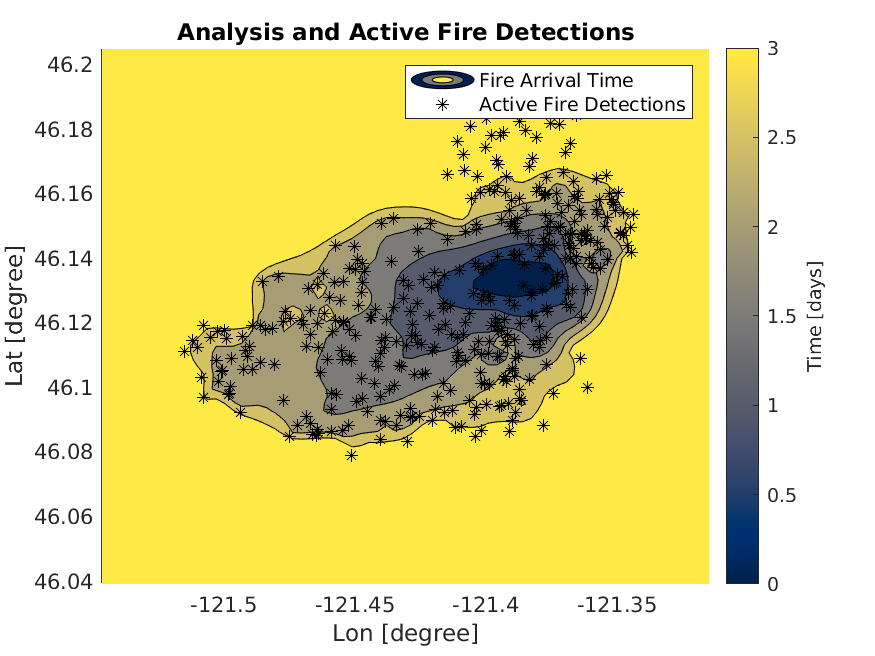}
       \includegraphics[width=0.45\textwidth]{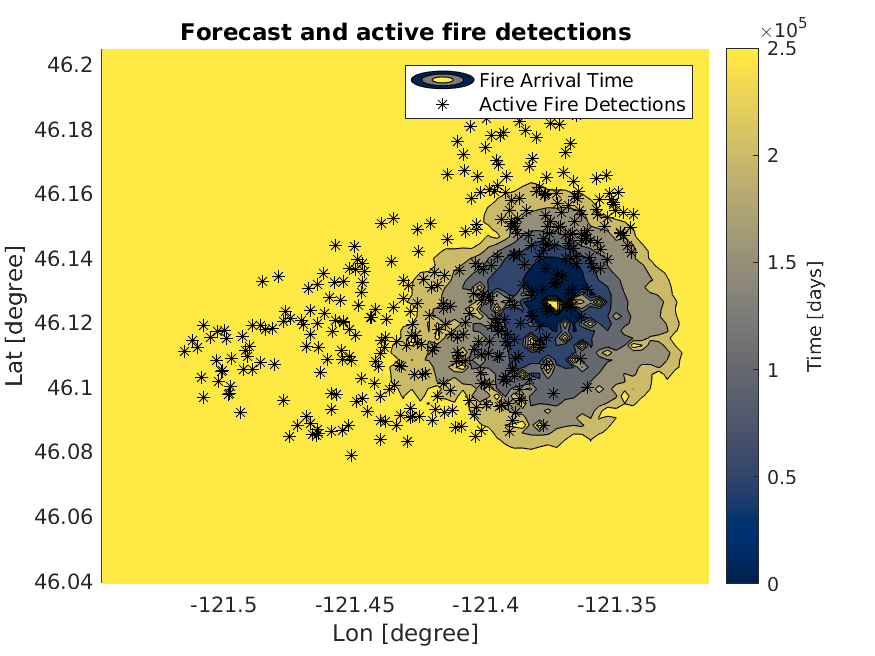}

     \caption[Illustrating the effect of using the data likelihood function for weighting the forecast fire arrival time and time of the active fire detections.]{Illustrating the effect of using the data likelihood function for weighting the forecast fire arrival time and time of the active fire detections. The top left and top right panels represent two versions on the analysis fire arrival time made without and with use of the data likelihood function, respectively. The bottom panel shows the forecast.}
\label{fig:likelihood_demo}
\end{figure}

The data assimilation method is similar to that used for estimation of fire arrival time from satellite data except that the initial estimate used is the model forecast and the data likelihood function regulates the adjustments made to the forecast. Both active fire detections and non-fire pixels are used to make adjustments to the forecast fire arrival time in order to make the analysis fire arrival time that blends forecast with satellite data. Figure \ref{fig:likelihood_demo} shows the difference that using the data likelihood function makes when using satellite data and an initial estimate of the fire arrival time to make an updated estimated. In the bottom panel of the figure are contours of an initial estimate of the fire arrival time, here taken to be the model forecast. In the upper left and the right are contours of two versions of the analysis fire arrival time. The analysis in upper left has been made without use of the data likelihood function and the iterative interpolation method has fit the satellite data closely as can be seen by the jagged fire perimeter. On the right, the analysis has been made with use of the data  likelihood function and the iterative interpolation has blended the forecast and data differently, achieving a smoother result.

\subsubsection{Active Fire Detections}

\begin{figure}[!ht]
\centering
       \includegraphics[width=0.45\textwidth]{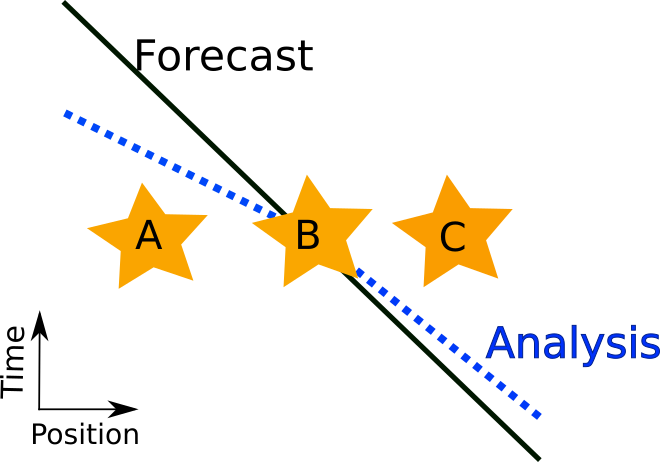}
     \caption[The data likelihood function determines how far the iterative interpolation process moves the forecast fire arrival time up or down towards the time of active fire detections.]{The data likelihood function determines how far the iterative interpolation process moves the forecast fire arrival time up or down towards the time of active fire detections.}
\label{fig:data_assimilation_moves}
\end{figure}

Active fire detections are used to adjust the forecast fire arrival time. The data likelihood function from Section \ref{sec:likelihood} controls how the forecast fire arrival time is adjusted at locations of satellite active fire detections in a manner similar to that used in Section \ref{sec:fire_arrival}.  Figure \ref{fig:data_assimilation_moves} shows how the data likelihood function determines how far the iterative interpolation process moves the forecast fire arrival time up or down towards the time of active fire detections. The orange stars indicate the time and location of active fire detections. The black line represents the forecast fire arrival time. Detection A has a time before the forecast fire arrival time, detection B has a time exactly coinciding with the forecast, and detection detection C has a time after the forecast. The analysis fire arrival time at detection A has the largest difference from the forecast at this location because the forecast is inconsistent with data. At location C, the difference between the forecast and analysis is smaller because the data and forecast fire arrival time are both plausible. The fire could have arrived at location C at the time of the forecast, but the satellite was not present to witness that arrival.

At each location of an active fire detection $d_i$, the analysis fire arrival time $T_a$ is a weighted average of the forecast fire arrival time $T_f$ and the time of the satellite fire detection $T(d_i)$ with a weight given by the data likelihood. The analysis fire arrival time is computed as
\begin{equation}
T_a = \alpha(T_f - T(d_i)) + T(d_i)
\end{equation}
where $\alpha$ is the computed from the exponential of the likelihood function of Equation \ref{eq:pos_detection}. Thus the weight $\alpha$ is a number between zero and one, given that the false detection rate used in the likelihood function is set to zero. If $\alpha=1$, then $T_a = T_f$ and $\alpha=0$ gives $T_a = T(d_1)$. In practice, the false detection rate is set to several percent and $\alpha$ is never zero, meaning that the satellite data is never fully trusted and the forecast fire arrival time always influences the the analysis fire arrival time $T_a$ at every location in the fire domain. 

With the analysis fire arrival time computed at each location of an active fire detection, the iterative interpolation technique from Section \ref{sec:fire_arrival} is then used to make adjustments to the fire arrival time at nearby locations with a local averaging method. Figure \ref{fig:ana_no_iter_interp} shows a sequence of panels depicting the method on a one-dimensional fire line. In this example, no additional data points have been interpolated along the paths. Figure \ref{fig:ana_iter_interp} shows an identical scenario except that additional points have been interpolated along the paths. The interpolation of more points helps this analysis better resolve the gradient information implied by the data that is seen on the left side of the figures at approximately time $t = 5.5.$

\subsubsection{Non-fire Pixels}
\begin{figure}[!ht]
\centering
       \includegraphics[width=0.45\textwidth]{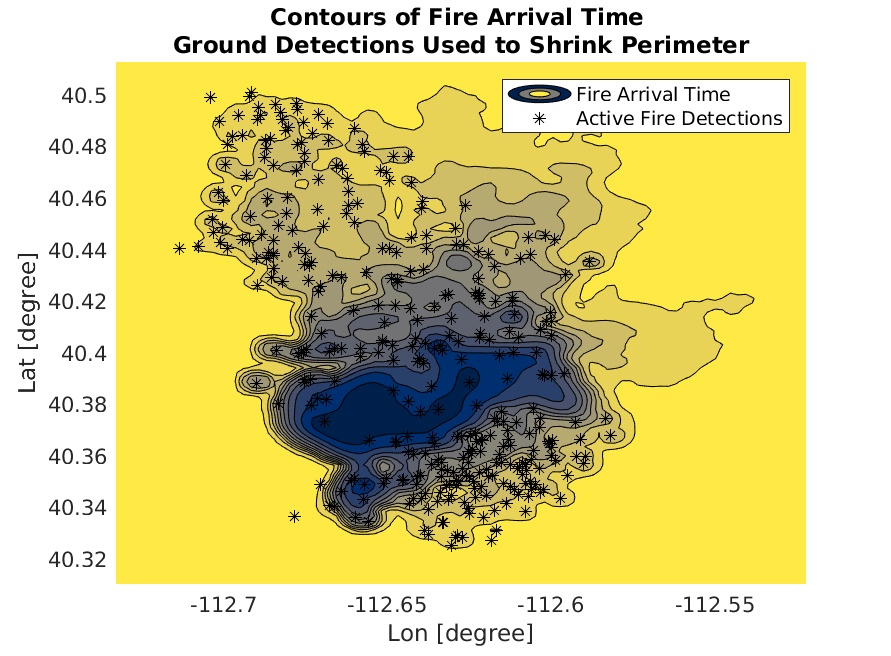}
       \includegraphics[width=0.45\textwidth]{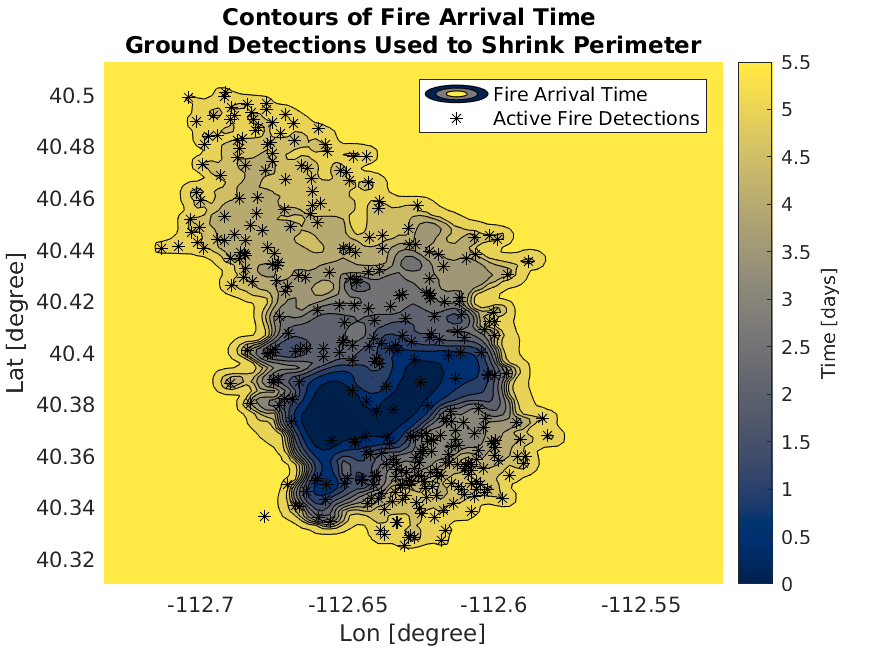}
       \includegraphics[width=0.45\textwidth]{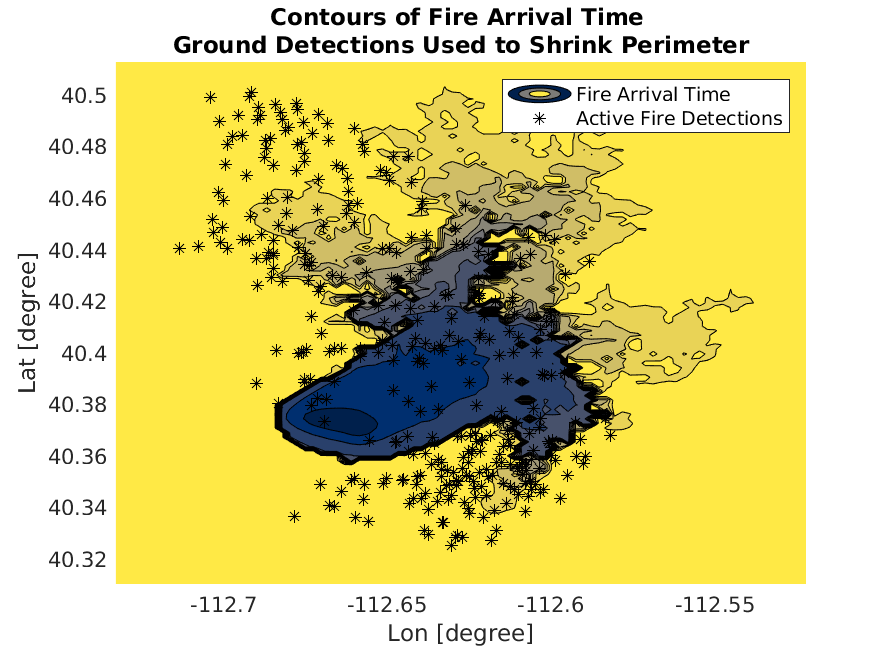}
     \caption[Comparison of the analysis made with and without use of non-fire pixel data.]{Comparison of the analysis made with and without use of non-fire pixel data. One the bottom are the original forecast fire arrival time and detection locations. The upper left panel shows analysis formed without use of non-fire pixels. Note the perimeters extend into regions not containing any fire detection pixels. The upper panel on the right shows the analysis formed using non-fire pixel data. The outer perimeter more closely matches the locations where active fire detections are located. }
\label{fig:exp_using_ground}
\end{figure}

In the same way that the data likelihood function determines the weighting of the forecast fire arrival time and the satellite fire detection time at the fire detection locations, the data likelihood determines how the analysis fire arrival time will be a weighted average of the forecast fire arrival time and the end-time of the estimation period. Figure \ref{fig:patch_ground_detects} shows how the forecast fire arrival time at the locations of non-fire pixels is moved upwards towards the ``flat top" of the fire arrival time cone that marks the end-time of the simulation. The non-fire pixels are only used to make adjustments to the forecast fire arrival outside of a polygon that is drawn around the active fire detections in the domain. Figure \ref{fig:exp_using_ground} shows the effect of using non-fire pixels to form the analysis fire arrival time. The bottom panel shows the forecast fire arrival time and the upper left and upper right panels show the analysis formed without using and with using ground non-fire pixels, respectively.

\begin{figure}[!ht]
\centering
       \includegraphics[width=0.45\textwidth]{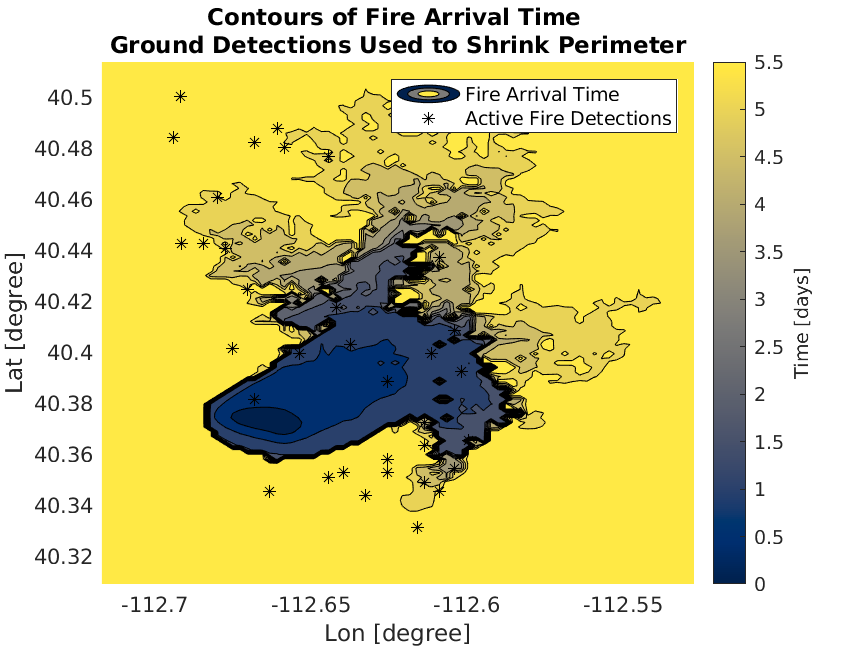}
       \includegraphics[width=0.45\textwidth]{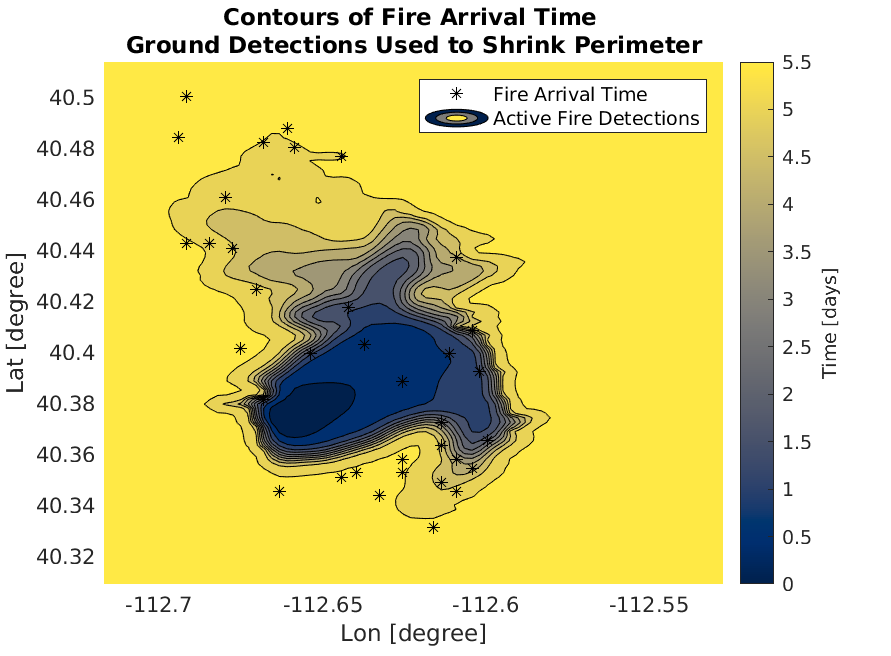}
     \caption[Using non-fire pixels to help constrain the size of a simulation or estimate of fire arrival time. ]{Using non-fire pixels to help constrain the size of a simulation or estimate of fire arrival time. The original perimeters of the forecast fire arrival time are on the left. Note that the contours lines in the northeast of the fire area extend into areas without any active fire detections. On the right is the effect of moving the fire arrival time upward for non-fire pixels outside of the polygon containing active fire detections. The iterative process shrinks the perimeter and smooths the fire arrival time.}
\label{fig:patch_shrink_perims}
\end{figure}

%
%
%
\subsection{Limitations of the Method}
The data assimilation method developed has some limitations in its utility due to the way its was designed and implemented. We outline some of those limitations below and offer potential remedies.

\subsubsection{Use of Non-fire Pixels}
The use of non-fire pixels to form the analysis fire arrival time is restricted to a region outside of a polygon  drawn around the locations of the active fire detections. This represents a loss of of information about the behavior of the fire. It is expected that some regions within the outer perimeter of the fire may remain unburned. Indeed, there may be regions contained within the fire perimeters, such as lakes or sand dunes, that contain no fuels to be burned. As implemented, the method developed here can assign a fire arrival time to a location that can never burn. One possible remedy for this shortcoming would be to use information about the fuels in the fire domain in addition to the non-fire pixels. The fire arrival time at locations where no fuels are found and non-fire pixels were recorded could be set to the end-time of the estimation period.

\subsubsection{Data Likelihood Usage}
\begin{figure}[!h]
\begin{center}
\includegraphics[width = 0.45\textwidth]{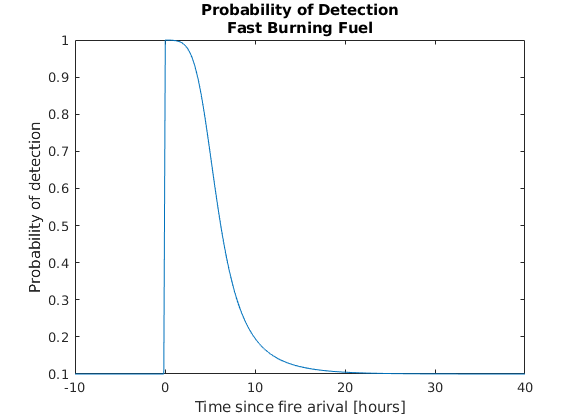}
\includegraphics[width = 0.45\textwidth]{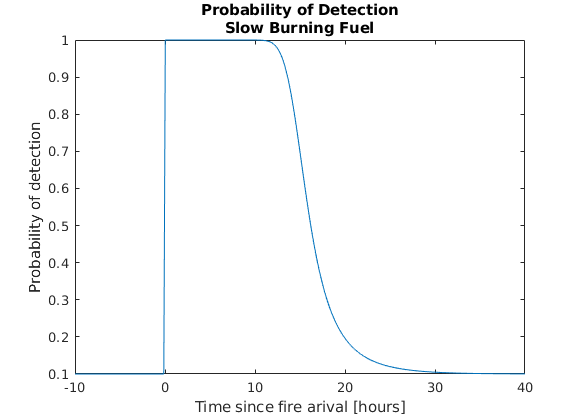}
  \caption[Illustrating the difference in the probability of satellite fire detection for regions with  different fuel types.]{Illustrating the difference in the probability of satellite fire detection for regions with  different fuel types.  On the left is the probability of detection for a region with a fast burning fuel. On the right is the probability of detection for a region with a slow burning fuel. }
  \label{fig:detect_prob_fuels}
  \end{center}
\end{figure}

For the examples in this thesis, the data likelihood used the same parameters for all locations in the fire domain, regardless of the effects of weather, terrain, and fuel properties. In particular, the fuel types in the fire domain would have an effect on the probability of the satellite detection that is a component of the data likelihood. For example, one would expect that fire in a dry, grassy region would be detectable by satellite for a shorter time than fire in a forest with large trees because the mass of fuel in the grassy region would be less and the fuel would be completely consumed by fire more rapidly. Figure \ref{fig:detect_prob_fuels} shows the difference in the shapes of curves giving the probability of detection since fire arrival time for a region with a fast burning fuel and for a region with a slow burning fuel. The curves are hypothetical, based on Equation \ref{eq:logistic_detection} and setting the parameter $a$, which can be determined by Equation \ref{eq:detect_prob_a}, so that the probability of detection being 20\% after 10 hours after the fire arrival time for the fast burning fuel and 20\% after 20 hours for the slow burning fuel. The data likelihood function described in this work does have the capability to account for such differences. Additional research is required to adapt it for the varying fuel types that are used by the fire spread model. 

\section{Operational Usages}\label{sec:operational}
In an operational setting, a fire simulation is initialized from an ignition point or from an estimate of the fire arrival time. After some time of running the model forward, new data becomes available and is used to make adjustments to the model output in a way that balances the uncertainties that exist in both the forecast and observational data. This analysis fire arrival time then becomes the starting point from which the next forecast is initialized. A typical simulation includes several cycles of making a model  forecast followed by performing data assimilation. 

In this section, two forecasting strategies using data assimilation will be demonstrated. In the first, only satellite observations will be used as a a data source. With rare exception, satellite data can be obtained for any fire to be simulated and this first strategy can be always be used. The second strategy will make use of infrared perimeter observations in addition to satellite data. In general, these types of observations will not be available for every fire. However, for large fires that pose a significant risk to life and property, they are often made available by the responding authorities. When available, these observations can be used to make an estimated fire arrival time that can be used to initialize a fire simulation. Both strategies are employed to simulate real-world fires and the results of those simulations are assessed by comparison with infrared perimeter observations, using the tools described in Section \ref{sec:assess_tools}.

\subsection{Forecast Cycling with Data Assimilation}
\label{cycling_da}
When only satellite observations are available, the forecast method that uses data assimilation starts with a simulation initialized from an ignition point or estimated fire arrival time and then periodically updates the model state with assimilation of satellite fire observations. A sequence of short-term forecasts are made in a cyclic manner where the forecasts are followed by updates. The basic method is outlined below.

\begin{enumerate}
\item Begin a simulation from an ignition point or a fire arrival time that has been estimated using satellite fire data. 
\item Run the model forward for a specified amount of time. In an operational setting, this might be a simulation spanning one day or up to a week. Typically, simulations for periods longer than two days begin to lose accuracy. See the results from Section \ref{sec:patch_data_assilation} to see some results.
\item Collect all the satellite data made available since the initial data collection in the first step. 
\item Construct the analysis fire arrival time from the directed graph with shortest paths as detailed in Section \ref{sec:data_assimilation}. Insert the analysis into the WRF-SFIRE restart file for the simulation.
\item Restart the simulation from the restart file, using the analysis in place of an ignition point or estimated fire arrival time as prescribed in step 1.
\item Go to step 2 and repeat steps 3 and 4 until the simulation end time.
\end{enumerate}

This procedure was used to simulate three real-world fires. In each case, forecast cycling with data assimilation produced forecasts that better matched infrared perimeter observations than simulations that made no attempt to adjust the course of an in-progress fire with data assimilation techniques. 

\subsubsection{Patch Springs Fire}\label{sec:patch_data_assilation}

The Patch Springs fire took place in August 2013 approximately 60 kilometers southwest of Salt Lake City, Utah. The fire is estimated to have started early in the evening of August 10 and was finally 100\% contained by October 23 \citep{Inciweb-2013-Patch} . Initially starting slowly, the fire advanced rapidly on August 14\citep{Gabbert-2019-UPS}. The following discussion pertains to modeling this fire with WRF-SFIRE during its first eight days.

\begin{figure}[!h]
\begin{center}
  \includegraphics[width = 0.9\textwidth]{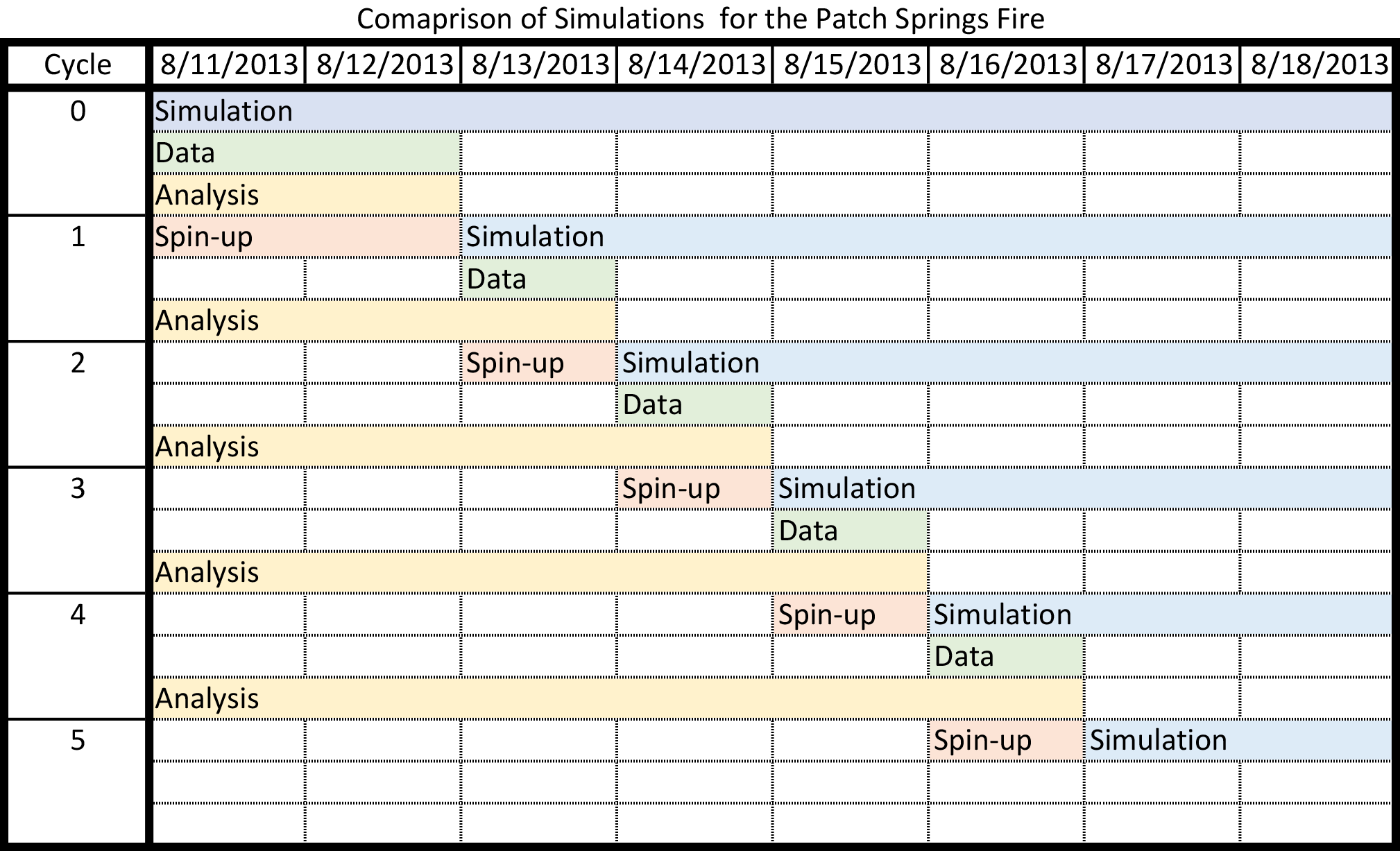}
  \caption[Schematic for several data assimilation cycles run consecutively for simulation of the Patch Springs fire.]{Schematic for several data assimilation cycles run consecutively for simulation of the Patch Springs fire. For the sake of comparison, all simulations were run from the end of the spin-up period the end time of the entire simulation period. Operationally, the cycle run would be terminated when new data was assimilated and a restart of the simulation initiated.}
  \label{fig:patch_spinup}
  \end{center}
\end{figure}

Using WRF-SFIRE, the Patch Springs fire was simulated over the period of August 11 to August 18. the ignition location was chosen as $40.37^{\circ} N, -112.64^{\circ}W$, with the ignition time of 06:00 UTC. The simulation was run on one computational domain with a relatively coarse grid spacing of 225 meters. Five data assimilation cycles were incorporated into the simulation. Figure \ref{fig:patch_spinup} illustrates the timing of the simulation periods. An explanation of the cycles of the simulation follows.

\begin{itemize}
\item \textbf{Cycle 0} - The simulation was started from an ignition point and run for eight days. Satellite data from the first two days was collected and combined with the first two days of the simulation forecast, forming the analysis fire arrival time that covered the first two days of the fire simulation.
\item \textbf{Cycle 1} - The analysis fire arrival time from cycle 0 was used to spin-up the model, replaying the first two days of the fire and bringing the state of the atmosphere into synchronization with the state of the fire. After the two days of spin-up time, the model took over and produced an additional six days of forecast. Satellite data from the first day of this forecast was collected and combined with the first day of the simulation forecast, forming the analysis fire arrival time that covered the first three days of the fire simulation.
\item \textbf{Cycle 2} - The analysis fire arrival time from cycle 1 was used to spin-up the model, replaying the third day of the fire. After the one day of spin-up time, the model took over and produced an additional five days of forecast. Satellite data from the first day of this forecast was collected and combined with the first day of the simulation forecast, forming the analysis fire arrival time that covered the first four days of the fire simulation.
\item \textbf{Cycle 3} - The analysis fire arrival time from cycle 2 was used to spin-up the model, replaying the fourth day of the fire. After the one day of spin-up time, the model took over and produced an additional four days of forecast. Satellite data from the first day of this forecast was collected and combined with the first day of the simulation forecast, forming the analysis fire arrival time that covered the first five days of the fire simulation.
\item \textbf{Cycle 4} -  The analysis fire arrival time from cycle 3 was used to spin-up the model, replaying the fifth day of the fire. After the one day of spin-up time, the model took over and produced an additional three days of forecast. Satellite data from the first day of this forecast was collected and combined with the first day of the simulation forecast, forming the analysis fire arrival time that covered the first six days of the fire simulation.
\item \textbf{Cycle 5} -  The analysis fire arrival time from cycle 4 was used to spin-up the model, replaying the fifth day of the fire. After the one day of spin-up time, the model took over and produced an additional two days of forecast.
\end{itemize}


\begin{figure}[!h]
\begin{center}
  \includegraphics[width = 0.45\textwidth]{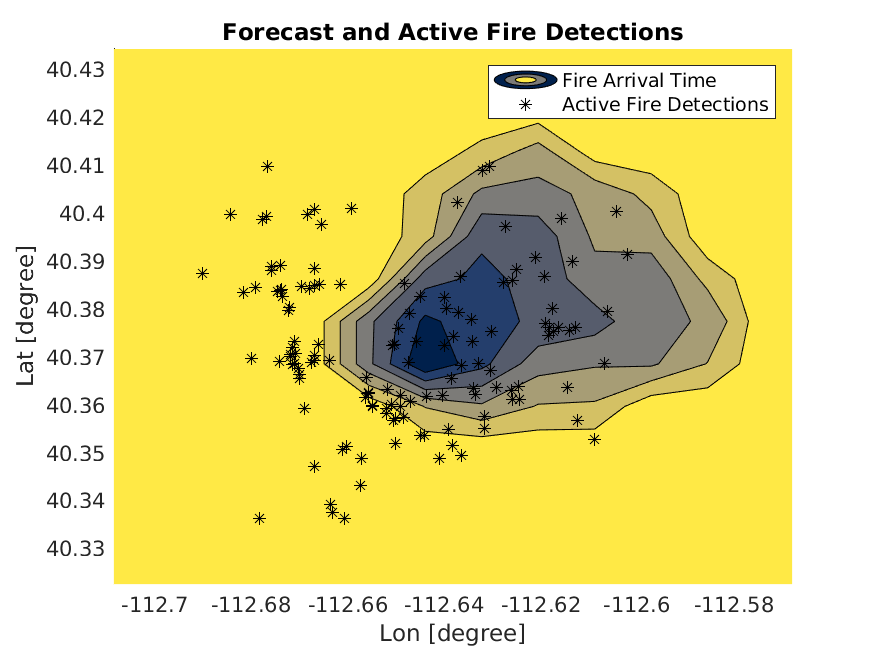}
    \includegraphics[width = 0.45\textwidth]{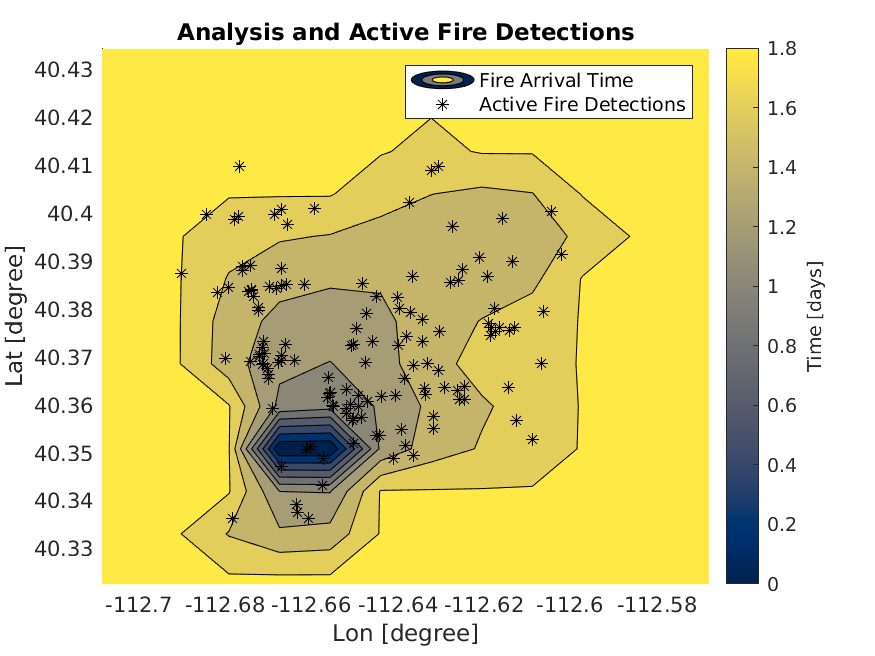}
  \caption[Contours of the forecast the fire arrival time (left) and the analysis (right) for the first cycle of the Patch Fire simulation.]{Contours of the forecast the fire arrival time (left) and the analysis (right) for the first cycle of the Patch Fire simulation. The data assimilation was performed on a computational grid with 250 meter spacing. Note that the ignition point has been moved towards the south by the process. The outer perimeter of the analysis fir arrival time was shifted eastward over much of the fire domain to better match the satellite fire detections.}
  \label{fig:patch_cycle_0}
  \end{center}
\end{figure}

\begin{figure}[!h]
\begin{center}
  \includegraphics[width = 0.45\textwidth]{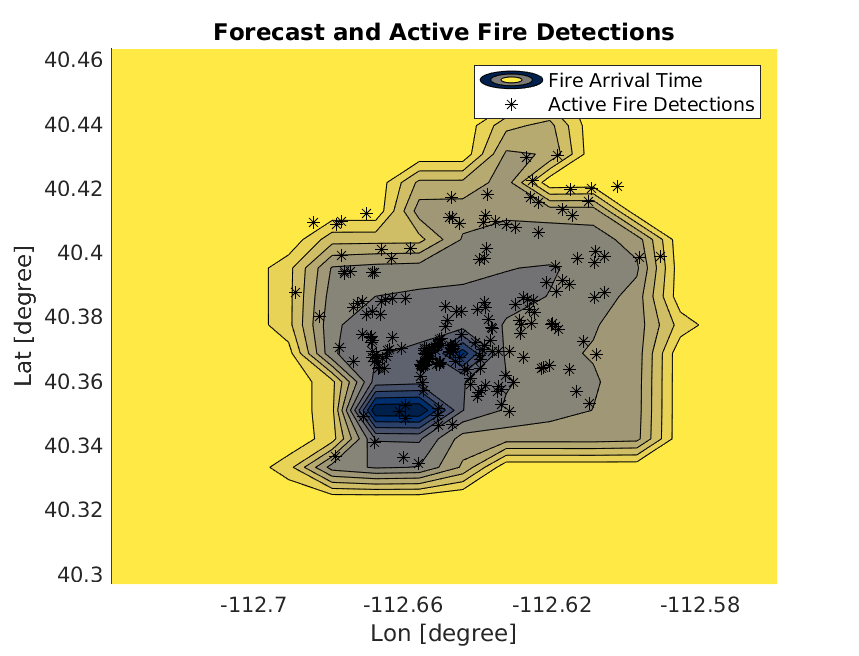}
    \includegraphics[width = 0.45\textwidth]{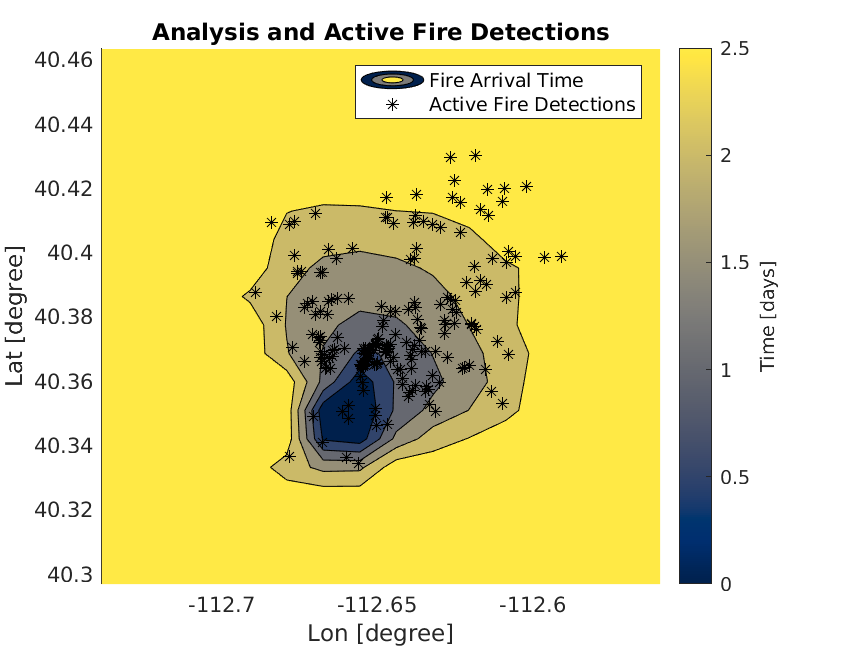}
  \caption{Contours of the forecast the fire arrival time (left) and the analysis (right) for the second cycle of the Patch Fire simulation. The data assimilation was performed on a computational grid with 250 meter spacing. Minimal changes were made to the forecast during this cycle.}
  \label{fig:patch_cycle_1}
  \end{center}
\end{figure}

\begin{figure}[!h]
\begin{center}
  \includegraphics[width = 0.45\textwidth]{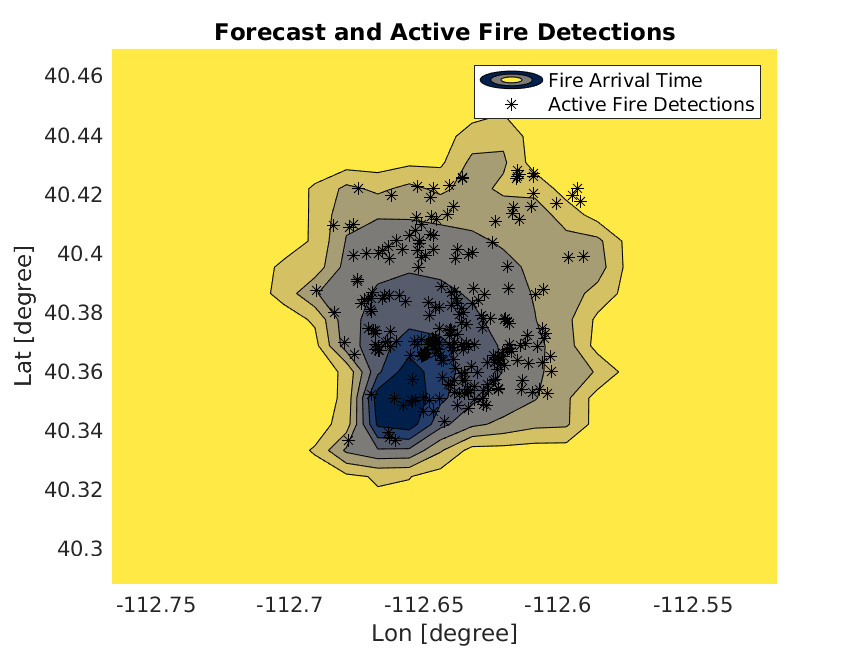}
    \includegraphics[width = 0.45\textwidth]{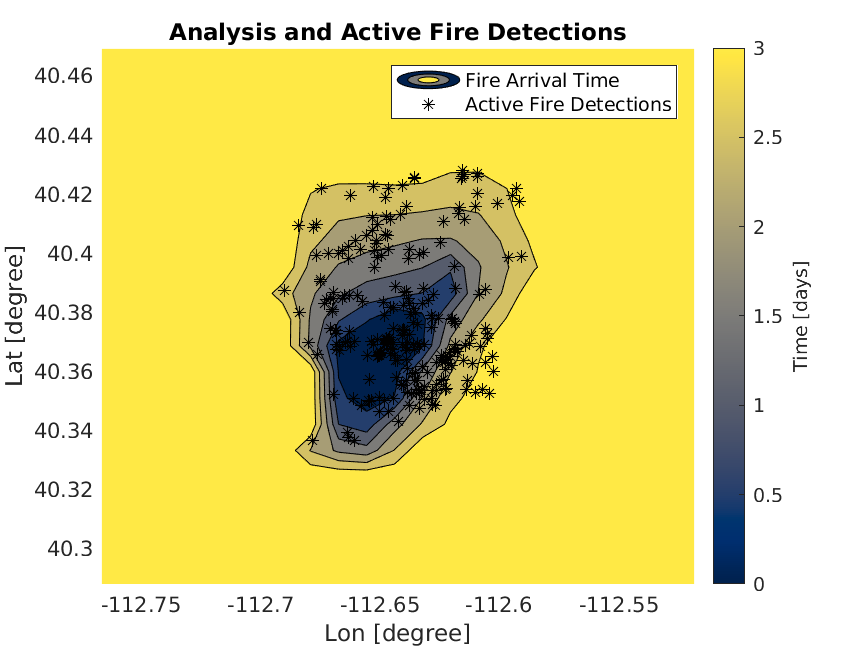}
  \caption{Contours of the forecast the fire arrival time (left) and the analysis (right) for the third cycle of the Patch Fire simulation. The data assimilation was performed on a computational grid with 250 meter spacing. Minimal changes were made to the forecast during this cycle.}
  \label{fig:patch_cycle_2}
  \end{center}
\end{figure}

\begin{figure}[!h]
\begin{center}
  \includegraphics[width = 0.45\textwidth]{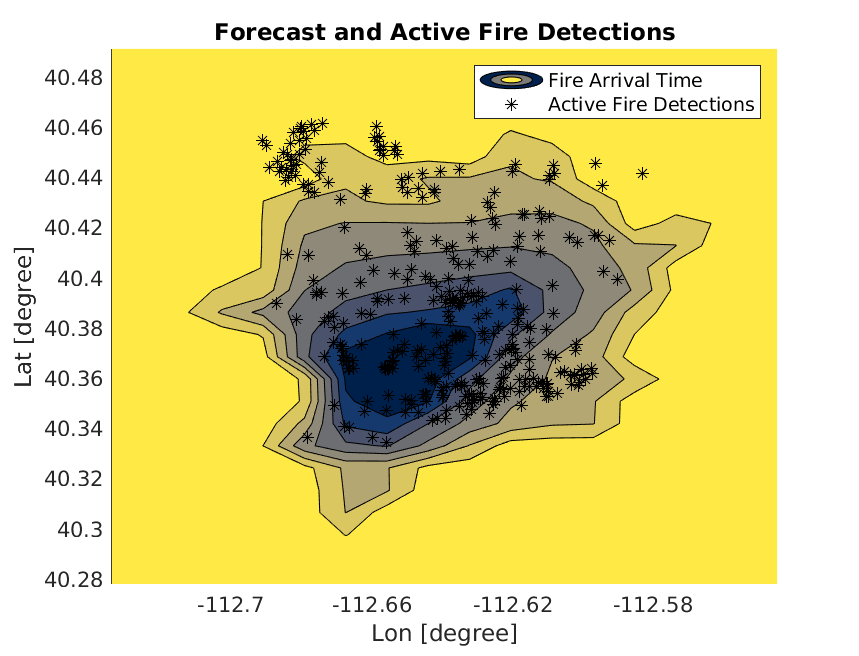}
    \includegraphics[width = 0.45\textwidth]{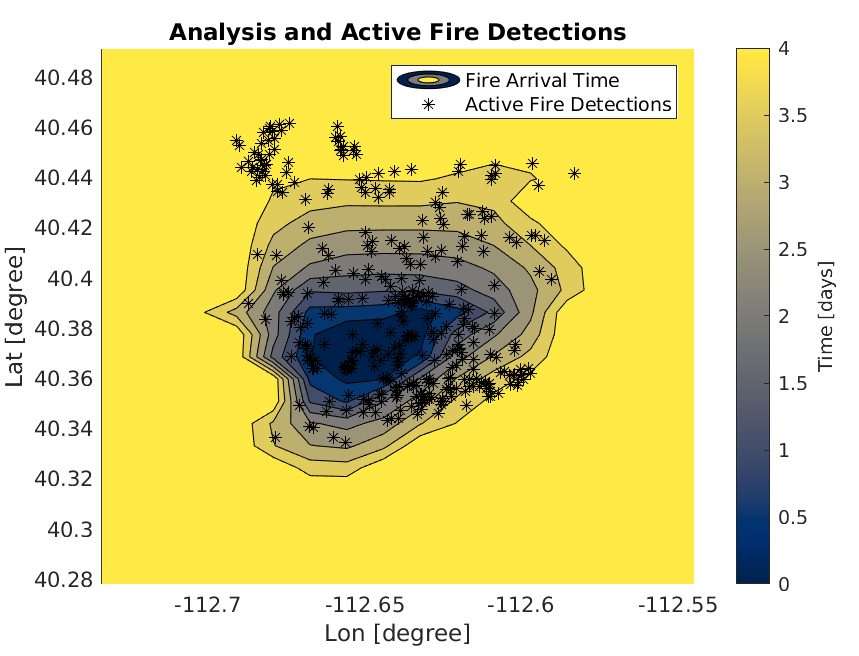}
  \caption{Contours of the forecast the fire arrival time (left) and the analysis (right) for the fourth cycle of the Patch Fire simulation. The data assimilation was performed on a computational grid with 250 meter spacing. The outer  perimeter of the forecast fire arrival time was expanded in the northwest region of the fire domain due to the presence of many active fire detections in the region. }
  \label{fig:patch_cycle_3}
  \end{center}
\end{figure}

\begin{figure}[!h]
\begin{center}
  \includegraphics[width = 0.45\textwidth]{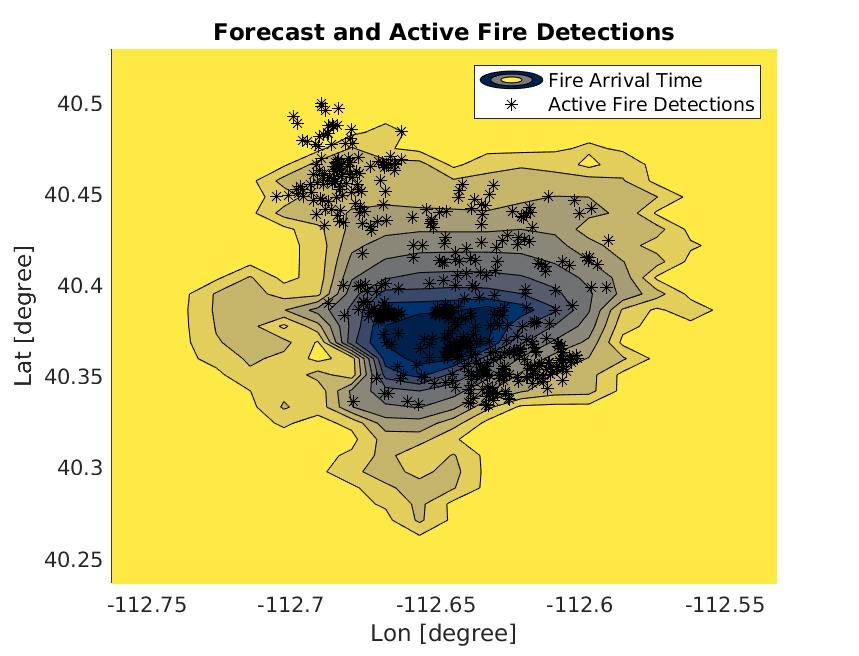}
    \includegraphics[width = 0.45\textwidth]{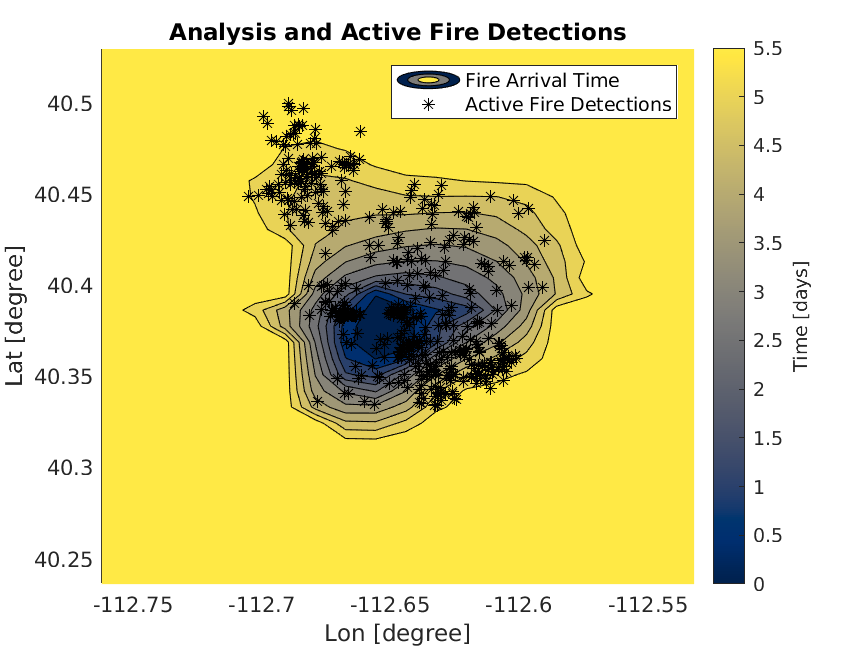}
  \caption[Contours of the forecast the fire arrival time (left) and the analysis (right) for the fifth cycle of the Patch Fire simulation.]{Contours of the forecast the fire arrival time (left) and the analysis (right) for the fifth cycle of the Patch Fire simulation. The outer  perimeter of the forecast fire arrival time was expanded in the northwest region of the fire domain due to the presence of many active fire detections in the region. }
  \label{fig:patch_cycle_4}
  \end{center}
\end{figure}

Figures \ref{fig:patch_cycle_0}-\ref{fig:patch_cycle_4} show how the data assimilation method was able to adjust the forecast fire arrival time to make use of the satellite data collected. As can be seen in Figure \ref{fig:patch_cycle_0}, the initial simulation required significant adjustment to to account for the large number of fire detections that were outside of its outer perimeter on the east side of the active fire region. During the remainder of the simulation, smaller adjustments were made to the fire arrival time by the data assimilation method. Each of the cyclic data assimilation adjustments incorporated 24 hours of additional satellite data. The minimal adjustments to the forecast indicate the model was providing good short-term forecasts.

To assess the method, an infrared perimeter observation was used to evaluate how well the forecasts matched the actual fire. Although the data assimilation routine made periodic adjustments and restarts of the model, every new forecast start was allowed to run to at least the time of the infrared perimeter observation on August 16, at 09:47 UTC. Table \ref{tbl:patch_moe} shows the results obtained as assessed using the MOE and S{\o}renson index. Graphically, the results are summarized in Figure \ref{fig:patch_moe_cycles}. The forecasts tended to overestimate the size of the fire in the northeast and southeast areas of the fire domain and underestimate it in the northwest. 

Also included in the table of results and in  Figure \ref{fig:patch_moe_cycles} are results for a simulation labeled ``Estimated Start." This model run was initialized from an estimated fire arrival time that was made using satellite fire detection data. This short-term forecasts did not exhibit the overestimation of the size of the fire in the northeast and southeast areas of the fire domain like the other forecasts did, but it also failed to predict the fire growth in the northeast are of the fire domain.

\begin{table}
\begin{tabular}{|c|c|c|c|c|c|}
\hline 
Cycle & Hours of Forecast & MOE X  & MOE Y & ||MOE|| & S{\o}renson \\ 
\hline 
0 & 130 & 0.4920 & 0.7629 & 0.9078 & 0.5982 \\ 
\hline 
1 & 82	& 0.7543 & 0.6413 &	0.9901 & 0.6932 \\ 
\hline 
2 & 	58	& 0.7009	 & 0.6476 & 	0.9543 & 0.6732 \\
\hline 
3 & 34	& 0.7733 & 	0.6394 & 1.0034	& 0.7000 \\
\hline 
4 & 10	& 0.8993 & 0.6297 & 	1.0978	& 0.7407  \\
\hline 
Estimated Start & 12.5 & 0.6143 & 0.9149 & 1.1020 & 0.7350 \\ 
\hline 
\end{tabular}
\caption{Scores for the cycling of the Patch Springs fire when compared with an infrared perimeter observation made August 16, at 09:47 UTC. The results in this table are summarized visually in Figure \ref{fig:patch_moe_cycles}}
\label{tbl:patch_moe}
\end{table}

\subsubsection{Cougar Creek Fire}
\label{sec:cougar_data_assilation}
The Cougar Creek fire took place in August, 2015. The fire is thought to have started naturally due to lightning strike at approximately 6:00 PM local time on August 10. Overall growth of the fire was slow due to high relative humidity and cloud cover. More than 400 fire fighting personnel were deployed and by September 14 the fire was 97\% contained. The fire was fueled by timber (some beetle-killed), light logging slash, and tall grass. The total size of the eventually grew to  more than 53,000 acres \citep{Inciweb-2015-Cougar}.

We simulated the first 4.5 days of the Cougar Creek fire. The simulation was performed on a three-domain computational grid, with the fire model running at a resolution of roughly 30 meters between grid nodes. For comparison, all cycles were run to produce a fire arrival time of the entire 4.5 days of simulation. Periodic restarts of the simulation were made after 24 hours of simulation time. The schematic in Figure \ref{fig:cougar_short_schematic} shows how the cycling was accomplished. Further explanation of the timing follows.

\begin{itemize}
\item \textbf{Cycle 0} - The simulation was started from an ignition point and run for 4.5 days. Satellite data from the first  day was collected and combined with the first  day of the simulation forecast, forming the analysis fire arrival time that covered the first day of the fire simulation.
\item \textbf{Cycle 1} - The analysis fire arrival time from cycle 0 was used to spin-up the model, replaying the first day of the fire and bringing the state of the atmosphere into synchronization with the state of the fire. After the one day of spin-up time, the model took over and produced an additional 3.5  days of forecast. Satellite data from the first day of this forecast was collected and combined with the first day of the simulation forecast, forming the analysis fire arrival time that covered the first two days of the fire simulation.
\item \textbf{Cycle 2} - The analysis fire arrival time from cycle 1 was used to spin-up the model, replaying the second day of the fire. After the one day of spin-up time, the model took over and produced an additional 2.5 days of forecast. Satellite data from the first day of this forecast was collected and combined with the first day of the simulation forecast, forming the analysis fire arrival time that covered the first three days of the fire simulation.
\item \textbf{Cycle 3} - The analysis fire arrival time from cycle 2 was used to spin-up the model, replaying the third day of the fire. After the one day of spin-up time, the model took over and produced an additional 1.5 days of forecast. Satellite data from the first day of this forecast was collected and combined with the first day of the simulation forecast, forming the analysis fire arrival time that covered the first four days of the fire simulation.
\item \textbf{Cycle 4} -  The analysis fire arrival time from cycle 3 was used to spin-up the model, replaying the fourth day of the fire. After the one day of spin-up time, the model took over and produced an half day of forecast. 
\end{itemize}

\begin{figure}[!h]
\begin{center}
  \includegraphics[width = 0.45\textwidth]{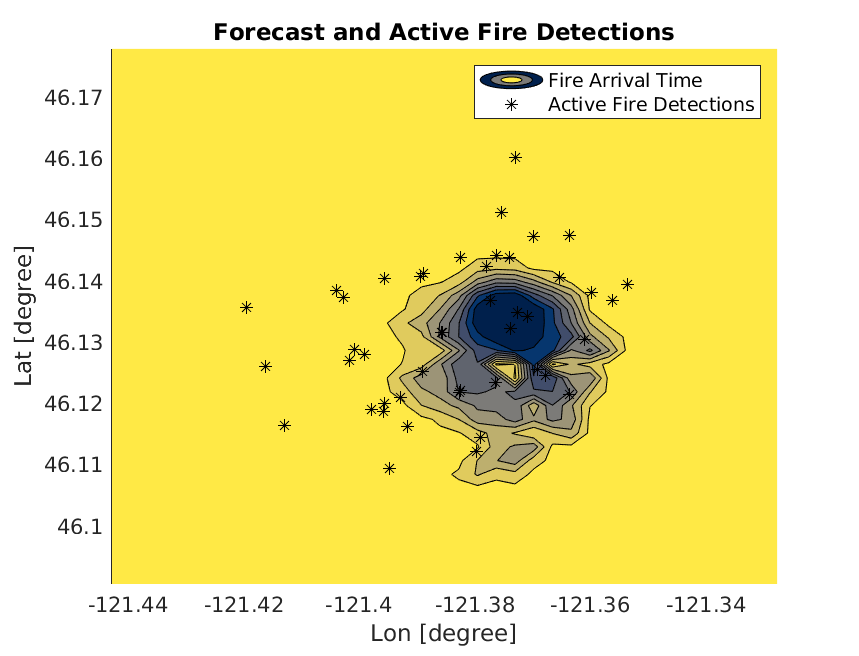}
    \includegraphics[width = 0.45\textwidth]{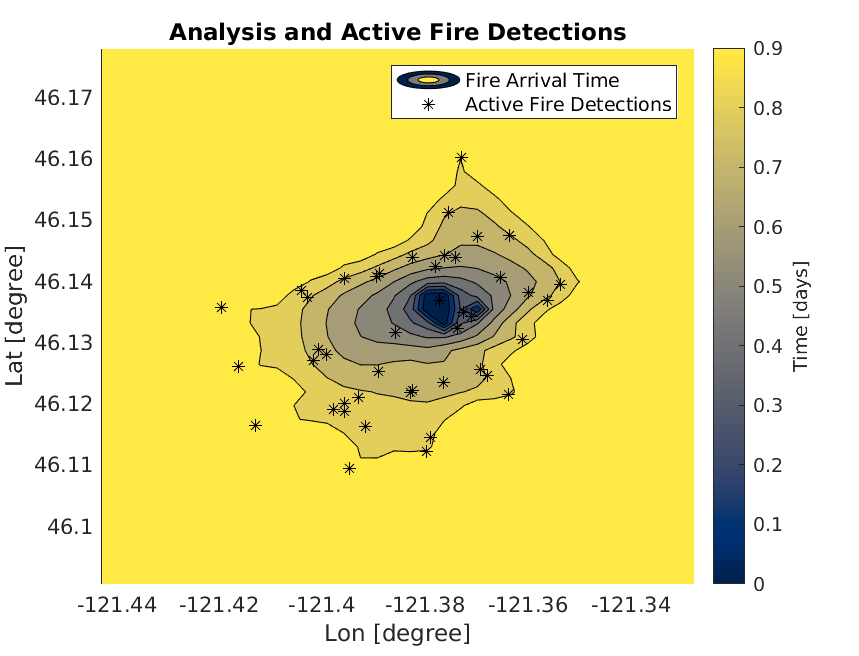}
  \caption{Contours of the forecast the fire arrival time (left) and the analysis (right) for the first cycle of the Cougar Creek Fire simulation. The data assimilation was performed on a computational grid with 250 meter spacing. The forecast had underestimated the size of the fire and the data assimilation expanded the outer perimeter to contain more of the active fire detections.}
  \label{fig:cougar_cycle_0}
  \end{center}
\end{figure}

\begin{figure}[!h]
\begin{center}
  \includegraphics[width = 0.45\textwidth]{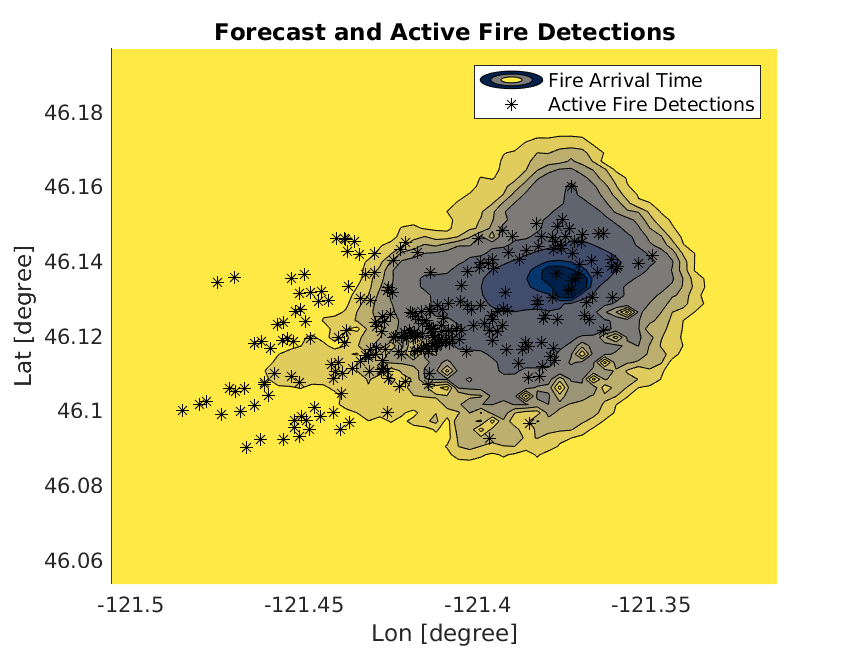}
    \includegraphics[width = 0.45\textwidth]{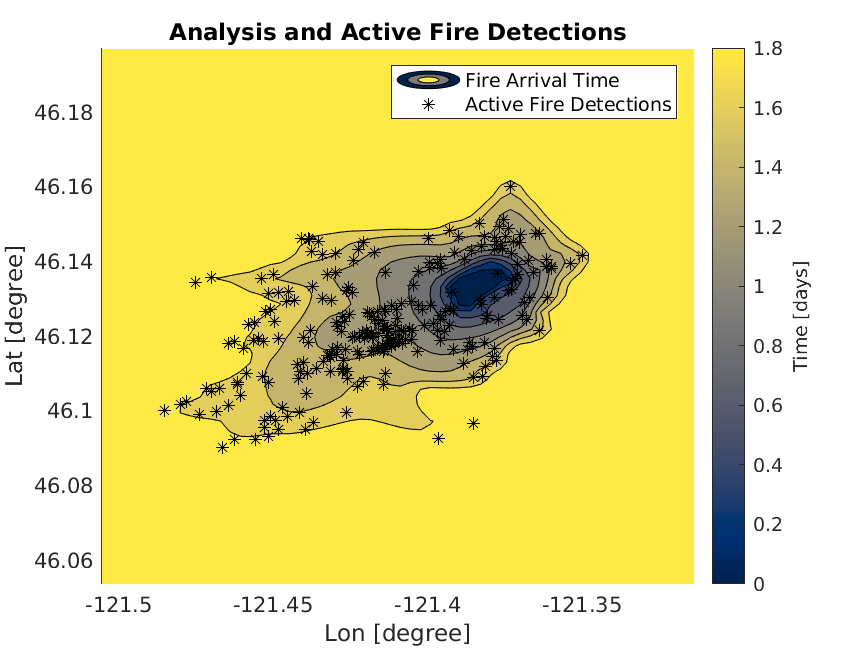}
  \caption{Contours of the forecast the fire arrival time (left) and the analysis (right) for the second cycle of the Cougar Creek Fire simulation. The data assimilation was performed on a computational grid with 250 meter spacing. Again, the outer perimeter was expanded to contain more of the active fire detections.}
  \label{fig:cougar_cycle_1}
  \end{center}
\end{figure}

\begin{figure}[!h]
\begin{center}
  \includegraphics[width = 0.45\textwidth]{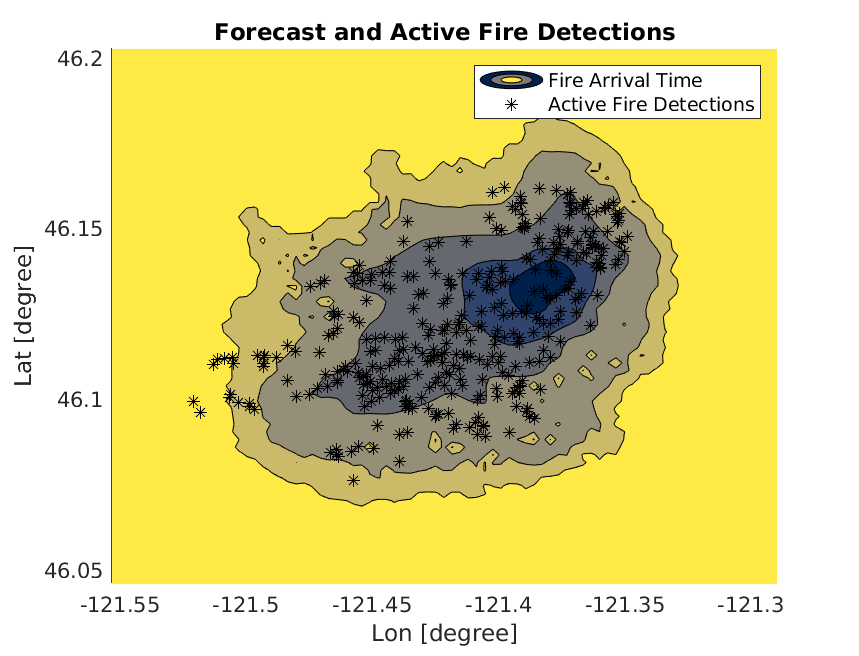}
    \includegraphics[width = 0.45\textwidth]{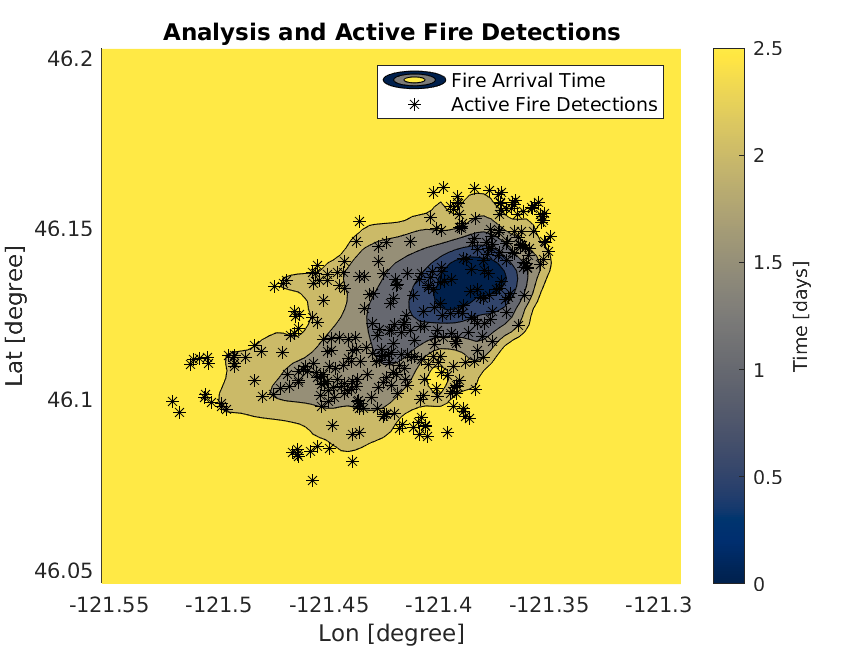}
  \caption{Contours of the forecast the fire arrival time (left) and the analysis (right) for the third cycle of the Cougar Creek Fire simulation. The data assimilation was performed on a computational grid with 250 meter spacing. The outer perimeter was contracted slightly by the data assimilation method in this cycle. The forecast had shown more growth in the northeast and southest sections of the fire region than was indicated by the satellite data. }
  \label{fig:cougar_cycle_2}
  \end{center}
\end{figure}

\begin{figure}[!h]
\begin{center}
  \includegraphics[width = 0.45\textwidth]{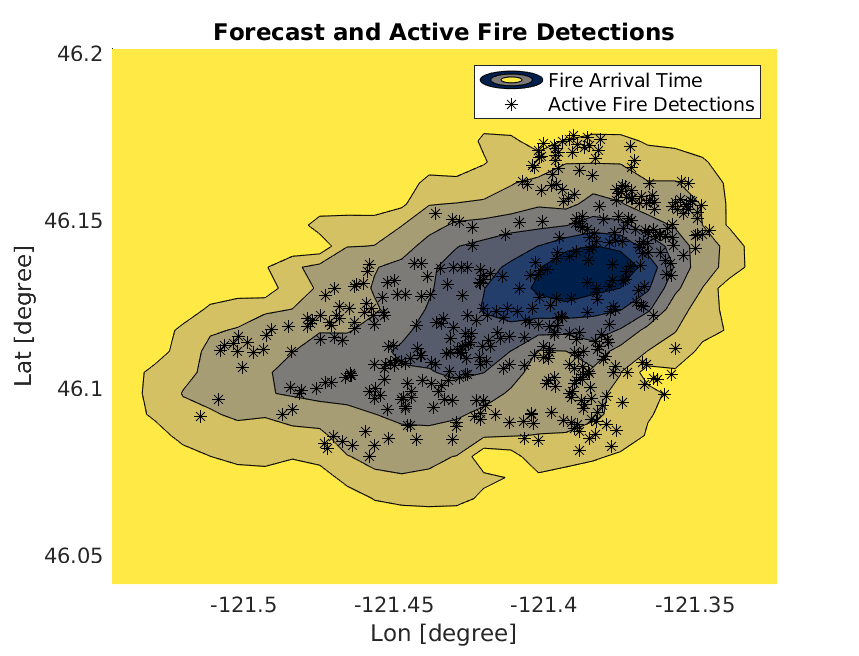}
    \includegraphics[width = 0.45\textwidth]{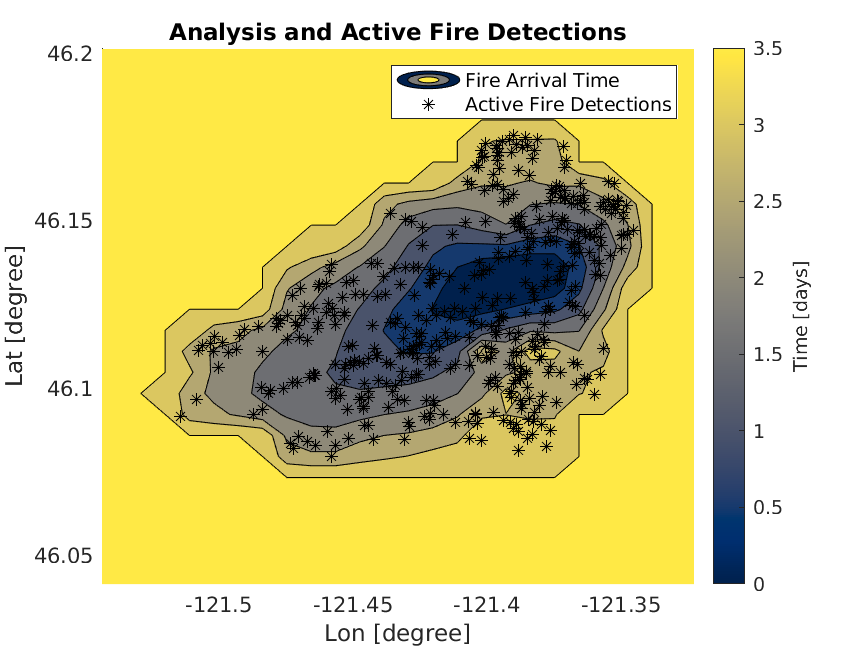}
  \caption{Contours of the forecast the fire arrival time (left) and the analysis (right) for the third cycle of the Cougar Creek Fire simulation. The data assimilation was performed on a computational grid with 250 meter spacing. Minimal adjustment were made to the fire arrival time during the data assimilation at the end of this cycle.}
  \label{fig:cougar_cycle_3}
  \end{center}
\end{figure}

Figures \ref{fig:cougar_cycle_0}-\ref{fig:cougar_cycle_3} show how the data assimilation made adjustments to the forecast fire arrival time. The first two cycles show fairly large adjustments being made to the forecast and the final to cycles show more moderate adjustments. Most likely this fire exhibited some uneven growth due to weather or fuel conditions the model was unable to account for. As was done in the case of the Patch Springs fire, each forecast from the cycles of the simulation was allowed to run for the entire duration of the simulation period. These forecasts were compared with an infrared perimeter observation made on August 15 at 10:33 UTC. The numeric results are summarized in Table \ref{tbl:cougar_creek_moe} and graphically represented in Figure \ref{fig:cougar_moe_cycles}. Each forecast from the cycles produced a better result, but most of the forecasts tended to overestimate the size of the fire. 

For comparison, the fire was also simulated by initializing the model with satellite detection data. Only 9 hours of forecast were produced by this simulation and it can be compared to the short forecast made as part of the cycling routine. The results of these two forecasts are quite similar. Most likely, the reason for the overestimation of the fire area by these forecasts was due to the fuel moisture content used by the model being too dry. Section \ref{sec:cougar_fmc_run} shows results from a simulation that used an adjusted FMC to achieve better results.

Another possible contributing factor to the overestimation of fire size when using data assimilation is related to the replay of the analysis fire arrival time during the restart of the fire simulation for the next forecast cycle. Although the analysis does provide an improved estimate of the state of the fire in terms of the fire arrival time, it does not necessarily give an accurate picture of where the fire is hottest and growing most rapidly. For example, in the simulation of the Cougar Creek fire, during the second forecast cycle, the forecast did not predict the growth in the southwest region of the fire, as is seen in Figure \ref{fig:cougar_cycle_1}. Apparently, the growth of the fire was mostly in a southwesterly direction during this time, with little growth appearing elsewhere. However, the data assimilation strategy treats all parts of the fire perimeter as if equal fire growth was occurring everywhere. This equal growth can be seen in the forecast produced in the third cycle of the Cougar Creek fire, shown in Figure \ref{fig:cougar_cycle_2}. In this forecast, the fire had continued to expand towards the northwest and southeast, despite some knowledge that most active are of the fire was in the southwest. Possibly this shortcoming of the data assimilation method could be overcome by adjusting the properties of the underlying fuels in the areas where the data assimilation has contracted the fire perimeter, indication the fire was less active in the region than the model had predicted.

\begin{table}
\begin{tabular}{|c|c|c|c|c|c|}
\hline 
Cycle & Hours of Forecast & MOE X  & MOE Y & ||MOE|| & Sorenson \\ 
\hline 
0 & 105 & 0.6638 & 0.6647 & 0.9394 & 0.6642 \\ 
\hline 
1 & 81 & 0.9787 & 0.4966 & 1.097 & 0.6589 \\ 
\hline 
2 & 57 & 0.98733 & 0.4773 & 1.097 & 0.6435 \\ 
\hline 
3 & 33 & 0.9580 & 0.5446 & 1.1019 & 0.6944 \\ 
\hline 
4 & 9 &  0.9533 & 0.6284 & 1.1418 & 0.7575 \\ 
\hline 
Estimated Start& 9 & 0.9560 & 0.6177 & 1.1382  & 0.7505  \\ 
\hline 
\end{tabular}
\caption{Scores for the cycling of the Cougar Creek fire when compared with a perimeter observation made August 15,  at 10:33 UTC.}
\label{tbl:cougar_creek_moe}
\end{table}

\subsubsection{Camp Fire}
\begin{figure}[!h]
\begin{center}
  \includegraphics[width = 0.45\textwidth]{camp_fire_paths.png}
  \includegraphics[width = 0.45\textwidth]{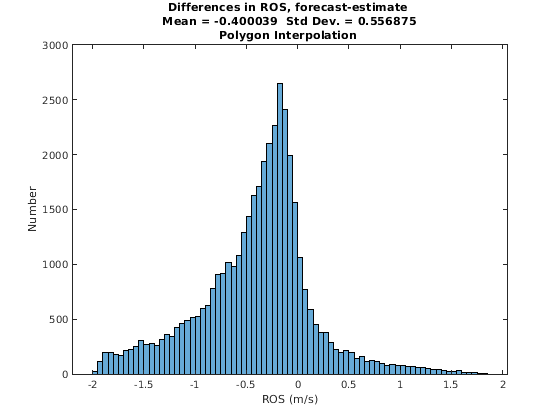}
  \caption[Two figures showing the high rate of growth of the Camp Fire during its early stages.]{Two figures showing the high rate of growth of the Camp Fire during its early stages. On the left, the shallow angles of paths in the graph indicate a high ROS. On the right, a histogram of the difference between the ROS forecast by the model and the ROS estimated from satellite data indicate the modeled fire was spreading much slower than real-word fire. }
  \label{fig:camp_fast_growth}
  \end{center}
\end{figure}

\label{{sec:camp_data_assilation}}
The Camp fire was a large and destructive fire that took place in California in 2018.  Drought conditions, dry fuels, and high winds combined to produce a fire that moved 25 kilometers in the first 12 hours \citep{Cal_fire-2018-GSB,Brewer-2020-CFM} and  consumed over 100,000 acres during its first 2 days \citep{Inciweb-2018-Camp}). The left panel of Figure \ref{fig:camp_fast_growth} shows the path structure derived from satellite detections during the first two days. The shallow angles of the paths indicate a high rate of spread was present.  It is thought that a major component of the spread mechanism of this fire was from the wind transport of burning embers from vegetation and man-made structures \citep{Brewer-2020-CFM,Syifa-2020-MPB}. The spot fires initiated by these wind-borne firebrands were observed to occur a mile in advance of the main bulk of the fire \citep{Cal_fire-2018-GSB}). This spread mechanism is not a part of the Rothermel model that WRF-SFIRE uses for computing the fire spread, making data assimilation an essential component of simulating such a fast-moving fire. Indeed, the right panel of Figure \ref{fig:camp_fast_growth} shows a histogram of the difference between the ROS implied by the satellite detections and that of the model forecast at those detection locations. One average, the model was underestimating the ROS by about 0.4 m/s.  

We used WRF-SFIRE to simulate the critical first two days of the Camp Fire. The simulation was run in WRF-SFIRE using three nested domains with resolutions of 5000m, 1667m, 500m. The simulation  used four data assimilation cycles to incorporate  observations from the MODIS and VIIRS active fire products. Figure \ref{fig:camp_schematic} shows the timing of the cycles. For assessment, the cycle forecasts were compared with an infrared perimeter observation made on November 10 at 07:00 UTC. More details of the individual forecast cycles follow.

\begin{itemize}
\item \textbf{Cycle 0} - The simulation was started from an ignition point and run for 43 hours. Satellite data from the first 16 hours of the simulation was collected and combined with the first 16 hours of the simulation forecast, forming the analysis fire arrival time that covered the first 16 hours of the fire simulation.
\item \textbf{Cycle 1} - The analysis fire arrival time from cycle 0 was used to spin-up the model, replaying the first 16 hours of the fire and bringing the state of the atmosphere into synchronization with the state of the fire. After the  spin-up period, the model took over and produced an additional 27 hours of forecast. Satellite data from the first 8 hours of this forecast was collected and combined with the first 8 hours of the simulation forecast, forming the analysis fire arrival time that covered the first 24 hours of the fire simulation.
\item \textbf{Cycle 2} - The analysis fire arrival time from cycle 1 was used to spin-up the model, replaying hours 16 to 24 of the fire. After the spin-up period, the model took over and produced an additional 19 hours of forecast. Satellite data from the first 8 hours of this forecast was collected and combined with the first 8 hours of the simulation forecast, forming the analysis fire arrival time that covered the first 32 hours of the fire simulation.
\item \textbf{Cycle 3} - The analysis fire arrival time from cycle 2 was used to spin-up the model, replaying hours 24 to 32 of the fire. After the spin-up period, the model took over and produced an additional 11 hours of forecast. Satellite data from the first 8 hours of this forecast was collected and combined with the first 8 hours of the simulation forecast, forming the analysis fire arrival time that covered the first 40 hours of the fire simulation.
\item \textbf{Cycle 4} -   The analysis fire arrival time from cycle 2 was used to spin-up the model, replaying hours 32 to 40 of the fire. After the spin-up period, the model took over and produced an additional 3 hours of forecast. 
\end{itemize}

\begin{figure}[!h]
\begin{center}
  \includegraphics[width = 0.45\textwidth]{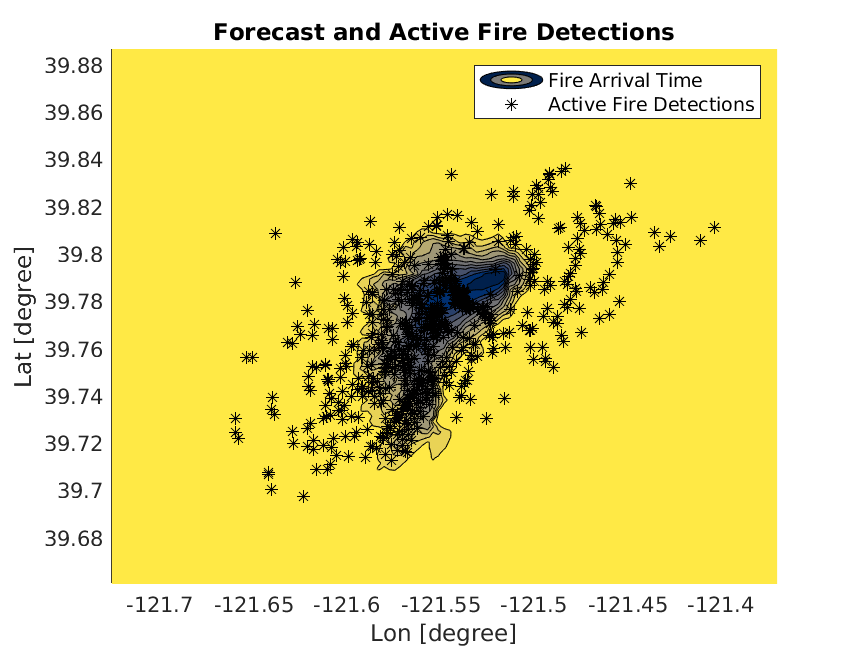}
  \includegraphics[width = 0.45\textwidth]{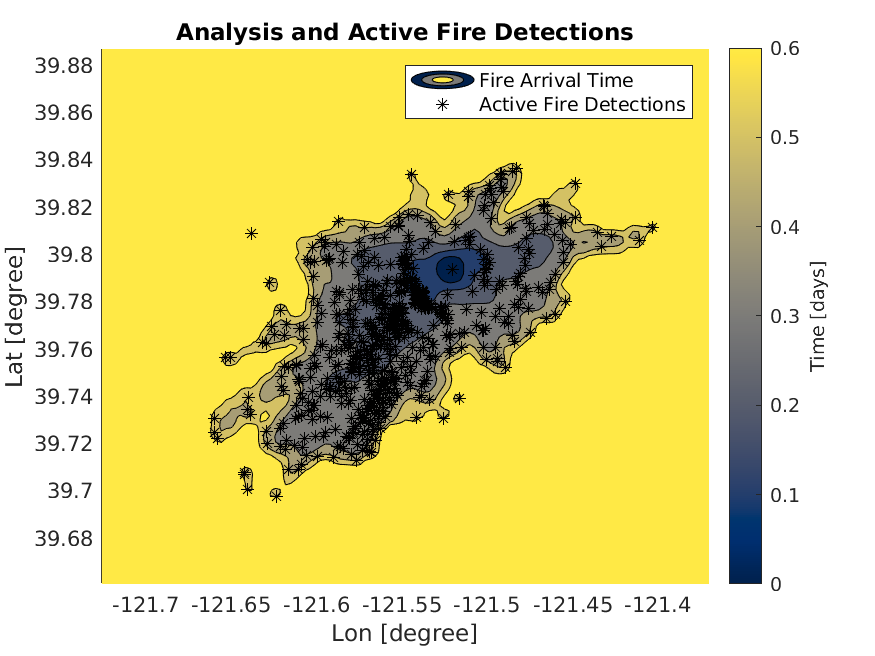}
  \caption{Contours of the forecast the fire arrival time (left) and the analysis (right) for the first cycle of the Camp Fire simulation. The data assimilation was performed on a computational grid with 250 meter spacing. The initial simulation underestimated the size of the fire, causing the analysis fire arrival time to have an expanded outer perimeter when compared to that of the forecast.}
  \label{fig:camp_cycle_0}
  \end{center}
\end{figure}

\begin{figure}[!h]
\begin{center}
  \includegraphics[width = 0.45\textwidth]{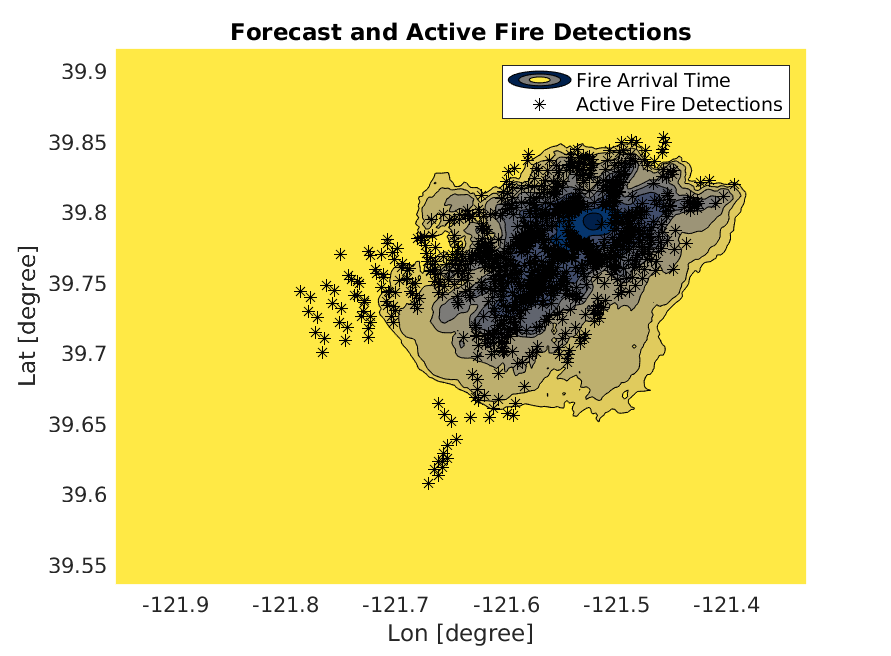}
  \includegraphics[width = 0.45\textwidth]{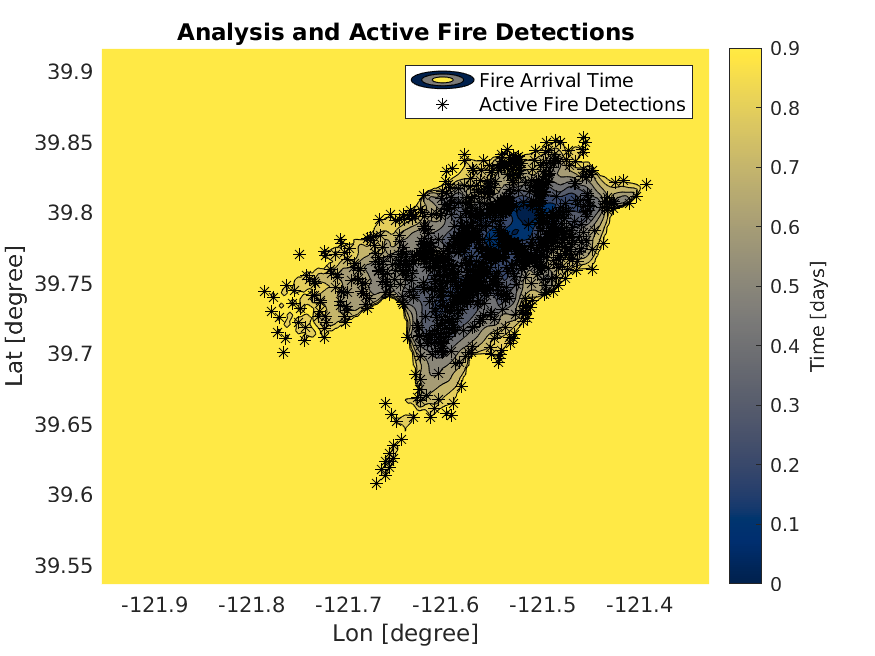}
  \caption{Contours of the forecast the fire arrival time (left) and the analysis (right) for the second cycle of the Camp Fire simulation. The data assimilation was performed on a computational grid with 250 meter spacing.}
  \label{fig:camp_cycle_1}
  \end{center}
\end{figure}

\begin{figure}[!h]
\begin{center}
  \includegraphics[width = 0.45\textwidth]{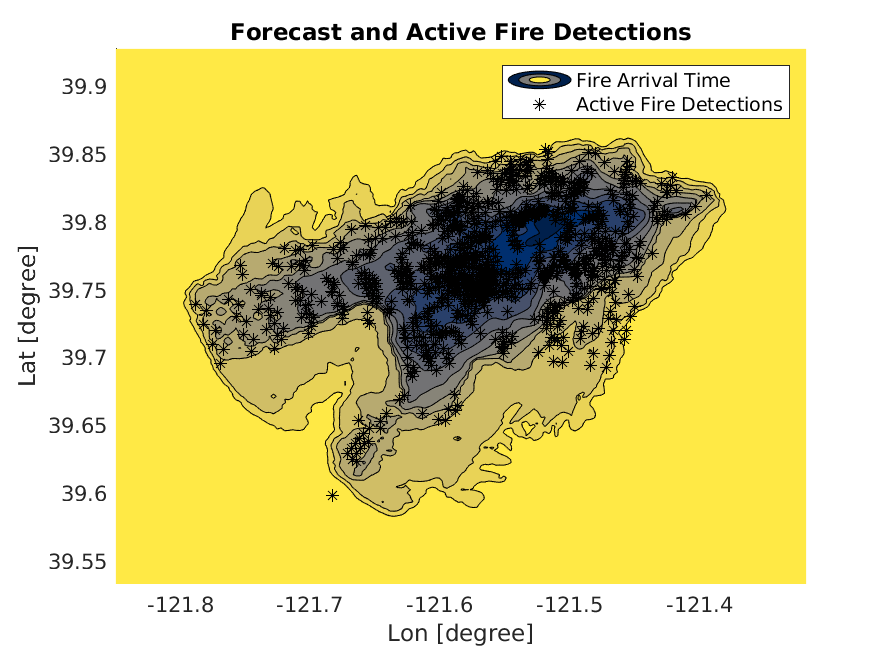}
  \includegraphics[width = 0.45\textwidth]{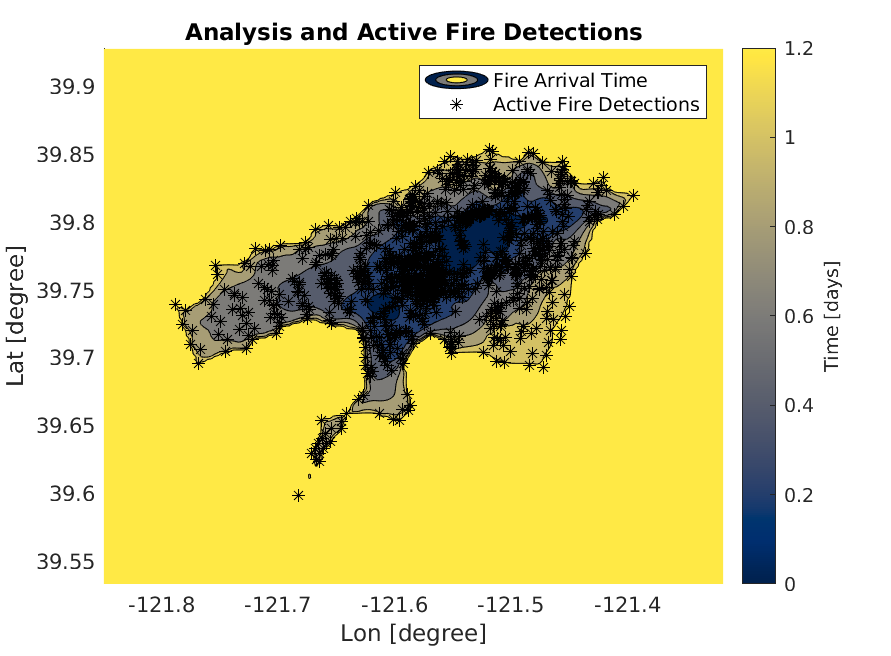}
  \caption{Contours of the forecast the fire arrival time (left) and the analysis (right) for the third cycle of the Camp Fire simulation. The data assimilation was performed on a computational grid with 250 meter spacing. According to satellite data, very little growth in the fire had occurred during this cycle, but the model continued to show expansive growth to the south of the main region of the fire domain. The data assimilation routine contracted the perimeter back to a similar position it was in at the end of the previous cycle.}
  \label{fig:camp_cycle_2}
  \end{center}
\end{figure}

\begin{figure}[!h]
\begin{center}
  \includegraphics[width = 0.45\textwidth]{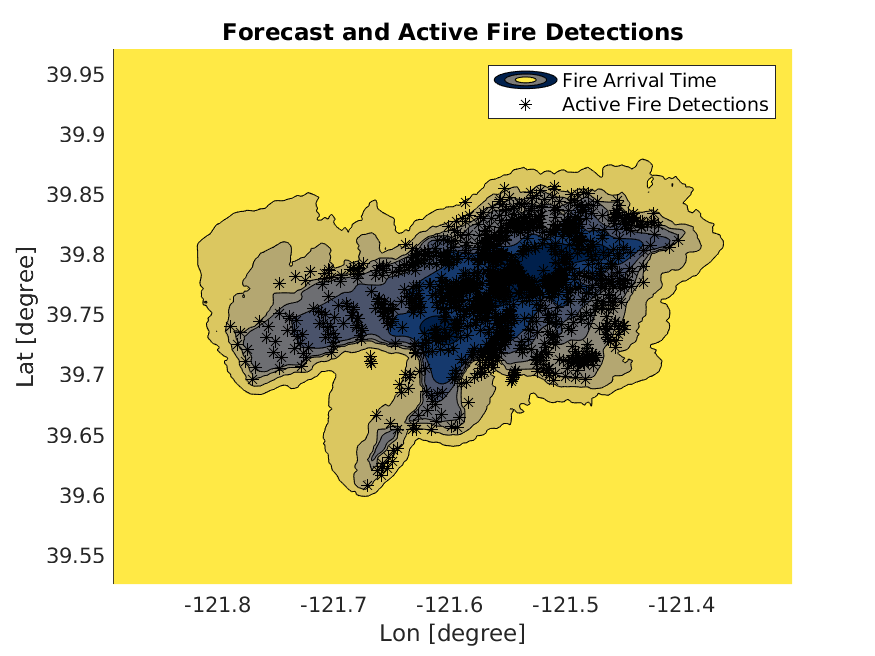}
  \includegraphics[width = 0.45\textwidth]{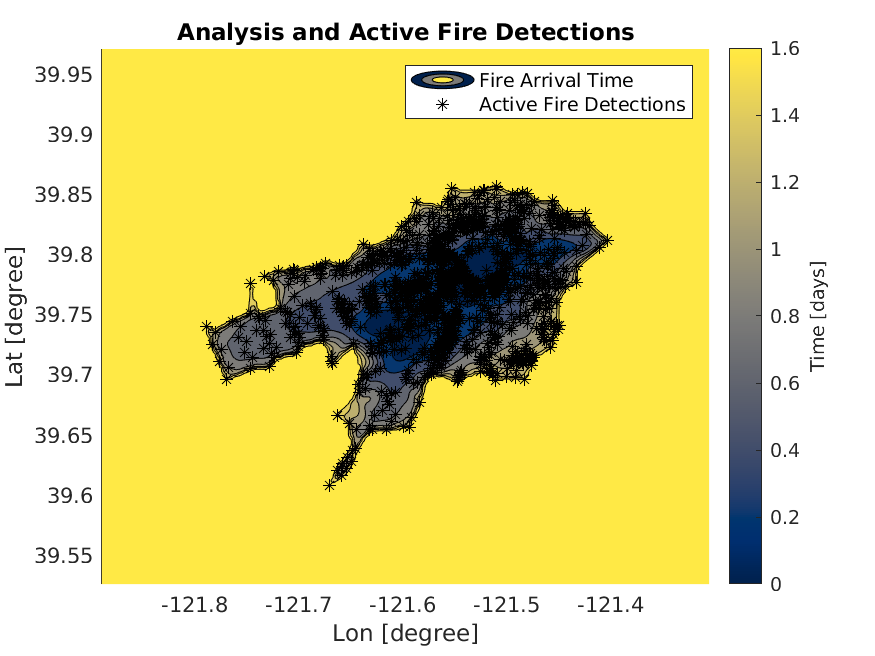}
  \caption{Contours of the forecast the fire arrival time (left) and the analysis (right) for the fourth cycle of the Camp Fire simulation. The data assimilation was performed on a computational grid with 250 meter spacing. In this final cycle, the forecast fire perimeter was again contracted to account for growth no observed by satellite.}
  \label{fig:camp_cycle_3}
  \end{center}
\end{figure}

Figures \ref{fig:camp_cycle_0}-\ref{fig:camp_cycle_3} show the adjustments made to the forecast fire arrival time by data assimilation. The initial cycle, Figure \ref{fig:camp_cycle_0}, shows a massive expansion of the fire perimeter to reflect the large number of active fire detections that extended beyond forecast perimeter. Indeed, this type of adjustment is necessary for a fire that spread largely by winds carrying embers ahead of the main fire front. The WRF-SFIRE model does not have the capability to advance fire by this mechanism and it is not surprising that it could not accurately predict the movement of the fire front during this period of rapid growth. In Figure \ref{fig:camp_cycle_1}, the forecast showed fire growth in the southern part of the fire domain that was not observed by satellite, leading to a contraction of the perimeter here during the data assimilation process. Interestingly, there's a large gap in the satellite fire detections  between two ``forks" of the fire in the south and southwest region of the fire domain. Most likely fire was present in this gap but was not observed. Possibly the region was obscured from satellite view by smoke. Table \ref{tbl:camp_moe} summarizes the results of comparison with the cycling forecasts with an infrared perimeter observation. Figure \ref{fig:camp_moe_cycles} give a graphical representation of these results. The forecasts tended to overestimate the size of the fire.

For comparison, a forecast of 19 hours duration, initialized from an estimated fire arrival time, was also produced. This short-term forecast ended at the same time of the simulations produced during the cycling routine and can be compared with the forecast produced in cycle 2 of the main simulation. As can be seen in the rightmost panels in Figure \ref{fig:camp_moe_cycles}, the results are similar. The simulations both overestimated the size of the fire.

\begin{table}
\centering
\begin{tabular}{|c|c|c|c|c|c|} 
\hline
Cycle    & Hours of Forecast & MOE X  & MOE Y  & \textbar{}\textbar{}MOE\textbar{}\textbar{} & Sorenson  \\ 
\hline
0        & 43                & 0.6111 & 0.8124 & 1.0166                                      & 0.6976    \\ 
\hline
1        & 27                & 0.9709 & 0.4664 & 1.0771                                      & 0.6301    \\ 
\hline
2        & 19                & 1.0000 & 0.4149 & 1.0827                                      & 0.5865    \\ 
\hline
3        & 11                & 0.9975 & 0.5393 & 1.1340                                      & 0.7001    \\ 
\hline
4        & 3                 & 0.9012 & 0.7658 & 1.1826                                      & 0.8283    \\ 
\hline
Estimate & 19                & 1.0000 & 0.4278 & 1.0877                                      & 0.5993    \\
\hline
\end{tabular}
\caption[Assessments of the simulation of the Camp Fire. Each cycle was run until November 10, 07:00 UTC so that a comparison with an infrared perimeter observation could be made.]{Assessments of the simulation of the Camp Fire. Each cycle was run until November 10, 07:00 UTC so that a comparison with an infrared perimeter observation could be made. Even very short forecasts had a difficult time with predicting the behavior of this fire. In general, data assimilation helped improve the model output, not as much as could be expected. The final row in the table represents the a simulation initialized from an estimated fire arrival time produced from satellite fire data. The output from this simulation is very similar to that of the cycle 2 simulation. Figure \ref{fig:camp_moe_cycles} gives a set of graphics that help visualize these results.}
\label{tbl:camp_moe}
\end{table}

\subsection{48-Hour Forecast Workflow}
\label{sec:48_hour}
This forecast strategy works by assimilating the most recent satellite data to update a short-term forecast initialized from an estimated fire arrival time made using infrared perimeter observations. The goal is to start a simulation from the best possible estimate of the state of the fire, derived from infrared perimeter observations, and then use the fire model and additional satellite fire observations to make minimal corrections to a short-term forecast. The model is then restarted from this updated estimate of the state of the fire and allowed to run for 48 hours. 

Typically, infrared perimeter observations are made by flying an aircraft over the fire around midnight, local time. Because of the accuracy of these observations, they provide the best available snapshot of the state of the fire and can be used to initialize a simulation. These perimeter observations may only become available to researchers sometime in the afternoon of the following day. During the period between when the infrared observations are made and when they become available to fire forecasters, paossibly many hours will have passed and additional satellite observations will have been made. The forecasts strategy in this section details a method to take advantage of these latest satellite observations by making a short-term forecast, initialized from the perimeter observations, and then assimilating the most recent satellite data using the same method previously outlined. The workflow for this forecast strategy is as follows.
\begin{enumerate}
\item At 5:00 PM local time, collect satellite and infrared perimeter data. Typically the infrared perimeter observations made around midnight the previous day become available at this time.
\item Estimate the fire arrival time from the data.
\item Run the model forward until the model time matches real time. This short-term forecast can be run in about two hours real time. 
\item Collect all the satellite data covering the time of the short-term forecast.
\item Make an analysis fire arrival time by using the data assimilation techniques from this chapter.
\item Restart the simulation from the analysis fire arrival time.
\end{enumerate}

\subsubsection{Patch Fire Example}

\begin{figure}[!h]
\begin{center}
  \includegraphics[width = 0.45\textwidth]{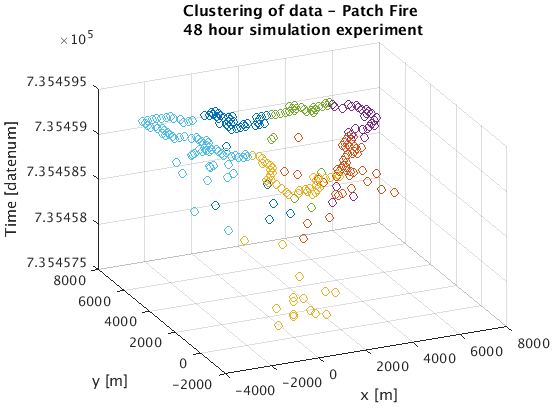}
  \includegraphics[width = 0.45\textwidth]{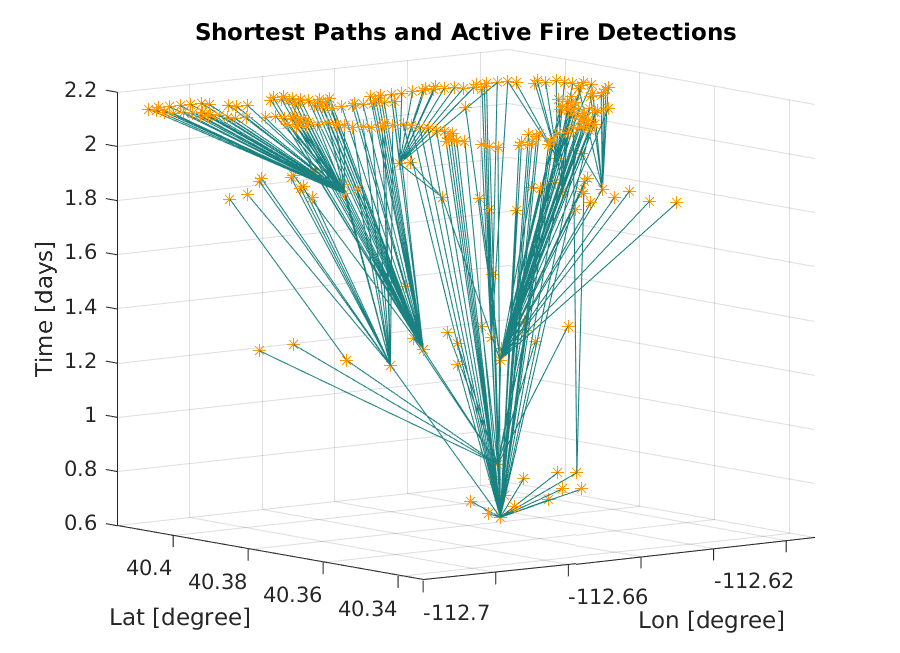}
  \includegraphics[width = 0.45\textwidth]{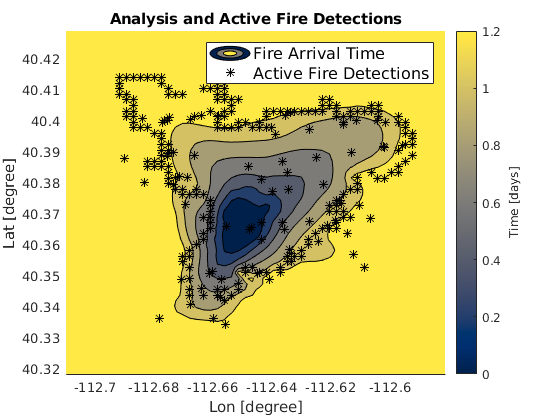}
  \caption{Satellite fire data organized into clusters and paths from the data used to initialize the initial simulation of the 48-hour workflow experiment for the Patch Springs Fire. The upper panels show the clustering and shortest paths. The bottom panel shows the estimated fire arrival time used to initialize the model.}
  \label{fig:patch_48_starts}
  \end{center}
\end{figure}

A simulation of the Patch Springs fire was run using the 48-hour forecast workflow. The simulation was started from an estimated fire arrival time derived from satellite fire data and an infrared perimeter observation. In total, 7 satellite granules and one infrared perimeter were used. The these observations spanned approximately the first two days of the fire. Figure \ref{fig:patch_48_starts} shows these observations and the estimated fore arrival time made for the simulation initialization. The simulation was run forward in time to produce several days of forecast, but in an operational setting, this simulation would produce only about 20 hours of forecast.

\begin{figure}[!h]
\begin{center}
\includegraphics[width = 0.45\textwidth]{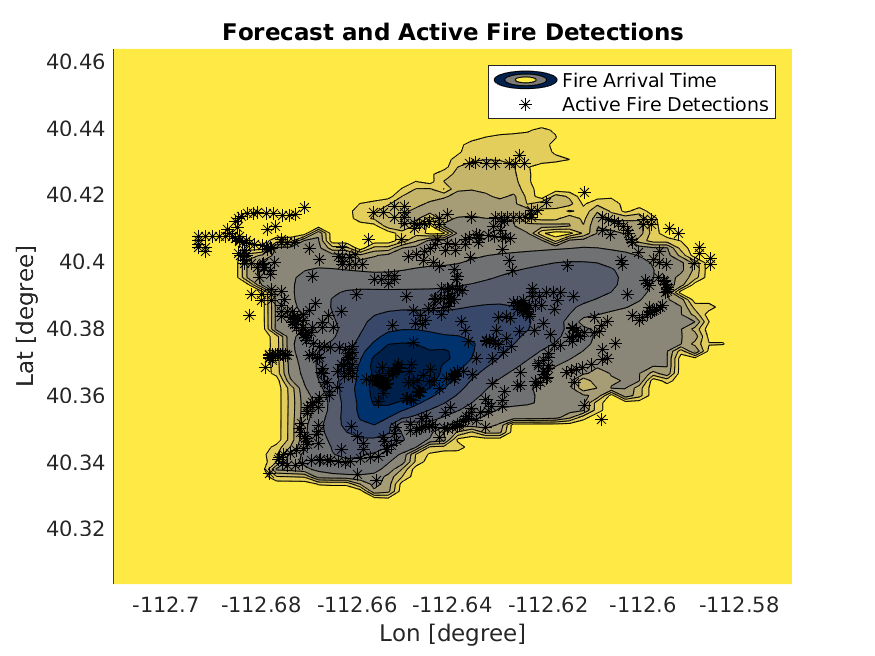}
\includegraphics[width = 0.45\textwidth]{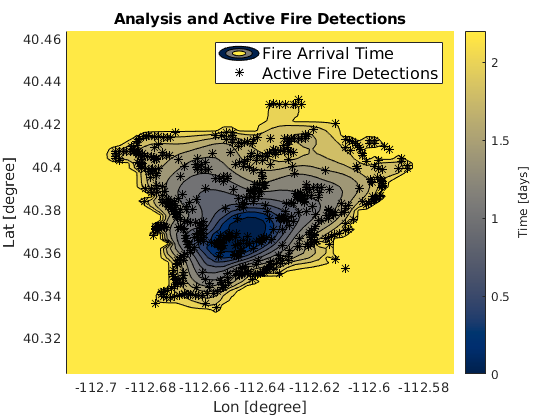}
\includegraphics[width = 0.45\textwidth]{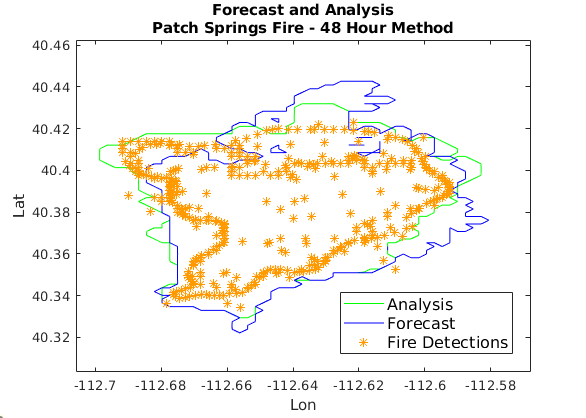}
  \caption[Data assimilation of 21 hours of satellite and perimeter data for the Patch Fire simulation.]{Data assimilation of 21 hours of satellite and perimeter data for the Patch Fire simulation. In the top panels, contours for the forecast and analysis fire arrival times can be seen. In the bottom, the final perimeters of both the forecast and analysis fire arrival times can be seen together. }
  \label{fig:patch_48_restarts}
  \end{center}
\end{figure}

To emulate the forecasting strategy, data assimilation was performed to make minor adjustments to the model output using the latest observations collected in the 21 hours since the time of the previous infrared perimeter. During this window, four granules of satellite data became available. Figure \ref{fig:patch_48_restarts} shows hows the effect of data assimilation. The Analysis has made minor changes to the fire arrival time, contracting the final perimeter over much of the fire domain, but expanding it slightly in northwest region of the fire. The forecast produced after assimilating 21 hours of data showed improvement over a the forecast made from the original initialization of the fire from the perimeter and satellite data when compared to an infrared perimeter. The results of this comparison are presented in Table \ref{tbl:patch_moe} and graphically in Figure \ref{fig:patch_48_moe}. The 48-hour strategy produced better results than the data assimilation cycling strategy from Section  \ref{sec:patch_data_assilation}. The results of this comparison are presented in Table \ref{tbl:patch_conclusion_results}.   

\begin{figure}[!h]
\begin{center}
  \includegraphics[width = 0.7\textwidth]{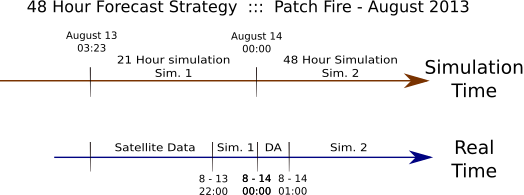}
  \caption[Timelines for the Patch Fire, 48-hour simulation experiment.]{Timelines for the Patch Fire, 48-hour simulation experiment. The orange arrow indicates timings of simulations and the blue arrow indicates the progression of real-time tasks taken to produce and process the simulations in the red timeline.}
  \label{fig:patch_48_timeline}
  \end{center}
\end{figure}

\begin{table}
\centering
\begin{tabular}{|c|c|c|c|c|c|} 
\hline
Cycle & Hours of Forecast & MOE X  & MOE Y  & \textbar{}\textbar{}MOE\textbar{}\textbar{} & Sorenson  \\ 
\hline
0     & 79                & 0.7027 & 0.6609 & 0.9647                                      & 0.6812    \\ 
\hline
1     & 58                & 0.6450 & 0.7559 & 0.9937                                      & 0.6961    \\
\hline
\end{tabular}
\caption[Results from the simulation of Patch Springs Fire using the 48-hour forecast strategy. Assesments were made by comparison was an infrared perimeter observation 10 hours after the the normal 48-hour forecast period.]{Results from the simulation of Patch Springs Fire using the 48-hour forecast strategy. Assesments were made by comparison was an infrared perimeter observation 10 hours after the the normal 48 hour forecast period. The assessment scores show that assimilating even a small number of additional satellite observations helped produce a better forecast.}
\label{tbl:patch_48_results} 
\end{table}

\subsubsection{Cougar Creek Fire Example}
\begin{figure}[!h]
\begin{center}
  \includegraphics[width = 0.45\textwidth]{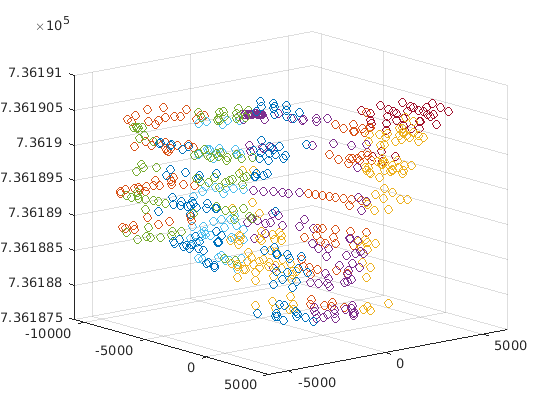}
    \includegraphics[width = 0.45\textwidth]{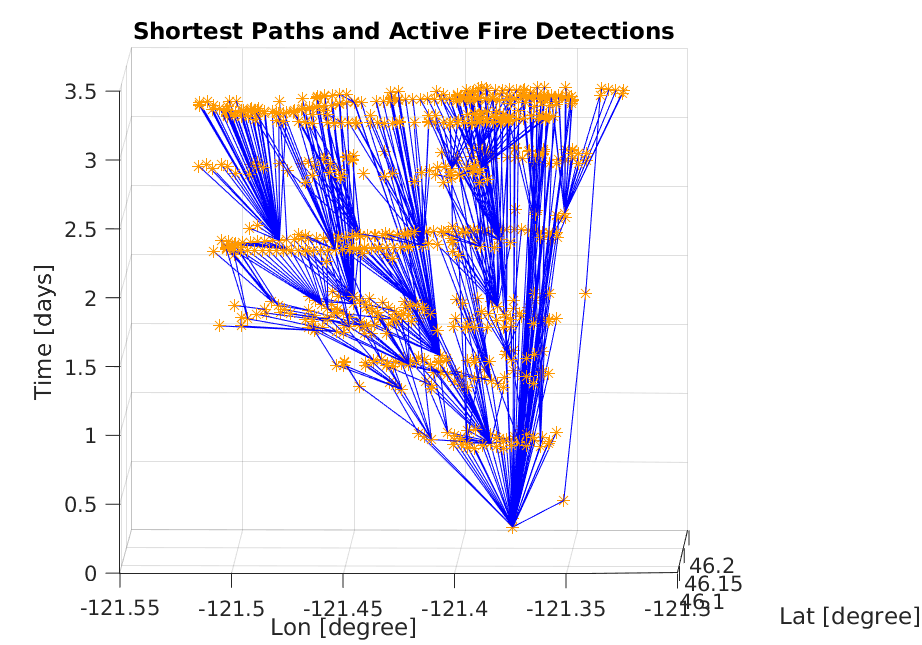}
      \includegraphics[width = 0.45\textwidth]{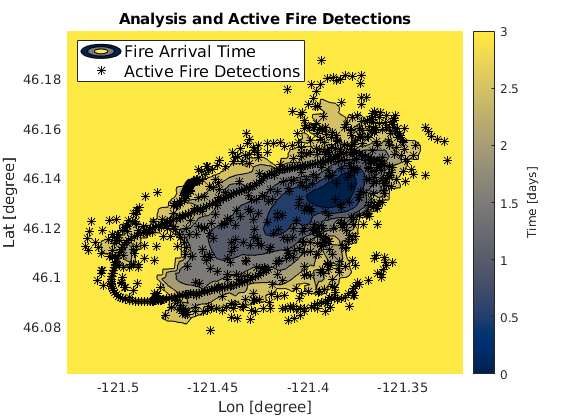}
  \caption{Satellite fire data organized into clusters and paths from the data used to initialize the initial simulation of the 48-hour workflow experiment for the Cougar Creek Fire. The upper panels show the clustering and shortest paths. The bottom panel shows the estimated fire arrival time used to initialize the model.}
  \label{fig:cougar_48_starts}
  \end{center}
\end{figure}

A simulation of the Cougar Creek Fire was run using the 48-hour forecast workflow. The simulation was started from an estimated fire arrival time derived from satellite fire data and an infrared perimeter observation. In total, 24 satellite granules and four infrared perimeters were used. The these observations spanned approximately the first two days of the fire. Figure \ref{fig:cougar_48_starts} shows these observations. The simulation was run forward in time to produce several days of forecast, but in an operational setting, this simulation would produce only about 20 hours of forecast.

\begin{figure}[!h]
\begin{center}
\includegraphics[width = 0.45\textwidth]{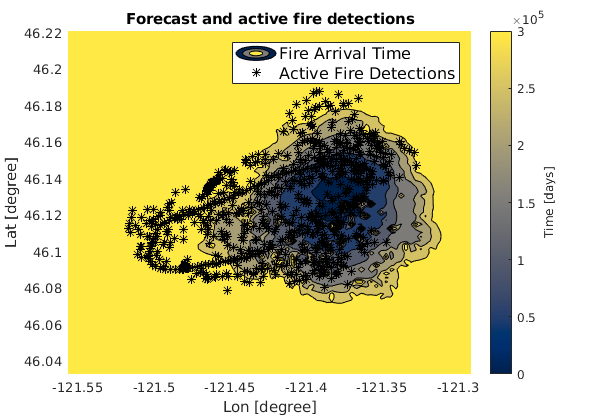}
\includegraphics[width = 0.45\textwidth]{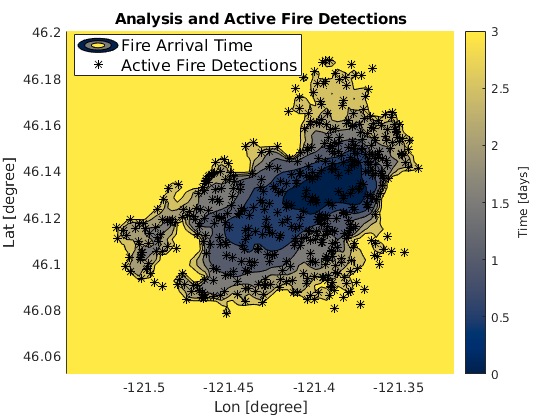}
\includegraphics[width = 0.45\textwidth]{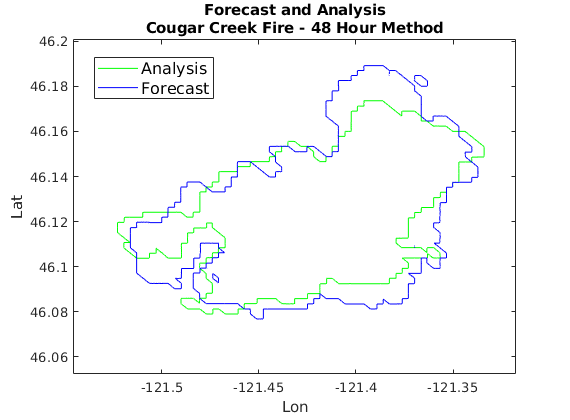}
  \caption[Data assimilation of 21 hours of satellite and perimeter data for the Cougar Creek Fire.]{Data assimilation of 21 hours of satellite and perimeter data for the Cougar Creek Fire. In the top panels, contours for the forecast and analysis fire arrival times can be seen. In the bottom, the final perimeters of both the forecast and analysis fire arrival times can be seen together. }
  \label{fig:cougar_48_restarts}
  \end{center}
\end{figure}

To emulate the forecasting strategy, data assimilation was performed to make minor adjustments to the model output using the latest observations collected in the 16 hours since the time of the previous infrared perimeter. During this window, three granules of satellite data containing 7 active fire detections became available. Figure \ref{fig:cougar_48_restarts} shows the effect of data assimilation. The analysis has made minor changes to the fire arrival time, contracting the final perimeter over much of the fire domain. Table \ref{tbl:cougar_48_results} gives the results of this test. Figure \ref{fig:cougar_48_moe} shows graphical representations of these results. Assessments were made by comparison was an infrared perimeter observation. Comparison between the cycle 0 and cycle 1 scores show that assimilating even a small number of additional satellite observations helped produce a better forecast.

\begin{figure}[!h]
\begin{center}
  \includegraphics[width = 0.7\textwidth]{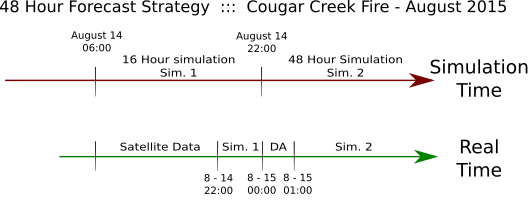}
  \caption{Timeline for Cougar Creek Fire, 48-hour simulation experiment. The orange arrow indicates timings of simulations and the blue arrow indicates the progression of real-time tasks taken to produce and process the simulations in the red timeline.}
  \label{fig:cougar_48_timeline}
  \end{center}
\end{figure}

\begin{table}
\centering
\begin{tabular}{|c|c|c|c|c|c|} 
\hline
Cycle & Hours of Forecast & MOE X  & MOE Y  & \textbar{}\textbar{}MOE\textbar{}\textbar{} & Sorenson  \\ 
\hline
0     & 61                & 0.7316 & 0.8349 & 1.1101                                      & 0.7799    \\ 
\hline
1     & 45                & 0.7305 & 0.8722 & 1.1377                                      & 0.7972    \\
\hline
\end{tabular}
\caption[Results from the simulation of Cougar Creek Fire using the 48-hour forecast strategy.]{Results from the simulation of Cougar Creek Fire using the 48-hour forecast strategy. Assessments were made by comparison was an infrared perimeter observation. The assessment scores show that assimilating even a small number of additional satellite observations helped produce a better forecast. Figure \ref{fig:cougar_48_moe} shows graphical representations of these results.}
\label{tbl:cougar_48_results} 
\end{table}

\subsection{Comparison of the Strategies}

Each forecast strategy outlined in this chapter was evaluated by comparing a final forecast perimeter with an infrared perimeter observation using the MOE and the S{\o}renson index. For simplicity, a score for each strategy was computed from the sum of these measures as

\begin{equation}
\text{SCORE} = \text{MOE X}+\text{MOE Y}+S.
\label{eq:strategy_score}
\end{equation}
where $\text{MOE X}$ and $\text{MOE Y}$ are the two components of the measure of effectiveness describing decreasing false negatives and false positives, respectively, and $S$ is the S{\o}renson index.

Additionally, for the purpose of comparison, evaluation of the initial forecast produced in the data assimilation cycling strategy was also evaluated. These forecasts used no data assimilation and serve as a baseline for comparison. The chief aim for data assimilation is to produce better forecasts than would otherwise be possible. That aim was met, as both the data assimilation cycling and the 48-hour forecast strategies produced better results. The results are summarized in Table \ref{tbl:operational_results}, according to Equation \ref{eq:strategy_score}.

\begin{table}
\centering
\begin{tabular}{|c|c|c|} 
\hline
Forecast Strategy         & Patch Springs & Cougar Creek  \\ 
\hline
Single Forecast           & 1.8940        & 1.9927        \\ 
\hline
Data Assimilation         & 2.0217        & 2.1081        \\ 
\hline
48-Hour Strategy          & 2.0970        & 2.3999        \\
\hline
\end{tabular}
\caption[Overall results of simulation strategies for Patch Springs and Cougar Creek Fire simulations.]{Overall results of simulation strategies for the Patch Springs and Cougar Creek Fire simulations. The scores were computed according to Equation \ref{eq:strategy_score}. For both fires, the 48-hour forecast strategy produced the best results. Using a single simulation produced the worst results. }
\label{tbl:operational_results}
\end{table}



\clearpage
\FloatBarrier

\chapter{\uppercase{Adjusting Fuel Moisture Content in the Model}} \label{chapter:05}
%

\section{Modifying Fuel Moisture Content}\label{sec:FMC}
Fire behavior is affected by weather, terrain, and the properties of the available fuels. If the fire model has accurate terrain maps and the weather component of the coupled model supplies accurate forecasts, improving the fidelity of the fuel properties used by the model is a plausible strategy  to produce better fire forecasts. In this section we will discuss a  method to improve the estimate of the fuel moisture content of the fuels in the domain of a fire simulation. The method  works by comparing the ROS that is computed from an estimated fire arrival time with the the ROS computed from the forecast fire arrival time. An adjustment to the fuel moisture content (FMC) is made  so that the ROS of the forecast fire arrival time better matches the ROS of the estimated fire arrival  time that  has been made with the use of satellite fire detections. 

The rate of spread in the model can be adjusted by various means. The WRF-SFIRE model has several parameters that can be changed to affect the way the factors like wind and terrain influence the ROS. Adjusting the ROS in this manner would be similar to using the multipliers in the equations determined by fire personnel at the fire front as was done by \citet{Rothermel-1983-FPV} and \citet{Finney-1998-FFA}. Another method for making adjustments to the ROS in the model in \citet{Cardil-2019-ARS} for the Wildfire Analyst software. Changes are made based on comparison of the forecast fire arrival time in specific locations where the fire has been observed. The method developed here is similar in the sense that comparison of model output and observational data is used to make affect the change of the ROS. However, the method presented here to change the ROS differs. The adjustment factors available in WRF-SFIRE are not changed; instead, the the fuel moisture content in the input files used by the WRF-SFIRE is adjusted. By adjusting the FMC, it is possible to make the ROS faster or slower, as required. For example, drier fuels will make fires grow faster than comparatively wetter fuels. 

Of particular concern is the FMC of dead fuels found in the fire domain. The FMC of these fuels is influenced greatly by the weather \citep{Mandel-2014-RAA}. These dead fuels are organized into classes defined by the time it takes for 63\% of the difference of the FMC and an equilibrium FMC (that depends on the weather) to vanish. Thus ``1 hour" fuels are small-sized fuels like dry grass or pine needles that react to the changes in the weather relatively quickly. The ``100 hour" fuels are the size of smaller tree branches, up to 10 cm in diameter, and react to changes in the weather more slowly. The fuel moisture estimates used in the WRF-SFIRE system are the product a data assimilation method developed in \citet{Vejmelka-2016-DAD}. %

\subsection{Model sensitivity to changes in the FMC}
\begin{figure}[!h]
\begin{center}
\includegraphics[width = 0.45\textwidth]{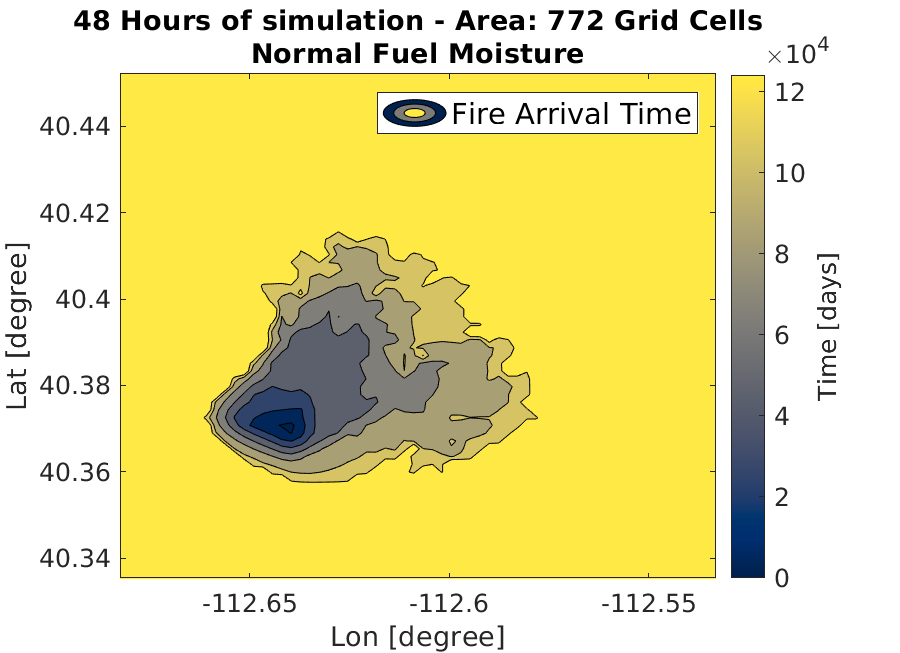}
\includegraphics[width = 0.45\textwidth]{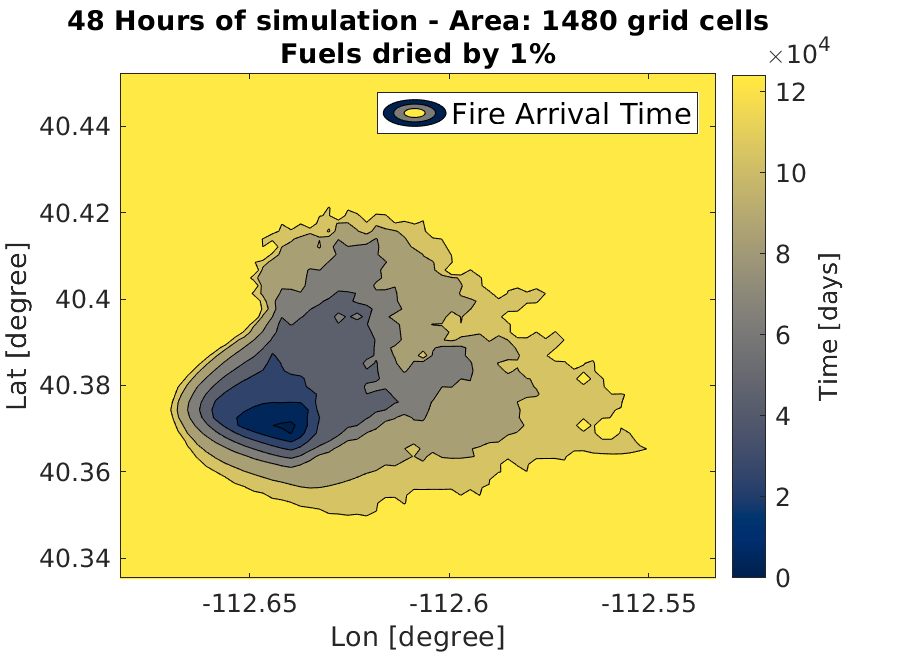}
        
\includegraphics[width = 0.45\textwidth]{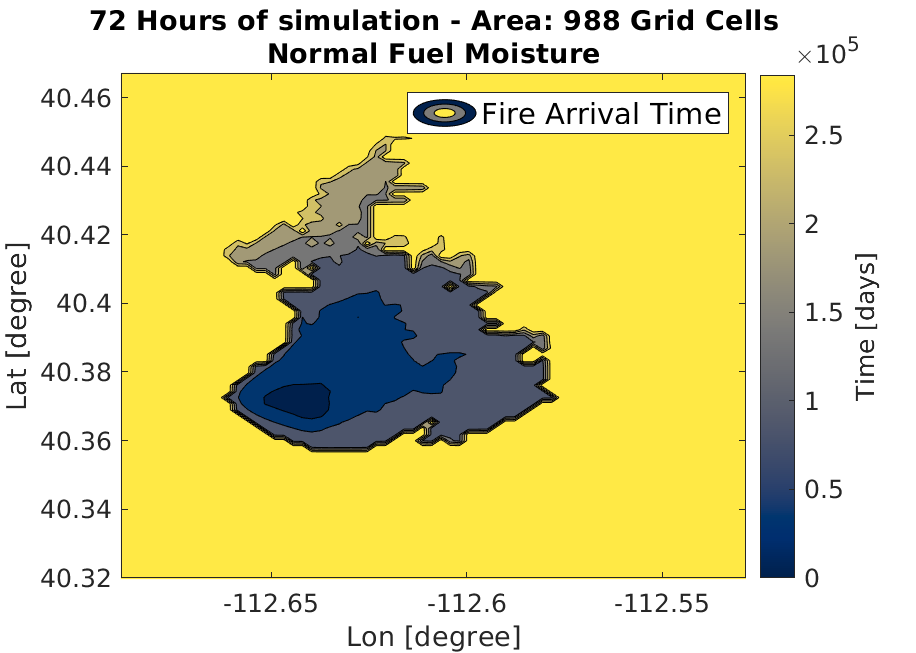}
\includegraphics[width = 0.45\textwidth]{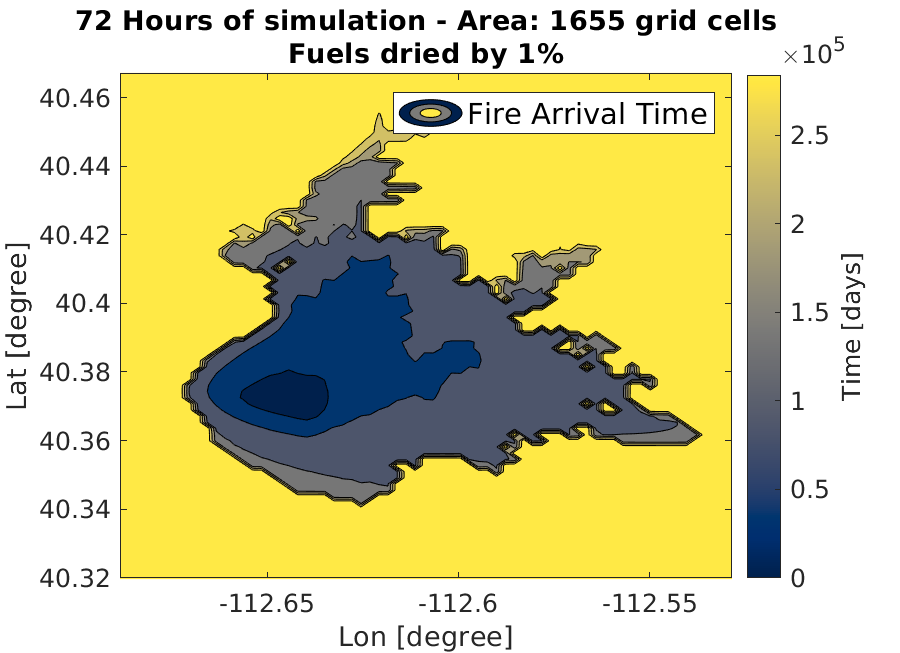}

    \caption[Effect of changing the FMC of the model output. Two simulations were run with identical initial conditions with the exception of the fuel moisture content.]{Effect of changing the FMC of the model output. Two simulations were run with identical initial conditions with the exception of the fuel moisture content. One the left are contours of the fire arrival time for the initial simulation. On the right are the contours of the fire arrival time for a simulation that had decreased the fuel moisture content by 1\% in all fuel categories. The drier fuels resulted in a fire whose area was nearly twice the size of the fire in the initial simulation.}
    \label{fig:patch_ros_sensitivity}
    \end{center}
\end{figure}

Even small changes in the fuel moisture content can lead to large changes in the behavior of the modeled fire. Therefore the proposed method takes a conservative approach to adjusting the FMC. To demosntarte the effect of changes to the FMC in the model, two simulations of the Patch Springs Fire were made using WRF-SFIRE, each having identical starting conditions except for the FMC. One simulation had a fuel moisture content 1\% drier than the other. The results for the fire simulations are shown Figure \ref{fig:patch_ros_sensitivity}. The figure shows the area of the fire forecasts after 48 and 72 hours of simulation time. These are simulation times that could be used in an operational setting. It can be seen that a change in the initial FMC has a large effect on the eventual size of the fire. Drying the fuel out by one percent resulted in a simulated fire that was nearly twice the size of the simulation using wetter fuels.

\begin{table}
\begin{tabular}{|c|c|c|}
\hline 
Fuel class & Base FMC & Revised FMC \\ 
\hline 
1 hr & 0.040 & 0.030 \\ 
\hline 
10 hr & 0.060 & 0.050 \\ 
\hline 
100 hr & 0.045 & 0.035 \\ 
\hline 
1000 hr & 0.050 & 0.040 \\ 
\hline 
Live Fuel & 0.270 & 0.26 \\ 
\hline 
\end{tabular}
\caption{Changes to the FMC for an experiment to show model sensitivity to changes in FMC. The left column shows the fuel class and the center and right columns show the initial and revised FMC contents.}
\label{tbl:patch_fmc_sensitivity}
\end{table}

\subsection{Method}
The proposed method for finding an adjustment to the FMC is simple and should be considered more of an initial  proof of concept rather than established an algorithm. The ROS can be influenced by many factors and this method works by comparing a forecasted ROS and a ROS estimated from satellite data. It is an open question how to best change the FMC in a way that accounts for the difference in these two rates of spread. Therefore, the method makes small, conservative changes only. Changes are made in the course of a simulation routine that uses data assimilation cycling. If a change to the fuel moisture is made, its effect will occur during the forecast period of the cycle following the data assimilation. The adjustment of the FMC is accomplished by the following steps.

\begin{figure}[!ht]
\centering
\includegraphics[width=0.45\textwidth]{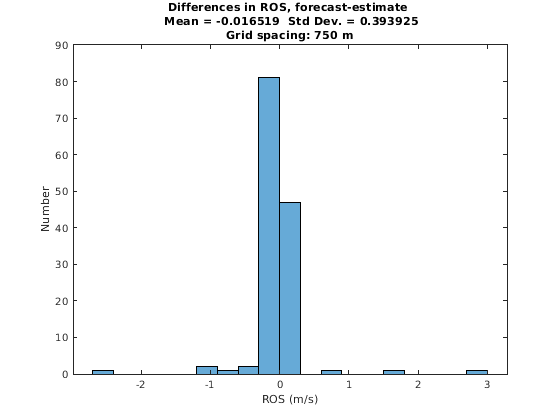}
       \hfill
\includegraphics[width = 0.49\textwidth]{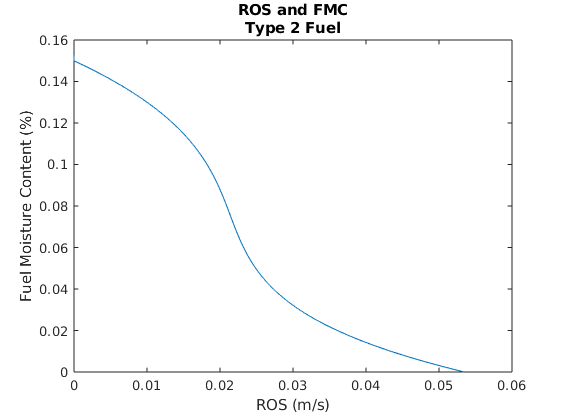}
     \caption[Tools for adjusting the FMC in the simulation.]{Tools for adjusting the FMC in the simulation. On the left, a histogram of the differences in the model ROS and the ROS implied by the data. The forecast was underestimating the ROS. On the right, curves giving the relationship between fuel moisture content and ROS can be used to adjust the fuel moisture content for the next simulation cycle.}
\label{fig:ros_diff_histogram}
\end{figure}

\begin{figure}[!h]
  \begin{center}
    \includegraphics[width = 0.45\textwidth]{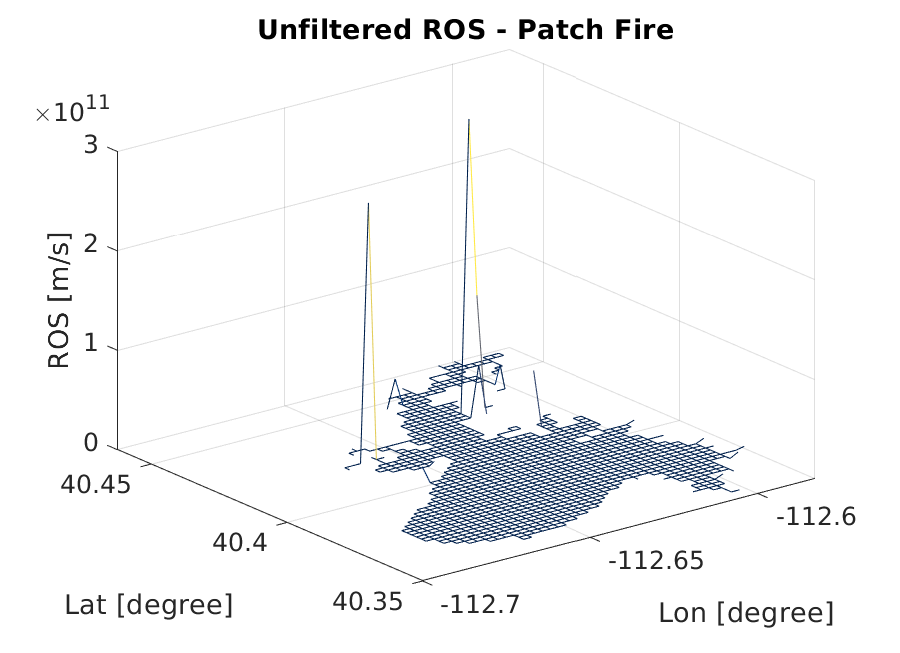}
    \includegraphics[width = 0.45\textwidth]{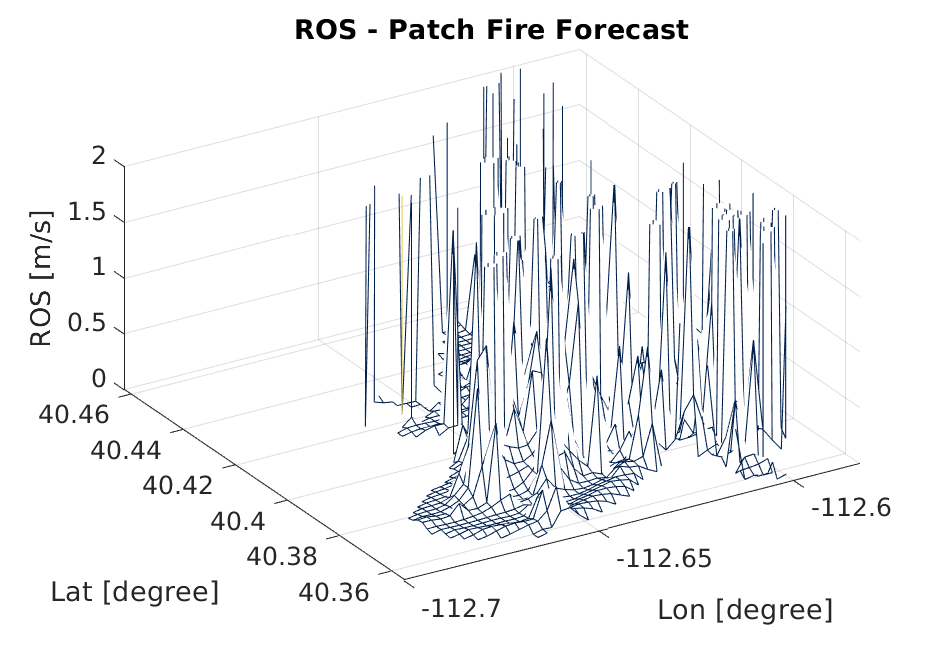}
    \includegraphics[width = 0.45\textwidth]{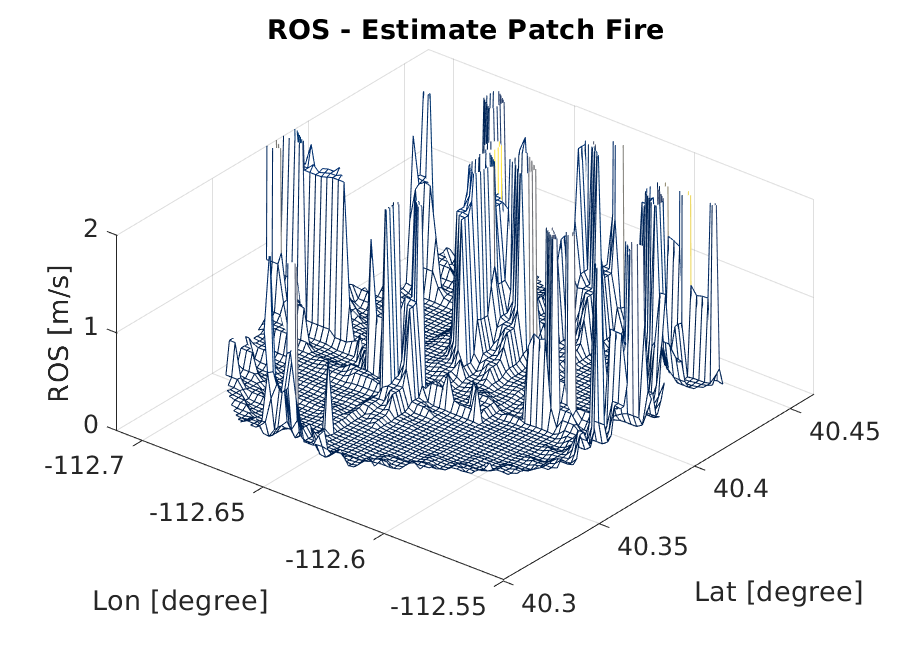}
    \caption[The ROS seen spatially in the fire domain.]{The ROS seen spatially in the fire domain. In  the upper left, several large spikes of the ROS in forecast fire arrival time can be seen. For the estimation of changes in the FMC, only ROS less than 2 m/s are considered. The upper right and bottom panel show the location of these lower rates of spread in the forecast and estimate from satellite data, respectively.}
    \label{fig:patch_ros_views}
    \end{center}
\end{figure}

\begin{enumerate}
\item Compute the final fire areas within the estimated and forecast fire arrival time perimeters. 
\item Compute the ROS in the fire domain from the gradient of the estimated fire arrival time and the forecast fire arrival time.
\item Consider only ROS less than 2 m/s. When computing the ROS from the gradient of the fire arrival time, it is typical that in many places the ROS will be greater than 2 m/s. These are very fast and not consistent with fires the spread by a mechanism modeled by WRF-SFIRE, so we only consider a ROS below the 2 m/s threshold. Additionally, when this threshold is not enforced, the presence of high rates of spread, as computed from the gradient of the fire arrival time, act as statistical outliers that can have a large effect on the average ROS. The upper left panel of Figure \ref{fig:patch_ros_views} shows a surface plot of the ROS containing locations in the fire domain where the computed ROS was more that $2\times 10^11$ m/s.
\item Compute the mean difference in the ROS for forecast and estimate fire arrival time in areas where both fires showed burning took place.
\item  There should be agreement between observations about the area of the fire and the differences in ROS to suggest making a change to the FMC. For example, if the the forecast area is smaller than the estimate area but the the estimated ROS in the forecast is greater than the estimated ROS, no changes should be made.
\item Collect fuel information about the fire domain and
find the inverse function for the burn curve in the primary fuel type found in along the paths in the graph. In this research, the inverse function was found by making a cubic spline of the function seen in Figure \ref{fig:ros_diff_histogram}.
\item Compute adjustment of FMC so that the mean ROS for the forecast matches that of the estimate, according to the burn curve.
\item Adjust the fuel moisture content used by the model. In WRF-SFIRE, this means adjusting the variable \texttt{FMC\_GC} in the input file used for the next cycle.
\end{enumerate}

\begin{figure}[!ht]
\centering
       \includegraphics[width=0.45\textwidth]{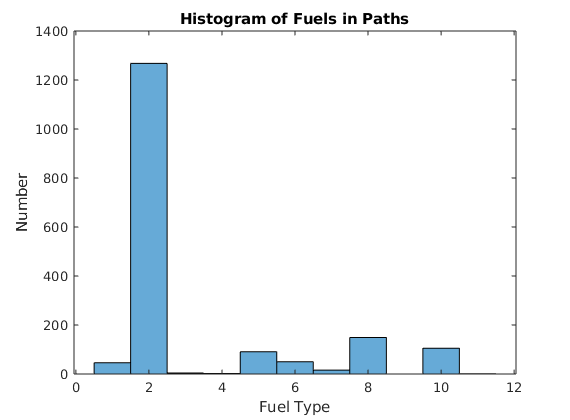}
     \hfill
       \includegraphics[width=0.45\textwidth]{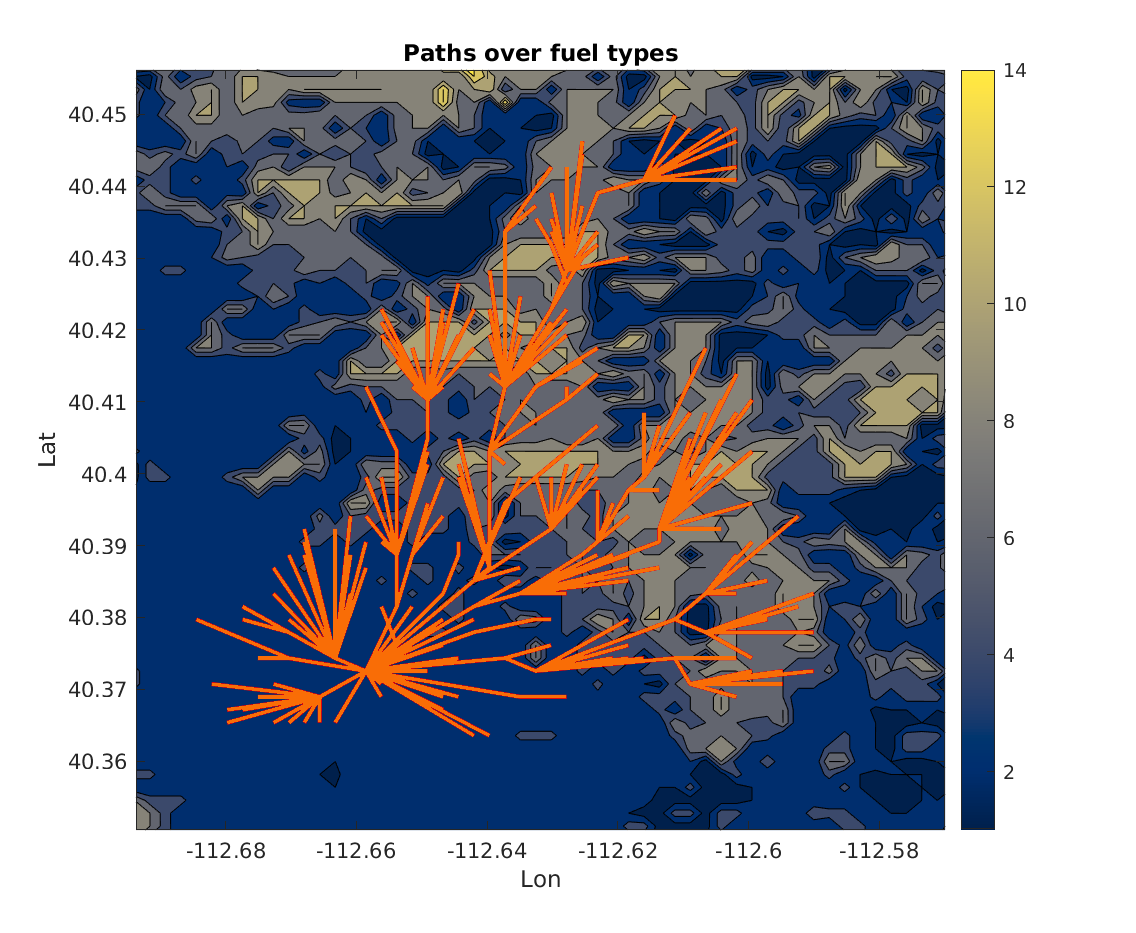}
     \caption[Fuels underlying the shortest paths in the detection graph for the Patch fire simulation.]{Fuels underlying the shortest paths in the detection graph for the Patch fire simulation. On the left, we see that the majority of the fire took place in the type 2 fuels. On the right, the paths though the detections placed on a map of the underlying fuels are seen.  The color bar indicates the fuel the type found at each location.}
\label{fig:patch_fuels}
\end{figure}

\subsubsection{ROS of the Estimate and Forecast}

\subsection{Examples of Forecasts Using Adjustment of Fuel Moisture Content}
Simulations of the Patch Springs and Cougar Creek fires were made using a workflow and starting conditions that were identical to those detailed in Chapter \ref{chapter:04}, with the exception that adjustments of the FMC were made during the data assimilation phase of each cycle. The results were mixed, in the case of the Patch Springs Fire, the assessment of the simulation with 58 hours of forecast time shows a result that was a little worse than that of a simulation where the FMC was not changed at any stage. The simulation of the Cougar Creek Fire using adjustment of the FMC was a little better than the simulation where no changes were made to the FMC. The following sections detail the experiments and show how these conclusions were derived.

\subsubsection{Patch Fire Simulation with FMC Adjustments}

\begin{figure}[!ht]
\centering
\includegraphics[width=0.45\textwidth]{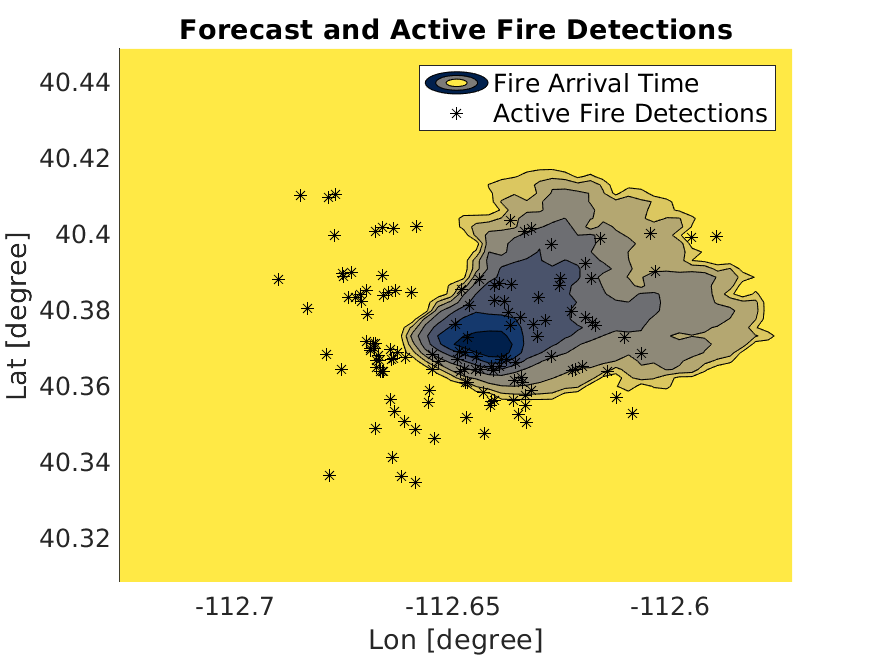}
\includegraphics[width=0.45\textwidth]{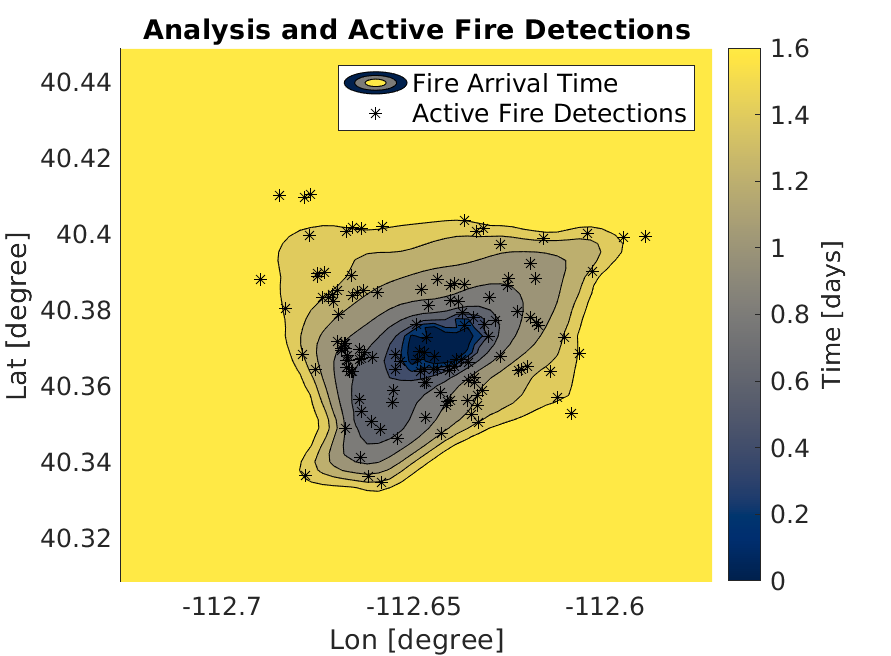}
\includegraphics[width=0.45\textwidth]{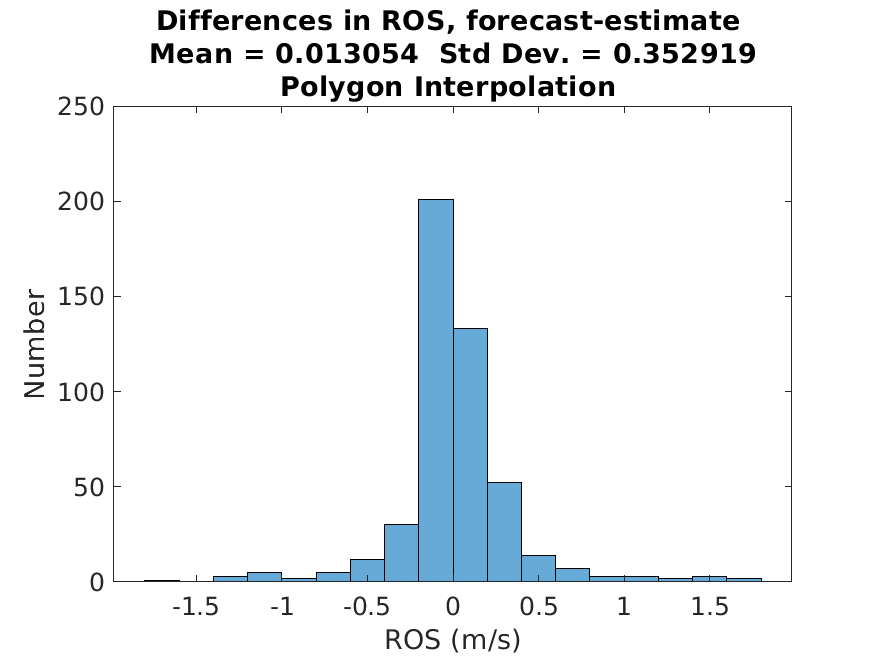}
     \caption[Cycle 0 of the Patch Fire simulation with adjustment of FMC]{Cycle 0 of the Patch Fire simulation with adjustment of FMC. In the upper left and right are contours of of the fire arrival time for the forecast and the analysis, respectively. On the bottom is the histogram showing the difference between the ROS at mesh points in the fire domain. The area of the forecast is smaller, but the ROS was higher in the forecast so that no changes to the FMC were made for the next simulation cycle.}
\label{fig:patch_fmc_cycle_0}
\end{figure}

A simulation of the Patch Springs fire was run using the cycling strategy as in Section \ref{sec:patch_data_assilation}, except that a change to the FMC was made during the data assimilation phase, according to the procedure outlined above. For simplicity, if a change to the FMC was suggested by the algorithm, it was applied to all fuels in the entire fire domain. Figure \ref{fig:patch_fmc_cycle_0} and Figures \ref{fig:patch_fmc_cycle_2}-\ref{fig:patch_fmc_cycle_4} show how the data assimilation made adjustment to the fire arrival time during the simulation cycles as well as a histogram showing how the ROS in the forecast differed from the estimated ROS. Table \ref{tbl:patch_fmc} shows a comparison of the forecast and estimated fire areas as well as the average ROS in the forecast and estimated fire arrival time for each cycle. When any adjustment to the FMC was made by the algorithm, that adjustment is also indicated.

Assessment of the simulation was made by comparison with and infrared perimeter and the results are summarized in Table \ref{tbl:patch_sim_fmc_moe_results}. Figure \ref{fig:patch_moe_fmc_cycles} gives a graphical representation of these results. When looking at the results from producing 58 hours of forecast, after the FMC adjustment,  the results are a little bit worse compared to using data assimilation without changing the FMC. The increase in the FMC caused the estimated fire area to be smaller than it should have been. However, as the simulation progressed, the scores for later cycles eventually became higher than those for the simulation without changes made to the FMC. It is likely that the adjustments made to the 100-hour fuels in the fire domain played a part of this improvement.

\begin{table}
\centering
\begin{tabular}{|c|c|c|c|c|c|} 
\hline
Cycle & Estimate Area & Forecast Area & Estimate ROS & Forecast ROS & FMC Adjustment \% \\ 
\hline
0     & 767       & 1126          & 0.1818   & 0.1949       & 0               \\ 
\hline
1     & 1603      & 1684          & 0.1755   & 0.2051       & 0.1967              \\ 
\hline
2     & 1998      & 1950          & 0.0994   & 0.1802       & 0               \\ 
\hline
3     & 2734      & 2343          & 0.0867   & 0.1048      & 0         \\ 
\hline
4     & 3602      & 3212          & 0.0912   & 0.1087       & 0          \\
\hline
\end{tabular}
\caption[Criticial data for making adjustments to the FMC in a simulation of the  Patch Springs fire.]{Criticial data for making adjustments to the FMC in a simulation of the  Patch Springs fire. Computations made on a 250m grid. No FMC adjustments were made during the initial cycles because the size of the forecast fire was too large but the ROS in the forecast was lower than that in the estimate from data.}
\label{tbl:patch_fmc}
\end{table}

\begin{table}
\centering
\begin{tabular}{|c|c|c|c|c|c|} 
\hline
Cycle & Hours of Forecast & MOE X  & MOE Y  & \textbar{}\textbar{}MOE\textbar{}\textbar{} & Sorenson  \\ 
\hline
0     & 130  & 0.4920 & 0.7629 & 0.9078                                      & 0.5982    \\ 
\hline
1     & 82 & 0.7543 & 0.6413 & 0.9901                                      & 0.6932    \\ 
\hline
2     & 58 & 0.6247 & 0.7157 & 0.9500                                      & 0.6671    \\ 
\hline
3     & 34 & 0.6873 & 0.6869 & 0.9717                                      & 0.6871    \\ 
\hline
4     & 10 & 0.8925 & 0.6684 & 1.1150                                      & 0.7643    \\
\hline
\end{tabular}
\caption{Assesment of the Patch Fire Simulation With Adjustment of FMC. Compare with the results in Table \ref{tbl:patch_moe}. The first two cycles had identical results since no changes were made to the FMC. The simulation with 58 hours of forecast was slightly worse than if no changes had been made to the FMC. The final forecasts were better, using the FMC adjustment.}
\label{tbl:patch_sim_fmc_moe_results}
\end{table}

\subsubsection{Cougar Creek Fire with FMC Adjustment}
\label{sec:cougar_fmc_run}

\begin{figure}[!ht]
\centering
\includegraphics[width=0.45\textwidth]{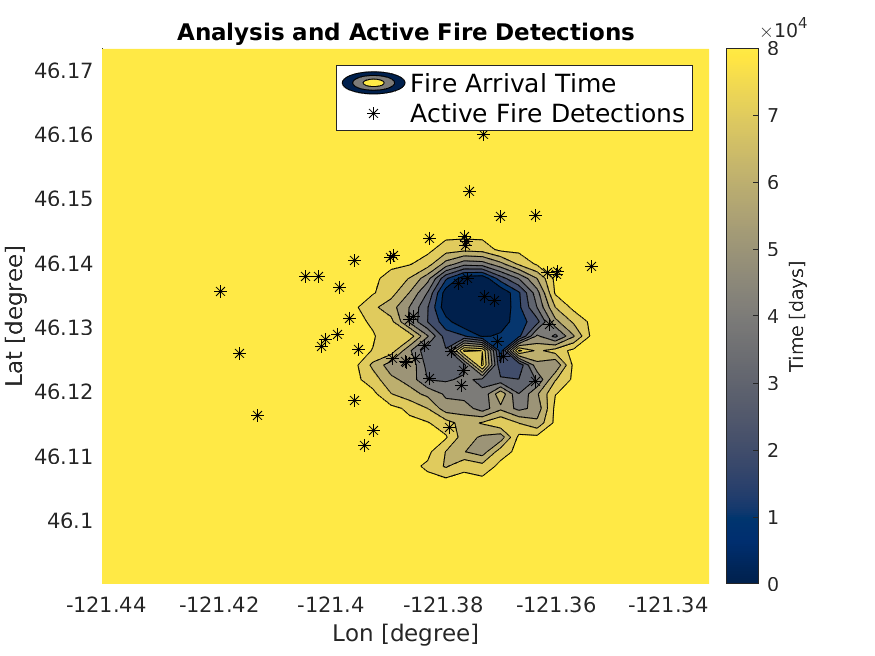}
\includegraphics[width=0.45\textwidth]{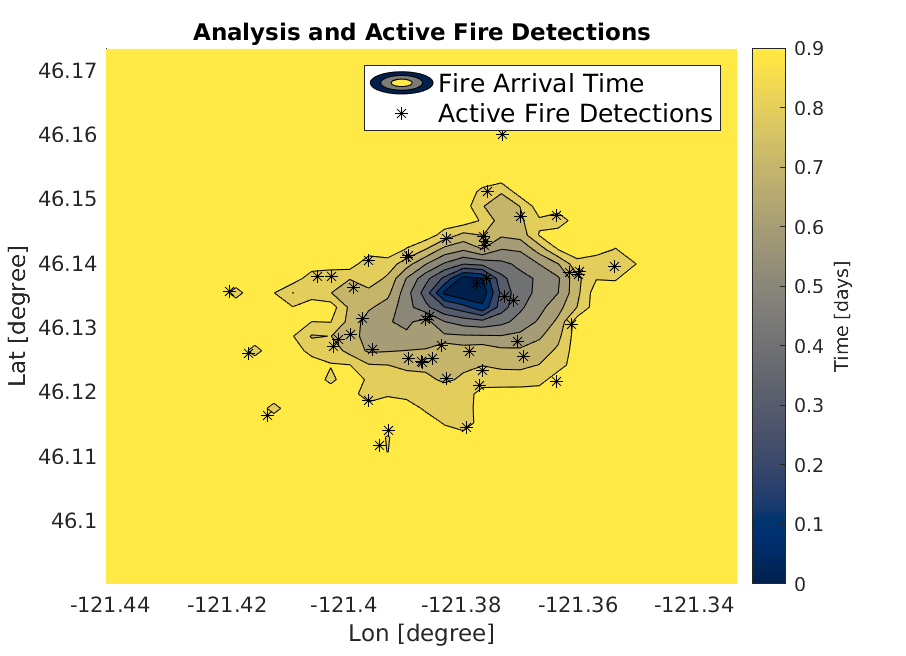}
\includegraphics[width=0.45\textwidth]{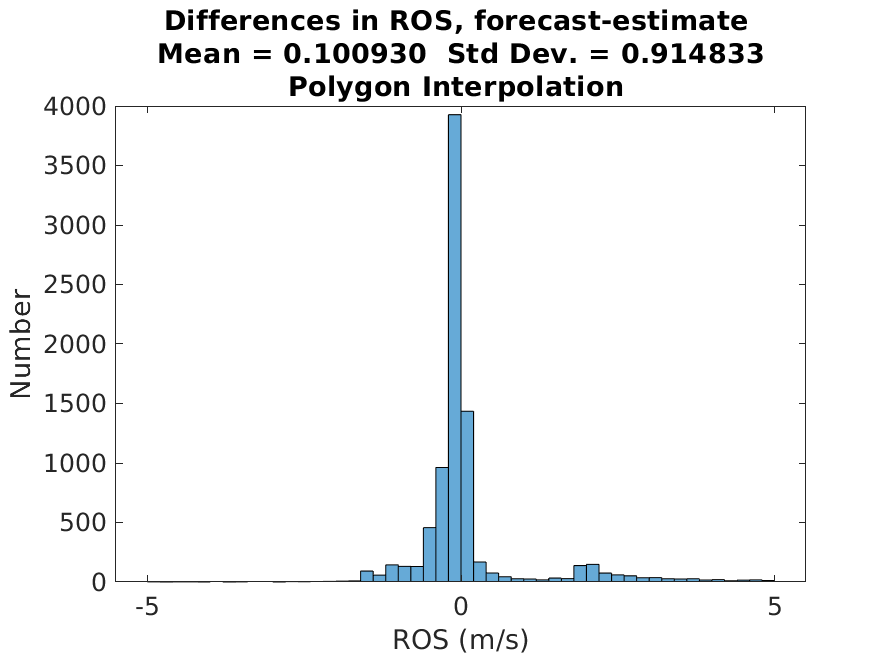}
     \caption{Cycle 0 of the Cougar Creek Fire simulation with adjustment of FMC. In the upper left and right are contours of of the fire arrival time for the forecast and the analysis, respectively. On the bottom is the histogram showing the difference between the ROS at mesh points in the fire domain. The forecast underestimated the size of the fire and the ROS so that the FMC was adjusted downward by -0.0574\%}
\label{fig:cougar_fmc_cycle_0}
\end{figure}

\begin{figure}[!ht]
\centering
\includegraphics[width=0.45\textwidth]{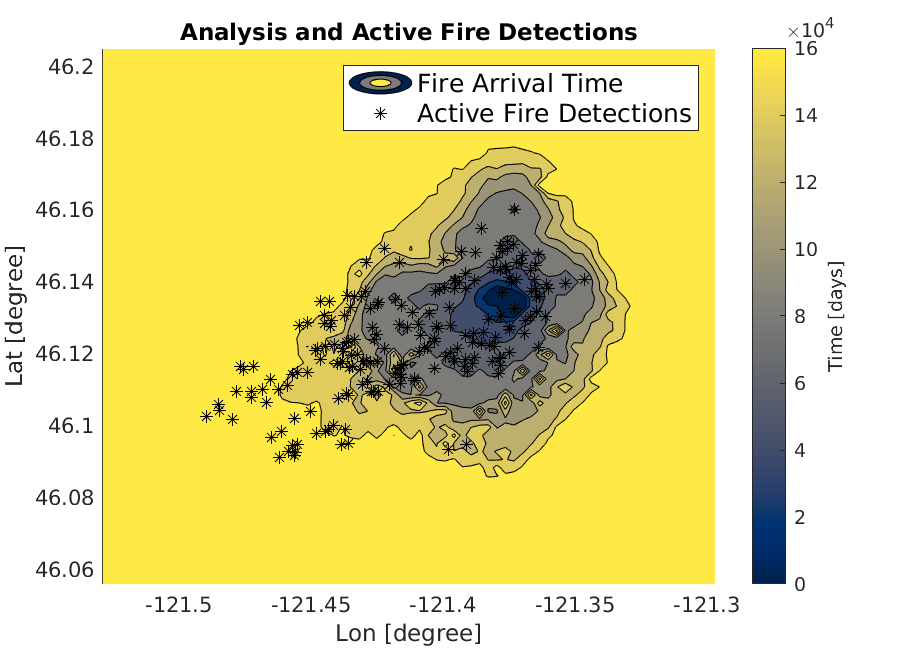}
\includegraphics[width=0.45\textwidth]{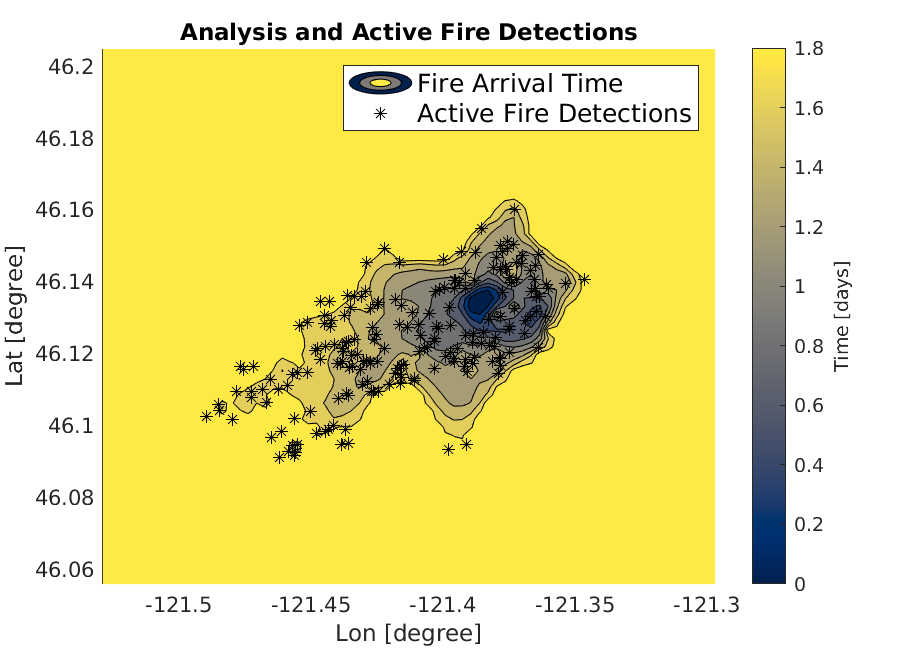}
\includegraphics[width=0.45\textwidth]{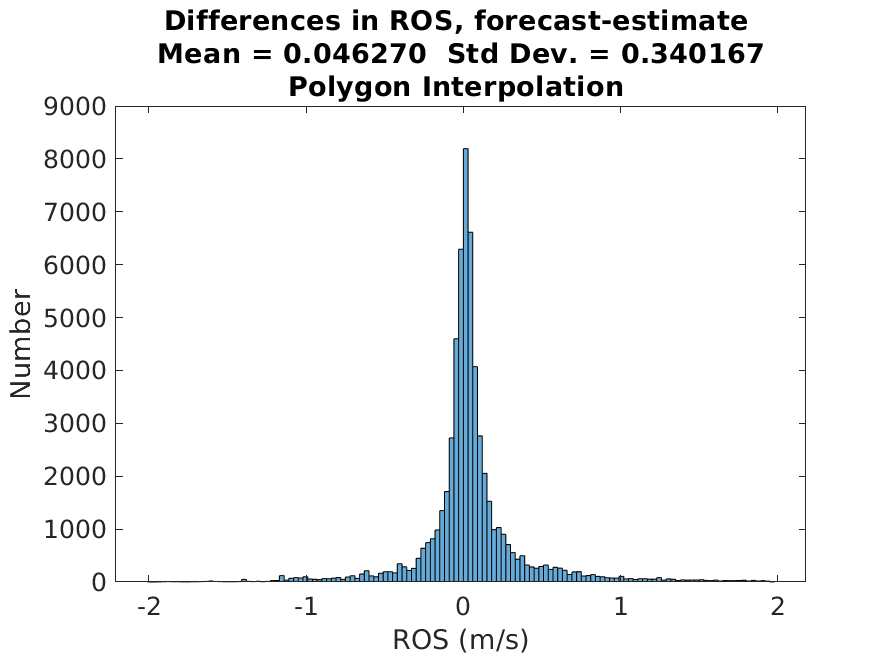}
     \caption{Cycle 1 of the Cougar Creek Fire simulation with adjustment of FMC. In the upper left and right are contours of of the fire arrival time for the forecast and the analysis, respectively. On the bottom is the histogram showing the difference between the ROS at mesh points in the fire domain. The forecast overestimated the size of the fire and the ROS so that the FMC was adjusted upward by 0.1947\%}
\label{fig:cougar_fmc_cycle_1}
\end{figure}

A simulation of the Cougar Creek fire was run using the cycling strategy as in Section \ref{sec:cougar_data_assilation} except that a change to the FMC was made during the data assimilation phase. If a change to the FMC was suggested by the algorithm, it was applied to all fuels in the entire fire domain. Figures \ref{fig:cougar_fmc_cycle_0},\ref{fig:cougar_fmc_cycle_1}, \ref{fig:cougar_fmc_cycle_2}, and \ref{fig:cougar_fmc_cycle_3} show how the data assimilation made adjustment to the fire arrival time during the simulation cycles as well as a histogram showing how the ROS in the forecast differed from the estimated ROS. Table \ref{tbl:cougar_fmc} shows a comparison of the forecast and estimated fire area as well as the average ROS in the forecast an estimate for each cycle. When any adjustment to the FMC was made by the algorithm, that adjustment was indicated too.

Assessment of the simulation was made by comparison with and infrared perimeter and the results are summarized in Table \ref{tbl:cougar_sim_fmc_moe_results}. Figure \ref{fig:cougar_moe_fmc_cycles} gives a graphical representation of these results. Compared to using data assimilation without changing the FMC, these results are a little better. In particular, the change of the FMC in the second cycle made for wetter fuels that helped constrain the growth of the fire, making for a better ultimate forecast.

\begin{table}
\centering
\begin{tabular}{|c|c|c|c|c|c|} 
\hline
Cycle & Data Area & Forecast Area & Data ROS & Forecast ROS & FMC Adjustment \% \\ 
\hline
0     & 18498     & 12924         & 0.1379   & 0.1006       & -0.0574         \\ 
\hline
1     & 78038     & 122154        & 0.1397   & 0.186        & 0.1947          \\ 
\hline
2     & 151647    & 148920        & 0.1111   & 0.1517       & 0               \\ 
\hline
3     & 161939    & 135567        & 0.0948   & 0.1066       & 0
\\ 
\hline
\end{tabular}
\caption[Adjustments made to the FMC in the simulation of the Cougar Creek Fire.]{Adjustments made to the FMC in the simulation of the Cougar Creek Fire. Adjustments were only made after cycle 0 and cycle 1. In cycles 2 and 3, the larger fire area implied by the data contradicted the larger average ROS found in the forecast. The algorithm makes no adjustments under these circumstances.}
\label{tbl:cougar_fmc}
\end{table}

\begin{table}
\centering
\begin{tabular}{|c|c|c|c|c|c|} 
\hline
Cycle & Hours of Forecast & MOE X  & MOE Y  & \textbar{}\textbar{}MOE\textbar{}\textbar{} & Sorenson  \\ 
\hline
0     & 105    & 0.6638 & 0.6647 & 0.9394                                      & 0.6642    \\ 
\hline
1     & 81    & 0.982  & 0.3736 & 1.0507                                      & 0.5413    \\ 
\hline
2     & 57     & 0.7025 & 0.7353 & 1.0169                                      & 0.7185    \\ 
\hline
3     & 33    & 0.6918 & 0.8971 & 1.1329                                      & 0.7812    \\ 
\hline
4     & 9     & 0.8919 & 0.8045 &  1.2011
& 0.8459    \\
\hline
\end{tabular}
\caption{Assessment of the Cougar Creek Fire simulation using adjustment of the FMC. Compare with the results in the test without adjustment of FMC in Table \ref{tbl:cougar_creek_moe}. }
\label{tbl:cougar_sim_fmc_moe_results}
\end{table}

\subsection{Limitations of the method}
The proposed method for adjusting the FMC used by the model  has some limitations and drawbacks that are explored in the following list. Some possible remedies are suggested as well. The goal of adjusting the FMC is to make the ROS in the simulation more closely match what is observed in the real world. Because fire behavior is such a complicated phenomena, other  approaches for producing a better ROS in the model could be explored as well as what was presented in this research. 
\begin{enumerate}
\item The method does not consider all inputs that effect the ROS of a fire such as fuel types, terrain, etc. Controlling for these other inputs would allow for a more precise, and likely more conservative, adjustment of the FMC. For example, if strong winds were present in the simulation, it could be the case that the effect of FMC on the ROS of the fire played a comparatively minor role when compared to the wind. 
\item Comparison between the forecast ROS and the estimated ROS is made at specific locations within the fire domain without regard to the changing properties of the fuels. For example, if the forecast fire arrival time at a specified location differs from the estimated fire arrival time at that location by  12 hours, it is likely that the underlying fuels in that area should be expected to have different FMC because of the diurnal cycle of FMC increasing during the cooler period of night and decreasing during hotter days. These diurnal changes have a larger affect on the the 1-hour and 10-hour fuels, but in principle, all fuels should be expected to have different properties at different times. The adjustment of FMC might be more accurate if comparison of the ROS was done only for location where the difference in the forecast and estimated fire arrival time was was not too large. A simpler method would be to only make adjustments to the FMC for the 100-hour and 1000-hour fuels that are less sensitive to changes in the weather and are observed more sparsely.
\item The cut-off of ROS above 2 m/s is arbitrary. No testing was done to arrive at this number. With certain combinations of weather, topography,and fuels properties some fires could exhibit such rapid growth. A better cut-off for the ROS could be possibly be found by using the estimates of the variance for the ROS found in Section \ref{sec:ros_uq}.
\end{enumerate}

\clearpage
\FloatBarrier

\chapter{\uppercase{Conclusion and Further Work}} \label{chapter:06}

\section{Conclusion}\label{sec:conclusion}
\subsection{Main results}
The goal of this research was to develop  techniques for using satellite data to provide improvements to wildfire modeling. The first of three three goals was to develop methods to obtain better initial conditions for simulating a wildfire by estimating either the ignition point or fire arrival time of a wildfire. The estimation of the fire arrival time should be considered the most useful of these two methods since it paves the way for starting better wildfire simulations and the estimate obtained may be used to infer other properties about a fire such as its ROS. The second goal was to develop a method for using satellite data to update a running fire simulation by data assimilation in a way that is justifiable by the known properties and uncertainties of the those data. The third goal was use satellite data to adjust the fuel moisture content used by the model to make better fire forecasts possible. The first two goals have been met, but the third goal needs more development before it can be called an unqualified success. Some of the main results are summarized in this section.

\subsection{Estimation of Igbition Point and Fire Arrival Time}
The first part of this thesis detailed  methods used to estimate the ignition point location and the fire arrival time from satellite fire data. Because the the ignition point location and the fire arrival time of real wildfires are not generally known, the methods were developed and tested by using artificial data with a known, ``ground truth" ignition point or  fire arrival time. 

For estimating the ignition point of a fire, an artificial fire arrival time was created and artificial fire detections were placed on the surface of the fire arrival cone. Using a grid search strategy, an ensemble of forecasts with varying locations and times of ignition was made and the data likelihood of each was computed. The estimated time and place of ignition was found from the ignition point of the forecast with the highest data likelihood. This method worked to identify the time and place of ignition for a test using a ``ground truth" artificial fire arrival time derived from the equation of a cone as well as in a test where the ``ground truth" was taken as the fire arrival time output from a WRF-SFIRE simulation. A third test of the method was made using data from a real wildfire. In this real-world example, the method was able to identify an ignition time within an hour of that surmised by fire investigators, but the estimated ignition location differed from that provided by investigators by around 3 km.

For estimating the fire arrival time, the testing took place in two stages. In the first stage, artificial fire detection data was created and the method for estimating the fire arrival time was used to attempt to reconstruct the fire arrival time from the sparse artificial data. Testing, using various strategies and key parameters for computer functions used in the method, was a key goal of this first step. Based on the results of the testing, a workable method for producing accurate estimates of the fire arrival time was devised. The testing also revealed some situations in which the method is likely to produce poor results. The method is likely to fail when only a small amount of observations are available for relatively small fires. The method also produced poor results when there was more than one fire in the domain. This shortcoming is more serious. During wildfire season, many fires can exist concurrently in a geographic region. The method presented here will attempt to treat these separate events as if they were a singular fire and the resulting estimate of the fire arrival time can be expected to show fire spread in regions where no data indicates such burning has occurred. 

The second stage of this testing was to determine how well the established method worked for more complicated ``ground truth" fire arrival times, taken from various forecasts produced by WRF-SFIRE. The goal was to use the method developed in the first stage in exactly the same way to see if similar results could be achieved, indicating the method can be generalized to fire scenarios other than those in the first stage of the testing. The results of the estimates in the second stage were a bit worse than those in the first stage. In particular, the method tended to underestimate the size of the fire and give a poorer estimate for the growth rate of the fire. 

Finally, the method was applied to estimating the fire arrival time of real-world fires. Because the actual fire arrival time of such fire is not known, assessment of the method was made by comparison with infrared perimeter observations. The method produced good results in terms of these comparisons with both dimensions of the MOE and the S{\o}renson index around 0.9. These scores are at least as good as those obtained in the hundreds of artificial fires that were used to test and develop the method. Despite the high scores for these real-world scenarios, it should be realized that the shortcomings previously mentioned are still of concern. The method is likely to deliver poor results during the early stage of a fire, when few satellite observations are available, and the presence of more than one fire in the domain will cause problems in the real-world situation just as it did in testing with artificial data.

With a good estimate of the fire arrival time established, it is possible to get an estimate of the rate of spread of the fire at all locations in the fire domain. In this work, we used the known uncertainties in the satellite observations to develop a sense of how reliable the estimates of the ROS are. Although used sparingly in this work, the method for for finding the ROS from satellite data has greater potential to impact wildfire modeling in a positive way. At the heart of wildfire model are the spread models like the Rothermel or Balbi models. Although widely used, other, improved models could be devised. Possibly the method for determining the ROS established here could be used in conjunction with the large amount of historical satellite data to develop a new model using machine learning techniques. 

\subsection{Data Assimilation and Adjustment of the Fuel Moisture Content}
Several different forecast strategies using data assimilation were discussed in this work and comparisons of simulations using these strategies follow. The Patch Springs Fire and Cougar Creek Fire are presented here. The Camp Fire also served as an example of the data assimilation techniques but will not be treated in the same way as the other two because during the simulation period its mechanism of spread was largely due to winds carrying burning embers ahead of the fire front. The WRF-SFIRE model does not currently model such behavior and the concept of adjusting the FMC content makes little sense under such circumstances. The foresting strategies to be compared follow. A single forecast, without using data assimilation is included in this comparison as a control. At a bare minimum, any forecast strategy using data assimilation techniques should be able to produce better results than this single forecast if it is to be worthy of consideration and further study.

\begin{enumerate}
\item A single forecast, started from an ignition point
\item A forecast using data assimilation cycling
\item A forecast with data assimilation cycling and adjustment of FMC
\item 48-Hour Forecast strategy
\end{enumerate} 

Each forecast strategy has been evaluated by comparing a final forecast perimeter with an infrared perimeter observation using the MOE and the S{\o}renson index. For simplicity, a score for each strategy was computed from the sum of these measures as

\begin{equation}
\text{SCORE} = \text{MOE X}+\text{MOE Y}+S.
\end{equation}
where $\text{MOE X}$ and $\text{MOE Y}$ are the two components of the measure of effectiveness describing decreasing false negatives and false positives, respectively, and $S$ is the S{\o}renson index. 

Table \ref{tbl:overall_score} gives the results of these scores for both the Patch Springs and Cougar Creek fires. In both cases, the 48-hour forecast strategy produced the best results, and the worst results were obtained by single forecasts from an ignition point that made no attempt to assimilate satellite data. Use of data assimilation and data assimilation with FMC adjustments gave mixed results, with both strategies falling in the middle. Part of the reason for the success of the 48-hour strategy is surely due to the extra perimeter data that was used in the process of finding an estimate of the fire arrival time that was used for initializing the model. Another potential advantage was that the data assimilation performed was over a relatively show period and didn't seek to make large changes to the forecast.  Further discussion of the particulars of each fire and its simulations are given below. 

\begin{table}
\centering
\begin{tabular}{|c|c|c|} 
\hline
Forecast Strategy         & Patch Springs & Cougar Creek  \\ 
\hline
Single Forecast           & 1.8940        & 1.9927        \\ 
\hline
Data Assimilation         & 2.0217        & 2.1081        \\ 
\hline
Data Assimilation and FMC & 2.0075        & 2.1563        \\ 
\hline
48-Hour Strategy          & 2.0970        & 2.3999        \\
\hline
\end{tabular}
\caption[Overall results of simulation strategies for Patch Springs and Cougar Creek Fire simulations.]{Overall results of simulation strategies for the Patch Springs and Cougar Creek Fire simulations. For both fires, the 48-hour forecast strategy produced the best results. Using a single simulation produced the worst results. }
\label{tbl:overall_score}
\end{table}

\subsubsection{Patch Springs Fire Simulations}

Comparison was made between four simulations of the Patch Springs Fire. The simulations using a single run, data assimilation cycling, and data assimilation with FMC adjustment were all started from an identical ignition point and the 48-hour strategy simulation was started from an estimated fire arrival time made using satellite and infrared perimeter data. The single simulation represents 130 hours of forecast time and the other simulations make 58 hours of forecast, each.

Table \ref{tbl:patch_conclusion_results} summarizes the results using the MOE and S{\o}renson index, and Figure \ref{fig:patch_moe_conclusion} shows a graphical representation of the results. The 48-hour forecast strategy produced the best result for this simulation. It should be noted that all forecast strategies failed to capture the rapid growth of the fire that occurred during the fourth day. The large red areas of false negative in the northeast section of the panels in Figure \ref{fig:patch_moe_conclusion} show the region where the observed fire growth was explosive during this period. The reason the model failed to predict this growth is unclear. Sudden changes in weather or an inaccurate accounting of the fuel properties in the are could explain the model's shortcomings. The simulation using data assimilation showed more burning in this region, but it also showed more burning in areas where no fire was actually observed. The simulation using data assimilation with FMC adjustment managed to control the size of the fire and produced a result similar to that of the simulation using the 48-hour forecast strategy.

\begin{table}
\centering
\begin{tabular}{|c|c|c|c|c|} 
\hline
Forecast Strategy         & MOE X  & MOE Y  & \textbar{}\textbar{}MOE\textbar{}\textbar{} & S{\o}renson  \\ 
\hline
Single Forecast  & 0.4920 & 0.7629 & 0.9078 & 0.5982  \\
\hline
Data Assimilation & 0.7009 & 0.6476 & 0.9543                                      & 0.6732    \\ 
\hline
Data Assimilation and FMC & 0.6247 & 0.7157 & 0.9500                                      & 0.6671    \\ 
\hline
48-Hour Strategy  & 0.6450 & 0.7559 & 0.9937                                      & 0.6961    \\
\hline
\end{tabular}
\caption{Results for the Patch Springs fire simulations. The 48-hour forecast strategy produced the best results. The worst results were from a single simulation without using any data assimilation. }
\label{tbl:patch_conclusion_results}
\end{table}

\subsubsection{Cougar Creek Fire}

The four strategies were applied to the Cougar Creek Fire Simulation. The single run, data assimilation cycling, and the data assimilation cycling with FMC adjustments were all started from an identical ignition point. The simulation using the 48-hour forecast strategy was started from an estimated fire arrival time made using satellite data and infrared perimeter observations. The single run simulation represents 105 hours of forecast time, the two simulations with data assimilation have 57 hours of forecast time, and the simulation using the 48-hour strategy has 57 hours of forecast time. 

Table \ref{tbl:cougar_conclusion_results} summarizes the results using the MOE and S{\o}renson index and Figure \ref{fig:cougar_moe_conclusion} shows a graphical representation of these results. The 48-hour forecast strategy produced the best result for this simulation. This strategy tended to underestimate the fire size, but the perimeter was close to the observed perimeter in many places. The worst result was obtained by running a single simulation. The overall size of the fire was similar to the observed fire, but the perimeter matched poorly. Using data assimilation made for a forecast that had too large of an area. Using data assimilation with FMC adjustment improved the forecast by slowing down the rate of spread and produced a forecast closer to that of the 48-hour strategy.

\begin{table}
\centering
\begin{tabular}{|c|c|c|c|c|} 
\hline
Forecast Strategy         & MOE X  & MOE Y  & \textbar{}\textbar{}MOE\textbar{}\textbar{} & S{\o}renson  \\ 
\hline
Single Forecast           & 0.6638 & 0.6647 & 0.9394                                      & 0.6642    \\ 
\hline
Data Assimilation         & 0.9873 & 0.4773 & 1.0966                                      & 0.6435    \\ 
\hline
Data Assimilation and FMC & 0.7025 & 0.7353 & 1.0169                                      & 0.7185    \\ 
\hline
48-Hour Strategy          & 0.7305 & 0.8722 & 1.1377                                      & 0.7972    \\
\hline
\end{tabular}
\caption{Results for the Cougar Creek Fire simulations.}
\label{tbl:cougar_conclusion_results}
\end{table}


\clearpage
\FloatBarrier

\renewcommand\bibname{REFERENCES}
\singlespacing

\bibliographystyle{ucdDissertation}
\bibliography{ref/geo.bib,ref/other,ref/web.bib}

\begin{thebibliography}{78}
\newcommand{\enquote}[1]{``#1''}
\providecommand{\natexlab}[1]{#1}
\providecommand{\url}[1]{\texttt{#1}}
\providecommand{\urlprefix}{URL }
\expandafter\ifx\csname urlstyle\endcsname\relax
  \providecommand{\doi}[1]{doi:\discretionary{}{}{}#1}\else
  \providecommand{\doi}{doi:\discretionary{}{}{}\begingroup
  \urlstyle{rm}\Url}\fi
\providecommand{\eprint}[2][]{\url{#2}}

\bibitem[{Albini(1976)}]{Albini-1976-EWB}
Albini FA (1976).
\newblock \enquote{Estimating wildfire behavior and effects.}
\newblock U. S. Forest Service. General Technical Report INT-30.
\newblock \url{http://www.treesearch.fs.fed.us/pubs/29574}.

\bibitem[{Andrews \emph{et~al.}(2007)Andrews, Finney, and
  Fischetti}]{Andrews-2007-PW}
Andrews P, Finney M, Fischetti M (2007).
\newblock \enquote{Predicting Wildfires.}
\newblock \emph{Scientific American}, \textbf{297}, 46--51, 54.
\newblock \doi{10.1038/scientificamerican0807-46}.

\bibitem[{Baker(2011)}]{Baker-2011-JPS}
Baker N (2011).
\newblock \enquote{Joint Polar Satellite System {(JPPS)} {VIIRS} Geolocation
  Algorithm Theoretical Basis Document.}
\newblock Available at
  \url{https://lpdaac.usgs.gov/sites/default/files/public/product_documentation/vnp03_v1_atbd.pdf},
  retrieved April 1, 2018.

\bibitem[{Balbi \emph{et~al.}(2009)Balbi, Morandini, Silvani, Filippi, and
  Rinieri}]{Balbi-2009-PMW}
Balbi JH, Morandini F, Silvani X, Filippi JB, Rinieri F (2009).
\newblock \enquote{A physical model for wildland fires.}
\newblock \emph{Combustion and Flame}, \textbf{156}(12), 2217--2230.
\newblock ISSN 0010-2180.
\newblock \doi{10.1016/j.combustflame.2009.07.010}.

\bibitem[{Benali \emph{et~al.}(2016)Benali, Russo, S{\'a}, Pinto, Price,
  Koutsias, and Pereira}]{Benali-2016-DFD}
Benali A, Russo A, S{\'a} ACL, Pinto RMS, Price O, Koutsias N, Pereira JMC
  (2016).
\newblock \enquote{Determining Fire Dates and Locating Ignition Points With
  Satellite Data.}
\newblock \emph{Remote Sensing}, \textbf{8}(4), Article number 326.
\newblock ISSN 2072-4292.
\newblock \doi{10.3390/rs8040326}.

\bibitem[{Brewer and Clements(2020)}]{Brewer-2020-CFM}
Brewer MJ, Clements CB (2020).
\newblock \enquote{The 2018 Camp Fire: Meteorological Analysis Using In Situ
  Observations and Numerical Simulations.}
\newblock \emph{Atmosphere}, \textbf{11}(1).
\newblock ISSN 2073-4433.
\newblock \doi{10.3390/atmos11010047}.

\bibitem[{{Cal Fire}(2018 (Accessed March 3, 2020))}]{Cal_fire-2018-GSB}
{Cal Fire} (2018 (Accessed March 3, 2020)).
\newblock \emph{Green Sheet: Burn Injuries, Camp Incident}.
\newblock
  \url{https://assets.documentcloud.org/documents/5628194/18-CA-BTU-016737-Camp-Green-Sheet.pdf}.

\bibitem[{Cardil \emph{et~al.}(2019{\natexlab{a}})Cardil, Monedero, Ramarez,
  and Silva}]{Cardil-2019-ARW}
Cardil A, Monedero S, Ramarez J, Silva CA (2019{\natexlab{a}}).
\newblock \enquote{Assessing and reinitializing wildland fire simulations
  through satellite active fire data.}
\newblock \emph{Journal of Environmental Management}, \textbf{231}, 996 --
  1003.
\newblock ISSN 0301-4797.
\newblock \doi{10.1016/j.jenvman.2018.10.115}.

\bibitem[{Cardil \emph{et~al.}(2019{\natexlab{b}})Cardil, Monedero, Silva, and
  Ramirez}]{Cardil-2019-ARS}
Cardil A, Monedero S, Silva CA, Ramirez J (2019{\natexlab{b}}).
\newblock \enquote{Adjusting the rate of spread of fire simulations in
  real-time.}
\newblock \emph{Ecological Modelling}, \textbf{395}, 39--44.
\newblock ISSN 0304-3800.
\newblock \doi{10.1016/j.ecolmodel.2019.01.017}.

\bibitem[{Clark \emph{et~al.}(1996)Clark, Jenkins, Coen, and
  Packham}]{Clark-1996-CAF}
Clark TL, Jenkins MA, Coen J, Packham D (1996).
\newblock \enquote{A Coupled Atmospheric-Fire Model: {C}onvective Feedback on
  Fire Line Dynamics.}
\newblock \emph{J. Appl. Meteor}, \textbf{35}, 875--901.
\newblock \doi{10.1175/1520-0450(1996)035<0875:ACAMCF>2.0.CO;2}.

\bibitem[{Coen and Schroeder(2013)}]{Coen-2013-USR}
Coen JL, Schroeder W (2013).
\newblock \enquote{Use of spatially refined satellite remote sensing fire
  detection data to initialize and evaluate coupled weather-wildfire growth
  model simulations.}
\newblock \emph{Geophysical Research Letters}, \textbf{40}, 1--6.
\newblock \doi{10.1002/2013GL057868}.

\bibitem[{Cunningham and Linn(2007)}]{Cunningham-2007-NSG}
Cunningham P, Linn RR (2007).
\newblock \enquote{Numerical simulations of grass fires using a coupled
  atmosphere-fire model: Dynamics of fire spread.}
\newblock \emph{Journal of Geophysical Research: Atmospheres}, \textbf{112},
  Art. D05108, 17pp.
\newblock ISSN 2156-2202.
\newblock \doi{10.1029/2006JD007638}.

\bibitem[{Dahl \emph{et~al.}(2015)Dahl, Xue, Hu, and Xue}]{Dahl-2015-CFA}
Dahl N, Xue H, Hu X, Xue M (2015).
\newblock \enquote{Coupled fire--atmosphere modeling of wildland fire spread
  using {DEVS-FIRE} and {ARPS}.}
\newblock \emph{Natural Hazards}, \textbf{77}(2), 1013--1035.
\newblock ISSN 1573-0840.
\newblock \doi{10.1007/s11069-015-1640-y}.

\bibitem[{Daley(1991)}]{Daley-1991-ADA}
Daley R (1991).
\newblock \emph{Atmospheric data analysis}.
\newblock Cambridge ; New York : Cambridge University Press.
\newblock ISBN 0521382157.

\bibitem[{Farguell \emph{et~al.}(2021)Farguell, Mandel, Haley, Mallia,
  Kochanski, and Hilburn}]{Farguell-2021-MLE}
Farguell A, Mandel J, Haley J, Mallia DV, Kochanski A, Hilburn K (2021).
\newblock \enquote{Machine Learning Estimation of Fire Arrival Time from
  Level-2 {Active Fires} Satellite Data.}
\newblock \emph{Remote Sensing}, \textbf{13}(11), 2203.
\newblock ISSN 2072-4292.
\newblock \doi{10.3390/rs13112203}.

\bibitem[{Ferguson(1967)}]{Ferguson-1967-MSD}
Ferguson TS (1967).
\newblock \emph{Mathematical statistics: a decision theoretic approach}.
\newblock Academic Press, London, England;New York, New York;.
\newblock ISBN 9781483221236;9781483182537;1483221237;1483182533;.

\bibitem[{Filippi \emph{et~al.}(2011)Filippi, Bosseur, Pialat, Santoni, Strada,
  and Mari}]{Filippi-2011-SCF}
Filippi JB, Bosseur F, Pialat X, Santoni P, Strada S, Mari C (2011).
\newblock \enquote{Simulation of coupled fire/atmosphere interaction with the
  {MesoNH-ForeFire} models.}
\newblock \emph{Journal of Combustion}, \textbf{2011}, Article ID 540390.
\newblock \doi{10.1155/2011/540390}.

\bibitem[{Finney(1998)}]{Finney-1998-FFA}
Finney MA (1998).
\newblock \enquote{{FARSITE}: {F}ire {A}rea {S}imulator -- model development
  and evaluation.}
\newblock Res. Pap. RMRS-RP-4 Revised 2004, Ogden, UT, USDA Forest Service,
  Rocky Mountain Research Station.
\newblock \doi{10.2737/RMRS-RP-4}.

\bibitem[{Finney(2006)}]{Finney-2006-OFF}
Finney MA (2006).
\newblock \enquote{An overview of {FlamMap} fire modeling capabilities.}
\newblock In PL~Andrews, BW~Butler (eds.), \emph{Fuels Management -- How to
  Measure Success}, pp. 213--220. USDA Forest Service.
\newblock Proceedings RMRS-P-41, available at
  \url{http://www.treesearch.fs.fed.us/pubs/25948}.

\bibitem[{Gabbert(2019)}]{Gabbert-2019-UPS}
Gabbert B (2019).
\newblock \enquote{Utah: Patch Springs Fire burns 13,000 acres southwest of
  Salt Lake City.}
\newblock
  \url{https://wildfiretoday.com/2013/08/15/utah-patch-springs-fire-burns-13000-acres-southwest-of-salt-lake-city/}.

\bibitem[{Giglio(2015)}]{Giglio-2015-MC6}
Giglio L (2015).
\newblock \enquote{{MODIS} Collection 6 Active Fire Product User's Guide
  Version 2.6.}
\newblock Department of Geographical Sciences, University of Maryland.
\newblock Available at
  \url{http://modis-fire.umd.edu/files/MODIS_C6_Fire_User_Guide_A.pdf},
  accessed March 2015.

\bibitem[{Giglio \emph{et~al.}(2016)Giglio, Schroeder, and
  Justice}]{Giglio-2016-C6M}
Giglio L, Schroeder W, Justice CO (2016).
\newblock \enquote{The collection 6 {MODIS} active fire detection algorithm and
  fire products.}
\newblock \emph{Remote Sensing of Environment}, \textbf{178}, 31--41.
\newblock ISSN 0034-4257.
\newblock \doi{10.1016/j.rse.2016.02.054}.

\bibitem[{Goss \emph{et~al.}(2020)Goss, Swain, Abatzoglou, Sarhadi, Kolden,
  Williams, and Diffenbaugh}]{Goss-2020-CCI}
Goss M, Swain DL, Abatzoglou JT, Sarhadi A, Kolden CA, Williams AP, Diffenbaugh
  NS (2020).
\newblock \enquote{Climate change is increasing the likelihood of extreme
  autumn wildfire conditions across California.}
\newblock \emph{Environmental Research Letters}, \textbf{15}(9), 094016.
\newblock \doi{10.1088/1748-9326/ab83a7}.

\bibitem[{Haley \emph{et~al.}(2018)Haley, Farguell~Caus, Mandel, Kochanski, and
  Schranz}]{Haley-2018-DLA}
Haley J, Farguell~Caus A, Mandel J, Kochanski AK, Schranz S (2018).
\newblock \enquote{Data Likelihood of Active Fires Satellite Detection and
  Applications to Ignition Estimation and Data Assimilation.}
\newblock In DX~Viegas (ed.), \emph{Advances in Forest Fire Research}.
  University of Coimbra Press.
\newblock Accepted. Available at \url{https://arxiv.org/abs/1808.03318}, July
  2018.

\bibitem[{Jaffe \emph{et~al.}(2020)Jaffe, O’Neill, Larkin, Holder, Peterson,
  Halofsky, and Rappold}]{Jaffe-2020-WPB}
Jaffe DA, O’Neill SM, Larkin NK, Holder AL, Peterson DL, Halofsky JE, Rappold
  AG (2020).
\newblock \enquote{Wildfire and prescribed burning impacts on air quality in
  the United States.}
\newblock \emph{Journal of the Air \& Waste Management Association},
  \textbf{70}(6), 583--615.
\newblock \doi{10.1080/10962247.2020.1749731}.
\newblock PMID: 32240055,
  \eprint{https://doi.org/10.1080/10962247.2020.1749731}.

\bibitem[{Jahn \emph{et~al.}(2011)Jahn, Rein, and Torero}]{Jahn-2011-FFG}
Jahn W, Rein G, Torero J (2011).
\newblock \enquote{Forecasting fire growth using an inverse zone modelling
  approach.}
\newblock \emph{Fire Safety Journal}, \textbf{46}(3), 81 -- 88.
\newblock ISSN 0379-7112.
\newblock \doi{https://doi.org/10.1016/j.firesaf.2010.10.001}.

\bibitem[{Jolly(2015)}]{Jolly-2015-CVG}
Jolly W (2015).
\newblock \enquote{Climate-induced variations in global wildfire danger from
  1979 to 2013.}
\newblock \emph{Nature Communications}, \textbf{6}, 7537.
\newblock \doi{10.1038/ncomms8537}.

\bibitem[{Lafore \emph{et~al.}(1998)Lafore, Stein, Asencio, Bougeault, Ducrocq,
  Duron, Fischer, H\'ereil, Mascart, Masson, Pinty, Redelsperger, Richard, and
  Vil\`a-Guerau~de Arellano}]{Lafore-1998-MHS}
Lafore JP, Stein J, Asencio N, Bougeault P, Ducrocq V, Duron J, Fischer C,
  H\'ereil P, Mascart P, Masson V, Pinty JP, Redelsperger JL, Richard E,
  Vil\`a-Guerau~de Arellano J (1998).
\newblock \enquote{The Meso-NH Atmospheric Simulation System. Part I: adiabatic
  formulation and control simulations.}
\newblock \emph{Annales Geophysicae}, \textbf{16}(1), 90--109.
\newblock \doi{10.1007/s00585-997-0090-6}.

\bibitem[{Linn \emph{et~al.}(2002)Linn, Reisner, Colman, and
  Winterkamp}]{Linn-2002-SWB}
Linn R, Reisner J, Colman JJ, Winterkamp J (2002).
\newblock \enquote{Studying wildfire behavior using {FIRETEC}.}
\newblock \emph{Int. J. of Wildland Fire}, \textbf{11}, 233--246.
\newblock \doi{10.1071/WF02007}.

\bibitem[{{Lloyd}(1982)}]{Loyd-1982-LSQ}
{Lloyd} S (1982).
\newblock \enquote{Least squares quantization in PCM.}
\newblock \emph{IEEE Transactions on Information Theory}, \textbf{28}(2),
  129--137.
\newblock \doi{10.1109/TIT.1982.1056489}.

\bibitem[{Loboda and Csiszar(2007)}]{Laboda-2007-RFS}
Loboda T, Csiszar I (2007).
\newblock \enquote{Reconstruction of fire spread within wildland fire events in
  Northern Eurasia from the MODIS active fire product.}
\newblock \emph{Global and Planetary Change}, \textbf{56}(3), 258 -- 273.
\newblock ISSN 0921-8181.
\newblock \doi{https://doi.org/10.1016/j.gloplacha.2006.07.015}.
\newblock Northern Eurasia Regional Climate and Environmental Change.

\bibitem[{Mandel \emph{et~al.}(2014{\natexlab{a}})Mandel, Amram, Beezley,
  Kelman, Kochanski, Kondratenko, Lynn, Regev, and Vejmelka}]{Mandel-2014-RAA}
Mandel J, Amram S, Beezley JD, Kelman G, Kochanski AK, Kondratenko VY, Lynn BH,
  Regev B, Vejmelka M (2014{\natexlab{a}}).
\newblock \enquote{Recent advances and applications of {WRF-SFIRE}.}
\newblock \emph{Natural Hazards and Earth System Science}, \textbf{14}(10),
  2829--2845.
\newblock \doi{10.5194/nhess-14-2829-2014}.

\bibitem[{Mandel \emph{et~al.}(2011)Mandel, Beezley, and
  Kochanski}]{Mandel-2011-CAF}
Mandel J, Beezley JD, Kochanski AK (2011).
\newblock \enquote{Coupled atmosphere-wildland fire modeling with {WRF} 3.3 and
  {SFIRE} 2011.}
\newblock \emph{Geoscientific Model Development}, \textbf{4}, 591--610.
\newblock \doi{10.5194/gmd-4-591-2011}.

\bibitem[{Mandel \emph{et~al.}(2008)Mandel, Bennethum, Beezley, Coen, Douglas,
  Kim, and Vodacek}]{Mandel-2008-WFM}
Mandel J, Bennethum LS, Beezley JD, Coen JL, Douglas CC, Kim M, Vodacek A
  (2008).
\newblock \enquote{A Wildland Fire Model with Data Assimilation.}
\newblock \emph{Mathematics and Computers in Simulation}, \textbf{79},
  584--606.
\newblock \doi{10.1016/j.matcom.2008.03.015}.

\bibitem[{Mandel \emph{et~al.}(2016)Mandel, Fournier, Jenkins, Kochanski,
  Schranz, and Vejmelka}]{Mandel-2016-ASA}
Mandel J, Fournier A, Jenkins MA, Kochanski AK, Schranz S, Vejmelka M (2016).
\newblock \enquote{Assimilation of Satellite Active Fires Detection Into a
  Coupled Weather-Fire Model.}
\newblock In \emph{Proceedings for the 5th International Fire Behavior and
  Fuels Conference April 11-15, 2016, Portland, Oregon, USA}, pp. 17--22.
  International Association of Wildland Fire, Missoula, Montana, USA.
\newblock
  \url{http://www.iawfonline.org/wp-content/uploads/2018/02/5th-Internatonal-Fire-Behavior-and-Fuel-Conference-Proceedings-Final-updated-1.9.2017-web.pdf},accessed
  March 3, 2020.

\bibitem[{Mandel \emph{et~al.}(2014{\natexlab{b}})Mandel, Kochanski, Vejmelka,
  and Beezley}]{Mandel-2014-DAS}
Mandel J, Kochanski AK, Vejmelka M, Beezley JD (2014{\natexlab{b}}).
\newblock \enquote{Data Assimilation of Satellite Fire Detection in Coupled
  Atmosphere-Fire Simulations by {WRF-SFIRE}.}
\newblock In DX~Viegas (ed.), \emph{Advances in Forest Fire Research}, pp.
  716--724. Coimbra University Press.
\newblock \doi{10.14195/978-989-26-0884-6_80}.

\bibitem[{Mandel \emph{et~al.}(2019)Mandel, Vejmelka, Kochanski, Farguell,
  Haley, Mallia, and Hilburn}]{Mandel-2019-IDH}
Mandel J, Vejmelka M, Kochanski AK, Farguell A, Haley JD, Mallia DV, Hilburn K
  (2019).
\newblock \enquote{An Interactive Data-Driven {HPC} System for Forecasting
  Weather, Wildland Fire, and Smoke.}
\newblock In \emph{2019 IEEE/ACM HPC for Urgent Decision Making (UrgentHPC),
  Denver, CO, USA}, pp. 35--44. IEEE.
\newblock \doi{10.1109/UrgentHPC49580.2019.00010}.

\bibitem[{MathWorks(2020 (accessed October 10, 2020))}]{Matlab-2020-CSS}
MathWorks (2020 (accessed October 10, 2020)).
\newblock \emph{Cubic smoothing spline}.
\newblock \url{https://www.mathworks.com/help/curvefit/csaps.html}.

\bibitem[{MathWorks(2021 (accessed April 28, 2021))}]{Matlab-2021-2GF}
MathWorks (2021 (accessed April 28, 2021)).
\newblock \emph{2-D Gaussian filtering of images - MATLAB imgaussfilt}.
\newblock \url{https://www.mathworks.com/help/images/ref/imgaussfilt.html}.

\bibitem[{MathWorks(2021 (accessed February 5, 2021))}]{Matlab-2021-SPB}
MathWorks (2021 (accessed February 5, 2021)).
\newblock \emph{Shortest path between two single nodes - {MATLAB}
  shortestpath}.
\newblock
  \url{https://www.mathworks.com/help/matlab/ref/graph.shortestpath.html}.

\bibitem[{MathWorks(2021 (accessed March 10, 2021))}]{Matlab-2021-GSA}
MathWorks (2021 (accessed March 10, 2021)).
\newblock \emph{Gradient, slope, and aspect of data grid - MATLAB gradientm}.
\newblock \url{https://www.mathworks.com/help/map/ref/gradientm.html}.

\bibitem[{McGrattan \emph{et~al.}(2013)McGrattan, McDermott, Weinschenk,
  Overholt, Hostikka, and Floyd}]{McGrattan-2013-FDS}
McGrattan K, McDermott R, Weinschenk C, Overholt K, Hostikka S, Floyd J (2013).
\newblock \enquote{Fire Dynamics Simulator User’s Guide.}
\newblock National Institute of Standards and Technology Special Publication
  1019, 262 pages (April 2013).

\bibitem[{Mell \emph{et~al.}(2007)Mell, Jenkins, Gould, and
  Cheney}]{Mell-2007-PAM}
Mell W, Jenkins MA, Gould J, Cheney P (2007).
\newblock \enquote{A physics-based approach to modelling grassland fires.}
\newblock \emph{Intl. J. Wildland Fire}, \textbf{16}, 1--22.
\newblock \doi{10.1071/WF06002}.

\bibitem[{NASA(2021)}]{Nasa-2021-EOD}
NASA (2021).
\newblock \enquote{Earth Observation Data | Earthdata.}
\newblock \url{https://earthdata.nasa.gov/earth-observation-data}.

\bibitem[{Nishihama(1997)}]{Nishihama-1997-ML1}
Nishihama M (1997).
\newblock \enquote{{MODIS} Level 1{A} {E}arth {L}ocation: Algorithm Theoretical
  Basis Document Version 3.0.}
\newblock Available at
  \url{https://modis.gsfc.nasa.gov/data/atbd/atbd_mod28_v3.pdf}, retrieved
  April 1, 2018.

\bibitem[{Ntaimo and Zeigler(2004)}]{Ntaimo-2004-EFC}
Ntaimo L, Zeigler BP (2004).
\newblock \enquote{Expression of a Forest Cell Model in Parallel {DEVS} and
  {Timed Cell-DEVS} Formalisms.}
\newblock In \emph{Proceedings of the 2004 Summer Computer Simulation
  Conference}, pp. 25--29. SCS.
\newblock Available at
  \url{http://www.acims.arizona.edu/PUBLICATIONS/PDF/NtaimoForestCellPaper.pdf}.

\bibitem[{Parks(2014)}]{Parks-2014-MDB}
Parks SA (2014).
\newblock \enquote{Mapping day-of-burning with coarse-resolution satellite
  fire-detection data.}
\newblock \emph{International Journal of Wildland Fire}, \textbf{23}(2),
  215--223.
\newblock \doi{10.1071/WF13138}.

\bibitem[{Pyne \emph{et~al.}(1996)Pyne, Andrews, and Laven}]{Pyne-1996-IWF}
Pyne S, Andrews PL, Laven RD (1996).
\newblock \emph{Introduction to Wildland Fire}.
\newblock Wiley, New York.

\bibitem[{Ram{\'\i}rez \emph{et~al.}(2011)Ram{\'\i}rez, Monedero, and
  Buckley}]{Ramirez-2011-NAF}
Ram{\'\i}rez J, Monedero S, Buckley D (2011).
\newblock \enquote{New approaches in fire simulations analysis with {Wildfire
  Analyst}, Sun City, South Africa, May 9--13, 2011.}
\newblock In \emph{7th International Conference on Forest Fire Research}.

\bibitem[{Reisner \emph{et~al.}(2000)Reisner, Wynne, Margolin, and
  Linn}]{Reisner-2000-CAM}
Reisner J, Wynne S, Margolin L, Linn R (2000).
\newblock \enquote{Coupled atmospheric-fire modeling employing the method of
  averages.}
\newblock \emph{Monthly Weather Review}, \textbf{128}(10), 3683--3691.
\newblock \doi{10.1175/1520-0493(2001)129<3683:CAFMET>2.0.CO;2}.

\bibitem[{Rios \emph{et~al.}(2014)Rios, Jahn, and Rein}]{Rios-2014-FWW}
Rios O, Jahn W, Rein G (2014).
\newblock \enquote{Forecasting wind-driven wildfires using an inverse modelling
  approach.}
\newblock \emph{Natural Hazards and Earth System Sciences}, \textbf{14}(6),
  1491--1503.
\newblock \doi{10.5194/nhess-14-1491-2014}.

\bibitem[{Rochoux \emph{et~al.}(2013)Rochoux, Delmotte, Cuenot, Ricci, and
  Trouv{\'e}}]{Rochoux-2013-RSW}
Rochoux MC, Delmotte B, Cuenot B, Ricci S, Trouv{\'e} A (2013).
\newblock \enquote{Regional-scale simulations of wildland fire spread informed
  by real-time flame front observations.}
\newblock \emph{Proceedings of the Combustion Institute}, \textbf{34},
  2641--2647.

\bibitem[{Rochoux \emph{et~al.}(2014)Rochoux, Emery, Ricci, Cuenot, and
  Trouve}]{Rochoux-2014-TPS}
Rochoux MC, Emery C, Ricci S, Cuenot B, Trouve A (2014).
\newblock \enquote{Towards predictive simulation of wildfire spread at regional
  scale using ensemble-based data assimilation to correct the fire front
  position.}
\newblock \emph{Fire Safety Science}, \textbf{11}, 1443--1456.
\newblock \doi{10.3801/IAFSS.FSS.11-1443}.

\bibitem[{Rothermel(1972)}]{Rothermel-1972-MMP}
Rothermel RC (1972).
\newblock \enquote{A Mathematical Model for Predicting Fire Spread in Wildland
  Fires.}
\newblock {USDA Forest Service Research Paper INT-115}.
\newblock \mbox{\url{https://www.fs.fed.us/rm/pubs_int/int_rp115.pdf}},
  accessed March 2018.

\bibitem[{Rothermel and Rinehart(1983)}]{Rothermel-1983-FPV}
Rothermel RC, Rinehart GC (1983).
\newblock \enquote{Field Procedures for Verification and Adjustment of Fire
  Behavior Predicitions.}
\newblock U.S. Forest Service General Technical Report INT-142. Ogden, UT.
\newblock \url{https://www.fs.fed.us/rm/pubs_int/int_gtr142.pdf}.

\bibitem[{S{\'a} \emph{et~al.}(2017)S{\'a}, Benali, Fernandes, S., Trigo,
  Salis, Russo, Jerez, Soares, Schroeder, and Pereira}]{Sa-2017-EFG}
S{\'a} ACL, Benali A, Fernandes PM, S PRM, Trigo RM, Salis M, Russo A, Jerez S,
  Soares PMM, Schroeder W, Pereira JMC (2017).
\newblock \enquote{Evaluating fire growth simulations using satellite active
  fire data.}
\newblock \emph{Remote Sensing of Environment}, \textbf{190}, 302--317.
\newblock ISSN 0034-4257.
\newblock \doi{10.1016/j.rse.2016.12.023}.

\bibitem[{Schoennagel \emph{et~al.}(2017)Schoennagel, Balch, Brenkert-Smith,
  Dennison, Harvey, Krawchuk, Mietkiewicz, Morgan, Moritz, Rasker, Turner, and
  Whitlock}]{Schoennagel-2017-AMW}
Schoennagel T, Balch JK, Brenkert-Smith H, Dennison PE, Harvey BJ, Krawchuk MA,
  Mietkiewicz N, Morgan P, Moritz MA, Rasker R, Turner MG, Whitlock C (2017).
\newblock \enquote{Adapt to more wildfire in western North American forests as
  climate changes.}
\newblock \emph{Proceedings of the National Academy of Sciences},
  \textbf{114}(18), 4582--4590.
\newblock ISSN 0027-8424.
\newblock \doi{10.1073/pnas.1617464114}.
\newblock \eprint{https://www.pnas.org/content/114/18/4582.full.pdf}.

\bibitem[{Schroeder and Giglio(2017)}]{Schroeder-2017-VII}
Schroeder W, Giglio L (2017).
\newblock \enquote{Visible {I}nfrared {I}maging {R}adiometer {S}uite
  ({V}{I}{I}{R}{S}) 750 m {A}ctive {F}ire {D}etection and {C}haracterization
  {A}lgorithm {T}heoretical {B}asis {D}ocument 1.0.}
\newblock Available at
  \url{https://lpdaac.usgs.gov/sites/default/files/public/product_documentation/vnp14_atbd_v1.pdf},
  retrieved August 28, 2018.

\bibitem[{Schroeder \emph{et~al.}(2008)Schroeder, Prins, Giglio, Csiszar,
  Schmidt, Morisette, and Morton}]{Schroeder-2008-VGM}
Schroeder W, Prins E, Giglio L, Csiszar I, Schmidt C, Morisette J, Morton D
  (2008).
\newblock \enquote{Validation of {GOES} and {MODIS} active fire detection
  products using {ASTER} and {ETM+} data.}
\newblock \emph{Remote Sensing of Environment}, \textbf{112}(5), 2711--2726.
\newblock \doi{10.1016/j.rse.2008.01.005}.

\bibitem[{Skamarock \emph{et~al.}(2008)Skamarock, Klemp, Dudhia, Gill, Liu,
  Berner, Wang, Powers, Duda, Barker, and Huang}]{Skamarock-2019-DAR}
Skamarock WC, Klemp JB, Dudhia J, Gill DO, Liu Z, Berner J, Wang W, Powers JG,
  Duda MG, Barker DM, Huang XY (2008).
\newblock \enquote{A Description of the {A}dvanced {R}esearch {WRF} Version 4.}
\newblock NCAR Technical Note 556.
\newblock \doi{10.5065/1dfh-6p97}.

\bibitem[{S{\o}rensen(1948)}]{Sorenson-1948-MEG}
S{\o}rensen T (1948).
\newblock \emph{A Method of Establishing Groups of Equal Amplitude in Plant
  Sociology}.
\newblock Biologiske skrifter. I kommission hos E. Munksgaard.

\bibitem[{Srivas \emph{et~al.}(2016)Srivas, Art\'{e}s, de~Callafon, and
  Altintas}]{Srivas-2016-WSP}
Srivas T, Art\'{e}s T, de~Callafon RA, Altintas I (2016).
\newblock \enquote{Wildfire Spread Prediction and Assimilation for {FARSITE}
  Using Ensemble Kalman Filtering1.}
\newblock \emph{Procedia Computer Science}, \textbf{80}, 897--908.
\newblock ISSN 1877-0509.
\newblock \doi{10.1016/j.procs.2016.05.328}.
\newblock International Conference on Computational Science 2016, ICCS 2016,
  6-8 June 2016, San Diego, California, USA.

\bibitem[{Stuart(2010)}]{Stuart-2010-IPB}
Stuart AM (2010).
\newblock \enquote{Inverse problems: A {B}ayesian perspective.}
\newblock \emph{Acta Numer.}, \textbf{19}, 451--559.
\newblock \doi{10.1017/S0962492910000061}.

\bibitem[{Sullivan(2009{\natexlab{a}})}]{Sullivan-2009-RWF1}
Sullivan AL (2009{\natexlab{a}}).
\newblock \enquote{A review of wildland fire spread modelling, 1990-present, 1:
  Physical and quasi-physical models.}
\newblock \emph{International Journal of Wildland Fire}, \textbf{18}, 347--368.
\newblock \doi{10.1071/WF06143}.

\bibitem[{Sullivan(2009{\natexlab{b}})}]{Sullivan-2009-RWF}
Sullivan AL (2009{\natexlab{b}}).
\newblock \enquote{A review of wildland fire spread modelling, 1990-present, 1:
  Physical and quasi-physical models, 2: Empirical and quasi-empirical models,
  3: Mathematical analogues and simulation models.}
\newblock \emph{International Journal of Wildland Fire}, \textbf{18}, 1:
  347--368, 2: 369--386, 3: 387--403.
\newblock \doi{10.1071/WF06143, 10.1071/WF06142, 10.1071/WF06144}.

\bibitem[{Sullivan(2009{\natexlab{c}})}]{Sullivan-2009-RWF3}
Sullivan AL (2009{\natexlab{c}}).
\newblock \enquote{A review of wildland fire spread modelling, 1990-present 3:
  Mathematical analogues and simulation models.}
\newblock \emph{International Journal of Wildland Fire}, \textbf{18}, 387--403.
\newblock \doi{10.1071/WF06144}.

\bibitem[{Syifa \emph{et~al.}(2020)Syifa, Panahi, and Lee}]{Syifa-2020-MPB}
Syifa M, Panahi M, Lee CW (2020).
\newblock \enquote{Mapping of Post-Wildfire Burned Area Using a Hybrid
  Algorithm and Satellite Data: The Case of the Camp Fire Wildfire in
  California, USA.}
\newblock \emph{Remote Sensing}, \textbf{12}(4).
\newblock ISSN 2072-4292.
\newblock \doi{10.3390/rs12040623}.

\bibitem[{Trucchia \emph{et~al.}(2020)Trucchia, D’Andrea, Baghino, Fiorucci,
  Ferraris, Negro, Gollini, and Severino}]{Trucchia-2020-POC}
Trucchia A, D’Andrea M, Baghino F, Fiorucci P, Ferraris L, Negro D, Gollini
  A, Severino M (2020).
\newblock \enquote{PROPAGATOR: An Operational Cellular-Automata Based Wildfire
  Simulator.}
\newblock \emph{Fire}, \textbf{3}(3).
\newblock ISSN 2571-6255.
\newblock \doi{10.3390/fire3030026}.

\bibitem[{USDA Forest~Service and Management(2013 (accessed April 20,
  2020))}]{Inciweb-2015-Cougar}
USDA Forest~Service F, Management A (2013 (accessed April 20, 2020)).
\newblock \emph{InciWeb the Incident Information System: Cougar Creek}.
\newblock
  \url{https://web.archive.org/web/20170522031551/https://inciweb.nwcg.gov/incident/4484/}.

\bibitem[{USDA Forest~Service and Management(2015 (accessed April 20,
  2020))}]{Inciweb-2013-Patch}
USDA Forest~Service F, Management A (2015 (accessed April 20, 2020)).
\newblock \emph{InciWeb the Incident Information System: Patch Springs
  Wildfire}.
\newblock
  \url{https://web.archive.org/web/20130825074158/http://inciweb.nwcg.gov/incident/3665}.

\bibitem[{USDA Forest~Service and Management(2018 (accessed April 20,
  2020))}]{Inciweb-2018-Camp}
USDA Forest~Service F, Management A (2018 (accessed April 20, 2020)).
\newblock \emph{Camp Fire Information - InciWeb the Incident Information
  System}.
\newblock
  \url{https://web.archive.org/web/20190228180400/https://inciweb.nwcg.gov/incident/6250/}.

\bibitem[{Vejmelka \emph{et~al.}(2016)Vejmelka, Kochanski, and
  Mandel}]{Vejmelka-2016-DAD}
Vejmelka M, Kochanski AK, Mandel J (2016).
\newblock \enquote{Data assimilation of dead fuel moisture observations from
  remote automatic weather stations.}
\newblock \emph{International Journal of Wildland Fire}, \textbf{25}, 558--568.
\newblock \doi{10.1071/WF14085}.

\bibitem[{Veraverbeke \emph{et~al.}(2014)Veraverbeke, Sedano, Hook, Randerson,
  Jin, and Rogers}]{Veraverbeke-2014-MDP}
Veraverbeke S, Sedano F, Hook SJ, Randerson JT, Jin Y, Rogers BM (2014).
\newblock \enquote{Mapping the daily progression of large wildland fires using
  {MODIS} active fire data.}
\newblock \emph{International Journal of Wildland Fire}, \textbf{23}, 655--667.
\newblock \doi{10.1071/WF13015}.

\bibitem[{Wang \emph{et~al.}(2017)Wang, Bruy\`ere, Duda, Dudhia, Gill,
  Kavulich, Keene, Chen, Lin, Michalakes, Rizvi, Zhang, Berner, Ha, Fossell,
  Beezley, Coen, Mandel, Chuang, McKee, Slovacek, Wolff, and
  Fossell}]{Wang-2017-AUG}
Wang W, Bruy\`ere C, Duda M, Dudhia J, Gill D, Kavulich M, Keene K, Chen M, Lin
  HC, Michalakes J, Rizvi S, Zhang X, Berner J, Ha S, Fossell K, Beezley JD,
  Coen JL, Mandel J, Chuang HY, McKee N, Slovacek T, Wolff J, Fossell K (2017).
\newblock \enquote{{ARW} Version 3 Modeling System User's Guide.}
\newblock Mesoscale \& Miscroscale Meteorology Division, National Center for
  Atmospheric Research.
\newblock
  \url{http://www2.mmm.ucar.edu/wrf/users/docs/user_guide_V3.8/ARWUsersGuideV3.8.pdf},
  retrieved February 2017.

\bibitem[{Warner \emph{et~al.}(2004)Warner, Platt, and Heagy}]{Warner-2004-UTM}
Warner S, Platt N, Heagy J (2004).
\newblock \enquote{User-Oriented Two-Dimensional Measure of Effectiveness for
  the Evaluation of Transport and Dispersion Models.}
\newblock \emph{Journal of Applied Meteorology - J APPL METEOROL}, \textbf{43},
  58--73.
\newblock \doi{10.1175/1520-0450(2004)043<0058:UTMOEF>2.0.CO;2}.

\bibitem[{WildfireAnalyst(2021 (accessed July 5, 2021))}]{WA-2021}
WildfireAnalyst (2021 (accessed July 5, 2021)).
\newblock \enquote{Homepage.}
\newblock \url{https://www.wildfireanalyst.com/}.

\bibitem[{Xue \emph{et~al.}(2012)Xue, Gu, and Hu}]{Xue-2012-DAU}
Xue H, Gu F, Hu X (2012).
\newblock \enquote{Data Assimilation Using Sequential {Monte Carlo} Methods in
  Wildfire Spread Simulation.}
\newblock \emph{ACM Trans. Model. Comput. Simul.}, \textbf{22}(4), 23:1--23:25.
\newblock ISSN 1049-3301.
\newblock \doi{10.1145/2379810.2379816}.

\bibitem[{Xue \emph{et~al.}(2001)Xue, Droegemeier, Wong, Shapiro, Brewster,
  Carr, Weber, Liu, and Wang}]{Xue-2001-ARP}
Xue M, Droegemeier K, Wong V, Shapiro A, Brewster K, Carr FH, Weber D, Liu Y,
  Wang D (2001).
\newblock \enquote{The Advanced Regional Prediction System (ARPS) – A
  multi-scale nonhydrostatic atmospheric simulation and prediction tool. Part
  II: Model physics and applications.}
\newblock \emph{Meteorology and Atmospheric Physics}, \textbf{76}, 143--165.

\end{thebibliography}

\doublespacing

\ucdappendix

\chapter{\uppercase{Workflow and details on the functions}} \label{chapter:07}
This section will add additional detail about the methods used and the computer functions written for this research.

\renewcommand{\thefigure}{\Alph{chapter}.\arabic{figure}}

\section{About the Code and Data}
This appendix offers and overview of some of the Matlab functions used to run the experiments performed during this research. Data is supplied and Matlab scripts are provided.

\subsection{Software Requirements and Dependencies}
\begin{itemize}
\item Operating system: The Matlab code make some system calls that need to be run in a Linux environment.
\item The oldest version of Matlab used in this research was R2017a. 
\item The following Matlab toolboxes are required: Curve Fitting, Mapping, Statistics and Machine Learning, Image Processing Toolbox, and Control System Toolbox.
\end{itemize}

Most of the the Matlab functions used for the task of data assimilation are kept in a Github repository maintained at \url{https://github.com/openwfm/wrf-fire-matlab}. The repository may be cloned and the functions needed exist in the branch 'cycling.' A list of the functions is available at \url{https://github.com/openwfm/wrf-fire-matlab/tree/master/cycling}.

\subsection{Data and Functions for Testing}
Much of the data and the Matlab functions used in this research are available for download. The download link contain one zip file that may be expanded into a directory structure containing functions and scripts that perform various tests or demonstrate some method developed in the thesis.

The data and functions have been added to the Auraria Institutional Repository. The homepage for this repository is \url{https://digital.auraria.edu/air}. The archive is hosted at:

\url{http://digital.auraria.edu/IR00000295/00001}

Search for 'James D. Haley','Assimilation of Satellite Active Fires Data' for possible additional uploads or in the case that the link provided fails. 

The data is also mirrored at the following address:
\url{https://figshare.com/articles/dataset/Assimilation_of_Satellite_Active_Fires_Data_-_Data_and_Code/15044337}

Within the compressed file are two directories.
\begin{enumerate}
\item \textbf{wrf-fire-matlab} This directory contains the Matlab functions that can be found in the 'cycling' branch of the git repository referenced above. Running the script \texttt{startup.m} within this directory will add the paths all sub-directories to the  Matlab environment.
\item \textbf{scripts} This directory contains data and functions to perform some of the tests that were part of this research. The artificial data sets used for a large part of the testing have been included. These data sets in include artificial fire arrival time and artificial fire detections. Another set contains the fire arrival times taken, from WRF-SFIRE output, and the artificial fire detections associated with them. The detection data and WRF-SFIRE output from a simulation of the Patch Springs fire has been included. This directory contains a set of many sub-directories containing individual tests, detailed below. In some cases, Matlab code has been copied into these sub-directories from the main code directory so that default values or graphics options can be set. The testing functions should be run in their respective sub-directories. The following subsections adds details about the contents purpose of those those sub-directories.
\end{enumerate}

\subsubsection{artificial\_data}

The files in this directory run an example case creating an estimated fire arrival time from artificial data that is also created here. The estimate is then assessed and scores computed. This can be run slowly if the artificial fire created is large and has many detection points. 

\subsubsection{data\_assimilation}

The files in this directory run an example case of the data assimilation routine. This shows how adjustment of the forecast fire arrival time is made. Matlab ``versions" of satellite data and files from WRF-SFIRE containing only necessary variables are included in this directory. This can be run quickly.

\subsubsection{estimate\_fire\_arrival\_time}

This creates an estimate of the fire arrival time from satellite data. An output file from WRF-SFIRE sets up the domain and a "Matlab Version" of satellite data is included. Uses the single pass and the multigrid approach for a comparison. 

\subsubsection{test\_artificial}

Testing artificial data files. The script in this directory steps through testing the process for estimating the fire arrival time, using 260 artificial fire arrival times. Plots of the MOE X, MOE Y, Sorenson, relative error are produced. By default, the test loops through only the first 6 artificial fire scenarios, but that can be changed to included all. This testing can take a long time. Testing emulates that from Section \ref{sec:ground_truth_1}, using a single pass method.

\subsubsection{test\_p}

Testing artificial data files. The script in this directory steps through testing,
using 260 artificial fire arrival times, the optimal value of $p$ to use in interpolating additional points along the shortest paths. This can take a long time. Only a few of the artificial data scenarios are tested by default. The testing produces figures comparable to Figure \ref{fig:p_test_artificial_figs} and Figure \ref{fig:p_test_patch_figs} 

\subsubsection{test\_wrf\_output}

Testing WRF-SFIRE output files used as ``ground truth."  The script in this directory steps through testing the process for estimating the fire arrival time, using 115 fire arrival times from WRF-SFIRE. Plots of the MOE X, MOE Y, Sorenson, relative error are produced. By default, the test loops through only the first 6 artificial fire scenarios, but that can be changed to included all. This testing can take a long time. Emulates the testing in Section\ref{wrf_out_testing}.

\section{Workflow}
\subsection{Data Assimilation with FMC Adjustment}

\begin{figure}[!h]
\begin{center}
  \includegraphics[width = 0.45\textwidth]{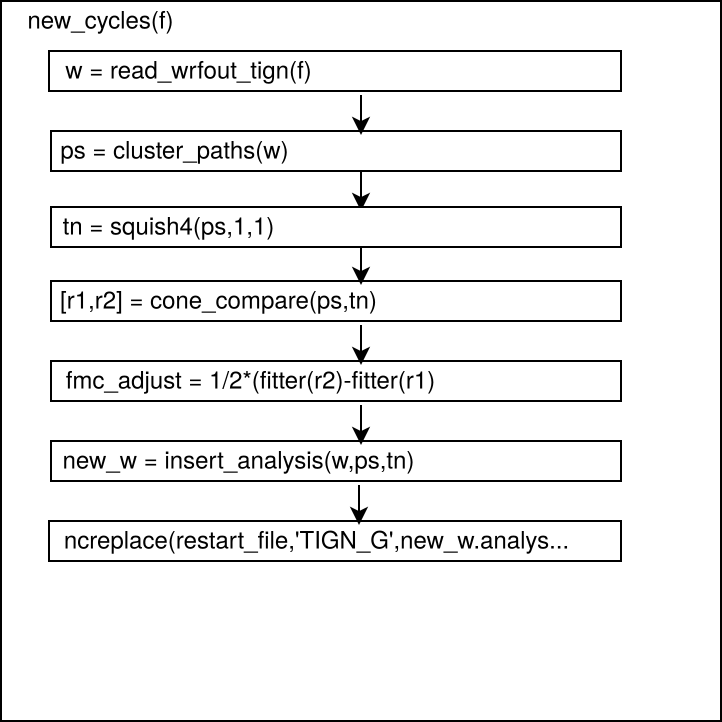}
\caption{Functions used for performing data assimilation with adjustment of fuel moisture. The large box contains the function \texttt{new\_cycles} and the boxes inside are the functions called from within it.}
\label{fig:new_cycles}
\end{center}
\end{figure}

The main function written for data assimilation with adjustment of the FMC is \texttt{new\_cycles}. The function is passed a string containing the relative path to an output file from WRF-SFIRE and gives output by replacing the fire arrival time and changing the FMC content in a rewrite file used by WRF-SFIRE during the next forecast cycle. Figure \ref{fig:new_cycles} shows the how the the function calls other functions that are described in Section \ref{sec:functions}.

\begin{figure}[!h]
\begin{center}
  \includegraphics[width = 0.45\textwidth]{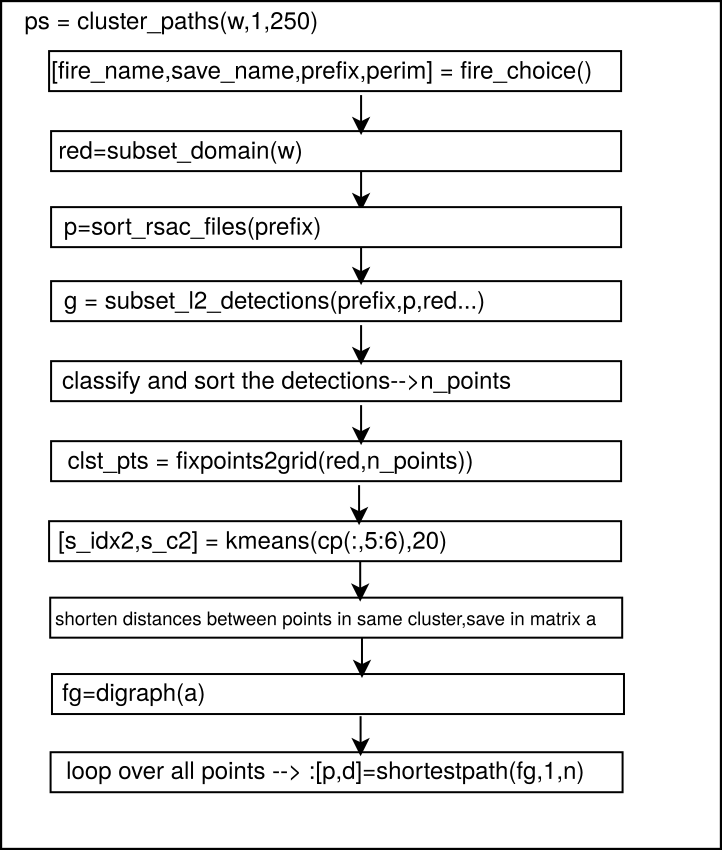}
\caption{Functions used for clustering fire detection data and construction of the shortest paths. The large box contains the function \texttt{cluster\_paths} and the boxes inside are the functions called from within it.}
\label{fig:cluster_paths}
\end{center}
\end{figure}

\subsection{Functions used}
\label{sec:functions}
\begin{enumerate}
\item \begin{verbatim} choose_time_step(f) \end{verbatim}  Helps choose the frame of a wrfout file. The input `f' is a string with path to a wrfout file. Output is a character string with the time of the wrfout frame.

\item \begin{verbatim} cone_compare(ps,tign2) \end{verbatim} Function compares the two fire arrival times. Used for adjusting the ROS in a simulation. Input 'ps' comes from \texttt{ps = cluster\_paths(w,1,250)} and is a struct with a forecast fire arrival time. Input 'tign2' will be the analysis fire arrival time obtained from  \texttt{tign2 = squish4(ps,gq,da)}. Typical use is \texttt{[r1,r2,adjr0,outer] = cone\_compare(ps,tign2)} where 'r1' and 'r2' are the average ROS in the forecast and analysis, respectively. The variable 'adjr0' is an experimental variable for use in adjusting the ROS in the model by the namelist.fire input file. The variable 'outer' is a struct with some statistics about the forecast and the analysis fire arrival times.

\item \begin{verbatim} cluster_paths(w,cull,grid_distance) \end{verbatim} Main function for organizing detection data and creating path structure. Calls many other functions in the process. Input `w' is a Matlab struct from the \texttt{read\_wrfout\_tign} function. The input `cull' can be used to limit the number of active fire detections used, \texttt{cull = 1} uses all detections. The input \texttt{grid\_distance} sets the spacing on the grid used for the interpolation routine. Output is a Matlab struct with fire detection data interpolated to a grid, a directed graph with the detection data, and shortest paths from an assumed ignition point to all other detection locations in the fire domain. Typical use is  \texttt{ps = cluster\_paths(w,1,250)}.

\item \begin{verbatim} detection_probability(tign)\end{verbatim}
The function produces the curve describing the probability of a satellite detecting a fire based on time difference between observation and fire arrival time. Input `tign' is a vector with times before and after fire arrival time.

\item \begin{verbatim} estimate_tign(ps) \end{verbatim}
The function makes makes an estimated fire arrival time from satellite data by drawing a succession of polygons around the active fire detections. Input `ps' is from \texttt{ps = cluster\_paths(w,1,250)}. Output is a matrix storing the fire arrival time. Typical usage is \texttt{t = estimate\_tign(ps)}.

\item \begin{verbatim} filled\_contours(ps,tign_new) \end{verbatim} Function plots contour maps of two fire arrival time. Input 'ps' is a path structure with forecast fire arrival time and input 'tign new' is the analysis fire arrival time in typical usage. no output.

\item \begin{verbatim} fire_choice()\end{verbatim} Selects from different fires simulated in the thesis. Sets up paths where detection and perimeter data is stored. No input required. Output are strings with with paths to fire data, fire name, and other particulars of the simulation.

\item \begin{verbatim}
fire_gradients(lon,lat,tign,unit)
\end{verbatim} Function returns the partial derivatives of the firearrival time. Inputs 'lat,lon,tign' can be taken as the latitude, longitiude, and fire arrival time output by WRF-SFIRE, respectively. Input 'unit' is an integer flag (unit =1 means yes) that returns partial derivatives as components of a unit vecotr. Typical use is \texttt{[dx,dy,nan\_msk] = fire\_gradients(lon,lat,tign,unit)}

\item \begin{verbatim} fixpoints2grid(red,pts) \end{verbatim} Interpolates the active fire detection data onto the computational grid. Calls function ``fixpt" to find nearest grid location for  a given latitude and longitude location. Input 'red' is from \texttt{red = subset\_domain(w)}. Input `pts' is an $n\times2$ matrix with rows containing the latitude and longitude of fire detections. 

\item \begin{verbatim}insert_analysis(w,ps,tn) \end{verbatim} The function creates a new Matlab struct from input `w' containing the variable 'analysis' which is the input `tn' that comes from the data assimilation function \texttt{tn = squish4(ps,1,1)}. Input `ps' is the path structure from \texttt{ps = cluster\_paths(w,1,250)}.

\item \begin{verbatim} interp_paths(ps,p_param)\end{verbatim} Interpolates new points along the shortest paths. Input `ps' is from \texttt{ps = cluster\_paths(w,1,250)}. Input \texttt{p\_param} is the smoothing parameter in the interpolation routine. The default is \texttt{p\_param = 0.9}

\item \begin{verbatim}  make_spline(time_bounds,num_pts)\end{verbatim}
The function produces splines used for making the data likelihood function and its derivatives. Input \texttt{time\_bounds} ,default value 72, is a scalar giving the with of the time window to create the spline. Input \texttt{num\_pts} gives the number of points to use when making the spline.

\item \begin{verbatim}
make_tign(ps,alpha,p_param,grid_fraction)
\end{verbatim}
The function makes an estimated fire arrival time using the iterative interpolation technique from Chapter \ref{chapter:03}. Input `ps' is from \texttt{ps = cluster\_paths(w,1,250)}. Input `alpha' controls blending between initial estimate for the fire arrival time and the interpolated fire arrival time from iterative interpolation. The default is \texttt{alpha = 1}, no weight is given to the initial estimate. 
Input \texttt{grid\_fraction} is used to interpolate data onto a different sized grid. 

\item \begin{verbatim} new_cycles(f) \end{verbatim}
The function runs the cycling routine for data assimilation. Input `f' is a string with path to a wrfout file. Many functions are called. Output is written into files needed to restart a simulation.

\item \begin{verbatim} read_wrfout_tign(f,ts) \end{verbatim}  Reads in a wrfout file to collect variables about the grid, fire arrival time, fuels, and other variables. The input `f' is a string with path to a wrfout file. The input `ts' is optional and is a character string with the time of a wrfout frame. Output is a Matlab struct file with variables describing the domain of the fire. Typical use is \texttt{w=read\_wrfout\_tign(f)}.

\item \begin{verbatim} smooth_up(tign,a,b) \end{verbatim} Function applies Gaussian smoothing to a fire arrival time, with more smoothin applied for earlier times and decreasing linear with the fire arrival time. Typical use is \texttt{sm\_up = smooth\_up(tign,a,b)}. Input 'tign' is a fire arrival time. Input 'a' controls how much smoothing is applied, it is the variance of the Gaussian, default is 300. Input 'b' is the proportion of the varaiance for smoothing the the fire arrival time at the end of the simulation period and the default is 1/3.

\item \begin{verbatim}  sort_rsac_files(prefix)\end{verbatim} Reads a directory with satellite detection data and produces list of fire fire and geolocation products organized by time of observation. Input `prefix' is a string with a path to satellite data products. Typical use is \texttt{p=sort\_rsac\_files(prefix)}.

\item \begin{verbatim} squish4(ps,gq,da)\end{verbatim} This functions performs the iterative interpolation. Output is the fire arrival time. Input 'ps' is from \texttt{ps = cluster\_paths(w,1,250)}. Input 'gq' is an integer flag (gq = 1 means yes) to determine if ground detections will be used.  Input 'da' is an integer flag (da = 1 means yes) whether to use likelihood to perform data assimilation. typical use is \texttt{tign\_new = squish4(ps,gq,da)}.

\item \begin{verbatim} subset_domain(w) \end{verbatim} Can be used to select only the portion of the fire domain where the fire is active. Converts the fire arrival time to datenum format. Input `w' is from \texttt{w=read\_wrfout\_tign(f)}. Typical use is \texttt{red = subset\_domain(w)}.

\item \begin{verbatim}  subset_l2_detections(prefix,p,red,time_bounds,fig)\end{verbatim} Compares the list of all detection data with start and end time of a simulation and outputs a Matlab struct file with active fire detection data for the granule within the fire domain. Input `prefix' is a string with a path to satellite data products. Input `p' is from \texttt{p=sort\_rsac\_files(prefix)}.Input `red' is from \texttt{red = subset\_domain(w)}. Typical use is \texttt{g = subset\_l2\_detections(prefix,p,red,time\_bounds,fig)}.

\item \begin{verbatim} subset_small(red,n,m,full_set)  \end{verbatim} Interpolates the fire grid onto a different size mesh. Uses scattered interpolant methods. Input `red' comes from the \texttt{subset\_domain} function. Inputs `m' and `n' are integers giving the size of the new grid to work on. Input \texttt{full\_set} is an integer indicating whether to interpolate all variables in the struct `red' if \texttt{full\_set = 1}. Output is a Matlab struct like the input `red.' Typical use is \texttt{r = subset\_small(red,n,m,full\_set)}


\end{enumerate}

\begin{figure}[!h]
\begin{center}
  \includegraphics[width = 0.45\textwidth]{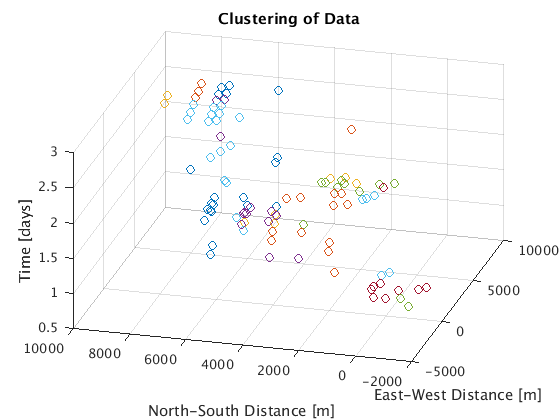}
  \includegraphics[width = 0.45\textwidth]{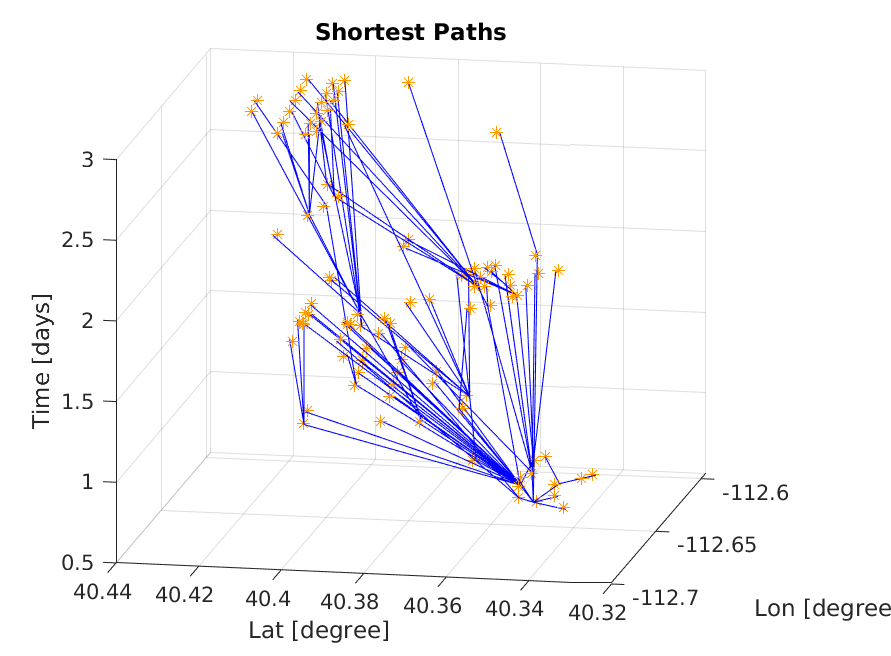}
\caption{Figures generated in the function \texttt{cluster\_paths}. On the left is figure 177, showing clustering of the active fire detection data. On the right figure 178, showing  the paths generated from the ignition to all other detections in the domain.}
\label{fig:cluster_paths_figs}
\end{center}
\end{figure} 

\begin{figure}[!h]
\begin{center}
  \includegraphics[width = 0.45\textwidth]{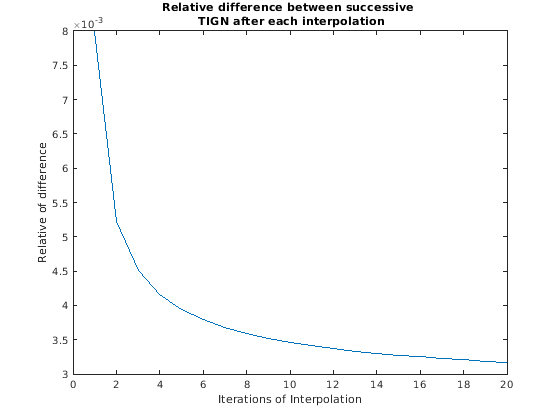}
  \includegraphics[width = 0.45\textwidth]{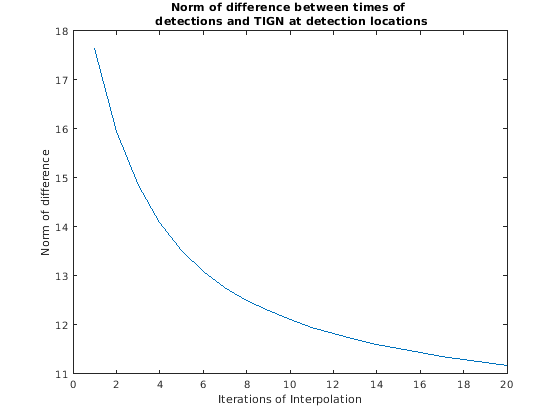}
\caption{On the left is the figure showing the decrease in the relative difference between successive iterations in the interpolation scheme implemented in the \texttt{squish4} function. On the right is the norm of the difference of the time of active fire detections and and the estimated fire arrival time at the detection locations. The figures are produced when running the function \texttt{squish4}.}
\label{fig:squish_converge_figs}
\end{center}
\end{figure}



\clearpage
\FloatBarrier
\chapter{\uppercase{Additional Figures, Tables, and Testing}} \label{chapter:08}
Additional figures, tables, and testing to supplement the discussion and summarize many of the tables of values found in the main text.
\renewcommand{\thefigure}{\Alph{chapter}.\arabic{figure}}
\renewcommand{\thetable}{\Alph{chapter}.\arabic{table}}

\section{Figures and Tables}
\begin{figure}[!h]
  \begin{center}
    \includegraphics[width = 0.45\textwidth]{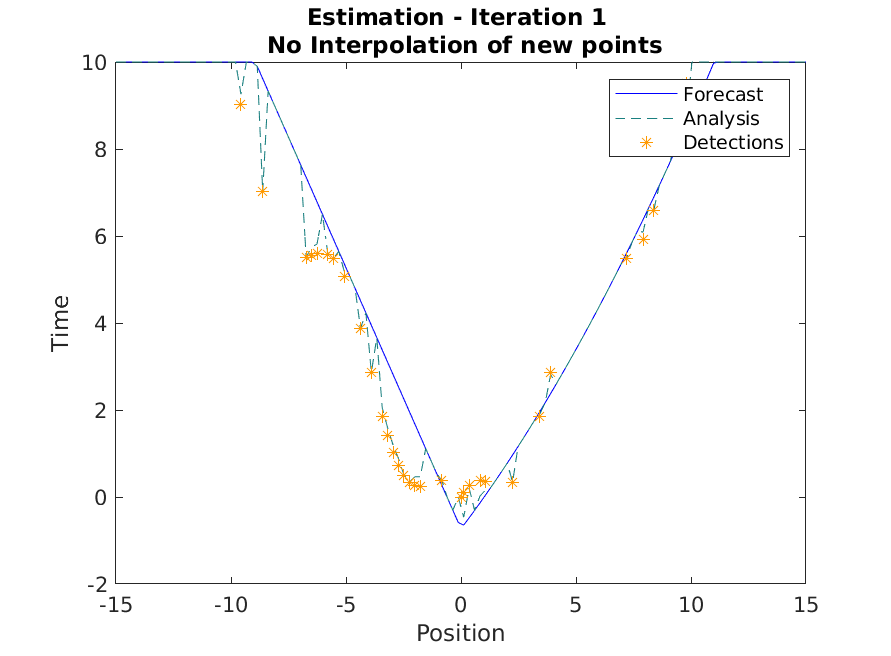}
       \includegraphics[width = 0.45\textwidth]{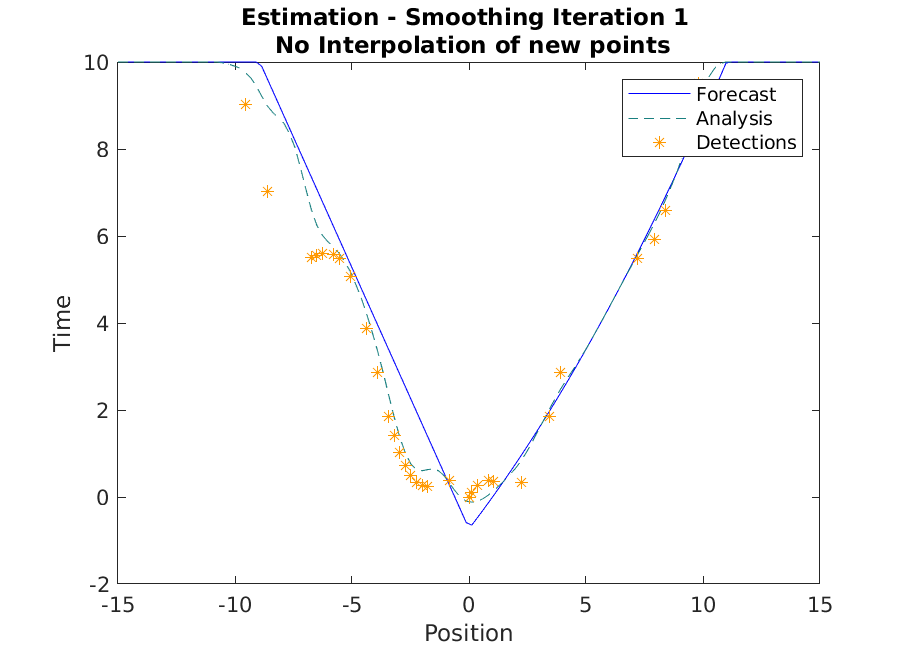}
       
           \includegraphics[width = 0.45\textwidth]{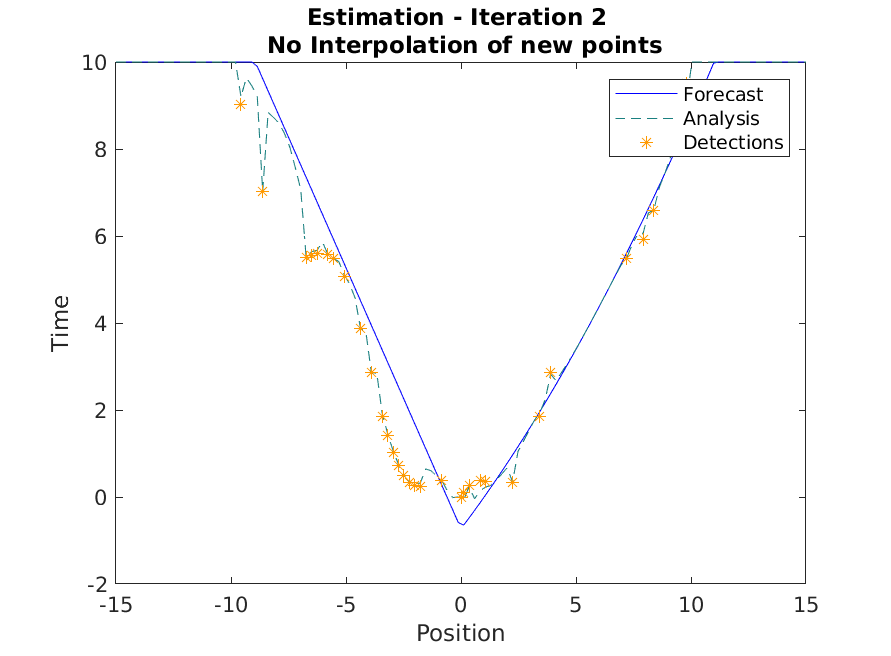}
       \includegraphics[width = 0.45\textwidth]{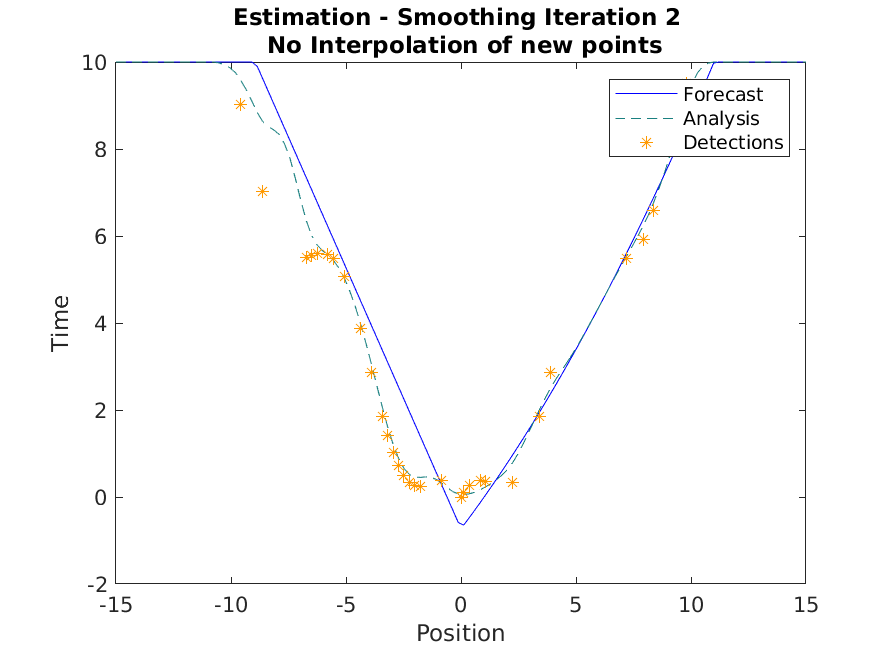}

           \includegraphics[width = 0.45\textwidth]{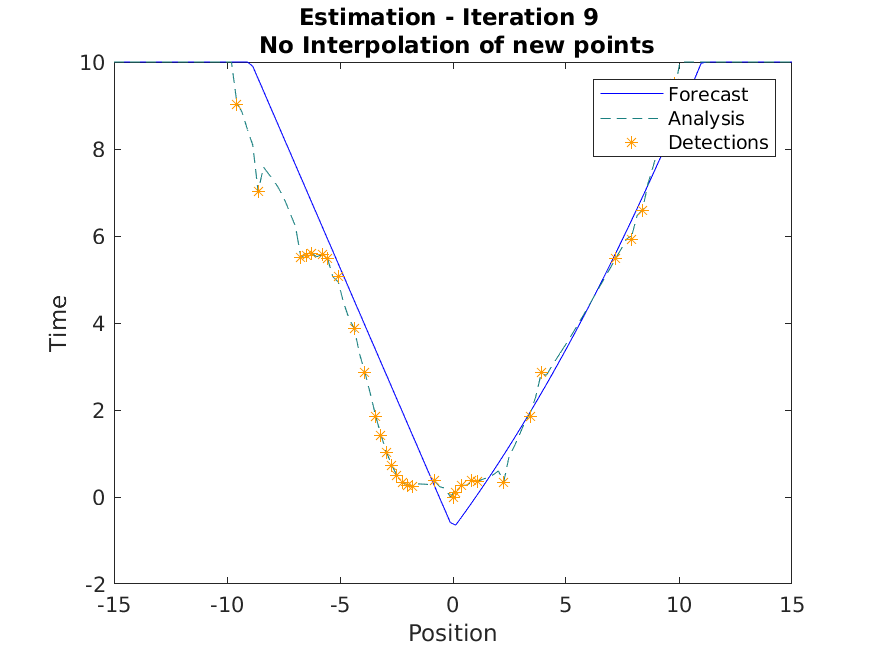}
       \includegraphics[width = 0.45\textwidth]{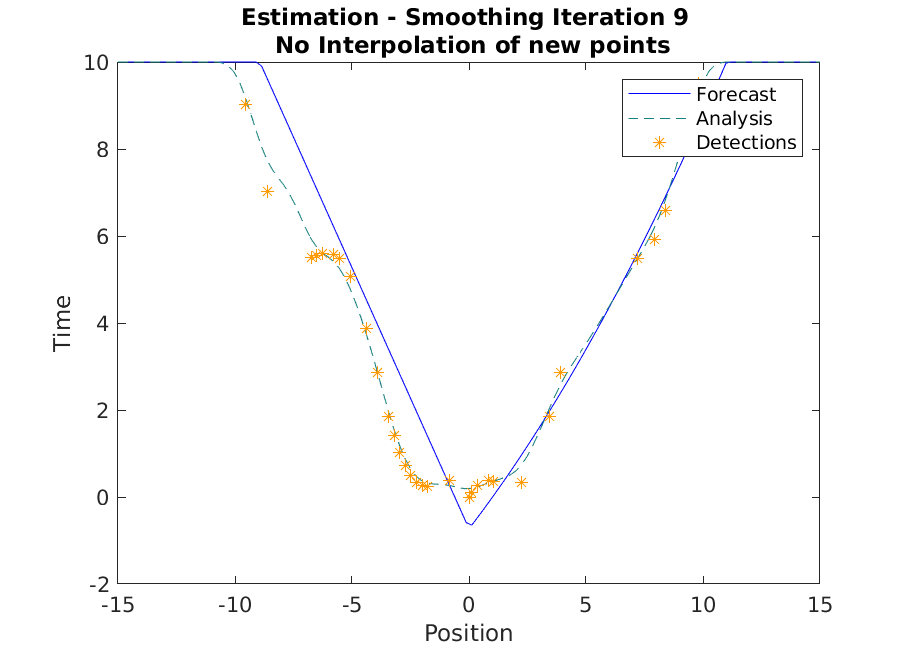}  
    \caption[Iterative interpolation without adding additional points. The sequence of panels shows the process of iterative interpolation applied on a one-dimensional fire line when making an estimated fire arrival time.]{Iterative interpolation. The sequence of panels shows the process of iterative interpolation applied on a one-dimensional fire line when making an estimated fire arrival time. At each detection location an estimate of the fire arrival time is made by adjusting the fire arrival time in places where detection data exists. A Gaussian smoothing is then applied to remove discontinuities in the data and to avoid over-fitting of uncertain data. Each row in the figure shows two iterations of the adjustment and smoothing steps. The left plot of each pair shows the effect of adjusting the fire arrival time at the location of a detection. The roughness introduced by the adjustment is then reduced by the smoothing operation and the result is shown in the right plot of the pair. }
    \label{fig:no_iter_interp}
    \end{center} 
\end{figure}

\begin{figure}[!h]
  \begin{center}
    \includegraphics[width = 0.45\textwidth]{interp_iter_1.png}
       \includegraphics[width = 0.45\textwidth]{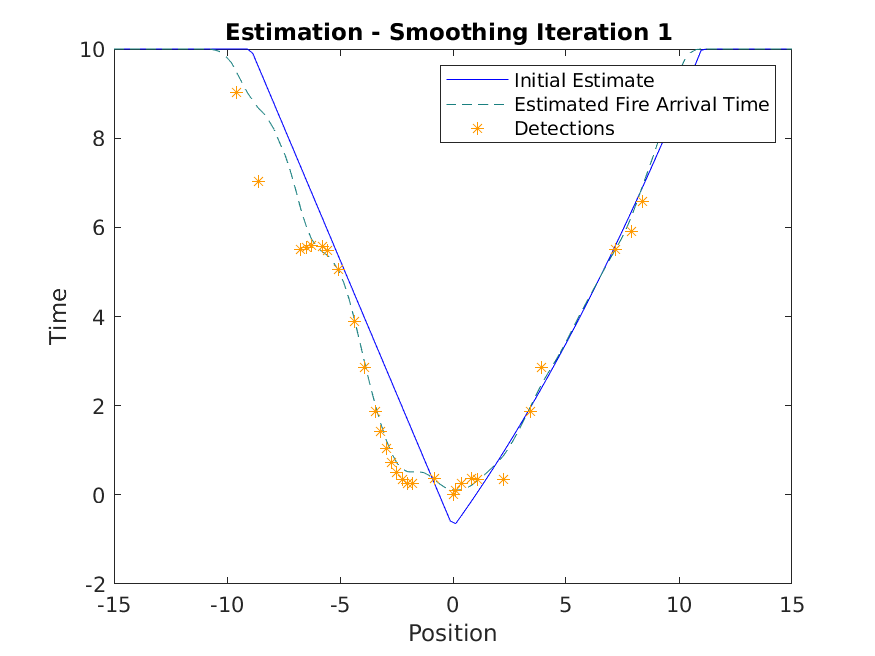}
       
           \includegraphics[width = 0.45\textwidth]{interp_iter_2.png}
       \includegraphics[width = 0.45\textwidth]{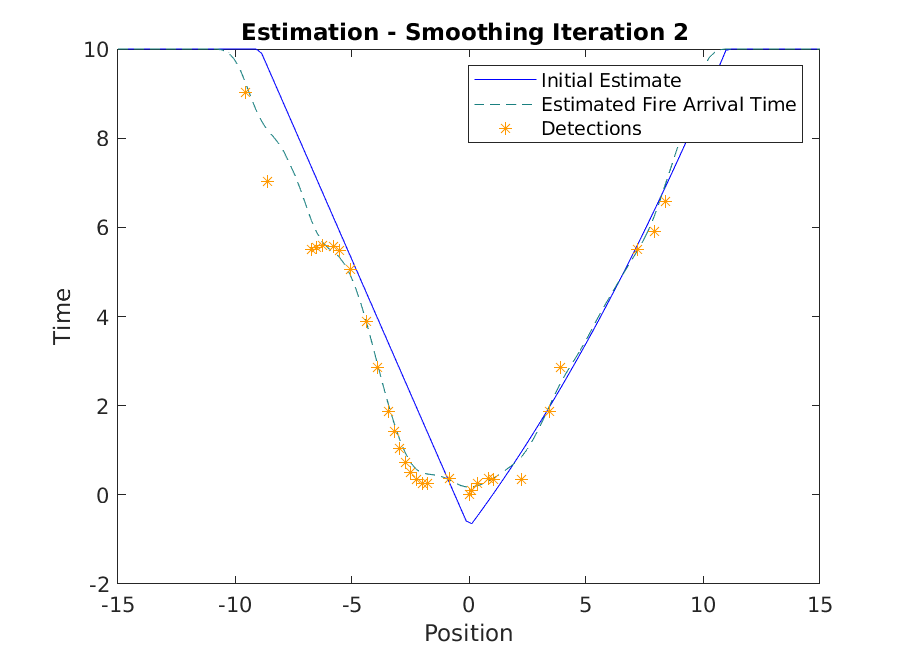}

           \includegraphics[width = 0.45\textwidth]{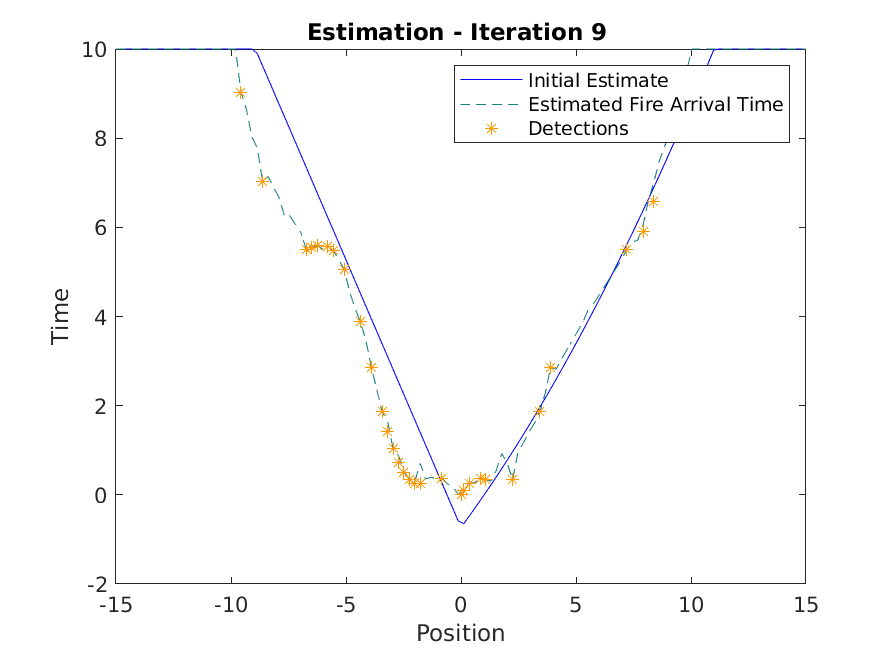}
       \includegraphics[width = 0.45\textwidth]{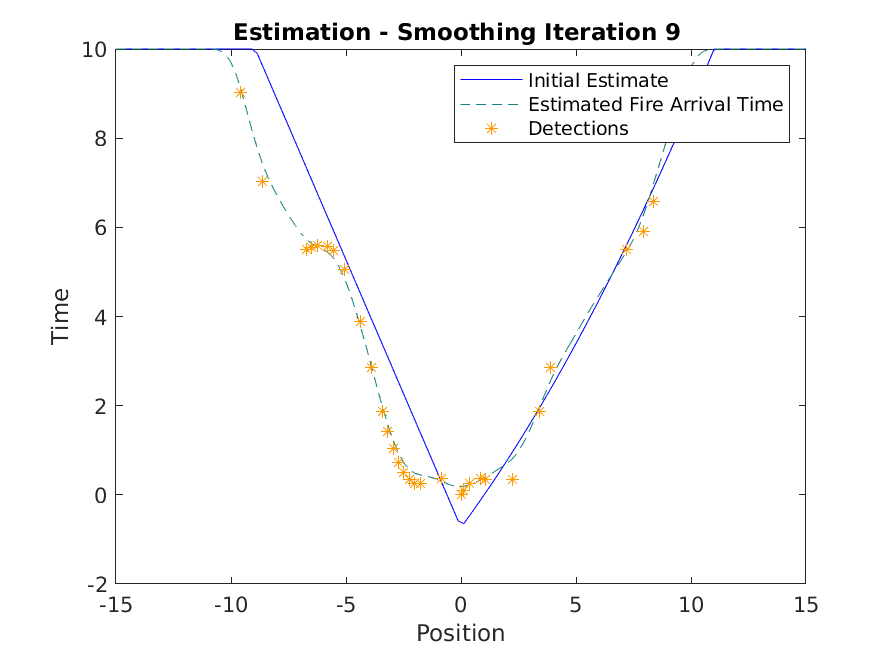}
       
    \caption[Iterative interpolation with adding additional points. The sequence of panels shows the process of iterative interpolation applied on a one-dimensional fire line when making an estimated fire arrival time.]{Iterative interpolation. The sequence of panels shows the process of iterative interpolation applied on a one-dimensional fire line when making an estimated fire arrival time. Extra data points were added at the midpoint between each of the original detections. Compare with the plots in Figure \ref{fig:no_iter_interp}. The addition of more points helps the estimate resolve the gradient information seen on the left side of the figures at approximately time $t = 5.5$.}
    \label{fig:est_iter_interp}
    \end{center} 
\end{figure}
\begin{figure}[!h]
  \begin{center}
    \includegraphics[width = 0.45\textwidth]{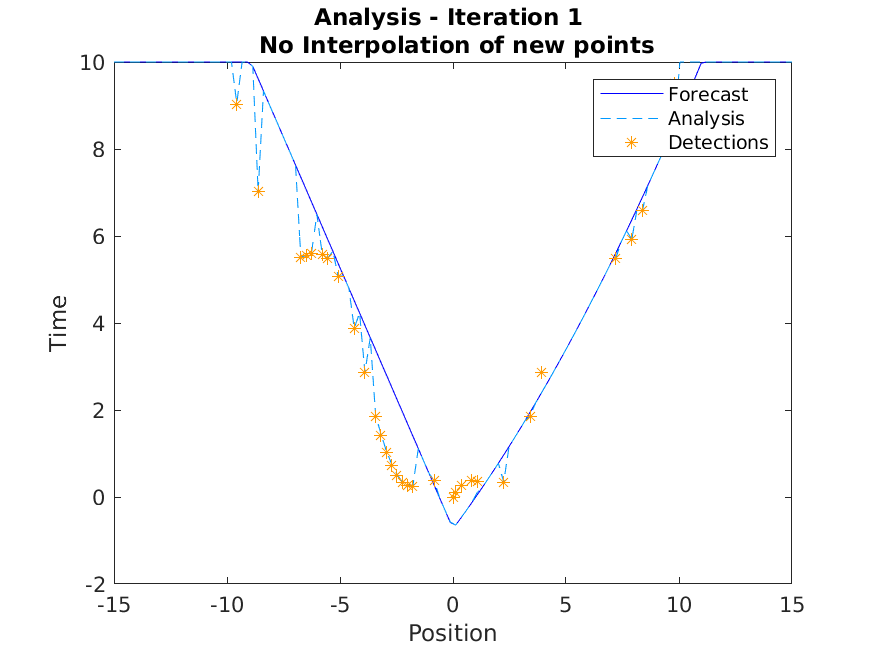}
       \includegraphics[width = 0.45\textwidth]{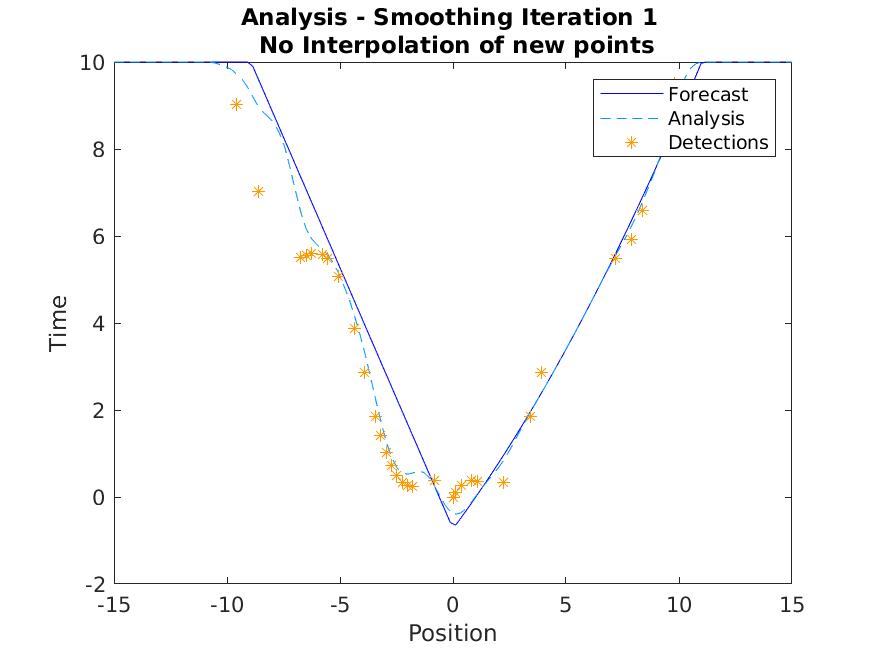}
           \includegraphics[width = 0.45\textwidth]{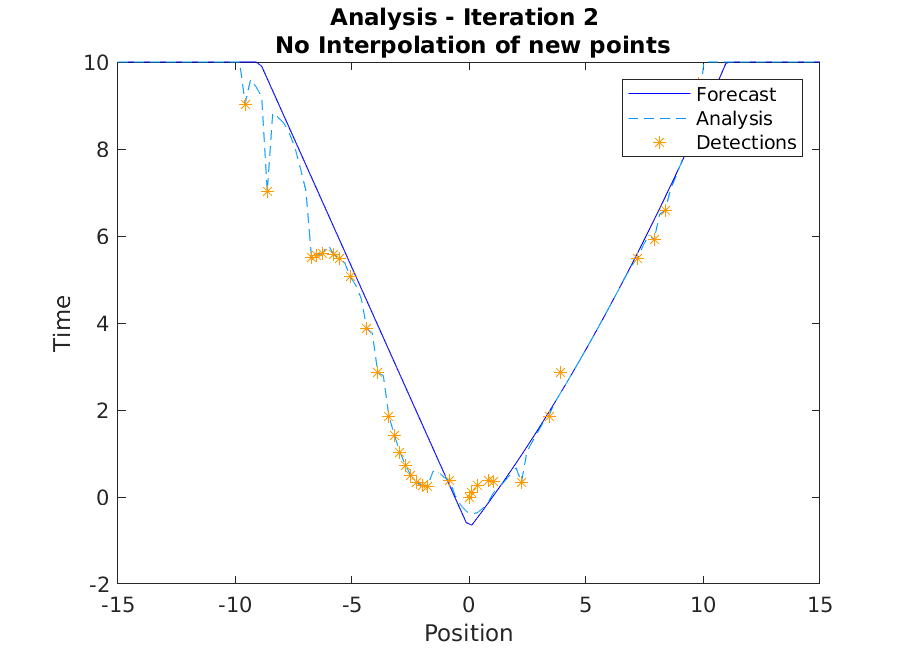}
       \includegraphics[width = 0.45\textwidth]{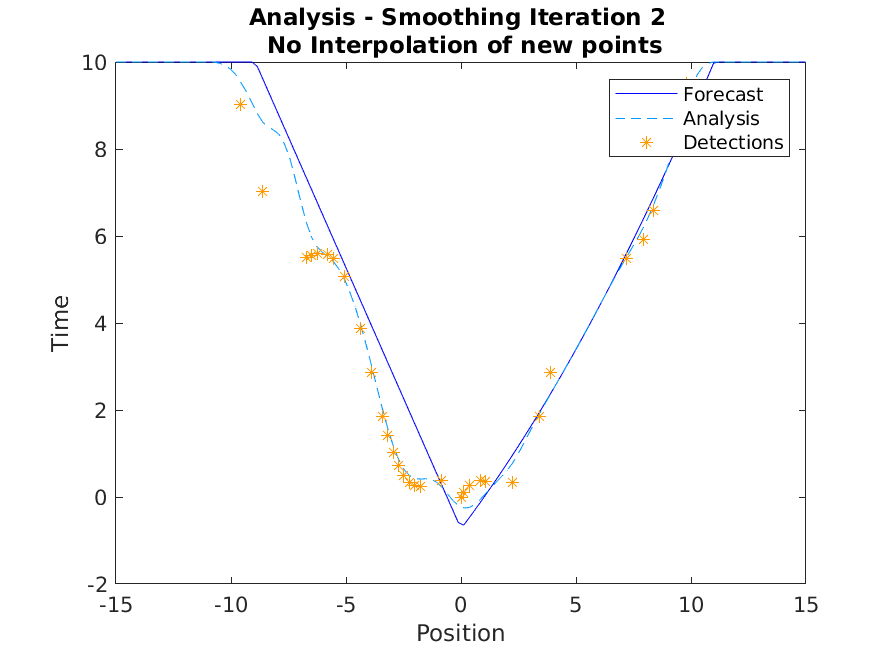}

           \includegraphics[width = 0.45\textwidth]{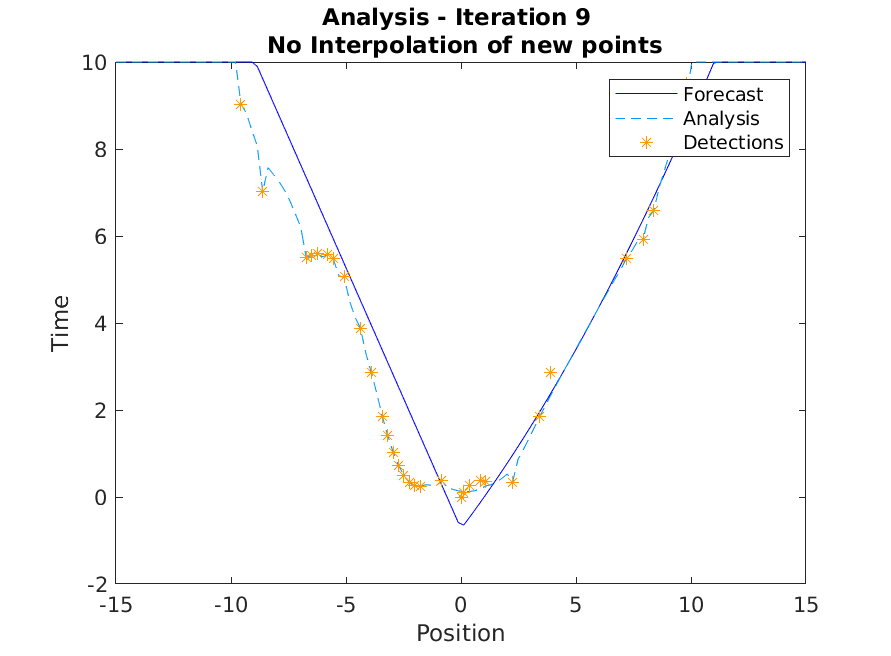}
       \includegraphics[width = 0.45\textwidth]{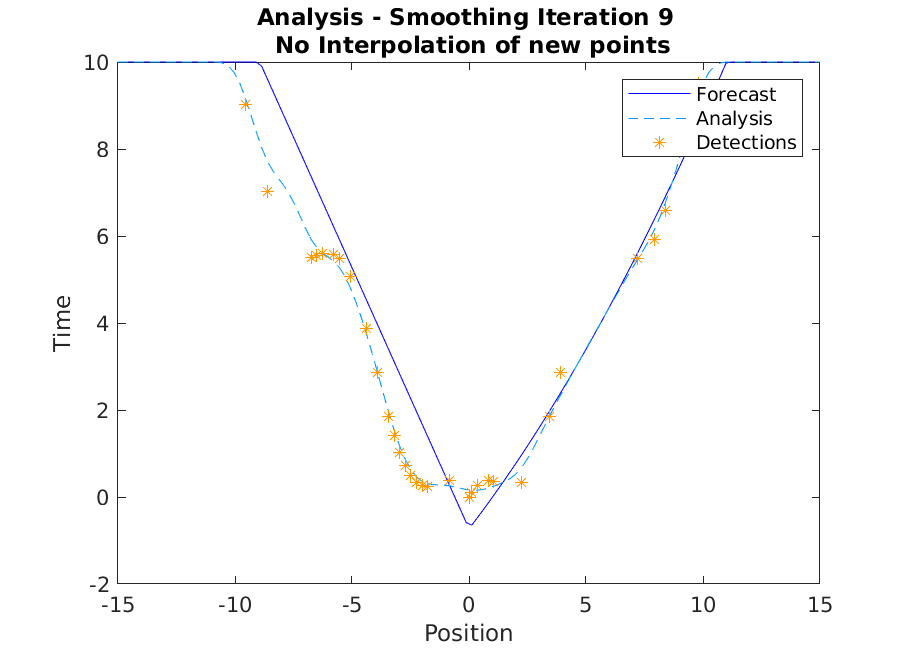}
       
    \caption[Iterative interpolation without adding additional points. The sequence of panels shows the process of iterative interpolation applied on a one-dimensional fire line when using detection data to adjust the fire arrival time.]{Iterative interpolation without adding additional points. The sequence of panels shows the process of iterative interpolation applied on a one-dimensional fire line when using detection data to adjust the fire arrival time. Each row of of plots in the figure shows two iterations of the adjustment and smoothing steps. The left plot of each pair shows the effect of adjusting the fire arrival time at the location of a detection. The roughness introduced by the adjustment is then reduced by the smoothing operation and the result is shown in the right plot of the pair. Note that in the first panel in the upper left that the data likelihood has allowed for large changes downward in the fire arrival time but only small changes, if any, upwards.}
    \label{fig:ana_no_iter_interp}
    \end{center} 
\end{figure}

\begin{figure}[!h]
  \begin{center}
    \includegraphics[width = 0.45\textwidth]{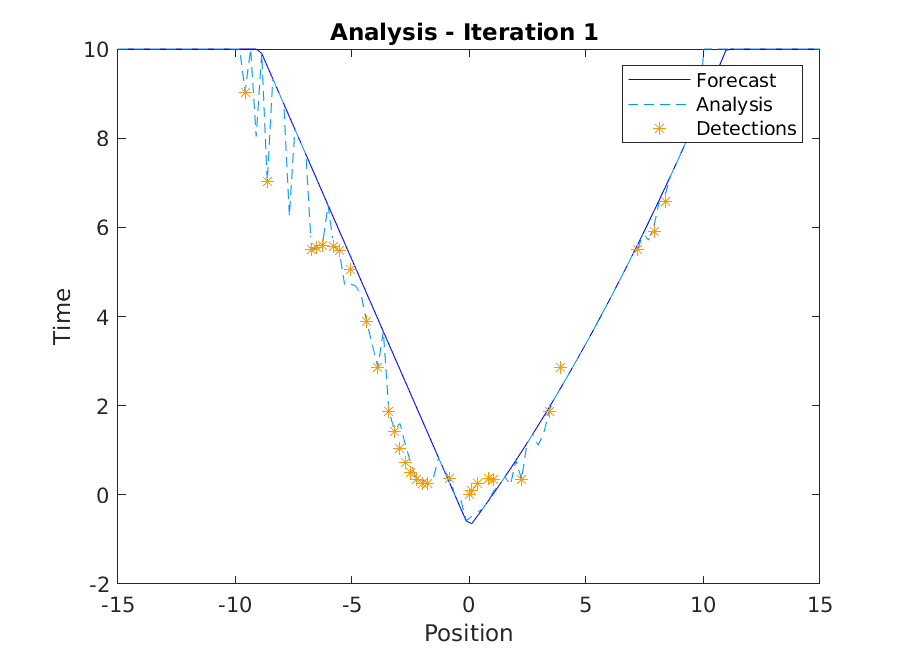}
       \includegraphics[width = 0.45\textwidth]{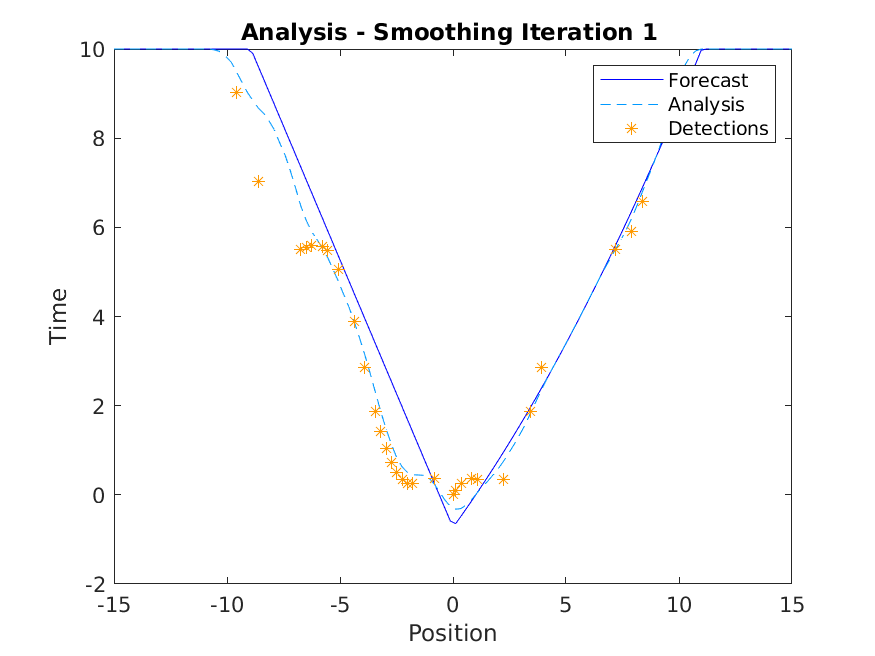}
       
           \includegraphics[width = 0.45\textwidth]{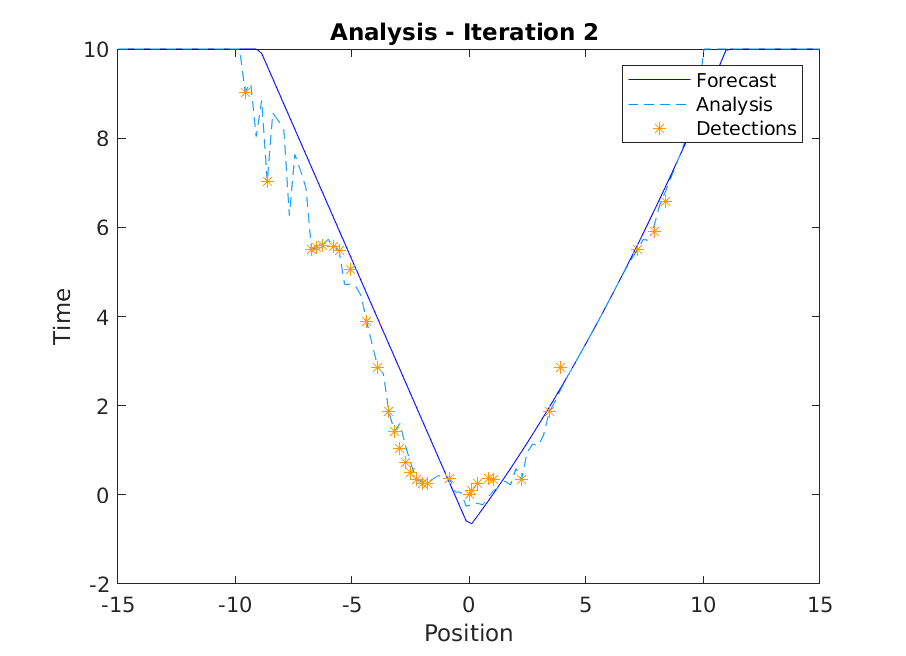}
       \includegraphics[width = 0.45\textwidth]{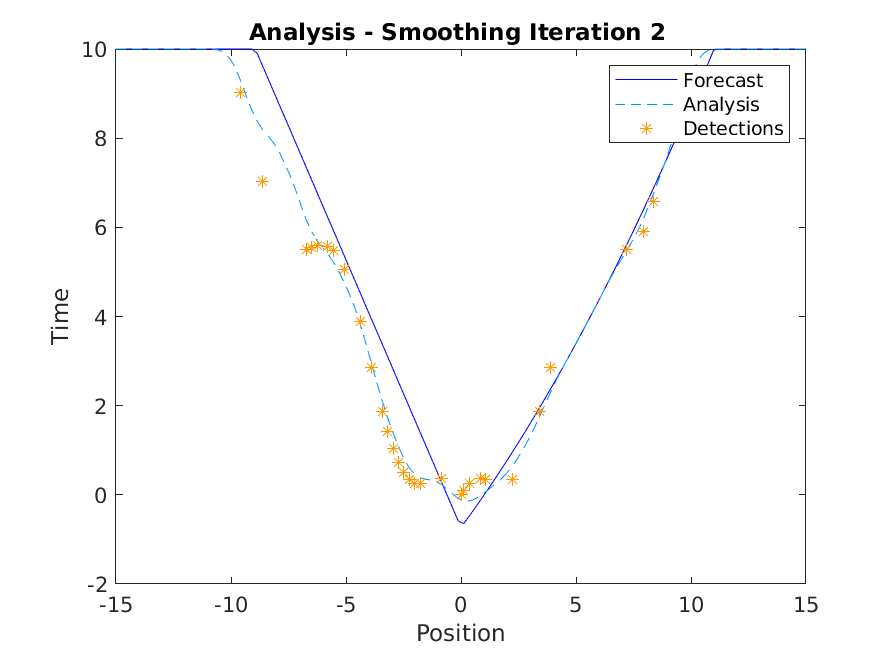}

           \includegraphics[width = 0.45\textwidth]{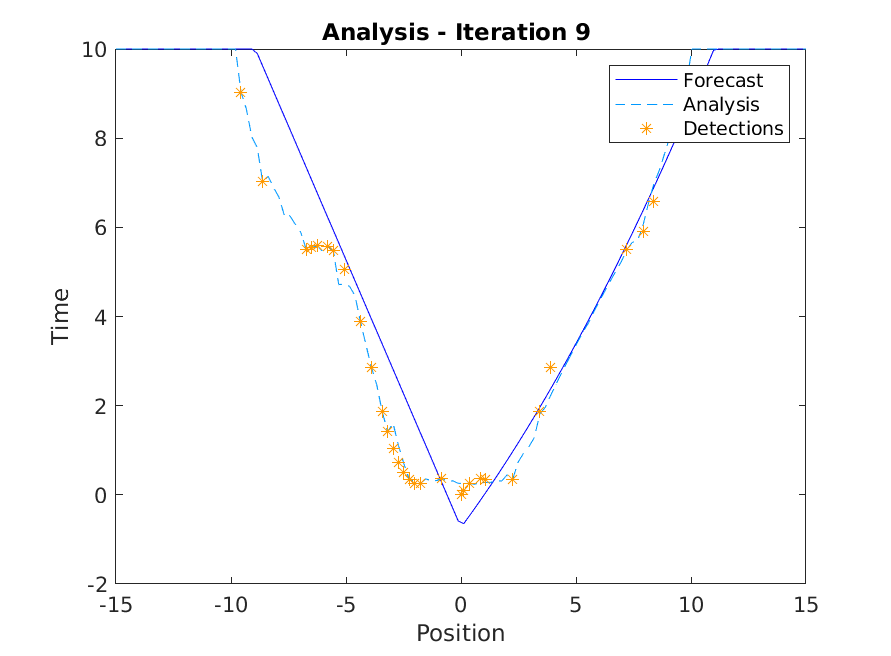}
       \includegraphics[width = 0.45\textwidth]{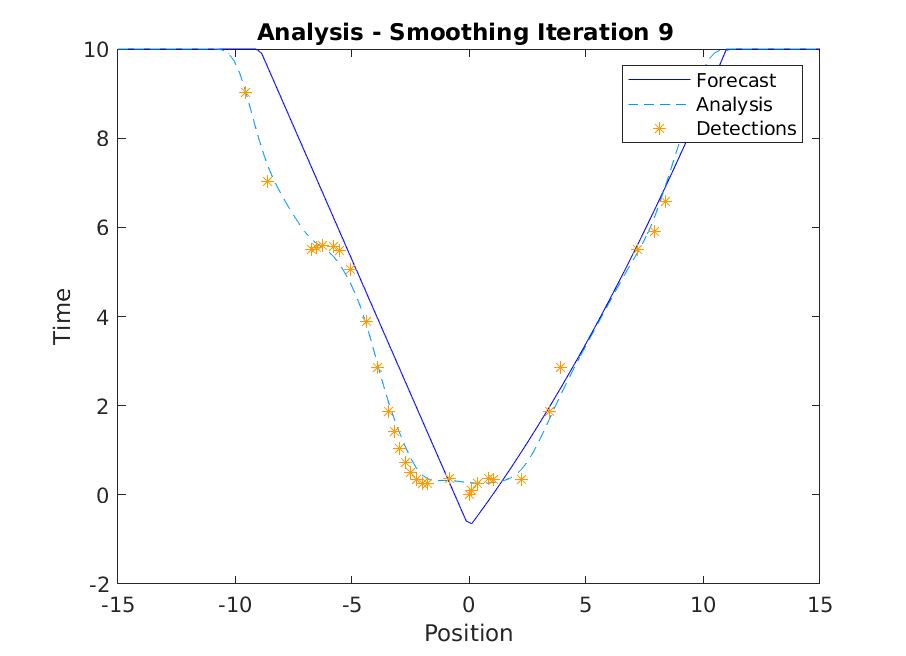}
       
    \caption[Iterative interpolation with adding additional points. The sequence of panels shows the process of iterative interpolation applied on a one-dimensional fire line when when using detection data to adjust the fire arrival time. Extra data points were added at the midpoint between each of the original detections.]{Iterative interpolation with adding additional points. The sequence of panels shows the process of iterative interpolation applied on a one-dimensional fire line when when using detection data to adjust the fire arrival time. Extra data points were added at the midpoint between each of the original detections. Compare with the plots in Figure \ref{fig:ana_no_iter_interp}. The addition of more points helps the estimate resolve the gradient information seen on the left side of the figures at approximately time $t = 5.5$.}
    \label{fig:ana_iter_interp}
    \end{center} 
\end{figure}

\begin{figure}[!ht]
\centering
\includegraphics[width = 0.45\textwidth]{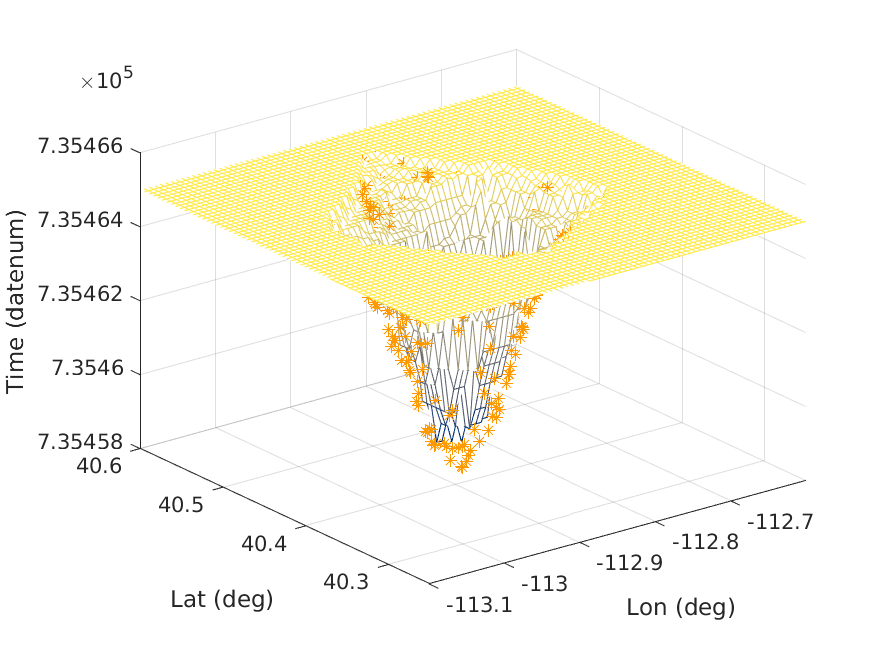}
\includegraphics[width = 0.45\textwidth]{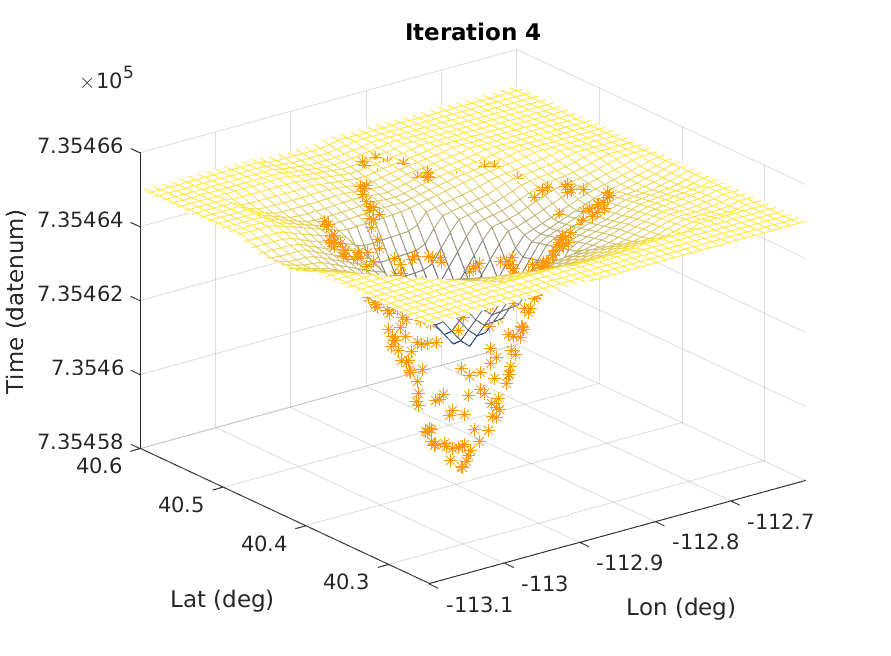}

\includegraphics[width = 0.45\textwidth]{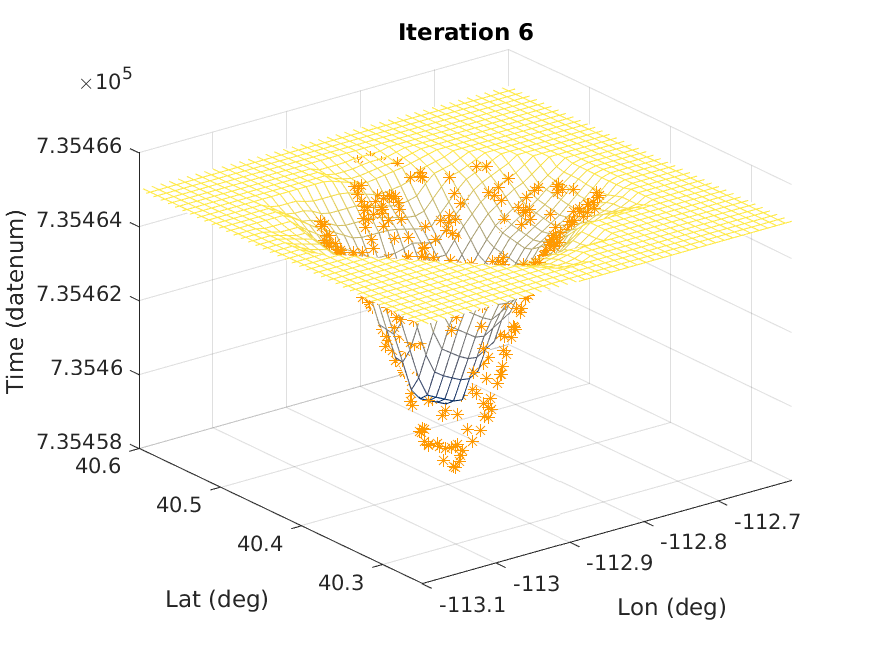}
\includegraphics[width = 0.45\textwidth]{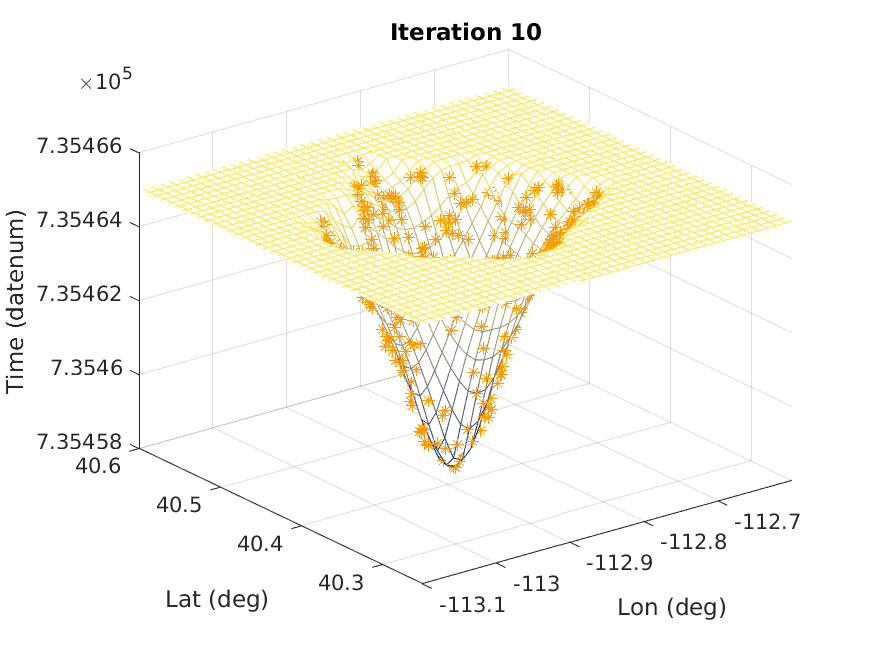}

     \caption[The sequence of plots illustrates the multigrid approach to estimating fire arrival time from satellite data.]{The sequence of plots illustrates the multigrid approach to estimating fire arrival time from satellite data. Each plot the figure shows an estimated fire arrival time and the satellite detection data interpolated onto a grid with 250 m spacing. In the upper left, the surface is an initial estimate of the fire arrival time made by drawing polygons around the detection active fire detections. The remaining figures show the estimates of the fire arrival time after being made on computational grids of decreasing grid spacing. The initial estimates are smooth and flat because interpolation of the satellite data onto a coarse grid causes the interpolation method to change the fire arrival time over as large geographic area around each detection. As the method decreases the working grid size, the fire arrival time in the area around the detections is moved closer to the time those detections were recorded.  }
\label{fig:multigrid_progression}
\end{figure}

\begin{figure}[!h]
\begin{center}
\includegraphics[scale = 0.5]{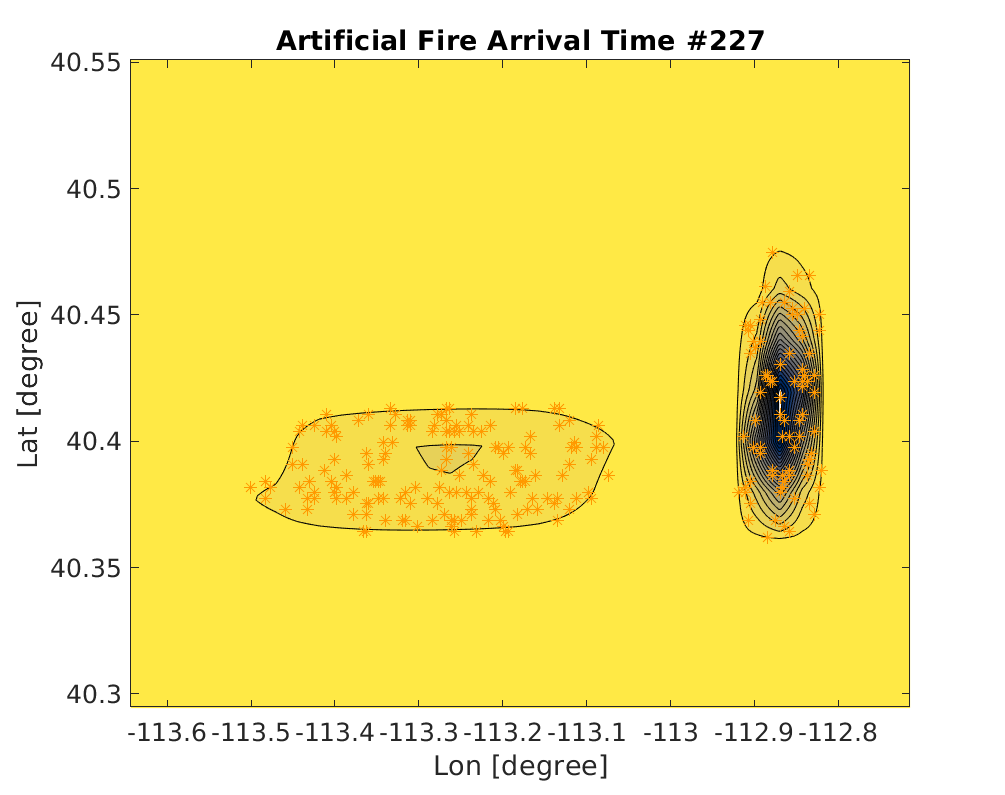}
  \caption[An example of a scenario for which the method cannot produce a satisfactory fire arrival time.]{An example of a scenario for which the method cannot produce a satisfactory fire arrival time. In all testing scenarios, the estimated ROS for this configuration was the worst. In the case of real detections, the data would suggest there are in fact two fires in the domain.}
  \label{fig:fat_227}
  \end{center}
\end{figure}

\begin{table}
\begin{tabular}{|c|c|c|c|c|c|c|c|c|c|}
\hline 
Scenrario & AGE & MRE & SRE & MOE X  & MOE Y  & ||MOE|| & Sorenson & Best & Worst \\ 
\hline 
Multi - NP & 0.1024 & 0.0083 & 0.0059 & 0.9801& 
0.7960 & 1.2656 & 0.8741 & 198 & 88 \\ 
\hline 
Multi - 500 & 0.2341 & 0.0034 & 0.0019 & 0.5713 &
0.9764 & 1.1407 & 0.7065 & 145 & 155 \\ 
\hline 
Multi - 1000 & 0.1154& 0.0043 & 0.0012 & 0.9610 &
0.8358 & 1.2755 & 0.8913 & 112 & 252 \\ 
\hline 
Multi - 2000 & 0.0045 & 0.0041 & 0.0019 & 0.9612 & 0.8421 & 1.2806 & 0.8938 & 70 & 220\\ 
\hline 
Multi - 3000 & 0.1156 & 0.0044 & 0.0013 & 0.9601 & 
0.8364 & 1.2755 & 0.8907 & 69 & 88 \\ 
\hline 
Multi - 4000 & 0.1971 & 0.0046 & 0.0013 & 0.9578 & 0.8362 & 1.2730 & 0.8907 & 54  & 88\\ 
\hline 
Single - NP & 0.1407  & 0.0162 & 0.0105 & 0.9720 & 0.7642 & 1.2395 & 0.8511 & 190 & 159 \\ 
\hline 
Single - 1000 & 0.3598 & 0.0163 & 0.0105 & 0.9642 &  0.7750 & 1.2426 & 0.8552 & 204 & 159 \\ 
\hline 
\end{tabular}
\caption[Averages of the assessment scores made for the 280 estimated fire arrival times generated from artificial data taken to be ``ground truth" for the purpose of comparison.]{Averages of the assessment scores made for the 280 estimated fire arrival times generated from artificial data taken to be ``ground truth" for the purpose of comparison. The columns for best and worst indicate the specific artificial data scenarios for which each strategy produced the best or worst estimate. There is little agreement between strategies in terms of which scenarios produced the best or worst estimates.}
\label{tbl:multi_test}
\end{table}

\begin{table}
\begin{tabular}{|c|c|c|c|c|c|c|c|c|}
\hline 
Scenario & MRD & SRD & MDD & SDD \\ 
\hline 
Multi - NP & -0.0348 & 0.1785 & 0.00089 & 0.1551 \\ 
\hline 
Multi - 500 & -0.0236 & 0.1753 & 0.0015 & 0.1524 \\ 
\hline 
Multi - 1000 & 0.0340 & 0.1772 & 0.0025 & 0.1568 \\ 
\hline 
Multi - 2000 & -0.0187 & 0.1716 & -0.00024 & 0.1568 \\ 
\hline 
Multi - 3000 & 0.0373 & 0.1803 & 0.0021 & 0.1294  \\ 
\hline 
Multi - 4000 & 0.0374 & 0.1797 & 0.0026 & 0.1259 \\ 
\hline 
Single - NP & -0.0852 & 0.1579 & 0.0020 & 0.1750  \\ 
\hline 
Single - 1000 & -0.0840 & 0.1645 & 0.0035 & 0.1800 \\ 
\hline 
\end{tabular} 
\caption[Results for the experiment computing the ROS and the direction of the gradient using 280 artificial fire arrival times as the ``ground truth" fire arrival time.]{Results for the experiment computing the ROS and the direction of the gradient using 280 artificial fire arrival times as the ``ground truth" fire arrival time.   The multigrid approach using additional points with 2000 meter spacing had the best performance in matching the true ROS of the simulated fire. }
\label{tbl:ros_test}
\end{table}

\begin{figure}[!h]
\begin{center}
\includegraphics[width = 0.45\textwidth]{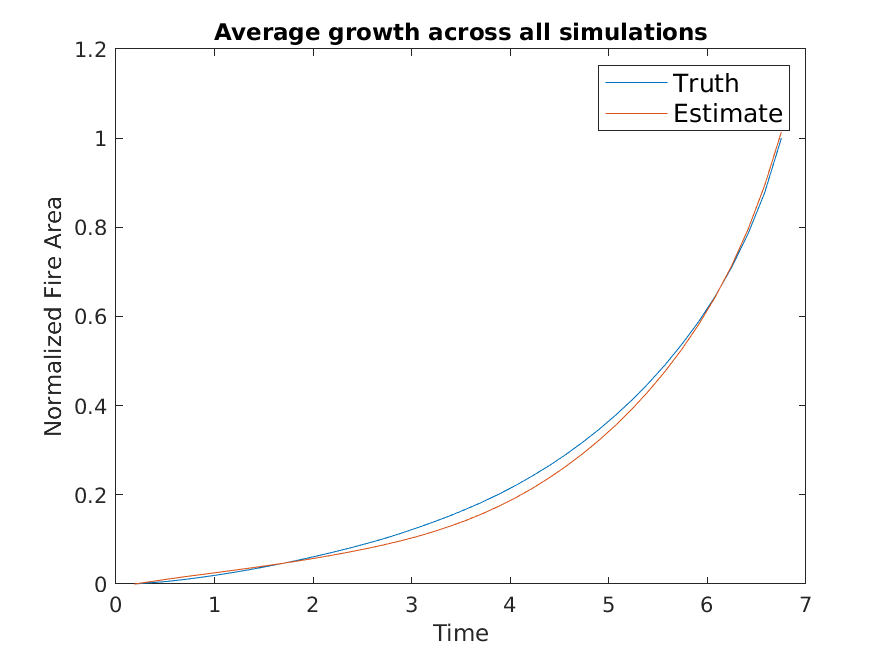}
\includegraphics[width = 0.45\textwidth]{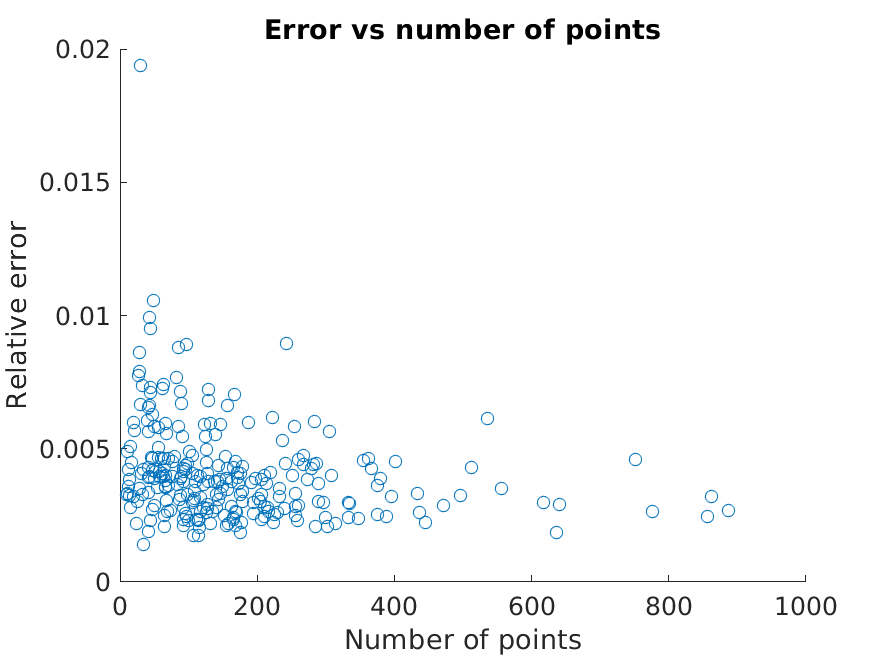}

\includegraphics[width = 0.45\textwidth]{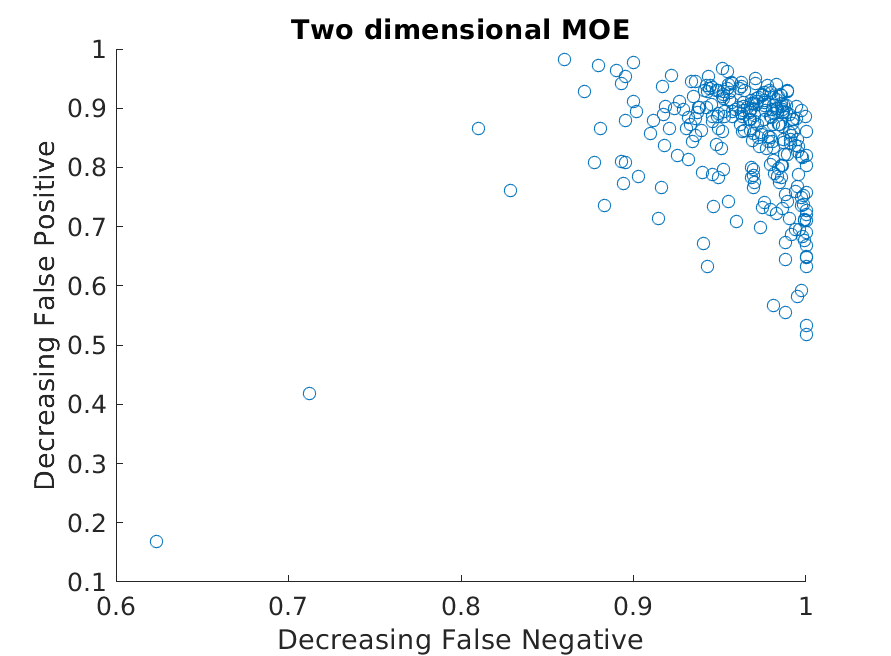}
\includegraphics[width = 0.45\textwidth]{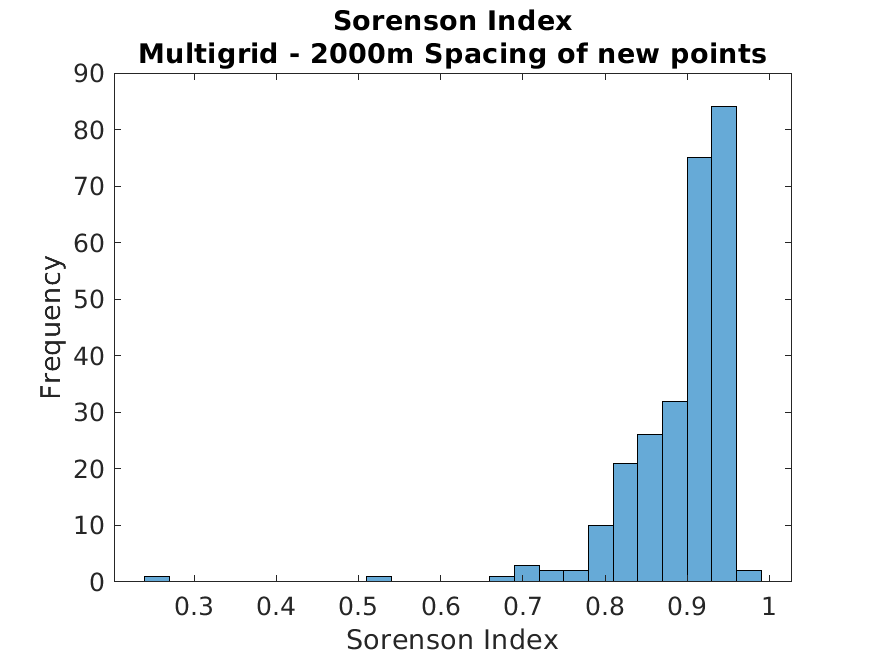}

\includegraphics[width = 0.45\textwidth]{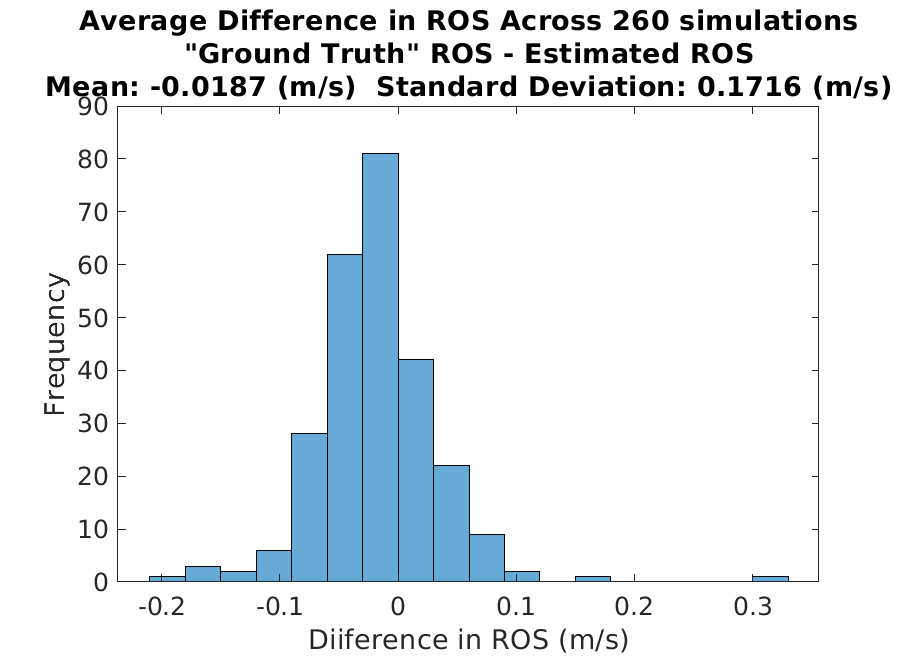}
\includegraphics[width = 0.45\textwidth]{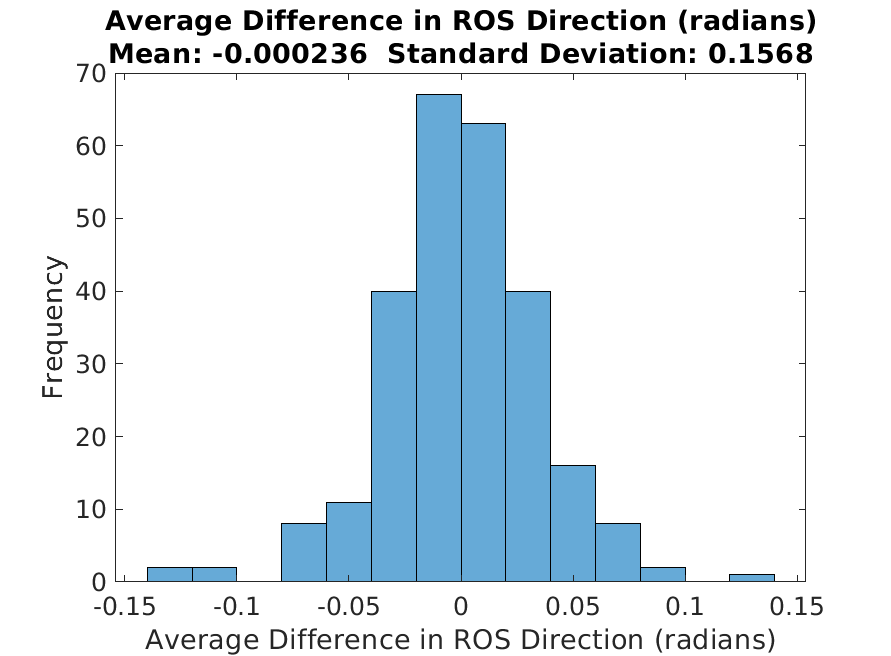}
  \caption[Test results from estimating the fire arrival time using the multigrid strategy with interpolation of extra points along the paths at a spacing of 2000 meters. This strategy produced the best results among those tested.]{Test results from estimating the fire arrival time using the multigrid strategy with interpolation of extra points along the paths at a spacing of 2000 meters. This strategy produced the best results among those tested. The upper left shows how the average growth rate of the estimated fire arrival time closely matched that of the ``ground truth." The upper right shows show the relative errors. The  middle left and right panels show the MOE and the  S{\o}renson index scores, respectively. The lower  right panels on the bottom show histograms of the comparison of the difference in the average ROS and the angle of the gradient for the simulations.}
  \label{fig:new_test_2000}
  \end{center}
\end{figure}

\begin{figure}[htbp]
\centering
          \includegraphics[width = 0.60\textwidth]{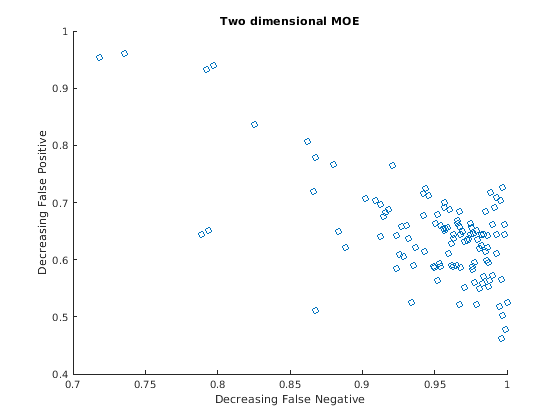}
 \caption[Scatter plot of the MOE for 115 estimated fire arrival times generated from artificial detections scattered on WRF-SFIRE simulation outputs.]{Scatter plot of the MOE for 115 estimated fire arrival times generated from artificial detections scattered on WRF-SFIRE simulation outputs. Most of the points have have a first coordinate close to 1, indicating a tendency to overestimate the size of the actual fire.}
 \label{fig:moe_wrf_test}
\end{figure}

\begin{figure}[htbp]
\centering
          \includegraphics[width = 0.60\textwidth]{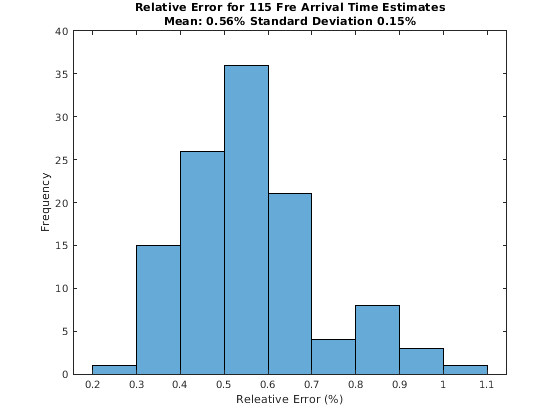}
 \caption{Histogram of the relative errors for 115 estimated fire arrival times generated from artificial detections scattered on WRF-SFIRE simulation outputs. The mean of the errors was 0.56\% and the standard deviation was 0.15\%. }
 \label{fig:re_wrf_test}
\end{figure}

\begin{figure}[htbp]
\centering
\includegraphics[width=0.60\textwidth]{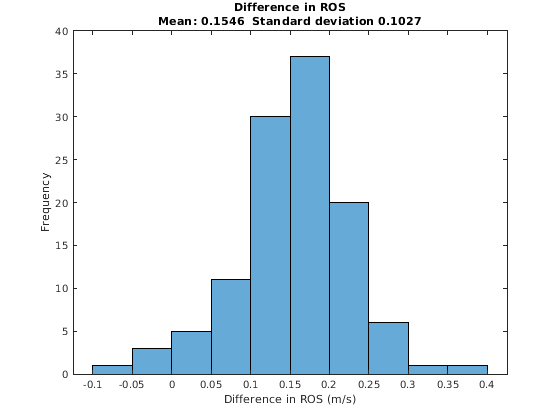}
 \caption{Histogram of the errors in estimated ROS  for 115 estimated fire arrival times generated from artificial detections scattered on WRF-SFIRE simulation outputs.}
 \label{fig:ros_wrf_test}
\end{figure}

\begin{figure}[htbp]
\centering
\includegraphics[width=0.45\textwidth]{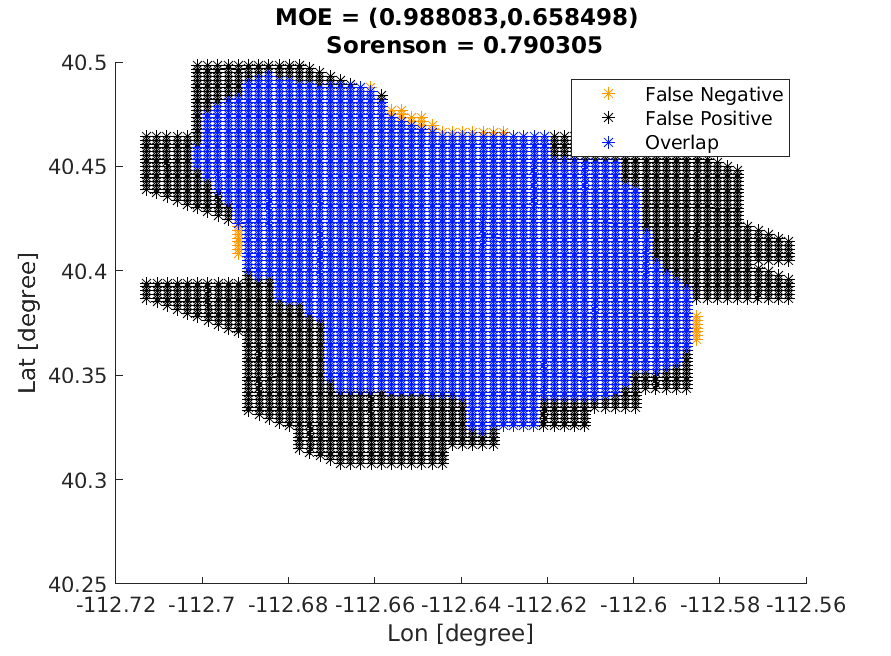}
\includegraphics[width=0.45\textwidth]{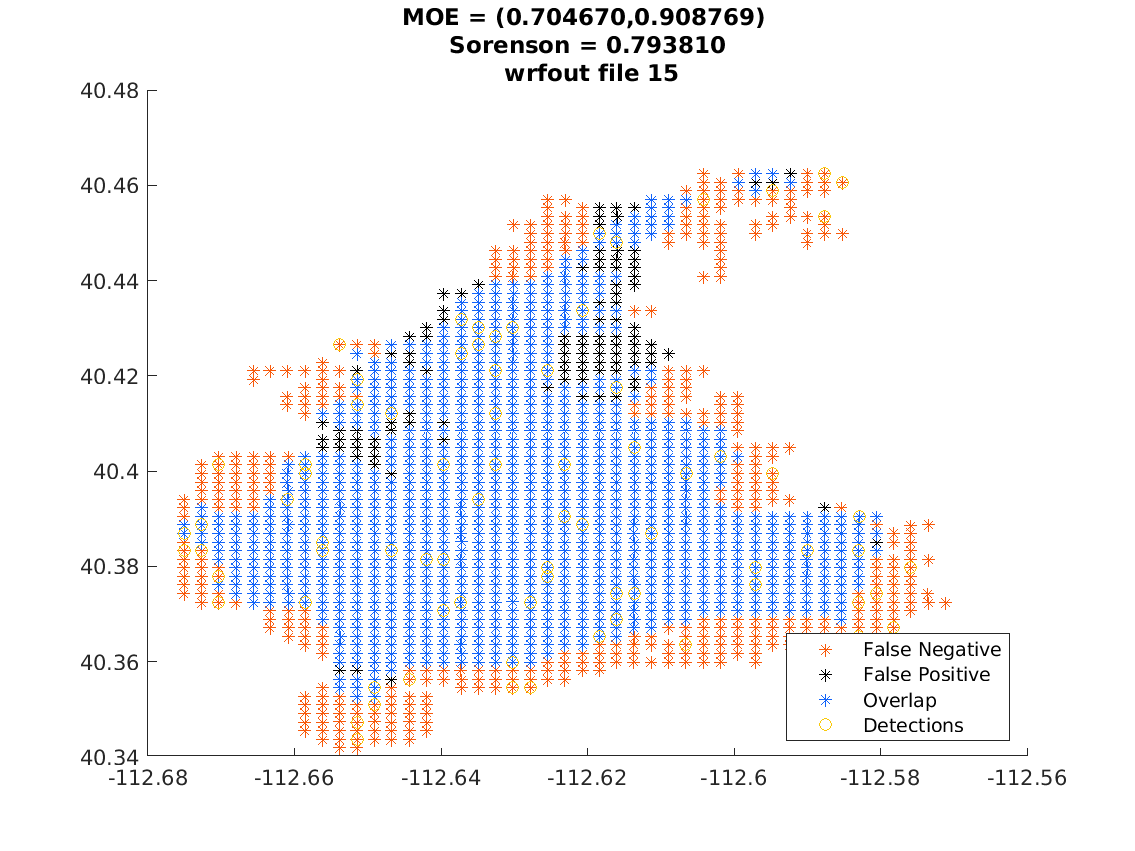}

\includegraphics[width=0.45\textwidth]{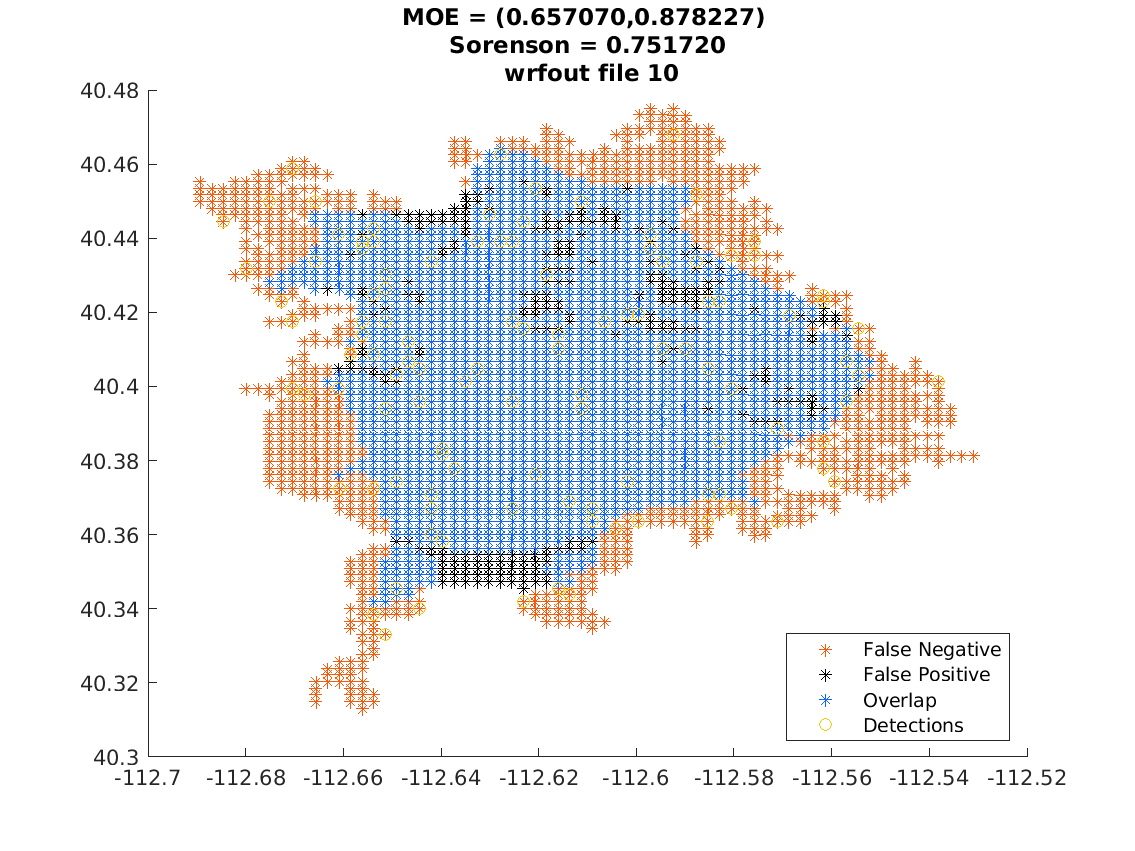}
\includegraphics[width=0.45\textwidth]{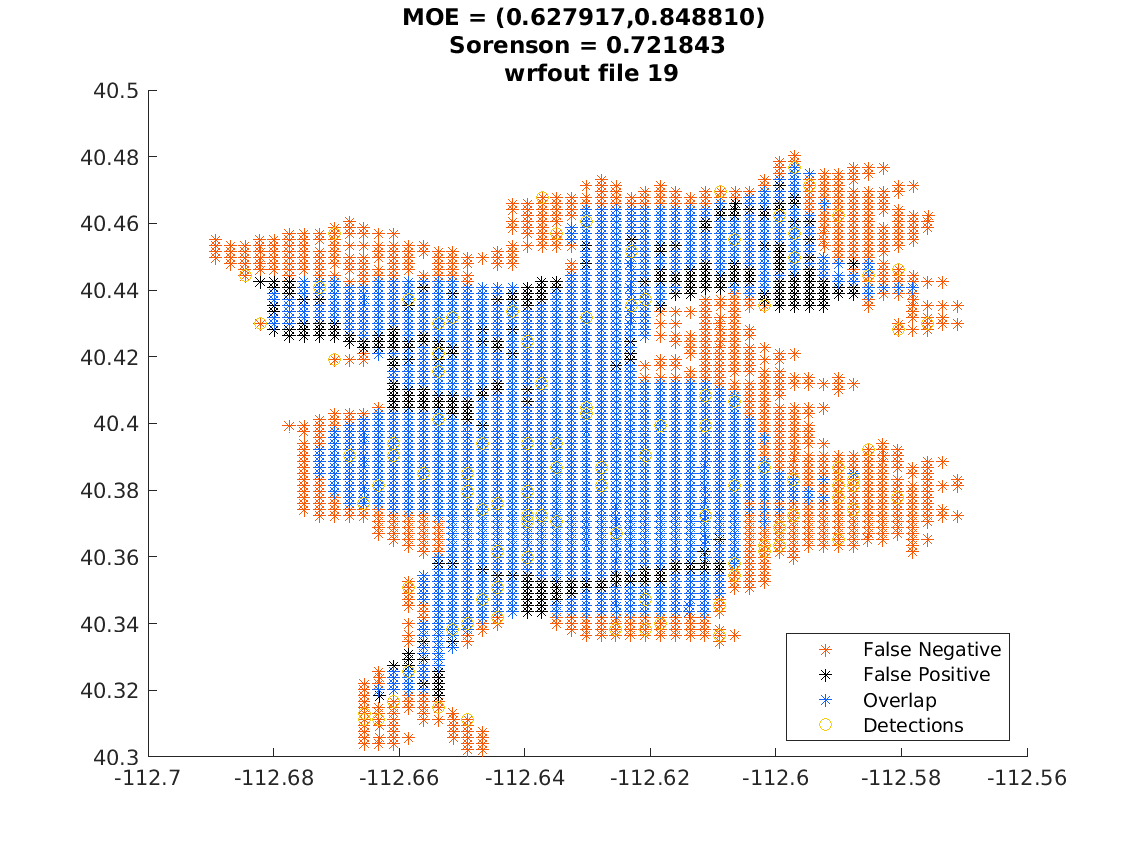}

\includegraphics[width=0.45\textwidth]{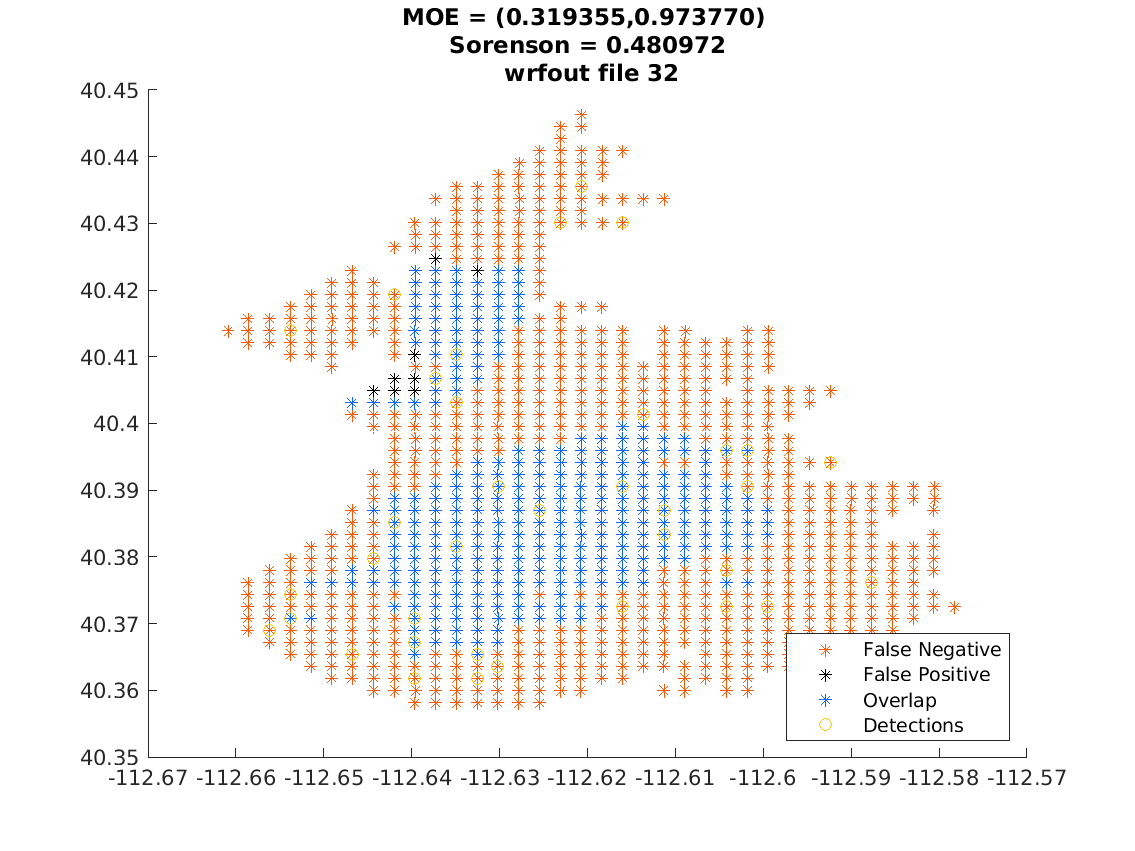}
\includegraphics[width=0.45\textwidth]{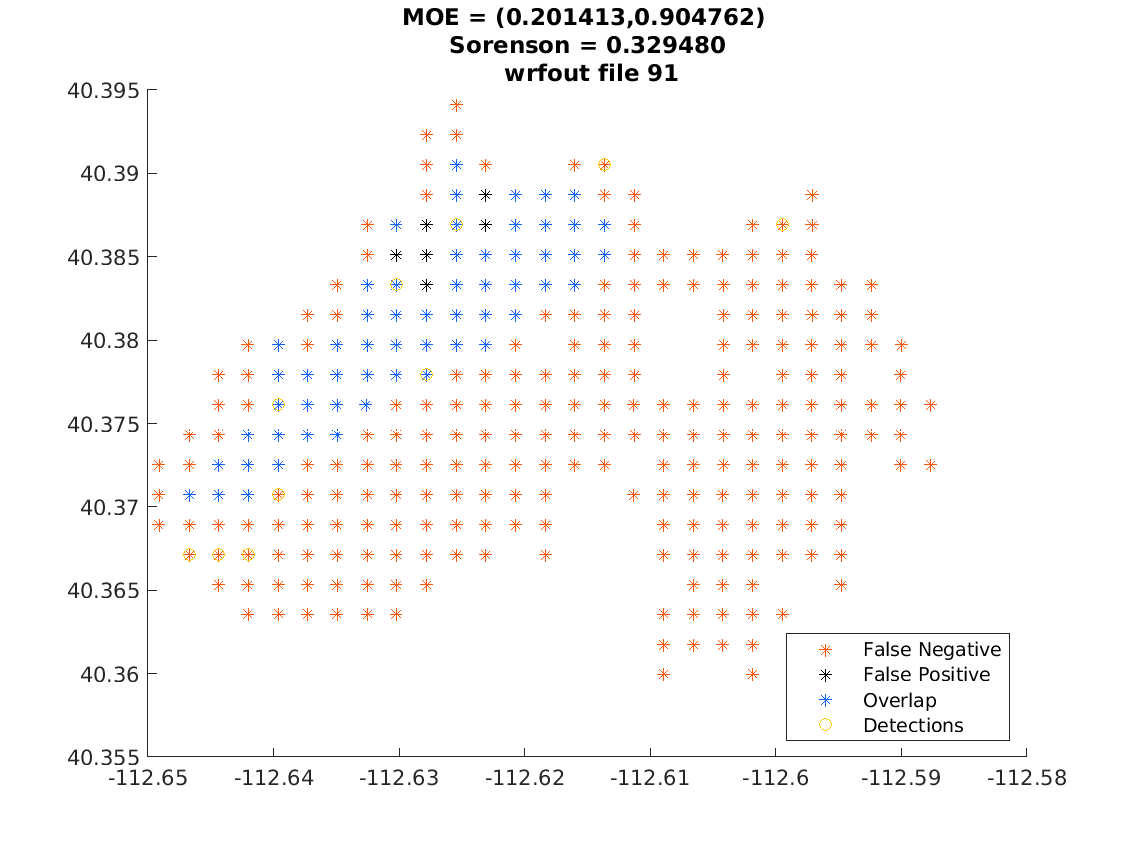}

 \caption[A gallery of some of the results from the test of estimating fire arrival time using output from WRF-SFIRE simulations as ``ground truth."]{A gallery of some of the results from the test of estimating fire arrival time using output from WRF-SFIRE simulations as ``ground truth." Assessment of plots in the top row show examples where the estimated fire arrival time produced good scores, the middle row show examples with score near the average, and the bottom row shows scores well below the average. The method consistently underestimated the the actual fire area. The worst scores were obtained for small fires with few satellite detections. }
 \label{fig:gallery_wrf_test}
\end{figure}

\begin{figure}[!ht]
\centering
\includegraphics[width = 0.48\textwidth]{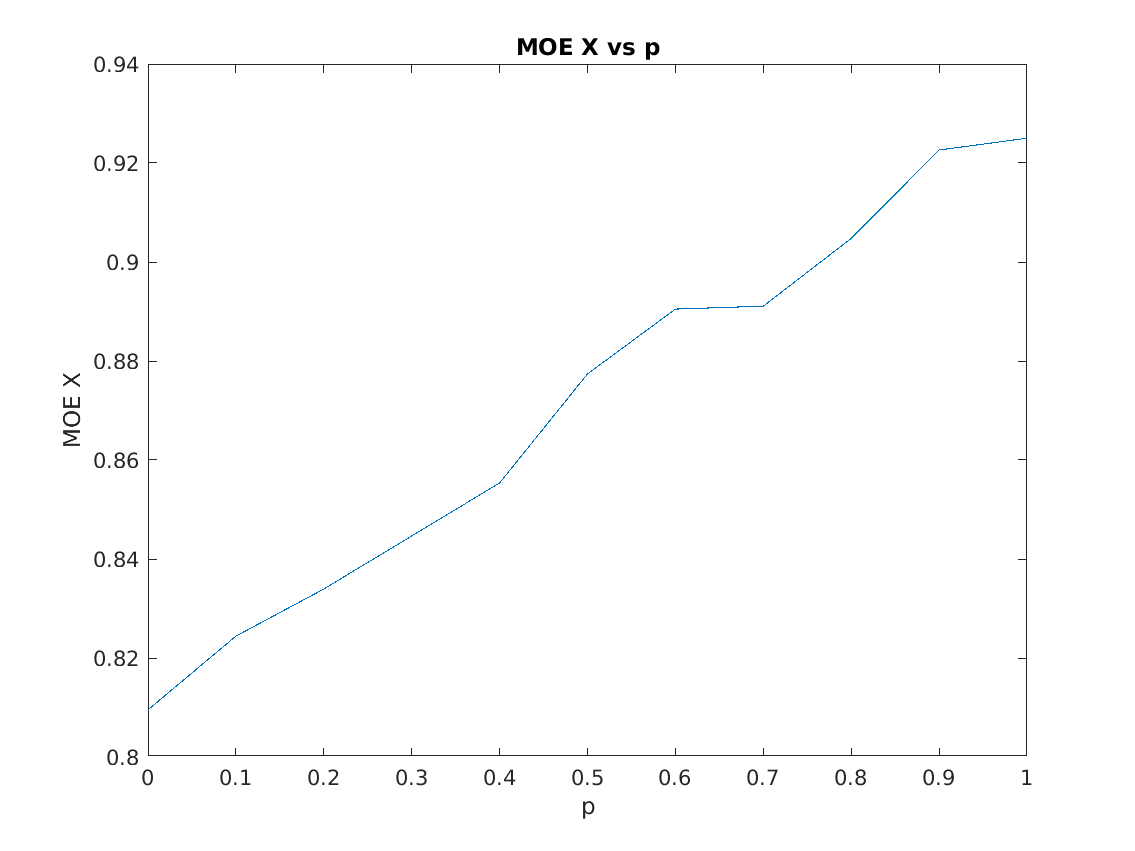}
\includegraphics[width = 0.48\textwidth]{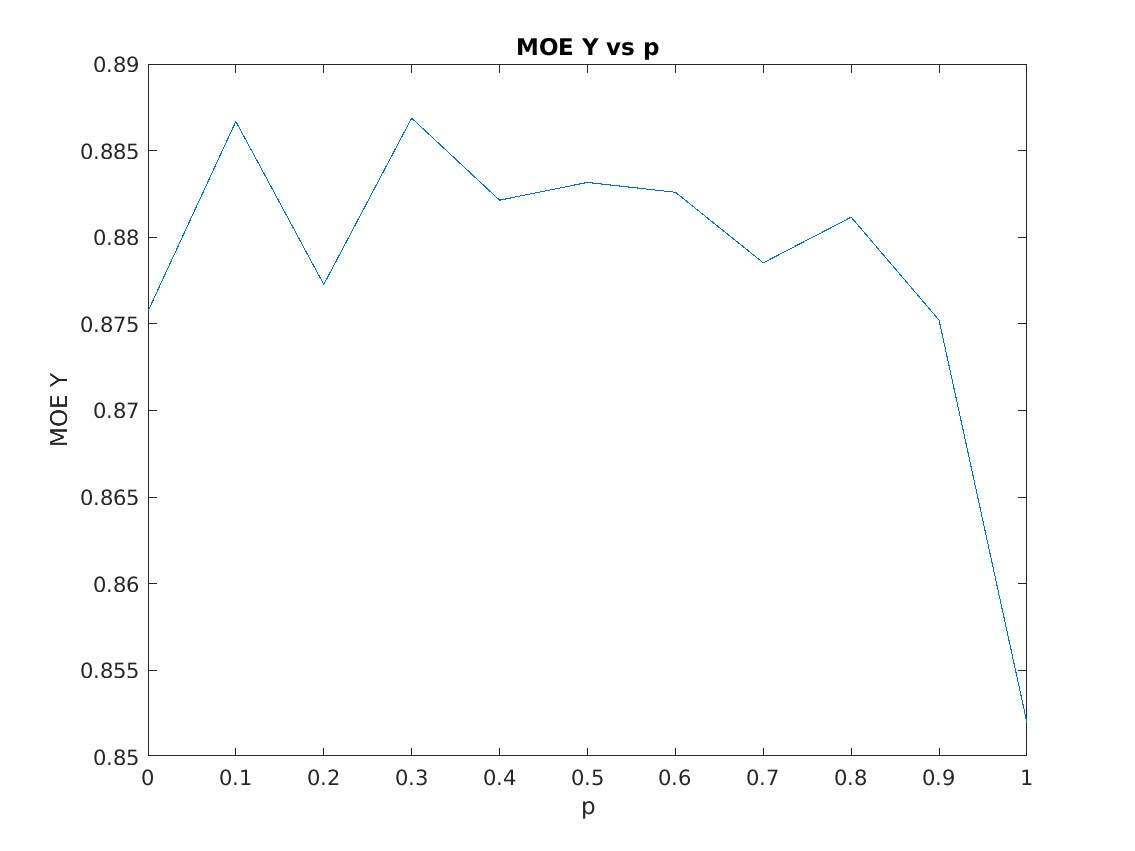}

\includegraphics[width = 0.48\textwidth]{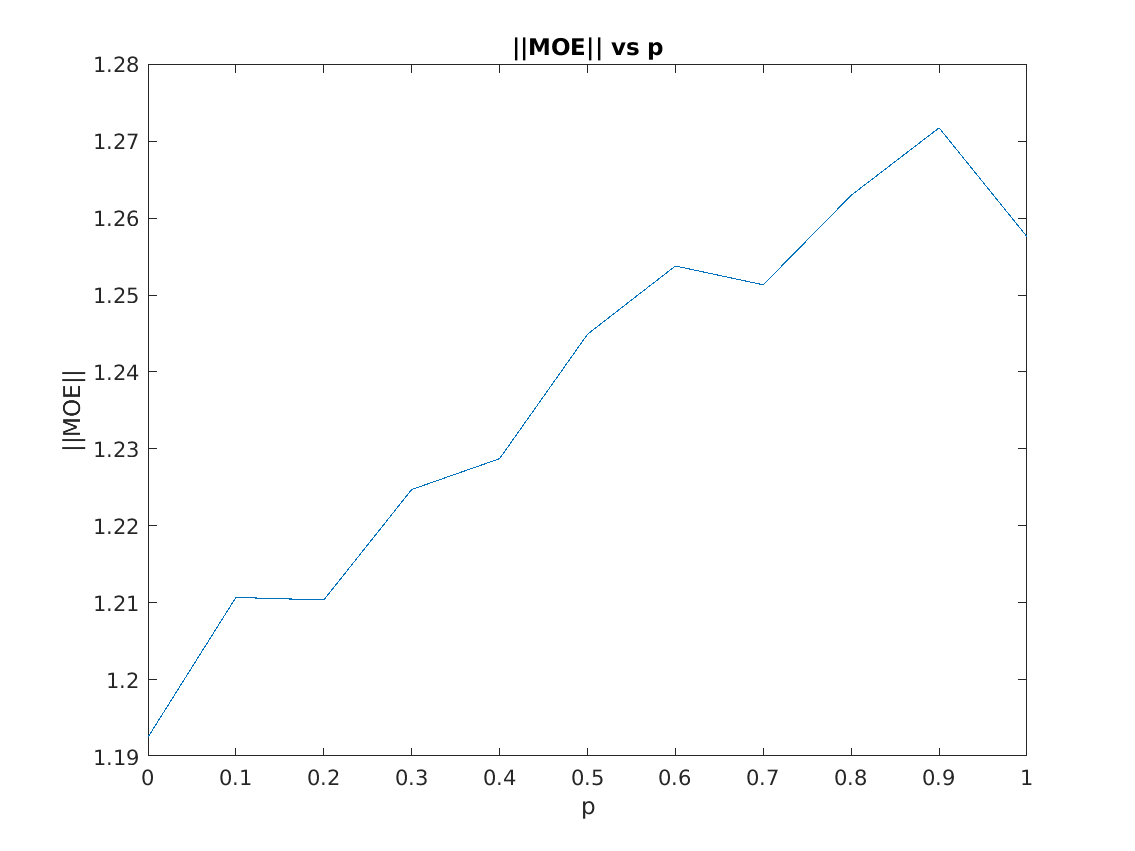}
\includegraphics[width = 0.48\textwidth]{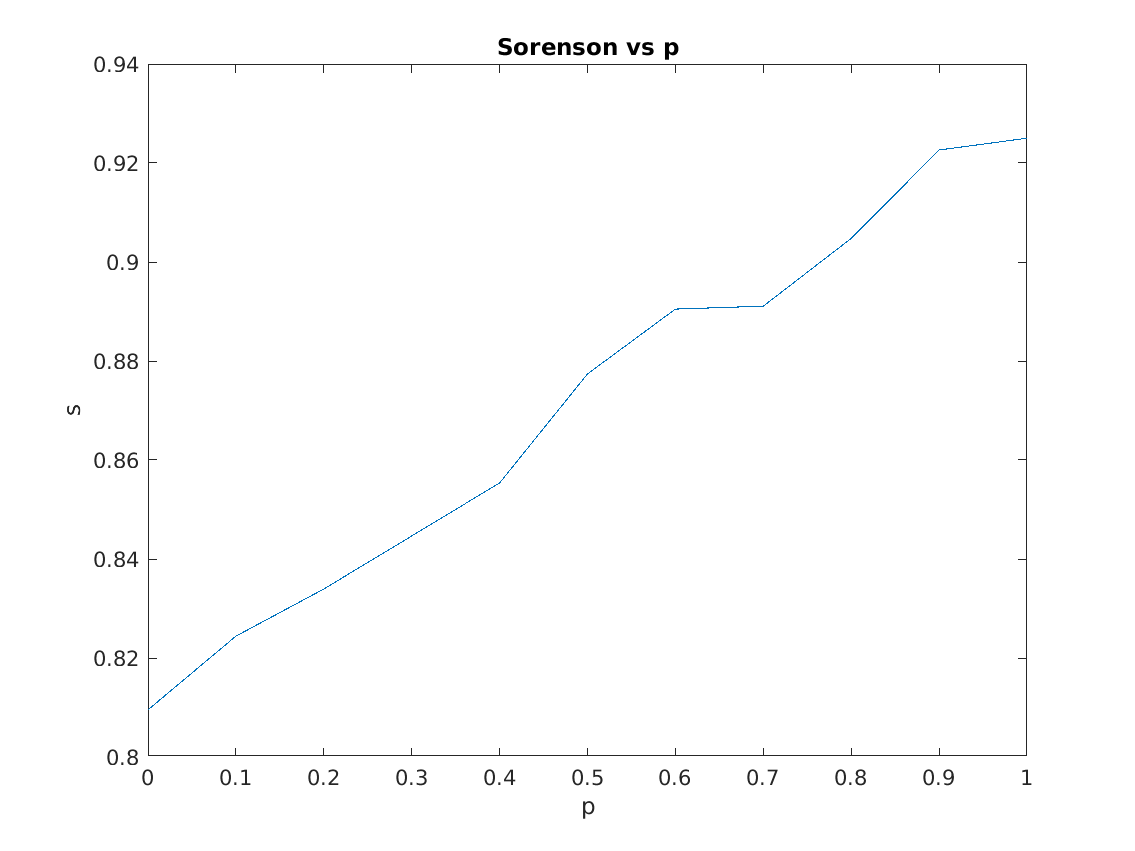}
     \caption[Assessment of estimated fire arrival times of the Patch Springs Fire using varied values of the interpolation parameter $p$.]{Assessment of estimated fire arrival times of the Patch Springs Fire using varied values of the interpolation parameter $p$. The figures summarize the results in Table \ref{tbl:patch_p_test}. On the top row, rom left to right are the plots of MOE X and MOE Y. On the bottom are the plots of  ||MOE||, and the S{\o}renson index. Scores generally increase with the value of $p$, but the drop in MOE Y when $p=1.0$ suggests that using $p=0.9$ is optimal.}
\label{fig:p_test_patch_figs}
\end{figure}

\begin{figure}[!ht]
\centering
\includegraphics[width = 0.48\textwidth]{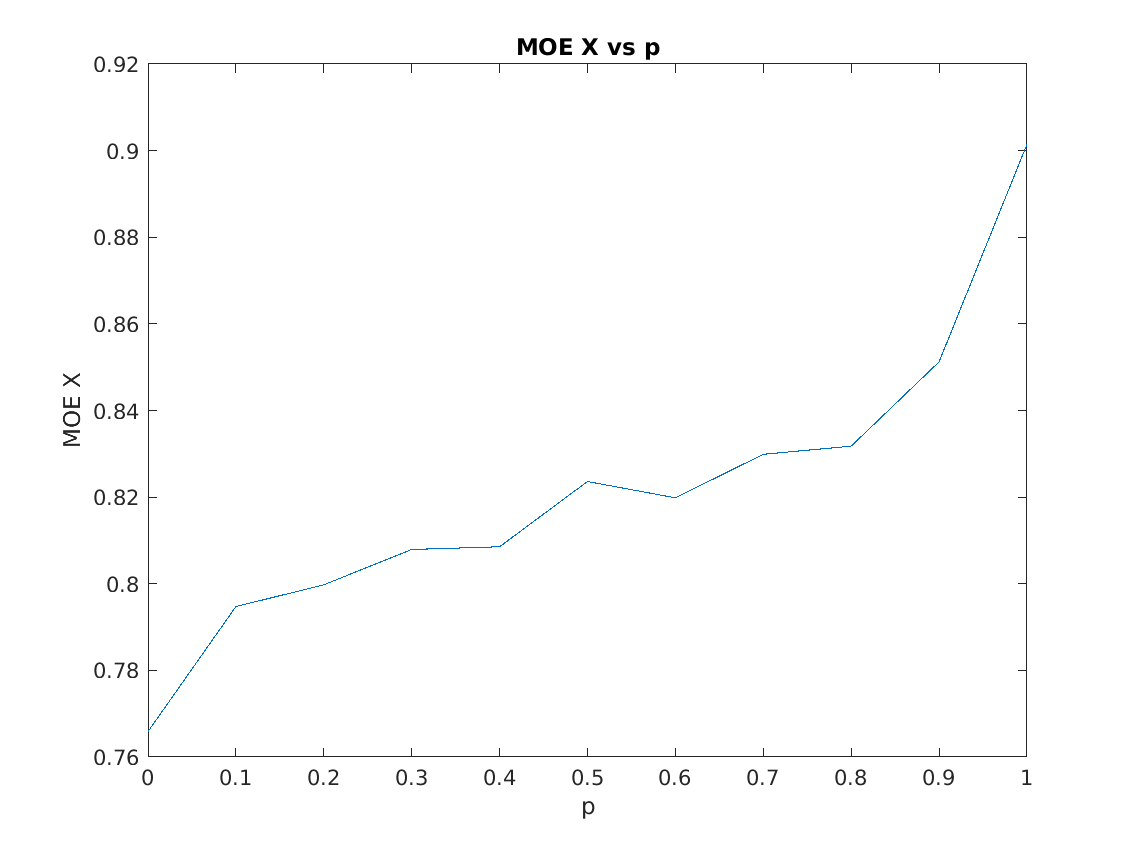}
\includegraphics[width = 0.48\textwidth]{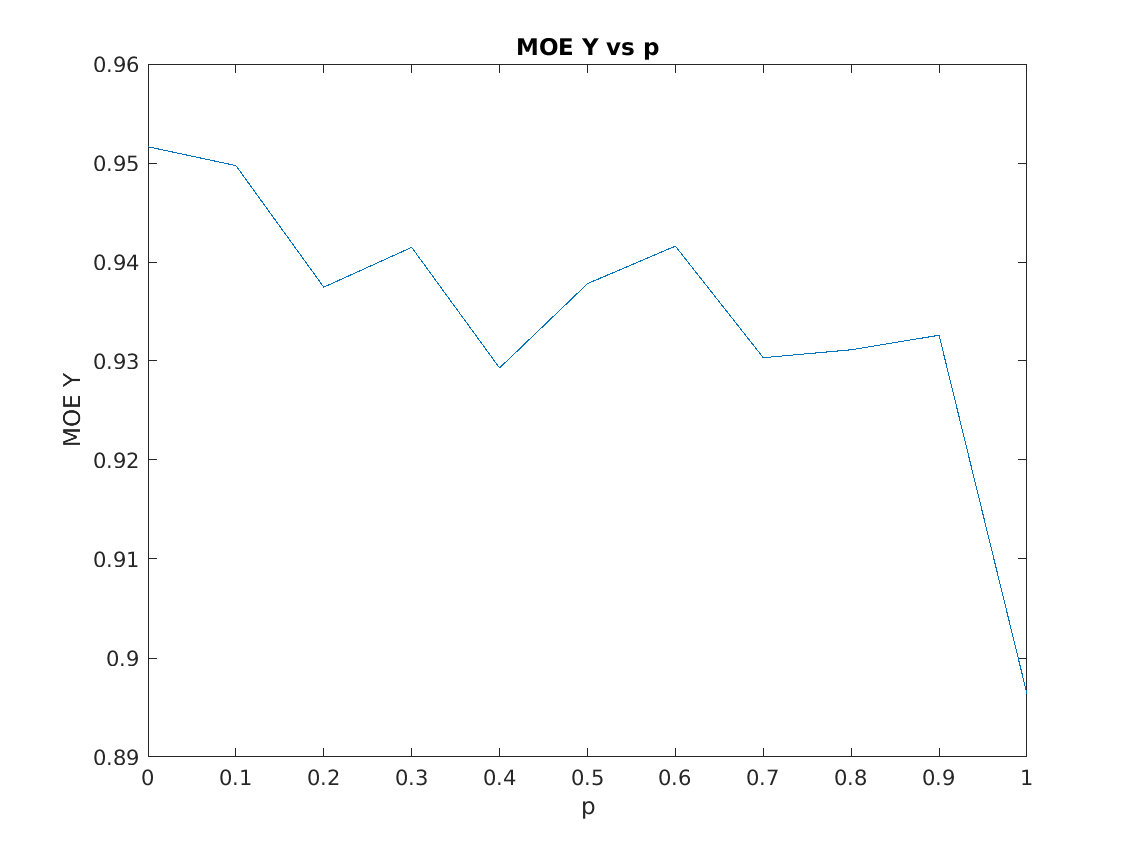}

\includegraphics[width = 0.48\textwidth]{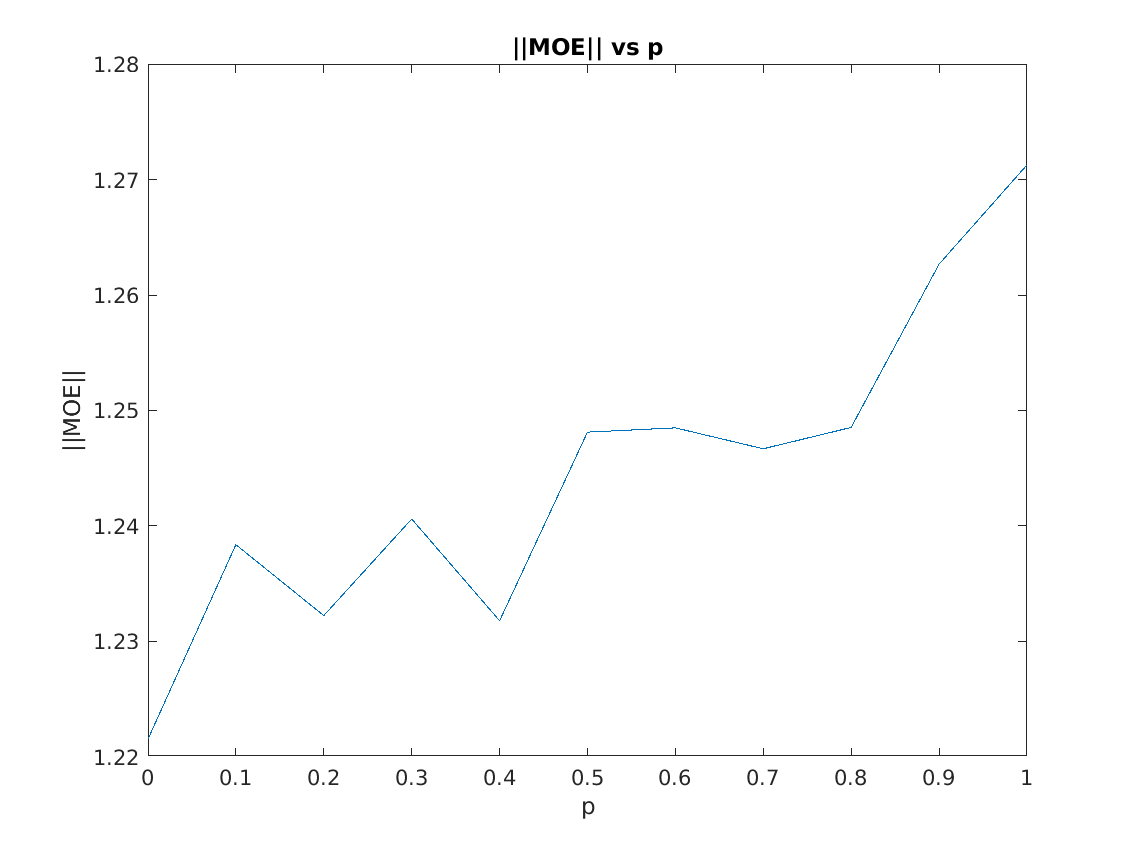}
\includegraphics[width = 0.48\textwidth]{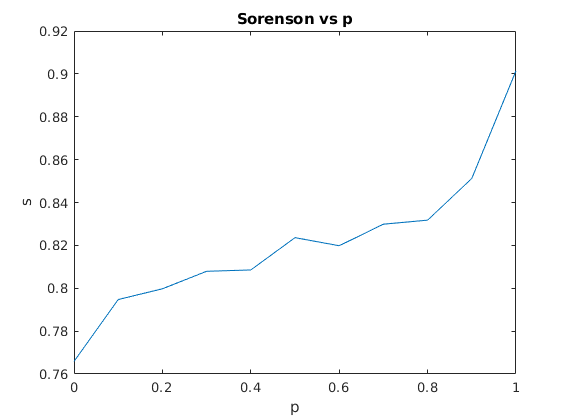}
\caption[Assessment of estimated fire arrival times of the Cougar Creek Fire using varied values of the interpolation parameter $p$.]{Assessment of estimated fire arrival times of the Cougar Creek Fire using varied values of the interpolation parameter $p$. The figures summarize the results in Table \ref{tbl:patch_p_test}. On the top row, rom left to right are the plots of MOE X and MOE Y. On the bottom are the plots of  ||MOE||, and the S{\o}renson index. Scores generally increase with the value of $p$, but the drop in MOE Y when $p=1.0$ suggests that using $p=0.9$ is optimal.}
\label{fig:p_test_cougar_figs}
\end{figure}

\begin{table}
\centering
\begin{tabular}{|c|c|c|c|c|} 
\hline
p   & MOE X  & MOE Y  & \textbar{}\textbar{}MOE\textbar{}\textbar{} & Sorenson  \\ 
\hline
No Interpolation  & 0.9095 & 0.8503 & 1.2451                                      & 0.8789    \\ 
\hline
0.0 & 0.8095 & 0.8757 & 1.1926                                      & 0.8413    \\ 
\hline
0.1 & 0.8244 & 0.8867 & 1.2107                                      & 0.8544    \\ 
\hline
0.2 & 0.8339 & 0.8773 & 1.2104                                      & 0.8551    \\ 
\hline
0.3 & 0.8446 & 0.8869 & 1.2247                                      & 0.8652    \\ 
\hline
0.4 & 0.8554 & 0.8821 & 1.2287                                      & 0.8685    \\ 
\hline
0.5 & 0.8774 & 0.8832 & 1.2449                                      & 0.8803    \\ 
\hline
0.6 & 0.8905 & 0.8826 & 1.2538                                      & 0.8865    \\ 
\hline
0.7 & 0.8911 & 0.8785 & 1.2513                                      & 0.8848    \\ 
\hline
0.8 & 0.9048 & 0.8812 & 1.2629                                      & 0.8928    \\ 
\hline
0.9 & 0.9226 & 0.8752 & 1.2717                                      & 0.8983    \\ 
\hline
1.0 & 0.9250 & 0.8520 & 1.2576                                      & 0.8870    \\
\hline
\end{tabular}
\caption[Results of varying parameter $p$ for estimates of the fire arrival time of the Patch Springs Fire. The first row shows the scores when no additional points are interpolated along the paths.]{Results of varying parameter $p$ for estimates of the fire arrival time of the Patch Springs Fire. The first row shows the scores when no additional points are interpolated along the paths. The original path structure contained 305 active fire detections. Interpolation with a spacing of 2000 meters added an additional 62 points to the path structures. The scores were made by comparison to an infrared perimeter observation from August 16, 09:47 UTC. The results are summarized in Figure \ref{fig:p_test_patch_figs}.}
\label{tbl:patch_p_test}
\end{table}

\begin{table}
\centering
\begin{tabular}{|c|c|c|c|c|} 
\hline
p   & MOE X  & MOE Y  & \textbar{}\textbar{}MOE\textbar{}\textbar{} & Sorenson  \\ 
\hline
No Interpolation & 0.8820 & 0.7893 & 1.1836                                      & 0.8331    \\ 
\hline
0.0 & 0.7659 & 0.9516 & 1.2215                                      & 0.8487    \\ 
\hline
0.1 & 0.7947 & 0.9497 & 1.2384                                      & 0.8653    \\ 
\hline
0.2 & 0.7997 & 0.9375 & 1.2322                                      & 0.8631    \\ 
\hline
0.3 & 0.8079 & 0.9415 & 1.2406                                      & 0.8696    \\ 
\hline
0.4 & 0.8085 & 0.9293 & 1.2318                                      & 0.8647    \\ 
\hline
0.5 & 0.8236 & 0.9378 & 1.2481                                      & 0.8770    \\ 
\hline
0.6 & 0.8198 & 0.9416 & 1.2485                                      & 0.8765    \\ 
\hline
0.7 & 0.8299 & 0.9303 & 1.2467                                      & 0.8772    \\ 
\hline
0.8 & 0.8318 & 0.9311 & 1.2485                                      & 0.8786    \\ 
\hline
0.9 & 0.8512 & 0.9326 & 1.2627                                      & 0.8901    \\ 
\hline
1.0 & 0.9014 & 0.8964 & 1.2713                                      & 0.8989    \\
\hline
\end{tabular}
\caption[Results of varying parameter $p$ for estimates of the fire arrival time of the Cougar Creek Fire. The first row shows the scores when no additional points are interpolated along the paths.]{Results of varying parameter $p$ for estimates of the fire arrival time of the Cougar Creek Fire. The first row shows the scores when no additional points are interpolated along the paths. The original path structure contained 432 active fire detections. Interpolation with a spacing of 2000 meters added an additional 24 points to the path structures. The scores were made by comparison to an infrared perimeter observation from August 15, 10:33 UTC.}
\label{tbl:cougar_p_test}
\end{table}

\begin{figure}[!h]
\begin{center}
\includegraphics[width = 0.45\textwidth]{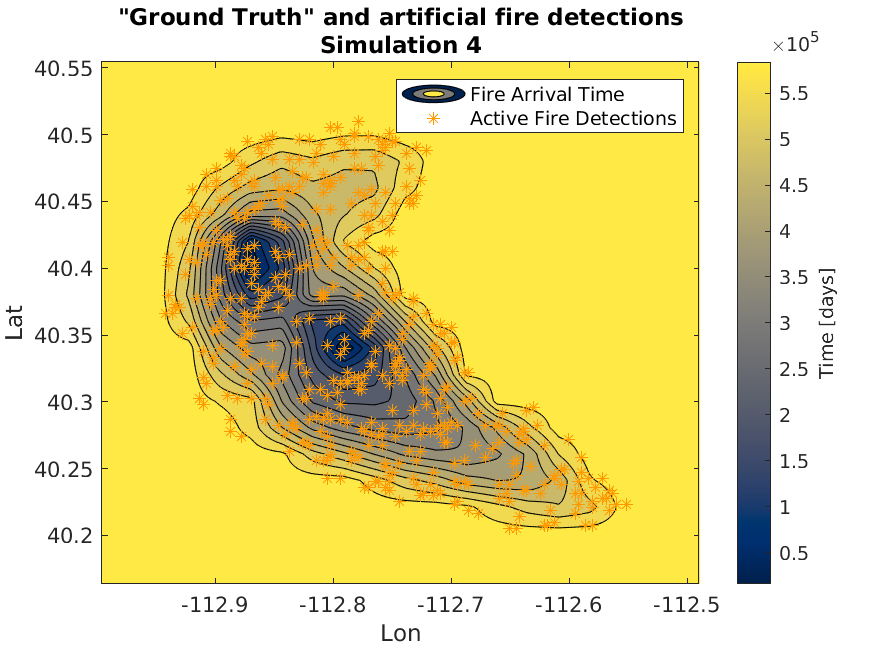}
\includegraphics[width = 0.45\textwidth]{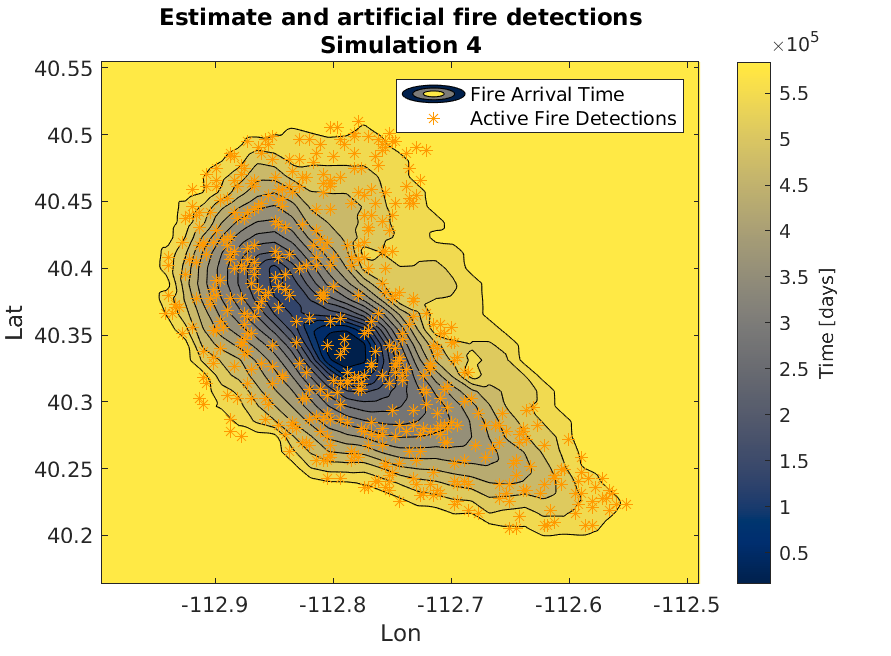}
\includegraphics[width = 0.45\textwidth]{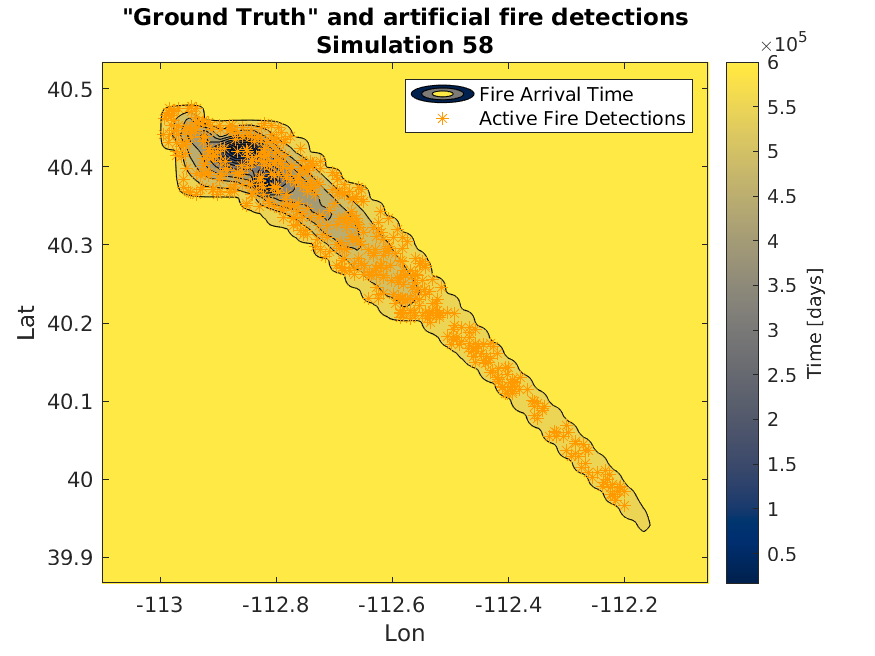}
\includegraphics[width = 0.45\textwidth]{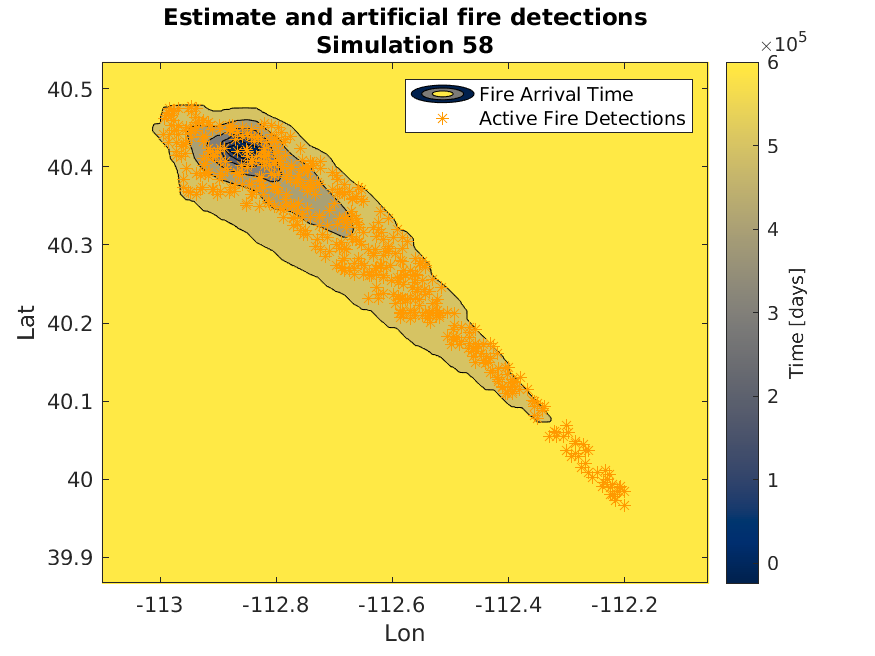}
\includegraphics[width = 0.45\textwidth]{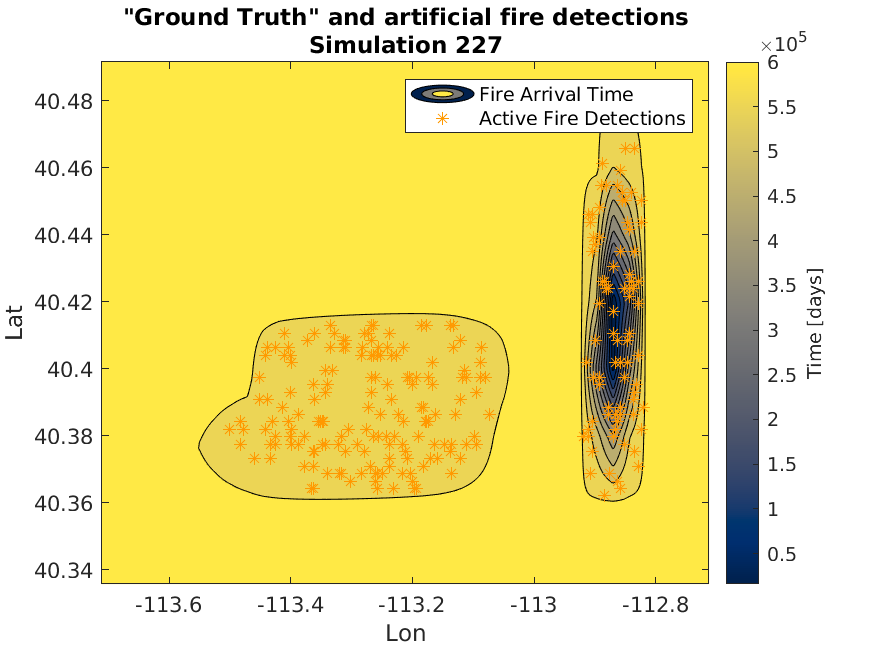}
\includegraphics[width = 0.45\textwidth]{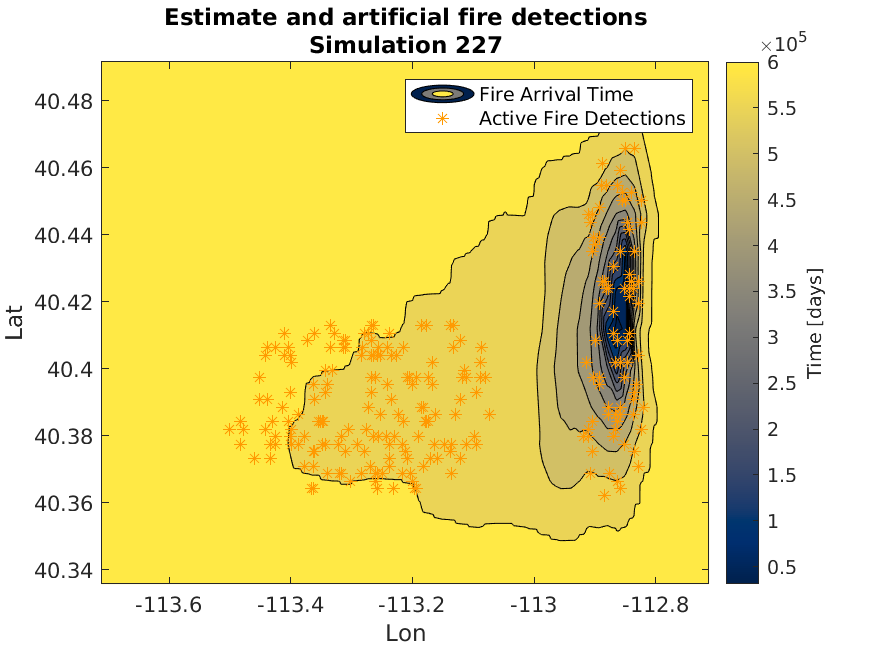}
  \caption[A sampling of some of the estimated fire arrival times generated in the experiment estimating the fire arrival time using artificial data. ]{A sampling of some of the estimated fire arrival times generated in the experiment estimating the fire arrival time using artificial data. on each row, the plots show comparisons between the ``ground truth" and estimated fire arrival times. All plots of the estimates were derived using the multigrid method with interpolation of additional points on the paths with 2000 meter spacing. The example on the bottom, simulation 227, had the worst ROS error for all estimates made using the multigrid approach, regardless of spacing of interpolated points on the paths.}
\label{fig:new_test_2000_contours}
\end{center}
\end{figure}

\begin{figure}[!ht]
\centering
\includegraphics[width = 0.48\textwidth]{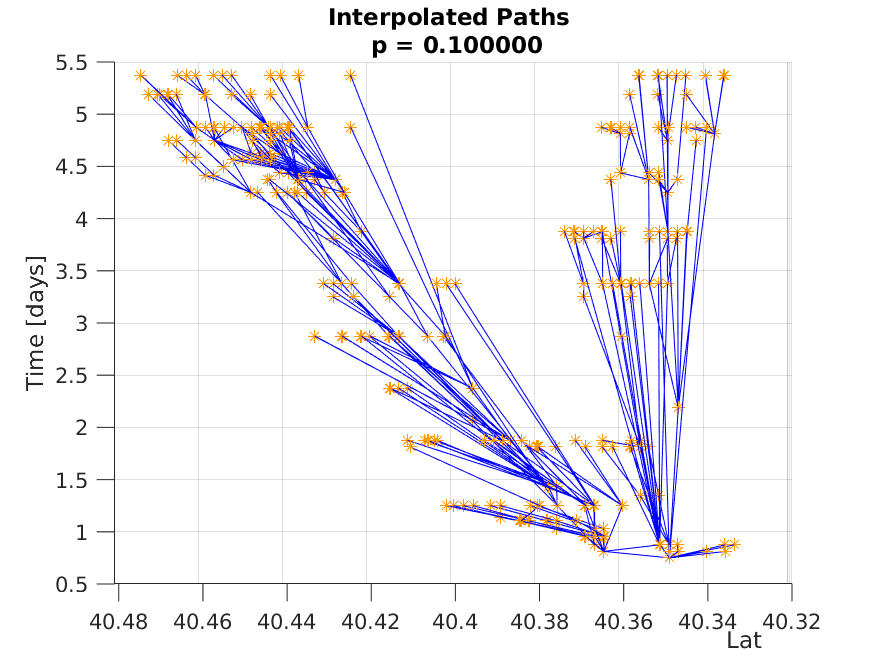}
\includegraphics[width = 0.48\textwidth]{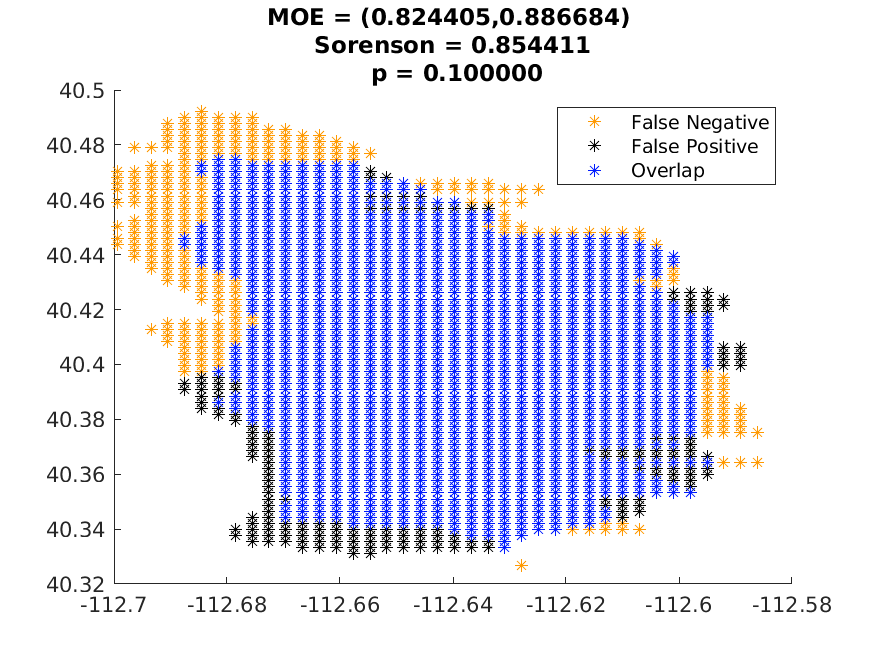}
\includegraphics[width = 0.48\textwidth]{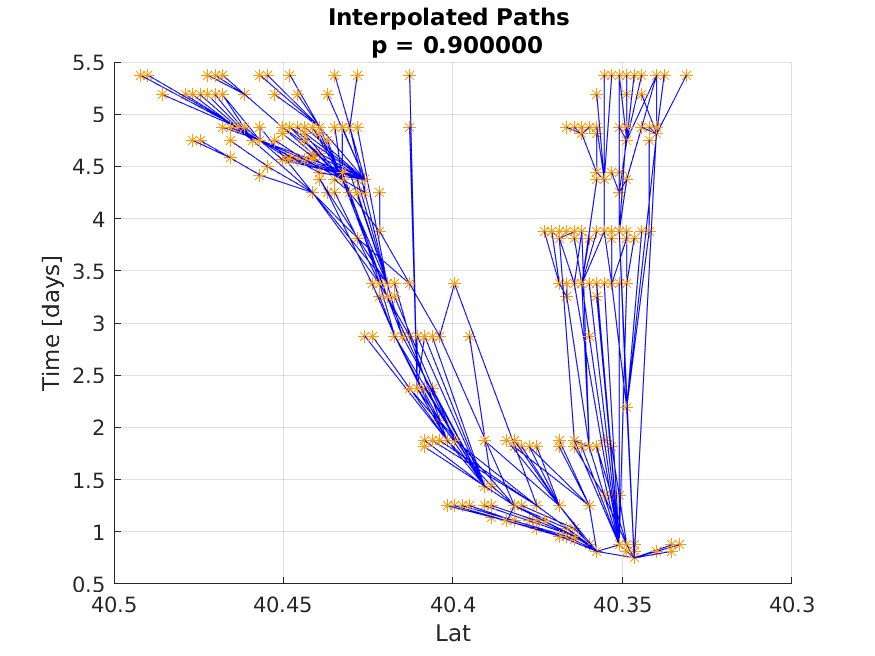}
\includegraphics[width = 0.48\textwidth]{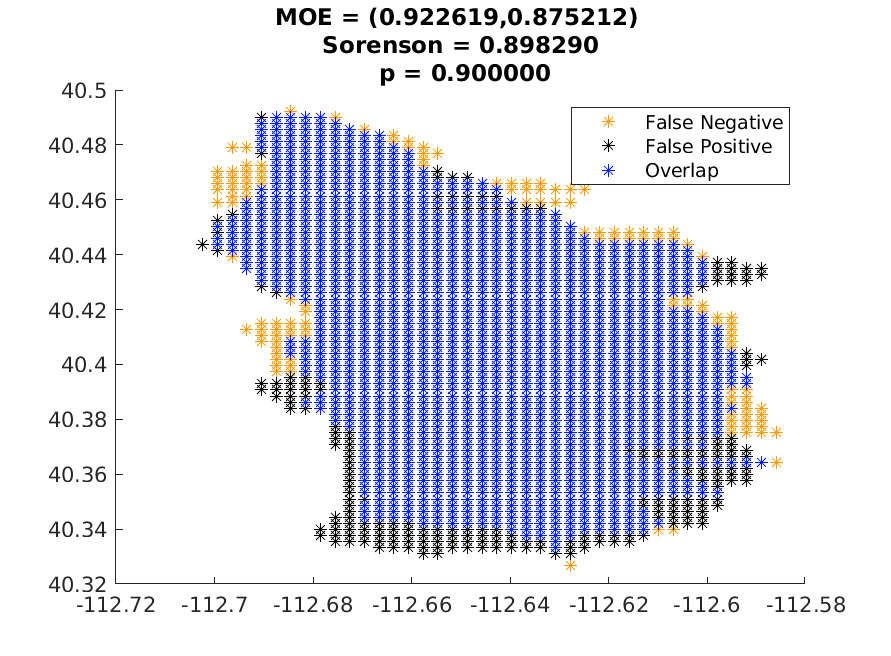}
\caption[Path structures using varying values of the interpolation parameter $p$ and assessment of the fire arrival times constructed from them. Each row shows the path structure and the assessment for a particular value of $p$.]{Path structures using varying values of the interpolation parameter $p$ and assessment of the fire arrival times constructed from them. Each row shows the path structure and the assessment for a particular value of $p$. The top row has $p=0.1$ of extra points and the bottom row has $p=0.9$.}
\label{fig:p_test_patch_paths_moe}
\end{figure}

%

\begin{figure}[!ht]
\centering
\includegraphics[width = 0.48\textwidth]{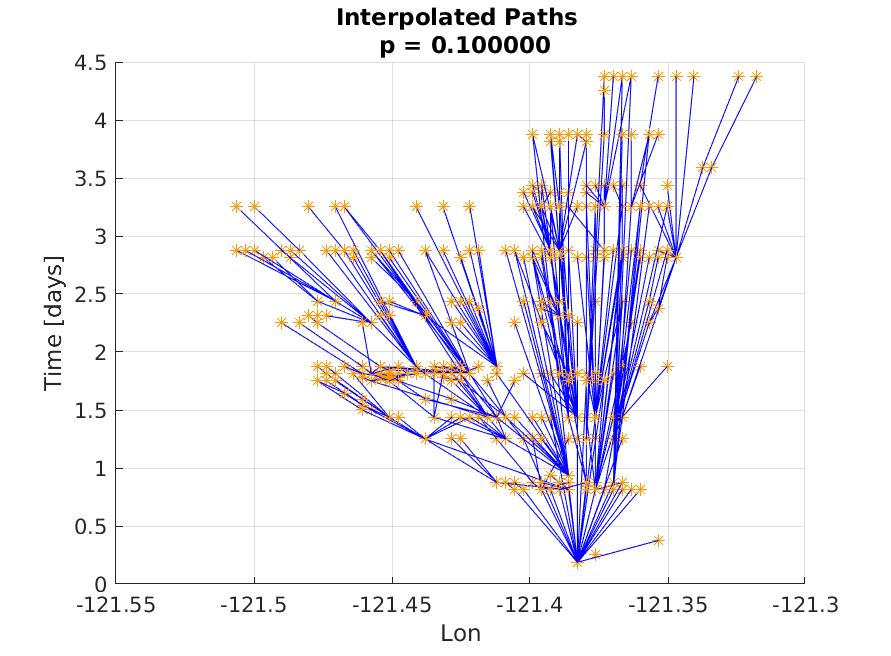}
\includegraphics[width = 0.48\textwidth]{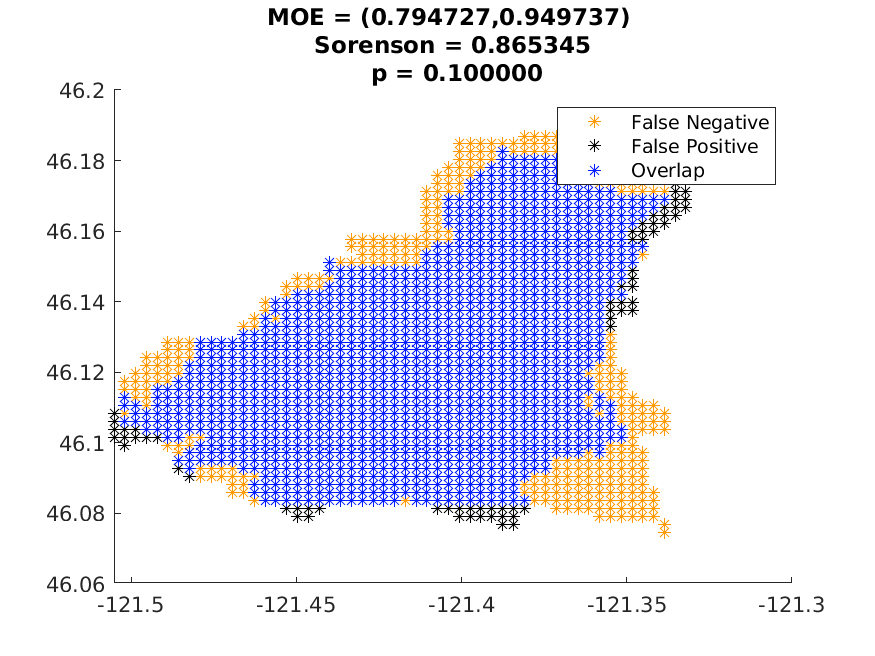}
\includegraphics[width = 0.48\textwidth]{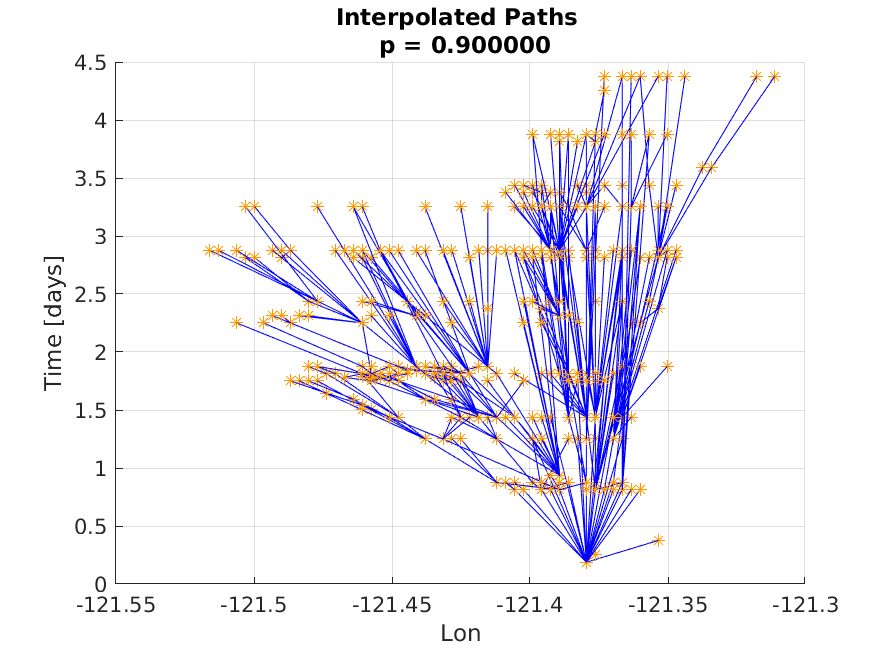}
\includegraphics[width = 0.48\textwidth]{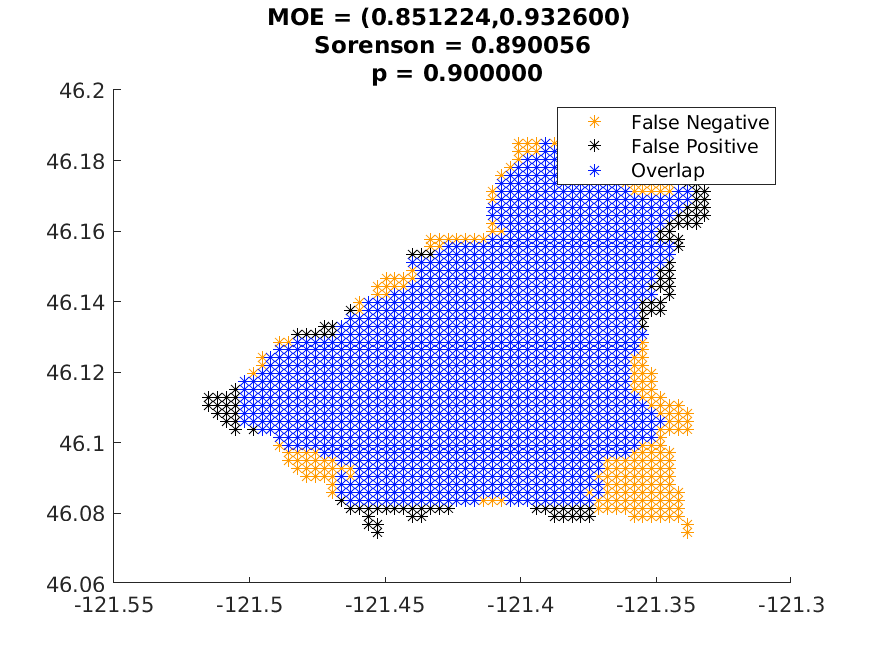}
\caption[Path structures using varying values of the interpolation parameter $p$ and assessment of the fire arrival times constructed from them. Each row shows the path structure and the assessment for a particular value of $p$.]{Path structures using varying values of the interpolation parameter $p$ and assessment of the fire arrival times constructed from them. Each row shows the path structure and the assessment for a particular value of $p$. The top row has $p=0.1$ of extra points and the bottom row has $p=0.9$.}
\label{fig:p_test_cougar_paths_moe}
\end{figure}

\begin{figure}[htbp]
\centering
\includegraphics[width=0.45\textwidth]{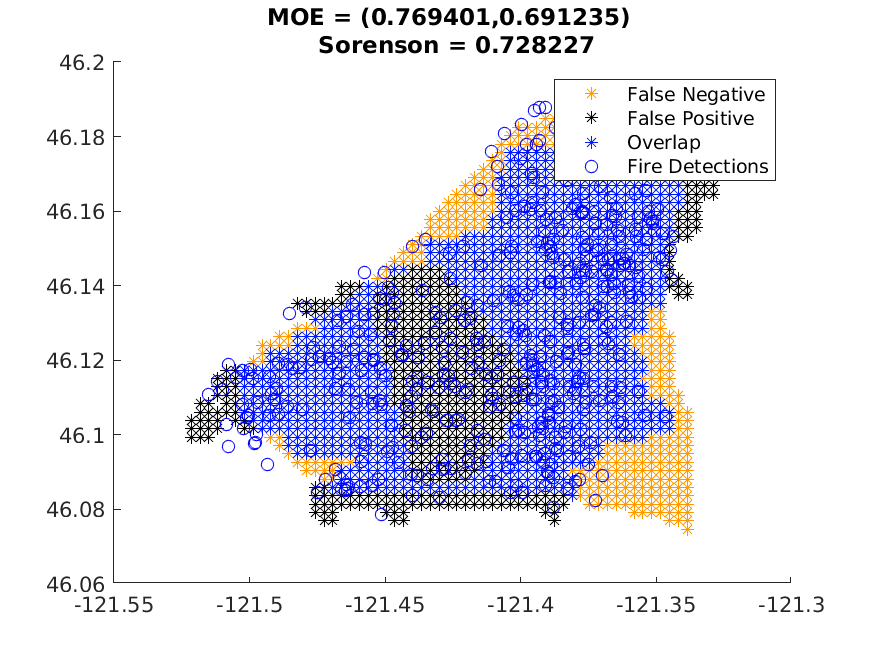}
\includegraphics[width=0.45\textwidth]{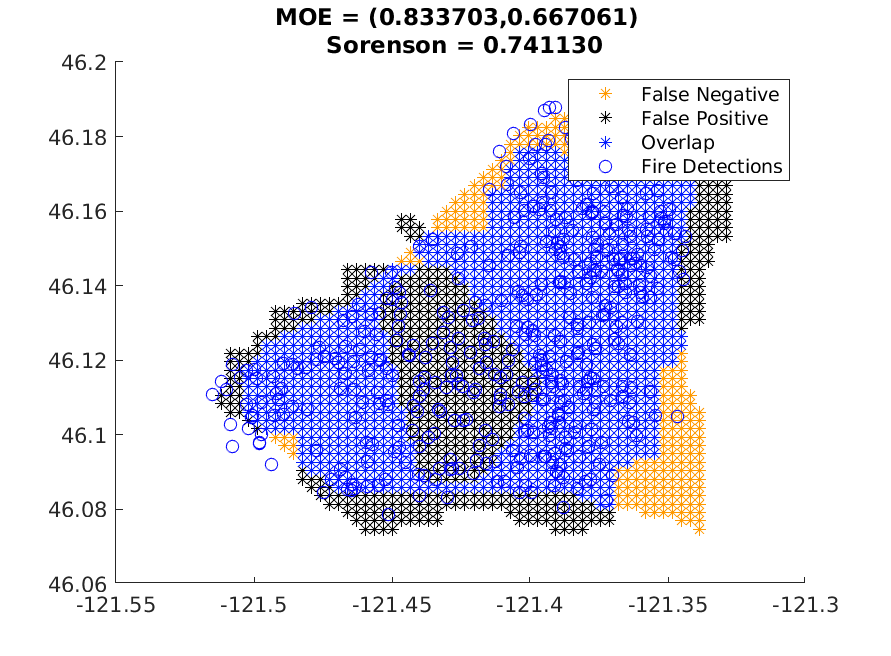}
 \caption[Assessing the estimated fire arrival time of the Cougar Creek Fire using a one-pass method (left) and multigrid method (right) by comparison with an infrared perimeter observation.]{Assessing the estimated fire arrival time of the Cougar Creek Fire using a one-pass method (left) and multigrid method (right) by comparison with an infrared perimeter observation. The estimates are derived for the available satellite data up to the perimeter time of August 15, 10:33:00 UTC. Both methods tend to underestimate the size of the fire. The routine for plotting the area within the infrared perimeter has failed, giving a larger black region within the fire that results in  lower scores than should be assigned to these estimations of the fire arrival time. }
 \label{fig:cougar_moe_perim_5}
\end{figure}

\begin{figure}[htbp]
\centering
\includegraphics[width=0.45\textwidth]{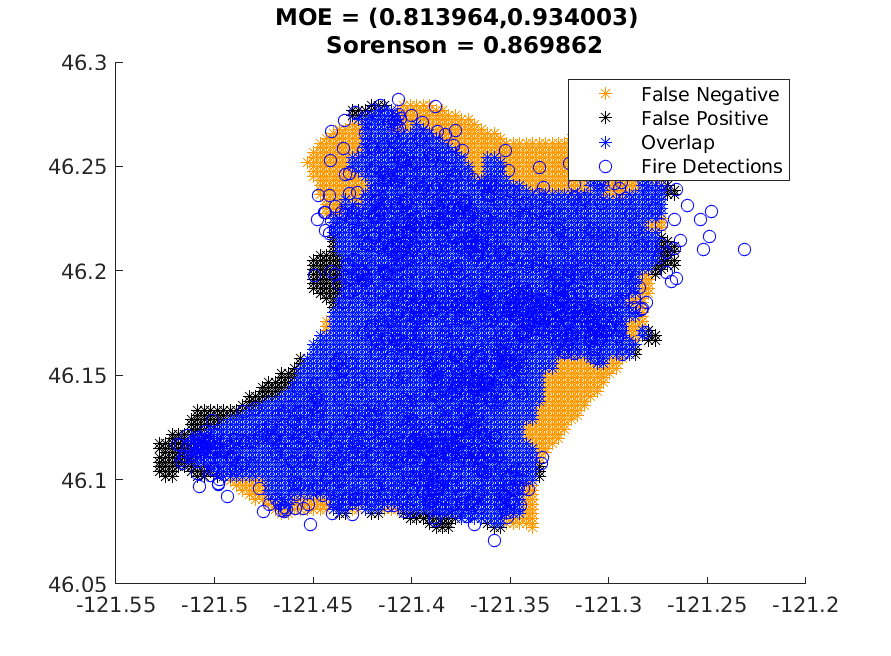}
\includegraphics[width=0.45\textwidth]{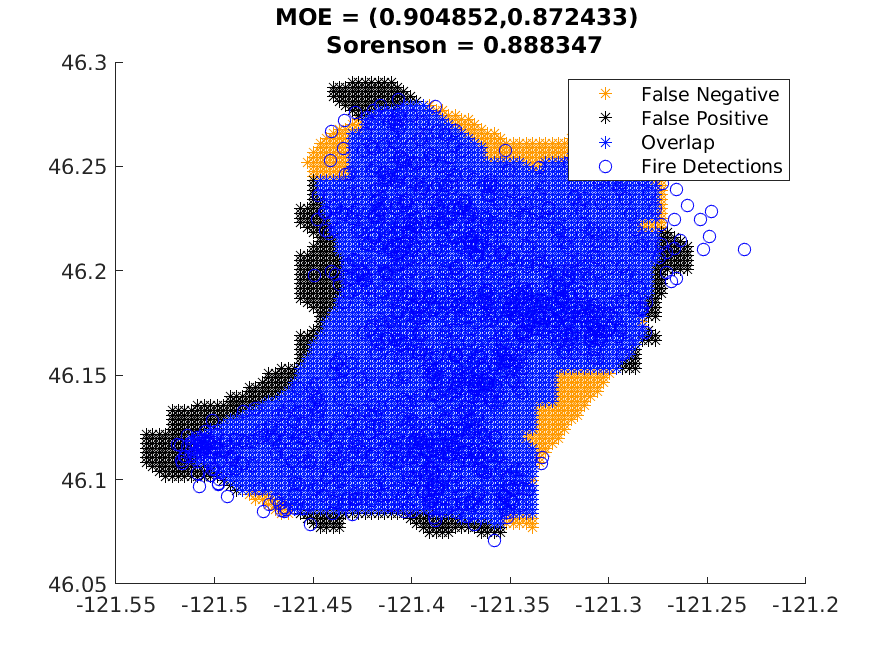}
 \caption[Assessing the estimated fire arrival time of the Cougar Creek Fire using a one-pass method (left) and multigrid method (right) by comparison with an infrared perimeter observation.]{Assessing the estimated fire arrival time of the Cougar Creek Fire using a one-pass method (left) and multigrid method (right) by comparison with an infrared perimeter observation. The estimates are derived for the available satellite data up to the perimeter time of September 4, 02:38:00 UTC. Both methods tend to underestimate the size of the fire. This is the final perimeter observation available. Both methods have similar results, but the multigrid method gives less false negatives and more false positives.}
 \label{fig:cougar_moe_perim_27}
\end{figure}

\begin{table}
\centering
\begin{tabular}{|c|c|c|c|c|c|c|c|c|} 
\hline
\# & Perimeter Time       & MOE\_X & MOE\_Y & \textbar{}MOE\textbar{} & S      \\ 
\hline
1  & 12-Aug 07:00:00 & 0.4855 & 0.8138 & 0.9476                  & 0.6082  \\ 
\hline
2  & 14-Aug 05:56:00 & 0.7794 & 0.9759 & 1.2490                  & 0.8667       \\ 
\hline
3  & 14-Aug 07:00:00 & 0.8493 & 0.9151 & 1.2485                  & 0.8810  \\ 
\hline
4  & 15-Aug 07:00:00 & 0.7903 & 0.9588 & 1.2425                  & 0.8664       \\ 
\hline
5  & 15-Aug 10:33:00 & 0.7694 & 0.6912 & 1.0343                  & 0.7282       \\ 
\hline
6  & 16-Aug 03:42:00 & 0.8601 & 0.8985 & 1.2438                  & 0.8789       \\ 
\hline
7  & 16-Aug 07:00:00 & 0.8287 & 0.9239 & 1.2411                  & 0.8737       \\ 
\hline
8  & 16-Aug 19:00:00 & 0.8668 & 0.9209 & 1.2647                  & 0.8930       \\ 
\hline
9  & 17-Aug 04:13:00 & 0.8369 & 0.7308 & 1.1111                  & 0.7803       \\ 
\hline
10 & 18-Aug 17:00:00 & 0.8305 & 0.9168 & 1.2370                  & 0.8715       \\ 
\hline
11 & 19-Aug 09:55:00 & 0.9545 & 0.8844 & 1.3012                  & 0.9181       \\ 
\hline
12 & 20-Aug 04:48:00 & 0.9564 & 0.8817 & 1.3008                  & 0.9175       \\ 
\hline
13 & 20-Aug 17:00:00 & 0.9447 & 0.8794 & 1.2907                  & 0.9109       \\ 
\hline
14 & 22-Aug 02:00:00 & 0.9118 & 0.8678 & 1.2588                  & 0.8893       \\ 
\hline
15 & 22-Aug 18:30:00 & 0.8376 & 0.9473 & 1.2645                  & 0.8891       \\ 
\hline
16 & 23-Aug 08:16:00 & 0.9226 & 0.8718 & 1.2693                  & 0.8965       \\ 
\hline
17 & 24-Aug 04:42:00 & 0.7913 & 0.9192 & 1.2128                  & 0.8504       \\ 
\hline
18 & 25-Aug 07:47:00 & 0.7913 & 0.9522 & 1.2381                  & 0.8643       \\ 
\hline
19 & 26-Aug 03:58:00 & 0.7693 & 0.9488 & 1.2215                  & 0.8497       \\ 
\hline
20 & 27-Aug 04:00:00 & 0.7846 & 0.9520 & 1.2336                  & 0.8602       \\ 
\hline
21 & 28-Aug 09:25:00 & 0.8704 & 0.9379 & 1.2796                  & 0.9029       \\ 
\hline
22 & 28-Aug 10:00:00 & 0.8207 & 0.9369 & 1.2455                  & 0.8749       \\ 
\hline
23 & 29-Aug 19:00:00 & 0.8309 & 0.9305 & 1.2475                  & 0.8779       \\ 
\hline
24 & 31-Aug 10:25:00 & 0.7844 & 0.9506 & 1.2325                  & 0.8595       \\ 
\hline
25 & 02-Sep 08:00:00 & 0.8151 & 0.9493 & 1.2512                  & 0.8771  \\ 
\hline
26 & 03-Sep 09:53:00 & 0.8250 & 0.9487 & 1.2573                  & 0.8825       \\ 
\hline
27 & 04-Sep 02:38:00 & 0.8140 & 0.9340 & 1.2389                  & 0.8699       \\
\hline
\end{tabular}
\caption[Assessment of estimated fire arrival time for the Cougar Creek Fire. Estimates were made  using  one pass of the adjustment and smoothing method on a single computational grid.]{Assessment of estimated fire arrival time for the Cougar Creek Fire. Estimates were made without using  the multigrid strategy.  Each estimate was made using all the available satellite data up to the the time of the infrared perimeter observation that is used to make the comparison. As can be seen by examining the area columns in the table, the fire experienced little to no growth after August 28. Figure \ref{fig:cougar_test_growth} shows how the area of the fire changed with time.}
\label{tbl:cougar_1_pass_results}
\end{table}

\begin{table}
\centering
\begin{tabular}{|c|c|c|c|c|c|c|c|c|} 
\hline
\# & Perimeter Time       & MOE\_X & MOE\_Y & \textbar{}MOE\textbar{} & S      \\ 
\hline
1  & 12-Aug 07:00:00 & 0.5821 & 0.7902 & 0.9814                  & 0.6704   \\ 
\hline
2  & 14-Aug 05:56:00 & 0.9045 & 0.9126 & 1.2849                  & 0.9086  \\ 
\hline
3  & 14-Aug 07:00:00 & 0.8725 & 0.8841 & 1.2421                  & 0.8783  \\ 
\hline
4  & 15-Aug 07:00:00 & 0.8909 & 0.9193 & 1.2802                  & 0.9049       \\ 
\hline
5  & 15-Aug 10:33:00 & 0.8337 & 0.6671 & 1.0677                  & 0.7411       \\ 
\hline
6  & 16-Aug 03:42:00 & 0.9215 & 0.8629 & 1.2625                  & 0.8913       \\ 
\hline
7  & 16-Aug 07:00:00 & 0.9196 & 0.8635 & 1.2615                  & 0.8907       \\ 
\hline
8  & 16-Aug 19:00:00 & 0.8954 & 0.8619 & 1.2428                  & 0.8783  \\ 
\hline
9  & 17-Aug 04:13:00 & 0.9434 & 0.6603 & 1.1515                  & 0.7769  \\ 
\hline
10 & 18-Aug 17:00:00 & 0.9198 & 0.8142 & 1.2284                  & 0.8638       \\ 
\hline
11 & 19-Aug 09:55:00 & 0.9741 & 0.7835 & 1.2501                  & 0.8685       \\ 
\hline
12 & 20-Aug 04:48:00 & 0.9606 & 0.7906 & 1.2441                  & 0.8674  \\ 
\hline
13 & 20-Aug 17:00:00 & 0.9462 & 0.8038 & 1.2416                  & 0.8692       \\ 
\hline
14 & 22-Aug 02:00:00 & 0.9709 & 0.7951 & 1.2549                  & 0.8743       \\ 
\hline
15 & 22-Aug 18:30:00 & 0.9136 & 0.8141 & 1.2237                  & 0.8610       \\ 
\hline
16 & 23-Aug 08:16:00 & 0.9662 & 0.8287 & 1.2729                  & 0.8922  \\ 
\hline
17 & 24-Aug 04:42:00 & 0.8694 & 0.8557 & 1.2198                  & 0.8625  \\ 
\hline
18 & 25-Aug 07:47:00 & 0.8602 & 0.8828 & 1.2326                  & 0.8714       \\ 
\hline
19 & 26-Aug 03:58:00 & 0.8214 & 0.9198 & 1.2332                  & 0.8678       \\ 
\hline
20 & 27-Aug 04:00:00 & 0.8492 & 0.9098 & 1.2445                  & 0.8785  \\ 
\hline
21 & 28-Aug 09:25:00 & 0.8808 & 0.9137 & 1.2691                  & 0.8969       \\ 
\hline
22 & 28-Aug 10:00:00 & 0.8989 & 0.8756 & 1.2549                  & 0.8871  \\ 
\hline
23 & 29-Aug 19:00:00 & 0.8844 & 0.8824 & 1.2493                  & 0.8834       \\ 
\hline
24 & 31-Aug 10:25:00 & 0.8837 & 0.9128 & 1.2705                  & 0.8980       \\ 
\hline
25 & 02-Sep 08:00:00 & 0.8735 & 0.8717 & 1.2340                  & 0.8726       \\ 
\hline
26 & 03-Sep 09:53:00 & 0.8830 & 0.8866 & 1.2513                  & 0.8848  \\ 
\hline
27 & 04-Sep 02:38:00 & 0.9049 & 0.8724 & 1.2569                  & 0.8883       \\
\hline
\end{tabular}
\caption[Assessment of estimated fire arrival time for the Cougar Creek Fire. Estimates were made using  the multigrid strategy.]{Assessment of estimated fire arrival time for the Cougar Creek Fire. Estimates were made without using  the multigrid strategy.  Each estimate was made using all the available satellite data up to the the time of the infrared perimeter observation that is used to make the comparison. As can be seen by examining the area columns in the table, the fire experienced little to no growth after August 28. Figure \ref{fig:cougar_test_growth} shows how the area of the fire changed with time.}
\label{tbl:cougar_multigrid_results}
\end{table}

\begin{figure}[htbp]
\centering
\includegraphics[width=0.7\textwidth]{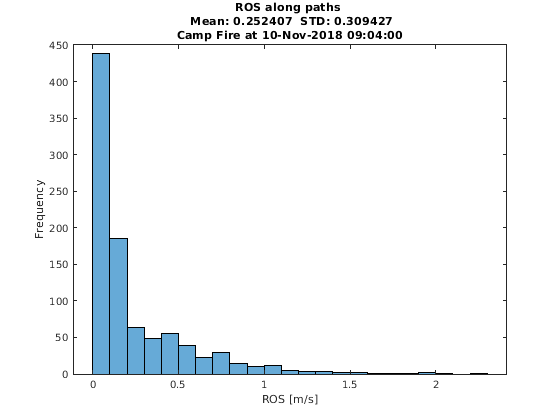}
 \caption{Histogram of the ROS computed along the paths in the directed graph of the Camp Fire. The mean rate of spread was 0.2524 m/s. This is a high number when compared with other fires studied in this research.}
 \label{fig:camp_path_ros}
\end{figure}

\begin{figure}[!ht]
\centering
       \includegraphics[width=0.45\textwidth]{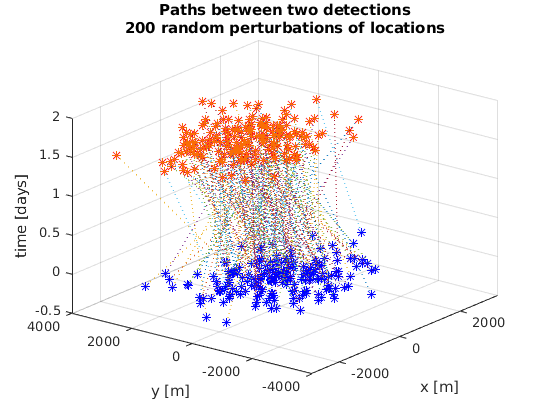}
       \includegraphics[width=0.45\textwidth]{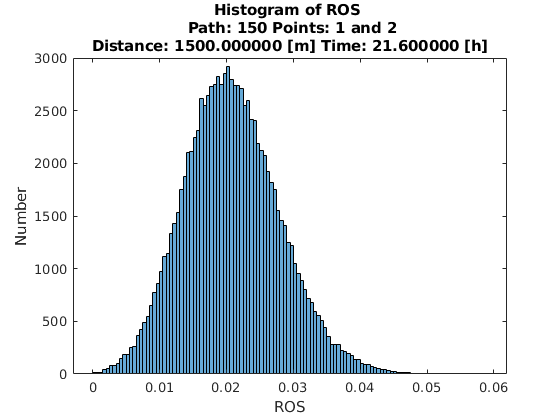}
       \includegraphics[width=0.45\textwidth]{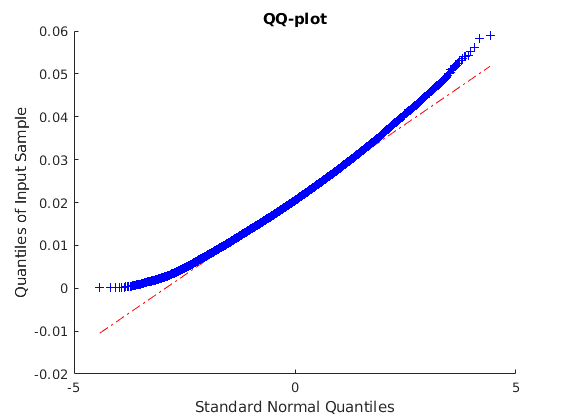}
     \caption[The uncertainty in the position and fire arrival time of active fire detections produces uncertainty in the computed ROS between consecutive detections in a path.]{The uncertainty in the position and fire arrival time of active fire detections produces uncertainty in the computed ROS between consecutive detections in a path. The panel on the left shows 200 perturbations of the time and locations of two fire detections connected in a path. The perturbation of the spatial locations was made following a Gaussian distribution to reflect the geolocation error. The perturbation of the points in time was made using a uniform distribution. In the upper right is a histogram of the ROS between two consecutive  and randomly perturbed points along a path used for an actual estimate of the fire arrival time of the Patch Springs Fire. The QQ plot in the panel on the bottom shows the ROS is not normally distributed.}
\label{fig:random_points}
\end{figure}

\begin{figure}[!h]
\begin{center}
  \includegraphics[width = 0.60\textwidth]{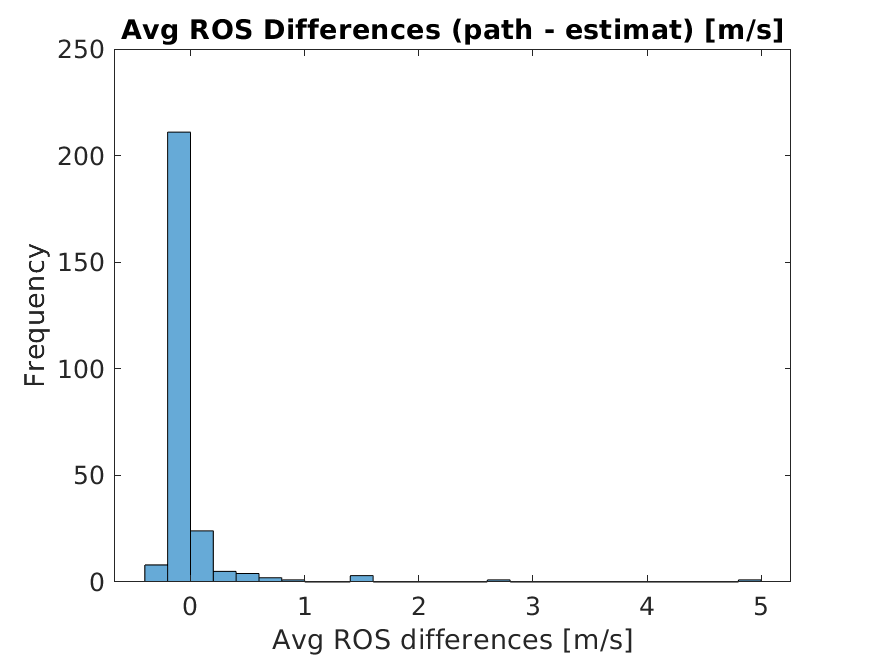}
\caption[The difference between the average ROS along paths in the graphs and in the fire arrival time cone estimates for the artificial ``ground truth" experiment, using the multigrid approach and a spacing between additional points interpolated along paths at a distance of 2000 meters.]{The difference between the average ROS along paths in the graphs and in the fire arrival time cone estimates for the artificial ``ground truth" experiment, using the multigrid approach and a spacing between additional points interpolated along paths at a distance of 2000 meters. 280 estimated fire arrival times were compared with the ``ground truth." The ROS along paths was on average 0.0053 m/s faster than in the estimate with a standard deviation of 0.4118 m/s. This figure gives a sense of how well the ROS compares on average in each simulation, but has no information about how the difference in the ROS can vary with location in individual scenarios.}
\label{fig:ros_diff}
\end{center}
\end{figure} 

\begin{figure}[htbp]
\centering
\includegraphics[width=0.45\textwidth]{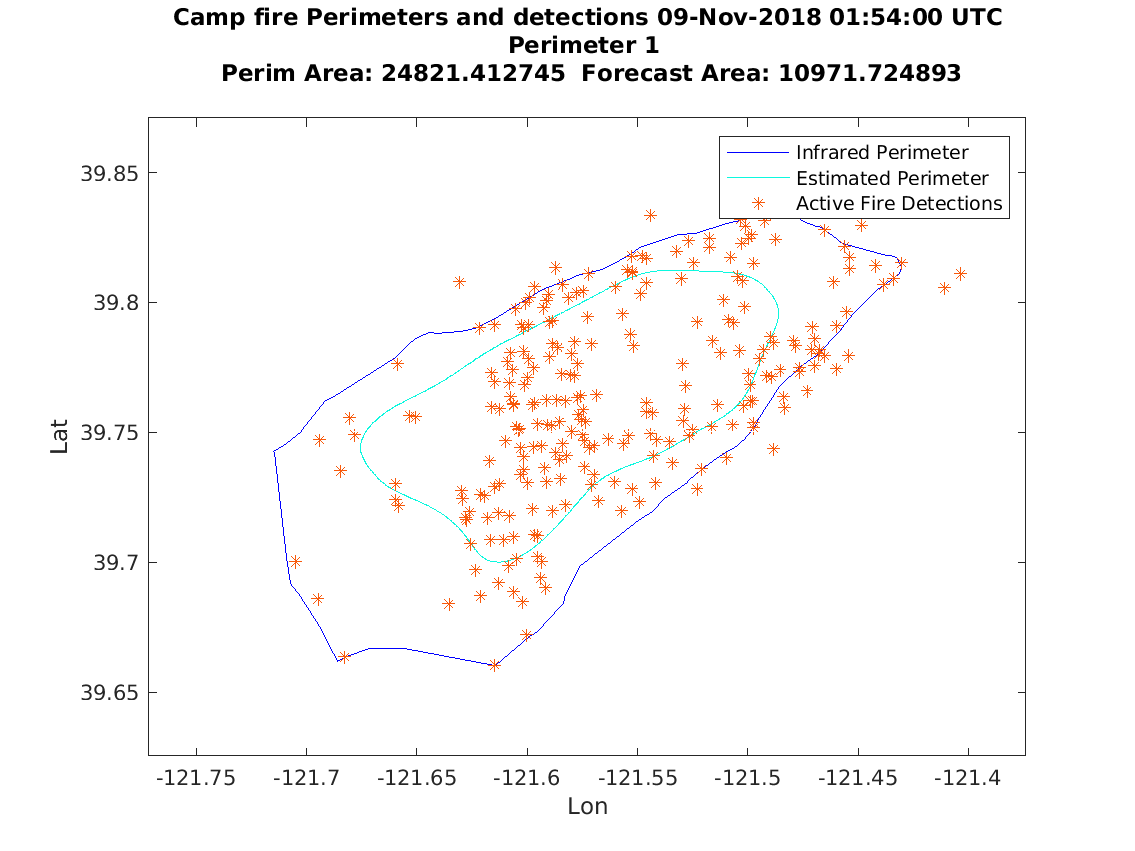}
\includegraphics[width=0.45\textwidth]{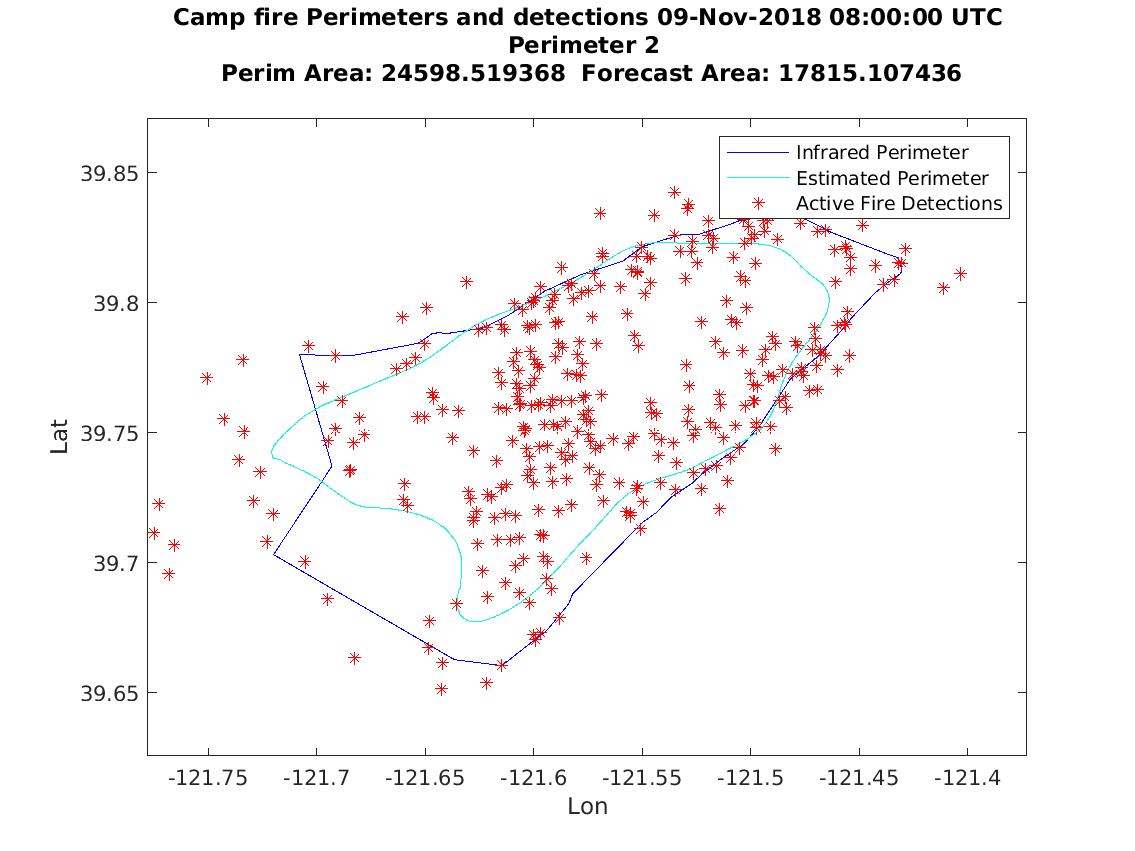}

\includegraphics[width=0.45\textwidth]{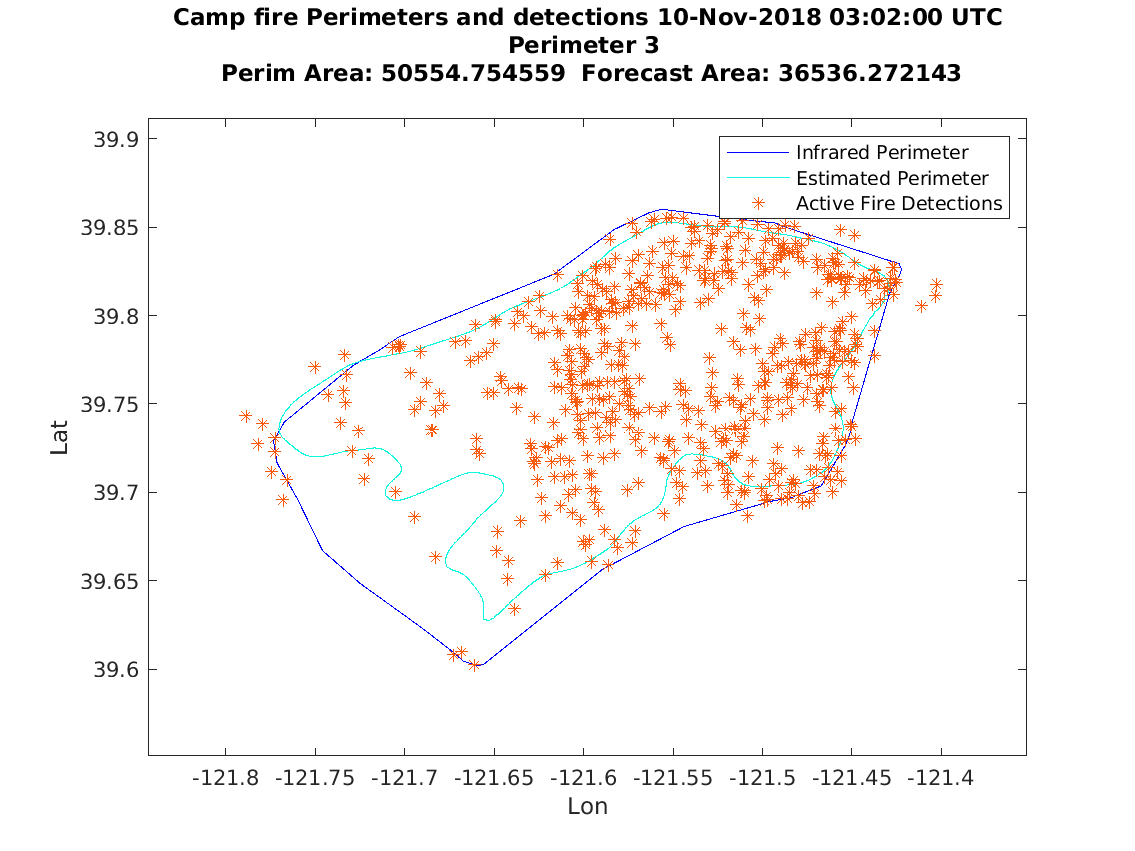}
\includegraphics[width=0.45\textwidth]{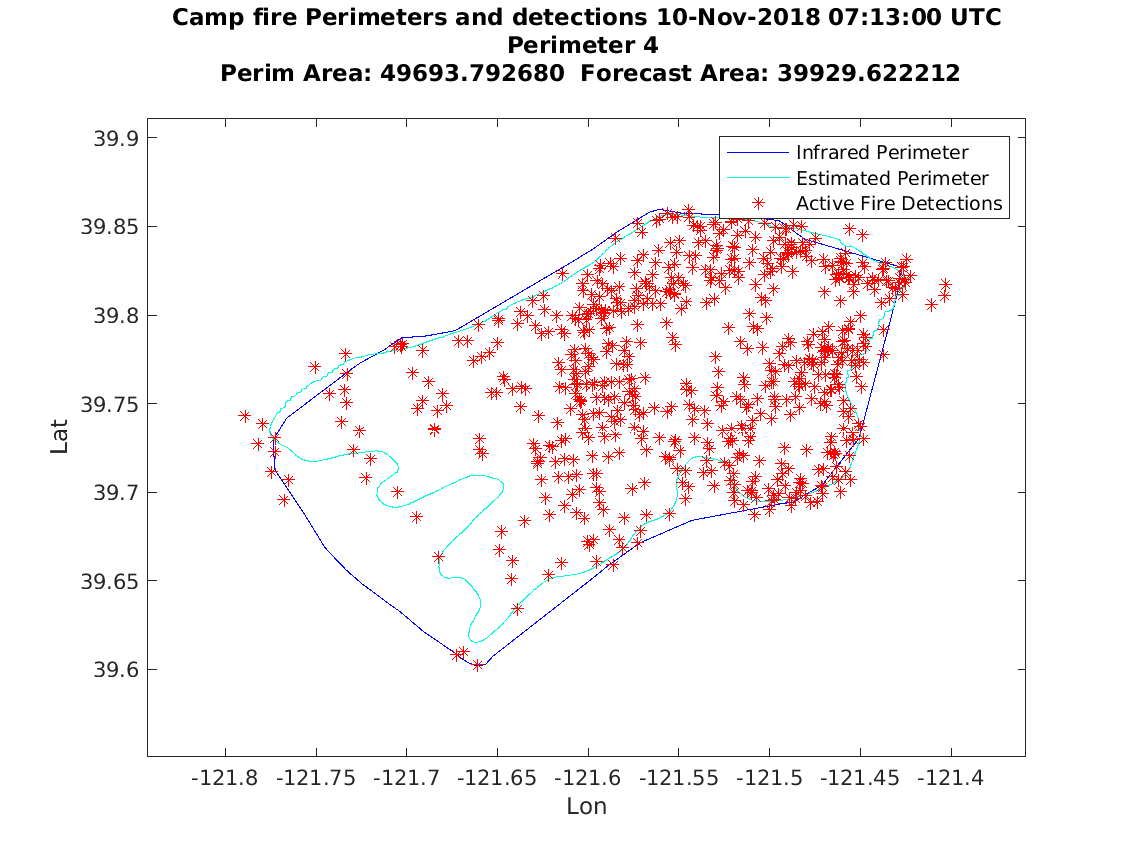}
 \caption{Comparison of infrared perimeters and the perimeters of an estimated fire arrival time for the Camp Fire. The progression of the fire during the first several days was estimated from satellite data. The top row shows the first two perimeter comparisons and the bottom row shows the third and fourth. Note that all the estimated perimeters underestimate the size of the fire. }
 \label{fig:camp_perim_sequence}
\end{figure}

\begin{figure}[!h]
\begin{center}
  \includegraphics[width = 0.45\textwidth]{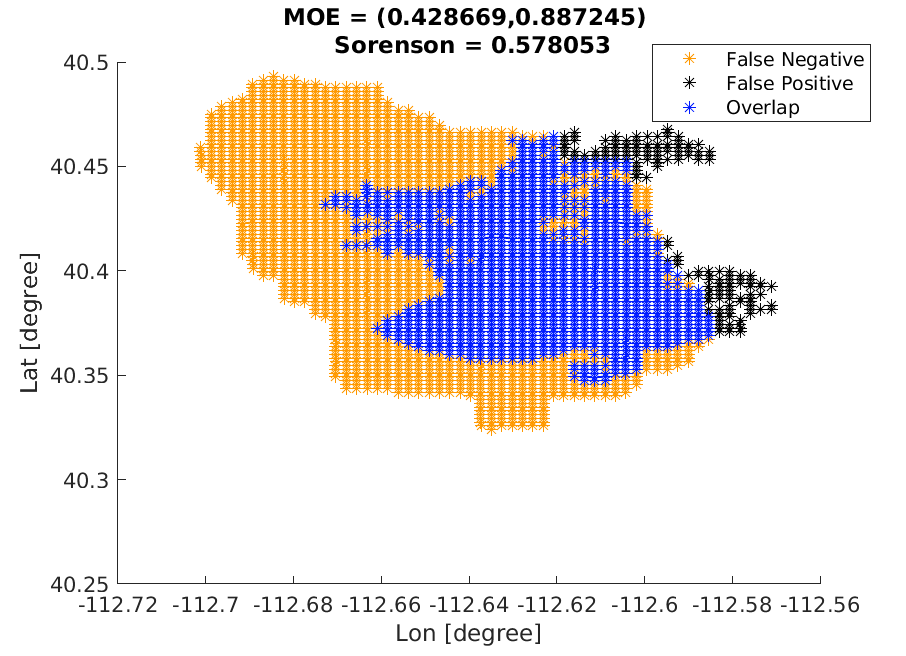}
  \includegraphics[width = 0.45\textwidth]{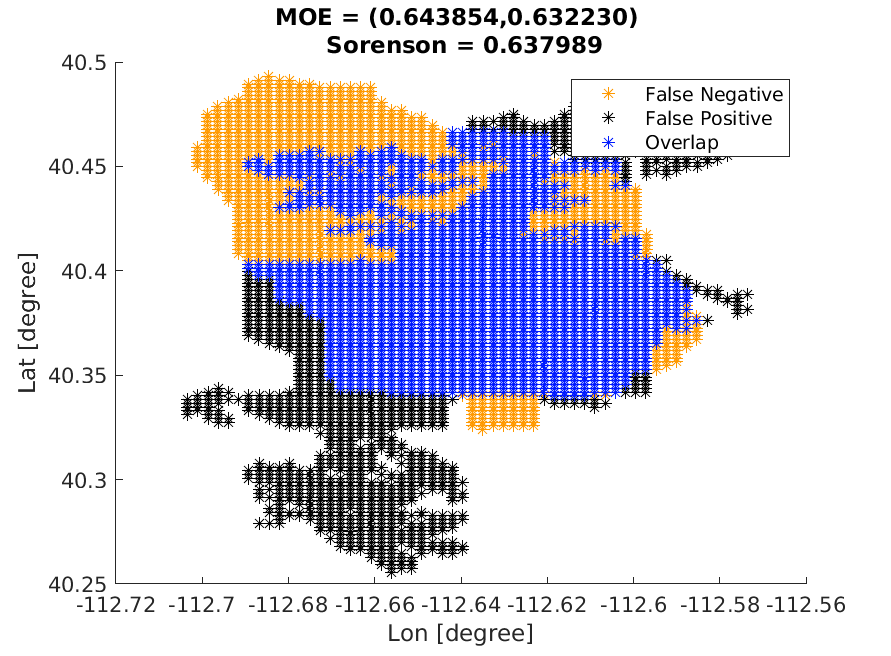}
  
  \includegraphics[width = 0.45\textwidth]{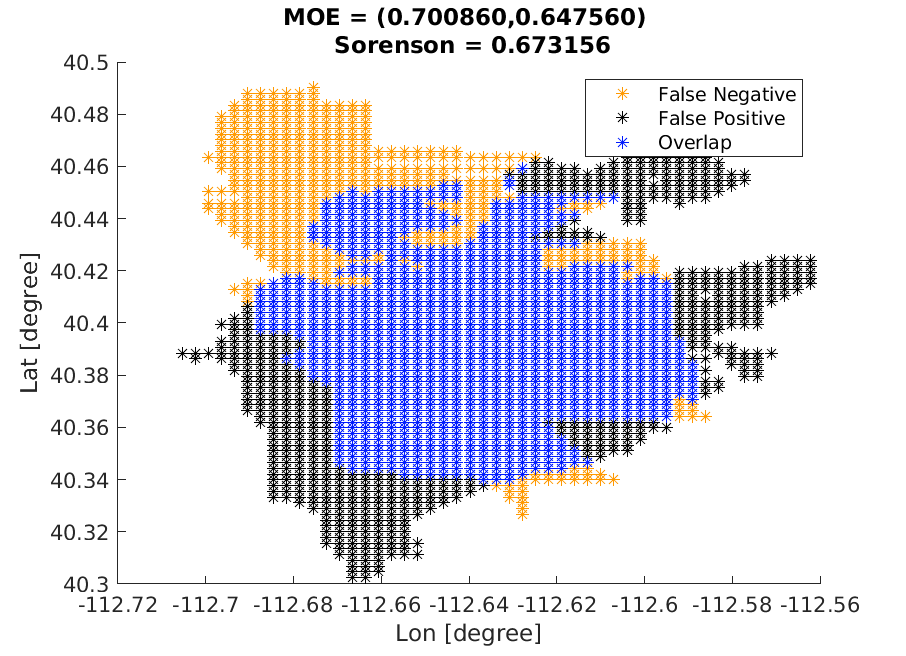}
  \includegraphics[width = 0.45\textwidth]{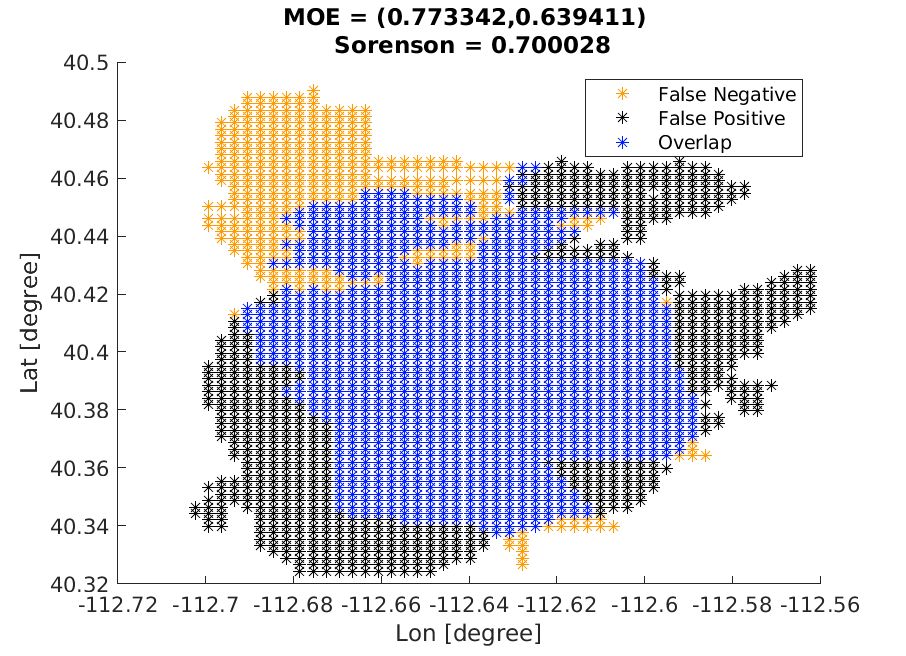} 
  
  \includegraphics[width = 0.45\textwidth]{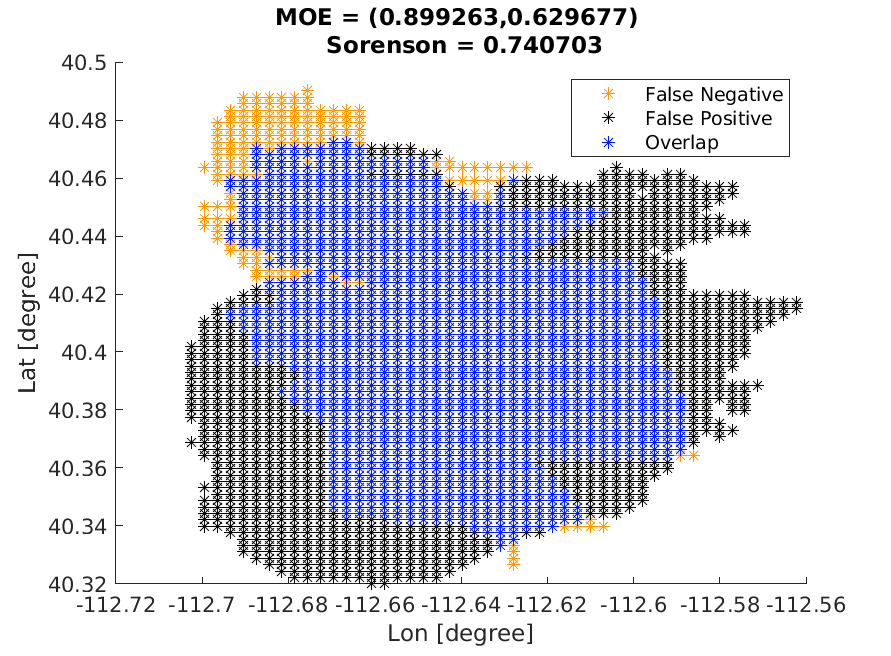}
  \includegraphics[width = 0.45\textwidth]{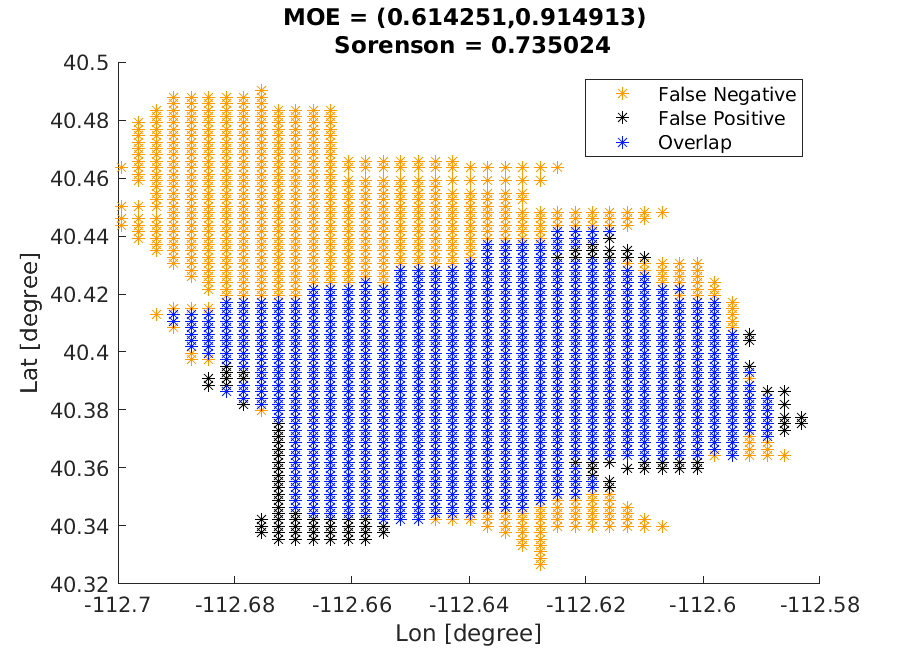}

  \caption[Assessment of the cycles of the Patch Fire simulation.]{Assessment of the cycles of the Patch Fire simulation. The graphics give a graphical representation of how the fire area of simulation cycles match a perimeter observation made on August 16 at 09:47 UTC. The panels from to to bottom and left to right cycles 0 through 4, with the lower right figure being the perimeter comparison for a simulation initialized using an estimated fire arrival time. The cycling method does produce better estimates as data is assimilated, but the large black and red areas in the figures indicate that model is not accurately predicting where the fire will be burning. }
  \label{fig:patch_moe_cycles}
  \end{center}
\end{figure}

\begin{figure}[!h]
\begin{center}
  \includegraphics[width = 0.9\textwidth]{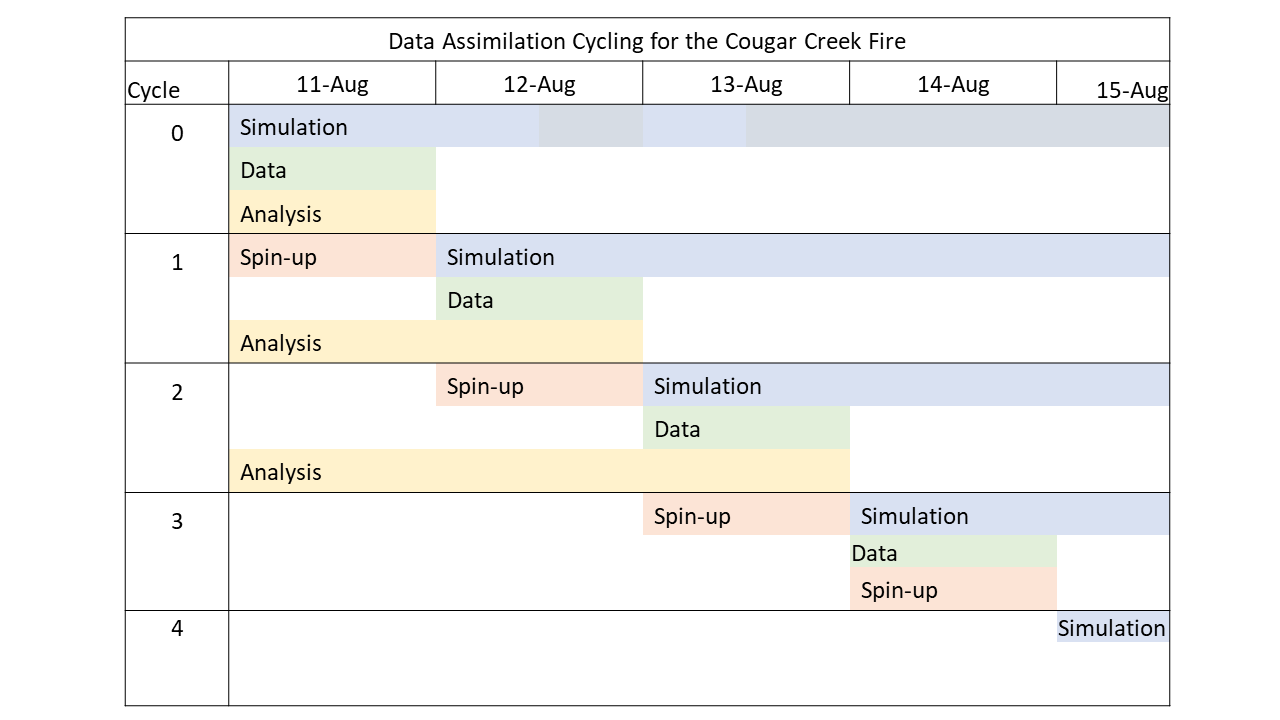}
  \caption{Schematic for several data assimilation cycles run consecutively to simulate the Cougar Creek fire of 2015. Each simulation was run until 09:00 UTC on August 15 for the sake of comparison. In typical operational usage, simulations would not extend more than two days beyond the end of the spin-up period of the previous cycle.}
  \label{fig:cougar_short_schematic}
  \end{center}
\end{figure}

\begin{figure}[!h]
\begin{center}
  \includegraphics[width = 0.9\textwidth]{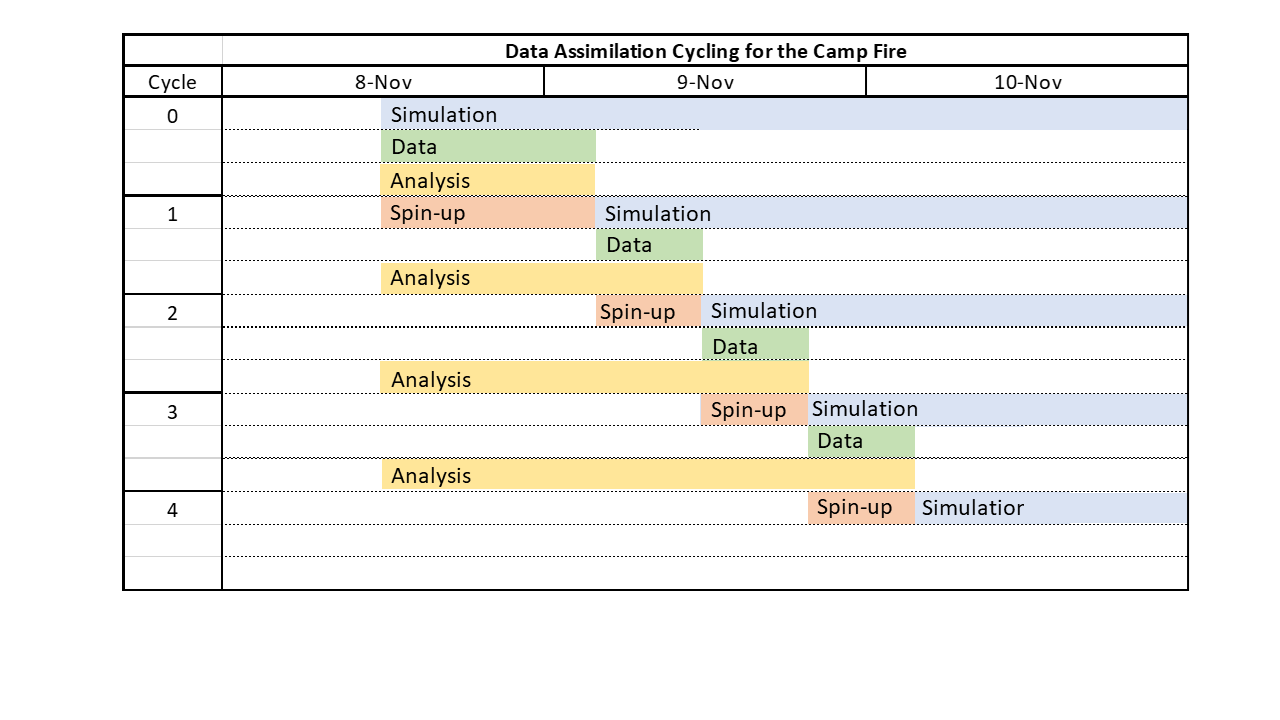}
  \caption{Schematic for several data assimilation cycles run consecutively in order to simulate the Camp fire of 2018.}
  \label{fig:camp_schematic}
  \end{center}
\end{figure}

\begin{figure}[!h]
\begin{center}
  \includegraphics[width = 0.45\textwidth]{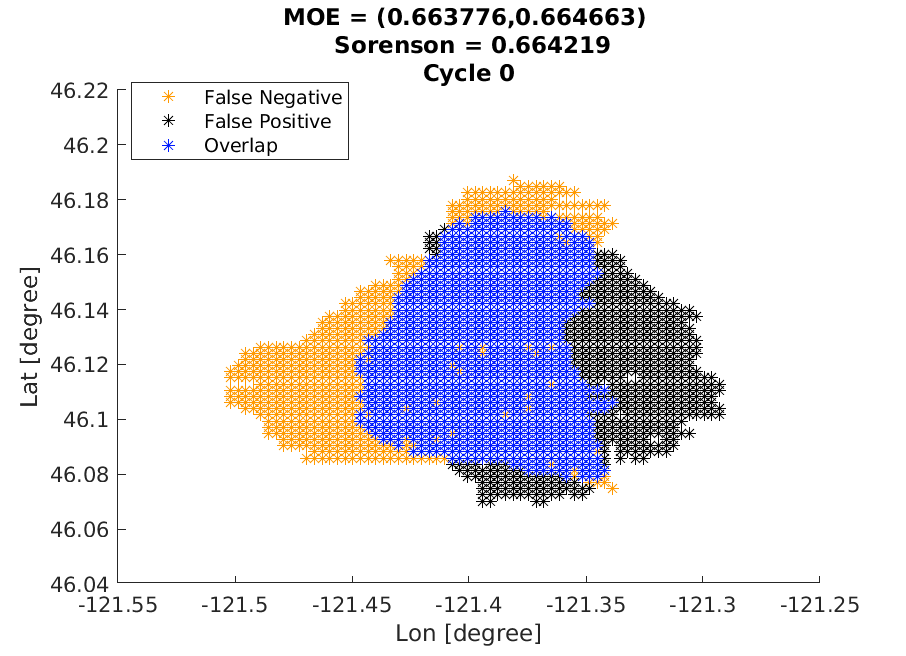}
  \includegraphics[width = 0.45\textwidth]{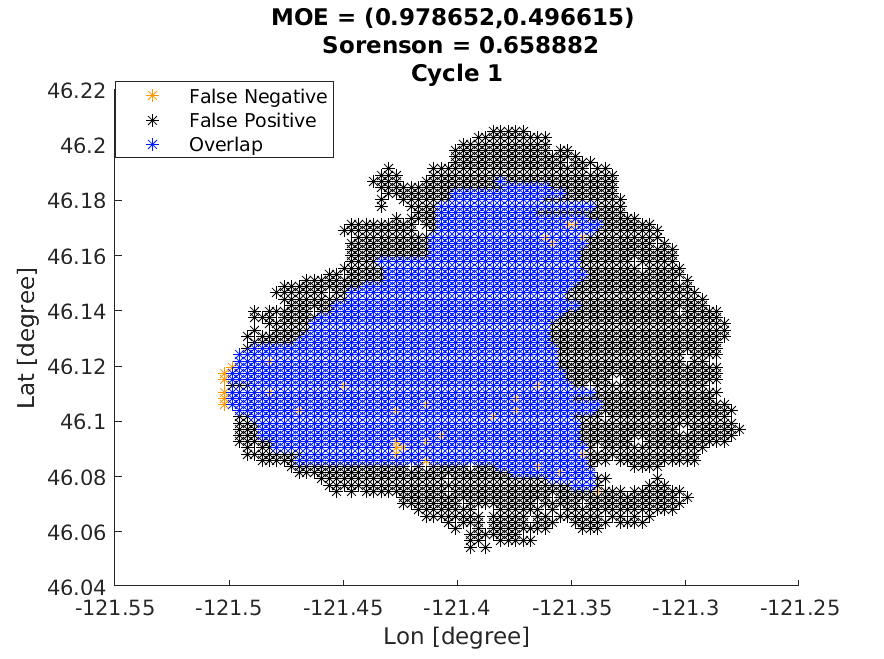}
  
  \includegraphics[width = 0.45\textwidth]{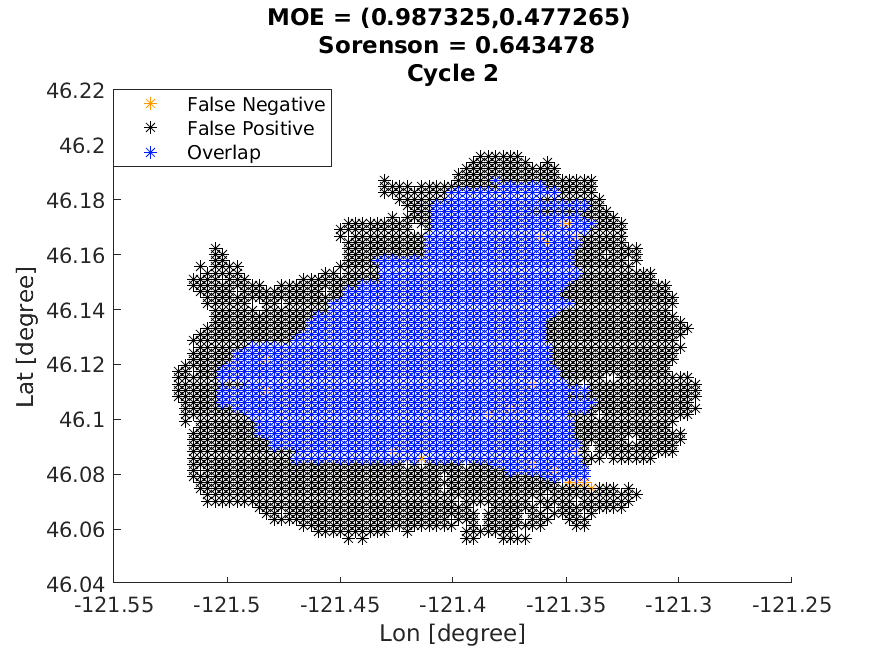}
  \includegraphics[width = 0.45\textwidth]{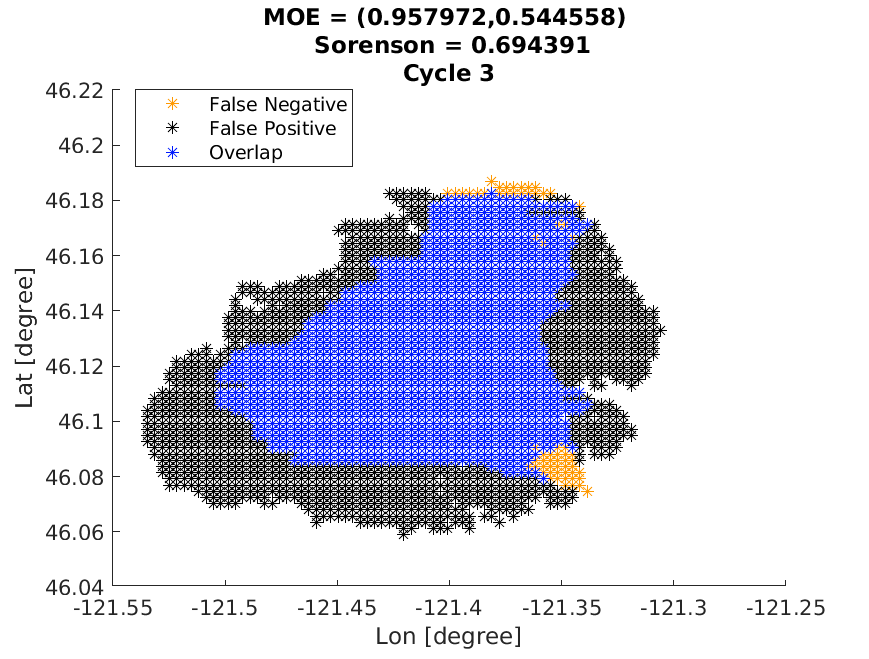}
  
  \includegraphics[width = 0.45\textwidth]{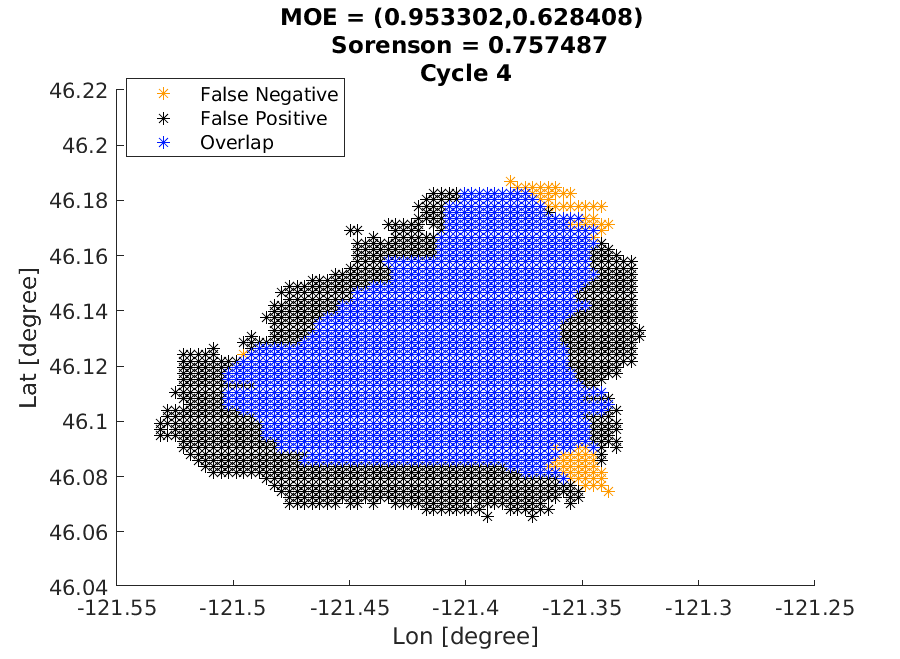}
  \includegraphics[width = 0.45\textwidth]{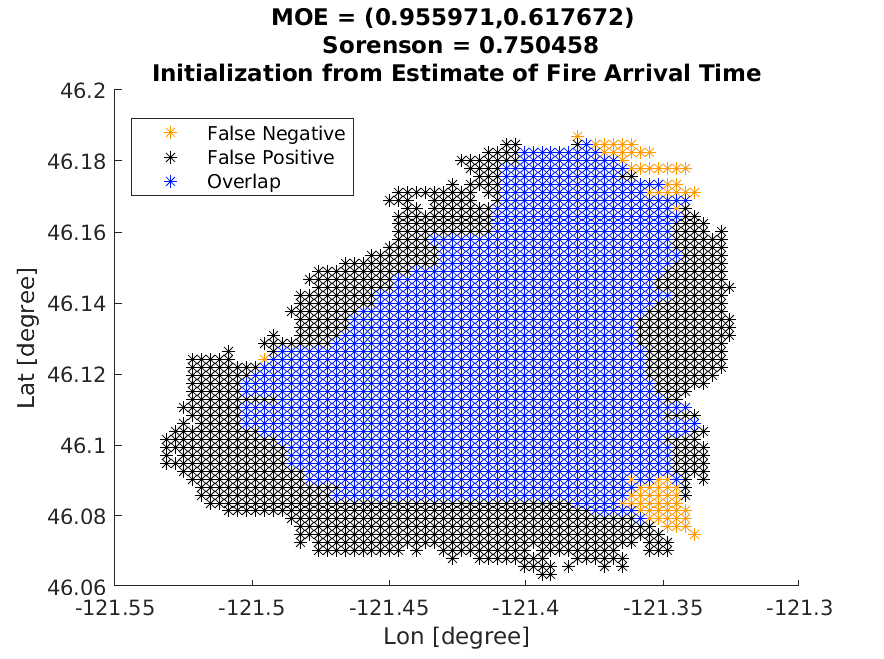}

  \caption{Assessment of the cycles of the Cougar Creek Fire simulation. The panel in the lower right shows the result from a simulation initialized from an estimate of the fire arrival time derived from satellite fire data.}
  \label{fig:cougar_moe_cycles}
  \end{center}
\end{figure}

\begin{figure}[!h]
\begin{center}
  \includegraphics[width = 0.45\textwidth]{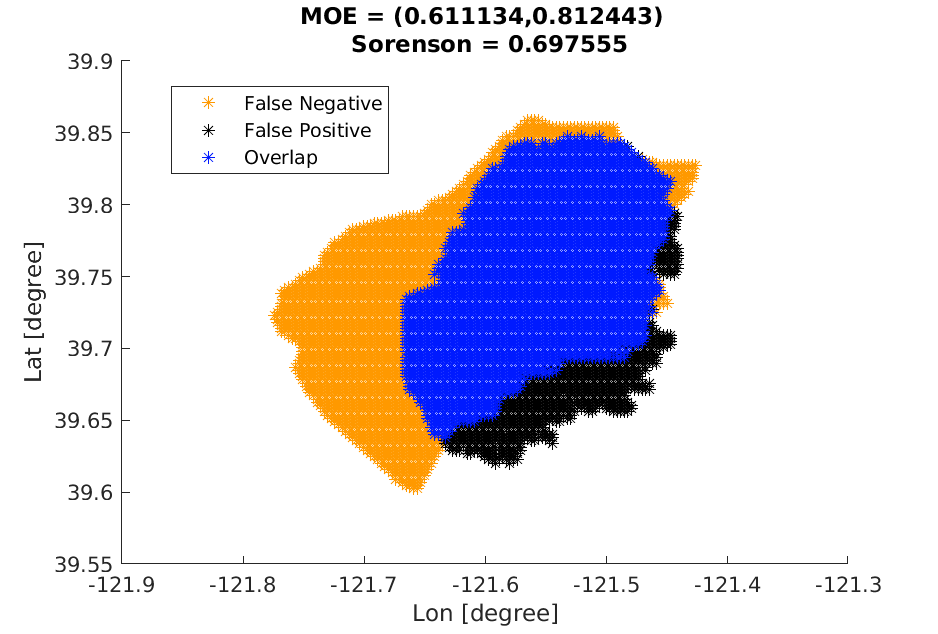}
  \includegraphics[width = 0.45\textwidth]{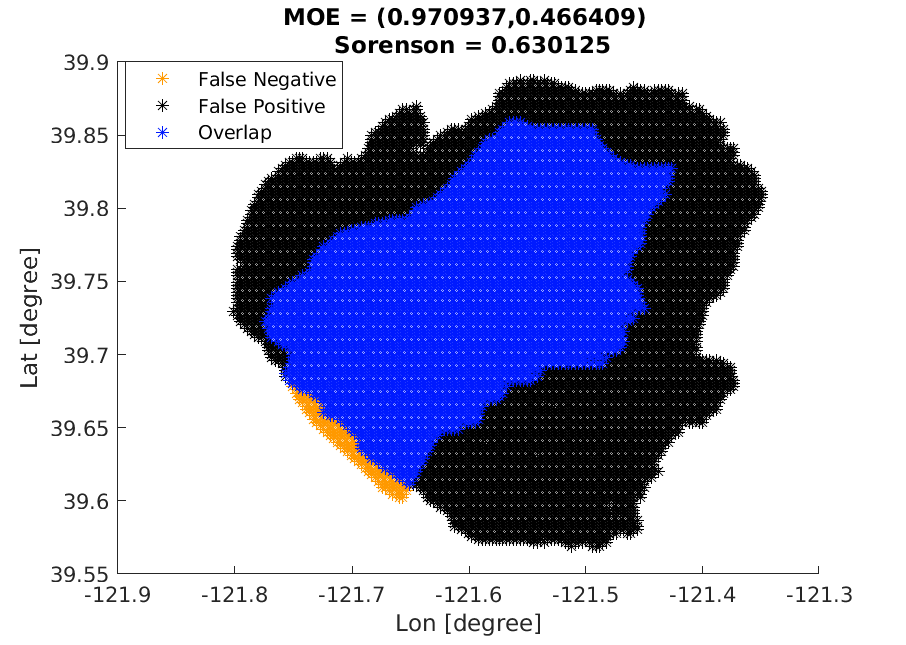}
  
  \includegraphics[width = 0.45\textwidth]{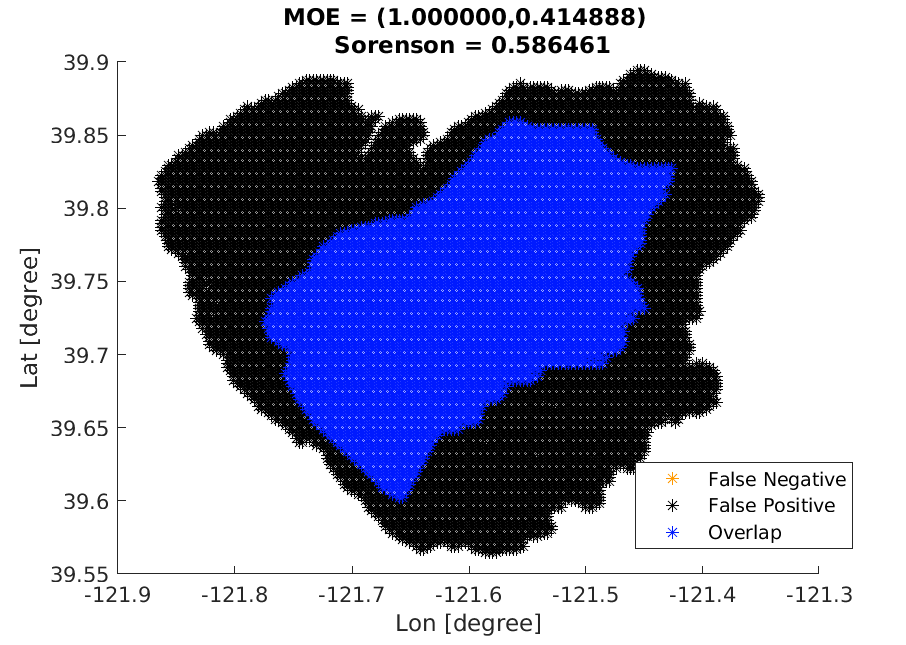}
  \includegraphics[width = 0.45\textwidth]{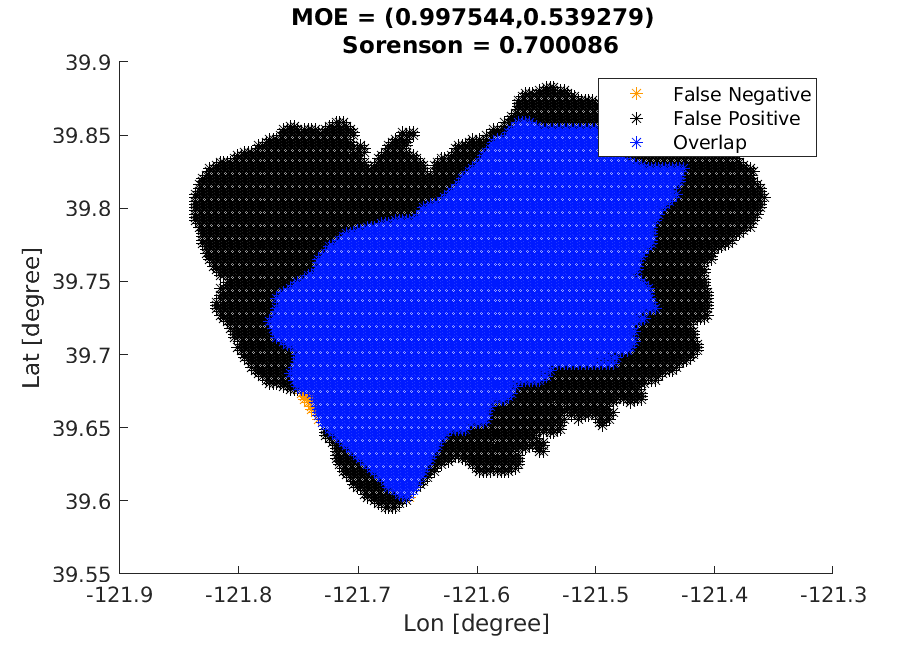}
  
  \includegraphics[width = 0.45\textwidth]{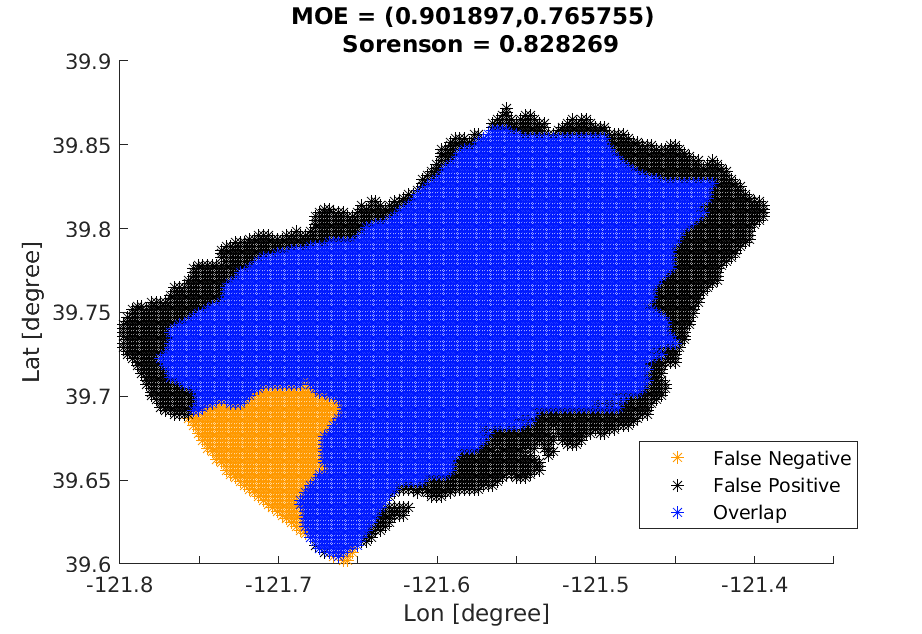}
  \includegraphics[width = 0.45\textwidth]{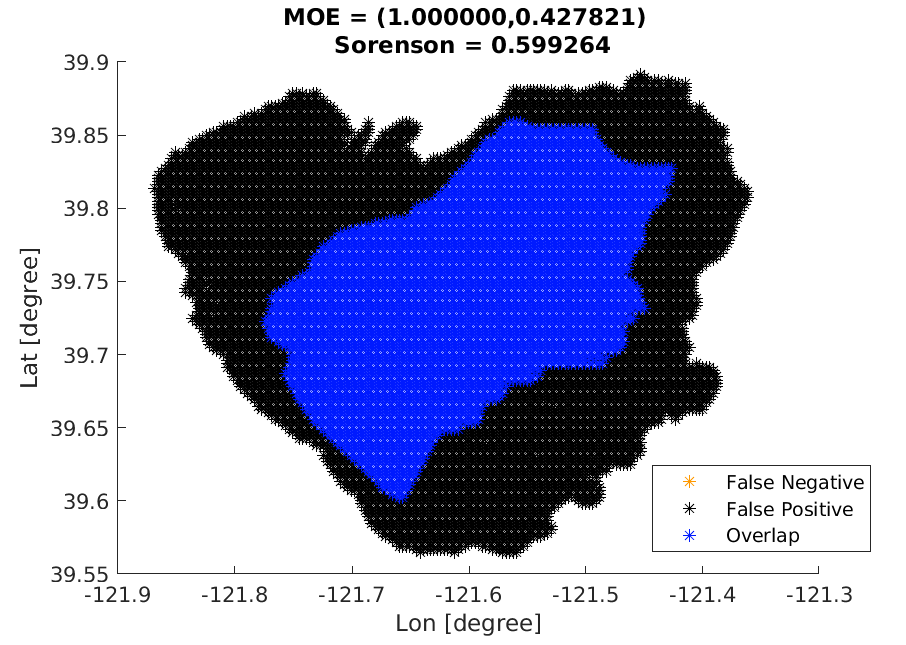}
  \caption{Assessment of the cycles of the Camp Fire simulation by comparison with an infrared perimeter observation made on November 10, 07:00 UTC. The panel in the lower right shows the result from a simulation initialized from an estimate of the fire arrival time derived from satellite fire data. Except for the initial cycle, the model tended to overestimate the fire size by expanding to the northwest and southeast more that was observed. These graphics summarize tha data recorded in Table \ref{tbl:camp_moe}.}
  \label{fig:camp_moe_cycles}
  \end{center}
\end{figure}

\begin{figure}[!h]
\begin{center}
\includegraphics[width = 0.45\textwidth]{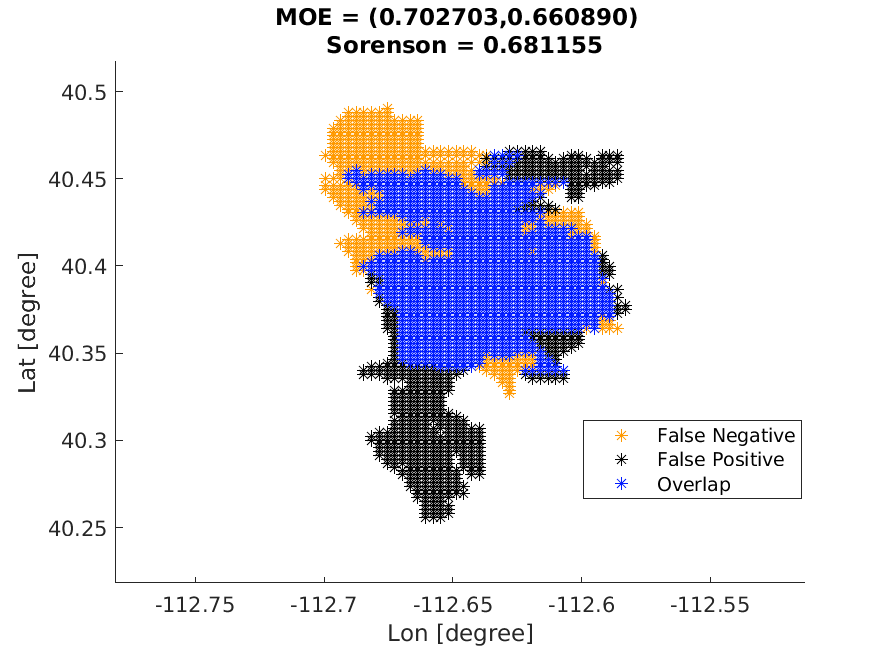}
\includegraphics[width = 0.45\textwidth]{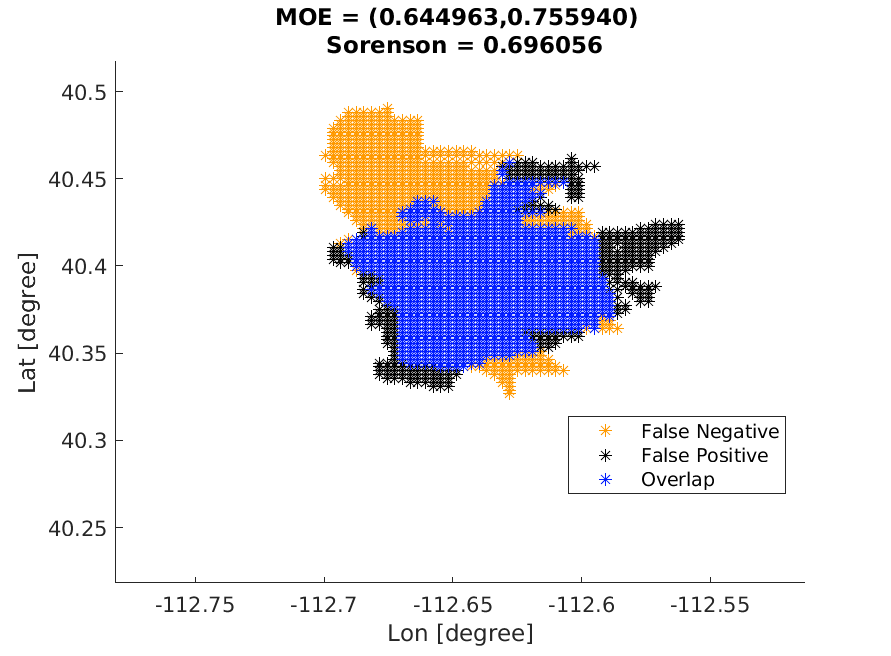}
\includegraphics[width = 0.45\textwidth]{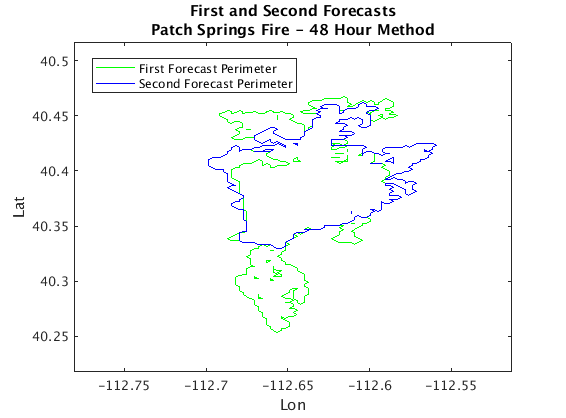}
  \caption[Assessment of the Patch Springs Fire simulation using the 48-hour forecast strategy.]{Assessment of the Patch Springs Fire simulation using the 48-hour forecast strategy. On the upper left and right are comparisons between the infrared perimeter observation and the first and second forecasts, respectively. On the bottom, the final perimeters of the first and second forecasts are displayed together. As can be seen, the data assimilation of a small number of satellite granules was able to slow the growth of the modeled fire and make for a better forecast.}
  \label{fig:patch_48_moe}
  \end{center}
\end{figure}

\begin{figure}[!h]
\begin{center}
\includegraphics[width = 0.45\textwidth]{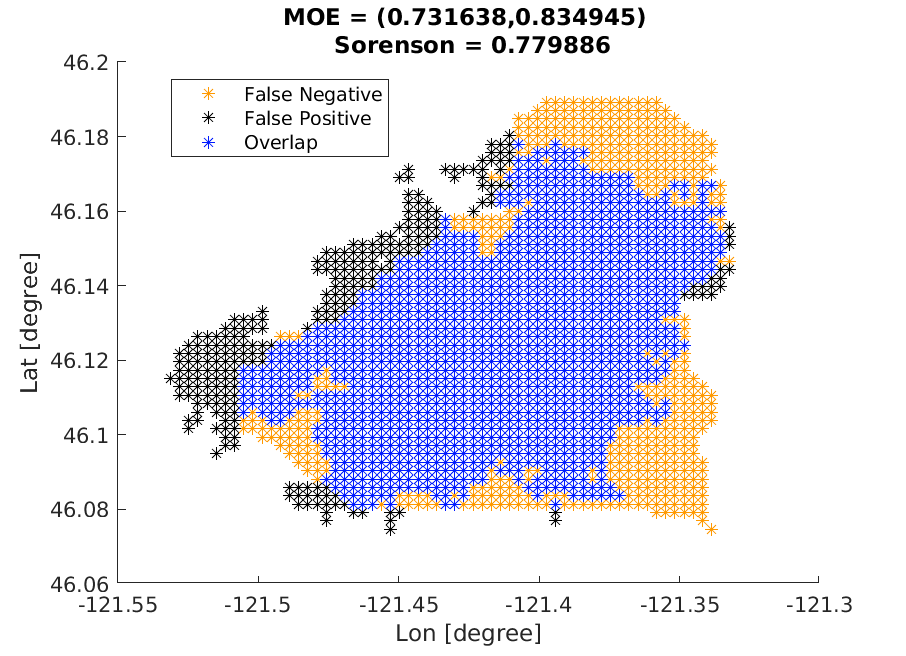}
\includegraphics[width = 0.45\textwidth]{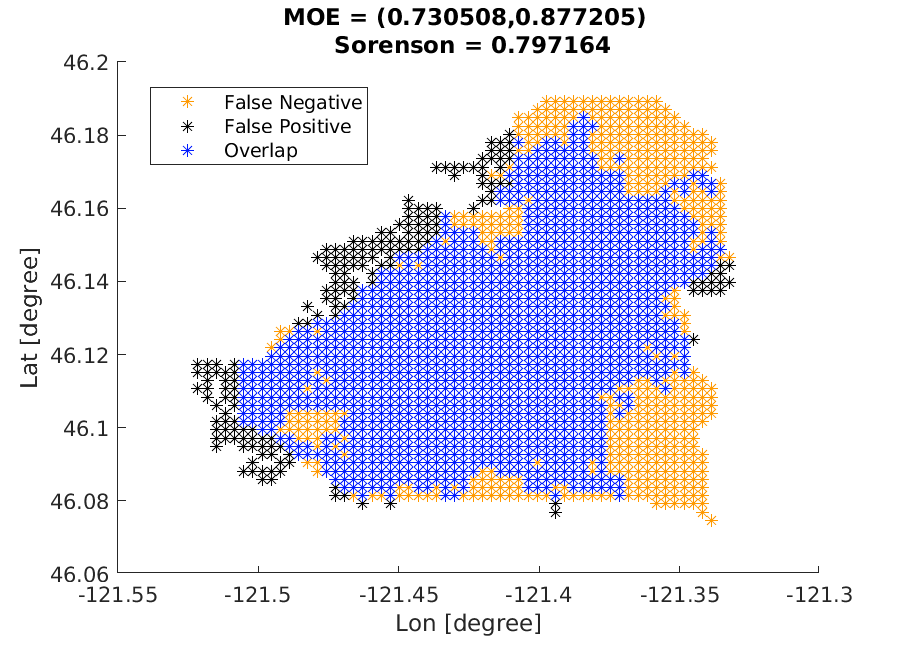}
\includegraphics[width = 0.45\textwidth]{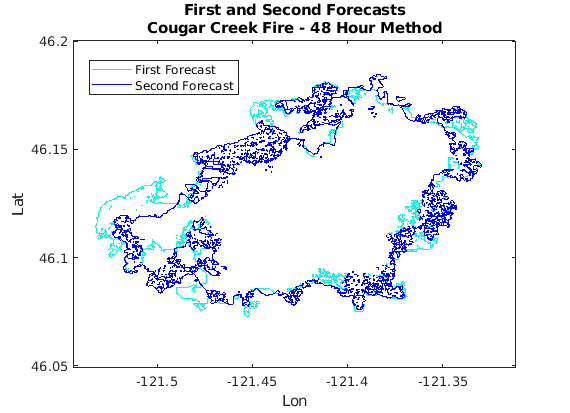}
  \caption[Assessment of the Cougar Creek Fire simulation using the 48-hour forecast strategy.]{Assessment of the Cougar Creek Fire simulation using the 48-hour forecast strategy. At the left and  center are comparisons between the infrared perimeter observation and the first and second forecasts, respectively. On the right, the final perimeters of the first and second forecasts are displayed together. As can be seen, the data assimilation of a small number of satellite granules was able to slow the growth of the modeled fire and make for a better forecast.}
  \label{fig:cougar_48_moe}
  \end{center}
\end{figure}

\begin{figure}[!ht]
\centering
\includegraphics[width=0.45\textwidth]{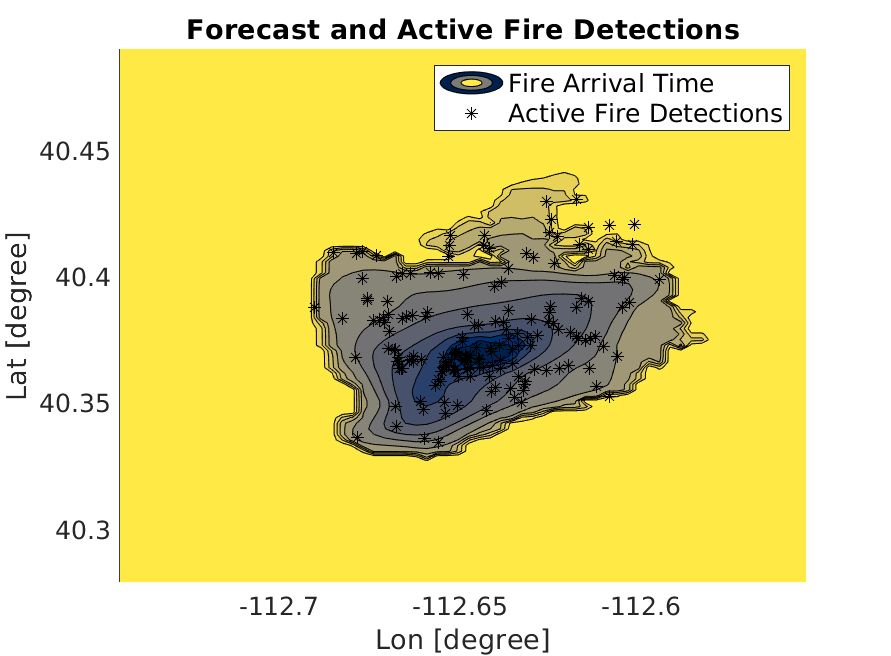}
\includegraphics[width=0.45\textwidth]{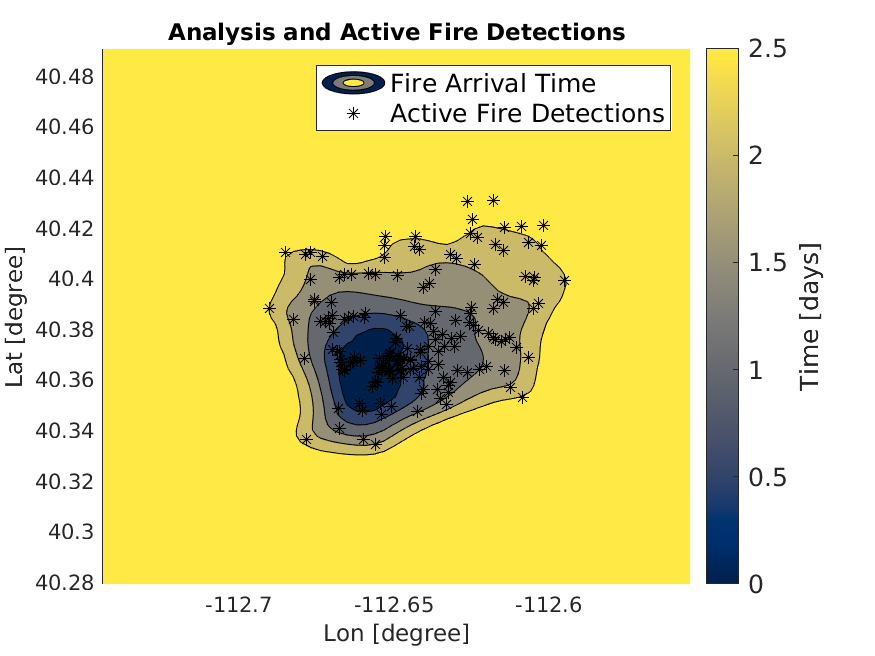}
\includegraphics[width=0.45\textwidth]{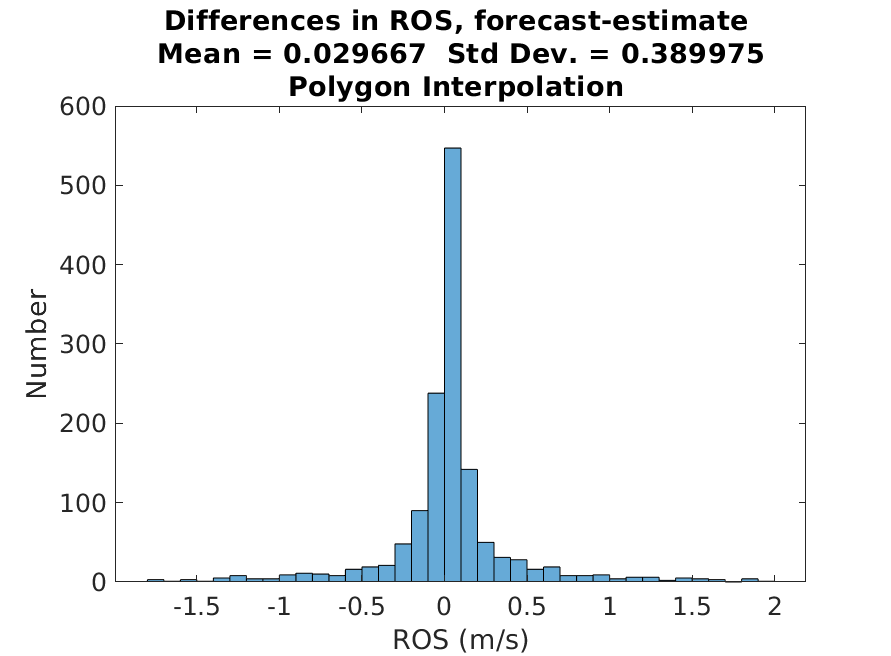}
     \caption[Cycle 1 of the Patch Fire simulation with adjustment of FMC.]{Cycle 1 of the Patch Fire simulation with adjustment of FMC. In the upper left and right are contours of of the fire arrival time for the forecast and the analysis, respectively. On the bottom is the histogram showing the difference between the ROS at mesh points in the fire domain.In this case the forecast area and the forecast ROS were higher than the estimate from the data, indicating an increase of the FMC by 0.1967\%}
\label{fig:patch_fmc_cycle_1}
\end{figure}

\begin{figure}[!ht]
\centering
\includegraphics[width=0.45\textwidth]{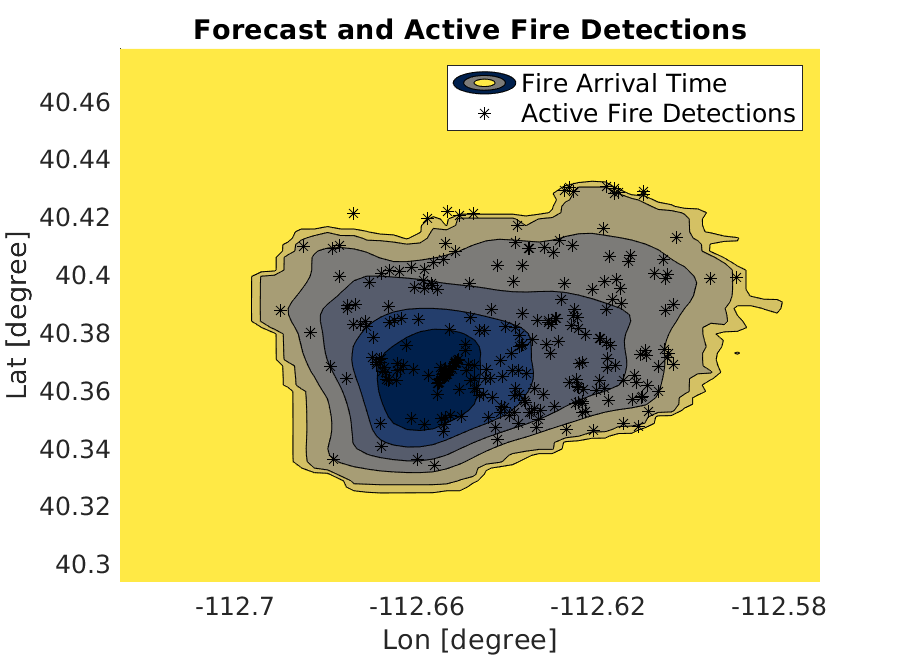}
\includegraphics[width=0.45\textwidth]{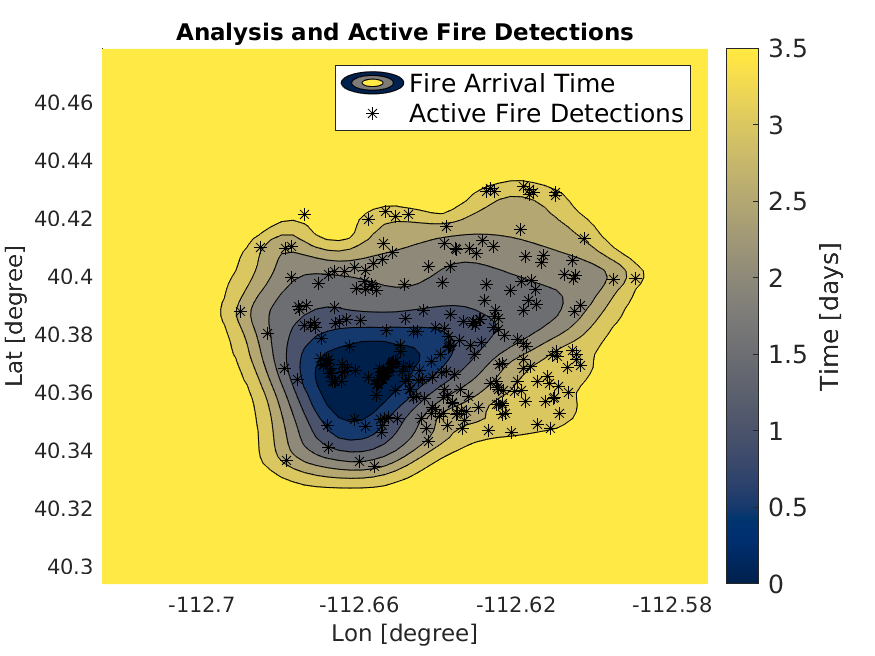}
\includegraphics[width=0.45\textwidth]{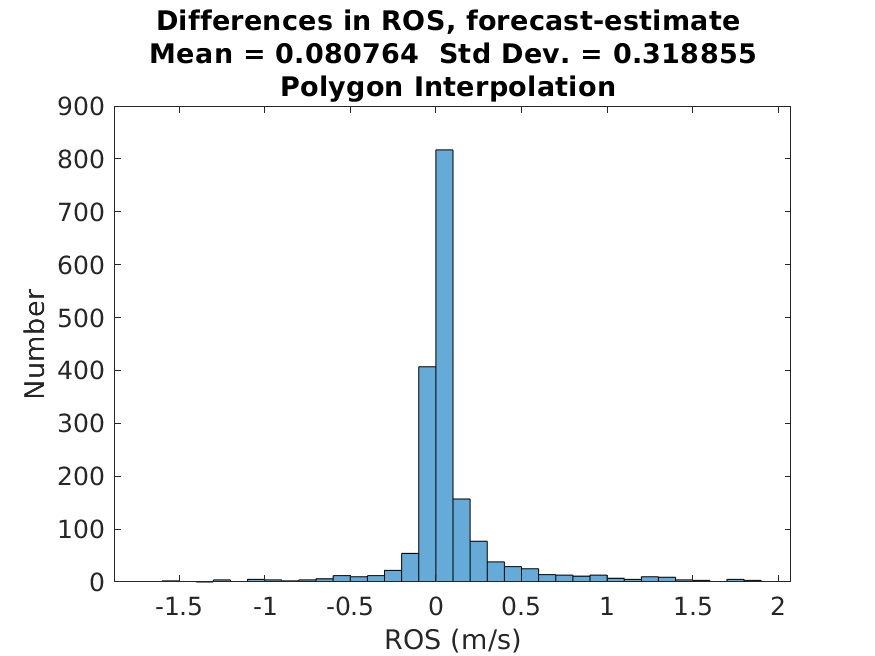}
     \caption[Cycle 2 of the Patch Fire simulation with adjustment of FMC.]{Cycle 2 of the Patch Fire simulation with adjustment of FMC. In the upper left and right are contours of of the fire arrival time for the forecast and the analysis, respectively. On the bottom is the histogram showing the difference between the ROS at mesh points in the fire domain. }
\label{fig:patch_fmc_cycle_2}
\end{figure}

\begin{figure}[!ht]
\centering
\includegraphics[width=0.45\textwidth]{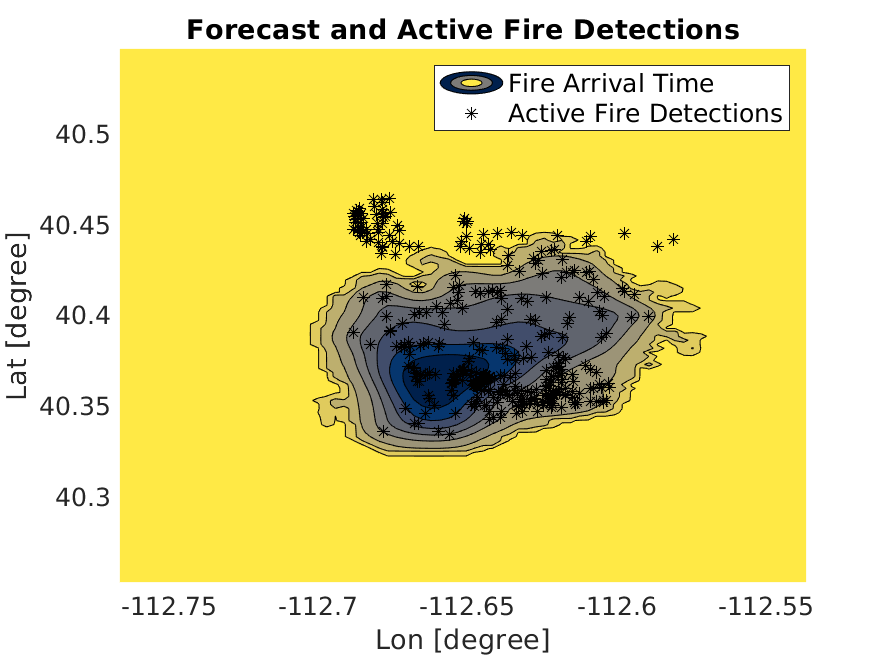}
\includegraphics[width=0.45\textwidth]{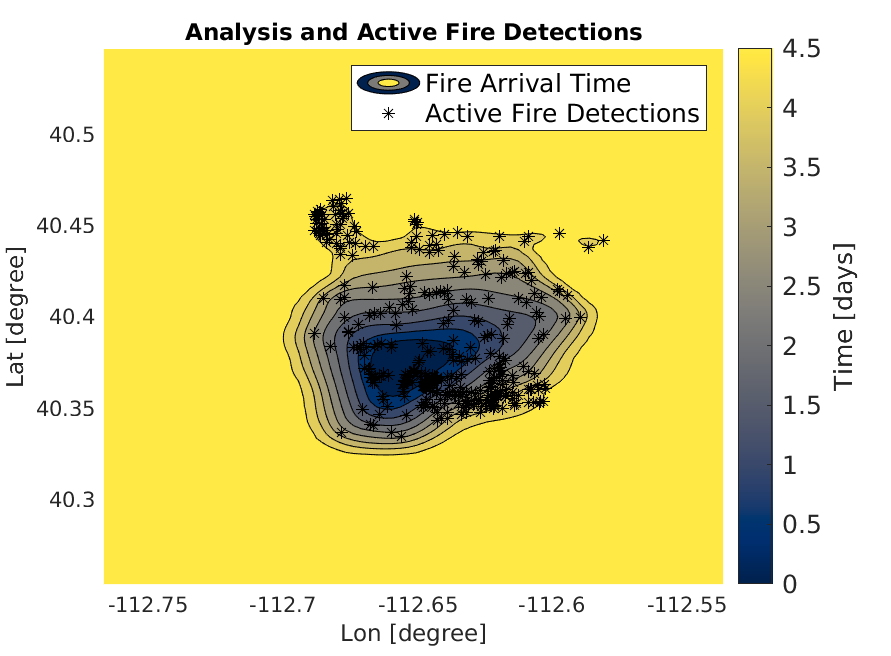}
\includegraphics[width=0.45\textwidth]{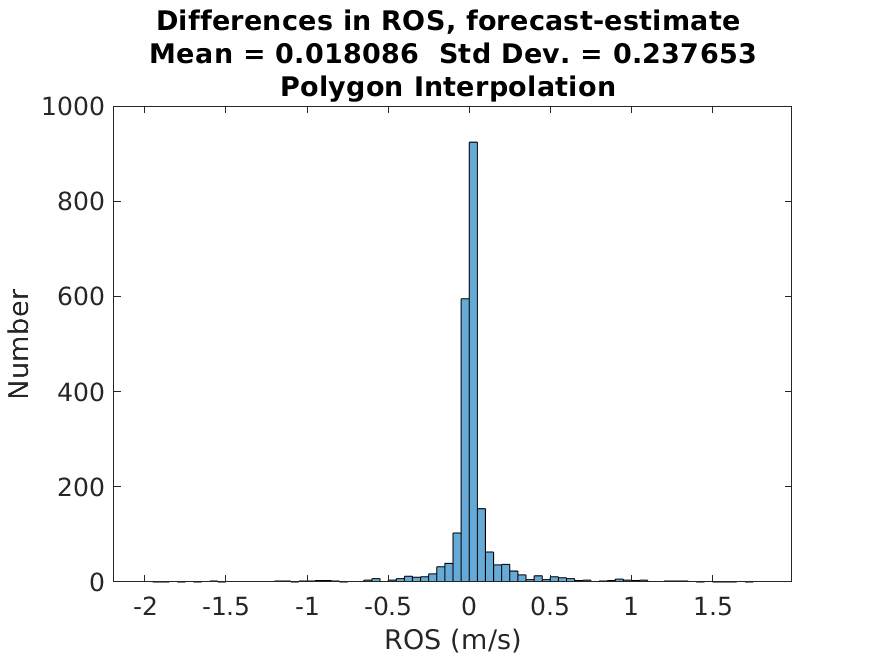}
     \caption[Cycle 3 of the Patch Fire simulation with adjustment of FMC.]{Cycle 3 of the Patch Fire simulation with adjustment of FMC. In the upper left and right are contours of of the fire arrival time for the forecast and the analysis, respectively. On the bottom is the histogram showing the difference between the ROS at mesh points in the fire domain. No changes to the FMC were made for the next simulation cycle.}
\label{fig:patch_fmc_cycle_3}
\end{figure}

\begin{figure}[!ht]
\centering
\includegraphics[width=0.45\textwidth]{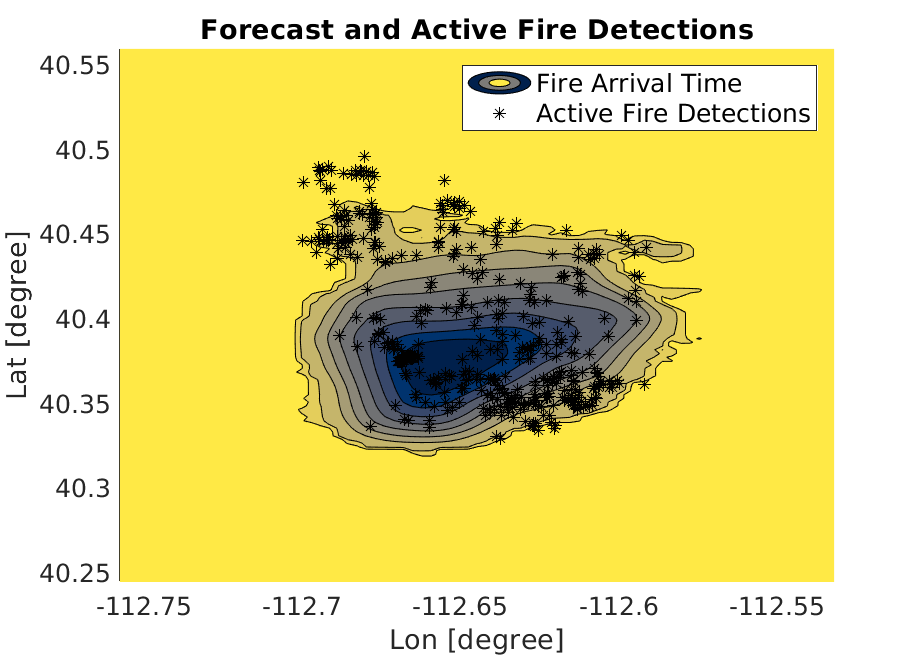}
\includegraphics[width=0.45\textwidth]{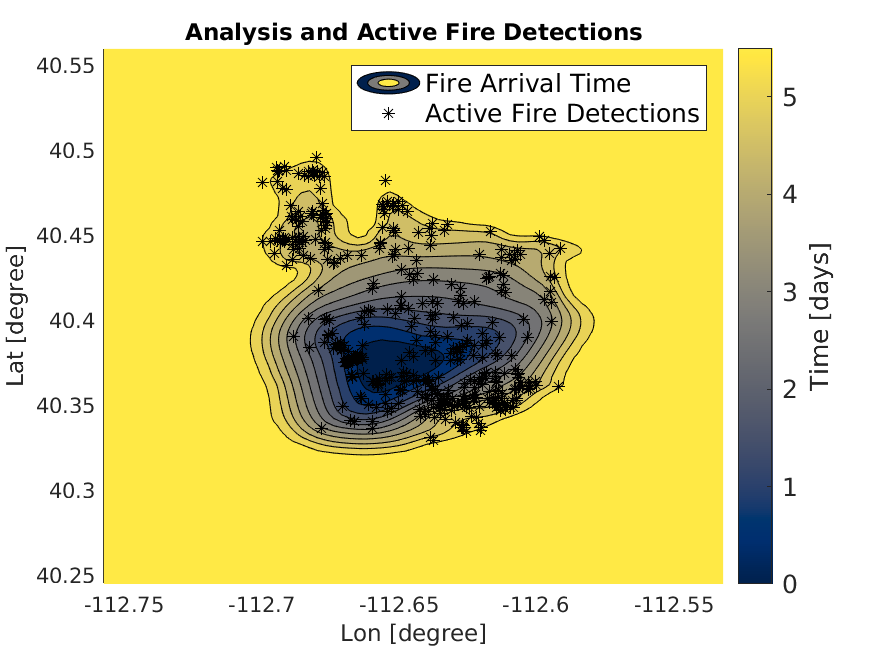}
\includegraphics[width=0.45\textwidth]{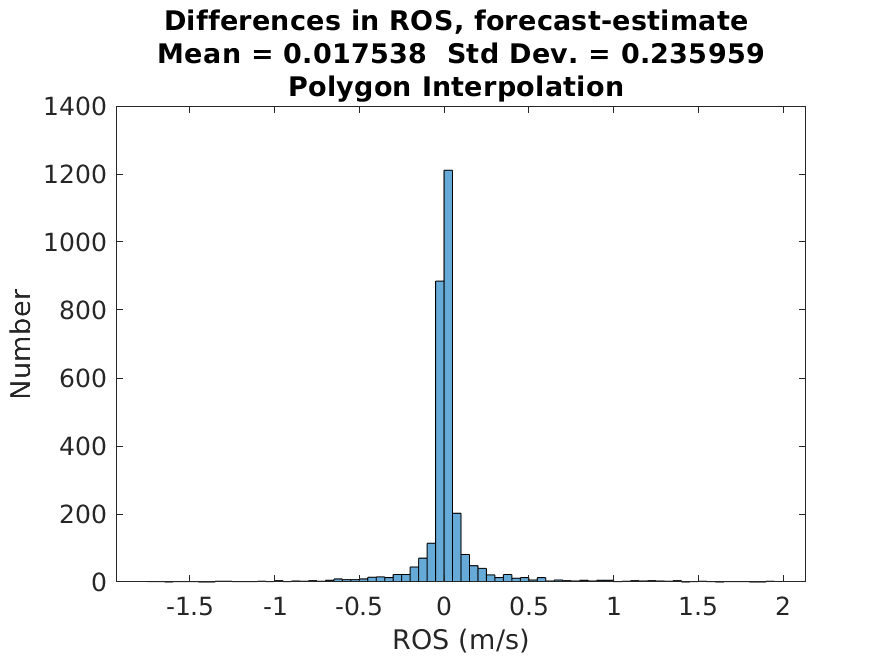}
     \caption[Cycle 4 of the Patch Fire simulation with adjustment of FMC.]{Cycle 4 of the Patch Fire simulation with adjustment of FMC. In the upper left and right are contours of of the fire arrival time for the forecast and the analysis, respectively. On the bottom is the histogram showing the difference between the ROS at mesh points in the fire domain.}
\label{fig:patch_fmc_cycle_4}
\end{figure}

\begin{figure}[!h]
\begin{center}
  \includegraphics[width = 0.45\textwidth]{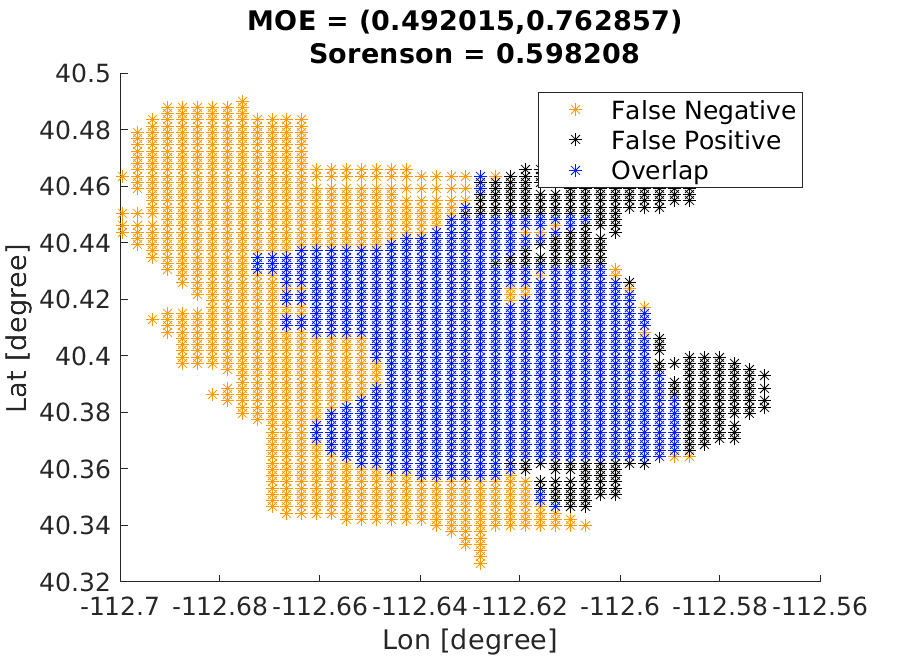}
  \includegraphics[width = 0.45\textwidth]{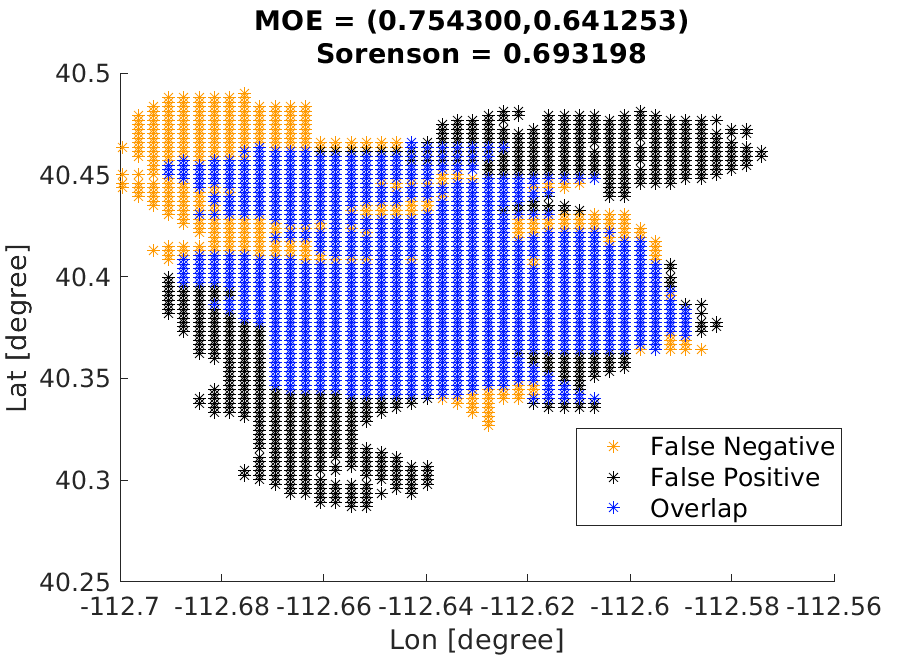}
  
  \includegraphics[width = 0.45\textwidth]{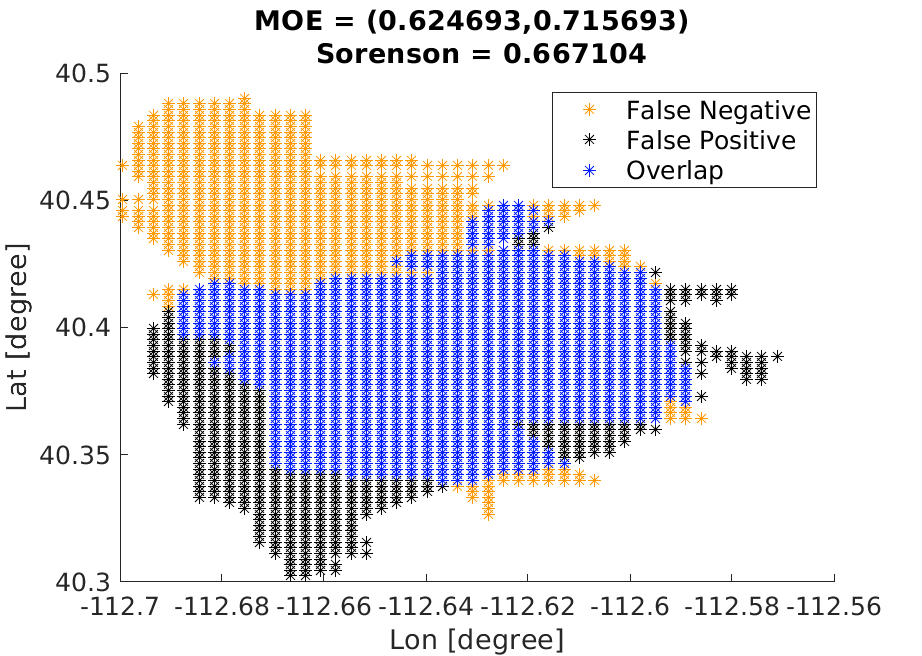}
  \includegraphics[width = 0.45\textwidth]{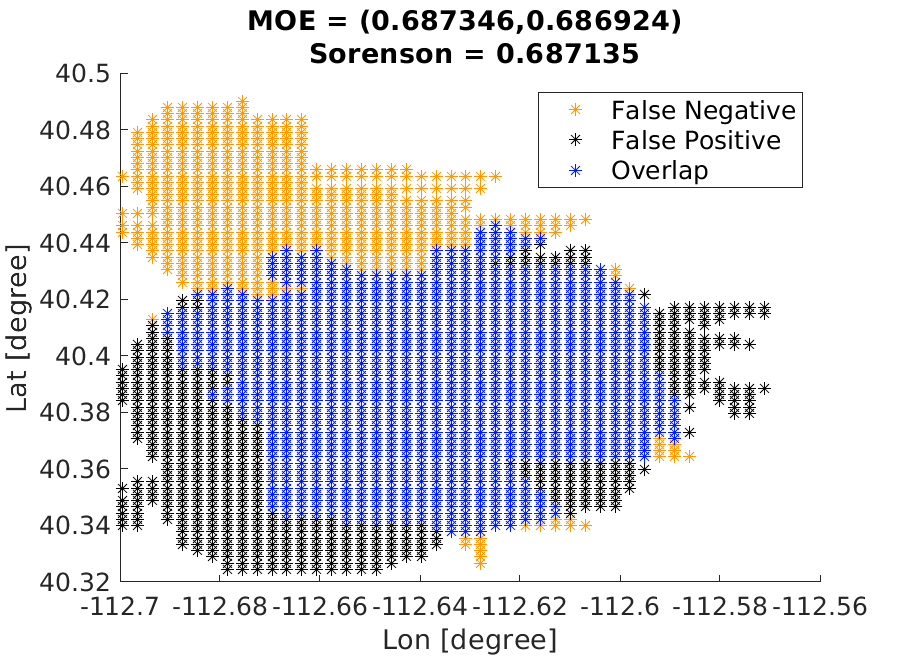}
  
  \includegraphics[width = 0.45\textwidth]{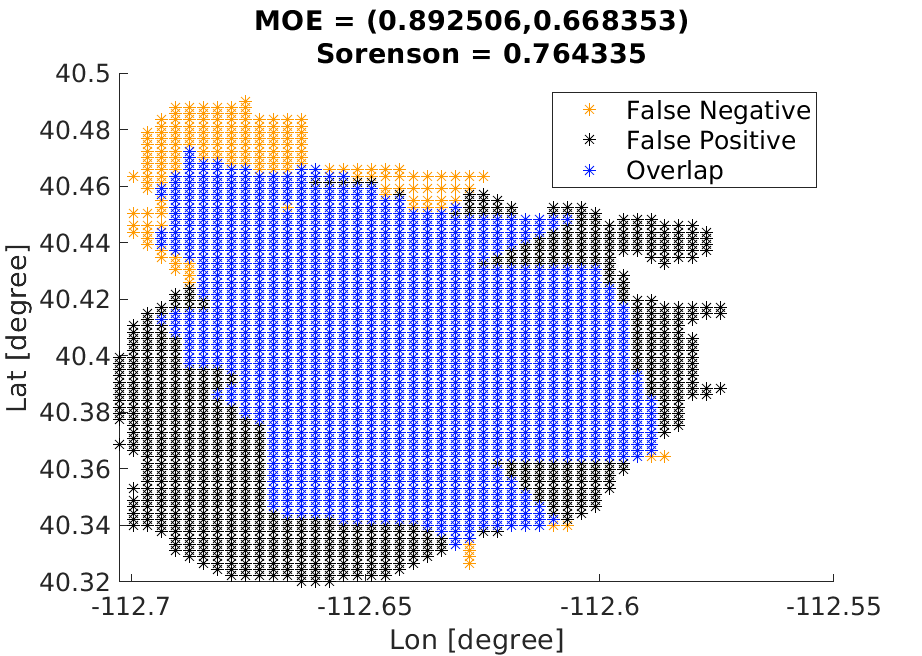}
  \includegraphics[width = 0.45\textwidth]{patch_moe_est.png}

  \caption{Assesment of the cycles of the Patch Fire simulation using adjustment of FMC. The panels show, from top to bottom and left to right, the assessments of cycles 0 to 4. The simulations overestimated the size of the fire, as was the case when no adjustments to the FMC had been made. The figure on the lower right shows the assessment of the simulation initialized from satellite data, the other figures show the cycles of a simulation started from an ignition point.}
  \label{fig:patch_moe_fmc_cycles}
  \end{center}
\end{figure}


\begin{figure}[!ht]
\centering
\includegraphics[width=0.45\textwidth]{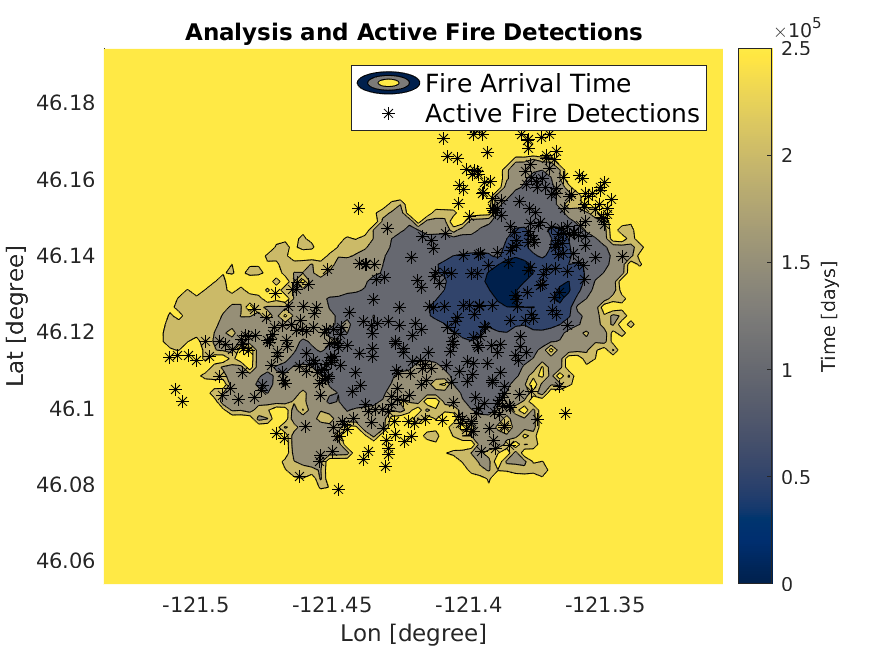}
\includegraphics[width=0.45\textwidth]{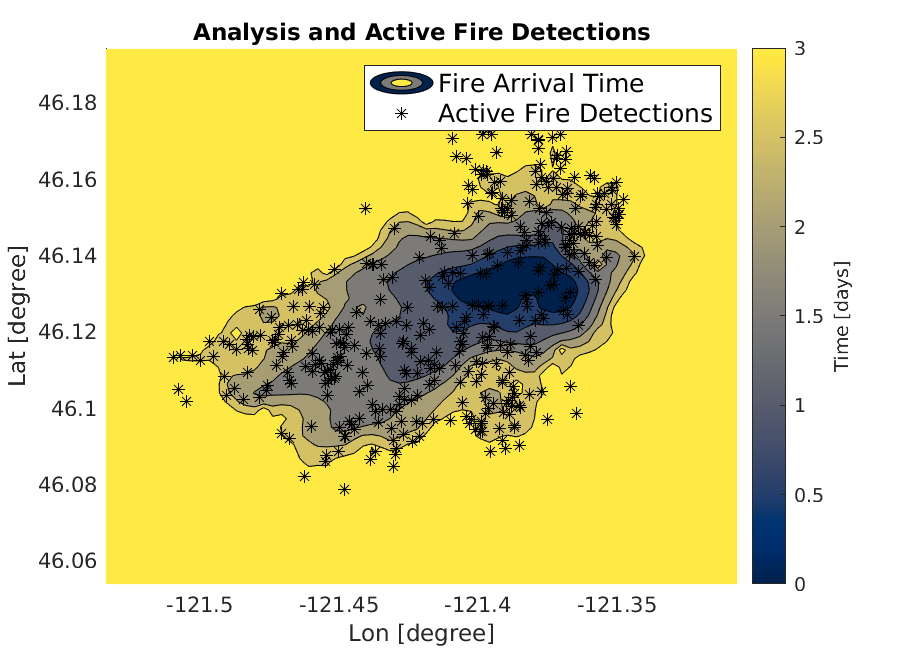}
\includegraphics[width=0.45\textwidth]{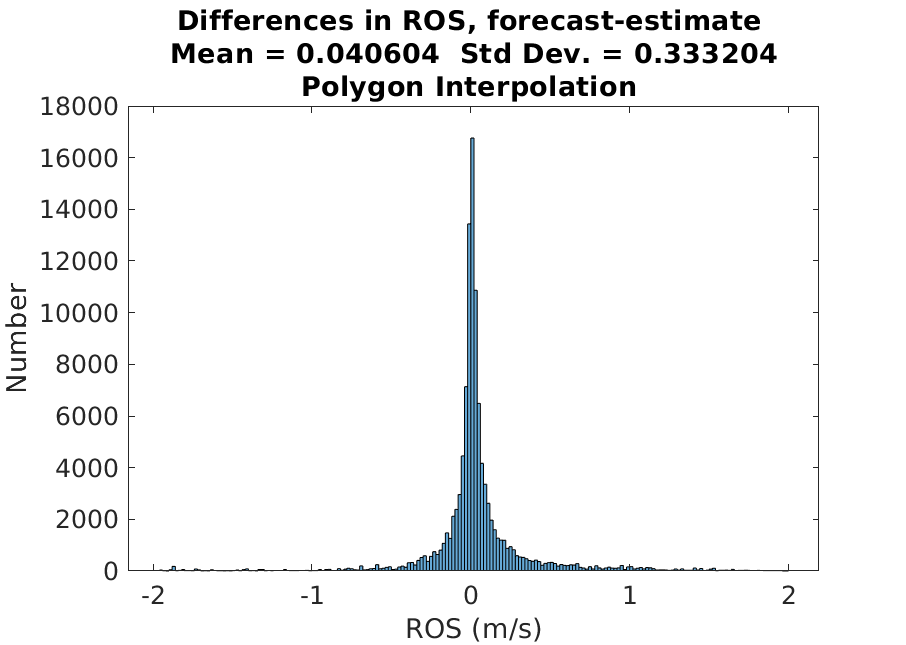}
     \caption{Cycle 2 of the Cougar Creek Fire simulation with adjustment of FMC. In the upper left and right are contours of of the fire arrival time for the forecast and the analysis, respectively. On the bottom is the histogram showing the difference between the ROS at mesh points in the fire domain. The forecast underestimated the size of the fire and overestimated the ROS so that the FMC was not adjusted.}
\label{fig:cougar_fmc_cycle_2}
\end{figure}

\begin{figure}[!ht]
\centering
\includegraphics[width=0.45\textwidth]{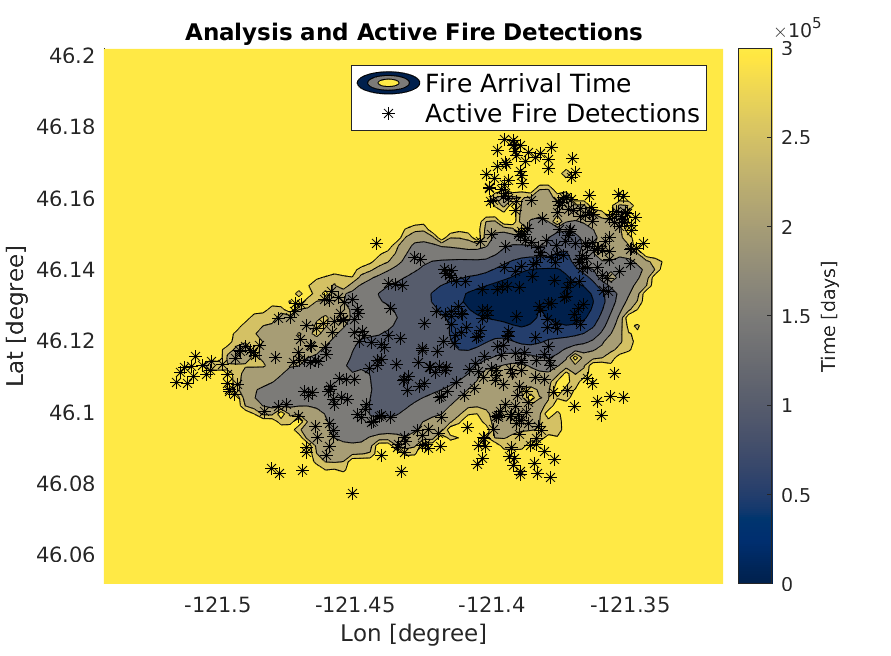}
\includegraphics[width=0.45\textwidth]{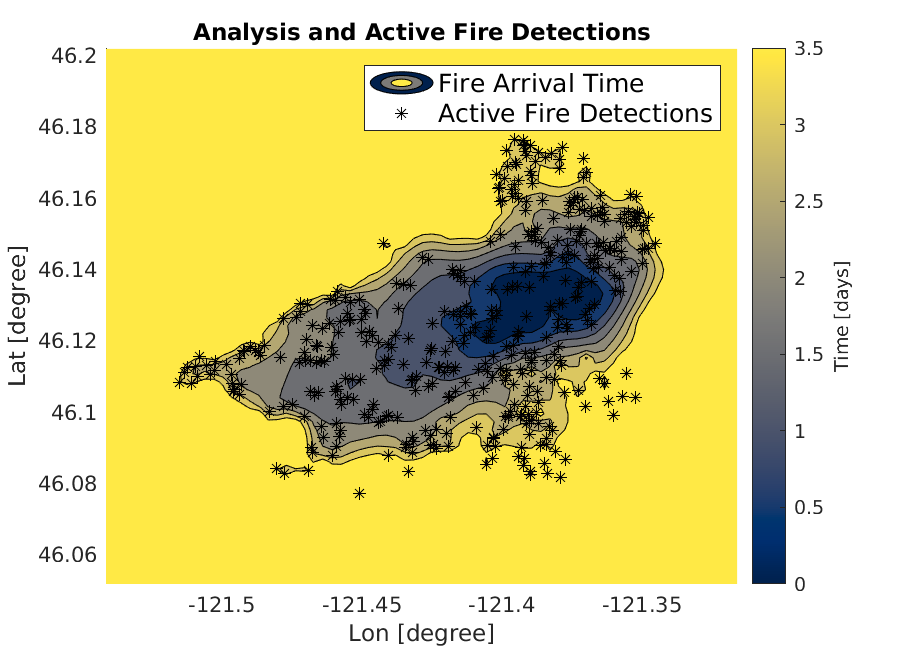}
\includegraphics[width=0.45\textwidth]{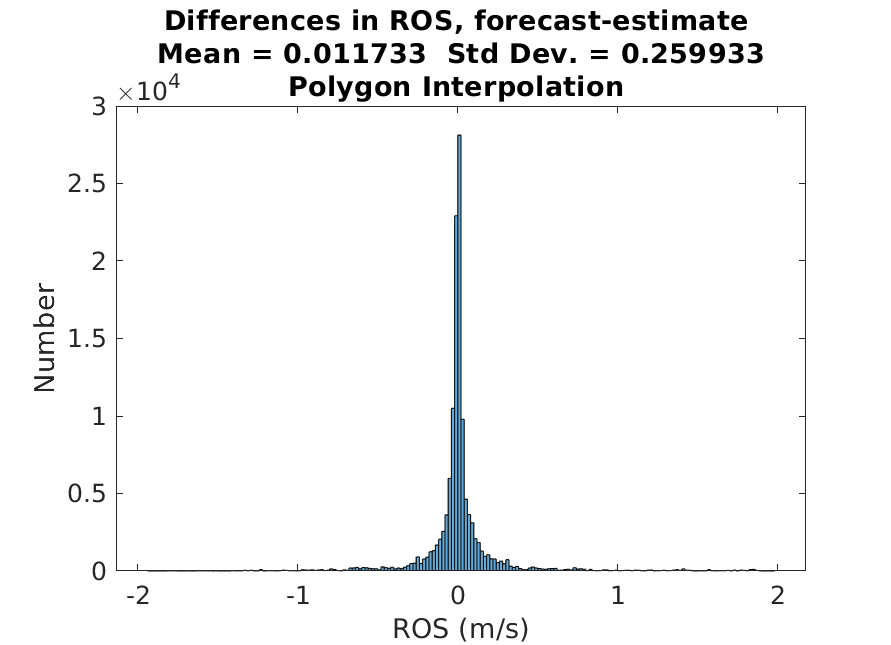}
     \caption{Cycle 3 of the Cougar Creek Fire simulation with adjustment of FMC. In the upper left and right are contours of of the fire arrival time for the forecast and the analysis, respectively. On the bottom is the histogram showing the difference between the ROS at mesh points in the fire domain. The forecast underestimated the size of the fire and overestimated the ROS so that the FMC was not adjusted.}
\label{fig:cougar_fmc_cycle_3}
\end{figure}

\begin{figure}[!h]
\begin{center}
  \includegraphics[width = 0.45\textwidth]{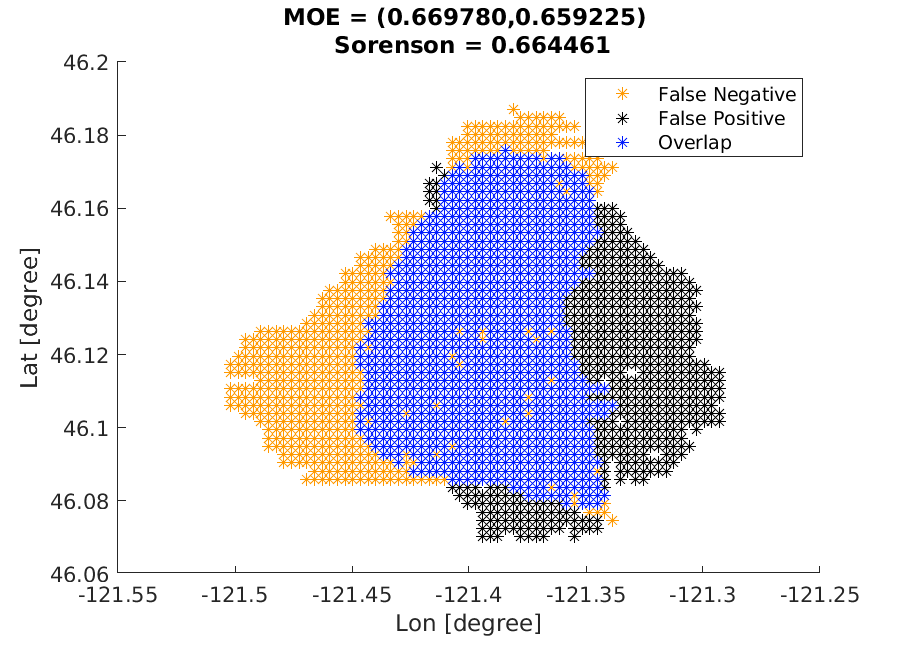}
  \includegraphics[width = 0.45\textwidth]{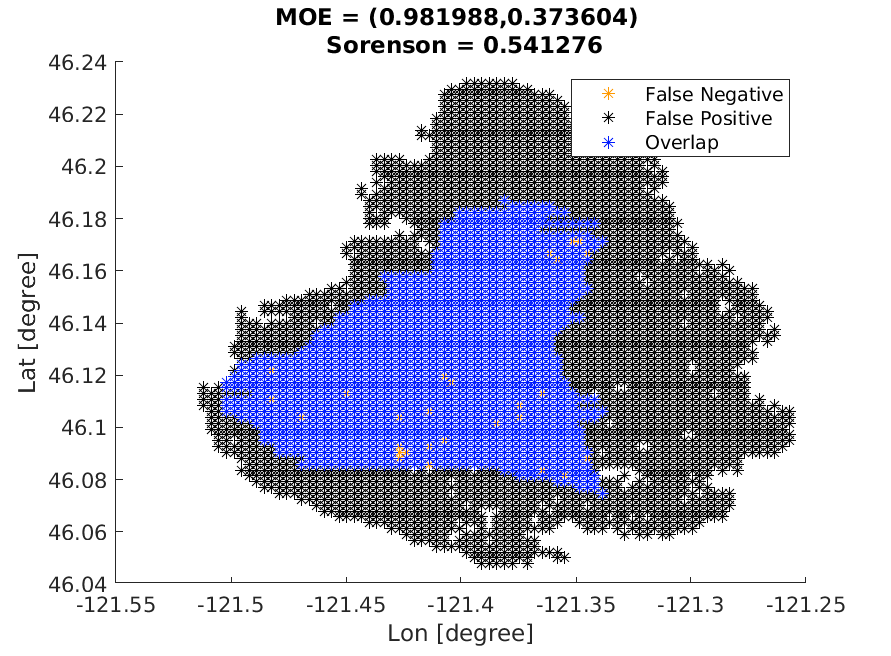}
  
  \includegraphics[width = 0.45\textwidth]{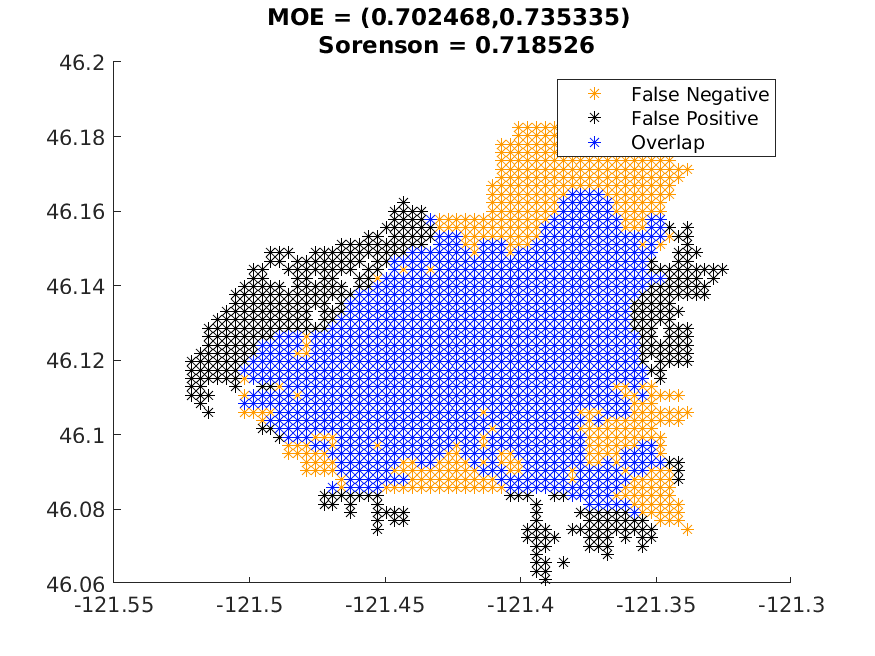}
  \includegraphics[width = 0.45\textwidth]{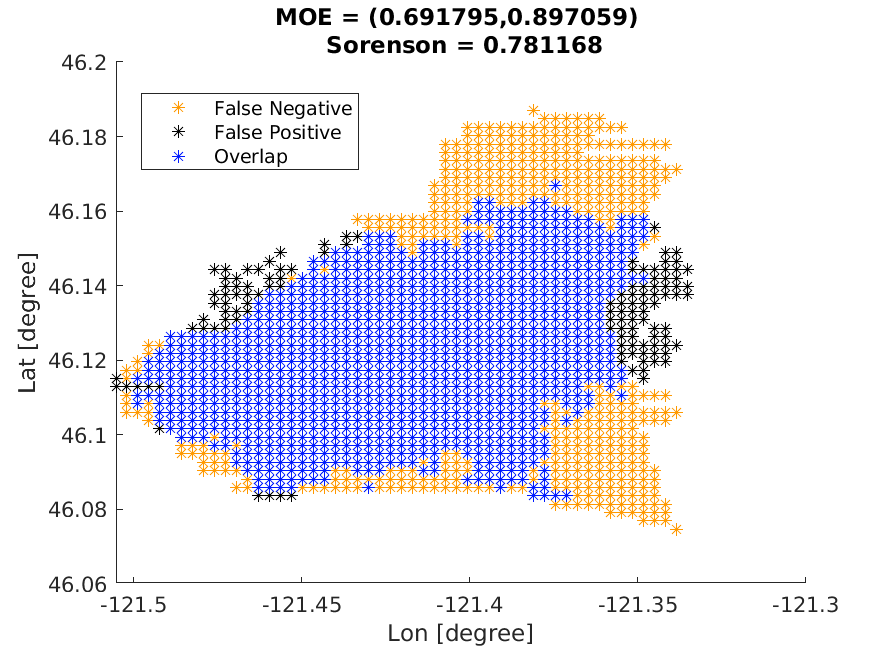}
  
  \includegraphics[width = 0.45\textwidth]{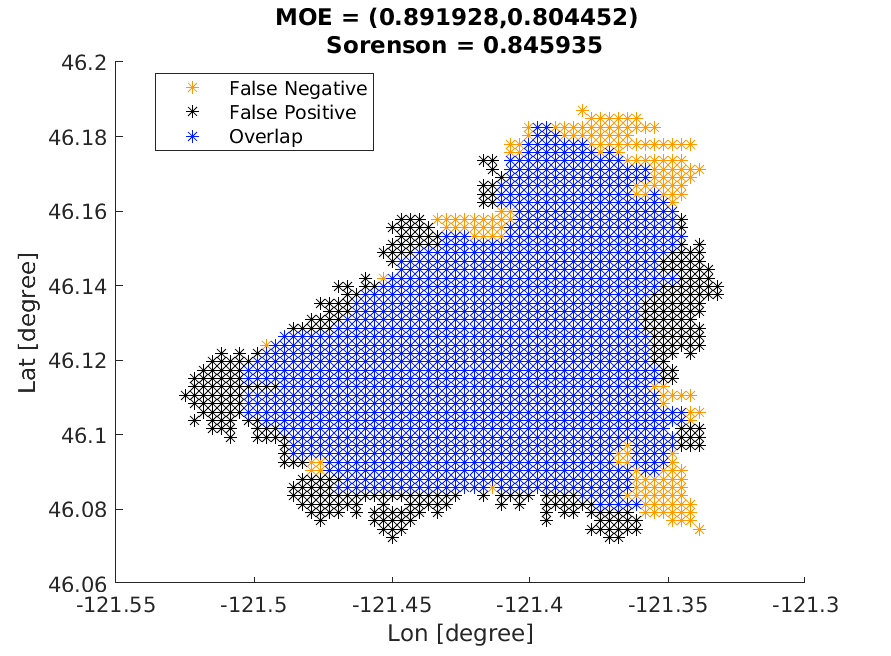}
  \includegraphics[width = 0.45\textwidth]{cougar_moe_estimate.png}

  \caption[Assessment of Cougar Creek Fire simulation using FMC adjustment]{Assessment of the cycles of the Cougar Creek Fire simulation that used adjustments of the FMC. TThe panels show, from top to bottom and left to right, the assessments of cycles 0 to 4. The panel in the lower right shows the result from a simulation initialized from an estimate of the fire arrival time derived from satellite fire data. Compare with Figure \ref{fig:cougar_moe_cycles}. The adjustment of the FMC has resulted in a better simulation in cycle 4.}
  \label{fig:cougar_moe_fmc_cycles}
  \end{center}
\end{figure}

\begin{figure}[!h]
\begin{center}
  \includegraphics[width = 0.45\textwidth]{patch_moe_0.png}
  \includegraphics[width = 0.45\textwidth]{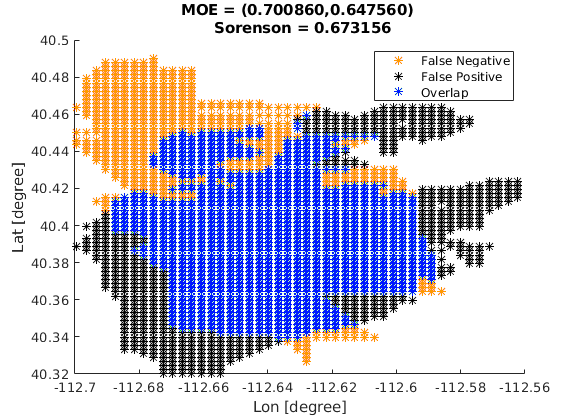}
  
  \includegraphics[width = 0.45\textwidth]{patch_fmc_moe_2.png}
  \includegraphics[width = 0.45\textwidth]{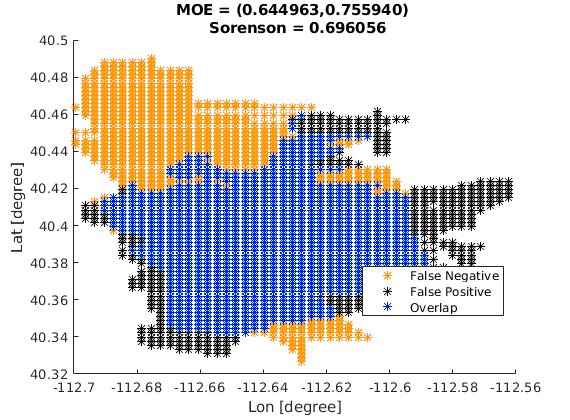}
  
\caption[Assessment of the cycles of the Patch Fire simulation using various forecasting strategies.]{Assessment of the cycles of the Patch Fire simulation using various forecasting strategies. The graphics give a graphical representation of how the fire area of simulation cycles match a perimeter observation made on August 16 at 09:47 UTC.}
\label{fig:patch_moe_conclusion}
\end{center}
\end{figure}

\begin{figure}[!h]
\begin{center}
\includegraphics[width = 0.45\textwidth]{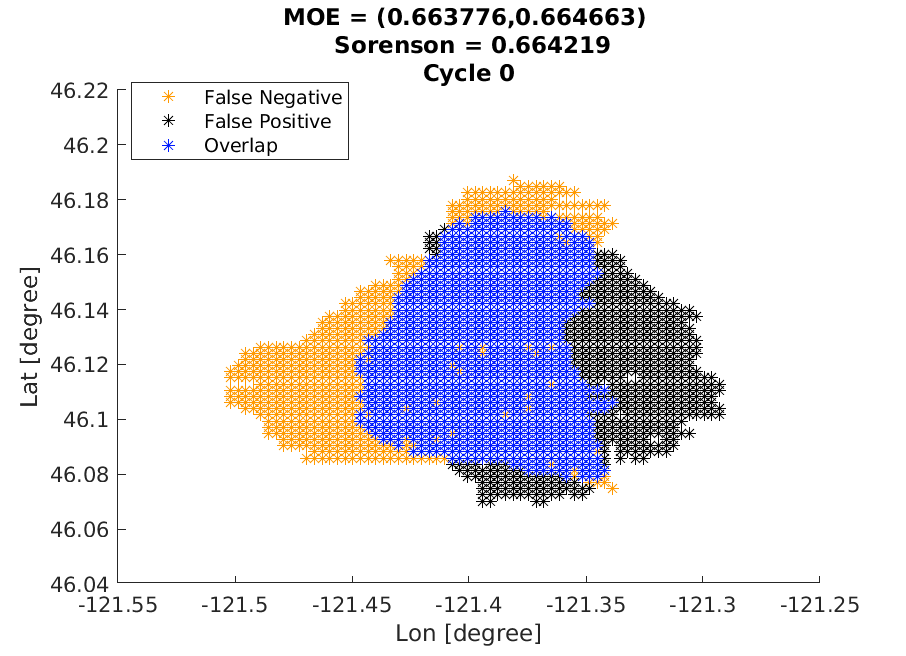}
\includegraphics[width = 0.45\textwidth]{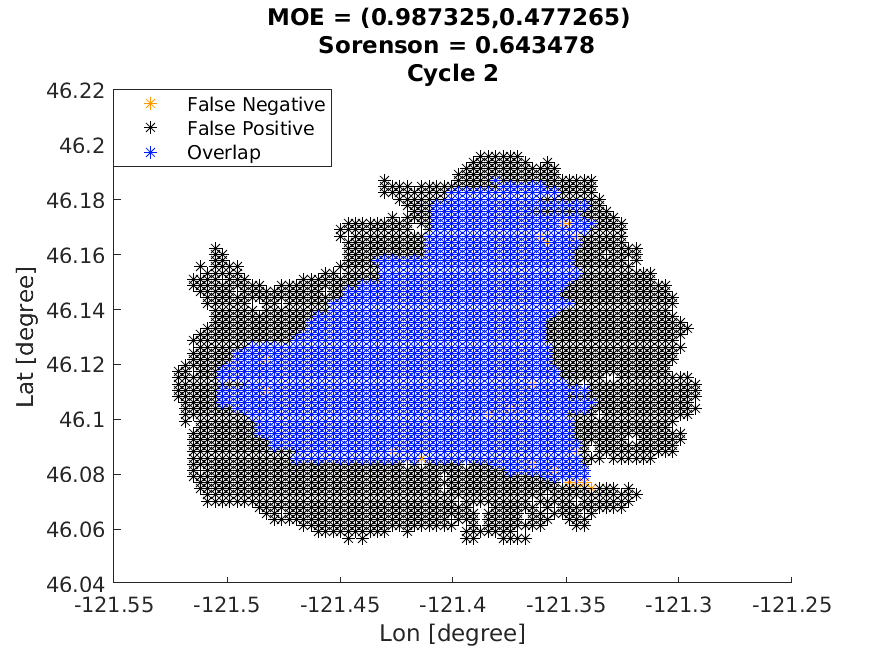}

\includegraphics[width = 0.45\textwidth]{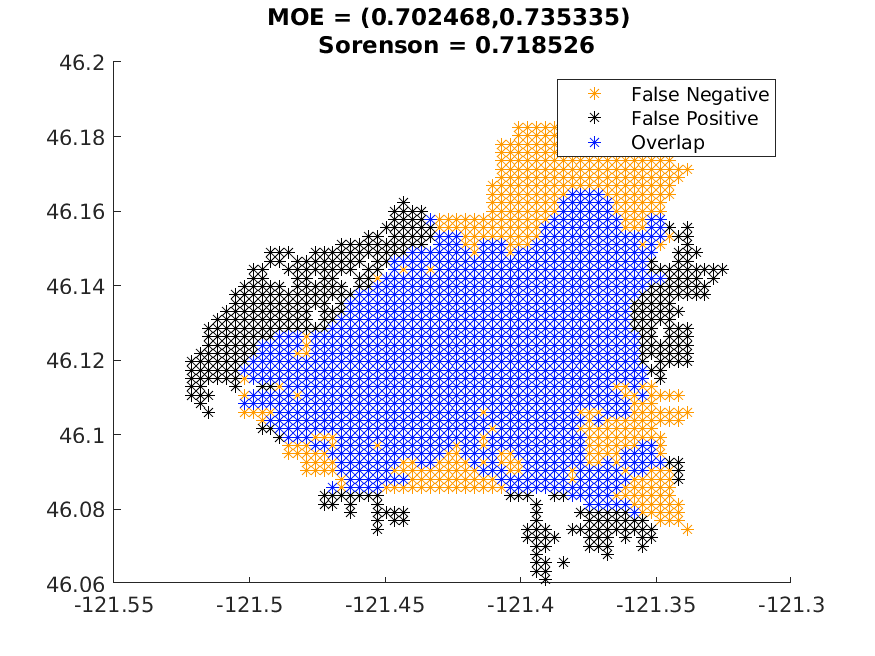}
\includegraphics[width = 0.45\textwidth]{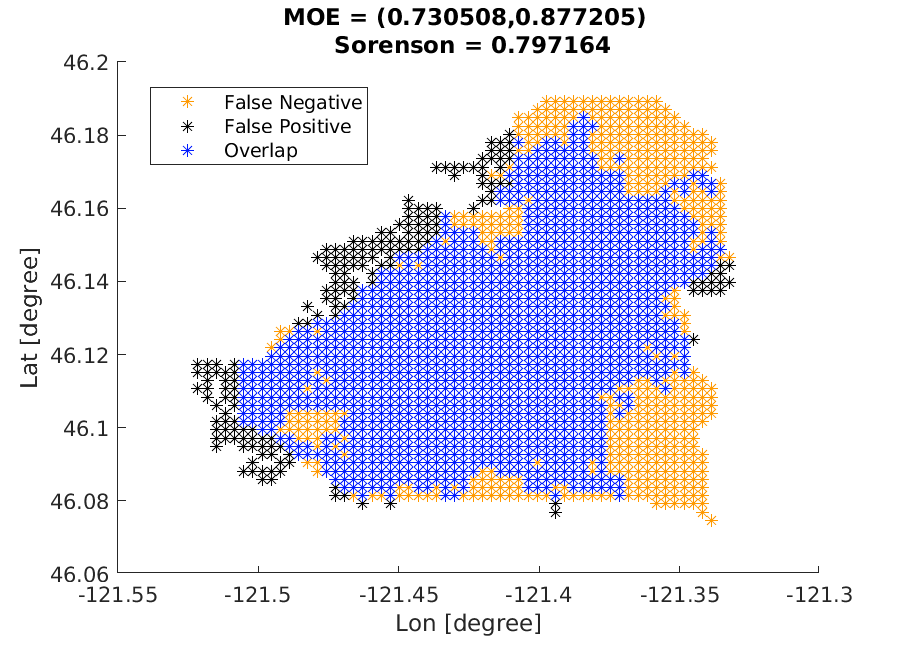}
  \caption[Assessment of the cycles of the Patch Fire simulation.]{Assessment of the cycles of the Patch Fire simulation.}
  \label{fig:cougar_moe_conclusion}
  \end{center}
\end{figure}


\begin{figure}[!h]
\begin{center}
  \includegraphics[width = 0.50\textwidth]{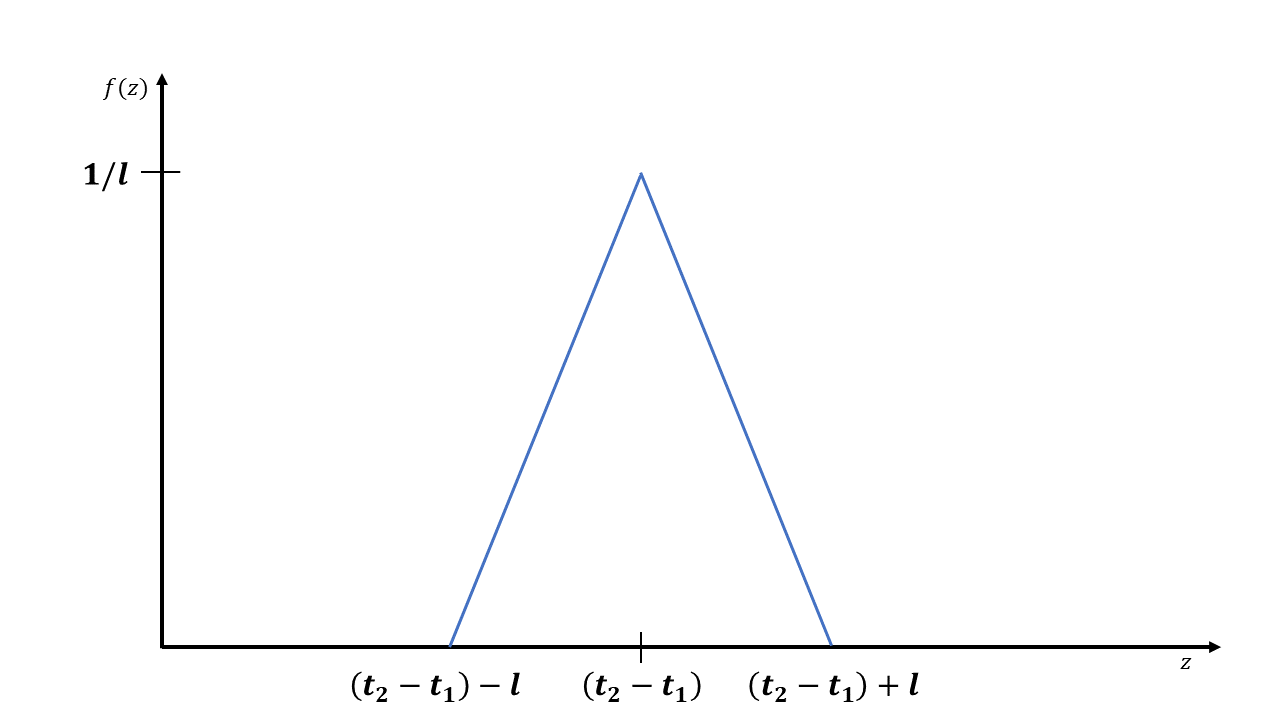}
\caption[Distribution of the difference of detection times.]{Distribution of the difference of detection times. The distribution has the familiar shape of a triangle resulting from the sum of two uniformly distributed random variables. The peak is $t_2-t_1$, the difference of the reported times of fire detections. The paramater $l$ gives the length of the support of the uniform random variables.}
\label{fig:uniform_diff}
\end{center}
\end{figure} 

\begin{figure}[!h]
\begin{center}
  \includegraphics[width = 0.79\textwidth]{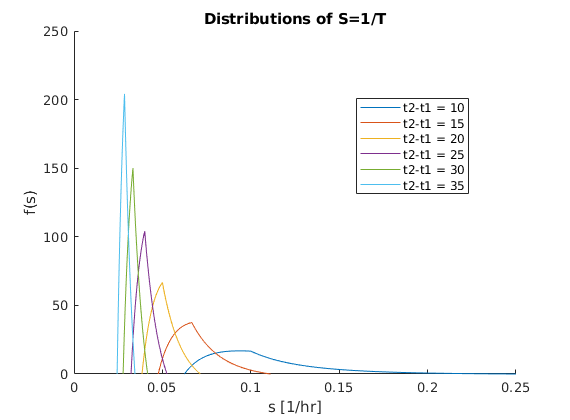}
\caption[Distributions of the reciprocals of difference of detection times.]{Distributions of the reciprocals of difference of detection times. As the time between detections increases, the distribution becomes narrower. Wider distributions result from closer detection times. In these cases, it was assumed that an individual detection time was a uniform random variable over the period spanning the six hours previous to the satellite overpass and detection.}
\label{fig:distributions}
\end{center}
\end{figure} 

\begin{figure}[!h]
\begin{center}
  \includegraphics[width = 0.50\textwidth]{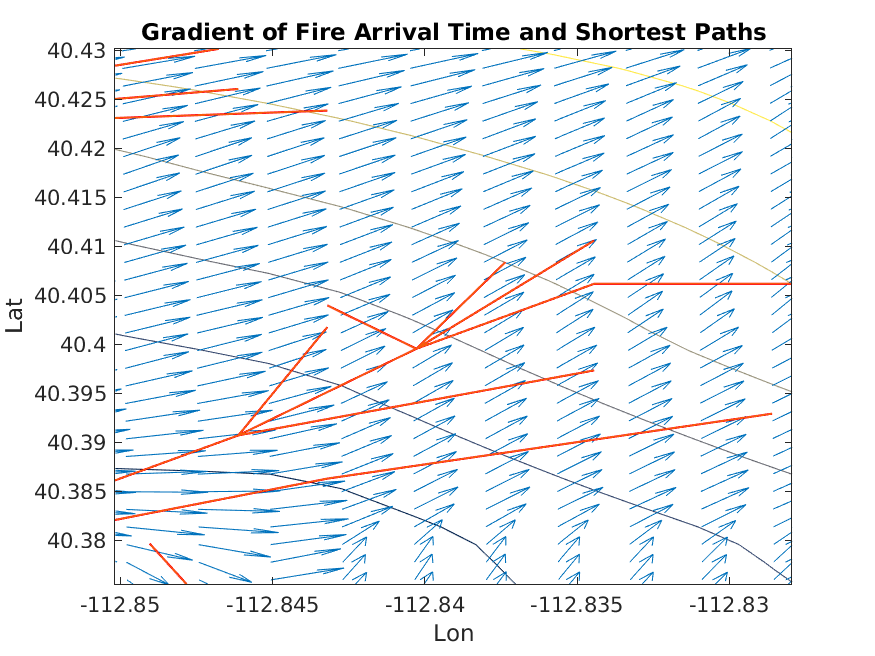}
\caption{Gradients in the estimated fire arrival time with shortest paths from directed graph of detection data. This is a detail of a small section in the fire domain to illustrate how the paths align with the gradient of the fire arrival time estimate.}
\label{fig:gradients_2d_detail}
\end{center}
\end{figure} 

\begin{figure}[!h]
\begin{center}
\includegraphics[width = 0.45\textwidth]{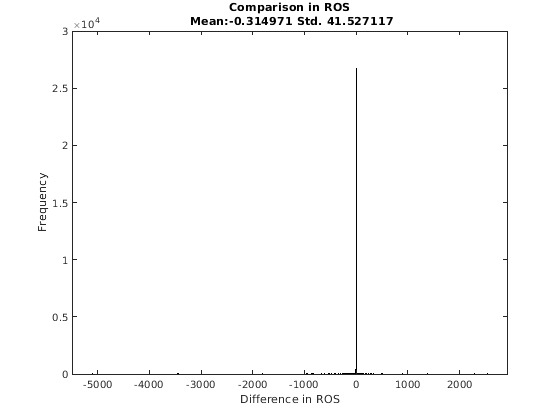}
\includegraphics[width = 0.45\textwidth]{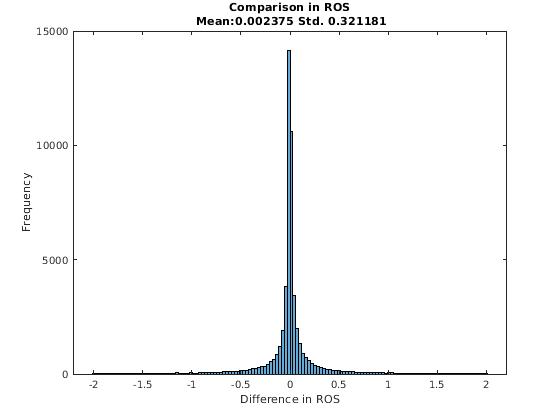}
\includegraphics[width = 0.45\textwidth]{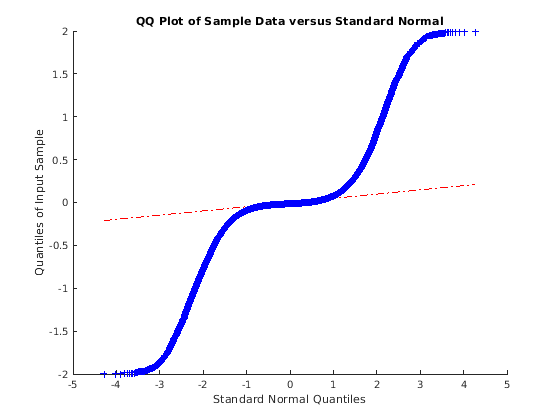}
\caption[The difference in ROS between  the ``ground truth" and the estimate made from data.]{The difference in ROS between the ``ground truth" and the estimate made from data. On the upper left, the histogram shows computed differences for all ROS in the fire domain. On the right, only the ROS differences less than 2 m/s in magnitude have been retained. On the bottom, a QQ plot confirms that the the differences in the ROS are not normally distributed.}
\label{fig:gradients_2d}
\end{center}
\end{figure} 
\FloatBarrier

\section{Additional Testing}
\subsection{Experiment: Effect of Using Perimeter Data on the Estimate of the Fire Arrival Time} 
\label{sec:fake_perim_test}

When infrared perimeter observations are available, they may be used in the same way as satellite fire detections to help derive an estimate of the fire arrival time. In this test, a comparison was made between estimates of the fire arrival time made from ``ground truth" artificial fire times whose surfaces had been scattered with artificial active fire detections. Two sets of estimates were made. One set was made using only artificial detections and the other was made using artificial perimeter observations as well. The following experiment uses a limited number of cases to get a first impression of whether using perimeter data could have a large effect on how well the method for estimating the fire arrival time works. In general, the use of perimeter data did result in better estimates, but the difference was not large. If there had been a large difference, then further testing would have been used to determine under what conditions and what strategies could be best used to arrive at the best possible estimate of the fire arrival time.

Four artificial fire arrival times were made. Each represented a fire burning for 6 days. An artificial perimeter of 100 points was made at the time corresponding to midnight on each day of the simulation. Estimates of the fire arrival time were made either using or not using the artificial perimeters. The estimates were then assessed using the methods described in Section \ref{sec:assess_tools}. The results, in Table \ref{tbl:perim_exp_results} show, that use of perimeter observations produces better estimates of the fire arrival time.
	
\begin{figure}[!ht]
\centering
\includegraphics[width=0.45\textwidth]{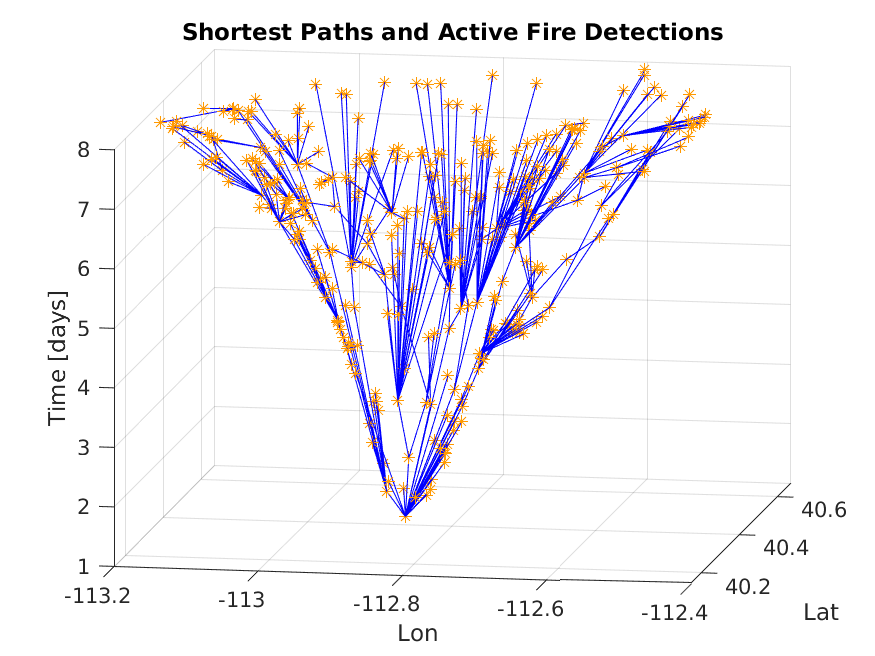}
\includegraphics[width=0.45\textwidth]{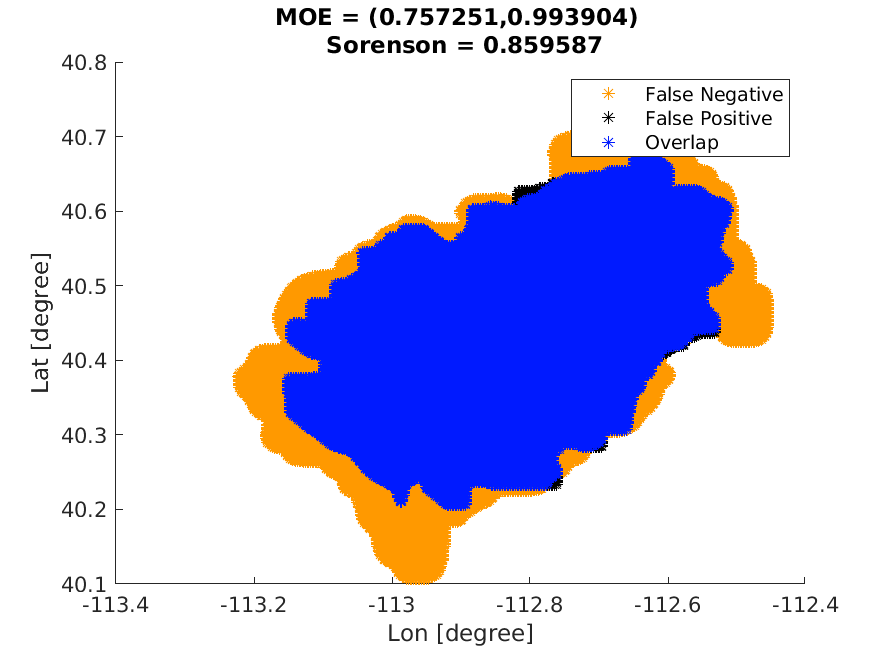}
              
\includegraphics[width=0.45\textwidth]{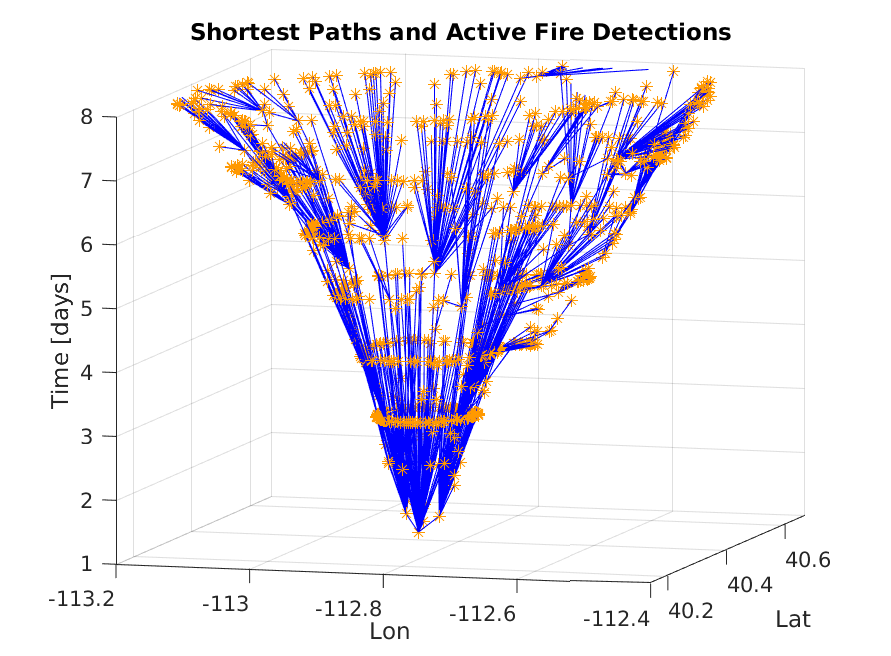}
\includegraphics[width=0.45\textwidth]{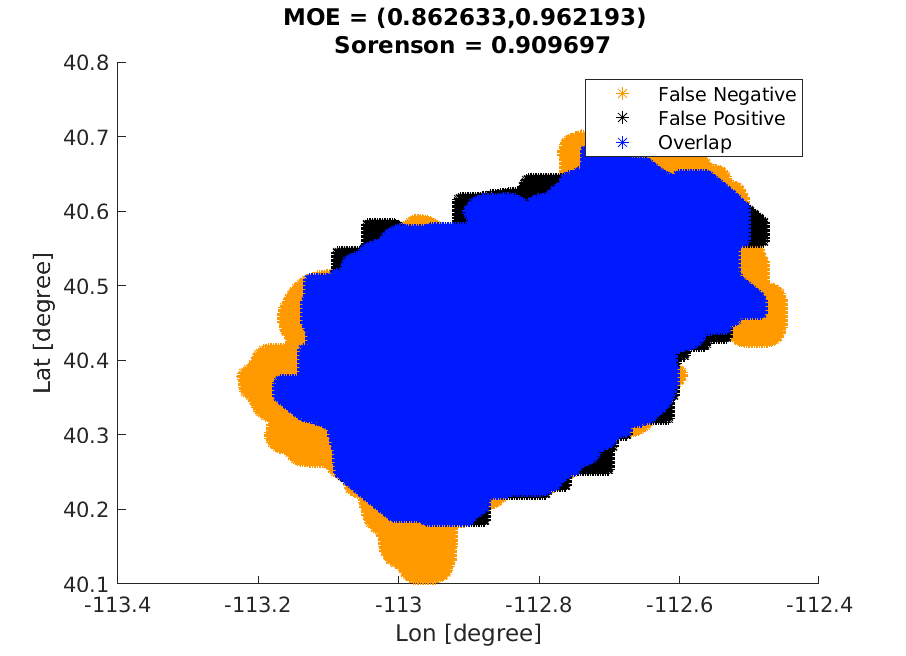}

     \caption[Using perimeter data as an additional source of fire detections in the method.]{Using perimeter data as an additional source of fire detections in the method. The top row shows the path structure made from a ``ground truth" fire arrival time and artificial detections and a plot of how the fire arrival time compared to the ground truth. The bottom row shows the path structure made from the detection data used in the paths in the top row as well as more data points created from artificial perimeter data. Several ``rings of detections" can be seen in the lower path structure. The plots on the right side of the figure show that the use of perimeter data makes for a bestter estimate of the fire arrival time.}
\label{fig:perim_exp_p4}
\end{figure}

\begin{table}
\centering
\begin{tabular}{|c|c|c|c|c|c|} 
\hline
\multicolumn{6}{|c|}{No Perimeters Used}                                         \\ 
\hline
Cone    & MOE X  & MORE Y & \textbar{}MOE\textbar{} & S{\o}renson & Relative Error  \\ 
\hline
1       & 0.5255 & 0.9990 & 1.1288                  & 0.6888   & 0.0085          \\ 
\hline
2       & 0.5449 & 1.0000 & 1.1388                  & 0.7054   & 0.0095          \\ 
\hline
3       & 0.6766 & 0.9974 & 1.2052                  & 0.8062   & 0.0064          \\ 
\hline
4       & 0.7573 & 0.9939 & 1.2495                  & 0.8596   & 0.0098          \\ 
\hline
Average & 0.6261 & 0.9976 & 1.1806                  & 0.7650   & 0.0085          \\ 
\hline
\multicolumn{6}{|c|}{Perimeters Used – 100 points in each}                       \\ 
\hline
Cone    & MOE X  & MORE Y & \textbar{}MOE\textbar{} & Sorenson & Relative Error  \\ 
\hline
1       & 0.6197 & 0.9986 & 1.1752                  & 0.7648   & 0.0085          \\ 
\hline
2       & 0.5317 & 1.0000 & 1.1326                  & 0.6942   & 0.0083          \\ 
\hline
3       & 0.7894 & 0.9595 & 1.2425                  & 0.8662   & 0.0063          \\ 
\hline
4       & 0.8626 & 0.9622 & 1.2923                  & 0.9097   & 0.0064          \\ 
\hline
Average & 0.7009 & 0.9801 & 1.2106                  & 0.8087   & 0.0074          \\
\hline
\end{tabular}
\caption{Comparison of estimates of fire arrival times made with or without using perimeter information in the path structure. Using perimeter information gives better spatial representations and has smaller relative errors when the estimated fire arrival times are compared to the ``ground truth" fire arrival times. }
\label{tbl:perim_exp_results}
\end{table}
\pagebreak
\FloatBarrier



\end{document}